\documentclass[11pt,letterpaper]{report}
\usepackage[width=150mm,top=35mm,bottom=35mm,bindingoffset=6mm]{geometry} 

\usepackage{multirow}
\usepackage[table,xcdraw]{xcolor}
\usepackage{amsmath,amssymb} 

\DeclareMathOperator*{\argmin}{arg\,min}

\usepackage[T1]{fontenc}
\usepackage{titlesec}

\titleformat{\section}
  {\normalfont\sffamily\Large\bfseries}
  {\thesection}{1em}{}

\newcolumntype{Z}{>{\centering\hspace{0pt}\arraybackslash}p{1.5cm}}

\usepackage[nottoc]{tocbibind}
	\settocbibname{References}
\usepackage[titletoc,title]{appendix}

\usepackage{fancyhdr}

\usepackage{multirow}
\usepackage{tabularx,tabulary}

\usepackage[newcommands]{ragged2e} 
\usepackage{booktabs} 
\usepackage{subcaption}
\usepackage{graphicx}
\usepackage{amsmath, amssymb,amsthm}

\usepackage{dsfont}
\usepackage[table,xcdraw]{xcolor}
\usepackage{arydshln}
\usepackage{textcomp}
\usepackage{ragged2e}

\usepackage{xfrac} 
\graphicspath{{images/}} 
\usepackage[hidelinks]{hyperref} 
\usepackage{pdflscape} 
\usepackage{afterpage}
\usepackage{color} 
\usepackage{mdwlist} 
\usepackage{diagbox} 
\usepackage{titlesec} 

\usepackage{mathtools}
\usepackage{bm}
\usepackage{nccmath}
\usepackage{algorithm}
\usepackage{algorithmic}
\usepackage{xcolor}
\usepackage{cite}
\usepackage{nccmath}
\usepackage{float}

\newcommand{\mat}[1]{\bm{\mathrm{#1}}}
\newcommand{\light}{\ell}
\newcommand{\lossfun}{\mathcal{L}}
\newcommand{\headercolor}{FFFFFF}
\newcommand{\bestcolor}{FFFC9E}
\newcommand{\secondbestcolor}{FFCCC9}

\newcommand{\modelweights}{\theta}
\newcommand{\waugmentation}{w/aug.}
\newcommand{\numinputs}{m}
\newcommand{\sobel}{\nabla}

\titleformat{\section}{\normalfont\Large\bfseries}{\thesection}{1em}{}

\titleformat{\chapter}{\normalfont\huge\bfseries}{\thechapter}{1em}{}

\newcommand{\im}{\mat{x}}
\newcommand{\imraw}{\im_{\mathrm{raw}}}

\newcommand{\imxyz}{\im_{\mathrm{xyz}}}
\newcommand{\imxyzgt}{\imxyz^*}
\newcommand{\imxyzpred}{\hat{\im}_{\mathrm{xyz}}}

\newcommand{\imglob}{\im_{\mathrm{glob}}}

\newcommand{\imsrgb}{\im_{\mathrm{srgb}}}
\newcommand{\imsrgbgt}{\imsrgb^*}
\newcommand{\imsrgbpred}{\hat{\im}_{\mathrm{srgb}}}

\newcommand{\imres}{\im_{\mathrm{res}}}

\newcommand{\pipe}{\mathcal{F}}
\newcommand{\invpipe}{\mathcal{G}}

\newcommand{\pipeglob}{\pipe_\mathrm{glob}}
\newcommand{\pipeloc}{\pipe_\mathrm{loc}}

\newcommand{\invpipeglob}{\invpipe_\mathrm{glob}}
\newcommand{\invpipeloc}{\invpipe_\mathrm{loc}}

\newcommand{\matt}{\mat{M}}
\newcommand{\matfwd}{\matt_\mathrm{fwd}}
\newcommand{\matinv}{\matt_\mathrm{inv}}

\newcommand{\pX}{\mat{x}}
\newcommand{\I}{\mat{I}}
\newcommand{\Iy}{\I_y}
\newcommand{\cR}{\mat{R}}
\newcommand{\cG}{\mat{G}}
\newcommand{\cB}{\mat{B}}

\newcommand{\abs}[1]{\left\lvert#1\right\rvert}

\newcommand{\kernelSixbyN}{\phi}
\newcommand{\reshape}{\psi}

\title{ \huge\textbf{IMAGE COLOR CORRECTION, \\ENHANCEMENT, AND EDITING}}
\author{ \Large{\color{black}MAHMOUD NASSER MOHAMMED AFIFI} \\[12pt]
\date{\vfill\normalsize{A DISSERTATION SUBMITTED TO THE FACULTY OF GRADUATE STUDIES IN PARTIAL FULFILLMENT OF THE REQUIREMENTS FOR THE DEGREE OF}\\ \vfill
\large{DOCTOR OF PHILOSOPHY} \\ \vfill
\normalsize{GRADUATE PROGRAM IN\\ELECTRICAL ENGINEERING AND COMPUTER SCIENCE\\
YORK UNIVERSITY\\
TORONTO, ONTARIO\\
April 2021\\ \vfill
\copyright \hspace{1mm} MAHMOUD AFIFI, 2021}}}

\begin{document}
\maketitle

\pagenumbering{roman}
\addtocounter{page}{1}

\chapter*{Abstract}
\addcontentsline{toc}{chapter}{Abstract}
This thesis presents methods and approaches to image color correction, color enhancement, and color editing. To begin, we study the color correction problem from the standpoint of the camera's image signal processor (ISP). A camera's ISP is hardware that applies a series of in-camera image processing and color manipulation steps, many of which are nonlinear in nature, to render the initial sensor image to its final photo-finished representation saved in the 8-bit standard RGB (sRGB) color space. As white balance (WB) is one of the major procedures applied by the ISP for color correction, this thesis presents two different methods for ISP white balancing. Afterwards, we discuss another scenario of correcting and editing image colors, where we present a set of methods to correct and edit WB settings for images that have been improperly white-balanced by the ISP.  Then, we explore another factor that has a significant impact on the quality of camera-rendered colors, in which we outline two different methods to correct exposure errors in camera-rendered images. Lastly, we discuss post-capture auto color editing and manipulation. In particular, we propose auto image recoloring methods to generate different realistic versions of the same camera-rendered image with new colors. Through extensive evaluations, we demonstrate that our methods provide superior solutions compared to existing alternatives targeting color correction, color enhancement, and color editing.

\clearpage
\begin{center}
\addcontentsline{toc}{chapter}{Dedication}
    \vspace*{\fill}
    \textit{To my wife, kids, parents, and aunt}
    \vspace*{\fill}
\end{center}


{\chapter*{Acknowledgment}
\addcontentsline{toc}{chapter}{Acknowledgment}
\begin{figure}[h!]
\vspace{-5mm}
\hfill\includegraphics[width=\textwidth]{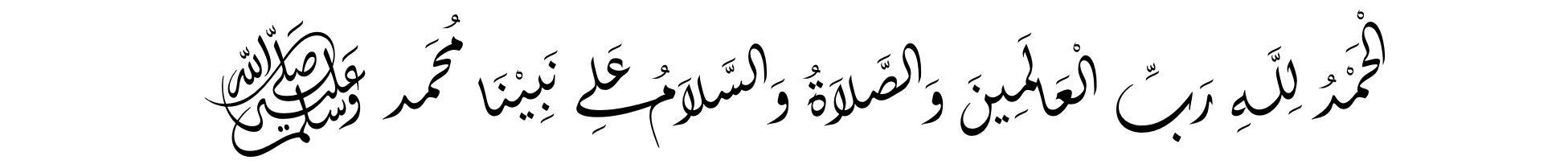}
\end{figure}
\noindent Praise be to Allah, Lord of the Worlds. Prayer and peace be upon
the Prophet Muhammad. I owe a great debt to my supervisor Michael S. Brown for his close guidance, support, and valuable advice. Sincere thanks to Marcus Brubaker, Richard Wildes, Brian Price, Scott Cohen, Konstantinos Derpanis, Francois Bleibel, Jonathan Barron, Chloe LeGendre, Yun-Ta Tsai for their help and support throughout this work. I thank my labmates at York, Abhijith Punnappurath, Abdelrahman Abdelhamed, Abdullah Abuolaim,  Hakki Can Karaimer, and Hoang Le for their valuable discussions.
I would like to express my deepest gratitude to my family. I thank my parents and my aunt for their love, support, and prayers. My wife, Zeinab, for providing me her unending love, patience, companionship, and encouragement to keep pushing ahead in my research when doubt crept in. My last thanks go to my kids, Malika and Yaseen, my most important contribution of all.
\clearpage}

{\addcontentsline{toc}{chapter}{Table of Contents}
\renewcommand{\contentsname}{Table of Contents}
\tableofcontents
\listoftables
\listoffigures
\clearpage}

{\chapter*{List of Acronyms}
\addcontentsline{toc}{chapter}{List of Acronyms}
\begin{tabbing}
AE ~~~~~~~~~~\= Auto Exposure \\
AC \> Auto-Color Function\\
AIM \> Advances in Image Manipulation  \\
AS \> Adobe Standard \\
AT \> Auto-Tone Function \\
AWB \> Auto White Balance \\

BCP \> Deep Bright-Channel Prior \\
BIN \> Batch-instance normalization \\
BN \> Batch Normalization  \\
BTF \> Brightness Transfer Function\\

Cat \> Category \\
CAM \> Color Adaptation Matrix \\
CC \> Color Constancy \\
CCC \> Convolutional Color Constancy \\
CFA \> Color Filter Array\\
NIS \> Natural Image Statistics \\
CIE \> International Commission on Illumination \\
CLAHE \> Contrast-Limited Adaptive Histogram Equalization \\
CNN \> Convolutional Neural Network\\
conv \> Convolutional \\
CPU \>  Central Processing Unit \\
CRF \> Camera Response Function\\
CS \> Camera Standard  \\
CST \> Color Space Transformation \\

dB \> Decibel \\
DCE \> Deep Curve Estimation \\
DNG \> Digital Negative \\
DNN \> Deep Neural Network \\
DS \> Deep Specialized\\
DSLR \> Digital Single-Lens Reflex \\
DoD \> Distribution of Color Distributions\\
DPE \> Deep Photo Enhancer\\

EGAN \> Enlighten GAN \\ 
EMD \> Earth Mover's Distance\\
EMoR \> Empirical Model of Response\\
EV \> Exposure Value\\

fc \> Fully Connected\\
FC4 \> Fully Convolutional Color Constancy with Confidence-Weighted Pooling\\
FFCC \> Fast Fourier Color Constancy\\

GAN \> Generative Adversarial Network \\
GE \> Gray Edges \\
GPU \>  Graphics Processing Unit \\
GT \> Ground Truth \\
GUI \> Graphical User Interface \\
GW \> Gray World \\

HDR \> High Dynamic Range \\
HE \> Histogram Equalization \\ 
Histo \> Histogram \\
HQEC \> High-Quality Exposure Correction \\

ILSVRC \> ImageNet Large Scale Visual Recognition Challenge \\
IoU \> Intersection-over-Union \\
ISP \> Image Signal Processor \\

JPEG \> Joint Photographic Experts Group \\

KinD \> Kindling the Darkness \\
KL \> Kullback-Leibler \\
KNN \> K Nearest Neighbor \\

LDR \> Low Dynamic Range \\
LCD \> Liquid Crystal Display\\
LIME \> Low-Light Image Enhancement \\
LOL \> LOw-Light \\
LReLU \> Leaky Rectified Linear Units \\
LUT \> Lookup Table \\

MAE \> Mean Angular Error \\
MB \> Megabyte \\
MoR \> Model of Response \\
MSE \> Mean Squared Error \\

NPE \> Naturalness Preserved Enhancement \\
nm \> Nanometers\\
NUS \> National University of Singapore\\

PCA \> Principal Component Analysis \\
PCC \> Polynomial Color Correction \\
PI \> Perceptual Index \\
PS \> Adobe Photoshop \\
PSNR \> Peak Signal-to-Noise Ratio \\
pxl-acc \> Pixel-Wise Accuracy \\

RAM \> Random-Access Memory \\
RBF \> Radial Basis Function \\
Rec. \> Reconstructed \\
ReHisto \> Recoloring Histogram \\
ReLU \> Rectified Linear Units \\
RHT \> HDR Transformation \\
RMSE \> Root Mean Squared Error \\
RNet \> RetinexNet \\
RPC \> Root-Polynomial Color Correction \\

SDK \> Software Development Kit \\
SIIE \> Sensor-Independent Illuminant Estimation \\
SIFT \> Scale-Invariant Feature Transform \\

SoG \> Shades of Gray \\
SPB \> MIT Scene Parsing Benchmark \\
SPD \> Spectral Power Distribution  \\
sRGB \> Standard RGB \\
SSIM \> Structrual Similarity \\
SURF \> Speeded-Up Robust Feature \\
SVD \> Singular Value Decomposition \\
SVR \> Support Vector Machine for Regression \\

t-SNE \> t-Distributed Stochastic Neighbor Embedding \\
TTL \> Through-the-Lens  \\

U CC \> Unsupervised Color Constancy \\
UPE \> Deep Underexposed Photo Enhancer \\ 
UPI \> Unprocessing Images \\

WB \> White Balance \\
wGE \> Weighted Gray Edges \\
WVM \> Weighted Variational Model \\

\end{tabbing}
\clearpage}

{\pagenumbering{arabic}
\part{Introduction and Prior Work\label{part:intro}}}
\chapter{Introduction \label{ch:intro}}
The human visual system has the ability to filter out the color cast caused by the dominating scene illumination~\cite{jameson1989essay, maloney1999physics}. This explains, in part, why an apple appears red under sunlight, incandescent light, and fluorescent light---even though these illuminations are significantly different in terms of their spectral profile. Camera sensors, however, do not have this ability and as a result, computational photography, or more specifically, computational color constancy (CC), is applied onboard cameras. In a photography context, this procedure is typically termed ``white balance'' (WB).

Most computational CC algorithms aim to achieve WB by correcting a scene's illumination to be ideal white light (i.e., the camera's sensor RGB responses to an achromatic object should lie along the achromatic ``white line'', that is R=G=B). In the literature, the scene illumination is usually assumed to be global and uniform (i.e., a single illuminant in the scene) \cite{gijsenijcomputational}. Under this assumption, the WB correction process is carried out using a simple $3\!\times\!3$ diagonal matrix in order to undo the effect of the estimated illuminant \cite{gijsenijcomputational}. 
One often overlooked issue with WB is that it is applied early in the processing chain directly to the sensor's RGB image---referred to as a raw image (or raw-RGB image). Cameras have a number of processing steps that convert the raw-RGB sensor response to the final output image. These collective steps result in a final output that is saved in a standard RGB (sRGB) image \cite{ramanath2005color, karaimer2016software}.

\begin{figure}[t]
\includegraphics[width=\linewidth]{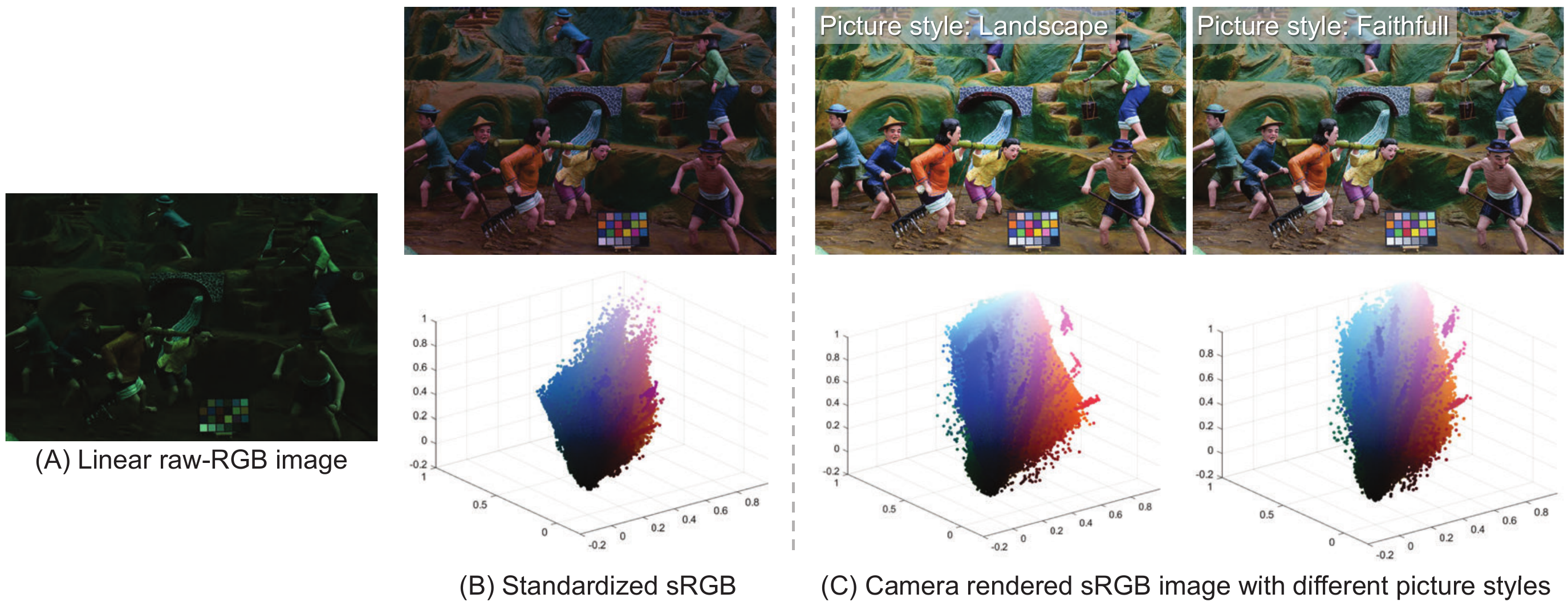}
\caption[This figure shows how the standardized sRGB \cite{anderson1996proposal} differs from what the cameras produce.]{This figure shows how the standardized sRGB \cite{anderson1996proposal} differs from what the cameras produce. (A) A raw-RGB image captured by Canon EOS-1Ds Mark III from NUS dataset \cite{cheng2014illuminant}. (B) Rendered sRGB image using a true sRGB encoding with only a single nonlinear gamma encoding applied~\cite{anderson1996proposal}. (C) Camera rendered sRGB images with different picture styles that include camera-specific nonlinear color rendering. For each sRGB image, we show the RGB histogram in the second row. It is clear that each camera is applying its own proprietary color manipulation.\label{fig:srgb-rendering}}
\end{figure}

While sRGB color space specifies a nonlinear gamma encoding as part of its encoding regime, cameras apply several other nonlinear operators that are not specified in the sRGB standard. These proprietary nonlinear color manipulations (also called photo-finishing or camera style) are typically unique to a particular make and model of a camera. Moreover, they are often tied to various camera settings used at the time the photo was captured. That means, for instance, the {\it Landscape} photo-finishing of a Canon camera applies a different color rendering than that applied by a Nikon camera even with a similar setting and imaging the exact same scene. Figure \ref{fig:srgb-rendering} shows an example of a raw-RGB image rendered to an sRGB image using a single gamma operation \cite{anderson1996proposal}. This image is compared to other images from cameras using their onboard picture styles. The camera sRGB-rendered images have notably different color distributions from the image's color distribution rendered by the sRGB \cite{anderson1996proposal}. Also, both camera rendered images have slightly different color distributions.

Given camera rendered final colors of photographs, it is nearly impossible to restore original linear scene-referred values (i.e., the raw-RGB image) without careful camera calibration to model the processing that was applied. Consequently, if the initial WB is computed incorrectly when capturing an image, it is challenging to undo afterwards in post-capture stage due to the nonlinear photo-finishing operations applied on the camera. In fact, WB errors are not the only camera errors that are hard to correct in post-capture stage. Exposure errors, for example, have a significant impact on the photographic quality of camera-rendered images; correcting such exposure errors in post-capture stage is more challenging than performing the correction in the linear sensor raw-RGB space as display-referred color spaces often have a smaller tonal range. Color correction and enhancement not only have crucial importance in photography aesthetics but also have a significant impact on different computer vision tasks, such as image retrieval, object tracking, texture classification, image forensics, and skin detection/classification \cite{gevers2000pictoseek, de2013exposing, yilmaz2006object, bianco2017improving, lui2020system, kakumanu2007survey}. 

\begin{figure}[t]
\includegraphics[width=\linewidth]{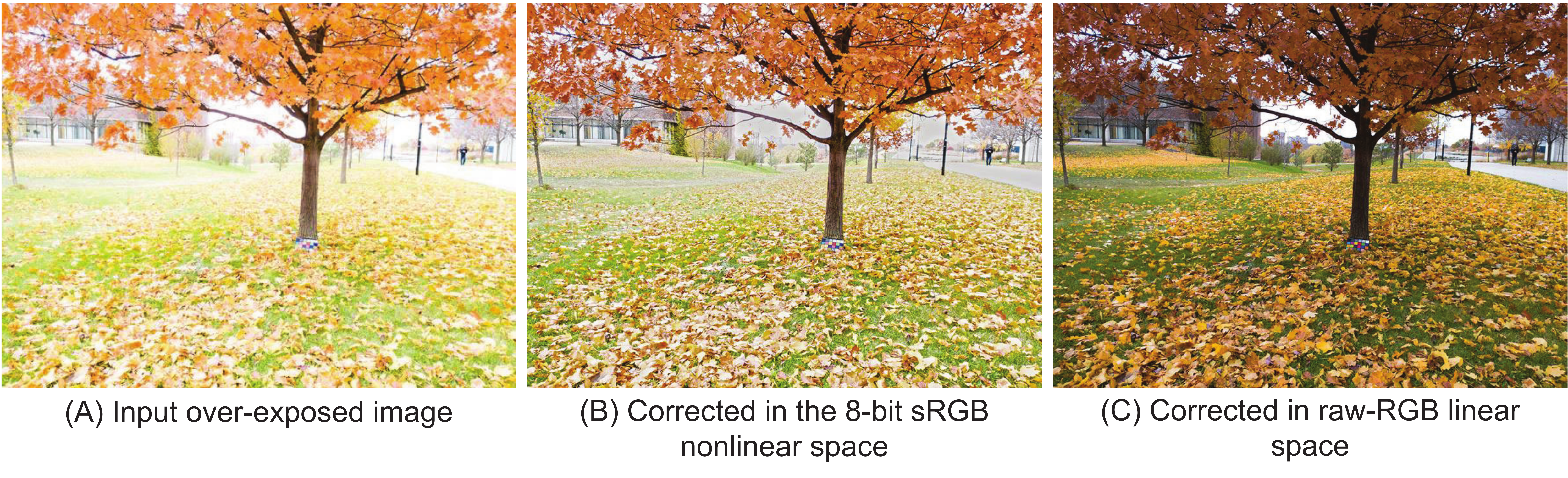}
\vspace{-9mm}
\caption[Correcting over-exposed photographs.]{Correcting over-exposed photographs. (A) Input over-exposed image. (B) Correction is applied in the 8-bit sRGB color space. (C) Correction is applied in raw-RGB linear space.\label{fig:intro_exposure_error}}
\end{figure}

Figure \ref{fig:intro_exposure_error} shows an example of post-capture color enhancement applied to an over-exposed photograph. As can be seen, the enhancement applied directly to the 8-bit sRGB display-referred image achieves a less pleasing result comparing with processing and re-rendering the linear raw-RGB space. 

Unfortunately, there is a misconception in the computer vision and image processing community that a simple inverse gamma operation can undo the nonlinear operations applied on such camera-rendered sRGB images. Instead, to properly undo the onboard color rendering, a detailed reverse engineering per camera is required. There is an entire research branch in computer vision, termed radiometric calibration, with the sole purpose to reverse the nonlinear processing applied within the camera pipeline (representative examples include \cite{grossberg2003what, kim2008robust, lin2004radiometric, lin2005determining, chakrabarti2014modeling, kim2012new, lin2011revisiting, xiong2012from}). Radiometric calibration is typically used for low-level computer vision tasks that require a linear response to scene radiance (e.g., photometric stereo, image deblurring, HDR imaging). When the necessary radiometric calibration data is available, it can be used to undo the photo-finishing in an sRGB image and effectively perform low-level computer vision tasks (e.g., WB correction) as demonstrated by~\cite{kim2012new}.  However, performing radiometric calibration requires a tedious calibrating procedure and as a consequence, radiometric calibration data is rarely available for most users.

\begin{figure}[t]
\includegraphics[width=\linewidth]{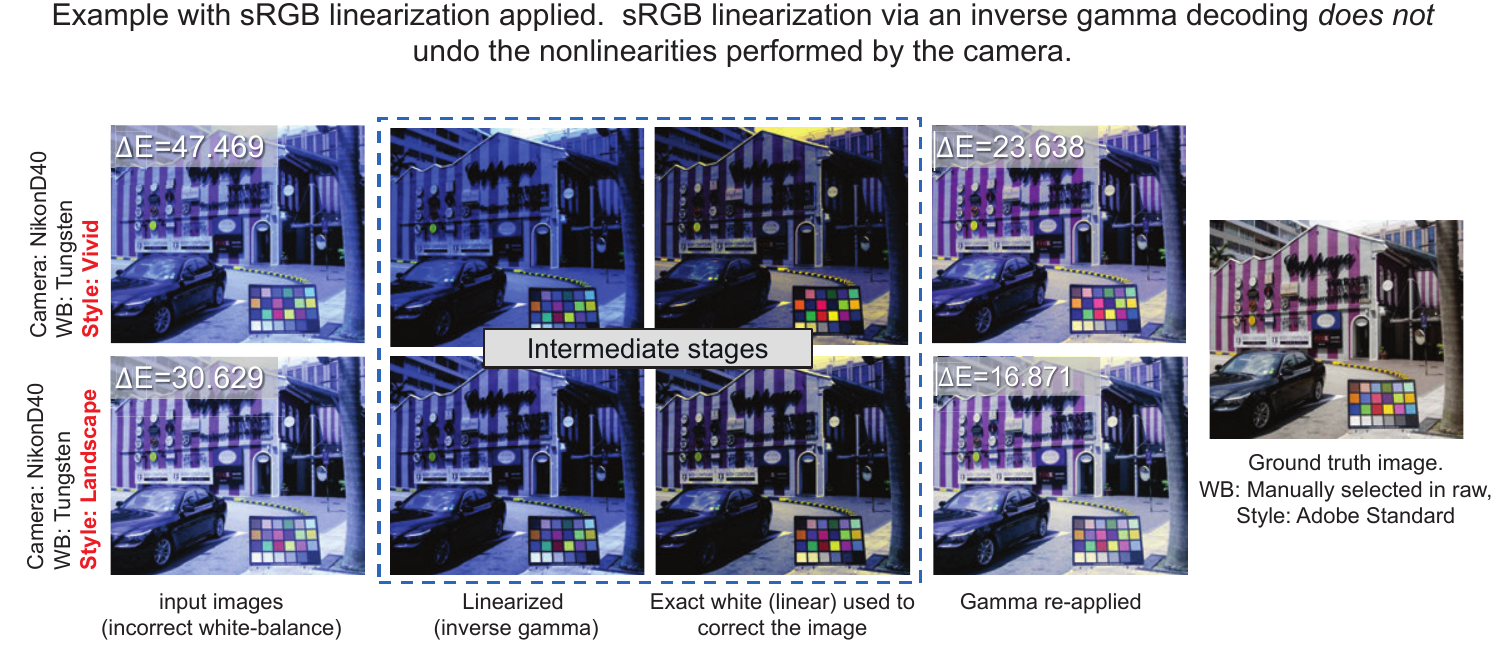}
\vspace{-8mm}
\caption[Shown are two input  images with incorrect WB rendered in sRGB with different picture styles.]{Shown are two input  images with incorrect WB rendered in sRGB with different picture styles. Intermediate steps with the \textit{inverse gamma} applied using the exact white values are shown. There are noticeable differences between the corrected images and the ground truth image that has been rendered to sRGB with the correct WB applied. This example is intended to show that trying to correct an sRGB image by first applying a simple gamma linearization is not sufficient. This strategy is a misconception commonly purported in the computer vision and image processing literatures.\label{fig:inverse-gamma}}
\end{figure}

To demonstrate the problem with the misconception that camera images only have a nonlinear gamma applied, we provide a visual example. Figure \ref{fig:inverse-gamma} shows two input images with incorrect WB rendered to sRGB with different camera picture styles. For each image, we apply the simple gamma linearization (i.e., we apply the inverse gamma specified by the sRGB encoding standard). After the gamma linearization, we apply a diagonal WB correction using the exact white values extracted from a color chart placed in the scene. Eventually, we re-apply the gamma operation to get the final sRGB image. There are noticeable differences between our corrected images and the ground truth image which is rendered with a correct WB applied to the original raw-RGB image before the photo-finishing step. 

\begin{figure}[t]
\includegraphics[width=\linewidth]{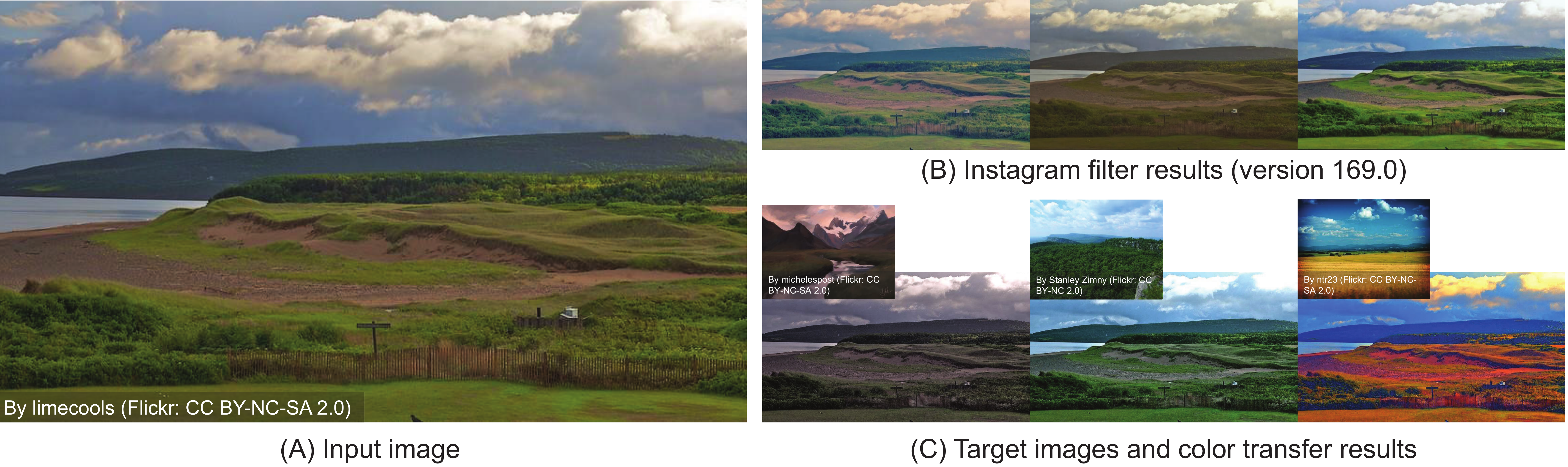}
\vspace{-7mm}
\caption[Example of static photo filter and dynamic color transfer results.]{Example of static photo filter and dynamic color transfer results. In this example, we recolored the input image in (A) using: Instagram photo filters, as shown in (B), and a color transfer method \cite{Pitie2007}  that accepts an additional target image to guide the recoloring process, as shown in (C).\label{fig:intro_color_filters}}
\end{figure}

While color correction and enhancement in photographs are more effective in scene-referred linear spaces (e.g., camera sensor raw-RGB space) than color processing in post-capture stage, post-capture color editing techniques can achieve impressive results in color editing to modify the photographic style of the captured image. One of the main goals of color editing is to transfer a new ``style and feel'' to the captured image. Photo filters that are widely offered by social media and smartphone applications achieve this goal by applying a pre-defined set of lookup tables (LUTs) to transfer a new style or feel to the input captured image. In the literature, there is a large body of work proposed for dynamic color mapping \cite{faridul2016colour}, where instead of relying on static LUTs, these methods propose more sophisticated approaches to achieve more accurate and compelling results of color mapping based on any arbitrary target colors. That is, these methods offer a more fixable way for color transfer comparing to popular LUT-based photo filters (see Fig.\ \ref{fig:intro_color_filters}). 

\section{Contributions}

This thesis provides five research contributions summarized as follows. First, we propose two methods for CC in camera raw-RGB images. Specifically, we propose two different lightweight deep learning models for CC. Unlike other CC deep learning models, our proposed methods are sensor-independent meaning that they can be deployed to new camera models without a need for retraining or calibration. 

Second, we propose the first framework that targets correcting improperly white-balanced camera-rendered images in post-capture. Our framework is enabled by generating a large dataset (over 60K images) of improperly white-balanced images rendered in the sRGB color space. We show that this framework, along with our dataset, can also be exploited to emulate WB errors in camera-rendered photographs and thus can be used to augment training data for different computer vision tasks. 

Third, three different methods for post-capture WB editing and manipulation are proposed. This part focuses in providing the user the freedom to edit the WB setting through interactive tools in post-capture stage. This WB editing and manipulation are required to match user preferences which do not always match accurate WB solutions.

Fourth, we propose two different methods for general color enhancement in photographs. Specifically, the goal of these methods is to enhance the colors of images rendered with exposure errors. As discussed earlier, correcting camera exposure errors in display-referred sRGB images is a challenging task due to the small tonal range of such sRGB images. This fact motivated us to propose a method for scene-referred image reconstruction. We show that this reconstruction can improve color enhancement of low-light and under-exposed camera-rendered images. Additionally, we generate a large dataset (over 24K images) with different exposure settings with broader exposure ranges. This generated dataset allows us to propose a deep learning method for directly correcting colors of under- and over-exposed photographs without a need to reconstruct such scene-referred images. 

Fifth, we discuss two different color editing techniques. Unlike our WB manipulation methods which provide global color editing in photographs, this part of our thesis aims at local image recoloring. Specifically, we propose model color distributions of several semantic objects in order to achieve object-aware image recoloring. We then extend this idea by proposing a generative adversarial network (GAN) to control colors in images. We show that our method can achieve auto image recoloring without a need for target images or any user interaction.

\section{Thesis Outline}
This thesis consists of eight parts. In Part \ref{part:intro}, we discuss the primary color rendering operations applied onboard cameras to render final sRGB photograph colors (Chapter\ \ref{ch:ch2}). Chapter\ \ref{ch:ch2} also discusses existing methods to linearize the sRGB image (i.e., from sRGB to linear RGB images) and their limitations. Next, Chapter\ \ref{ch:ch3} reviews prior work for color correction and manipulation. This survey discuss methods proposed for color correction (CC and WB), color enhancement in photographs rendered with exposure errors, and post-capture color manipulation. Part \ref{part:cc} of this thesis includes two chapters (Chapters \ref{ch:ch5} and \ref{ch:ch6}) that present two different sensor-independent illuminant estimation methods. Afterwards, we discuss our framework for correcting improperly white-balanced images in Part \ref{part:wb-correction}. We first outline the details of our WB correction framework in Chapter\ \ref{ch:ch7}. Then, we show how this framework can be extended to improve the robustness of computer vision tasks against WB errors in photographs (Chapter\ \ref{ch:ch8}). Part \ref{part:enhancement} includes two chapters (Chapters \ref{ch:ch9} and \ref{ch:ch10}) that present our methods for enhancing colors of under- and over-exposed photographs. The last contribution of this thesis is discussed in Part \ref{part:recoloring}, where our image recoloring methods are presented in Chapters \ref{ch:ch11} and \ref{ch:ch12}. Thesis conclusions and future work are discussed in Part \ref{part:conclusion} (Chapter\ \ref{ch:ch13}). Lastly, bibliography and appendices are given in Parts \ref{part:reference} and \ref{part:appendex}, respectively.

\chapter[Preliminaries]{Preliminaries \label{ch:ch2}}
The sRGB color space is the primary color space used to save images captured by digital cameras. As a result, sRGB images are the primary format used by many computer vision systems.  In the color science community, sRGB is known as an ``output-referred'' or ``display-referred'' color space as it is intended for use on display devices (monitors, LCD screens and even printers).   In this chapter, we present an overview of the formation of sRGB images through the lens of the camera imaging pipeline. To begin, we present a brief review of different standard color spaces. We then overview the in-camera image processing pipeline that is responsible for converting sensor raw-RGB images to the corresponding display-referred sRGB images. Lastly, we discuss the possibility of converting the sRGB colors back to the original linear format.

\section{Standard Color Spaces}

Colors are not physical characteristics of objects. Colors are words we use to describe sensations that arise from our perception of objects based on both the material characteristic of an object (e.g., surface reflectance and specularity) and the incoming visible light, which is within a certain band of the electromagnetic spectrum, called the visual spectrum\footnote{The visual spectrum is roughly from 380 to 780 nanometers (nm).} for typical human eyes \cite{ebner2007color}.

\begin{figure}[!t]
\begin{center}
\includegraphics[width=\textwidth]{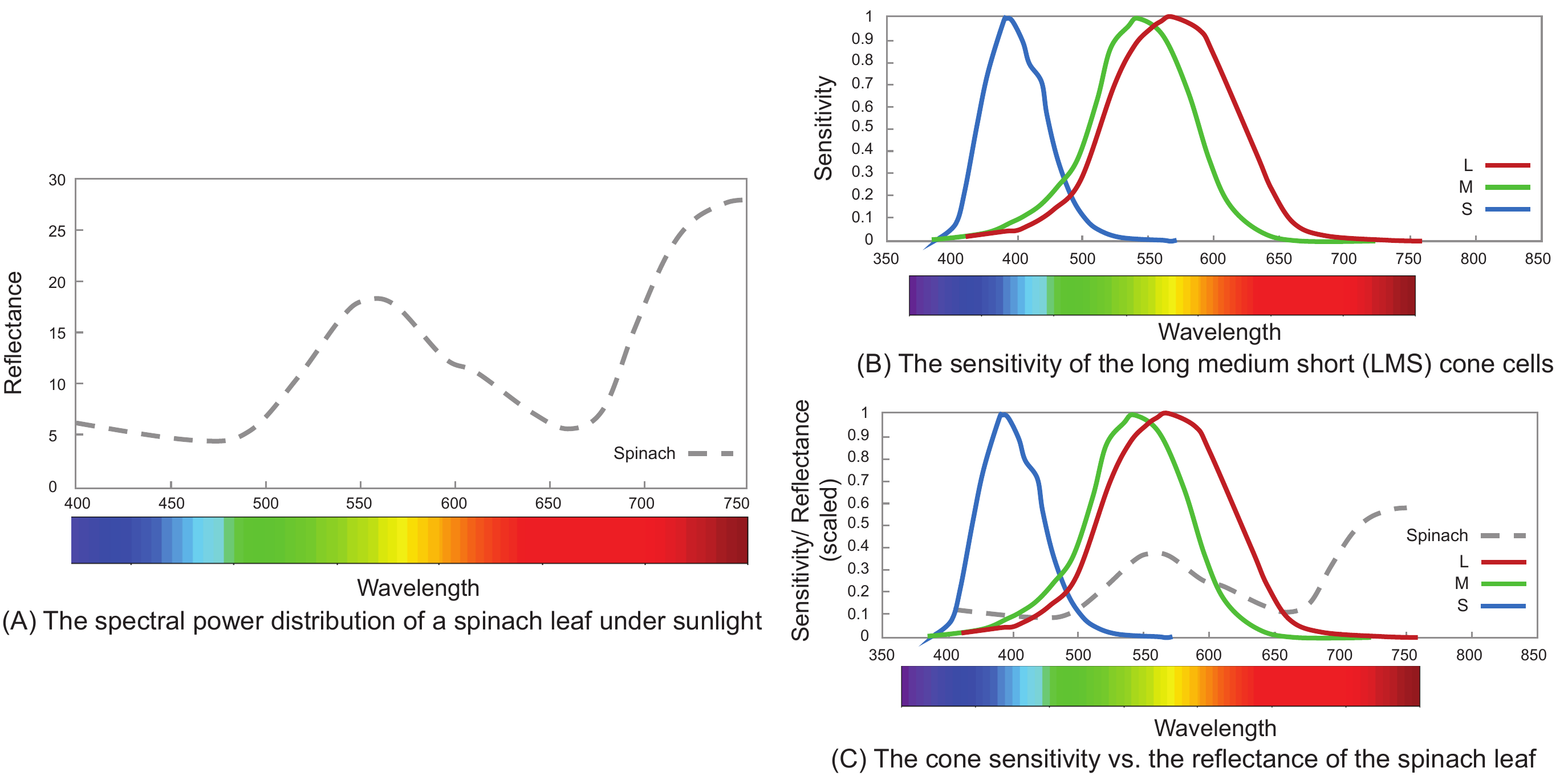}
\end{center}
\vspace{-5mm}
\caption[In this example, we show the spectral power distribution (SPD) of a spinach leaf captured under sunlight (A).]{In this example, we show the spectral power distribution (SPD) of a spinach leaf captured under sunlight (A). In the retina, there are three cone cells, namely long (L), medium (M), and long (L) cone cells. Each cone is sensitive for only a certain wavelength range that match a range of colors as shown in (B). The reason of why the spinach leaf appears greenish is that the reflected visible light of it mostly matches the medium cone cell, as shown in (C).} \label{fig:chapter2_1}
\end{figure}

When the human eye receives visible light spectra, there are three spectral sensors in the retina, known as cone cells, responsible for the color vision process. The cones were named according to the ordering of peak wavelengths, where the three cones are: (i) L: long cone (560-580 nm), (ii) M: medium cone (530-540 nm), and (iii) S: short cone (420-440 nm) \cite{fairchild2013color}. When these cones receive a visible light, their responses can be modeled by the following equation:
\begin{equation}
\label{eq:light}
c_i =\int_{\gamma} \rho(\lambda)s_{i}(\lambda) d\lambda,
\end{equation}
\noindent
where $\gamma$ is the visible spectrum, $s(\cdot)$ is the spectral sensitivity of the $i^{\text{th}}$ cone cell at wavelength $\lambda$, and $\rho(\cdot)$ is the incoming spectral power distribution (SPD) emitted or reflected from an object---this combines both the object's reflective properties and the scene's illumination. The final color is perceived after mixing the received colors by each cone, see Fig. \ref{fig:chapter2_1}.

According to Eq. \ref{eq:light}, two different SPDs can be perceived identically due to the accumulation effect of the three cones. This phenomenon is referred to as metamerism and color samples that are perceived identically under the same lighting conditions are called metamers. From Eq. \ref{eq:light}, we can also notice that any color can be matched by a linear combination of the three independent responses \cite{grassmann1853theorie}.

\begin{figure}[t]
\begin{center}
\includegraphics[width=\textwidth]{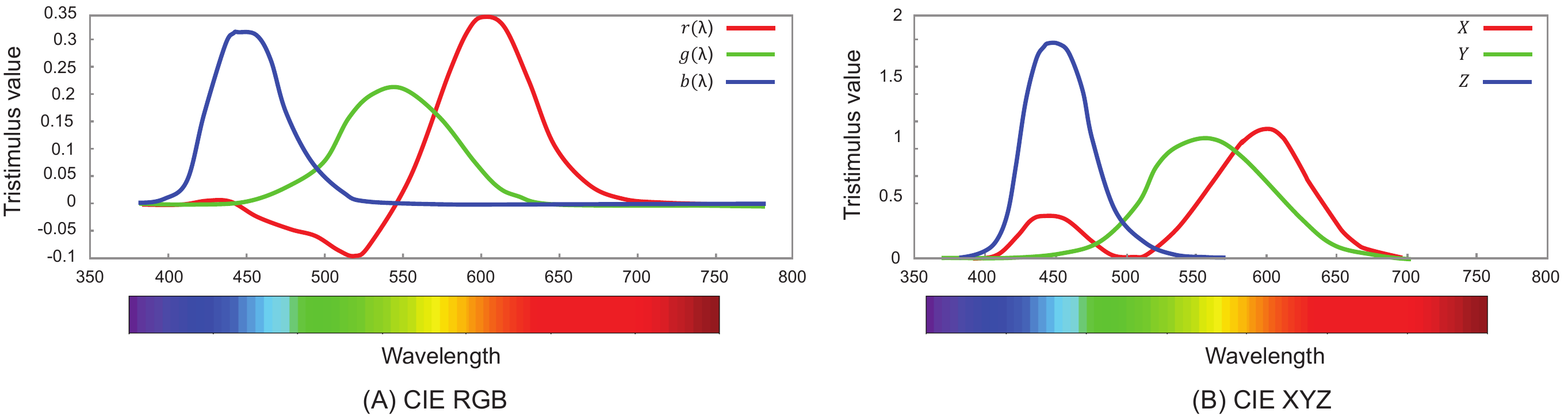}
\end{center}
\vspace{-4mm}
\caption{The tristimulus values through the visible spectrum of: (A) CIE RGB \cite{wright1929re} and (B) CIE XYZ spaces.} \label{fig:chapter2_2}
\end{figure}

Even before the three spectral sensitivities were physiologically discovered, a set of psycho-physical
experiments were carried out in order to establish a standard color space \cite{wright1929re}. These experiments determined the color mapping functions of a human observer through a mixing of relative amounts of three standard ``primary'' colors. This formed the basis of the CIE RGB color space---the term CIE refers to the Commission Internationale de l'Eclairage in French, also known as The International Commission on Illumination in English. One problem with the CIE RGB color space is that the proposed RGB primaries did not span the full visible range. As a consequence, some of the primaries were mixed in with negative values until matching the target colors, see Fig. \ref{fig:chapter2_2}-(A). To overcome this issue in CIE RGB color space, the CIE derived a new color space from the CIE RGB color space with no negative points, as shown in Fig. \ref{fig:chapter2_2}-(B). This color space was called 1931 CIE XYZ space \cite{cie1932commission} and is now widely accepted as a canonical device-independent color space \cite{agoston2013color}. The variables $X$, $Y$, $Z$ were used as they do not correspond to known color sensations. This canonical color space is also used to define standard illuminants based on their SPDs. For example, the standard CIE illuminant A represents incandescent light; while standard CIE illuminant series D is defined for natural daylight light. The latter includes different standard illuminants, such as D50 (horizon light) and illuminant D65 (noon light).

The CIE 1931 XYZ color space has remained the dominant color space used by the vast majority of imaging devices.  Many other color spaces common in academic and engineering research are derived directly from the CIE XYZ. These include NTSC, YUV, YIQ, CIE $\texttt{L}^{*}\texttt{a}^{*}\texttt{b}^{*}$, sRGB, and Adobe RGB color spaces. Among the existing color spaces, the 24-bit sRGB color model is the most dominant in consumer electronic systems. This color space was introduced in 1996 by Hewlett-Packard (HP) and Microsoft as a standardized and universal color space for all devices (e.g., monitors, printers, scanners, and digital cameras), through representing each color channel by 8 bits within the sRGB gamut \cite{anderson1996proposal}. The sRGB gamut is shown in Fig. \ref{fig:sRGBGAMUT}-(A). Converting the CIE XYZ values to the corresponding sRGB tristimulus values, according to the standard conversion introduced in 1996, can be performed by the following steps:
 \begin{itemize}
\item Converting from CIE XYZ values ($X$, $Y$, $Z$) to the corresponding linear sRGB values ($sR_l$, $sG_l$, $sB_l$) using the following equation:
\begin{equation}
\label{eq:fromxyz2rgb}
\begin{bmatrix} sR_l \\ sG_l \\sB_l\end{bmatrix}  = \mat{T}_{XYZ2sRGB}
\begin{bmatrix} X \\ Y \\Z \end{bmatrix},
\end{equation}

\noindent where $\mat{T}_{XYZ2sRGB}$ is a nonsingular transformation matrix that maps between the CIE XYZ and the corresponding linear sRGB values.

\begin{figure}[!t]
\includegraphics[width=\textwidth]{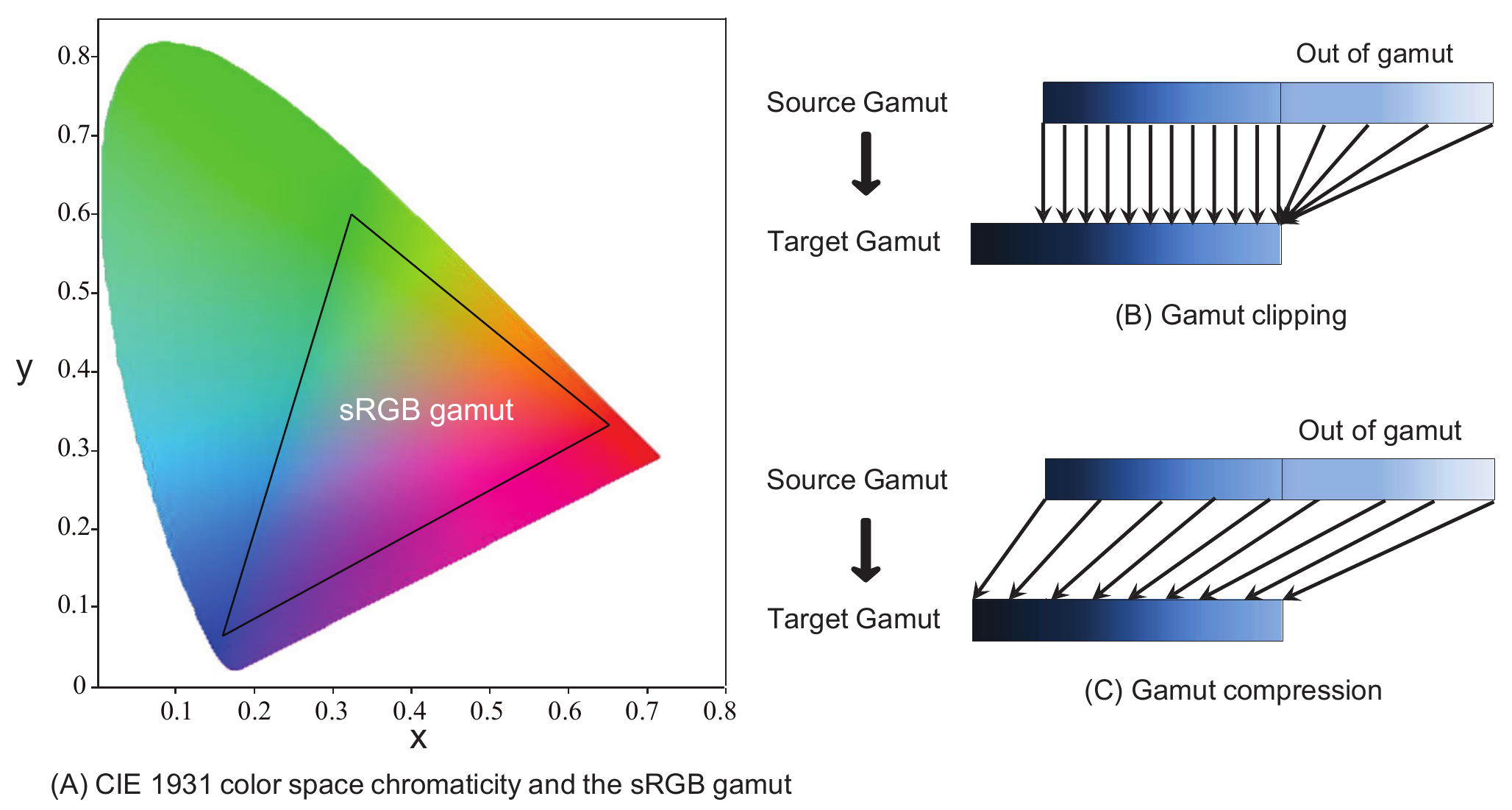}
\caption[The CIE 1931 x-y chromaticity diagram and the sRGB gamut.]{(A) The CIE 1931 x-y chromaticity diagram (i.e., a 2D projection of the CIE 1931 XYZ values onto the plane represented by $X+Y+Z = 1$) and the sRGB gamut. (B) and (C) show two gamut mapping approaches, namely gamut clipping and compression, respectively. This figure is adapted from \cite{haiting2013new}.} \label{fig:sRGBGAMUT}
\end{figure}

\item Converting from the linear sRGB values to the nonlinear sRGB values based on the following equation:

\end{itemize}

\begin{equation}
sR = \left\{\begin{matrix}
\hspace{35pt} sR_l \times 12.92 \hspace{90pt} \text{, if} \hspace{5pt} sR_l\leqslant 0.00304\\
\left(sR_{l}^{(1.0/2.4)} \times 1.055\right) - 0.055 \hspace{10pt} \text{, otherwise.}
\end{matrix}\right.
\end{equation}

The above equation is applied to the other linear color channels (i.e., $sG_l$ and $sB_l$) to get the final sRGB values. The result of this equation is represented by 24 bits for each color (8-bits/channel) with a close fit to the 2.2 gamma curve. The idea behind fitting the 2.2 gamma curve, also known as gamma encoding, is to exploit the nonlinearity of the human perceptual system whose sensitivity to brightness can be approximately represented by a power function---the human perceptual system is more sensitive to darker tones than lighter ones \cite{poynton2012digital}. In this way, the usage of 8 bits per channel is optimized in order to represent a wide range of different perceptual colors.

One important benefit of adhering to the standardized approach of converting CIE XYZ values to their corresponding nonlinear sRGB colors is the ability of computing the inverses of this operation to get the original CIE XYZ values---meaning that any two pixels with the same sRGB value should have the same CIE XYZ value, and consequently they represent, ``in theory'', exactly the same perceptual scene color. This conversion can simply be performed using the following equations:

\begin{equation}
\label{eq:invgamma}
sR_l = \left\{\begin{matrix}
\hspace{55pt}\left(sR/255\right)/ 12.92 \hspace{75pt} \text{, if} \hspace{5pt} sR/255 \leqslant 0.03928\\
\left(\left(sR/255 + 0.055\right)/1.055\right)^{2.4} \hspace{15pt} \text{, otherwise,}
\end{matrix}\right.
\end{equation}

\begin{equation}
\label{eq:fromsrgb2xyz}
\begin{bmatrix} X \\ Y \\Z\end{bmatrix}  = \mat{T}_{XYZ2sRGB}^{-1}
\begin{bmatrix} sR_l \\ sB_l \\sG_l \end{bmatrix}.
\end{equation}

Despite the success of the idea of having a standardized color space and the adoption of many devices of it (e.g., cameras, displays), no device follows this standard convention in reality, as will be explained in the next section.

\section{Camera Imaging Pipeline}
\label{pipeline}
\begin{figure}[!t]
\includegraphics[width=\textwidth]{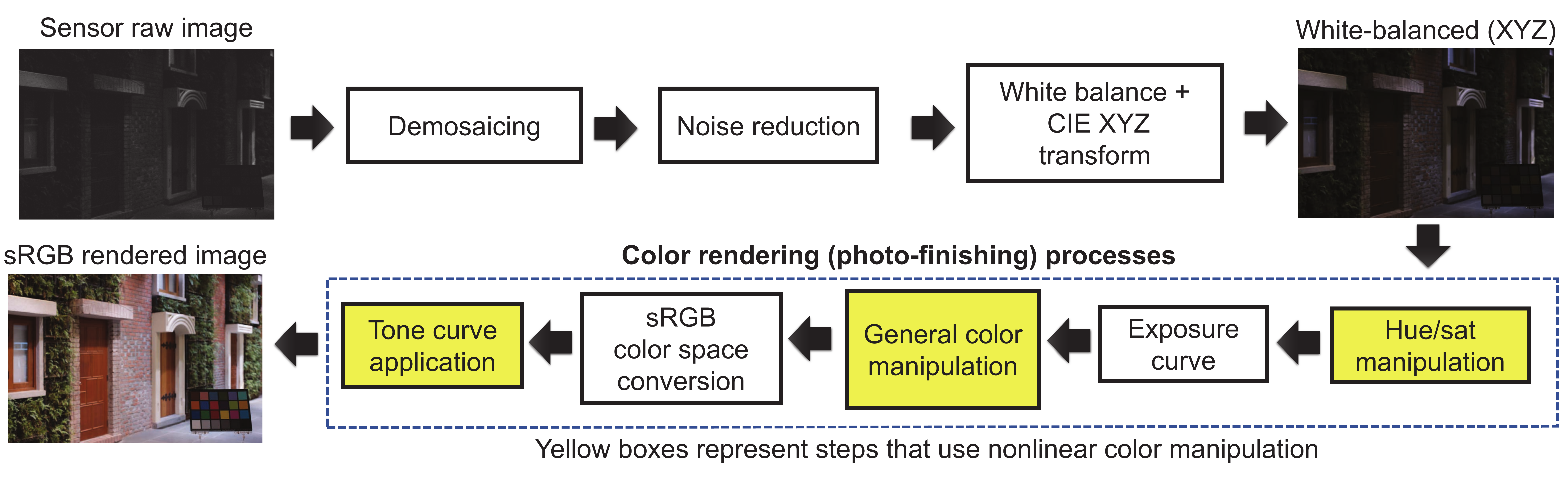}
\vspace{-7mm}
\caption[This figure shows a simplified the camera processing pipeline.]{This figure shows a simplified the camera processing pipeline, where a set of image processing/low-level computer vision operations are applied early in the pipeline (e.g., demosaicing, denoising, white balancing). Afterwards, a number of nonlinear color manipulations (highlighted in yellow) are applied to obtain the sRGB output image. Note that this arrangement and the details of the shown camera pipeline steps may differ based on the camera manufacturer and model producing different colors by each camera capturing the same scene based on its model and settings.\label{fig:camerapipeline}}
\end{figure}

Digital cameras apply a series of processing routines to convert a captured raw-RGB sensor image to the final sRGB output image.  These routines are part of the image signal processor (ISP) hardware on the camera.  While each camera manufacturer has its own customized ISP, researchers have developed a reasonably representative ISP model that includes the main components of camera imaging pipelines \cite{ramanath2005color, karaimer2016software}. Figure \ref{fig:camerapipeline} shows these main components. In the following part of this section, we will explain each component in detail.

\paragraph{Reading and Pre-processing the Raw Image}
The first input to the camera pipeline is the mosaiced raw image. This image contains digital values that are linear with respect to the amount of physical light irradiance that fell on the camera sensor for some given exposure. The sensor's photodiodes are covered with a color filter array (CFA) that is arranged on a square grid. The format of this CFA can vary based on the camera manufacturer, but the mosaiced Bayer pattern is widely used by digital cameras.

Most sensors have a number of defective pixels. Subsequently, defective pixels (e.g., dead/dark pixels or bright pixels, also called hot pixels) are masked out using a pixel mask.  The values for these corrupted pixels are interpolated based on their neighboring values. Due to thermal noise on the sensor, a pixel receiving no light might still output positive values which is known as the black level. Cameras perform a black-level subtraction and then normalize pixel values based on the maximum value (also known as the saturation level) to be in the [0-1] range using 10-16 bits. Due to the effect of the lens distortion within many cameras, flat-field correction is performed to reduce the vignetting effect.

\paragraph{Demosaicing} After the black-level and vignetting corrections, camera ISP performs an interpolation process to produce R, G, B raw digital values per pixel. This process is commonly referred to as demoasicing or debayering. This process can be performed using a simple nearest neighbor interpolation or using more sophisticated algorithms, such as pixel grouping \cite{pixelGrouping}, interpolation using alternating projections \cite{gunturk2002color}, or even using deep learning models \cite{liu2020joint}.  This process produces a full three channel raw-RGB output.

\paragraph{Noise Reduction} This step aims to reduce any noise that naturally occurs on the camera sensor.  A review of noise reduction is outside the scope of this thesis, but it should be noted that there is a large body of literature dedicated to this topic (for surveys on image denoising, see \cite{motwani2004survey, milanfar2013tour}).

\paragraph{White Balance and Color Space Conversion}
At this stage, WB correction is applied to the demosaiced raw-RGB image.
Some cameras allow the user to select a preset WB from the camera's settings, or more commonly an auto WB (AWB) routine is used. AWB involves estimating the scene illuminant (represented as a vector $\in \mathbb{R}^3$) and then applying the WB correction (typically is performed using a diagonal matrix based on the estimated illuminant vector). In the next chapter, we will discuss further details on the WB process. Based on the estimated illuminant color, the correlated color temperature is computed (more details are given in Appendix \ref{ch:appendix1}) and a colorimetric conversion is then applied\footnote{Some camera manufacturers omit this colorimetric conversion and convert white-balanced sensor colors to the sRGB space directly.} to map the white-balanced raw-RGB values to a canonical perceptual color space---namely, the CIE 1931 XYZ space. Usually this is performed using a $3\!\times\!3$ full color space transformation matrix (CST). Note that if the WB is applied incorrectly, the image will have a strong color cast and the resulting CIE XYZ values will be wrong.

\paragraph{Hue/Saturation Manipulation} Up to this stage, the processed image in the camera pipeline is in a canonical perceptual color space (CIE XYZ or one of its derivatives) and is directly associated to the physical scene image (i.e., scene-referred). However, this link is broken starting from this stage due to the camera-specific nonlinear operations that aim to produce a ``pleasing'' representation of the captured scene. The hue/saturation manipulation is one of these nonlinear operations and it is usually implemented as a 3D LUT.

\begin{figure}[!t]
\includegraphics[width=\textwidth]{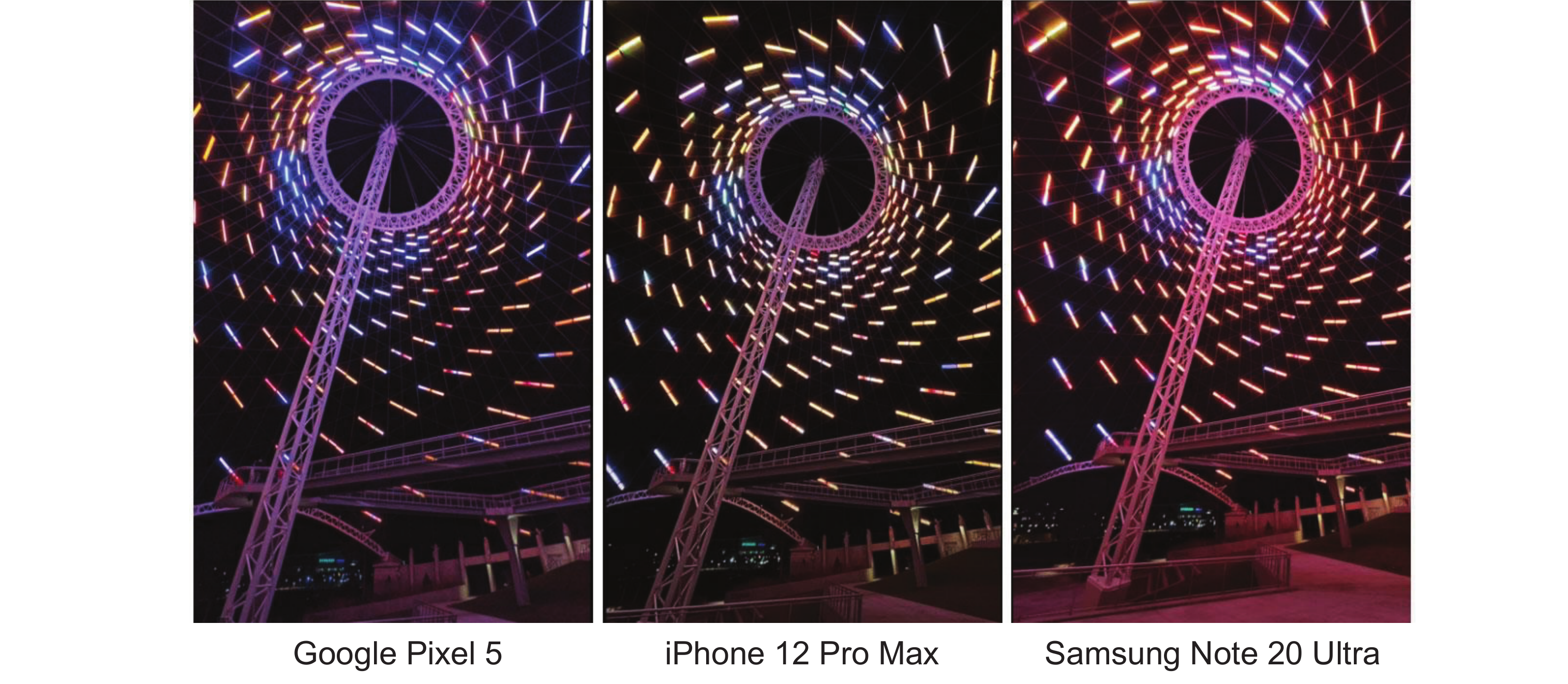}
\vspace{-7mm}
\caption[Three images captured the same scenes using different camera models, however each camera produced different colors.]{Three images captured the same scenes using different camera models, however each camera produced different colors. Photo credit: Max Tech on YouTube. \label{fig:diff_cam_responses}}
\end{figure}

\paragraph{Exposure Compensation} Beside the physical exposure control (i.e., shutter speed and aperture size), digital exposure can be applied to the pixel intensities using a simple linear gain.

\paragraph{General Color Manipulation} Each camera has its own color manipulation function that is usually represented by a 3D LUT and it is applied in order to get more visually pleasing colors.  

\paragraph{sRGB Color Space Conversion} At this stage, a $3\!\times\!3$ full matrix (e.g., $\mat{T}_{XYZ2sRGB}$) is used to convert pixel values from the previous stage to the final sRGB color space. During this stage, a gamut mapping operation is performed to map the out-of-gamut pixels to the sRGB gamut. The simple approach is gamut clipping \cite{gamutclip}, illustrated in Fig. \ref{fig:sRGBGAMUT}-(B), while gamut compression can be used for a better mapping, illustrated in Fig. \ref{fig:sRGBGAMUT}-(C).

\paragraph{Tone Curve Application}
Before rendering the final image, a camera-specific tone map operation is applied. This tone map may have different effects based on the selected camera style before capture. It is worth noting that this tone mapping operation may include local operations that are dynamically changed based on the current scene context \cite{nam2017modelling, hasinoff2016burst, HDRNET}.  Because the sRGB encoding standards recommends a 2.2 gamma encoding, it is often erroneously assumed that the gamma encoding is the tone curve. However, virtually no camera applies merely this simple gamma encoding \cite{ramanath2005color, kim2012new, karaimer2016software, hasinoff2016burst, nam2017modelling}.

Lastly, it is important to emphasize that the parameters, the arrangement, and the details of the previously described camera pipeline steps may differ based on the camera manufacturer and model producing different colors by each camera capturing the same scene based on its model and settings. Figure \ref{fig:diff_cam_responses} shows three images of the same scene captured by three different cameras. It is apparent that each image has different colors.

\section {From sRGB to Linear RGB}~As mentioned earlier, the camera imaging pipeline contains several set of nonlinear operations applied to generate a more visually pleasing sRGB image. Applying the standard inverse gamma operation (Eq. \ref{eq:invgamma}) is a long standing misconception found on wiki pages and provided by even well-known computing environments and libraries, such as Matlab and OpenCV, as a solution to linearize \textit{any} sRGB image. This is a serious problem not only for computing the CIE XYZ values, but also for converting the sRGB colors to any color space derived from the CIE XYZ (e.g., CIE $\texttt{L}^{*}\texttt{a}^{*}\texttt{b}^{*}$). Moreover, many of the photometric-based post-processing procedures, applied to sRGB images, are subject to a considerable amount of error \cite{nguyen2017you}. For these reasons, there is a large body of radiometric calibration literature to obtain a more accurate reconstruction of the linear RGB image.

By definition, radiometry refers to quantitative measurements of electromagnetic radiation (either the light source radiance or the surface irradiance). Radiometric calibration aims to invert the nonlinear operations applied to the sRGB-rendered images to reconstruct the image irradiance $\mat{I}$ weighted by the spectral sensitivities of the R, G, B filters on the camera \cite{grossberg2004modeling}. In other words, the goal of radiometric calibration is to model the camera response function (CRF), $f_{\text{CRF}}: \mat{I}\rightarrow \mat{I}_{\texttt{sRGB}}$, and the model of response (MoR), $f_{\text{CRF}}^{-1}: \mat{I}_{\texttt{sRGB}} \rightarrow \mat{I}$.

\begin{figure}[!t]
\includegraphics[width=\textwidth]{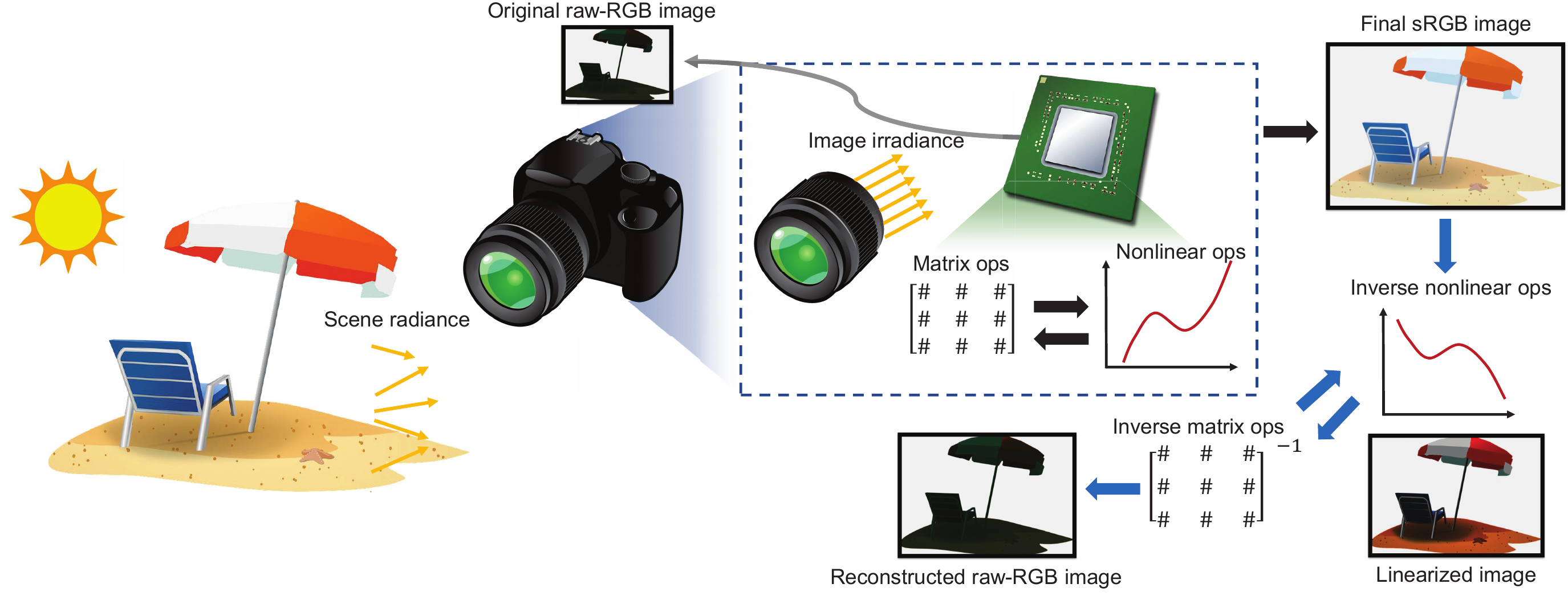}
\vspace{-7mm}
\caption[Reconstruction of raw-RGB images requires inverting the operations performed in the sRGB image formation in order to reconstruct a demosaiced form of the image irradiance.]{Reconstruction of raw-RGB images requires inverting the operations performed in the sRGB image formation in order to reconstruct a demosaiced form of the image irradiance. The linearization process aims only to undo the nonlinear effects applied onboard the camera to obtain a linearized representation of the image.} \label{fig:radiometric}
\end{figure}

Most of the conventional radiometric calibration algorithms approximate the CRF to linearize the sRGB image. This linearization process, however, does not reconstruct the original raw-RGB image \cite{kim2012new}. Instead, its goal is to invert the nonlinear functions applied onboard the camera without considering the effects of the other components in the camera imaging pipeline (e.g., WB, color space conversion, or gamut mapping); see Fig. \ref{fig:radiometric}. As a response, we organize this section into two parts. First, we discuss the \textit{full} reconstruction process of the raw-RGB image. Second, we review the main strategies of the existing radiometric calibration methods.

\subsection[Raw-RGB Image Reconstruction]{Raw-RGB Image Reconstruction}

A simplified sRGB image formation model can be represented by the following equation:

\begin{equation}\label{raw_srgb}
\mat{I}_{\texttt{sRGB}} = f_{\text{CRF}}(\mat{I}).
\end{equation}
We can formulate the problem of reconstructing the raw-RGB image as:

\begin{equation}\label{srgb_raw}
\mat{I} = f_{\text{CRF}}^{-1}(\mat{I}_{\texttt{sRGB}}).
\end{equation}

Due to the complexity of the nonlinear camera pipeline's processes, it is non-trivial to find $f_{\text{CRF}}^{-1}$ without having a prior knowledge of the camera model and the capture settings.  To perform proper radiometric calibration, it is necessary to capture many images of a color rendition chart, or other calibration device, under a controlled environment. One approach could be to measure the camera responses inside a white sphere to a varying incoming light in order to model the correspondence between scene radiance and measured pixel values \cite{szeliski2010computer}. Simpler solutions can be given by capturing a color rendition chart under uniform lighting conditions \cite{grossberg2003determining}.

Grossberg and Nayar \cite{grossberg2004modeling} found that real-world cameras have a bounded space of CRFs. Thus, they proposed an empirical model of response (EMoR) obtained based on 201 real camera responses. In their work, they approximated the CRF as a nonlinear response of the camera without considering other factors (e.g., WB) through a brightness transfer function (BTF) $f_{\text{BTF}}$ that maps the image from some linear form to the final pixel brightness in the sRGB space. Their EMoR is represented as a principle component analysis (PCA)-based model of $f_{\text{BTF}}^{-1}$. Specifically, the EMoR is formatted by the following equation:

\begin{equation}\label{PCA_prior}
\hat{f}_{\text{BTF}}^{-1} = \mu(f_{\text{BTF}}^{-1}) + \mat{L}\mat{b},
\end{equation}

\noindent where $\mu(\cdot)$ represents the mean of the BTFs of different real-world cameras, $\mat{L}$ is a matrix whose columns contains the first $g$ eigenvectors, and $\mat{b} \in \mathbb{R}^{g}$ is the PCA coefficient vector. This EMoR representation gives the ability to approximate the complete BTF using a fewer number of parameters (e.g., $g = 5$ used in \cite{lin2004radiometric, li2017radiometric}).

Kim \textit{et al.} \cite{lin2011revisiting, kim2012new} proposed a new in-camera imaging model in order to reconstruct an accurate raw-RGB image from the given sRGB image. In their model, they can effectively reconstruct the MoR by formulating the problem as follows:

\begin{equation}\label{accurate_representation}
\mat{I}_{\texttt{sRGB}} = f_{\text{CRF}}(\mat{I}) =  f_{\text{BTF}}(h(\mat{I} \text{ } \texttt{diag}(\mat{\light}^{*}) \mat{T}_\text{CAM})),
\end{equation}
\noindent where $h(\cdot)$ is a nonlinear 3D gamut mapping function, $\texttt{diag}(\mat{\light}^{*})$ is a $3\!\times\!3$ diagonal matrix for WB correction, and  $\mat{T}_\text{CAM}$ is a $3\!\times\!3$ full color transformation matrix that converts from the camera space to the linear RGB space---this matrix combines the CAM and the XYZ-RGB conversion matrix.  This formulation allowed them to obtained an accurate reconstruction of raw-RGB images by calibrating the camera model in order to define $\mat{T}_\text{CAM}^{-1}$, $\texttt{diag}(\mat{\light}^{*})^{-1}$, $h^{-1}$, and $f_{\text{BTF}}^{-1}$. This calibration process was performed in a three-stage manner. First, they estimated $f_{\text{BTF}}$ based on the PCA model of camera responses \cite{grossberg2004modeling} from non-saturated pixels that are unaffected by the gamut mapping process. Second, they calibrated the $\mat{T}_\text{CAM}$ and $\texttt{diag}(\mat{\light}^{*})$ matrices from the linearized sRGB values after applying the calibrated $f_{\text{BTF}}^{-1}$. Third, $h^{-1}$ is modeled by a non-parametric model based on a point interpolation using radial basis functions
(RBFs). Xiong \textit{et al.} \cite{xiong2012from} extended this idea to provide a distribution of the possible raw-RGB image colors using a probabilistic model with an uncertainty prediction. A more recent deep learning-based solution was proposed by Nam and Kim~\cite{nam2017modelling}. Their model relies on the image's scene context and color distribution in order to reconstruct the original raw-RGB image.

Despite the accurate reconstruction, these methods require the presence of the camera models used to capture the sRGB image in order to either calibrate the camera \cite{lin2011revisiting, kim2012new} or train a reconstruction model \cite{nam2017modelling}.

Recently, Jiang \textit{et al.} \cite{jiang2017learning} showed that the camera imaging pipeline can be represented as a set of affine transformation matrices to map the raw-RGB image to the sRGB image. They achieved this by clustering the sensor data into various classes based on their spatial and color information followed by learning the transformation matrix for each class.

Nguyen and Brown \cite{nguyen2018raw} proposed to embed a small memory overhead to keep the necessary metadata in order to reconstruct the original raw-RGB image. This metadata includes information for tone mapping, WB, color space transform, saturation pixels, and the color manipulation for gamut mapping. To keep the overhead small, some assumptions and approximations were made. First, they assumed the tone mapping operation affects only the chromatic colors, so they employed only the V channel of the HSV color space to reduce the required information for the reconstruction process. The 3D sRGB color histogram was approximated by scattered points in order to undo the gamut mapping operation. Lastly, the original values of the saturated pixels are saved to avoid the problem of overexposed pixels. At the end, their approach effectively encoded useful metadata for reconstructing the raw-RGB image from the sRGB-JPEG image. Unfortunately, such metadata is not provided by the existing camera models \cite{punnappurath2019learning}.

\subsection{Radiometric Calibration}

\begin{figure}[!t]
\includegraphics[width=\textwidth]{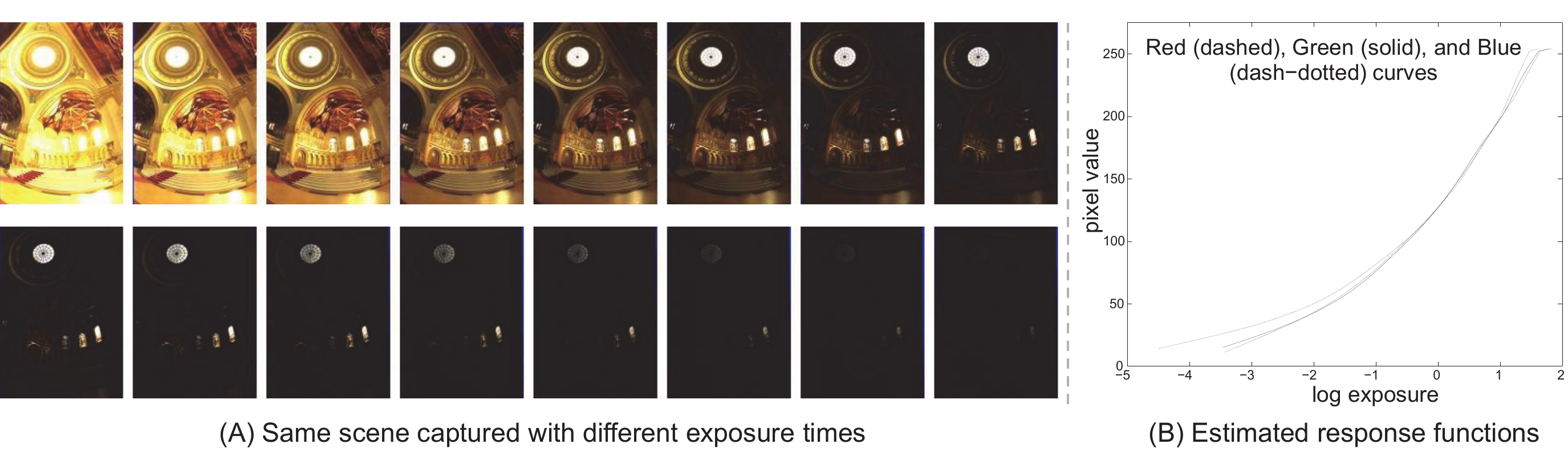}
\vspace{-7mm}
\caption[Reconstructed response functions from multiple images captured with different exposures using Canon 35mm SLR camera.]{Reconstructed response functions from multiple images captured with different exposures using Canon 35mm SLR camera, adapted from~\cite{debevec1997recovering}. (A) Aligned sRGB images captured with different exposure times. (B) Estimated response functions for Canon 35mm SLR camera by Debevec and Malik~\cite{debevec1997recovering}. } \label{fig:radio_1}
\end{figure}

The settings required to reconstruct the original raw-RGB image are tedious and impractical in many scenarios. As a result, most of the conventional radiometric methods try to model BTFs instead of the complete CRFs. In other words, they aim to undo the effect of the BTF without the need for a calibration object in order to reconstruct a radiometrically linear representation $\mat{I}_{l}$ of the sRGB image $\mat{I}_\texttt{sRGB}$ rather than the original raw-RGB image $\mat{I}$. Now, the problem is usually formulated as the following equation \cite{kim2008radiometric}:

\begin{equation}\label{radio_eq}
\mat{I}_{\texttt{sRGB}} = f_{\text{BTF}}( \varphi \mat{I}_{l} \mat{\light}),
\end{equation}

\noindent where $\varphi$ is the exposure value used during capturing $\mat{I}_{l}$. One solution is to use multiple aligned sRGB images of the same scene with different known exposure values under a constant lighting condition to construct a matrix of the ``brightness'' values. This matrix can be used to reconstruct the linearized image by fitting the brightness/exposure data \cite{mann1994beingundigital, debevec1997recovering}.
Figure \ref{fig:radio_1} shows an example of the nonparametric recovered BTF obtained by Debevec and Malik \cite{debevec1997recovering}. Based on the  idea of having varying exposure, many modifications were proposed (e.g., iterative polynomial-based solving \cite{mitsunaga1999radiometric}, spatially varying optical mask \cite{nayar2000high}, prior-based probabilistic model \cite{pal2004probability}).

Another solution is based on obtaining a set of images of the same scene under different lighting conditions by employing the EMoR \cite{grossberg2004modeling} to estimate the PCA coefficients using pixels with the same lighting conditions in the scene \cite{kim2008radiometric}. The vignetting function also can be integrated into Eq. \ref{radio_eq} as well, while assuming a radial symmetry vignetting to estimate the radiometric response function from a sequence of sRGB images \cite{kim2008robust}.

The main limitation of these methods is the need to capture a set of images under certain settings. Also, since the BTF can be a scene-dependent nonlinear function, the assumption of a fixed response function per channel is not sufficient \cite{grossberg2004modeling}. As a solution to all these problems, scene geometric calibrations were performed for a large set of training images to benefit from the estimated normal vectors of scene surfaces \cite{diaz2011radiometric, mo2017radiometric}. This enables the ability to estimate the response function of a given sRGB-rendered image. However, these methods require a large amount of training data that is pre-geometrically calibrated.

\section{Summary}

In this chapter, we have provided an overview of the sRGB color space and its formation. We have explained the difference between the standardized approach for generating sRGB colors and the existing camera imaging pipeline operations. We have shown that the current camera models apply a sequence of nonlinear operations in order to make the captured scene more pleasing regardless of the effects on the relation between the image colors and the real scene colors. These nonlinear operations make it hard to reconstruct the original linear colors. Accordingly, a full radiometric calibration is required. However, we have discussed how full radiometric calibration requires tedious image processing steps  or the embedding of necessary metadata to help in the reconstruction process. We have also reviewed representative examples of radiometric calibration methods that aim to linearize the sRGB image in a more efficient way than the simple inverse gamma operation. We have shown that these models did not consider the effect of main components of the camera imaging pipeline. Additionally, they require certain conditions to work properly.

\chapter{Prior Work \label{ch:ch3}}
This chapter reviews prior work proposed for color correction and editing in photographs. Specifically, we discuss in more detail the image white balancing process, which is one of the major procedures that are responsible for color correction and manipulation on board cameras (Sec.\ \ref{sec:image_whitebalancing}). Then, we will elaborate on why correcting colors in the post-capture stage is more challenging; especially, if the camera-rendered images have some errors in WB (Sec.\ \ref{Sec:white_balance}). Afterward, we will discuss other factors that directly contribute to the quality of camera-rendered image colors. In particular, we will discuss exposure errors in cameras and how such errors have a significant impact on the final rendered colors by cameras (Sec.\ \ref{sec:exposure_errors}). Lastly, we will briefly review post-capture image color editing techniques (Sec.\ \ref{sec:post_capture_color_manipulation}). 

\section{Image White Balancing} \label{sec:image_whitebalancing}

WB is applied as an approximation to color constancy (CC), described earlier in Chapter~\ref{ch:intro}, that is the term given to the human visual system's ability to perceive an object's color as the same when viewed under different illumination~\cite{Gilchrist07seeing}. Camera sensors lack this ability and unprocessed raw-RGB camera images contain noticeable color cast due to the scene's illumination.  WB, or more generally computational CC, is a fundamental processing step applied onboard cameras to compensate for scene illumination.

\begin{figure}[!t]
\begin{center}
\includegraphics[width=\linewidth]{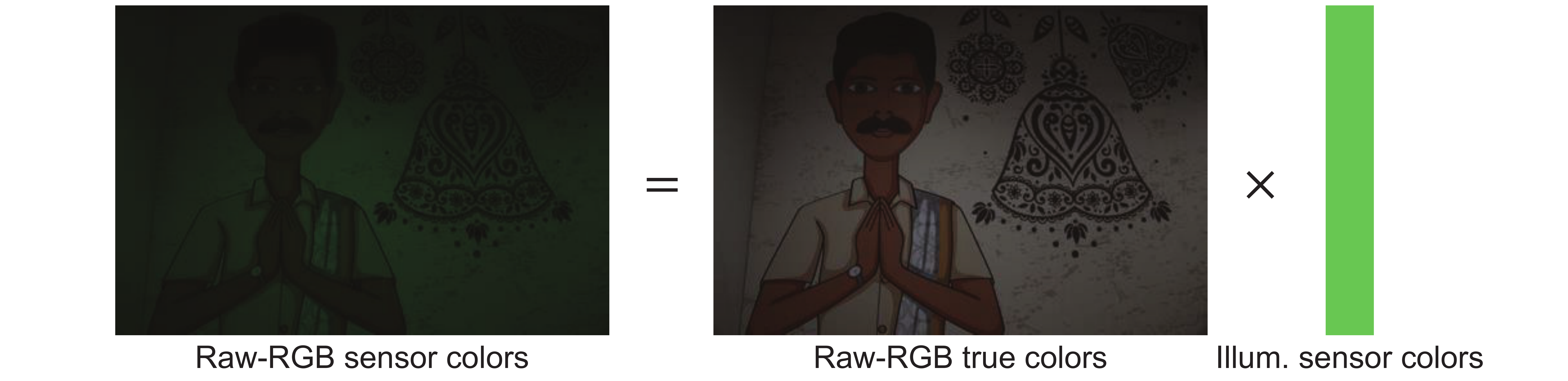}
\end{center}
\vspace{-4mm}
\caption[Problem formulation of color constancy with the assumption of a single uniform lighting condition.]{Problem formulation of color constancy with the assumption of a single uniform lighting condition. The shown image is from the INTEL-TAU dataset \cite{laakom2019intel}.\label{fig:intro_wb_illustration}}
\end{figure}

We can formally describe WB in terms of the image formation process.
Let $\mat{I} = \{\mat{I}_r, \mat{I}_g, \mat{I}_b\}$ denote an image captured in the linear raw-RGB space. The value of each color channel $c = \{\text{R}, \text{G}, \text{B}\}$ for a pixel located at $x$ in $\mat{I}$ is given by the following equation \cite{basri2003lambertian}:

\begin{equation}
\label{eq0}
\mat{I}_c(x) =\int_{\gamma} \rho(x,\lambda)R(x,\lambda)S_{c}(\lambda) d\lambda,
\end{equation}
\noindent
where $\gamma$ is the visible spectrum, $\rho(\cdot)$ is the illuminant spectral power distribution, $R(\cdot)$ is the captured objects' body reflectance (i.e., diffuse reflection component), and $S(\cdot)$ is the sensor response function at wavelength $\lambda$. According to this simple model, the surface appears the same from all viewing directions---assuming there is no specular reflection. The problem can be simplified more by assuming a single uniform illuminant. Hence, the problem can be written as:

\begin{equation}
\label{eq1}
\mat{I}_c =  \mat{R}_{c} \times \mat{\light}_{c},
\end{equation}
\noindent
where $\mat{\light}_{c}$ is the color channel $c$ of this single illuminant (see Fig.\ \ref{fig:intro_wb_illustration}. We assume that black-level subtraction has been applied to $\mat{I}$. Now, the problem can be solved by a simple linear model (i.e., a $3\!\times\!3$ diagonal matrix) to make $\mat{\light}_{\text{R}} = \mat{\light}_{\text{G}} = \mat{\light}_{\text{B}}$ (i.e., white illuminant).

Typically, $\mat{\light}$ is unknown and should be estimated from the \textit{linear raw-RGB images}. Illumination estimation is one of the fundamental processes performed onboard cameras as a part of their WB feature. Illumination estimation methods predict the color of the scene's illumination from a captured image in the form of an R, G, B vector in the sensor's raw-RGB color space. 

The straightforward  way to do this is to capture an image of an object that acts as a pure reflector---for example, an achromatic (i.e., white or neutral) object.  Under ideal white light, the camera's sensor response to the achromatic object should lie along the achromatic ``white line''; that is, R=G=B.  The scale of the R, G, B values depend on the intensity of the reflected light from the object.  Under non-white illuminations, the camera's response to a pure reflector would \textit{not} lie along the achromatic line and the R, G, B responses would, therefore, represent the measurement of the illumination in the sensor's color space.

In this section, we present a  survey on the existing WB methods from raw-RGB images. Then, we examine the ability of existing WB techniques to correct improperly white-balanced sRGB-rendered images.

The WB process consists of two steps: (i) estimating the color of the illumination in the camera's sensor space and (ii) correcting the image based on the estimated illumination.

\subsection{Illuminant Estimation}\label{Sec:illuminant_estimation}

In practice, we do not always have neutral patterns in our scenes and the color of the illumination must instead be estimated directly from captured images. This illuminant estimation is a challenging problem, because it is fundamentally under-constrained: an infinite family of white-balanced images and global color casts can explain the same observed image. Illuminant estimation is, therefore, often framed in terms of inferring the most likely illuminant color given some observed image and some prior knowledge of the spectral properties of the camera's sensor. 

We can categorize illumination estimation methods into two different categories: (A) single-illuminant scene estimation and (B) multi-illuminant scene estimation. 

While there are few attempts proposed for illuminant estimation of multi-illuminant scene (e.g., \cite{finlayson1995color, barnard1997color, ebner2004color, kawakami2005consistent, riess2011illuminant, beigpour2014multi, hussain2018color}), the majority of prior work adopted the single illuminant assumption. Generally, single-illuminant scene estimation methods fall roughly into four categories: (i) statistical methods, (ii) physics-based methods, and (iii) learning-based methods.

\subsubsection{Statistical-Based Methods} Statistical-based methods operate using statistics from an image's color distribution and spatial layout. Most statistical-based methods are based on one or more assumptions in order to apply a set of generic statistics to estimate the illuminant color vector.

The \textit{gray world} (GW) assumption \cite{GW}, for instance, assumes that the mean of image irradiance is achromatic (i.e., ``gray''). That is, the algorithm computes the mean of the given raw-RGB image in order to estimate the illuminant color (i.e., $\overline{\mat{I}_c} \propto \mat{\light}_{c}$). Smoothing the camera responses can be applied to compute an initial local averaging before employing the gray world assumption \cite{gijsenijcomputational}. This smoothing operation is usually performed using a Gaussian filter. This method is called general gray world (GGW). Potential improvements can be achieved by using a weighted gray world, such that the mean of each color channel is adapted based on its standard deviation \cite{lam2004automatic, pan2014improved}.

From another perspective, other methods assumed the presence of white objects with larger intensity values than other pixels in the captured scene. This assumption is called the \textit{white patch} hypothesis which can be implemented by computing the maximum response of each color channel (i.e., max-RGB) \cite{maxRGB}. This hypothesis is extended later in the bright pixels algorithm (BP) \cite{BP} by considering the gamut of the bright pixels instead of the simple max-RGB method. Finlayson and Trezzi \cite{SoG} showed that the max-RGB and gray world algorithms are special cases of a more generic algorithm for computational CC, referring to it as shades of gray (SoG), which assumes the mean of the Minkowski $p$-norm of the scene is shades of gray. Specifically, the gray world and max-RGB algorithms can be represented by the following equation:

\begin{equation}
\label{Eq.shades}
\frac{{\lVert\mat{I}_c\rVert}_p}{N^p} \propto \mat{\light}_c,
\end{equation}

\noindent where ${\lVert \cdot \rVert}_p$ is the $p$-norm and $N$ is the total number of pixels in $\mat{I}$. For $p=1$, Eq. \ref{Eq.shades} represents the gray world algorithm, while if $p=\infty$, Eq. \ref{Eq.shades} computes the max-RGB. For $p \in ]1,\infty[$, the equation refers to the SoG algorithm. They found that the best results obtained with $p=6$.

Unlike the previous assumptions which rely merely on the color information, \textit{gray edges} (GE) assumption assumes that the mean reflectance differences in a scene is achromatic \cite{GE}. In this context, Eq. \ref{Eq.shades} can be modified to be:

\begin{equation}
\label{Eq.grayedges}
\frac{{\lVert\bigtriangledown\texttt{blur}(\mat{I}_c,\sigma_b)\rVert}_p}{N^p} \propto \mat{\light}_c,
\end{equation}

\noindent where $\bigtriangledown$ denotes the gradient magnitude of the Gaussian blurred version of $\mat{I}_c$ with standard deviation $\sigma_b$. Similarly to the weighted gray world, the weighted gray edges (wGE) algorithm~\cite{gijsenij2012improving} assigns weights to the image's edges based on their  photometric properties (i.e., shadow edges, material edges, etc.) to improve the accuracy.

In contrast, Cheng \textit{et al.} \cite{cheng2014illuminant} showed that relying only on the color distribution of the image is sufficient to estimate the illuminant color without the need for any spatial information in the image. This work shows that the reason behind the spatial-based methods' success is the ability of obtaining large color differences from edges in the image's scene content. Hence, they showed that the scene illuminant can be obtained from the vector which maximizes the variance of the projected dark and bright pixels into one dimension (i.e., the first PCA vector).

The advantages of the statistical-based methods can be summarized as follows: (i) simplicity, (ii) speed, and (iii) few number of parameters---mostly less than three parameters that are usually fine-tuned for each camera model.

Despite the considerable merits of the statistical-based methods, their results are not always satisfactory. Zakizadeh \textit{et al.} \cite{zakizadeh2015hybrid} experimentally showed that most statistical-based methods have a systematic failure due to the reliance on the statistics of scene content. In particular, they showed that there are certain types of images that are consistently difficult for different statistical-based methods including GW, max-RGB, SoG, and GE. There are a few attempts to improve the accuracy of such statistical-based methods by introducing a post-correction transformation to correct the ``bias'' error produced by these methods (e.g., \cite{MomentCorrection, royalsociety, afifi2019projective}). However, such methods requires a large set of training data with corresponding ground-truth illuminant colors to \textit{learn} the bias-correction function. With the reliance on labeled training data, we can think of these post-correction methods as a learning-based mechanism. We will discuss prior learning-based CC methods later in this chapter.

\subsubsection{Physics-Based Methods}
\begin{figure}[!t]
\begin{center}
\includegraphics[width=\linewidth]{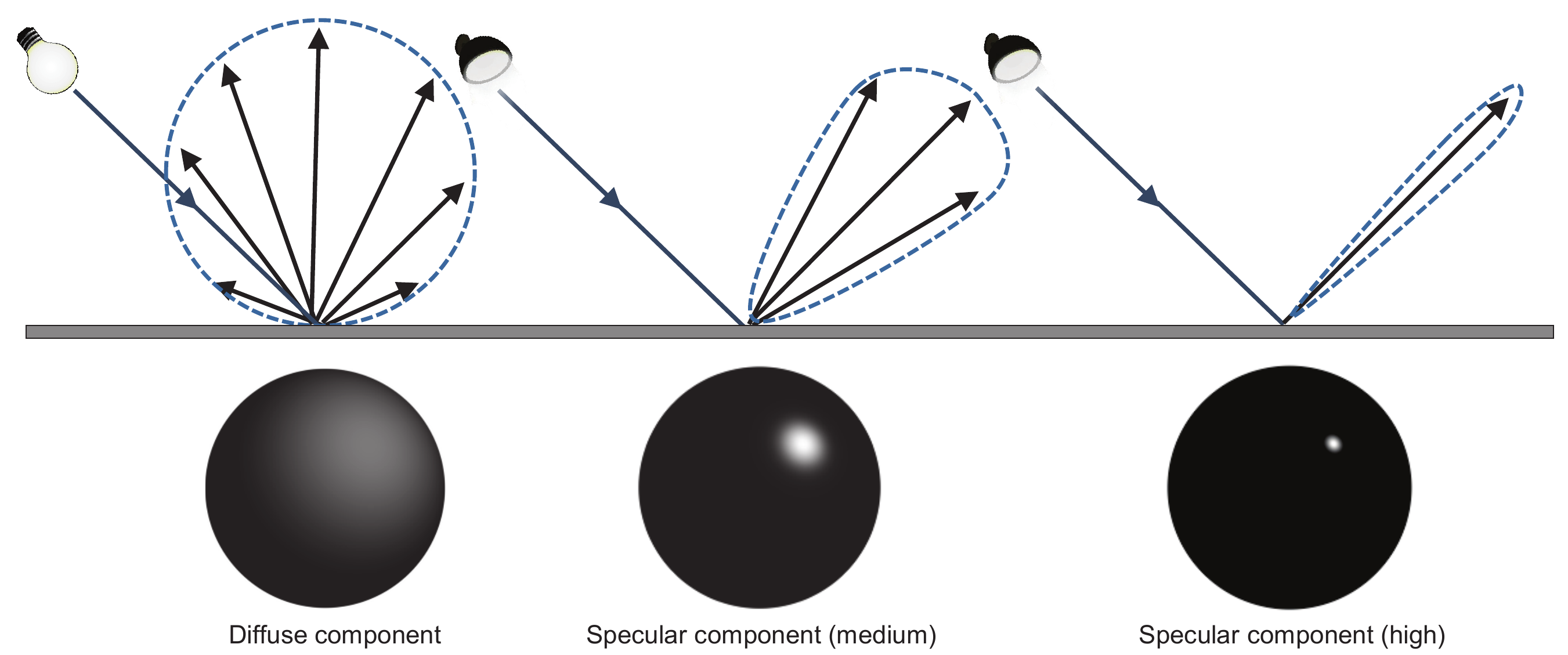}
\end{center}
\vspace{-4mm}
\caption[Lambertian model \cite{basri2003lambertian} vs. the dichromatic reflection model \cite{shafer1985using}.]{Lambertian model \cite{basri2003lambertian} vs. the dichromatic reflection model \cite{shafer1985using}. Lambertian model considers only the diffuse component of the object, while the dichromatic reflection model considers both the specular and diffuse components.}
\label{fig:lamb_dichromatic}
\end{figure}

Physics-based methods usually depend on a more complex model than the Lambertian model (described in Eq. \ref{eq1}) to estimate the illumination vector based on the physical interaction between the illumination source and the scene's surfaces. According to the dichromatic reflection model \cite{shafer1985using}, also known as neutral interface reflection assumption, the value of each color channel $c = \{\text{R}, \text{G}, \text{B}\}$ for a pixel located at $x$ in $\mat{I}$ is given by two components, as described in the following equation:

\begin{equation}
\label{eq:dichromatic}
\mat{I}_c(x) =\int_{\gamma} m_b(x)\rho(x,\lambda)R(x,\lambda)S_{c}(\lambda) d\lambda +    m_s(x)\rho(x,\lambda)R^{'}(x,\lambda)S_{c}(\lambda) d\lambda,
\end{equation}
\noindent
where $m_b(\cdot)$ and $m_s(\cdot)$ are independent scale factors depending on: (i) the angle of the viewpoint, (ii) direction of the light source, (iii) and surface orientation, and $R^{'}(\cdot)$ is the surface specular reflectance. Figure \ref{fig:lamb_dichromatic} illustrates the diffuse reflectance component, $R(\cdot)$, and specular reflectance component, $R^{'}(\cdot)$.

Based on this model, the color of specular highlights is the best cue for the illuminant color in the scene. The simplest approach is to estimate the pixel values at $m_b(\cdot) = 0$ (i.e., specular regions should be brighter than non-specular regions in the image). This gives us the max-RGB algorithm \cite{maxRGB}. This simple assumption, however, is not accurate enough to estimate the specular pixels. Tan \textit{et al.} \cite{tan2008color} found that the correlation between the pixel intensities and chromaticity is not linear for the specular component. Thus, the specular pixels can be represented by a curve in the chromaticity-intensity space. On the other hand, the diffuse pixels construct a vertical line (i.e., the diffuse chromaticity and the total image intensity values are independent).

Li and Lee \cite{li2017auto} proposed to search through a predefined set of light sources in order to undo the effect of the specular component. This cancellation condition can be satisfied if the specular component is projected on a perpendicular plane to the scene light source. To that end, they optimized the following minimization function:

\begin{equation}
\label{eq:physical}
\underset{\iota, \varrho}{\argmin} \left(\texttt{Area}\left(\iota, \varrho\right)\omega\left(\iota, \varrho\right)\right),
\end{equation}

\noindent where $\iota= G/R$ and $\varrho = G/B$, $\texttt{Area}(\cdot)$ refers to the projected area on the plane spanned by $\mat{\upsilon}_1\left(\iota,\varrho\right)$ and $\mat{\upsilon}_2\left(\iota,\varrho\right)$ vectors, and $\omega(\cdot)$ is a learned bias function that maximizes the corrected image with a low perceptual error from the ground truth image. The perceptual error was calculated using $\bigtriangleup$E, where $\bigtriangleup$E is defined as the Euclidean norm between the CIELAB color value of each pixel in the corrected image and the CIELAB color value of the corresponding pixel in the ground truth image.

Although these methods are assumed, in theory, to be more accurate, they are difficult to computed in practice \cite{hordley2006scene}.

\subsubsection{Learning-Based Methods} Learning-based methods rely on training data with examples where the ``true'' scene illumination is known (e.g., by placing a neutral object in the scene) and use various strategies to estimate or predict the illumination for new unseen images. In the following part of this section, we review examples from learning-based models, such as gamut mapping, probabilistic models, and deep learning models.

\textit{Gamut mapping} depends on finding a prior knowledge to guide a fixed model for estimating a set of plausible scene illuminants. These illuminants are determined based on their ability of mapping the entire color distribution of the testing image inside the ``canonical'' gamut (i.e., the prior). This idea was first introduced by Forsyth \cite{forsyth1990novel}, where it was found that the gamut of 180 color chips under white lighting conditions can be represented as a convex hull in the color space representing the canonical gamut under the ideal light. The final estimated illuminant vector is generated from this set of plausible scene illuminants. Finlayson and Hordley \cite{finlayson2000improving} showed that the median of these candidate illuminants works better than averaging them.

An efficient modification of the gamut mapping idea can be implemented by defining a set of all possible illuminants and the associated gamut for each one to constrain the solution space. The main idea lies in the fact that the solution space is bounded in terms of illuminant chromaticity, and can be represented by a finite set of chromaticity gamuts. Given the input image's gamut, a brute-force search through the predefined chromaticity gamuts can be used to solve the problem. This is how Finlayson \textit{et al.} \cite{Gamut} proposed a gamut-based constrained solution by minimizing the error between the chromaticity of the given image and the predefined chromaticity gamuts in order to estimate the illuminant in the scene.

Similar to the physical-based approaches, Bianco and Schettini \cite{bianco2012color} relied on physical characteristics of skin colors to guide their method to estimate the illuminant vector of a given raw-RGB image. They found that skin colors can be clustered in the color space. Specifically, the skin ``canonical'' gamut can be computed from measured samples of skin tones in $\text{Y}^{'}\text{CbCr}$ color space with a roughly constant value of luma. In their experiments, Bianco and Schettini used 697 samples with $\mu({\text{Y}^{'}}) = 0.5$. This skin canonical gamut is represented as a convex hull of these skin samples. Based on this new canonical gamut, the feasible solution is the illuminant vector that can completely map the gamut of the given image's skin pixels inside the skin canonical gamut.

Instead of relying on non-parametric models, a more efficacious way can treat the problem as a \textit{probability-based} problem to find the most likely illuminant parameters given the observed data (i.e., the given raw-RGB image's colors). This can be implemented by computing the posterior probability for each illuminant vector $\hat{\mat{\light}}_{i}$ using Bayesian estimation, such that:

\begin{equation}\label{prob}
\texttt{p}(\hat{\mat{\light}}_{i}| \mat{I}) \propto \left|\texttt{diag}(\hat{\mat{\light}}_{i})^{-1}\right|^N \texttt{p}(\texttt{diag}(\hat{\mat{\light}}_{i})^{-1} \text{ } \mat{I}) \texttt{p}(\hat{\mat{\light}}_{i}),
\end{equation}

\noindent where $\texttt{p}(\cdot)$ denotes the probability, $\texttt{diag}(\hat{\mat{\light}}_{i})^{-1}$ is the correction matrix that reconstructs the image without the effect of $\hat{\mat{\light}}_{i}$\footnote{To simplify, we represent the correction matrix using the inverse of the diagonal matrix, but more details are given in Sec. \ref{subsec:crhomatic_adaptation}}, and $\mat{I}$ is represented as $3\!\times\!N$ matrix, where $N$ is the total number of pixels in the given raw-RGB image. In Eq. \ref{prob}, the determinant term $\left|\texttt{diag}(\hat{\mat{\light}}_{i})^{-1}\right|^N = \prod_c 1/\mat{\light}_i(c)^N$ is used for normalization \cite{rosenberg2004bayesian}.

By assuming that the probability distribution of the illuminant $\texttt{p}(\hat{\mat{\light}}_{i})$ is a uniform distribution (i.e., constant), the posterior probability can be computed using maximum likelihood estimation (e.g., \cite{finlayson1999colour, rosenberg2001color}).

To define the prior of the object's reflectances (i.e., $\mat{R} = \texttt{diag}(\mat{\light})^{-1} \text{ } \mat{I}$), and $\hat{\mat{\light}}_{i}$ in the case of the Bayesian-based models, different approaches were adapted. Brainard et al. \cite{brainard1997bayesian}, for instance, assumed that the prior of the object reflectances can be represented by a Gaussian distribution. Another solution proposes to use the estimated illuminants of other algorithms (e.g., statistical-based methods) as a proxy for the ground truth illuminants to get the color distribution under a ``white'' illumination \cite{rosenberg2001color, rosenberg2004bayesian}. For the sake of accuracy, Gehler \textit{et al.} \cite{gehler2008bayesian} obtained more precise priors by collecting a set of raw-RGB images with an achromatic surface as a reference object.

Since then, many raw-RGB images have been publicly available, and consequently, more accurate learning-based algorithms were presented. For instance, Cheng \textit{et al.} \cite{Effective} proposed a fast framework that comprises a bank of regression trees to rectify the initial estimation of four different statistical-based methods (GW, max-RGB, histogram-based color palette, and histogram-based dominant color) followed by computing the median of these corrected illuminant vectors. Gijsenij \textit{et al.} \cite{gijsenij2011color} suggested incorporating semantic information, represented by the edge distributions, in order to select the proper statistical-based method. During the training stage, they clustered the training data into groups and evaluate different statistical-based methods on each cluster. When testing, the given image is assigned to the closest cluster and the best illuminant estimation method is used.

Recently, several researchers have introduced convolutional neural network (CNN)-based solutions to solve the problem, due to the impressive results obtained using the \textit{deep neural neural networks} (DNNs) in many computer vision problems \cite{deep}. Even before the evolution of DNNs in the recent years, there were a few attempts of training shallow networks in order to estimate the scene illuminant (e.g., one hidden layer was trained by Funt \textit{et al.} \cite{funt1996learning} and two hidden layers were used by Cardei \textit{et al.} \cite{cardei2002estimating}).

As the CC problem usually is posed as a regression problem, the majority of the CNNs models  were trained to predict the parameters of the global illuminant vector (e.g., \cite{BMVC1, CCC, hu2017fc}). There are a few methods (e.g., \cite{Seoung}) that approximates the solution as a classification problem to benefit from the well-established CNN-based frameworks for image classification tasks.

The straightforward adaptation of the existing CNN architectures can be performed by fine-tuning one of the pre-trained models (e.g., the pre-trained AlexNet \cite{krizhevsky2012imagenet} using ImageNet dataset \cite{deng2009imagenet}) after replacing the last fully connected (fc) layer with a new fc layer (e.g., a new fc with three neurons, each of which represents a color channel value of the estimated illuminant color). Lou \textit{et al.} \cite{BMVC1} suggested to tackle the problem using the deep learning power by feeding the network a set of images with the corresponding ground truth illuminant vectors. They used $\texttt{L}2$ loss function to fine-tune their AlexNet-based network outperforming several of the previous statistical-based methods.

Seoung and Kim \cite{Seoung} approximated the solution space by a set of clustered illuminant vectors to derive benefit from the CNN models for the image classification problem. They adopted the AlexNet architecture by fine-tuning the network's weights to classify the given raw-RGB image based on these clusters. The final estimated illuminant vector is represented by a weighted summation of the cluster centers using the score of the softmax layer.

Finlayson and Hordley \cite{finlayson2001color} showed that the illuminant vector $\mat{\light} \in \mathbb{R}^3$ can be represented by two component in the $uv$ log-chrominance space in which the problem of estimating $\mat{\light}$ can be reformulated by two unknown values instead of three. Thus, Barron \cite{CCC} represented the raw-RGB image by a 2D histogram of its log-chrominance components to treat the problem as a localization problem (correcting the image's colors now can be representing by translating the 2D histogram in the log-chrominance space). Based on this idea, Barron and Tasi \cite{FFCC} later proposed to detect the illuminant ``location'' in the histogram space through a learnable convolutional filter in the frequency domain (read Chapter \ref{ch:ch6} for more details).

Unlike Barron's methods \cite{CCC, FFCC} which depend only on the histogram feature, Shi \textit{et al.} \cite{DSNET} relied on $\mat{I}_{\text{u}}$ and $\mat{I}_{\text{v}}$ in order to train their CNN for a regression/classification task. They designed a novel architecture, called deep specialized network (DS-Net), consisting of two interacting networks---one network for regression and the second one for classification. The first network, they called it hypotheses network (HypNet), was designed to estimate two hypotheses of the illuminant vector in the $\text{UV}$ space with two output branches: branch (A) and branch (B). Each branch estimates an illuminant vector; we denote them as  $\hat{\mat{\light}}_{\text{A}}$ and $\hat{\mat{\light}}_{\text{B}}$. The selection of the best candidate is performed by the second network called selection network (SelNet). Such network is responsible for selecting the best candidate from the suggested illuminant vectors produced by HypNet (i.e., classifying the estimated illuminant to pick the best). This network is trained separately after training HypNet, where it receives $\mat{I}_\text{u}$, $\mat{I}_\text{v}$, $\hat{\mat{\light}_{\text{A}}}$, and $\hat{\mat{\light}_{\text{B}}}$. Due to the huge number of the networks' parameters, DS-Net was designed to accept only $44\!\times\!44$ patches. The final response is generated by applying median pooling on the local illuminant vectors estimated for the image's patches.

Similar to DS-Net \cite{DSNET}, Bianco \textit{et al.} \cite{bianco2017single} designed a three-stage framework. At the first stage, the given raw-RGB image is divided into a grid of $32\!\times\!32$ patches to predict the local illuminant vector using a CNN model. The second stage was designed to classify the captured scene as a multi-illuminant scene or a single-illuminant scene. At that point, the angular error between each pair of the local illuminant vectors is computed followed by a thresholding process to determine whether the scene has a single illuminant or more. In the case of a global illuminant, the third stage contains a support vector machine for regression (SVR) model, with RBF kernel, which was trained based on the angular error between the response and the ground truth illuminant vectors.

A major challenge of such patch-based methods is determining the usefulness of these patches---some local patches reflect useful information about the scene illuminant, while others do not. Hu \textit{et al.} \cite{hu2017fc} suggested to estimate a feature map, they referred to it as ``the confidence map'', that can be learned from the semantic context of the local patches to know which patch is more reliable than others. To be able to learn such confidence maps, they proposed a fully CNN that estimates a four-channel output, such that the first three channels represent the downscaled pixel-wise local estimated illuminants, while the last channel represents the confidence weights of this patch. This confidence map is used to produce the weighted estimated local illuminants. They used the angular error as a loss function that penalizes the network based on the angle between the aggregated weighted local illuminants and the ground truth illuminant. We can interpret their method as a local weighted GW with learnable weights (i.e., the confidence map) and learnable local illuminants. They found that objects with a bounded range of innate colors (e.g., faces) have higher confidence weights compared to other objects. Interestingly, this finding matches the previous work in \cite{bianco2012color} that relies on faces as a cue to estimate the scene illuminant.

Noticeably, adopting CNN architectures that had been originally designed for other tasks is commonly used for the illuminant estimation problem (for example, AlexNet was used by \cite{BMVC1, Seoung, hu2017fc}). Such architectures were basically designed to filter out the unnecessary information from the given image producing a strong feature vector (i.e., deep features). Typically, this deep feature extraction process is carried out through a stacked set of traditional conv filters followed by fc layers with learnable weights. Then, a regression/classification model is fed by these deep features in order to tackle a certain problem.  One of the important reasons for adopting this conv-based architecture is to effectively extract structural features from the spatial information of the image.

\subsection{Chromatic Adaptation Transform} \label{subsec:crhomatic_adaptation}

Chromatic adaption transform (CAT) is an essential process in color balancing to map image colors, captured under scene illumination source, to the corresponding colors under a different illumination source. Usually, CAT is employed for WB correction to eliminate undesirable color casts, so that neutral objects that perceptually appear white in reality are rendered white (i.e., R=G=B) in the final output image regardless the lighting conditions. According to the von Kries coefficient law \cite{fairchild2013color}, transforming the color response under one illuminant to another can be achieved using a simple scaling operation. In other words, a $3\!\times\!3$ diagonal matrix is sufficient to normalize the illumination's colors by mapping them to the achromatic line in the camera's raw-RGB color space \cite{cheng2015beyond, karaimer2016software}.  Here, the von Kries coefficient law is applied in the raw-RGB color space. We refer to it as the {\it standard approach} that can be represented by the following equation \cite{gijsenijcomputational}:
 \begin{equation}
 \label{sub:eq1_raw}
 \mat{I}_{\text{corr}} = \texttt{diag} \left(\mat{\light}^{*}\right)\textrm{  }\mat{I}_{\textrm{in}},
 \end{equation}
where $\mat{I}_{\textrm{in}}$ and $\mat{I}_{\text{corr}}$ are the raw-RGB input and corrected images represented as a $3\!\times\!N$ matrices, respectively, and $\texttt{diag}\left(\mat{\light}^{*}\right)$ is a $3\!\times\!3$ diagonal matrix constructed as follows:
\begin{equation}
\label{sub:eq1_diag_raw}
\texttt{diag}\left(\mat{\light}^{*}\right)=\texttt{diag}\left( \left[\frac{\hat{\mat{\light}}_{(\texttt{G})}}{\hat{\mat{\light}}_{(\texttt{R})}}, \frac{\hat{\mat{\light}}_{(\texttt{G})}}{\hat{\mat{\light}}_{(\texttt{G})}}, \frac{\hat{\mat{\light}}_{(\texttt{G})}}{\hat{\mat{\light}}_{(\texttt{B})}}\right]\right),
\end{equation}

\noindent where $\hat{\mat{\light}}$ is the estimated illuminant vector. If the color space of $\mat{I}_{\textrm{in}}$ is the CIE XYZ space, this approach is referred as XYZ scaling \cite{djordjevic2010comparison}. It is worth noting that the best solution using this method can be obtained if $\hat{\mat{\light}}$ is defined manually by picking a known neutral color in the scene (e.g., achromatic colors in a color rendition chart).

The modern models, such as von Kries transform \cite{fairchild2013color}, apply WB correction in post-adaptation cone responses related to biological vision (i.e., tristimulus responses of the L, M, S cone in the human eye \cite{stockman2008physiologically}). Note that adopting von Kries coefficient law in such new spaces is referred as ``wrong von Kries'' \cite{fairchild2013color, susstrunk2005evaluating}.

\begin{table}[!t]
\centering
\caption{Examples of chromatic adaption transform (CAT) models.}
\scalebox{0.75}{
\begin{tabular}{|c|c|}
\hline
\textbf{CAT model} & \textbf{CAT entries} \\ \hline
\rule{0pt}{8ex} XYZ scaling& \begin{tabular}[c]{@{}c@{}}$\begin{bmatrix} 1.00000 & 0.00000 & 0.00000\\ 0.00000 & 1.00000 & 0.00000\\ 0.00000 & 0.00000 & 1.00000 \end{bmatrix}$\end{tabular} \\ \hline
\rule{0pt}{8ex} von Kries \cite{fairchild2013color}&
\begin{tabular}[c]{@{}c@{}}$\begin{bmatrix}  0.40024 & 0.70760 & -0.08081\\ -0.22630 & 1.16532 & 0.04570\\ 0.0000 & 0.0000 & 0.91822 \end{bmatrix}$\end{tabular} \\ \hline \rule{0pt}{6ex}
\rule{0pt}{8ex} Bradford \cite{hunt2011metamerism} & \begin{tabular}[c]{@{}c@{}}$\begin{bmatrix}  0.89510 & 0.26640 & -0.16140\\ -0.75020 & 1.71350 & 0.03670\\ 0.03890 & -0.06850 & 1.02960 \end{bmatrix}$\end{tabular} \\ \hline
\rule{0pt}{8ex} Sharp \cite{finlayson1994spectral, finlayson2000performance} &  \begin{tabular}[c]{@{}c@{}}$\begin{bmatrix}  1.26940 & -0.09880 & -0.17060\\ -0.83640 & 1.80060 & 0.03570\\ 0.02970 & -0.03150 & 1.00180 \end{bmatrix}$\end{tabular} \\ \hline
\rule{0pt}{8ex} CMCCAT2000 \cite{li2002cmc}&  \begin{tabular}[c]{@{}c@{}}$\begin{bmatrix}  0.79820 & 0.33890 & -0.13710\\ -0.59180 & 1.55120 & 0.04060\\ 0.00080 & 0.23900 & 0.97530 \end{bmatrix}$\end{tabular} \rule{0pt}{8ex}\\  \hline
\end{tabular}
\label{tableCAT}
}
\end{table}

In particular, these models transform the colors from the CIE XYZ space into another space in which the diagonal model is assumed to work better---these models assume that the WB correction can be performed in either the CIE XYZ space or a transformed space from the CIE XYZ space, which is not the case in the existing camera imaging pipelines. According to these models, the correction process can be represented as \cite{chromatic, bianco2010two}:
 \begin{equation}
  \label{sub:eq3_raw}
  \mat{I}_{\text{corr(XYZ)}} = \left(\mat{E}^{-1}\texttt{diag}\left(\mat{\light}^{**}\right)\textrm{ }\mat{E} \right) \textrm{ } \mat{I}_{\textrm{in(XYZ)}},
  \end{equation}
\noindent
where $\mat{I}_{\textrm{in(XYZ)}}$ and $\mat{I}_{\text{corr(XYZ)}}$ are the input and corrected images in the CIE XYZ space \cite{susstrunk2005evaluating}, respectively, $\mat{E}$ is a nonsingular $3\!\times\!3$ CAT matrix, and $\texttt{diag}\left(\mat{\light}^{**}\right)$ is a diagonal matrix containing the scaling factors. The scaling vector $\mat{\light}^{**}$ is given by the following equations:

\begin{ceqn}
\begin{eqnarray}
\begin{gathered}
\\\mat{\light}^{**} = \left[\frac{\mat{\light}_{\texttt{r(R)}}}{\mat{\light}^{'}_{\texttt{e(R)}}}, \frac{\mat{\light}_{\texttt{r(G)}}}{\mat{\light}^{'}_{\texttt{e(G)}}}, \frac{\mat{\light}_{\texttt{r(B)}}}{\mat{\light}^{'}_{\texttt{e(B)}}}\right],\\
\left[\mat{\light}_{\texttt{r(R)}}, \mat{\light}_{\texttt{r(G)}}, \mat{\light}_{\texttt{r(B)}}\right] =   \left[\mat{\light}_{\texttt{r(X)}}, \mat{\light}_{\texttt{r(Y)}}, \mat{\light}_{\texttt{r(Z)}}\right] \texttt{ }\mat{E}^{T},
\\
\left[\mat{\light}^{'}_{\texttt{e(R)}}, \mat{\light}^{'}_{\texttt{e(G)}}, \mat{\light}^{'}_{\texttt{e(B)}}\right] =   \left[\hat{\mat{\light}}_{\texttt{(X/Y)}}, 1, \hat{\mat{\light}}_{\texttt{(Z/Y)}}\right] \texttt{ }\mat{E}^{T}, 
\end{gathered}
\end{eqnarray}
\end{ceqn}
where $\mat{\light}_{\texttt{r(XYZ)}}$ is a $1\!\times\!3$ row vector of a standard illuminant in CIE XYZ space (e.g., CIE standard illuminant D65).

Table \ref{tableCAT} shows examples of CAT matrices proposed in the literature. Von Kries transform \cite{fairchild2013color}, for example, is based on assuming the independent gain control of the LMS cone responses, and as a consequence, the XYZ scaling is based on the ratio of the LMS cone responses of the illumination sources, where $\mat{E}_{\text{vonKries}}$ is defined to convert the CIE XYZ values to the corresponding values in the LMS. This conversion is described in the following equation:

\begin{equation} \label{eq:von}
\begin{bmatrix}
\texttt{L}
\\
\texttt{M}
\\
\texttt{S}
\end{bmatrix} =
\begin{bmatrix}
 0.40024 & 0.70760 & -0.08081\\ -0.22630 & 1.16532 & 0.04570\\ 0.0000 & 0.0000 & 0.91822
\end{bmatrix}
\begin{bmatrix}
\texttt{X}
\\
\texttt{Y}
\\
\texttt{Z}
\end{bmatrix}.
\end{equation}

Note that the entries of $\mat{E}_{\text{vonKries}}$ in Eq. \ref{eq:von} are normalized to CIE standard illuminant D65.

\begin{figure}[!t]
\includegraphics[width=\linewidth]{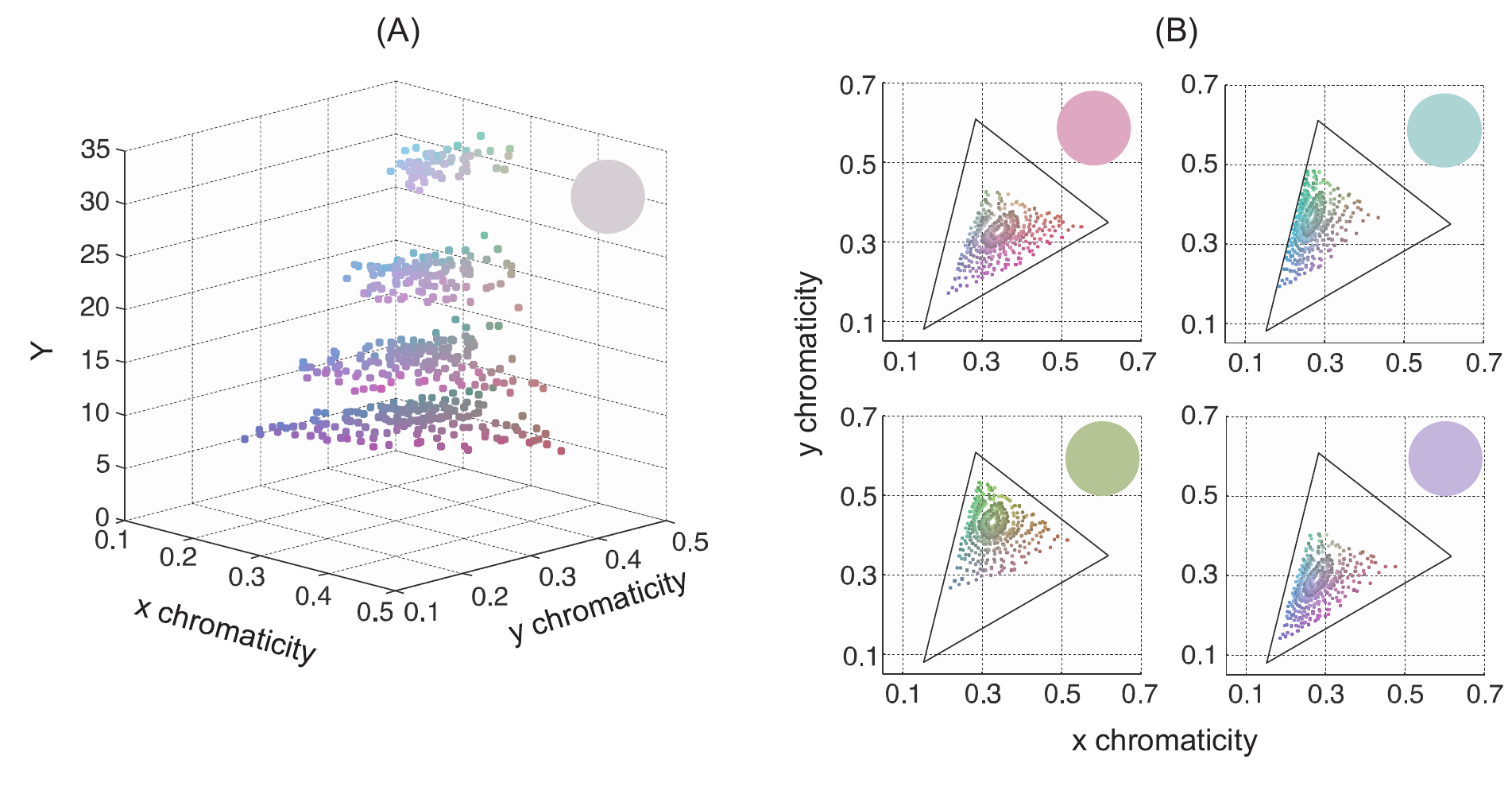}
\vspace{-8mm}
\caption[Example of Munsell chip collection.]{Example of Munsell chip collection,  adapted from~\cite{olkkonen2009categorical}.  (A) Munsell chips  under a neutral illuminant in the  CIE xyY space.
Each symbol's color represents the reflected color  of each chip. (B) shows the  projected chip  chromaticities under four illuminations. }
\label{Munsell}
\end{figure}

Usually, advanced CAT matrices (e.g., CMCCAT2000 \cite{li2002cmc}) were derived based on minimizing the error of illumination mapping (e.g., from CIE C to CIE D65) using the corresponding-color datasets \cite{susstrunk2005evaluating}. Such corresponding-color datasets include pairs of  CIE XYZ color tristimulus values under two different illumination sources based on physical stimulus \cite{luo1999corresponding, pridmore2005theory, oleari2014corresponding}. Figure \ref{Munsell} shows an example of a corresponding-color dataset.

The Bradford transform  \cite{hunt2011metamerism} is a widely used CAT matrix. The Bradford transform was derived empirically using a set of corresponding-colors of 58 dyed wool samples with different CC under CIE A and CIE D65.

\begin{figure}[!t]
\begin{center}
\includegraphics[width=0.8\linewidth]{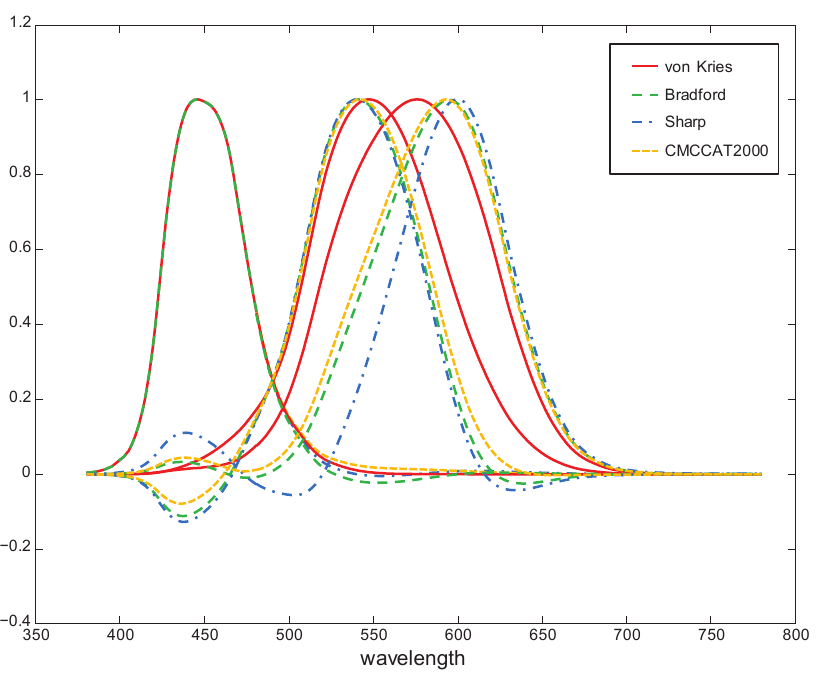}
\end{center}
\vspace{-5mm}
\caption[Normalized von Kries \cite{fairchild2013color}, Bradford  \cite{hunt2011metamerism}, Sharp  \cite{finlayson2000performance} and CMCCAT2000 \cite{li2002cmc} sensors.]{Normalized von Kries \cite{fairchild2013color}, Bradford  \cite{hunt2011metamerism}, Sharp  \cite{finlayson2000performance} and CMCCAT2000 \cite{li2002cmc} sensors. This figure is adapted from~\cite{finlayson2000performance}. }
\label{fig_Cat}
\end{figure}

Another CAT matrix is the Sharp adaption transform \cite{finlayson1994spectral}. The idea of the Sharp transform is based on the narrow cone space of the Bradford sensors---namely, the linear combination of the XYZs after applying the Bradford transform. Bradford sensors are more de-correlated and represented by a narrowed cone space compared to the relative LMS cone
responses, as shown in Fig. \ref{fig_Cat}. Accordingly, Finlayson \textit{et al.} \cite{finlayson1994spectral, finlayson2000performance} found that there is a potential improvement in the performance of CAT models, if the data is pre-processed by applying Sharp transform $\mat{E}_{sharp}$, such that:
\begin{equation}
  \label{sub:sharpining}
  \mat{C}_{\mat{\light}_1} = \mat{C}_{\mat{\light}_2} \mat{E}_{\texttt{sharp}} \texttt{diag}\left(\mat{\light}^{**}\right),
  \end{equation}
where $\mat{C}_{\mat{\light}_1}$ and $\mat{C}_{\mat{\light}_2}$ are $n\!\times\!3$ matrices of the CIE XYZ values under the first illumination $\mat{\light}_1$ and second illumination $\mat{\light}_2$, respectively. The sharpening transform can be obtained through eigenvector decomposition of the matrix $\mat{S}$ given in the following equation:
\begin{equation}
\mat{S} = (\mat{C}_{\mat{\light}_2}^{T} \mat{C}_{\mat{\light}_2})^{-1}\mat{C}_{\mat{\light}_2}^{T}\mat{C}_{\mat{\light}_1},
\end{equation}
\begin{equation}
\mat{S} = \mat{E}_{\texttt{sharp}}\texttt{ diag}\left(\mat{\light}^{**}\right)\text{ }\mat{E}_{\texttt{sharp}}^{-1}.
  \end{equation}

Another example of CAT matrices is CMCCAT2000, which is derived using an iterative optimization process to minimize the error between predicted and observed colors over a set of eight color datasets in the CIELAB color space \cite{li2002cmc}.

Several studies in the literature were carried out to evaluate different CATs \cite{susstrunk2000chromatic, susstrunk2005evaluating, djordjevic2010performance}. These studies aimed to find the best CAT that obtains the smallest color differences between transformed source and target color values.

To the best of our knowledge, there is no clear criterion for what the best CAT is. For example, Sharp adaptation transform outperforms Bradford transform using particular datasets, while Bradford performs better in another set of color pairs \cite{finlayson2000performance}. S$\ddot{\text{u}}$sstrunk and Finlayson \cite{susstrunk2005evaluating} found that CMCCAT2000 outperforms the Sharp transform. In another experiment, it is found that Bradford transform is exhibited the lowest error compared to von Kries, XYZ scaling,
and CMCCAT2000 using 8,190 color patches containing different pairs of illuminants (D50-A, D50-D65, and D65-A) \cite{djordjevic2010performance}.

Note that these CAT models are applied to the image's CIE XYZ values. Computing the CIE XYZ values from the raw-RGB image, however, is performed through a $3\!\times\!3$ full CST which is computed based on an accurate predetermination of the correct scene illuminant value and color temperature; meaning that in order to get correct CIE XYZ values, the scene illuminant should be determined first---a chicken-and-egg problem. Thus, existing camera imagining pipelines apply the WB correction to the raw-RGB image (i.e., the standard approach) before converting it to the CIE XYZ space \cite{brownimproving}.

In the case of multi-illuminant scenes, Yang and Shevell's study \cite{yang2003surface} shows that in the case of two different illuminants, CC is improved if only the specular highlight cues of both illuminants are consistent. They found, however, that CC is reduced if the scene's objects have cues from two distinct illuminants. Accordingly, the chromatic adaption methods mostly assume a single light source (i.e., correct only the dominant light source).

In the literature, considering multiple light sources is performed in two different ways. The first approach is to correct the raw-RGB image for each estimated illuminant separately, then applying a blending post-processing process to produce the final ``corrected'' image. This approach was adopted by Cheng \textit{et al.} \cite{cheng2016two} who assessed the user preference and found that ``warmer'' (reddish) results are more preferable for the outdoor-illuminant scenes captured under two distinct illuminants (i.e., warm and cold tones). They achieved that by blending two images, each of which is corrected using the standard approach, with a blending factor for the ``cold'' image tone $\in\{0.25,0.5\}$. 

The second approach is applying a pixel-wise diagonal-based correction for multi-illuminant scenes. This pixel-wise estimation is mostly approximated by patch-wise estimation, followed by post-processing procedures to get approximated pixel-wise illuminant vectors \cite{beigpour2014multi, hussain2018color}. For instance, Hsu \textit{et al.} \cite{hsu2008light} proposed a method to correct images with two mixed illuminants ($\mat{\light}_{(1)}$ and $\mat{\light}_{(2)}$) by assuming that the illuminant vectors are specified by the user and the captured scene has a small number of material colors (i.e., a sparse set of colors). According to their case, the image formation can be expressed by the following equation:

\begin{equation}
\label{eq1_mixed_two}
\mat{I}_c(x) = \mat{R}_{c}\times (\beta_1\mat{\light}_{(1)c}  + \beta_2\mat{\light}_{(2)c}),
\end{equation}
\noindent
where $\beta_1$ and $\beta_2$ are unknown scalar factors representing the influence on the pixel located at $x$ of light sources $\mat{\light}_{(1)}$ and $\mat{\light}_{(2)}$, respectively.  Thus, the $3\!\times\!3$ diagonal correction matrix's elements $\mat{\light}^{*}$ can be expressed as follows:

\begin{equation}
\label{eq2_mixed_two}
\mat{\light}^{*}_{c} = \frac{\beta_1 + \beta_2}{\beta_1\mat{\light}_{(1)c}  + \beta_2\mat{\light}_{(2)c}} = \frac{1}{\beta*\mat{\light}_{(1)c}  + (1-\beta*)\mat{\light}_{(2)c}},\hspace{3pt} \text{with} \hspace{10pt} \beta* = \frac{\beta_1}{\beta_1+\beta_2}.
\end{equation}

To estimate the value of $\beta*$, they first estimate potential material colors in the captured scene by adopting a greedy voting approach. This voting approach works in a $32\times 32$ projected chromaticity space (i.e., $\text{R}/\text{B}$, $\text{B}/\text{R}$) with log spacing by assigning each pixel to the nearest bin. That is, given an estimated material color $\hat{\mat{R}}$ and the associated pixels to this material in $\mat{I}$, Eq. \ref{eq1_mixed_two} can be solved. Despite the impressive results obtained by this method, it is constrained by specific conditions.

\subsection{Research Directions}

Following up on the main research areas discussed in this section, we can summarize promising research directions as: (i) lightweight learning-based CC and (ii) sensor-independent learning-based CC. 

\subsubsection{Lightweight Learning-Based CC}
As discussed earlier, statistical-based illuminant estimation methods operate using statistics from an image's color distribution and spatial layout to estimate the scene illuminant. These methods are fast and easy to implement; however, their results are not always satisfactory. On the other hand, learning-based methods rely on training data with examples where the illumination is known (e.g., by placing a neutral object in the scene) and use various strategies to estimate or predict the illumination. In recent years, learning-based methods employing deep-learning techniques have shown state-of-the-art performance. Learning-based methods, however, suffer from a substantial increase in complexity, with deep network architectures requiring millions of parameters (see Table \ref{tableCNNparam}).

\begin{table}[!t]
\caption{Comparison between number of parameters required by examples of statistical-based and CNN-based methods.}
\label{tableCNNparam}
\centering
\scalebox{0.6}{
\begin{tabular}{|c|c|c|c|c|c|c|c|c|c|}
\hline
\multirow{2}{*}{\textbf{Method}} & \multicolumn{4}{c|}{\textbf{Statistical-based methods}} & \multicolumn{5}{c|}{\textbf{CNN-based methods}} \\ \cline{2-10}
 & GW \cite{GW} & SoG \cite{SoG}& GE \cite{GE}& PCA \cite{cheng2014illuminant} & DS-Net \cite{DSNET} & SCC \cite{semantic18}& AlexNet-FC4 \cite{hu2017fc} & DOCC \cite{hold2017deep}& Quasi U CC \cite{bianco2019quasi} \\ \hline
 \textbf{\begin{tabular}[c]{@{}c@{}}Number of\\  parameters\end{tabular}} & 0 & 2 & 2 & 1 & $\sim$17 millions & $\sim$14 millions & $\sim$4 millions & $\sim$4 millions & $\sim$80 millions \\ \hline
 \end{tabular}
 }
 \end{table}
 
In the absence of specialized chips or GPUs, the computational and memory requirements associated with running these methods onboard the camera are still prohibitive. As a result, cameras currently rely on simple statistical methods even though these methods are not as accurate as their learning-based counterparts. A promising research direction could include improving the accuracy of such simple statistical-based methods by learning a post-process enhancement mechanism.

\subsubsection{Sensor-Independent Learning-Based CC}
\begin{figure}[t]
\includegraphics[width=\textwidth]{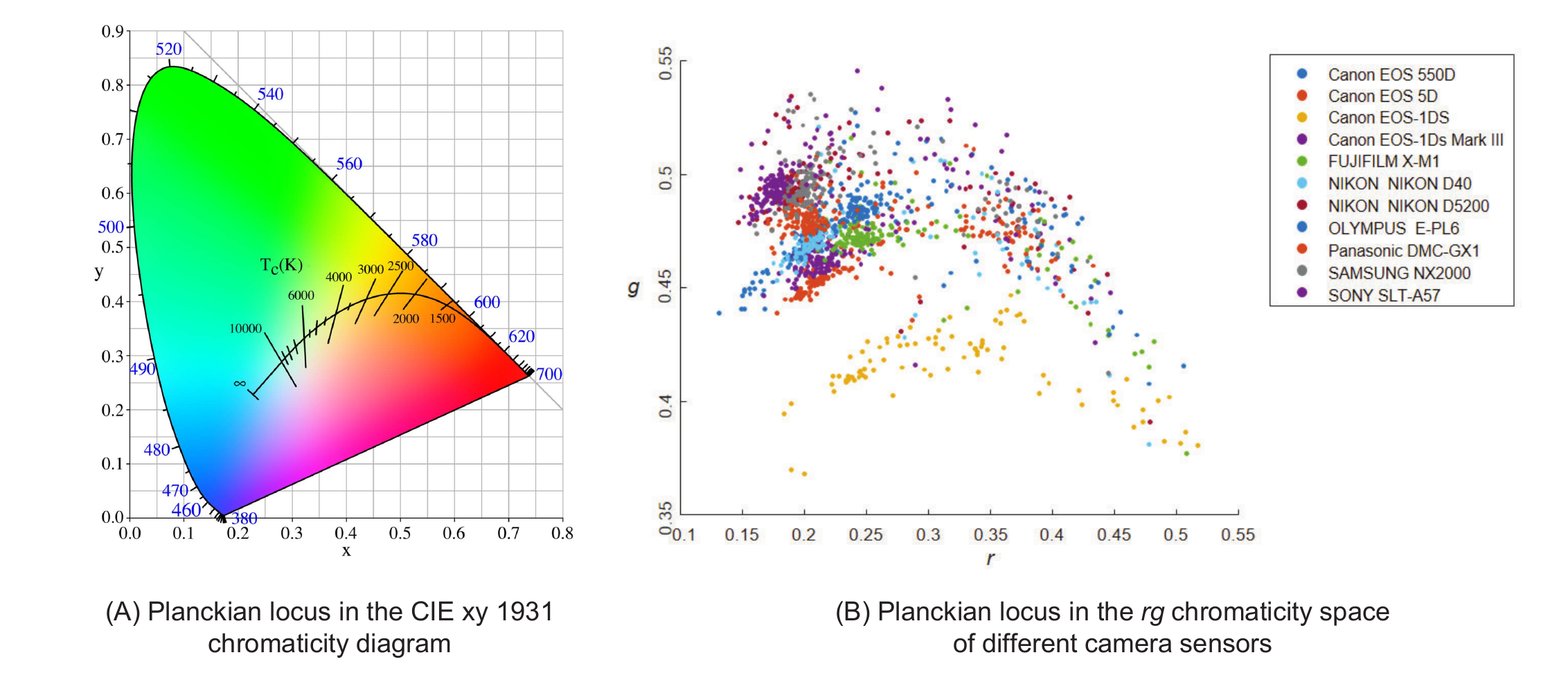}
\vspace{-7mm}
\caption[This figure shows the ideal path that a black body color takes in a chromaticity space as the black body temperature changes (i.e., Planckian locus).]{This figure shows the ideal path that a black body color takes in a chromaticity space as the black body temperature changes (i.e., Planckian locus). (A) Planckian locus in the device-independent CIE xy 1931 chromaticity space. (B) Planckian locus in the $rg$ chromaticity space of different camera sensors, where $r = \text{R}/(\text{R} + \text{G} + \text{B})$ and $g = \text{G}/(\text{R} + \text{G} + \text{B})$.\label{fig:sensor_spectral_senesitvity}}
\end{figure}

Learning-based illuminant estimation models outperform statistical-based methods by training sensor-specific models on training examples provided with the labeled images with ground-truth illumination. These training images are captured with the sensor make and model being trained. The obvious drawback of these methods is that they do not generalize well for arbitrary camera sensors without re-training/fine-tuning on samples captured by testing camera sensor. The reason behinds this is in Eq. \ref{eq0}, where the sensor's spectral sensitivity function has a direct contribution to the captured colors by cameras. Figure \ref{fig:sensor_spectral_senesitvity}-(A) shows the black body locus (also called Planckian locus) in an ideal device-independent space chromaticity diagram (i.e., CIE xy 1931 chromaticity) for a wide range of temperatures. This Planckian curve, however, does not be represented similarly in different camera sensor spaces (see Fig.\ \ref{fig:sensor_spectral_senesitvity}-[B]) due to the differences in the spectral sensitivity functions. A promising research direction, thus, is to develop a sensor-independent learning-based illuminant estimation method that is explicitly designed to generalize well for unseen camera sensors without the need to re-train/tune our model. One could think of mapping camera raw-RGB sensor responses to a perceptual color space. As discussed in Chapter \ref{ch:ch2}, this process is applied onboard digital cameras to map the captured sensor-specific raw-RGB image to a standard device-independent ``canonical'' space (e.g., CIE XYZ) \cite{ramanath2005color, karaimer2016software}. Usually this conversion is performed using a $3\!\times\!3$ matrix and requires an accurate estimation of the scene illuminant \cite{can2018improving}. It is important to note that this mapping to CIE XYZ requires that white-balance procedure first be applied.  As a result, it is not possible to use CIE XYZ as the canonical color space to perform illumination estimation. Work by Nguyen~et al. \cite{nguyen2014raw} studied several transformations to map responses from a source camera sensor to a target camera sensor, instead of mapping to a perceptual space. In their study, a color rendition reference chart is captured by both source and target camera sensors in order to compute the raw-to-raw mapping function. Learning a mapping transformation between responses of two different sensors is also adapted in \cite{gao2017improving}. However, the work in~\cite{nguyen2014raw, gao2017improving} has no mechanism to map an unseen sensor to a canonical working space without explicit calibration. 

Recently, few-shot and multi-domain learning techniques \cite{xiao2020multi, mcdonagh2018formulating} have been proposed to reduce the effort of re-training camera-specific learned color constancy models. These methods require only a small set of labeled images for a new camera unseen during training. Another strategy has been proposed to white balance the input image with several illuminant color candidates and learn the likelihood of properly white-balanced images \cite{hernandez2020multi}. Such a Bayesian framework requires prior knowledge of the target camera model's illuminant colors to build the illuminant candidate set. Despite promising results, these methods all require labeled training examples from the target camera model: raw images paired with ground-truth illuminant colors. As mentioned earlier, collecting such training examples is a tedious process, as certain conditions must be satisfied---i.e., for each image to have a single uniform lighting and a calibration object to be present in the scene \cite{cheng2014illuminant}.

\section{Color Correction for Camera-Rendered Images}
\label{Sec:white_balance}

\begin{figure}[!t]
\begin{center}
\includegraphics[width=\textwidth]{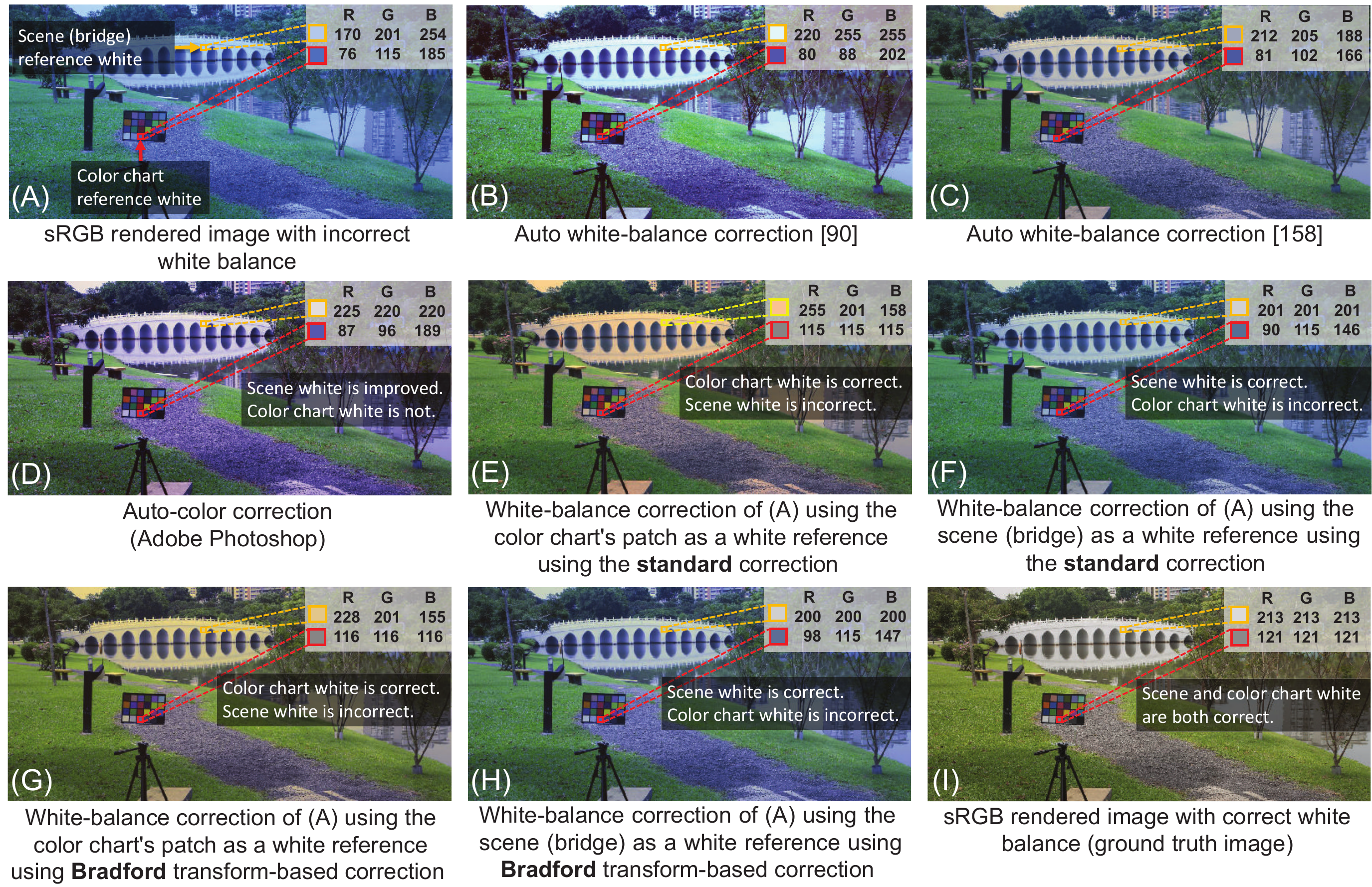}
\end{center}
\vspace{-5mm}
\caption[(A) An sRGB image rendered with an incorrect WB applied. (B) and (C) show results obtained using two different AWB algorithms \cite{tai2012automatic, huo2006robust}. (D) shows the result of auto-color correction from Adobe Photoshop. (E) and (F) show standard WB correction applied to the sRGB-rendered image using different reference white points. (G) and (H) show the Bradford transform-based correction. (I) Ground truth sRGB image with the correct WB applied.]{(A) An sRGB image rendered with an incorrect WB applied. There are two achromatic regions highlighted in red and yellow. (B) and (C) show results obtained using two different AWB algorithms \cite{tai2012automatic, huo2006robust}. (D) shows the result of auto-color correction from Adobe Photoshop. (E) and (F) show
standard WB correction applied to the sRGB-rendered image using different reference white points. (G) and (H) show the Bradford transform-based correction as described in Eq. \ref{sub:eq3} using the same reference white points. (I) Ground truth sRGB image with the correct WB applied. \label{fig:wrongWB}}
\end{figure}

Following up our discussion on image white balancing, we now assume that the given image was rendered with an incorrect WB setting in the sRGB color space (see Fig.~\ref{fig:wrongWB}-[A] for example). As can be seen, the shown image has a strong color cast due to the WB error. As shown in Fig. \ref{fig:wrongWB}-(A), there are two highlighted achromatic regions in the scene: (i) a patch from a white bridge and (ii) a neutral patch from the color rendition chart. The same scene was rendered to sRGB with the correct WB in Fig.~\ref{fig:wrongWB}-(I). As shown, because the WB was applied correctly in the proper space (i.e., raw-RGB), both the scene achromatic regions lie on the white line (i.e., R=G=B).

\begin{figure}[t]
\includegraphics[width=\textwidth]{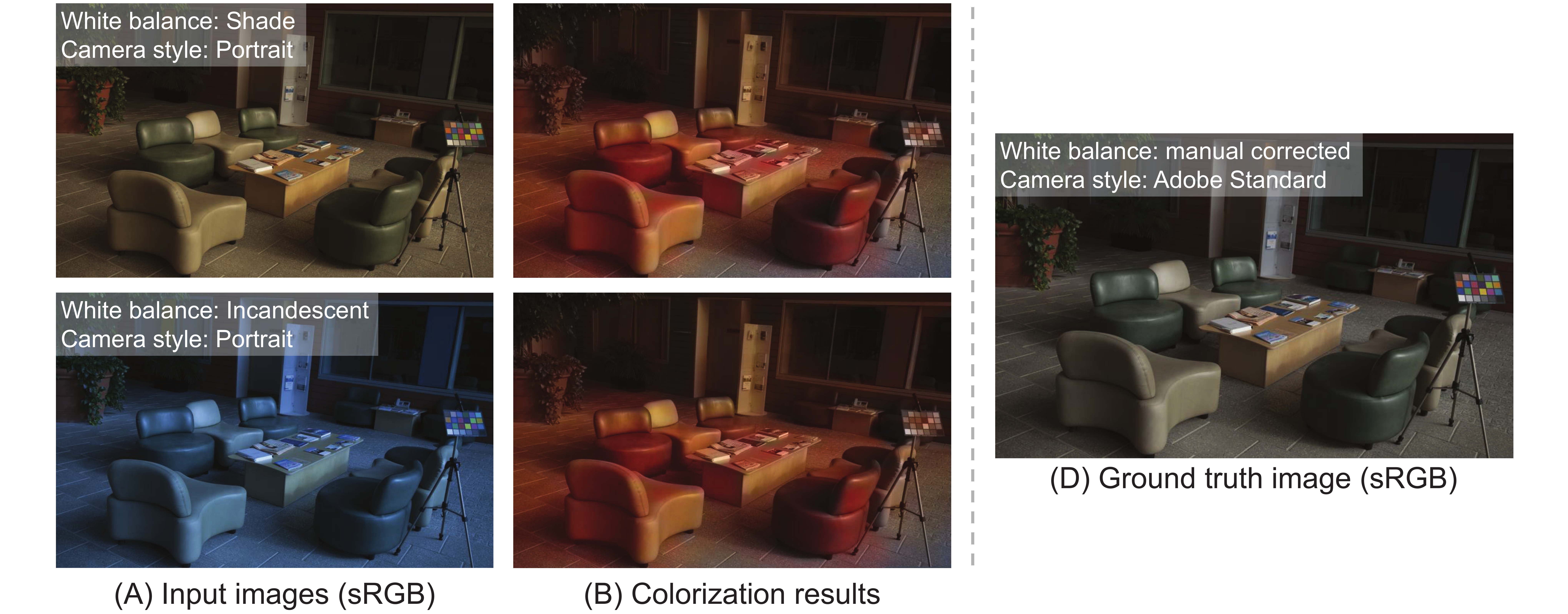}
\vspace{-7mm}
\caption[(A) Input images with a wrong WB rendered in the sRGB color space. (B) Colorized images using Zhang \textit{et al.}'s method \cite{zhang2016colorful}. (C) Ground truth image with an accurate WB correction rendered in the sRGB color space.]{(A) Input images with a wrong WB rendered in the sRGB color space. (B) Colorized images using Zhang \textit{et al.}'s CNN-based method \cite{zhang2016colorful}. (C) Ground truth image with an accurate WB correction rendered in the sRGB color space. Note that the input for the colorization method was the $L^*$ channel of the input image. \label{fig:colortransfer_colorization}}
\end{figure}

One solution to correct the colors of such images that were rendered with WB errors is by estimating the target color distribution of the given image using a trained CNN for relevant problems, such as image colorization. Although the plausible colors are produced by the recent CNN-based colorization methods (e.g., \cite{zhang2016colorful}), they usually consider only the spatial information, regardless of the input image's color distribution, in order to produce the output image. Consequently, the estimated colors are consistent and too far from the ground truth images regardless of the level of degradation of the given image's color distribution. Figure \ref{fig:colortransfer_colorization} shows the results obtained by employing a colorization method to correct camera-rendered images with WB errors. As can be seen, the results of colorization is promising in terms of colorizing a given image; however, they are still far from the ground truth images.

Another possible direction is considering simple histogram-based operations, such as applying histogram stretching for each color channel \cite{wang2011fast}. The results of the stretching operation can be blended with the results of statistical-based methods to give room for improvements \cite{tai2012automatic}. However, these simple operations are inadequate to deal with the  high degradation resulting from the nonlinear camera imaging operations applied after an incorrect WB setting, as shown in Fig. \ref{fig:wrongWB}-(B). Even the color correction functions provided by different commercial software packages cannot properly work with such cases. As an example, Adobe Photoshop has two functions for color correction---namely, auto-color and auto-tone. The auto-tone function automatically adjusts the black and white points of the given sRGB image's tone curve. The process includes clipping parts of the shadow and highlights, and mapping the darkest and lightest values of each color channel to pure black and white \cite{adopePhotoshopGuide}. The auto-color function adjusts both the contrast and colors of the given sRGB image by searching through the image's color to identify shadows, midtones, and highlights in order to map the midtones to 128 gray levels followed by clipping the shadows and highlights by 0.5\% \cite{adopePhotoshopGuide}. Fig.~\ref{fig:wrongWB}-(D) shows the results of Adobe Photoshop's auto-color. As shown, the auto-color function reduces the color cast in Fig.~\ref{fig:wrongWB}-(A), but in the perspective of WB, the result still needs more improvement.

\begin{figure}[!t]
\begin{center}
\includegraphics[width=\textwidth]{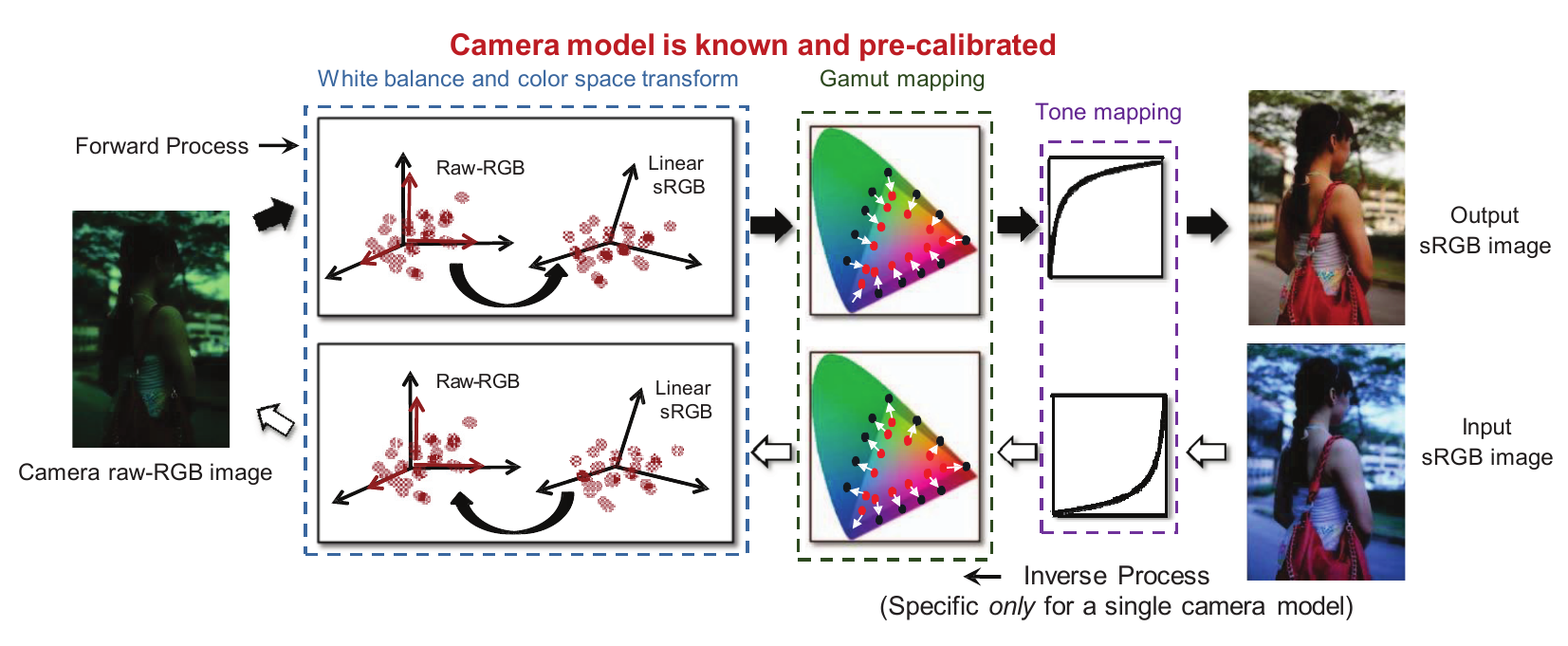}
\end{center}
\vspace{-7mm}
\caption[An accurate WB correction for sRGB-rendered images can be obtained with calibrated camera models, in which the full model of response (MoR) is defined and the original raw-RGB image reconstruction is applicable.]{An accurate WB correction for sRGB-rendered images can be obtained with calibrated camera models, in which the full model of response (MoR) is defined and the original raw-RGB image reconstruction is applicable. This figure shows the inverse process of the in-camera pipeline after calibrating the required parameters and settings of the imaging process. This calibration is done by using training images captured by the same camera used to photograph the testing image. This figure is adapted from \cite{kim2012new}. \label{fig:inverse_pipeline}}
\end{figure}

We can think of utilizing the rich amount of research done for illuminant estimation and color correction in the linear space by extending the existing methods to work in the sRGB color space. Here, we refer to the sRGB image as $\mat{I}$ instead of $\mat{I}_{\texttt{sRGB}}$ for simplicity. Mathematically, if the given image is rendered in the sRGB color space, the value of each color channel, according to the Lambertian model in Eq. \ref{eq1}, is now represented as

\begin{equation}
\label{eq222}
\mat{I}_{c} = f_{\text{CRF}}(\mat{R}_{c} \times \mat{\light}_{c}).
\end{equation}

Currently, the available accurate solutions to calculate $f_{\text{CRF}}$ and $f_{\text{CRF}}^{-1}$ require applying a serious radiometric calibration (e.g., \cite{kim2012new}) or embedding metadata onboard cameras (e.g., \cite{nguyen2018raw}) to reconstruct the original raw-RGB image. After reconstruction, the diagonal WB correction can be applied, then the corrected image is converted back to the sRGB color space. Figure \ref{fig:inverse_pipeline} shows an example of Kim \textit{et al.}'s in-camera model to tackle the problem of post color correction of improperly white-balanced rendered sRGB images. Although this solution gives accurate results, this process is impractical in many scenarios, as it requires calibrating the camera model used in photographing the image.

One possible solution is to use another color space,  instead of the raw-RGB or sRGB color spaces, to apply the WB correction. For example, if we aim to apply the correction in the CIE XYZ color space, the problem now can be reformulated as

\begin{equation}
\label{eq2}
\mat{I}_{c} = f^{-1}(\mat{I}_{(XYZ)c} \times \mat{\light}_{c}),
\end{equation}

\noindent where $f: \left[\texttt{R}, \texttt{G}, \texttt{B}\right] \rightarrow \left[\texttt{X}, \texttt{Y}, \texttt{Z}\right]$ is a nonlinear function that ``linearizes'' the sRGB color triplet to get the corresponding CIE XYZ values \cite{ebner2007color}. We would emphasize that the $f$ function is defined only if the image was rendered to the sRGB color space using the standardized sRGB that assumes the gamma value of 2.2 with no picture style applied \cite{anderson1996proposal}. Otherwise, $f$ is undefined and a radiometric calibration process is required.

Here, most of the conventional radiometric calibration solutions are not applicable as a result of the absence of any information related to the camera model or the ability to re-capture several images for the same scene as required by most of the radiometric calibration methods.  The simplest solution is to adopt the inverse of the gamma operation defined in the standardized sRGB conversion \cite{anderson1996proposal} as an approximation to the true nonlinear function applied on the image.

If we defined the $f$ function as a single gamma operation, the standard approach for correction can be updated according to

\begin{equation}
 \label{sub:eq1_srgb}
 \mat{I}_{\texttt{corr}} = f^{-1}\left( \texttt{diag}\left(\mat{\light}^{*}\right) \textrm{ } f\left(\mat{I}_{\textrm{in}}\right)\right).
\end{equation}

Note that the values of the illuminant should also be in the CIE XYZ space by obtaining the illuminant vector after converting the sRGB image to the CIE XYZ color space. Applying modern CAT models models can also be adjusted to deal with sRGB images according to

\begin{equation}
\label{sub:eq3}
\mat{I}_{\texttt{corr}} = f^{-1}\left( \left(\mat{E}^{-1}\texttt{diag}\left(\mat{\light}^{**}\right)\textrm{ }\mat{E} \right) \textrm{ } f\left(\mat{I}_{\textrm{in}}\right)   \right).
\end{equation}

Figure \ref{fig:wrongWB}-(G) and (H) show results obtained by the Matlab function for WB correction using the Bradford transform~\cite{hunt2011metamerism} with the inverse of the gamma operations defined in \cite{anderson1996proposal}. There are noticeable differences between the corrected images and the ground truth image.

Now, we examine applying the standard diagonal approach for WB correction while intentionally ignoring the nonlinear color manipulations applied onboard cameras. As an example, Huo \textit{et al.} \cite{huo2006robust} proposed to convert the sRGB image into the YUV space to correct its colors iteratively using the standard diagonal approach after estimating a set of estimated gray pixels at each iteration; see Fig. \ref{fig:wrongWB}-(C).

Instead of estimating the illuminant vector, the best solution (i.e., the ground truth) can be obtained if $\hat{\mat{\light}}$ is defined manually by picking a known neutral color in the scene. As shown in Fig. \ref{fig:wrongWB}-(A), there are two reference white points in the scene: (i) a gray patch in the color rendition chart $p(1)$ and (ii) a white patch from the bridge in the scene $p(2)$. In Fig. \ref{fig:wrongWB}-(E), we applied Eq. \ref{sub:eq1_raw} directly to correct the input sRGB image in Fig. \ref{fig:wrongWB}-(A) using the color rendition chart's patch as a reference white.
By plugging the values of $\texttt{diag}\left(\mat{\light}^{*}\right)$ into Eq. \ref{sub:eq1_raw}, we can correct the reference achromatic point $p(1)$ as follows:
\begin{equation}
\label{sub:eq2}
p(1)= \begin{bmatrix}
1.5132 & 0 & 0\\
0 & 1 & 0\\
0 & 0 & 0.6216
\end{bmatrix} \textrm{ }  \begin{bmatrix} 76\\ 115\\ 185 \end{bmatrix} = \begin{bmatrix} 115\\ 115\\ 115\end{bmatrix}.
\end{equation}

However, applying the same diagonal correction matrix to the second reference white point $p(2)$ results in incorrect WB (i.e., $p(2)=\left[255, 201, 158\right]$). Figure \ref{fig:wrongWB}-(F) shows another attempt using the bridge scene region as a reference white. As exhibited, the same problem appears in the color rendition chart's reference point.

\begin{figure}[!t]
\begin{center}
\includegraphics[width=0.85\textwidth]{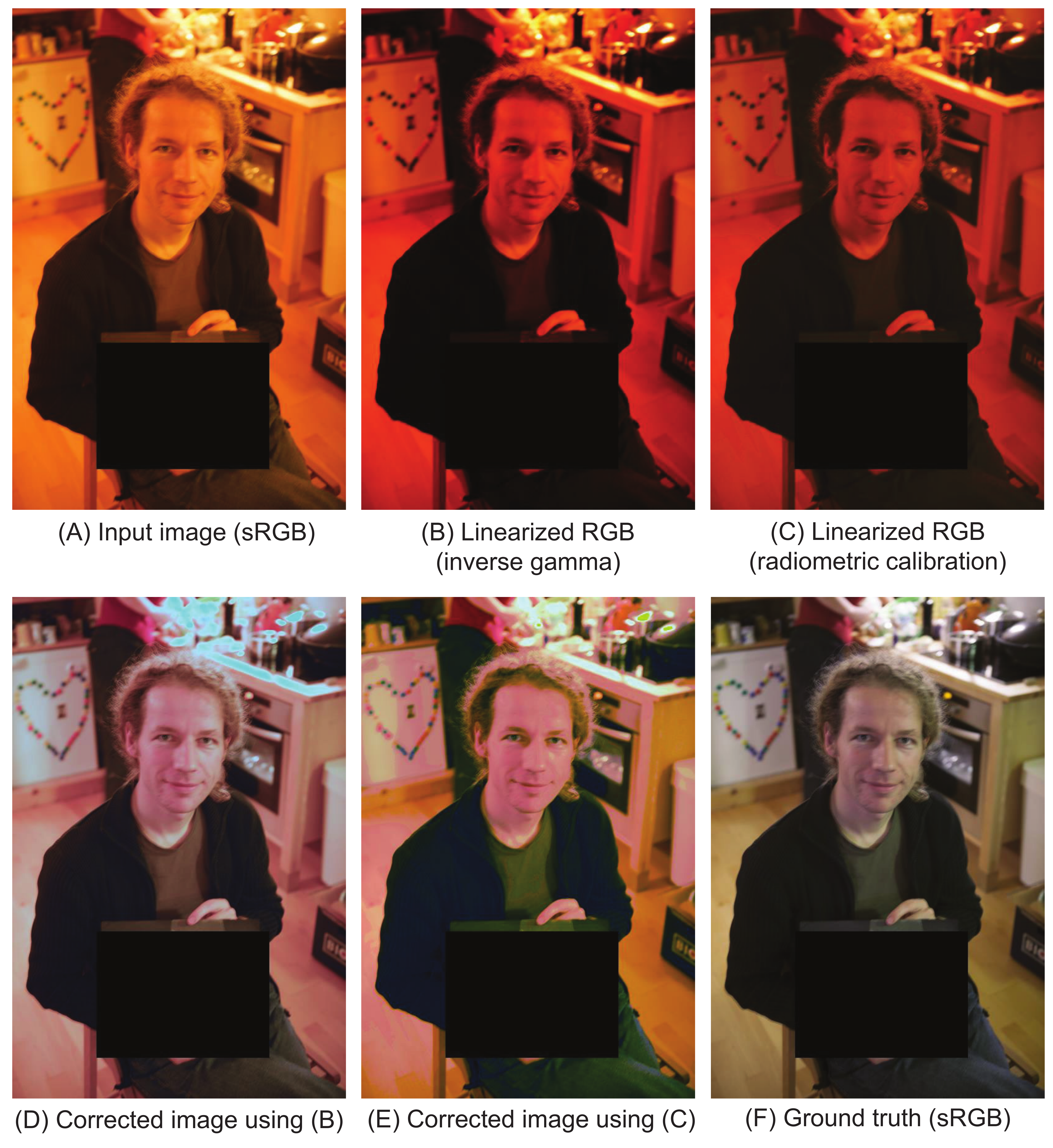}
\end{center}
\vspace{-5mm}
\caption[Example of applying the ``linearization'' process before the WB correction for sRGB images.]{Example of applying the ``linearization'' process before the WB correction for sRGB images. (A) Input sRGB image is linearized using the simple inverse gamma operation \cite{anderson1996proposal} in (B) and the recent radiometric calibration for face images \cite{li2017radiometric} in (C). The achromatic region in the provided color rendition chart (hidden after selecting the achromatic region manually) was used to correct images in (B) and (C) by applying WB correction using Bradford transform \cite{hunt2011metamerism}, as shown in (D) and (E), respectively. (F) Ground-truth image.  \label{fig:wrongWB_linearized}}
\end{figure}

Based on this discussion, a practical possible solution for WB sRGB-rendered images taken by uncalibrated or unknown camera models is to adopt the current chromatic adaption methods with a single estimated illuminant vector with or without the gamma-based linearization process. This can lead in some cases to out-of-gamut colors, because such chromatic adaption methods are meant to work in linear spaces. As a consequence, applying them to the improperly white-balanced sRGB images usually leads to undesirable results---even with attempts to linearize the sRGB image using the simple inverse of the gamma operation, as shown in Fig. \ref{fig:wrongWB_linearized}-(D), or one of the applicable radiometric calibration methods, as shown in Fig. \ref{fig:wrongWB_linearized}-(E).

\subsection{Research Directions}
Based on the previous discussion, it is obvious that there is a need for a solution to correct image colors that were rendered with WB errors. Another research direction is to allow the user to interactively manipulate the WB settings in the post-capture stage (i.e., applying accurate chromatic adaptation to different color temperatures in the sRGB color space). This post-capture WB editing may act as a beneficial tool to satisfy different user preferences (which may not always match camera AWB correction). Editing the WB settings in camera-rendered images also allow the user to manipulate the colors of captured images of challenging scenes, such as multi-illuminant scenes, where camera raw diagonal WB correction cannot produce correct colors for the entire scene.

\section{Exposure Errors in Camera-Rendered Images} \label{sec:exposure_errors}
As discussed in Chapter \ref{ch:intro}, exposure settings have a significant impact on the quality of the final rendered colors. This section of the thesis discusses potential exposure errors that may occur in the camera's ISP when capturing an image. We then present a brief review of related methods for correcting images rendered with exposure errors. 

Photographic exposure refers to the amount of received light to camera sensor. The exposure used at capture time directly affects the overall brightness of the final rendered photograph. Digital cameras control exposure using three main factors: (i) capture shutter speed, (ii) f-number, which is the ratio of the focal length to the camera aperture diameter, and (iii) the ISO value to control the amplification factor of the received pixel signals. In photography, exposure settings are represented by exposure values (EVs), where each EV refers to different combinations of camera shutter speeds and f-numbers that result in the same exposure effect---also referred to as `equivalent exposures' in photography.

Digital cameras can adjust the exposure value of captured images for the purpose of varying the brightness levels. This adjustment can be controlled manually by users or performed automatically in an auto-exposure (AE) mode. When AE is used, cameras adjust the EV to compensate for low/high levels of brightness in the captured scene using through-the-lens (TTL) metering that measures the amount of light received from the scene \cite{peterson2016understanding}.

Exposure errors can occur due to several factors, such as errors in measurements of TTL metering, hard lighting conditions (e.g., very low lighting and backlighting), dramatic changes in the brightness level of the scene, and errors made by users in the manual mode. Such exposure errors are introduced early in the capture process and are thus hard to correct after rendering the final 8-bit image. This is due to the highly nonlinear operations applied by the ISP afterwards to render the final 8-bit sRGB image \cite{karaimer2016software}.

\begin{figure}[!t]
\includegraphics[width=\textwidth]{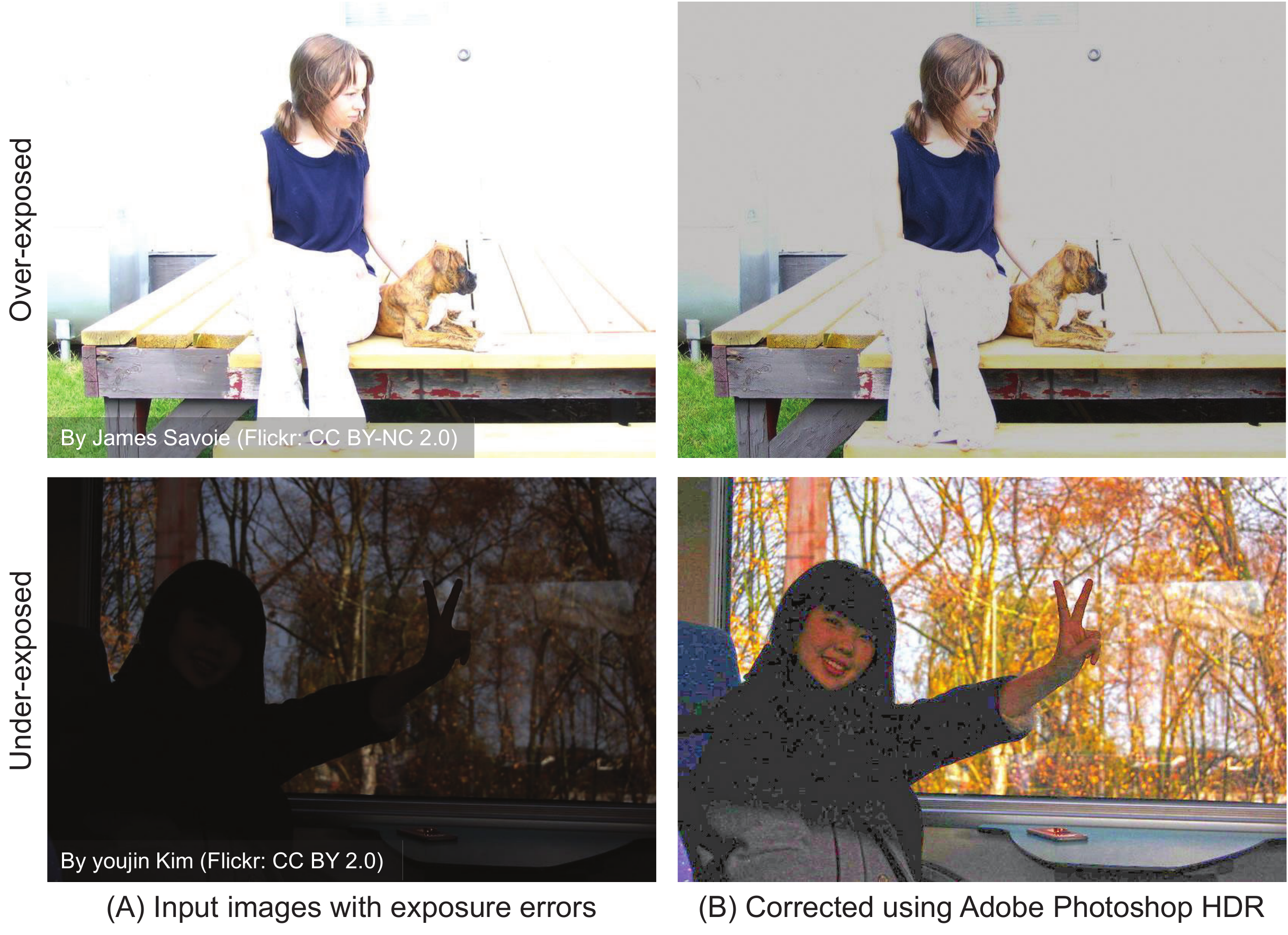}
\vspace{-7mm}
\caption[Examples of camera-rendered images with exposure errors.]{Examples of camera-rendered images with exposure errors. (A) Input images. (B) Corrected images using Adobe Photoshop HDR tool \cite{dayley2010photoshop}. \label{EXPOSURE_ERROR_EXAMPLE}}
\end{figure}

Figure \ref{EXPOSURE_ERROR_EXAMPLE}-(A) shows typical examples of images with exposure errors. In Fig.\ \ref{EXPOSURE_ERROR_EXAMPLE}, exposure errors result in either very bright colors, due to overexposure, or very dark colors, caused by underexposure errors, in the final rendered images. Correcting images with such errors is a challenging task even for well-established image enhancement software packages, see Fig.\ \ref{EXPOSURE_ERROR_EXAMPLE}-(B). Although both over- and underexposure errors are common in photography, most prior work is mainly focused on correcting underexposure errors \cite{guo2017lime, HQEC, Chen2018Retinex, zhang2019kindling, DeepUPE} or generic image quality enhancement \cite{HDRNET, DPE}.

As our focus is on correcting exposure errors in camera-rendered 8-bit per channel sRGB images, we refer the reader to \cite{chen2018learning, hu2018exposure, hasinoff2016burst, liba2019handheld} for representative examples for rendering linear raw-RGB images captured with low light or exposure errors.

\subsection{Exposure Correction }  Traditional methods for exposure correction and contrast enhancement rely on image histograms to re-balance image intensity values \cite{10.5555/559707, pizer1987adaptive, adaptivehisteq, 5773086, lee2013contrast}.  Alternatively,
tone curve adjustment is used to correct images with exposure errors. This process is performed by relying either solely on input image information \cite{yuan2012automatic} or trained deep learning models \cite{yu2018deepexposure, park2018distort, guo2020zero, moran2020deeplpf}. The majority of prior work adopts the Retinex theory \cite{land1977retinex} by assuming that improperly exposed images can be formulated as a pixel-wise multiplication of target images, captured with correct exposure settings, by illumination maps. Mathematically, these methods formulate the problem as follows:

\begin{equation} \label{eq:retinex_theory}
\mat{I} = \mat{S} \odot \mat{\tilde{I}},
\end{equation}

\noindent where $\mat{I}$ is the sRGB camera-captured image, which was captured with some exposure errors or under low-lighting conditions, $\mat{S}$ is an unknown illumination map, and $\mat{\tilde{I}}$ is the reflectance image that was captured with correct exposure settings under normal-lighting conditions. Thus, the goal of most Retinex-based methods is to predict the illumination map, $\mat{S}$, to recover the well-exposed target images. Representative Retinex-based methods include \cite{land1977retinex, jobson1997multiscale, wang2013naturalness, meylan2006high, guo2017lime, HQEC, zhang2019dual} and the most recent deep learning ones \cite{Chen2018Retinex, zhang2019kindling, DeepUPE}. Most of these methods, however, are restricted to correcting underexposure errors \cite{guo2017lime, HQEC, Chen2018Retinex, zhang2019kindling, DeepUPE, yang2020fidelity, xu2020learning, zhu2020eemefn} due to the fact that over-exposed images usually require introducing new content in the corrupted input images, which usually have missing contents as being over-exposed (see Fig.\ \ref{EXPOSURE_ERROR_EXAMPLE}). 

\subsection{HDR Restoration and Image Enhancement } Over-exposed images can be corrected by restoring the high dynamic range (HDR) of the captured-image by reconstructing scene radiance HDR values from one or more low dynamic range (LDR) input images. Prior work either require access to multiple LDR images \cite{mertens2009exposure, kalantari2017deep, endoSA2017} or use a single LDR input image, which is converted to an HDR image by hallucinating missing information~\cite{HDRCNN, moriwaki2018hybrid}. We experimentally found that, however, this single-HDR reconstruction approach is not able to properly deal with images with over-exposure errors (experimental evaluations are given in Chapter\ \ref{ch:ch13}). 

Ultimately, these reconstructed HDR images are mapped back to LDR for perceptual visualization. This mapping can be directly performed from the input multi-LDR images \cite{debevec1997recovering,cai2018learning}, the reconstructed HDR image \cite{yang2018image}, or directly from the single input LDR image without the need for radiance HDR reconstruction \cite{HDRNET, DPE}. There are also methods that focus on general image enhancement that can be applied to enhancing images with poor exposure.  In particular, work by~\cite{DPED, WESPE} was developed primarily to enhance images captured on smartphone cameras by mapping captured images to appear as high-quality images captured by a DSLR.

\subsection{Datasets}
Paired datasets are crucial for supervised learning for image enhancement tasks. Existing paired datasets for exposure correction focus only on low-light underexposed images. Representative examples include Wang et al.'s dataset \cite{DeepUPE} and the low-light (LOL) paired dataset \cite{Chen2018Retinex}. One could think of HDR datasets as an alternative option to train models for exposure correction. However, most of the available HDR datasets have a limited number of scenes that can negatively affect the generalization of deep learning models. For example, Funt et al.'s HDR dataset \cite{funt2010rehabilitation} has only 105 HDR scenes. A relatively large HDR dataset is the HDR+ dataset \cite{hasinoff2016burst}, which has 3,640 bursts of artificially aligned linear raw-RGB images and the corresponding HDR ``ground-truth'' image. These images were \textit{not} intentionally captured with exposure errors, similarly to image quality enhancement datasets, such as the DSLR Photo Enhancement dataset (DPED) \cite{DPED}. Moreover, the ground-truth images in the HDR+ dataset were generated by Google Camera's HDR+ algorithm \cite{hasinoff2016burst, HDRNET}. Thus, it is also arguable that supervision training on this dataset would result in models that learn to mimic procedures applied in \cite{hasinoff2016burst, HDRNET}, rather than learning the underlying mechanisms to correct exposure errors. 

\subsection{Research Directions}
Both under- and over-exposure greatly affect the colors in the image and the overall visual appeal. The problem becomes more challenging when the image is rendered in 8-bit format by unknown camera model. A promising research direction towards enhancing colors in images captured with exposure errors is to consider an accurate image linearization process, that  models the commonly applied camera pipeline procedures, to reconstruct scene-referred images. With an accurate image linearization, one could expect that low-image enhancement algorithms could achieve better results compared to applying the same algorithms to the 8-bit sRGB camera-rendered images. Another promising research direction is to consider over-exposure errors in images. This direction of research requires generating a dataset of \textit{both} exposure errors -- namely, under- and over-exposure errors -- in camera-rendered images. 

\section{Post-Capture Color Editing} \label{sec:post_capture_color_manipulation}

As discussed in Chapter\ \ref{ch:ch2}, camera ISPs apply a set of nonlinear color manipulations as a part of the color rendering process. Once the image is rendered, post-capture color editing can be performed using photo editing techniques and filters. One of the fundamental research directions related to color editing is ``color transfer'', which aims at transferring the colors of a given input image to share the same ``feel'' with another target image colors \cite{faridul2016colour}.

\begin{figure}[!t]
\begin{center}
\includegraphics[width=\textwidth]{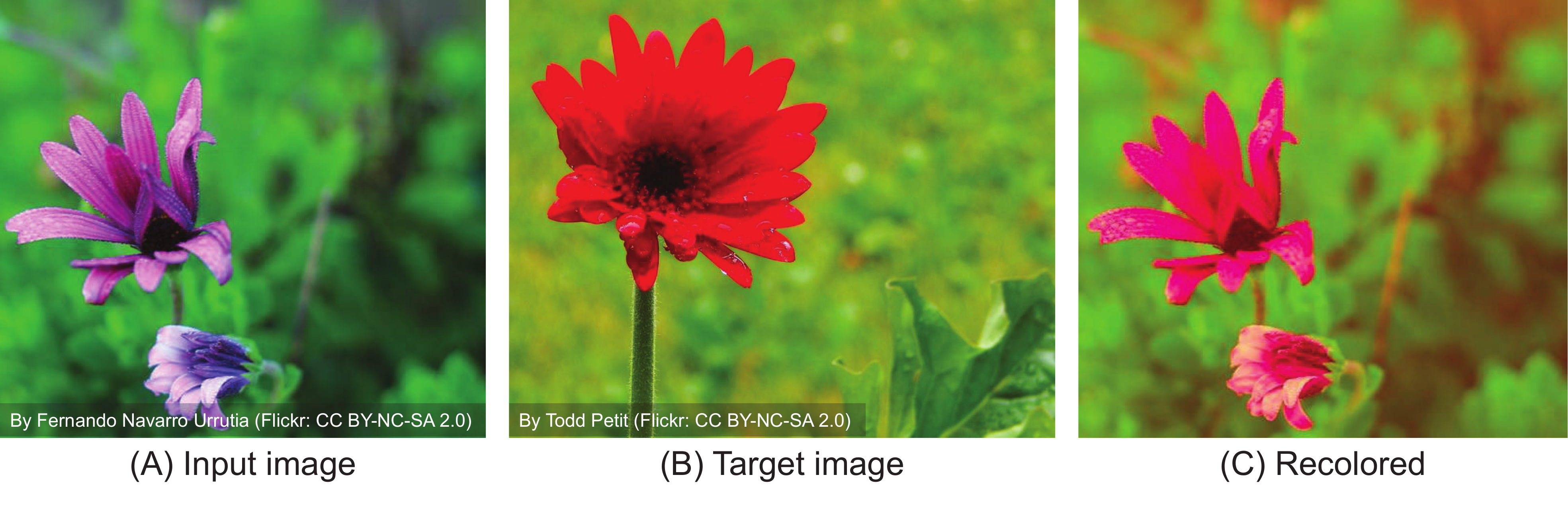}
\end{center}
\vspace{-7mm}
\caption{Examples of color transfer. In this example, we use the color transfer method proposed by Reinhard et al. \cite{reinhard2001color}. (A) Input image. (B) Target image. (C) Recolored image. \label{color_transfer_figure}}
\end{figure}

Figure \ref{color_transfer_figure} shows a typical example of transferring colors from a target image to the input image. As it can be seen, color transfer produces similar effect to image filters provided in photo editing applications and capturing filters provided in smartphone cameras (e.g., \cite{myfilter}). In contrast, recent color transfer methods can achieve more compelling results with the ability to accurately adjusted based on the target image.

Color transfer methods can be categorized into the following categories: (i) geometry-based methods, (ii) statistical-based methods, (iii) learning-based methods, and (iv) color palette-based methods.

\subsection{Geometry-Based Methods}
This category of color transfer methods (e.g., \cite{yamamoto2006color, tehrani2010iterative}) aims to find semantic similarities between input and target images to achieve realistic color mapping. These methods usually rely on feature detection methods, such as the scale-invariant feature transform (SIFT) \cite{lowe2004distinctive} or the speeded-up robust feature (SURF) \cite{bay2006surf}, to extract features of both images---namely, the input and target images. Then, a matching technique (e.g., \cite{beis1997shape}) is used to determine a set of candidate correspondences. Once the candidate correspondences are defined, one simple approach is to build an LUT for color mapping. 

One drawback of these methods is that they mainly rely on the accuracy of finding the candidate correspondences. In many cases, however, it is hard to find reasonable feature correspondences between objects present in both input and target images.

\subsection{Statistical-Based Methods}
Statistical-based methods, on the other hand, aims at transferring statistical properties from the target image to the input image. A simple statistical-based method was proposed by Reinhard et al. \cite{reinhard2001color}, where color mapping was attained by transferring simple statistical moments (mean and standard deviation) between each channel of both images. Specifically, given two images $\mat{I}_\textrm{in}$ and $\mat{I}_\textrm{target}$, transferring the colors of $\mat{I}_\textrm{target}$ to $\mat{I}_\textrm{in}$ to produce a recolored image $\mat{I}_\textrm{recol}$ can be computed as follows \cite{reinhard2001color}:

\begin{equation}
\label{eq_colorTransfer}
\begin{gathered}
\mat{I'}_{\textrm{in}_{L^{*}}} = \mat{I}_{\textrm{in}_{L^{*}}} - \mu_{\textrm{in}_{L^*}},
\hspace{4mm}
\mat{I'}_{\textrm{in}_{a^{*}}} = \mat{I}_{\textrm{in}_{a^{*}}} - \mu_{\textrm{in}_{a^{*}}},
\\
\mat{I'}_{\textrm{in}_{b^{*}}} = \mat{I}_{\textrm{in}_{b^{*}}} - \mu_{\textrm{in}_{b^{*}}},
\\
\mat{I}_{{\texttt{recol}}_{L^*}} = \left(\sigma_{\textrm{target}_{L*}} / \sigma_{\textrm{in}_{L^*}}\right) \text{ } \mat{I'}_{\textrm{in}_{L^{*}}} + \mu_{\textrm{target}_{L^*}},
\\
\mat{I}_{{\texttt{recol}}_{a^*}} =  \left(\sigma_{\textrm{target}_{a*}} / \sigma_{\textrm{in}_{a^*}}\right) \text{ } \mat{I'}_{\textrm{in}_{a^{*}}} + \mu_{\textrm{in}_{a^{*}}},
\\
\mat{I}_{{\texttt{recol}}_{b^*}} =  \left(\sigma_{\textrm{target}_{b*}} /\sigma_{\textrm{in}_{b^*}}\right) \text{ } \mat{I'}_{\textrm{in}_{b^{*}}} + \mu_{\textrm{in}_{b^{*}}},
\end{gathered}
\end{equation}

\noindent where $\mat{I}_{L^{*}}$, $\mat{I}_{a^{*}}$, and $\mat{I}_{b^{*}}$ are the CIE $\texttt{L}^{*}$, $\texttt{a}^{*}$, $\texttt{b}^{*}$ components of an image $\mat{I}$, respectively, $\mu$ is the arithmetic mean, and $\sigma$ is the standard deviation.  Despite the various methods proposed to improve the baseline color transfer method \cite{reinhard2001color} described in Eq. \ref{eq_colorTransfer} (e.g., \cite{pitie2005n, pitie2007automated, xiao2006color, nguyen2014illuminant, finlayson2017color, he2017neural}), all statistical-based methods aims to find a better histogram mapping between images.

For instance, Piti\'e et al. \cite{pitie2005n, pitie2007automated} proposed to map the entire 3-dimensional histogram of the source image using a 3D rotation matrix computed iteratively. Nguyen et al. \cite{nguyen2014illuminant} he showed that WB both images (input and target) provides a fast alignment mechanism for color mapping. Thus, a heuristic technique based on image white balancing was proposed by iteratively computing a linear transformation matrix for histogram mapping. Though such statistical-based methods have a less restriction compared to geometry-based methods, the results are not always realistic as such methods do not consider semantic information in both images. 

\subsection{Learning-Based Methods}
Despite the efficiency of using color histograms for color mapping, recent deep learning methods mostly use images as an input without an explicit reliance on color histograms. These methods do not only transfer colors between images, but also consider the texture information \cite{gatys2015neural, gatys2016image, johnson2016perceptual, ulyanov2016instance, isola2017image, luan2017deep, sheng2018avatar}. That is, the goal of these methods is to achieve image ``style'' transfer. Such learning-based style transfer methods can be categorized to: (i) image-optimization-based methods and (ii) model-optimization-based methods \cite{jing2019neural}. 

The majority of neural style transfer work belongs to the first category, where an online optimization process is performed in order to transfer the style of a target image to the input image (e.g., \cite{gatys2015neural, gatys2017controlling, risser2017stable, sheng2018avatar}). The first deep learning endeavor to produce images with artistic-style is Deep Dream \cite{deepdream}. Deep Dream works by reversing a CNN, pre-trained for image classification, using image-optimization. That is, the process begins with a noise image, which is iteratively updated through an optimization process to make the CNN predict a certain output class.

Inspired by Deep Dream, Gatys et al. \cite{gatys2015neural} proposed to transfer the style of target image to the source image by transferring statistics of the target image representations from intermediate layers of a pre-trained network (i.e., deep features) to the corresponding input image deep features' statistics. In order to retain the source image content in the output image, while transferring only the style from the target image, the optimization process minimizes a dual objective function. This objective function mainly includes a content loss (i.e., similarity between input and output images' contents) and style loss (i.e., similarity between target and output images' styles). The former can be computed using the squared Euclidean distance between deep features of the input and output images. The latter, however, requires a mechanism to model the visual texture to encourage the network to transfer the style of the target image to the source image. To that end, the Gram matrix is used to encode the correlations between deep features of different layers in the pre-trained classification network \cite{gatys2015neural}. 

\begin{figure}[!t]
\includegraphics[width=\textwidth]{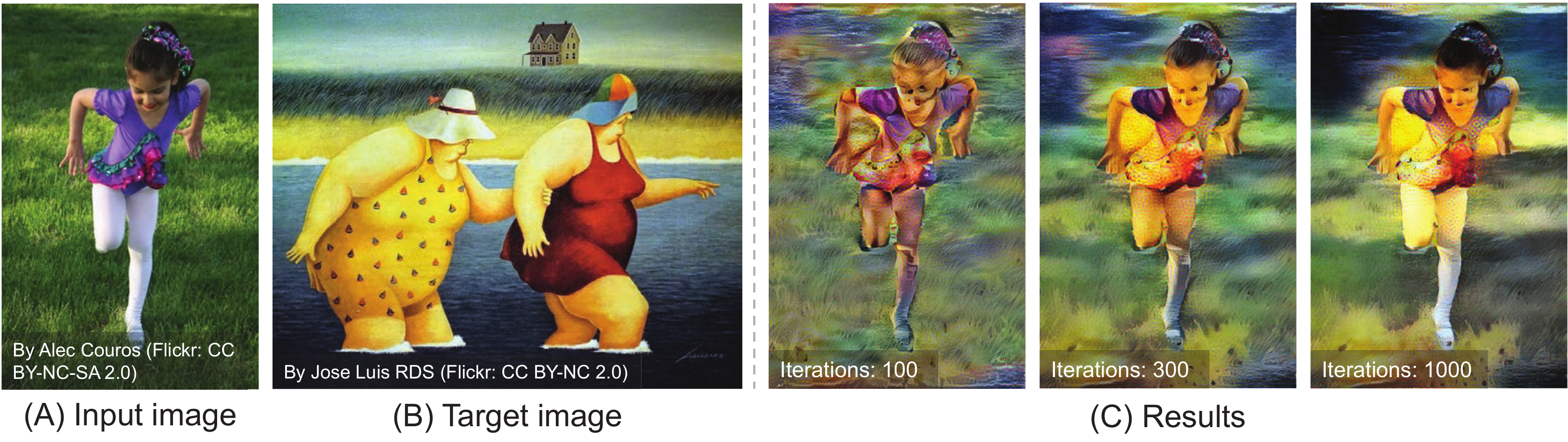}
\vspace{-7mm}
\caption{Examples of neural style transfer. In this example, we show the results of the method proposed by Gaty et al. \cite{reinhard2001color}. (A) Input image. (B) Target image. (C) Results of optimization after different number of iterations. \label{style_transfer_example}}
\end{figure}

In spite of the impressive results achieved by the image-optimization-methods (see Fig.\ \ref{style_transfer_example}), these methods are usually inefficient for interactive applications. For example, Gaty et al.'s optimization method \cite{gatys2015neural} takes $\sim$4 minutes to process a single image using a single GTX 1080 GPU.  On the other hand, model-optimization-based methods for neural style transfer offers a more efficient solution. However, these model-optimization-based methods are mostly limited to transfer a single style per model -- i.e., a new model should be re-trained for any new target style -- (representative examples include  \cite{zhang2017style, johnson2016perceptual, ulyanov2016texture, ulyanov2017improved}). There are a few attempts to achieve multiple-style transfer per model \cite{li2017universal} or arbitrary-style transfer per-model \cite{li2018closed}. The quality of the results produced by these methods, however, mostly depends on the matching degree of the semantic content and/or the colors between both images---namely, the input and target images \cite{he2019progressive}. 

\subsection{Color Palette-Based Methods}
Color palette-based methods offer another type of color mapping, where the target image is replaced by a target color palette. Color palette is a compact representation of the main colors of a given image \cite{chang2015palette}. Palette-based methods assume that colors of a given image $\mat{I}$ can be approximately represented by a linear combination of a small number of colors (color palettes) as follows:
\begin{equation}\label{palette_eq_related_work}
\mat{I}^{(p)} = \sum_{i=1}^{k} \mat{w}_i^{(p)} \mat{C}_i,
\end{equation}
\noindent where $\mat{C}_i$ is the $i^{\text{th}}$ color in the color palette and $k$ is the total number of colors in this color palette, $\mat{w}_i^{(p)}$ is a weighting factor that represent the contribution of the color $\mat{C}_i$ in forming the color in the original image $\mat{I}$ at pixel $p$. As shown in Eq.\ \ref{palette_eq_related_work}, by defining the unknowns $\mat{C}$ and $\mat{w}$, the user can interactively change the palette colors $\mat{C}_i$ ($i\in{1, ..., k}$) to recolor the input image $I$. Palette-based methods (e.g., \cite{chang2015palette, tan2016decomposing, aksoy2016interactive, aksoy2017unmixing, tan2018efficient, afifi2019dynamic}) introduced different ways in order to find the most effective set of colors in the color palette and utilized regularized color decomposition optimization techniques to compute the weighting factors.  

Similar to other color mapping techniques, color palette-based approaches may introduce artifacts (e.g., color bleeding) in the recolored images based on the changes in the target color palette. 

\subsection{Research Directions}
By definition, color transfer requires a target image/color palette in order to transfer the colors of this target image to the input image. These image exemplars, however, have a direct impact on the quality of the color transfer process \cite{faridul2016colour}. This motivated a few methods towards achieving automatic color transfer (e.g., \cite{laffont2014transient, huang2014learning}). These methods, however, are usually restricted to deal with certain type of images (e.g., outdoor images \cite{laffont2014transient}). Prior work shows the significant role of WB on changing the global colors of images \cite{nguyen2014illuminant}. A potential research direction is to use WB to achieve image color manipulation in the post-capture stage. However, as discussed earlier, WB editing, or chromatic adaptation generally, is applied on linear images (i.e., raw or CIE XYZ images). The challenge would be to achieve accurate WB manipulation on camera-rendered images. Another interesting research direction could be auto image recoloring, where different recolored versions of the input image are produced automatically without any user interaction required. This auto recoloring should consider semantic content present in the image in order to produce realistic recolored images.

\section{Summary}

We have provided a survey of different approaches for color correction and editing. We began by reviewing existing methods for image white balancing including illuminant estimation and chromatic adaptation techniques for scene-referred linear images. Furthermore, we have explained why the current solutions cannot deal with improperly white-balanced camera-rendered images due to the nonlinearity applied on board cameras. Afterward, we briefly discussed other factors that have a significant effect on the quality of camera-rendered image colors. Specifically, we have discussed exposure errors in photographs and reviewed existing methods to enhance images rendered with low-light conditions/exposure errors. Then, we have reviewed post-capture color transfer methods. This chapter also has discussed promising research directions for color correction and editing, which motivated our work presented in the next chapters of this thesis.

\part{Computational Color Constancy\label{part:cc}}

\chapter{Sensor-Independent Color Constancy \label{ch:ch5}}


The previous chapter introduced a learning-based procedure to improve CC. While learning-based methods for illuminant estimation (especially the modern deep neural networks) achieve state-of-the-art results, it is currently necessary to train a separate DNN for each type of camera sensor. This means when a camera manufacturer uses a new sensor, it is necessary to re-train an existing DNN model with training images captured by the new sensor. This chapter addresses this problem by introducing a novel sensor-independent illuminant estimation framework \footnote{This work was published in \cite{afifi2019sensor, afifi2021systems}: Mahmoud Afifi and Michael S. Brown. Sensor-Independent Illumination Estimation for DNN Models. In 
British Machine Vision Conference (BMVC), 2019.}. Our method learns a sensor-independent {\it working space} that can be used to canonicalize the RGB values of any arbitrary camera sensor. Our learned space retains the linear property of the original sensor raw-RGB space and allows unseen camera sensors to be used on a single DNN model trained on this working space.  We demonstrate the effectiveness of this approach on several different camera sensors and show it provides performance on par with state-of-the-art methods that were trained per sensor. The source code of this work is available on GitHub: \href{https://github.com/mahmoudnafifi/SIIE}{https://github.com/mahmoudnafifi/SIIE}.

\section{Introduction}
\label{siie:sec:intro}
Recall that in Chapter \ref{ch:ch3}, we have described the computational CC in terms of the physical image formation process as follows. Let $\mat{I} = \{\mat{I}_r, \mat{I}_g, \mat{I}_b\}$ denote an image captured in the linear raw-RGB space. The value of each color channel $c = \{\text{R}, \text{G}, \text{B}\}$ for a pixel located at $x$ in $\mat{I}$ is given by the following equation \cite{basri2003lambertian}:
\begin{equation}
\label{siie:eq0}
\mat{I}_c(x) =\int_{\gamma} \rho(x,\lambda)R(x,\lambda)S_{c}(\lambda) d\lambda,
\end{equation}
\noindent
where $\gamma$ is the visible light spectrum (approximately 380nm to 780nm), $\rho(\cdot)$ is the illuminant spectral power distribution, $R(\cdot)$ is the captured scene's spectral reflectance properties, and $S(\cdot)$ is the camera sensor response function at wavelength $\lambda$. The problem can be simplified by assuming a single uniform illuminant in the scene as follows:
\begin{equation}
\label{siie:eq1}
\mat{I}_c =  \mat{\light}_{c}\!\times\!\mat{R}_{c},
\end{equation}
\noindent
where $\mat{\light}_{c}$ is the scene illuminant value of color channel $c$. A standard approach to this problem is to use a linear model (i.e., a $3\!\times\!3$ diagonal matrix) to such that $\mat{\light}_{\text{R}} = \mat{\light}_{\text{G}} = \mat{\light}_{\text{B}}$ (i.e., white illuminant).

\begin{figure}
\includegraphics[width=\linewidth]{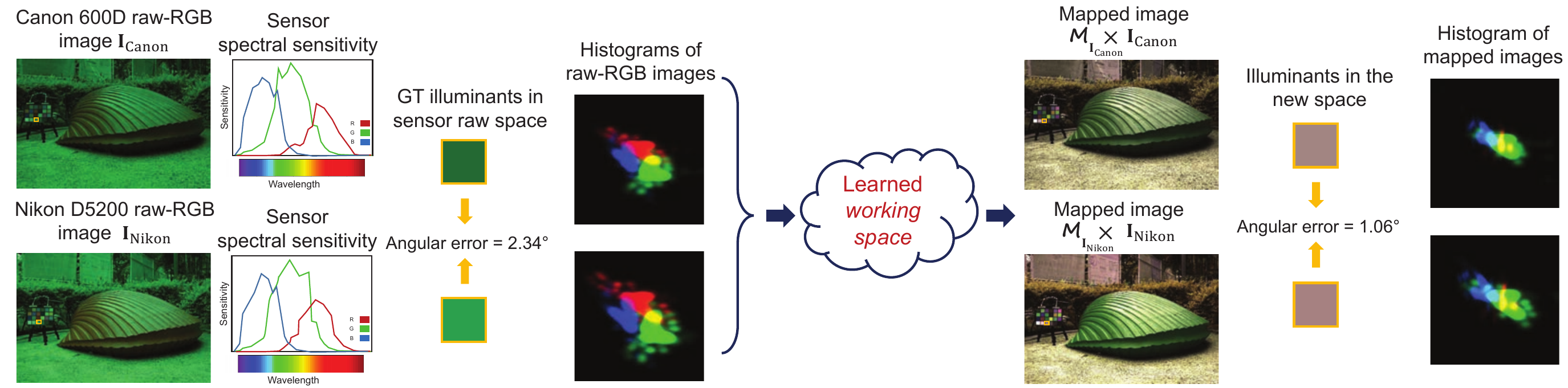}
\vspace{-7mm}
\caption[A scene captured by two different camera sensors results in different ground truth illuminants due to different camera sensor responses.]{A scene captured by two different camera sensors results in different ground truth illuminants due to different camera sensor responses. We learn a device-independent {\it working space} that reduces the difference between ground truth illuminants of the same scenes.}
\label{siie:fig:fig1}
\end{figure}

Typically, $\mat{\light}$ is unknown and should be defined to obtain the true objects' body reflectance values $\mat{R}$ in the input image $\mat{I}$. The value of $\mat{\light}$ is specific to the camera sensor response function $S(\cdot)$, meaning that the same scene captured by different camera sensors results in different values of $\mat{\light}$. Figure \ref{siie:fig:fig1} shows an example.

Illuminant estimation methods aim to estimate the value $\mat{\light}$ from the sensor's raw-RGB image. Recently, DNN methods have demonstrated state-of-the-art results for the illuminant estimation task. These approaches, however, need to train the DNN model per camera sensor. This is a significant drawback. When a camera manufacture decides to use a new sensor, the DNN model will need to be retrained on a new image dataset captured by the new sensor. Collecting such datasets with the corresponding ground-truth illuminant raw-RGB values is a tedious process. As a result, many AWB algorithms deployed on cameras still rely on simple statistical-based methods even though the accuracy is not comparable to those obtained by the learning-based methods.

\begin{figure}
\includegraphics[width=\linewidth]{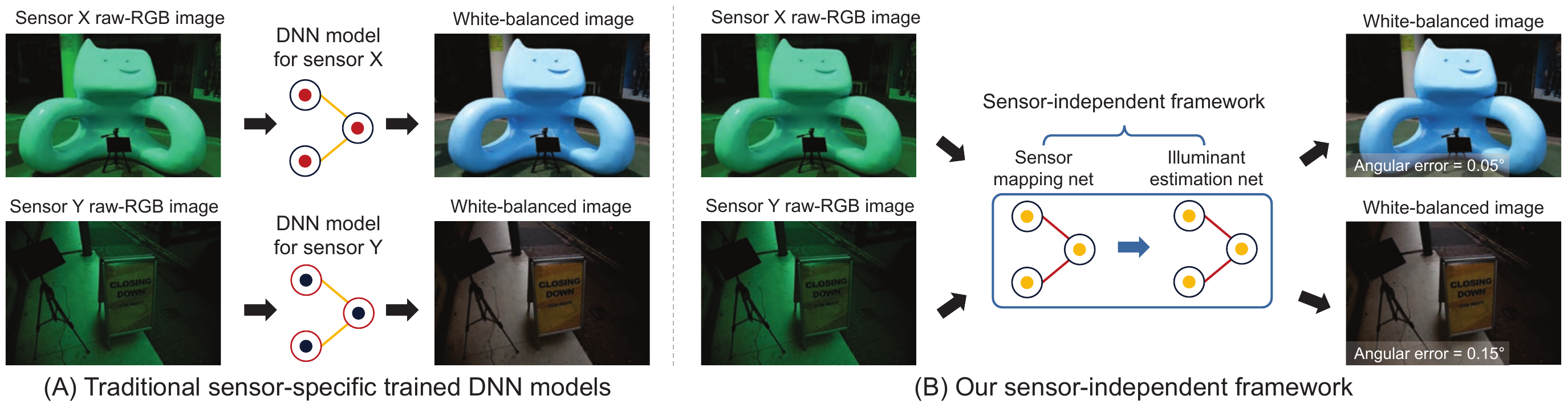}
\vspace{-7mm}
\caption[(A) Traditional learning-based illuminant estimation methods train or fine-tune a model per camera sensor. (B) Our method can be trained on images captured by different camera sensors and generalizes well for unseen camera sensors.]{(A) Traditional learning-based illuminant estimation methods train or fine-tune a model per camera sensor. (B) Our method can be trained on images captured by different camera sensors and generalizes well for unseen camera sensors. Shown images are rendered in the sRGB color space by the camera imaging pipeline in \cite{karaimer2016software} to aid visualization.}
\label{siie:fig:teaser}
\end{figure}

\paragraph{Contribution}~In this chapter, we introduce a sensor-independent learning framework for illuminant estimation. The idea is similar to the color space conversion process applied onboard cameras that maps the sensor-specific RGB values to a perceptual-based color space -- namely, CIE XYZ. The color space conversion process estimates a color space transform (CST) matrix to map white-balanced sensor-specific raw-RGB images to CIE XYZ~\cite{ramanath2005color, karaimer2016software}.  This process is applied onboard cameras {\it after} the illuminant estimation and white-balance step, and relies on the estimated scene illuminant to compute the CST matrix~\cite{can2018improving}. Our solution, however, is to learn a new space that is used \textit{before} the illuminant estimation step. Specifically, we design a novel unsupervised deep learning framework that learns how to map each input image, captured by arbitrary camera sensor, to a non-perceptual sensor-independent {\it working space}. Mapping input images to this space, allows us to train our model using training sets captured by different camera sensors achieving good accuracy and generalizing well for unseen camera sensors as shown in Fig. \ref{siie:fig:teaser}.

\section{Proposed Method}
\label{siie:sec:method}

\begin{figure}
\includegraphics[width=\linewidth]{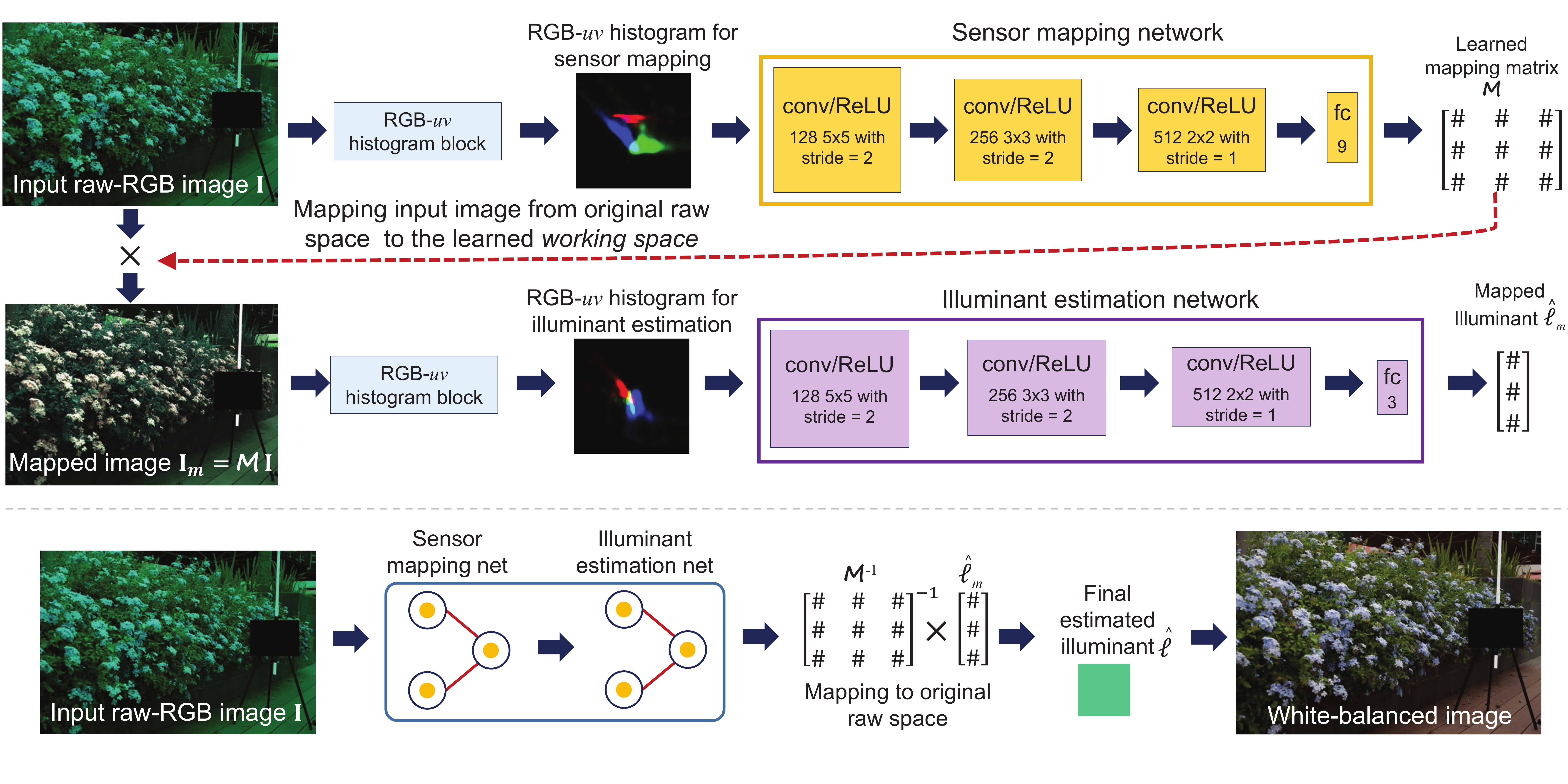}
\vspace{-7mm}
\caption[Our proposed sensor-independent illuminant estimation (SIIE) framework consists of two networks: (i) a sensor mapping network and (ii) an illuminant estimation network.]{Our proposed sensor-independent illuminant estimation (SIIE) framework consists of two networks: (i) a sensor mapping network and (ii) an illuminant estimation network. Our networks are trained jointly in an end-to-end manner to learn an image-specific mapping matrix (resulting from the sensor mapping network) and scene illuminant in the learned space (resulting from the illuminant estimation network). The final estimated illuminant is produced by mapping the result illuminant from our learned space to the input image's camera-specific raw space.}
\label{siie:fig:main}
\end{figure}

Figure \ref{siie:fig:main} provides an overview of our sensor-independent illuminant estimation (SIIE) framework. Our SIIE accepts thumbnail ($150\!\times\!150$ pixels) linear raw-RGB images, captured by an arbitrary camera sensor and estimates scene illuminant RGB vectors in the same space of input images. We decided to make SIIE accepting thumbnail images instead of full-sized images because such thumbnail images are often already produced by camera ISPs, and processing these downsized images mostly produces illuminant estimation accuracy that is on par with the results using full-sized images, but with less memory and computational power \cite{FFCC}.

We rely on color distribution of input thumbnail image $\mat{I}$ to estimate an image-specific transformation matrix that maps the input image to our working space. This mapping allows us to accept images captured by different sensors and estimate scene illuminant values in the original space of input images.

We begin with the formulation of our problem followed by a detailed description of our framework components and the training process. Note that we will assume input raw-RGB images are represented as $3\!\times\!n$ matrices, where $n=150\!\times\!150$ is the total number of pixels in the thumbnail image and the three rows represent the R, G, and B values.

\subsection{Problem Formulation}
\label{siie:subsec:overview}
We propose to work in a new learned space for illumination estimation. This space is sensor-independent and retains the linear property of the original raw-RGB space. To that end, we introduce a learnable $3\!\times\!3$ matrix $\mat{\mathcal{M}}$ that maps an input image $\mat{I}$ from its original sensor-specific space to a new working space. We can reformulate Eq. \ref{siie:eq1} as follows:
\begin{equation}
\label{siie:eq2}
\mat{\mathcal{M}}^{-1}\mat{\mathcal{M}}\mat{I} =  \texttt{diag}(\mat{\mathcal{M}}^{-1}\mat{\mathcal{M}}\mat{\light}) \mat{R},
\end{equation}
\noindent
where $\texttt{diag}(\cdot)$ is a diagonal matrix and  $\mat{\mathcal{M}}$ is a learned matrix that maps arbitrary sensor responses to a sensor-independent space.

Given a mapped image $\mat{I}_m = \mat{\mathcal{M}}\mat{I}$ in our learned space, we aim to estimate the mapped vector $\mat{\light}_m = \mat{\mathcal{M}}\mat{\light}$ that represents the scene illumination values of $\mat{I}_m$ in the new space. The original scene illuminant (represented in the original sensor raw-RGB space) can be reconstructed by the following equation:
\begin{equation}
\label{siie:eq3}
\mat{\light} = \mat{\mathcal{M}}^{-1}\mat{\light}_m.
\end{equation}

\subsection{RGB--$uv$ Histogram Block}
\label{siie:subsec:RGB-uvHist}
Prior work has shown that the illumination estimation problem is related primarily to the image's color distribution~\cite{CCC, FFCC}.  Accordingly, we use the image's color distribution as an input for our SIIE. Representing the image using a full 3D RGB histogram requires significant amounts of memory -- for example, a $256^3$ RGB histogram requires more than 16 million entries. Down-sampling the histogram -- for example, to 64-bins -- still requires a considerable amount of memory. Thus, we use a projected histogram feature. Our histogram feature is inspired by prior work~\cite{CCC, FFCC}, in which we construct our feature in the log-chrominance space \cite{drew2003recovery, finlayson2001color}, which represents the color distribution of an image $\mat{I}$ as an $m\!\times\!m\!\times\!3$ tensor that is parameterized by $uv$. We refer to this as an RGB-$uv$ histogram. 

We use two learnable parameters to control the contribution of each color channel in the generated histogram and the smoothness of histogram bins. Specifically, our RGB-$uv$ histogram block represents the color distribution of an image $\mat{I}$ as a three-layer histogram $\mat{H}(\mat{I})$  represented as an $m\!\times\!m\!\times 3$ tensor. The produced histogram $\mat{H}(\mat{I})$ is parameterized by $uv$ and computed as follows:
\begin{equation}
\label{siie:eq2_UVHist}
\begin{gathered}
\mat{I}_y(i) = \sqrt{\mat{I}_{\textrm{R}(i)}^{2} + \mat{I}_{\textrm{G}(i)}^{2} + \mat{I}_{\textrm{B}(i)}^{2}},
\\
\mat{I}_{u1(i)} = \log{\left(\frac{ \mat{I}_{\textrm{R}(i)}}{\mat{I}_{\textrm{G}(i)}} + \epsilon\right)} \textrm{ , }  \mat{I}_{v1(i)} = \log{\left(\frac{ \mat{I}_{\textrm{R}(i)}}{\mat{I}_{\textrm{B}(i)}}+ \epsilon\right)},
\\
\mat{I}_{u2(i)} = \log{\left(\frac{ \mat{I}_{\textrm{G}(i)}}{\mat{I}_{\textrm{R}(i)}}+ \epsilon\right)} \textrm{ , }  \mat{I}_{v2(i)} = \log{\left(\frac{ \mat{I}_{\textrm{G}(i)}}{\mat{I}_{\textrm{B}(i)}}+ \epsilon\right)},
\\
\mat{I}_{u3(i)} = \log{\left(\frac{ \mat{I}_{\textrm{B}(i)}}{\mat{I}_{\textrm{R}(i)}}+ \epsilon\right)} \textrm{ , }  \mat{I}_{v3(i)} = \log{\left(\frac{ \mat{I}_{\textrm{B}(i)}}{\mat{I}_{\textrm{G}(i)}}+ \epsilon\right)},
\\
\mat{H}\left(\mat{I}\right)_{(u,v,c)} = \left(s_{c} \sum_{i} \mat{I}_{y(i)}  \exp{\left(-\left| \mat{I}_{uc(i)} - u \right|/\sigma_c^2\right)}   \exp{\left(-\left| \mat{I}_{vc(i)} - v \right|/\sigma_c^2\right)}\right)^{1/2},
\end{gathered}
\end{equation}
\noindent
where  $i = \{1,...,n\}$, $c \in {\{1, 2, 3\}}$ represents each color channel in $\mat{H}$, $\epsilon$ is a small positive constant added for numerical stability, and $s_{c}$ and $\sigma_c$ are learnable scale and fall-off parameters, respectively. The scale factor $s_{c}$ controls the contribution of each layer in our histogram, while the fall-off factor $\sigma_c$ controls the smoothness of the histogram's bins of each layer. The values of these parameters (i.e., $s_{c}$ and $\sigma_c$) are learned during the training phase.

\subsection{Network Architecture}
\label{siie:subsec:net}

As shown in Fig. \ref{siie:fig:main}, our framework consists of two networks: (i) a sensor mapping network and (ii) an illuminant estimation network. The input to each network is the RGB-$uv$ histogram feature produced by our histogram block. The sensor mapping network accepts an RGB-$uv$ histogram of a thumbnail raw-RGB image $\mat{I}$ in its original sensor space, while the illuminant estimation network accepts RGB-$uv$ histograms of the mapped image $\mat{I}_m$ to our learned space. In our implementation, we use $m=61$ and each histogram feature is represented by a $61\!\times\!61\!\times\!3$ tensor.

We use a simple network architecture for each network. Specifically, each network consists of three conv/ReLU layers followed by a fully connected (fc) layer. The kernel size and stride step used in each conv layer are shown in Fig. \ref{siie:fig:main}.

In the sensor mapping network, the last fc layer has nine neurons. The output vector $\mat{v}$ of this fc layer is reshaped to construct a $3\!\times\!3$ matrix $\mat{V}$, which is used to build $\mat{\mathcal{M}}$ as described in the following equation:
\begin{equation}
\mat{\mathcal{M}} =  \frac{1}{\lVert\mat{V}\rVert_1 + \epsilon} \left|\mat{V}\right|,
\end{equation}
\noindent where $\left|\cdot\right|$ is the modulus (absolute magnitude), $\lVert\cdot\rVert_1$ is the matrix 1-norm, and  $\epsilon = 10^{-8}$ is added for numerical stability. The modulus step is necessary to avoid negative values in the mapped image $\mat{I}_m$, while the normalization step is used to avoid having extremely large values in $\mat{I}_m$. Note the values of $\mat{\mathcal{M}}$ are image-specific, meaning that its values are produced based on the input image's color distribution in the original raw-RGB space.

There are three neurons in the last fc layer of the illuminant estimation network to produce illuminant vector $\hat{\mat{\light}}_m$ of the mapped image $\mat{I}_m$. Note that the estimated vector $\hat{\mat{\light}}_m$ represents the scene illuminant in our learned space. The final result is obtained by mapping $\hat{\mat{\light}}_m$ back to the original space of $\mat{I}$ using Eq. \ref{siie:eq3}.

\begin{figure}
\includegraphics[width=\linewidth]{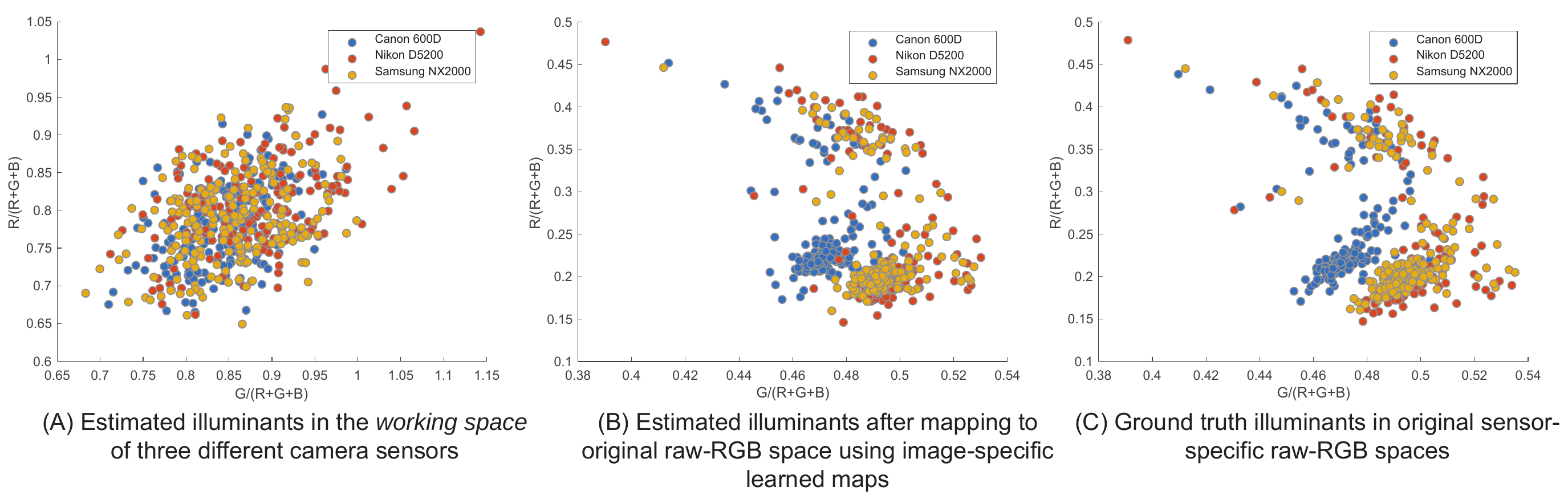}
\vspace{-7mm}
\caption[Raw-RGB images capture the same set of scenes using three different cameras taken from the NUS 8-Cameras dataset \cite{cheng2014illuminant}.]{Raw-RGB images capture the same set of scenes using three different cameras taken from the NUS 8-Cameras dataset \cite{cheng2014illuminant}. (A) Estimated illuminants resulted from the illuminant estimation network in our learned {\it working space}. (B) Estimated illuminants after mapping to the original raw-RGB space. This mapping is performed by multiplying each illuminant vector by the inverse of the learned image-specific mapping matrix (resulting from the sensor mapping network). (C) Corresponding ground truth illuminants in the original raw-RGB space of each image.}
\label{siie:fig:points}
\end{figure}

\begin{figure}[t]
\includegraphics[width=\linewidth]{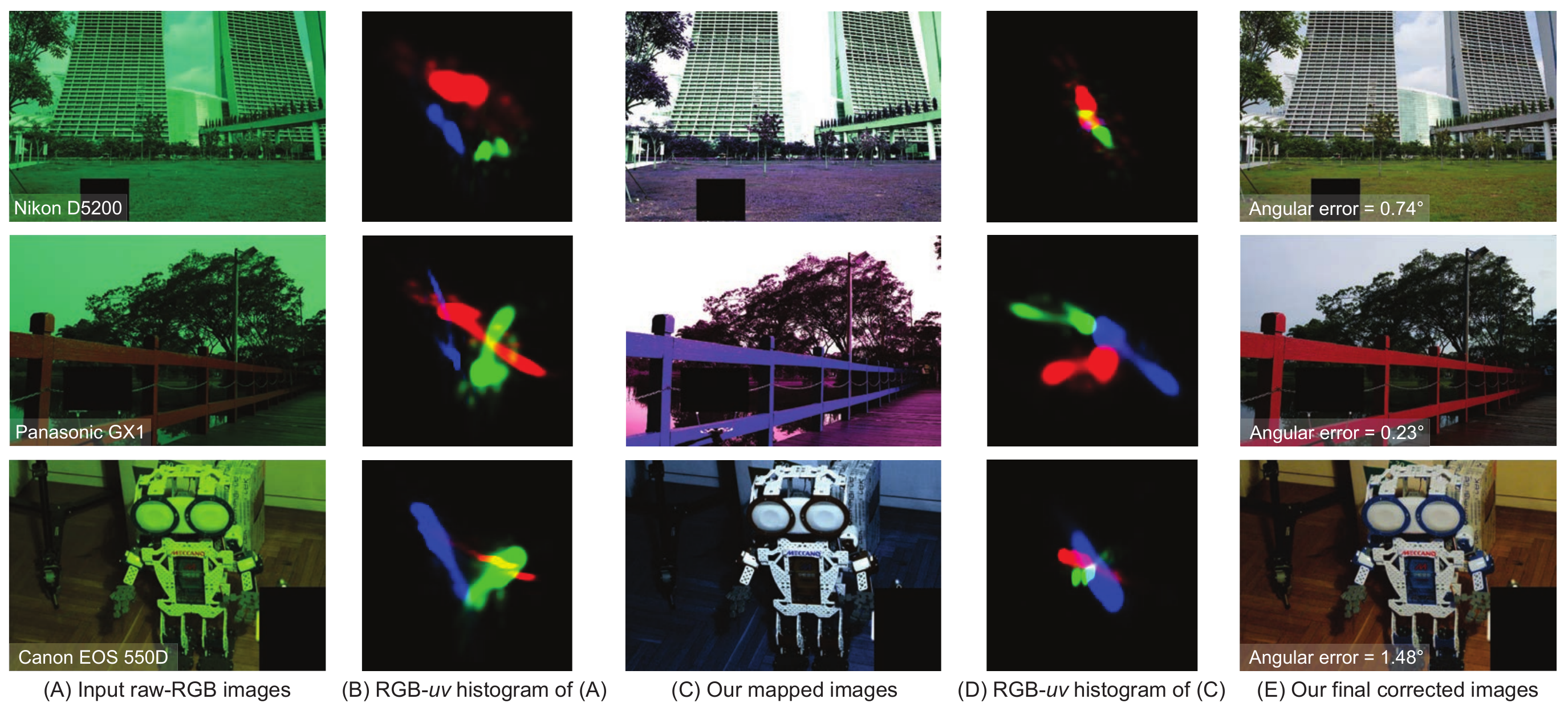}
\vspace{-7mm}
\caption[Example of our generated RGB-$uv$ histograms.]{Example of our generated RGB-$uv$ histograms. (A) Input raw-RGB images. (B) Generated histograms of images in (A). (C) After mapping images in (A) to the learned space. (D)
Generated histograms of images in (C).
(E) After correcting images in (A) based on our estimated illuminants. Shown images are rendered in the sRGB color space by the camera imaging pipeline in \cite{karaimer2016software} to aid visualization.}
\label{siie:fig:hists}
\end{figure}

\subsection{Training}
\label{siie:subsec:training}
We jointly train our sensor mapping and illuminant estimation networks in an end-to-end manner using the adaptive moment estimation (Adam) optimizer \cite{kingma2014adam} with a decay rate of gradient moving average $\beta_1=0.85$, a decay rate of squared gradient moving average $\beta_2=0.99$, and a mini-batch with eight observations at each iteration. We initialized both network weights with Xavier initialization \cite{glorot2010understanding}. The learning rate was set to $10^{-5}$ and decayed every five epochs.

We adopt the recovery angular error (referred to as the angular error) as our loss function \cite{hordley2004re}. The angular error is computed between the ground truth illuminant $\mat{\light}$ and our estimated illuminant $\hat{\mat{\light}}_m$ after mapping it to the original raw-RGB space of training image $\mat{I}$. The loss function can be described by the following equation:
\begin{equation}
\lossfun(\hat{\mat{\light}}_m, \mat{\mathcal{M}}) = \cos^{-1}\left( \frac{ \mat{\light} \cdot \left(\mat{\mathcal{M}}^{-1} \hat{\mat{\light}}_m\right)}{\lVert \mat{\light} \rVert \lVert  \mat{\mathcal{M}}^{-1} \hat{\mat{\light}}_m\rVert}\right),
\end{equation}
where $\lVert\cdot\rVert$ is the Euclidean norm, and ($\cdot$) is the vector dot-product.

As the values of $\mat{\mathcal{M}}$ are produced by the sensor mapping network, there is a possibility of producing a singular matrix output. In this case, we add small  offset $\mathcal{N}(0,1)\!\times\!10^{-4}$ to each parameter in $\mat{\mathcal{M}}$ to make it invertible.

At the end of the training process, our framework learns an image-specific matrix $\mat{\mathcal{M}}$ that maps an input image taken by an arbitrary sensor to the learned space. Figure \ref{siie:fig:points} shows an example of three different camera responses capturing the same set of scenes. As shown in Fig. \ref{siie:fig:points}-(A), the estimated illuminants of these sensors are bounded in the learned space. These illuminants are mapped back to the original raw-RGB sensor space of the corresponding input images using Eq. \ref{siie:eq3}. As shown in Fig. \ref{siie:fig:points}-(B) and Fig. \ref{siie:fig:points}-(C), our final estimated illuminants are close to the ground truth illuminants of each camera sensor. Figure \ref{siie:fig:hists} shows examples of the generated histograms of input images in original raw-RGB space and our learned space.

\section{Experimental Results}
\label{siie:sec:results}

In our experiments, we used all cameras of three different datasets, which are: (i) NUS 8-Camera \cite{cheng2014illuminant}, (ii) Gehler-Shi \cite{gehler2008bayesian}, and (iii) Cube+ \cite{banic2017unsupervised} datasets. In total, we have 4,014 raw-RGB images captured by 11 different camera sensors.

We followed the leave-one-out cross-sensor validation scheme to evaluate our method. Specifically, we excluded all images captured by one camera for testing and trained a model with the remaining images. This process was repeated for all cameras. 

We also tested our SIIE on the Cube dataset. In this experiment, we used a trained model on images from the NUS and Gehler-Shi datasets, and excluded all images from the Cube+ dataset. 

The calibration objects (i.e., X-Rite color chart or SpyderCUBE) were masked out in both training and testing processes.

\begin{table}[]
\caption[Angular errors on the NUS 8-Cameras \cite{cheng2014illuminant} and Gehler-Shi \cite{gehler2008bayesian} datasets.]{Angular errors on the NUS 8-Cameras \cite{cheng2014illuminant} and Gehler-Shi \cite{gehler2008bayesian} datasets. Methods highlighted in gray are trained/tuned for each camera sensor (i.e., sensor-specific models). The lowest errors are highlighted in yellow. \label{siie:Table1}}
\parbox{.48\linewidth}{
\centering
\scalebox{0.5}{
\begin{tabular}{l|cccc}
\textbf{\begin{tabular}[c]{@{}l@{}}\textbf{NUS 8-Cameras Dataset}\\\textbf{Method} \end{tabular}}& \textbf{Mean} & \textbf{Med.} & \textbf{\begin{tabular}[c]{@{}l@{}}Best \\ 25\%\end{tabular}} & \textbf{\begin{tabular}[c]{@{}l@{}}Worst \\ 25\%\end{tabular}} \\ \hline
White-Patch \cite{maxRGB} & 9.91 & 7.44 & 1.44 & 21.27 \\
Pixel-based Gamut \cite{PixelGamut} & 5.27 & 4.26 & 1.28 & 11.16 \\
GW \cite{GW} & 4.59 & 3.46 & 1.16 & 9.85 \\
Edge-based Gamut \cite{PixelGamut}& 4.40 & 3.30 & 0.99 & 9.83 \\
SoG \cite{SoG}& 3.67 & 2.94 & 0.98 & 7.75 \\
\cellcolor[HTML]{EFEFEF}Bayesian \cite{gehler2008bayesian} & 3.50 & 2.36 & 0.78 & 8.02 \\
Local Surface Reflectance \cite{gao2014efficient}& 3.45 & 2.51 & 0.98 & 7.32 \\
2nd-order GE \cite{GE} & 3.36 & 2.70 & 0.89 & 7.14 \\
1st-order GE \cite{GE}& 3.35 & 2.58 & 0.79 & 7.18 \\
Quasi-U CC \cite{bianco2019quasi} & 3.00 & 2.25 & - & - \\
\cellcolor[HTML]{EFEFEF}Corrected-Moment \cite{MomentCorrection} & 2.95 & 2.05 & 0.59 & 6.89 \\
PCA-based B/W Colors \cite{cheng2014illuminant}& 2.93 & 2.33 & 0.78 & 6.13 \\
Grayness Index \cite{GI}& 2.91 & 1.97 & 0.56 & 6.67 \\

\cellcolor[HTML]{EFEFEF}Color Dog \cite{colorDog} & 2.83 & 1.77 & 0.48 & 7.04 \\
\cellcolor[HTML]{EFEFEF}APAP using GW \cite{afifi2019projective}& 2.40 & 1.76 & 0.55 & 5.42 \\
\cellcolor[HTML]{EFEFEF}CCC \cite{CCC}& 2.38 & 1.69 & 0.45 & 5.85 \\
\cellcolor[HTML]{EFEFEF}Effective Regression Tree \cite{Effective}& 2.36 &  1.59 & 0.49 & 5.54 \\
\cellcolor[HTML]{EFEFEF}Deep Specialized Net \cite{DSNET}& 2.24 & 1.46 & 0.48 & 6.08 \\
\cellcolor[HTML]{EFEFEF}Meta-AWB w 20 tuning images \cite{mcdonagh2018meta}& 2.23 & 1.49 & 0.49 & 5.20 \\
\cellcolor[HTML]{EFEFEF}SqueezeNet-FC4 & 2.23 & 1.57 & 0.47 & 5.15 \\
\cellcolor[HTML]{EFEFEF}AlexNet-FC4 \cite{hu2017fc}& 2.12 & 1.53 & 0.48 & 4.78 \\
\cellcolor[HTML]{EFEFEF}FFCC -- thumb, 2 channels \cite{FFCC} & 2.06 & 1.39 & 0.39 & 4.80 \\
\cellcolor[HTML]{EFEFEF}FFCC -- full, 4 channels \cite{FFCC}& 1.99 & \cellcolor[HTML]{FFFC9E}1.31 & \cellcolor[HTML]{FFFC9E}0.35 & 4.75 \\
\cellcolor[HTML]{EFEFEF}Quasi-unsupervised CC (tuned) \cite{bianco2019quasi} & \cellcolor[HTML]{FFFC9E}1.97 & 1.41 & - & - \\
\hline
 Avg. result for sensor-independent  & 4.26 & 3.25 & 0.99 & 9.43 \\
 Avg. result for sensor-dependent & 2.40 & 1.64 & 0.50 & 5.75\\
  \hdashline
SIIE (Ours) & 2.05 & 1.50 & 0.52 & \cellcolor[HTML]{FFFC9E}4.48

\end{tabular}
}
}
\hfill
\parbox{.48\linewidth}{
\centering
\scalebox{0.5}{
\begin{tabular}{l|cccc}
\textbf{\begin{tabular}[c]{@{}l@{}}\textbf{Gehler-Shi Dataset}\\\textbf{Method} \end{tabular}} & \textbf{Mean} & \textbf{Med.} & \textbf{\begin{tabular}[c]{@{}c@{}}Best \\ 25\%\end{tabular}} & \textbf{\begin{tabular}[c]{@{}c@{}}Worst \\ 25\%\end{tabular}} \\ \hline
White-Patch \cite{maxRGB}& 7.55 & 5.68 & 1.45 & 16.12 \\
Edge-based Gamut \cite{PixelGamut}& 6.52 & 5.04 & 5.43 & 13.58 \\
GW \cite{GW}& 6.36 & 6.28 & 2.33 & 10.58 \\
1st-order GE \cite{GE}& 5.33 & 4.52 & 1.86 & 10.03 \\
2nd-order GE \cite{GE}& 5.13 & 4.44 & 2.11 & 9.26 \\
SoG \cite{SoG}& 4.93 & 4.01 & 1.14 & 10.20 \\
\cellcolor[HTML]{EFEFEF}Bayesian \cite{gehler2008bayesian} & 4.82 & 3.46 & 1.26 & 10.49 \\
\cellcolor[HTML]{EFEFEF}Pixels-based Gamut \cite{PixelGamut}& 4.20 & 2.33 & 0.50 & 10.72 \\
Quasi-U CC \cite{bianco2019quasi} & 3.46 & 2.23 & - & - \\
PCA-based B/W Colors \cite{cheng2014illuminant}& 3.52 & 2.14 & 0.50 & 8.74 \\
\cellcolor[HTML]{EFEFEF}NetColorChecker \cite{BMVC1} & 3.10 & 2.30 & - & - \\
Grayness Index \cite{GI}& 3.07 & 1.87 & 0.43 & 7.62 \\
\cellcolor[HTML]{EFEFEF}Meta-AWB w 20 tuning images \cite{mcdonagh2018meta}& 3.00 & 2.02 & 0.58 & 7.17 \\
\cellcolor[HTML]{EFEFEF}Quasi-unsupervised CC \cite{bianco2019quasi} (tuned) & 2.91 &  1.98 & - & - \\
\cellcolor[HTML]{EFEFEF}Corrected-Moment \cite{MomentCorrection}& 2.86 & 2.04 & 0.70 & 6.34 \\
\cellcolor[HTML]{EFEFEF}APAP using GW \cite{afifi2019projective}& 2.76 & 2.02 & 0.53 & 6.21 \\
\cellcolor[HTML]{EFEFEF}Bianco et al.'s CNN \cite{bianco2015color} & 2.63 & 1.98 & 0.72 & 3.90 \\
\cellcolor[HTML]{EFEFEF}Effective Regression Tree \cite{Effective}& 2.42 & 1.65 & 0.38 & 5.87 \\
\cellcolor[HTML]{EFEFEF}FFCC - thumb, 2 channels \cite{FFCC}& 2.01 & 1.13 & 0.30 & 5.14 \\
\cellcolor[HTML]{EFEFEF}CCC \cite{CCC}& 1.95 & 1.22 & 0.35 & 4.76 \\
\cellcolor[HTML]{EFEFEF}Deep Specialized Net \cite{DSNET}& 1.90 & 1.12 & 0.31 & 4.84 \\
\cellcolor[HTML]{EFEFEF}FFCC - full, 4 channels \cite{FFCC}& 1.78 & \cellcolor[HTML]{FFFC9E}0.96 & \cellcolor[HTML]{FFFC9E}0.29 & 4.62 \\
\cellcolor[HTML]{EFEFEF}AlexNet-FC4 \cite{hu2017fc}& 1.77 & 1.11 & 0.34 & 4.29 \\
\cellcolor[HTML]{EFEFEF}SqueezeNet-FC4 \cite{hu2017fc}& \cellcolor[HTML]{FFFC9E}1.65 & 1.18 & 0.38 & \cellcolor[HTML]{FFFC9E}3.78 \\
\hline
 Avg. result for sensor-independent  & 5.10 & 4.03 & 1.91 & 10.77\\
 Avg. result for sensor-dependent & 2.62 & 1.75  & 0.50 & 5.95\\
 \hdashline
SIIE (Ours) & 2.77 & 1.93 & 0.55 & 6.53
\end{tabular}}
}
\end{table}

Unlike results reported by existing learning methods which use three-fold cross-validation for evaluation, our reported results were obtained by models that were {\it not} trained on any example of the testing camera sensor.

In Tables \ref{siie:Table1}--\ref{siie:Table2}, the mean, median, best 25\%, and the worst 25\% of the angular error between our estimated illuminants and ground truth are reported. The best 25\% and worst 25\% are the mean of the smallest 25\% angular error values and the mean of the highest 25\% angular error
values, respectively. We highlight learning methods (i.e., models trained/tuned for the testing sensor) with gray in the shown tables. It is notable that our SIIE performs better than all statistical-based methods and outperforms some sensor-specific learning methods. We obtain results on par with the state-of-the-art results in the NUS 8-Camera dataset (Table \ref{siie:Table1}). We would like to emphasize that these state-of-the-art results are obtained by \textit{sensor-specific} methods and reported using three-fold cross-validation on images taken by the same sensor.

\begin{table}
\caption[Angular errors on the Cube and Cube+ datasets \cite{banic2017unsupervised}.]{Angular errors on the Cube and Cube+ datasets \cite{banic2017unsupervised}. Methods highlighted in gray are trained/tuned for each camera sensor (i.e., sensor-specific models). The lowest errors are highlighted in yellow. \label{siie:Table2}}
\parbox{.48\linewidth}{
\centering
\scalebox{0.59}
{
\begin{tabular}{l|cccc}
\textbf{\begin{tabular}[c]{@{}l@{}}\textbf{Cube Dataset}\\\textbf{Method} \end{tabular}} & \textbf{Mean} & \textbf{Med.} & \textbf{\begin{tabular}[c]{@{}c@{}}Best \\ 25\%\end{tabular}} & \textbf{\begin{tabular}[c]{@{}c@{}}Worst \\ 25\%\end{tabular}} \\ \hline
White-Patch \cite{maxRGB} & 6.58 & 4.48 & 1.18 & 15.23 \\
GW \cite{GW} & 3.75 & 2.91 & 0.69 & 8.18 \\
SoG \cite{SoG} & 2.58 & 1.79 & 0.38 & 6.19 \\
2nd-order GE \cite{GE}& 2.49 & 1.60 & 0.49 & 6.00 \\
1st-order GE \cite{GE}& 2.45 & 1.58 & 0.48 & 5.89 \\

\cellcolor[HTML]{EFEFEF}APAP using GW \cite{afifi2019projective}& 1.55  & 1.02 & 0.28 &  3.74 \\

\cellcolor[HTML]{EFEFEF}Color Dog \cite{colorDog}&  \cellcolor[HTML]{\bestcolor}1.50 & \cellcolor[HTML]{\bestcolor}0.81 &  \cellcolor[HTML]{\bestcolor}0.27 & \cellcolor[HTML]{\bestcolor}3.86 \\

\cellcolor[HTML]{EFEFEF}Meta-AWB (20) \cite{mcdonagh2018meta}& 1.74 & 1.08 & 0.29 & 4.28 \\
\hline
 Avg. result for sensor-independent  &  3.57 & 2.47 & 0.64 & 8.30\\
 Avg. result for sensor-dependent & 1.54 & 0.92 & 0.26 & 3.85\\
  \hdashline
SIIE (Ours) & 1.98 & 1.36 & 0.40 & 4.64

\end{tabular}
}
}
\hfill
\parbox{.48\linewidth}{
\centering
\scalebox{0.575}
{
\begin{tabular}{l|cccc}
\textbf{\begin{tabular}[c]{@{}l@{}}\textbf{Cube+ Dataset}\\\textbf{Method} \end{tabular}} & \textbf{Mean} & \textbf{Med.} & \textbf{\begin{tabular}[c]{@{}c@{}}Best \\ 25\%\end{tabular}} & \textbf{\begin{tabular}[c]{@{}c@{}}Worst \\ 25\%\end{tabular}} \\ \hline
White-Patch \cite{maxRGB}& 9.69 & 7.48 & 1.72 & 20.49 \\
GW \cite{GW}& 7.71 & 4.29 & 1.01 & 20.19 \\
\cellcolor[HTML]{EFEFEF}Color Dog \cite{colorDog}& 3.32 & 1.19 & 0.22 & 10.22 \\
SoG \cite{SoG}& 2.59 & 1.73 & 0.46 & 6.19 \\
2nd-order GE \cite{GE}& 2.50 & 1.59 & 0.48 & 6.08 \\
1st-order GE \cite{GE}& 2.41 & 1.52 & 0.45 & 5.89 \\
\cellcolor[HTML]{EFEFEF}APAP using GW \cite{afifi2019projective}& 2.01  &  1.36 & 0.38 & 4.71 \\
\cellcolor[HTML]{EFEFEF}Color Beaver \cite{kovsvcevic2019color}& \cellcolor[HTML]{\bestcolor}1.49 & \cellcolor[HTML]{\bestcolor}0.77 & \cellcolor[HTML]{\bestcolor}0.21 & \cellcolor[HTML]{\bestcolor}3.94 \\

\hline
 Avg. result for sensor-independent  & 4.98 & 3.32 & 0.82 & 11.77\\

 Avg. result for sensor-dependent & 2.04 & 1.02 & 0.25 & 5.58\\
  \hdashline
SIIE (Ours) & 2.14 & 1.44 & 0.44 &  5.06
\end{tabular}

}
}
\end{table}

\begin{table}[]
\caption[Our results (angular errors) on each camera of the NUS 8-Camera \cite{cheng2014illuminant}.]{Our results (angular errors) on each camera of the NUS 8-Camera \cite{cheng2014illuminant}. \label{siie:tab:NUS_2}}
\scalebox{0.73}
{
\begin{tabular}{ccccccccc}
\textbf{} & \multicolumn{8}{c}{\textbf{NUS 8-Cameras Dataset}} \\
\multicolumn{1}{c|}{\textbf{Camera}} & \textbf{\begin{tabular}[c]{@{}c@{}}Canon EOS \\ 1Ds MrkIII\end{tabular}} & \textbf{\begin{tabular}[c]{@{}c@{}}Canon EOS \\ 600D\end{tabular}} & \textbf{\begin{tabular}[c]{@{}c@{}}Fujifilm \\ XM1\end{tabular}} & \textbf{\begin{tabular}[c]{@{}c@{}}Nikon \\ D5200\end{tabular}} & \textbf{\begin{tabular}[c]{@{}c@{}}Olympus \\ EPL6\end{tabular}} & \textbf{\begin{tabular}[c]{@{}c@{}}Panasonic\\ GX1\end{tabular}} & \textbf{\begin{tabular}[c]{@{}c@{}}Samsung \\ NX2000\end{tabular}} & \textbf{\begin{tabular}[c]{@{}c@{}}Sony \\ SLT-A57\end{tabular}} \\ \hline
\multicolumn{1}{c|}{\textbf{Mean}} & 2.07 & 1.99 & 2.08 & 2.06 & 2.26 & 1.82 & 1.71 & 2.29 \\
\multicolumn{1}{c|}{\textbf{Median}} & 1.59 & 1.43 & 1.46 & 1.51 & 1.73 & 1.41 & 1.32 & 1.78 \\
\multicolumn{1}{c|}{\textbf{Best 25\%}} & 0.48 & 0.56 & 0.56 & 0.55 & 0.63 & 0.49 & 0.41 & 0.54 \\
\multicolumn{1}{c|}{\textbf{Worst 25\%}} & 4.51 & 4.43 & 4.63 & 4.44 & 4.70 & 3.83 & 3.71 & 5.16
\end{tabular}
}
\end{table}

In Table \ref{siie:tab:NUS_2}, we show our results on each camera of the NUS 8-Camera dataset. We report the mean, median, best 25\%, and the worst 25\% of the angular error between our estimated illuminants and ground truth.

\begin{table}
\caption[Angular and reproduction angular errors \cite{Reproduction} on the Cube+ challenge \cite{challenge}.]{Angular and reproduction angular errors \cite{Reproduction} on the Cube+ challenge \cite{challenge}. The methods are sorted by the median of the errors (shown in bold), as ranked in the challenge \cite{challenge}. Methods highlighted in gray are sensor-specific models. We show our results w/wo training on Cube+ dataset. The lowest errors over all methods are highlighted in yellow. \label{siie:Table3}}
\parbox{.48\linewidth}{
\scalebox{0.54}
{
\begin{tabular}{l|cccc}
\textbf{\begin{tabular}[c]{@{}l@{}}\textbf{Cube+ challenge (angular error)}\\\textbf{Method} \end{tabular}} & \textbf{Mean} & \textbf{Med.} & \textbf{\begin{tabular}[c]{@{}c@{}}Best \\ 25\%\end{tabular}} & \textbf{\begin{tabular}[c]{@{}c@{}}Worst \\ 25\%\end{tabular}} \\ \hline

GW \cite{GW} & 4.44 & \textbf{3.50} & 0.77 & 9.64 \\
1st-order GE \cite{GE} & 3.51 & \textbf{2.3} & 0.56 & 8.53 \\
 V Vuk et al., \cite{challenge}& 6 & \textbf{1.96} & 0.99 & 18.81 \\
\cellcolor[HTML]{EFEFEF} Y Qian et al., (1) \cite{challenge}& 2.48 & \textbf{1.56} & 0.44 & 6.11 \\
\cellcolor[HTML]{EFEFEF} K Chen et al., \cite{challenge}& \cellcolor[HTML]{\bestcolor}1.84 & \cellcolor[HTML]{\bestcolor}\textbf{1.27} & \cellcolor[HTML]{\bestcolor}0.39 & \cellcolor[HTML]{\bestcolor}4.41 \\
\cellcolor[HTML]{EFEFEF} Y Qian et al., (3) \cite{challenge}& 2.27 & \textbf{1.26} & 0.39 & 6.02 \\
\cellcolor[HTML]{EFEFEF} FFCC \cite{FFCC}& 2.1 & \textbf{1.23} & 0.47 & 5.38\\
\cellcolor[HTML]{EFEFEF} A Savchik et al., \cite{challenge}& 2.05 & \textbf{1.2} & 0.41 & 5.24 \\
\hdashline
SIIE (ours) trained wo/ Cube+ & 2.89 & \textbf{1.718} & 0.71 & 7.061 \\
\cellcolor[HTML]{EFEFEF} SIIE (ours) trained w/ Cube+ & 2.1 & \textbf{1.23} & 0.47 & 5.38
\end{tabular}
}
}
\hfill
\parbox{.48\linewidth}{
\scalebox{0.54}
{
\begin{tabular}{l|cccc}
\textbf{\begin{tabular}[c]{@{}l@{}}\textbf{Cube+ challenge (reproduction error)}\\\textbf{Method} \end{tabular}} & \textbf{Mean} & \textbf{Med.} & \textbf{\begin{tabular}[c]{@{}c@{}}Best \\ 25\%\end{tabular}} & \textbf{\begin{tabular}[c]{@{}c@{}}Worst \\ 25\%\end{tabular}} \\ \hline

GW \cite{GW} & 5.74 & \textbf{4.60} & 1.12 & 12.21 \\
1st-order GE \cite{GE} & 4.57 & \textbf{3.22} & 0.84 & 10.75 \\
V Vuk et al., \cite{challenge}& 6.87 & \textbf{2.1} & 1.06 & 21.82 \\
\cellcolor[HTML]{EFEFEF} Y Qian et al., (1) \cite{challenge}& 6.87 & \textbf{2.09} & 0.61 & 8.18 \\
\cellcolor[HTML]{EFEFEF} K Chen et al., \cite{challenge}& \cellcolor[HTML]{\bestcolor}2.49 & \textbf{1.69} & 0.52 & \cellcolor[HTML]{\bestcolor}6.00 \\
\cellcolor[HTML]{EFEFEF} Y Qian et al., (3) \cite{challenge}& 2.93 & \cellcolor[HTML]{\bestcolor}\textbf{1.64} & \cellcolor[HTML]{\bestcolor}0.50 & 7.78 \\
\cellcolor[HTML]{EFEFEF} FFCC \cite{FFCC}& 2.48 & \textbf{1.59} & 0.58 & 7.27 \\
\cellcolor[HTML]{EFEFEF} A Savchik et al., \cite{challenge}& 2.65 & \textbf{1.51} &\cellcolor[HTML]{\bestcolor} 0.50 & 6.85 \\
\hdashline
SIIE (ours) trained wo/ Cube+ & 3.97 & \textbf{2.31} & 0.86 & 10.07\\
\cellcolor[HTML]{EFEFEF} SIIE (ours) trained w/ Cube+ & 2.8 & \textbf{1.54} & 0.58 &  7.27
\end{tabular}
}
}
\end{table}


\begin{table}

\caption[This table shows the angular and reproduction angular errors \cite{Reproduction} obtained on the Cube+ challenge \cite{challenge} using our trained models.]{This table shows the angular and reproduction angular errors \cite{Reproduction} obtained on the Cube+ challenge \cite{challenge} using our trained models. We did not use any example from the Cube+ challenge testing set in the training/validation sets. The reported results in Table \ref{siie:Table3} are highlighted in green.\label{tab:challenge_angular}}
\parbox{.48\linewidth}{
\scalebox{0.55}
{
\begin{tabular}{l|cccc}
\textbf{\begin{tabular}[c]{@{}l@{}}\textbf{Cube+ challenge}\\\textbf{Method} \end{tabular}} & \textbf{Mean} & \textbf{Med.} & \textbf{\begin{tabular}[c]{@{}c@{}}Best \\ 25\%\end{tabular}} & \textbf{\begin{tabular}[c]{@{}c@{}}Worst \\ 25\%\end{tabular}} \\ \hline
\cellcolor[HTML]{B7E589}Trained wo/ Canon EOS 550 D (Cube+) & 2.89 & 1.72 & 0.71 & 7.06\\ 
Trained wo/ Canon 1Ds MkIII (NUS) & 1.98 & 1.22 & 0.43 & 4.89\\
Trained wo/ Canon 600D (NUS) & 1.96 & 1.31 & 0.44 & 4.72 \\ 
Trained wo/ Fujifilm XM1 (NUS) & 2.31 & 1.61 & 0.52 & 5.36\\ 
Trained wo/ Nikon D5200 (NUS) & 1.97 & 1.22 & 0.47 & 4.75 \\ 
Trained wo/ Olympus EPL6 (NUS) & 2.4 & 1.92 & 0.58 & 5.21\\ 
Trained wo/ Panasonic GX1 (NUS) & 2.21 & 1.44 & 0.65 & 5.14\\ 
Trained wo/ Samsung NX2000 (NUS) & 2.02 & 1.38 & 0.38 & 4.92\\ 
\cellcolor[HTML]{B7E589}Trained wo/ Sony SLT-A57 (NUS) & 2.1 & 1.23 & 0.47
& 5.38
\\ 
Trained wo/ Sony Canon 5D (Gehler-Shi) & 2.02 & 1.27 & 0.432 & 4.927 \\ 
\end{tabular}
}
}
\hfill
\parbox{.48\linewidth}{
\scalebox{0.55}
{
\begin{tabular}{l|cccc}
\textbf{\begin{tabular}[c]{@{}l@{}}\textbf{Cube+ challenge}\\\textbf{Method} \end{tabular}} & \textbf{Mean} & \textbf{Med.} & \textbf{\begin{tabular}[c]{@{}c@{}}Best \\ 25\%\end{tabular}} & \textbf{\begin{tabular}[c]{@{}c@{}}Worst \\ 25\%\end{tabular}} \\ \hline
\cellcolor[HTML]{B7E589}Trained wo/ Canon EOS 550 D (Cube+) & 3.97 & 2.31 & 0.86 & 10.07
 \\ 
Trained wo/ Canon 1Ds MkIII (NUS) & 2.65 & 1.59 & 0.54 & 6.54 \\ 
Trained wo/ Canon 600D (NUS) & 2.59 & 1.69 & 0.53 & 6.248 \\
Trained wo/ Fujifilm XM1 (NUS) & 3.08 & 2.19 & 0.67 & 7.1 \\ 
Trained wo/ Nikon D5200 (NUS) & 2.62 & 1.73 & 0.57 & 6.29\\ 
Trained wo/ Olympus EPL6 (NUS) & 3.23 & 2.59 & 0.76 & 6.97\\ 
Trained wo/ Panasonic GX1 (NUS) & 2.89 & 1.86  & 0.74 & 6.86  \\ 
Trained wo/ Samsung NX2000 (NUS) & 2.7& 1.89& 0.48& 6.51\\ 
\cellcolor[HTML]{B7E589}Trained wo/ Sony SLT-A57 (NUS) & 2.8& 1.54& 0.58& 7.27\\ 
Trained wo/ Sony Canon 5D (Gehler-Shi) & 2.69 & 1.68 & 0.54 & 6.59 \\ 
\end{tabular}
}
}
\end{table}

We also examined our trained models on the Cube+ challenge \cite{challenge}. This challenge introduced a new testing set of 363 raw-RGB images captured by Canon EOS 550 D (the same camera model used in the original Cube+ dataset \cite{banic2017unsupervised}). In our results, we did not include any image from the testing set in the training/validation processes. Instead, we used the same models trained for the evaluation on the other datasets (Tables \ref{siie:Table1}--\ref{siie:Table2}). Table \ref{siie:Table3} shows the angular error and reproduction angular errors \cite{Reproduction} obtained by our models and the top-ranked methods that participated in the challenge. Additionally, we show results obtained by other methods \cite{GW, SoG}.

We report results of two trained models using our method. The first one was trained without examples from Cube+ camera sensor (i.e., trained on all camera models in NUS and Gehler-Shi datasets). The second model was originally trained to evaluate our method on one camera of the NUS 8-Cameras dataset (i.e., trained on 7 out of the 8 camera models in NUS 8-Cameras dataset, the Cube+ camera model, and the Gehler-Shi camera models). The latter model is provided to demonstrate the ability of our method to use different camera models beside the target camera model during the training phase.

\begin{figure}
\begin{center}
\includegraphics[width=\linewidth]{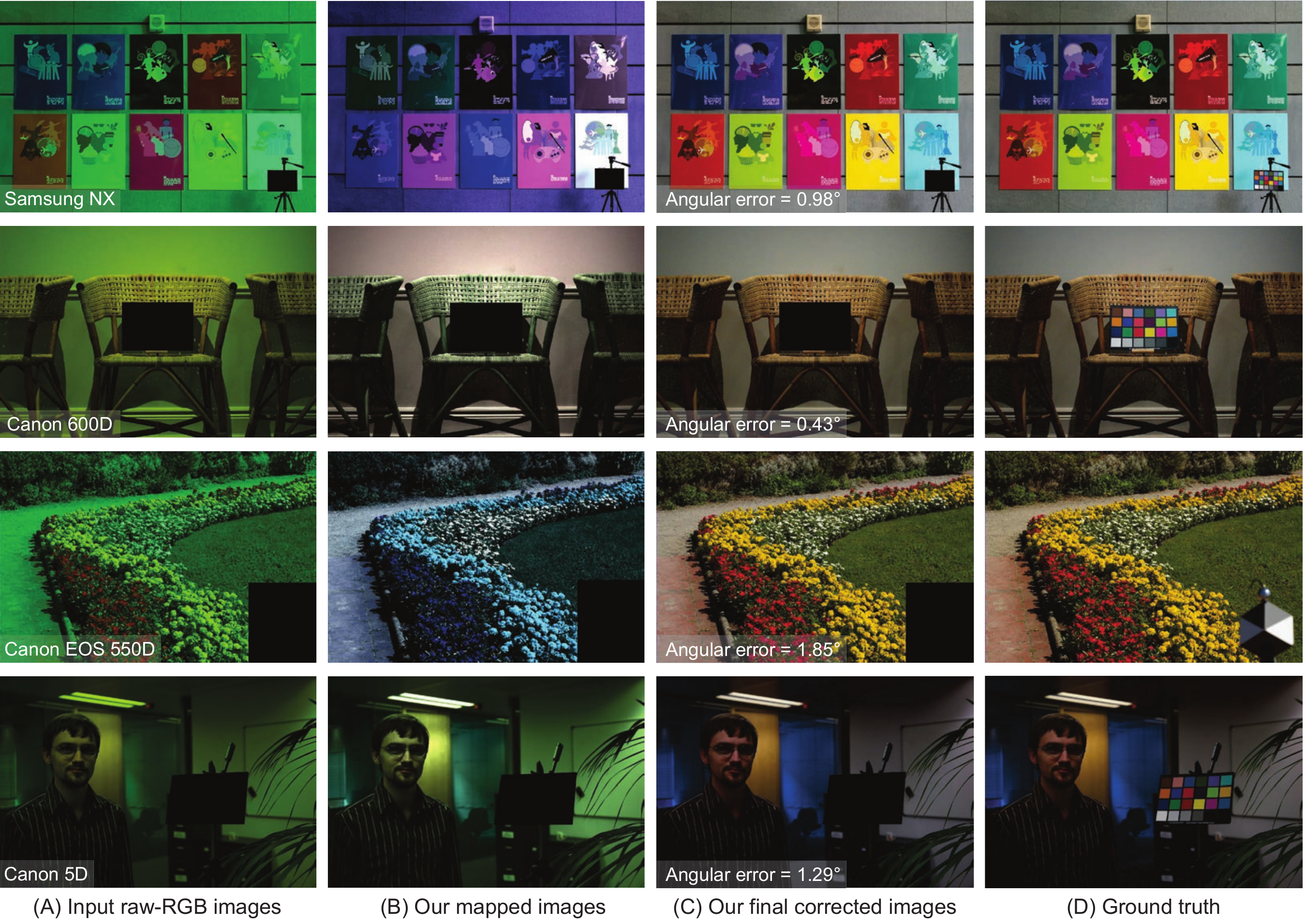}
\end{center}
\vspace{-4mm}
\caption[Qualitative results of our method.]{Qualitative results of our method. (A) Input raw-RGB images. (B) After mapping images in (A) to the learned space. (C) After correcting images in (A) based on our estimated illuminants. (D) Corrected by ground truth illuminants. Shown images are rendered in the sRGB color space by the camera imaging pipeline in \cite{karaimer2016software} to aid visualization.}
\label{siie:fig:results}
\end{figure}

Table \ref{tab:challenge_angular} provides our results obtained on the Cube+ challenge \cite{challenge} using different trained models. The models were originally trained for evaluation on NUS 8-Camera \cite{cheng2014illuminant}, Gehler-Shi \cite{gehler2008bayesian}, and  Cube+ \cite{banic2017unsupervised} datasets using the leave-one-out cross-sensor validation scheme. We did not use any example from the Cube+ challenge testing set in the training/validation phases.

We further tested our trained models on the INTEL-TUT dataset \cite{TUT}, which includes DSLR and mobile phone cameras that are not included in the NUS 8-Camera, Gehler-Shi, and Cube+ datasets. Table \ref{siie:Table4} shows the obtained results by the proposed method trained on DSLR cameras from the NUS 8-Camera, Gehler-Shi, and Cube+ datasets.

Finally, we show qualitative examples in Fig. \ref{siie:fig:results}. For each example, we show the mapped image $\mat{I}_m$ in our learned intermediate space. In the shown figure, we rendered the images in the sRGB color space by the camera imaging pipeline in \cite{karaimer2016software} to aid visualization.

\begin{table}[]
\caption[Angular errors on the INTEL-TUT dataset \cite{TUT}.]{Angular errors on the INTEL-TUT dataset \cite{TUT}. Methods highlighted in gray are trained/tuned for each camera sensor (i.e., sensor-specific models). The lowest errors are highlighted in yellow. \label{siie:Table4}}\vspace{2mm}
\scalebox{0.62}
{
\begin{tabular}{c|cccccc:cc}
\textbf{\begin{tabular}[c]{@{}l@{}}\textbf{INTEL-TUT Dataset} \end{tabular}} & GW \cite{GW}& \begin{tabular}[c]{@{}c@{}} SoG \cite{SoG}\end{tabular}& \begin{tabular}[c]{@{}c@{}}2nd-order\\  GE \cite{GE}\end{tabular} & \begin{tabular}[c]{@{}c@{}}PCA-based \\ B/W Colors \cite{cheng2014illuminant}\end{tabular} & \begin{tabular}[c]{@{}c@{}}1st-order \\ GE \cite{GE}\end{tabular} & \cellcolor[HTML]{EFEFEF}\begin{tabular}[c]{@{}c@{}}
APAP \cite{afifi2019projective}\end{tabular} & \begin{tabular}[c]{@{}c@{}}Ours \\trained on NUS \\ and Cube+\end{tabular} & \begin{tabular}[c]{@{}c@{}}Ours\\ trained on NUS\\  and Gehler-Shi\end{tabular} \\ \hline
\textbf{Mean} & 4.77 & 4.99 & 4.82 & 4.65 & 4.62 & 4.30 &  \cellcolor[HTML]{\bestcolor}3.76 & 3.82 \\
\textbf{Median} & 3.75 & 3.63 & 2.97 & 3.39 & 2.84 & 2.44 &  \cellcolor[HTML]{\bestcolor}2.75 &  2.81 \\
\textbf{Best 25\%} & 0.99 & 1.08 & 1.03 & 0.87 & 0.94 & 0.69 &  \cellcolor[HTML]{\bestcolor}0.81 & 0.87 \\
\textbf{Worst 25\% }& 10.29 & 11.20 & 11.96 & 10.75 & 11.46 & 11.30 & \cellcolor[HTML]{\bestcolor}8.40 & 8.65
\end{tabular}
}
\end{table}

\section{Summary}
\label{siie:sec:conclusion}
We have proposed a deep learning framework for illuminant estimation. Unlike other learning-based methods, our method is a sensor-independent and can be trained on images captured by different camera sensors. To that end, we have introduced an image-specific learnable mapping matrix that maps an input image to a new sensor-independent space. Our method relies only on color distributions of images to estimate scene illuminants. We adopted a compact color histogram that is dynamically generated by our new RGB-$uv$ histogram block. Our method achieves good results on images captured by new camera sensors that have not been used in the training process.

\chapter[Sensor-Independent Convolutional Color Constancy]{Sensor-Independent Convolutional \\Color Constancy \label{ch:ch6}}
In this chapter, we present ``Cross-Camera Convolutional Color Constancy'' (C5)\footnote{Work was done while Mahmoud was an intern at Google; this work was accepted in IEEE International Conference on Computer Vision (ICCV) 2021. A preprint version is available in \cite{afifi2020cross}: Mahmoud Afifi, Jonathan T. Barron, Chloe LeGendre, Yun-Ta Tsai, and Francois Bleibel. Cross-Camera Convolutional Color Constancy. arXiv preprint 2020.}, a learning-based method, trained on images from multiple cameras, that accurately estimates a scene's illuminant color from raw images captured by a new camera previously unseen during training.
C5 is a hypernetwork-like extension of the convolutional color constancy (CCC) approach: C5 learns to generate the weights of a CCC model that is then evaluated on the input image, with the CCC weights dynamically adapted to different input content.
Unlike prior cross-camera color constancy models, which are usually designed to be agnostic to the spectral properties of test-set images from unobserved cameras, C5 approaches this problem through the lens of transductive inference: additional unlabeled images are provided as input to the model at test time, which allows the model to calibrate itself to the spectral properties of the test-set camera during inference.
C5 achieves state-of-the-art accuracy for cross-camera color constancy on several datasets, is fast to evaluate ($\sim$7 and $\sim$90 ms per image on a GPU or CPU, respectively), and requires little memory ($\sim$2 MB), and, thus, is a practical solution to the problem of calibration-free automatic white balance for mobile photography. The source code of this work is available on GitHub: \href{https://github.com/mahmoudnafifi/C5}{https://github.com/mahmoudnafifi/C5}.

\section{Introduction} \label{C5:sec.intro}

\begin{figure}
\includegraphics[width=\linewidth]{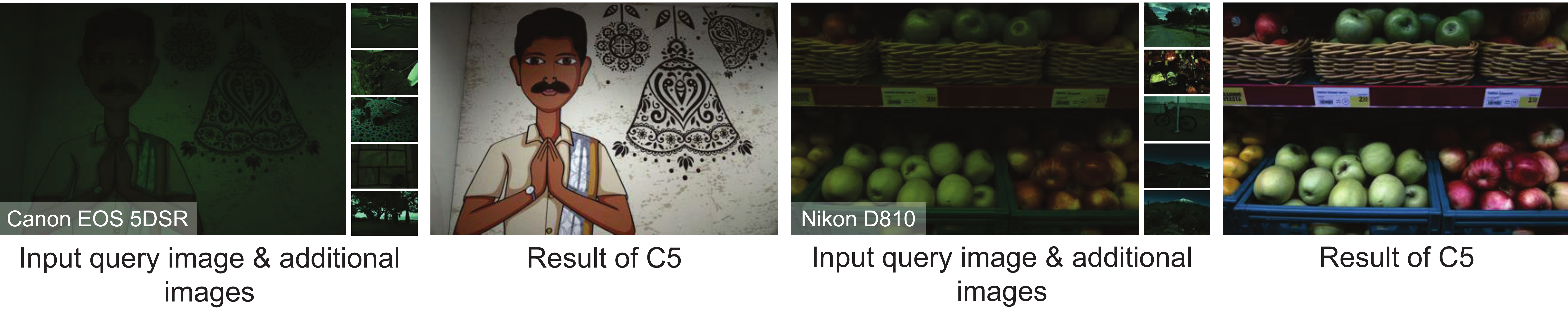}
\vspace{-7mm}
\caption[Our C5 model exploits the colors of additional images captured by the new camera model to generate a specific color constancy model for the input image.]{Our C5 model exploits the colors of additional images captured by the new camera model to generate a specific color constancy model for the input image. The shown images were captured by \textit{unseen} DSLR and smartphone camera models \cite{laakom2019intel} that were not included in the training stage. \label{C5:teaser}}
\end{figure}

As discussed in Chapter \ref{ch:ch3}, one simple heuristic applied to the color constancy problem is the GW assumption: that colors in the world tend to be neutral gray and that the color of the illuminant can, therefore, be estimated as the average color of the input image~\cite{GW}. This GW method and its related techniques have the convenient property that they are \emph{invariant} to much of the spectral sensitivity differences among camera sensors and, therefore, very well-suited to the cross-camera task. If camera A's red channel is twice as sensitive as camera B's red channel, then a scene captured by camera A will have an average red intensity that is twice that of the scene captured by camera B, and so GW will produce identical output images (though this assumes that the spectral response of A and B are identical up to a scale factor, which is rarely the case in practice).

However, current state-of-the-art learning-based methods for color constancy rarely exhibit this property, because they often learn things like the precise distribution of likely illuminant colors (a consequence of black-body illumination and other scene lighting regularities) and are, therefore, sensitive to any mismatch between the spectral sensitivity of the camera used during training and that of the camera used at test time.

\begin{figure}[t]
\begin{center}
\includegraphics[width=0.94\linewidth]{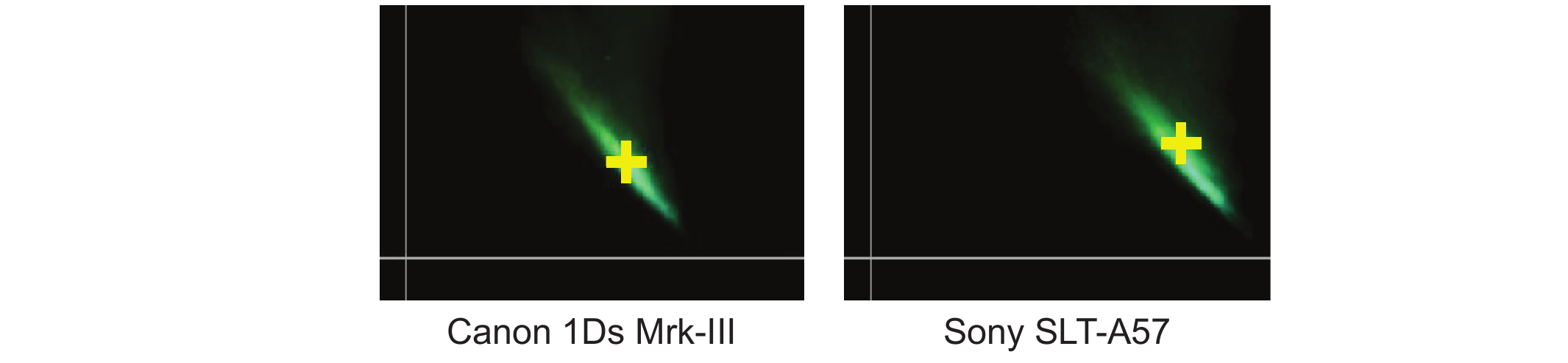}
\end{center}
\vspace{-7mm}
\caption[A visualization of $uv$ log-chroma histograms of images from two different cameras averaged over many images, as well as the $uv$ coordinate of the mean of ground-truth illuminants over the entire scene set.]{
A visualization of $uv$ log-chroma histograms ($u = \log(g / r)$, $v = \log(g / b)$) of images from two different cameras averaged over many images (green), as well as the $uv$ coordinate of the mean of ground-truth illuminants over the entire scene set (yellow)~\cite{cheng2014illuminant}.
The ``positions'' of these histograms change significantly across the two camera sensors because of their different spectral sensitivities, which is why many color constancy models generalize poorly across cameras.
\label{C5:fig:idea}}
\end{figure}

As discussed in Chapter \ref{ch:ch5}, there is often significant spectral variation across camera models (as shown in Fig.~\ref{C5:fig:idea}), this sensitivity of existing methods is problematic when designing practical white-balance solutions. Training a learning-based algorithm for a new camera requires collecting hundreds, or thousands, of images with ground-truth illuminant color labels (in practice: images containing a color chart), a burdensome task for a camera manufacturer or platform that may need to support hundreds of different camera models. However, the GW assumption still holds surprisingly well across sensors---if given several images from a particular camera, one can do a reasonable job of estimating the range of likely illuminant colors (as can also be seen in Fig.~\ref{C5:fig:idea}).

\paragraph{Contribution}~This chapter presents C5, a cross-camera convolutional color constancy model. Our model addresses this problem of high-accuracy cross-camera color constancy through the use of two concepts. First, our system is constructed to take as input not just a single test-set image, but also a small set of additional images from the test set, which are arbitrarily-selected, unlabeled, and not white balanced. This allows the model to calibrate itself to the spectral properties of the test-time camera during inference. We make no assumptions about these additional images except that they come from the same camera as the ``target'' test set image and they contain some content (not all black or white images). In practice, these images could simply be randomly chosen images from the photographer's ``camera roll'', or they could be a fixed set of ad hoc images of natural scenes taken once by the camera manufacturer---because these images do not need to be annotated, they are abundantly available. Second, our system is constructed as a \emph{hypernetwork}~\cite{ha2016hypernetworks} around an existing color constancy model. The target image and the additional images are used as input to a deep neural network whose output is the weights of a smaller color constancy model, and those generated weights are then used to estimate the illuminant color of the target image. Our approach is also closely related to the work on domain adaptation~\cite{daume2006domain, saenko2010adapting} and transfer learning~\cite{pan2009survey}, both of which attempt to enable learning-based models to cope with differences between training and test data. Our system is trained using labeled (and unlabeled) images from multiple cameras, but at test time our model is able to look at a set of (unlabeled) test set images from a new camera. Our hypernetwork is able to infer the likely spectral properties of the new camera that produced the test set images (much as the reader can infer the likely illuminant colors of a camera from only looking at aggregate statistics, as in Fig.~\ref{C5:fig:idea}) and produce a small model that has been dynamically adapted to produce accurate illuminant estimates when applied to the target image. By learning the weights of an FFCC-like model using other test-set images, C5 is able to dynamically adapt to the domain of unseen camera models, thereby allowing a single learned color constancy model to be applied to a wide variety of disparate datasets (see Fig.~\ref{C5:teaser}). By leveraging the fast convolutional approach already in-use by FFCC, C5 is able to retain the computational efficiency and low memory footprint of FFCC, while achieving state-of-the-art results compared to other camera-independent color constancy methods.

\section{Methodology} \label{C5:sec.method}

We call our system ``cross-camera convolutional color constancy'' (C5), because it builds upon the existing ``convolutional color constancy'' (CCC) model~\cite{CCC} and its successor ``fast Fourier color constancy'' (FFCC)~\cite{FFCC}, but embeds them in a multi-input hypernetwork to enable accurate cross-camera performance. These CCC/FFCC models work by learning to perform localization within a log-chroma histogram space, such as those shown in Fig.~\ref{C5:fig:idea}. Here, we present a convolutional CC model that is a simplification of those presented in the original work \cite{CCC} and its FFCC follow-up~\cite{FFCC}. This simple convolutional model will be a fundamental building block that we will use in our larger neural network.

The image formation model behind CCC/FFCC (and most CC models) is that each pixel of the observed image is assumed to be the element-wise product of some ``true'' white-balanced image (or equivalently, the observed image if it were imaged under a white illuminant) and some illuminant color:
\begin{equation}
\forall_k\,\mat{c}^{(k)} = \mat{w}^{(k)} \circ \mat{\light},
\end{equation}
where $\mat{c}^{(k)}$ is the observed color of pixel $k$, $\mat{w}^{(k)}$ is the true color of the pixel, and $\mat{\light}$ is the color of the illuminant, all of which are 3-vectors of RGB values.
CC algorithms traditionally use the input image $\{\mat{c}^{(k)}\}$
to produce an estimate of the illuminant $\hat{\mat{\light}}$ that is then divided (element-wise) into each observed color to produce an estimate of the true color of each pixel $\{ \hat{\mat{w}}^{(k)} \}$.

CCC defines two log-chroma measures for each pixel, which are simply the log of the ratio of two color channels:
\begin{equation}
u^{(k)} = \log\big( c^{(k)}_g / c^{(k)}_r\big), \quad v^{(k)} = \log\big(c^{(k)}_g / c^{(k)}_b\big)\,.
\end{equation}

As discussed in Chapter \ref{ch:ch3} Finlayson and Hordley showed that this log-chrominance representation of color means that illuminant changes (i.e.\, element-wise scaling by $\mat\light$) can be modeled simply as additive offsets to this $uv$ representation~\cite{finlayson2001color}.
We then construct a 2D histogram of the log-chroma values of all pixels:
\begin{equation}
\mat{N}_0(u,v) = \displaystyle \sum_k || \mat{c}^{(k)} ||_2 \left[ \left|u^{(k)} - u \right| \leq \epsilon \, \wedge\,  \left| v^{(k)} - v \right| \leq \epsilon \right]\,. \label{C5:eq:hist}
\end{equation}

This is simply a histogram over all $uv$ coordinates of size ($64 \times 64)$ written out using Iverson brackets, where $\epsilon$ is the width of a histogram bin, and where each pixel is weighted by its overall brightness under the assumption that bright pixels provide more actionable signal than dark pixels. We use a histogram bin width $\epsilon = (b_{\texttt{max}} - b_{\texttt{min}})/n$, where $b_{\texttt{max}}$ and $b_{\texttt{min}}$ are the histogram boundary values. In our experiments, we set $b_{\texttt{min}}$ and $b_{\texttt{max}}$ to -2.85 and 2.85, respectively.
As was done in FFCC, we construct two histograms: one of pixel intensities $\mat{N}_0$, and one of gradient intensities $\mat{N}_1$ (constructed analogously to Equation~\ref{C5:eq:hist}).

These histograms of log-chroma values exhibit a useful property: element-wise multiplication of the RGB values of an image by a constant results in a \emph{translation} of the resulting log-chrominance histograms. The core insight of CCC is that this property allows CC to be framed as the problem of ``localizing'' a log-chroma histogram in this $uv$ histogram-space~\cite{CCC}---because every $uv$ location in $N$ corresponds to a (normalized) illuminant color $\mat{\light}$, the problem of estimating $\mat{\light}$ is \emph{reducible} (in a computability sense) to the problem of estimating a $uv$ coordinate. This can be done by discriminatively training a ``sliding window'' classifier much as one might train, say, a face-detection system: the histogram is convolved with a (learned) filter and the location of the argmax is extracted from the filter response, and that argmax corresponds to $uv$ value that is (the inverse of) an estimated illumination location. 

We adopt a simplification of the convolutional structure used by FFCC~\cite{FFCC}:
\begin{equation}
\mat{P} = \operatorname{softmax}\bigg(\mat{B} + \sum_i\big(\mat{N}_i * \mat{F}_i\big)\bigg), \label{C5:eq:CCC}
\end{equation}
where $\{ \mat{F}_i \}$ and $\mat{B}$ are filters and a bias, respectively, which have the same shape as $\mat{N}_i$ (unlike FFCC, we do not include a ``gain'' multiplier, as it did not result in a uniformly improved performance). Each histogram $\mat{N}_i$ is convolved with each filter $\mat{F}_i$ and summed across channels (a ``conv'' layer) and $\mat{B}$ is added to that summation, which collectively biases inference towards $uv$ coordinates that correspond to common illuminants, such as black body radiation.
As was done in FFCC, this convolution is accelerated through the use of FFTs, though, unlike FFCC, we use a non-wrapped histogram and, thus, non-wrapped filters and bias. This sacrifices speed for simplicity and accuracy and avoids the need for the complicated ``de-aliasing'' scheme used by FFCC which is not compatible with the convolutional neural network structure that we will later introduce.

The output of the softmax $\mat{P}$ is effectively a ``heat map'' of what illuminants are likely, given the distribution of pixel and gradient intensities reflected in $\mat{N}$ and in the prior $\mat{B}$, from which, we extract a ``soft argmax'' by taking the expectation of $u$ and $v$ with respect to $\mat{P}$:
\begin{equation}
    \hat{\light}_u = \sum_{u,v} u \mat{P}(u,v)\,,\quad
    \hat{\light}_v = \sum_{u,v} v \mat{P}(u,v). \label{C5:eq:argmax}
\end{equation}

Equation\ \ref{C5:eq:argmax} is equivalent to estimating the mean of a fitted Gaussian, in the $uv$ space, weighted by $\mat{P}$. Because the absolute scale of $\mat{\light}$ is assumed to be irrelevant or unrecoverable in the context of CC, after estimating $(\hat{\light}_u, \hat{\light}_v)$, we produce an RGB illuminant estimate $\hat{\mat{\light}}$ that is simply the unit vector whose log-chroma values match
\mbox{our estimate}:
\begin{gather}
\hat{\mat{\light}} = \left( \exp\left(-\hat{\light}_u\right) / z, \, 1 / z, \, \exp\left(-\hat{\light}_v\right) / z \right), \\
z = \sqrt{\exp\left(-\hat{\light}_u\right)^2 + \exp\left(-\hat{\light}_v\right)^2 + 1}. \label{C5:eq:light_rgb}
\end{gather}

A convolutional color constancy model is then trained by setting $\{ \mat{F}_i \}$ and $\mat{B}$ to be free parameters which are then optimized to minimize the difference between the predicted illuminant $\hat{\mat{\light}}$ and the ground-truth illuminant $\mat{\light}^*$.

\subsection{Architecture} \label{C5:subsec:architecture}

\begin{figure}[!t]
\includegraphics[width=\linewidth]{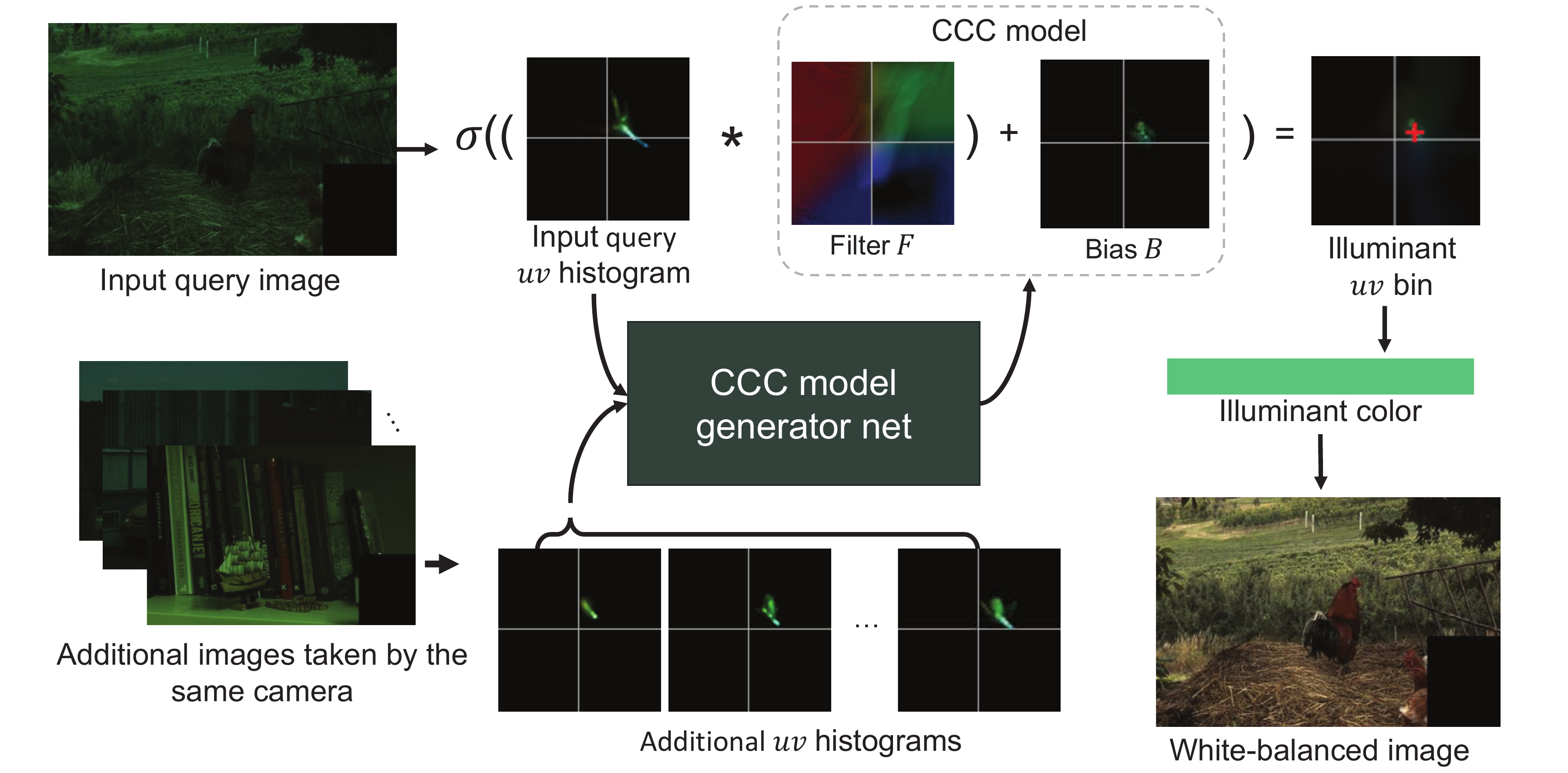}
\vspace{-7mm}
\caption[An overview of our C5 model.]{An overview of our C5 model. The $uv$ histograms for the input query image and a variable number of additional input images taken from the same sensor as the query are used as input to our neural network, which generates a filter bank $\{\mat{F}_i\}$ (here shown as one filter) and a bias $B$, which are the parameters of a conventional CCC model~\cite{CCC}.
The query $uv$ histogram is then convolved by the generated filter and shifted by the generated bias to produce a heat map, whose argmax is the estimated illuminant~\cite{CCC}.
\label{C5:fig:main}}
\end{figure}

With our baseline CCC/FFCC-like model in place, we can now construct our cross-camera convolutional color constancy model (C5), which is a deep architecture in which CCC is a component. Both CCC and FFCC operate by learning a single fixed set of parameters consisting of a single filter bank $\{ \mat{F}_i \}$ and bias $\mat{B}$. In contrast, in C5 the filters and bias are parameterized as the output of a deep neural network (parameterized by weights $\modelweights$) that takes as input not just log-chrominance histograms for the image being color-corrected (which we will refer to as the ``query'' image), but also log-chrominance histograms from several other randomly selected input images (but with no ground-truth illuminant labels) from the test set. 
By using a generated filter and bias from additional images taken from the query image's camera (instead of using a fixed filter and bias as was done in previous work) our model is able to automatically ``calibrate'' its CCC model to the specific sensor properties of the query image. This can be thought of as a hypernetwork~\cite{ha2016hypernetworks}, wherein a deep neural network emits the ``weights'' of a CCC model, which is itself a shallow neural network. This approach also bears some similarity to a Transformer approach, as a CCC model can be thought of as ``attending'' to certain parts of a log-chroma histogram, and so our neural network can be viewed as a sort of self-attention mechanism~\cite{vaswani2017attention}.
See Fig.~\ref{C5:fig:main} for a visualization of this data flow.

\begin{figure}[!t]
\includegraphics[width=\linewidth]{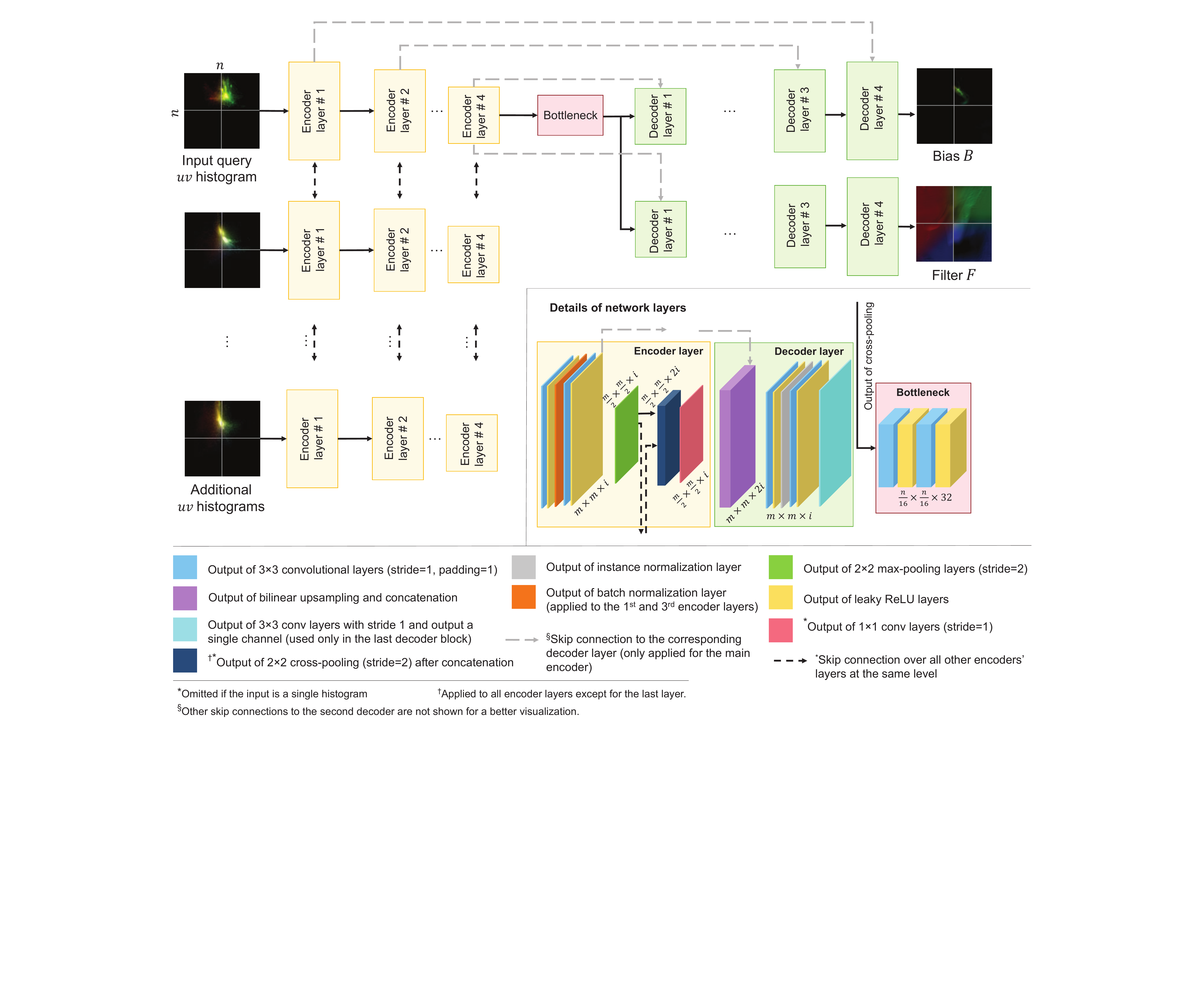}
\vspace{-7mm}
\caption[An overview of neural network architecture that emits CCC model weights.]{An overview of neural network architecture that emits CCC model weights. The $uv$ histogram of the query image along with additional input histograms taken from the same camera are provided as input to a set of multiple encoders. The activations of each encoder are shared with the other encoders by performing max-pooling across encoders after each block. The cross-pooled features at the last encoder layer are then fed into two decoder blocks to generate a bias and filter bank of an CCC model for the query histogram. Each scale of the decoder is connected to the corresponding scale of the encoder for query histogram with skip connections. \label{C5:fig:architecture}}
\end{figure}

At the core of our model is the deep neural network that takes as input a set of log-chroma histograms and must produce as output a CCC filter bank and bias map.  For this we use a multi-encoder-multi-decoder U-Net-like architecture \cite{ronneberger2015u}. The first encoder is dedicated to the ``query'' input image's histogram, while the rest of the encoders take as input the histograms corresponding to the additional input images. To allow the network to reason about the set of additional input images in a way that is insensitive to their ordering, we adopt the permutation invariant pooling approach of Aittala et al. \cite{aittala2018burst}: we use max pooling \emph{across} the set of activations of each branch of the encoder. This ``cross-pooling'' gives us a single set of activations that are reflective of the set of additional input images, but are agnostic to the particular ordering of those input images. At inference time, these additional images are needed to allow the network to reason about how to use them in challenging cases. Due to the ``cross-pooling," the activations of the encoders for the additional images depend on the query image, and so cannot be pre-computed for a given sensor.
The cross-pooled features of the last layer of all encoders are then fed into two decoder blocks. Each decoder produces one component of our CCC model: a bias map $B$ and two filters, $\{ \mat{F}_0, \mat{F}_1 \}$ (which correspond to pixel and edge histograms $\{ \mat{N}_0, \mat{N}_1 \}$, respectively). 
As per the traditional U-Net structure, we use skip connections between each level of the decoder and its corresponding level of the encoder with the same spatial resolution, but only for the encoder branch corresponding to the query input image's histogram.
Each block of our encoder consists of a set of interleaved $3\!\times\!3$ conv layers, leaky ReLU activation, batch normalization, and $2\!\times\!2$ max pooling, and each block of our decoder consists of $2\times$ bilinear upsampling followed by interleaved $3\!\times\!3$ conv layers, leaky ReLU activation, and instance normalization.
When passing our 2-channel (pixel and gradient) log-chroma histograms to our network, we augment each histogram with two extra ``channels'' comprising of only the $u$ and $v$ coordinates of each histogram, as in CoordConv~\cite{liu2018intriguing}. This augmentation allows a convolutional architecture on top of log-chroma histograms to reason about the absolute ``spatial'' information associated with each $uv$ coordinate, thereby allowing a convolutional model to be aware of the absolute color of each histogram bin.
See Figure~\ref{C5:fig:architecture} for a detailed visualization of our architecture.

\subsection{Training} \label{C5:training}

Our model is trained by minimizing the angular error \cite{hordley2004re} between the predicted unit-norm illuminant color $\hat{\mat{\light}}$ and the ground-truth illuminant color $\mat{\light}^*$, as well as an additional loss that regularizes the CCC models emitted by our network. Our loss function $\lossfun(\cdot)$ is:
\begin{equation}\label{C5:loss:Eq.1}
\lossfun\left(\mat{\light}^*,  \hat{\mat{\light}} \right) = \cos^{-1}\left(\frac{\mat{\light}^* \cdot \hat{\mat{\light}}}{\lVert\mat{\light}^*\rVert}\right) + S\left(\{ \mat{F}_i(\modelweights) \}, \mat{B}(\modelweights)\right)\,,
\end{equation}
where $S(\cdot)$ is a regularizer that encourage the network to generate smooth filters and biases, which reduces over-fitting and improves generalization:
\begin{align}
S\left(\{ \mat{F}_i \}, \mat{B}\right) = \lambda_{B} (&\lVert \mat{B} \ast \sobel_u \rVert^2 + \lVert \mat{B} \ast \sobel_v \rVert^2 )  \nonumber \\
+ \lambda_{F} \sum_i (&\lVert \mat{F}_i \ast \sobel_u \rVert^2 + \lVert \mat{F}_i \ast \sobel_v \rVert^2 ) \,, \label{C5:loss:Eq.2}
\end{align}
\noindent where $\sobel_u$ and $\sobel_v$ are $3\!\times\!3$ horizontal and vertical Sobel filters, respectively, and $\lambda_{F}$ and $\lambda_{B}$ are multipliers that control the strength of the smoothness for the filters and the bias, respectively. This regularization is similar to the total variation smoothness prior used by FFCC~\cite{FFCC}, though here we are imposing it on the filters and bias generated by a neural network, rather than on a single filter bank and bias map. We set the multiplier hyperparameters $\lambda_{F}$ and $\lambda_{B}$ to 0.15 and 0.02, respectively (see Sec. \ref{C5:sec:ablation} for ablation studies). 
In addition to regularizing the CCC model emitted by our network, we additionally regularize the weights of our network themselves, $\modelweights$, using L2 regularization (i.e., ``weight decay'') with a multiplier of $5\!\times\!10^{-4}$. This regularization of our network serves a different purpose than the regularization of the CCC models emitted by our network---regularizing $\{ \mat{F}_i(\modelweights) \}$ and $\mat{B}(\modelweights)$ prevents over-fitting by the CCC model emitted by our network, while regularizing $\modelweights$ prevents over-fitting by the model \emph{generating} those CCC models.

Training is performed using the Adam optimizer \cite{kingma2014adam}
with hyperparameters $\beta_1=0.9$, $\beta_2=0.999$, for 60 epochs. We use a learning rate of $5\!\times\!10^{-4}$ with a cosine annealing schedule \cite{loshchilov2016sgdr} and increasing batch-size (from 16 to 64) \cite{smith2017don, masters2018revisiting} which improve the stability of training. When training our model for a particular camera model, at each iteration we randomly select a batch of training images (and their corresponding ground-truth illuminants) for use as query input images, and then randomly select $m$ additional input images for each query image from the training set for use as additional input images.

\begin{figure}[!t]
\includegraphics[width=\linewidth]{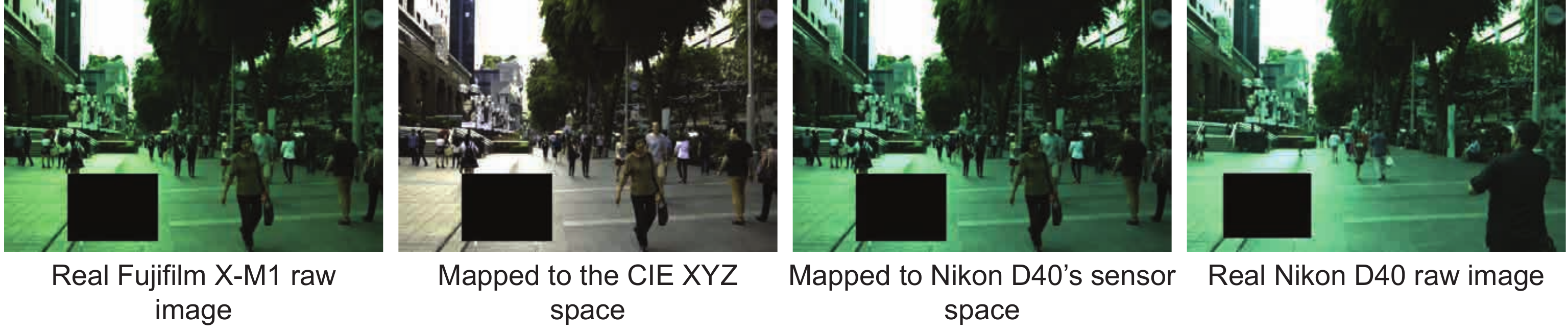}
\vspace{-7mm}
\caption[An example of the image mapping used to augment training data.]{An example of the image mapping used to augment training data. From left to right: a raw image captured by a Fujifilm X-M1 camera; the same image after white-balancing in CIE XYZ; the same image mapped into the Nikon D40 sensor space; and a real image captured by a Nikon D40 of the same scene for comparison \cite{cheng2014illuminant}.\label{C5:fig:sensor_mapping_example}}
\end{figure}

\section{Experiments and Discussion} \label{C5:results}

Similar to our work in Chapter \ref{ch:ch5}, we used downsized raw images after applying the black-level normalization and masking out the calibration object to avoid any ``leakage'' during the evaluation. Excluding histogram computation time (which is difficult to profile accurately due to the expensive nature of scatter-type operations in deep learning frameworks), our method runs in $\sim$7 milliseconds per image on a NVIDIA GeForce GTX 1080, and $\sim$90 milliseconds on an Intel Xeon CPU Processor E5-1607 v4 (10M Cache, 3.10 GHz). Because our model exists in log-chroma histogram space, the uncompressed size of our entire model is $\sim$2 MB, small enough to easily fit within the narrow constraints of limited compute environments such as mobile phones.

\subsection{Data Augmentation}\label{C5:subsec:dataaug}
Many of the datasets we use contain only a few images per distinct camera model (e.g. the NUS dataset \cite{cheng2014illuminant}) and this poses a problem for our approach as neural networks generally require significant amounts of training data. To address this, we use a data augmentation procedure in which images taken from a ``source'' camera model are mapped into the color space of a ``target'' camera.
To perform this mapping, we first white balance each raw source image using its ground-truth illuminant color, and then transform that white-balanced raw image into the device-independent CIE XYZ color space \cite{cie1932commission} using the CST matrices provided in each DNG file \cite{DNG}. Then, we transform the CIE XYZ image into the target sensor space by inverting the CST of an image taken from the target camera dataset. Instead of randomly selecting an image from the target dataset, we use the correlated color temperature of each image and the capture exposure setting to match source and target images that were captured under roughly the same conditions. This means that ``daytime'' source images get warped into the color space of ``daytime'' target images, etc., and this significantly increases the realism of our synthesized data.

\begin{table}[t]
\caption[Results of ablation studies.]{Results of ablation studies. The shown results were obtained by training our network on the NUS \cite{cheng2014illuminant} and the Gehler-Shi datasets \cite{gehler2008bayesian} with augmentation, and testing on the Cube+ dataset \cite{banic2017unsupervised}. In this set of experiments, we used seven encoders (i.e., six additional histograms). Note that none of the training data includes any scene/sensor from the Cube+ dataset \cite{banic2017unsupervised}. For each set of experiments, we highlight the lowest errors in yellow.\label{C5:table:ablation}}
\centering
\scalebox{0.65}{

\begin{tabular}{lccccc}
& \textbf{Mean} & \textbf{Med.} & \textbf{B. 25\%} & \textbf{W. 25\%} & \textbf{Tri.} \\ \cline{2-6} 
& \multicolumn{5}{c}{\cellcolor[HTML]{\headercolor}Histogram bin size, $n$} \\ \hline
\multicolumn{1}{l|}{$n=16$} &  2.28$\pm$0.01 & 1.81$\pm$0.03 & 0.65$\pm$0.01 & 4.72$\pm$0.02 & 1.91$\pm$0.02\\ \hline
\multicolumn{1}{l|}{$n=32$} &  2.02$\pm$0.01 & 1.44$\pm$0.01 & 0.44$\pm$0.01 & 4.66$\pm$0.01 & 1.86$\pm$0.03  \\ \hline
\multicolumn{1}{l|}{$n=64$} &  \cellcolor[HTML]{\bestcolor}1.87$\pm$0.00 & \cellcolor[HTML]{\bestcolor}1.27$\pm$0.01 & 0.41$\pm$0.01 & \cellcolor[HTML]{\bestcolor}4.36$\pm$0.01 & \cellcolor[HTML]{\bestcolor}1.40$\pm$0.01\\ \hline
\multicolumn{1}{l|}{$n=128$} & 2.03$\pm$0.00 & 1.42$\pm$0.01 & \cellcolor[HTML]{\bestcolor}0.40$\pm$0.00 & 4.70$\pm$0.01 & 1.54$\pm$0.01 \\ \hline

& \multicolumn{5}{c}{\cellcolor[HTML]{\headercolor}Smoothness factors, $\lambda_{B}$ and $\lambda_{F}$ ($n=64$)} \\ \hline
\multicolumn{1}{l|}{$\lambda_B=0, \lambda_F = 0$} & 2.07$\pm$0.01 & 1.42$\pm$0.01 & 0.47$\pm$0.01 & 4.67$\pm$0.01 & 1.57$\pm$0.01  \\ \hline

\multicolumn{1}{l|}{$\lambda_B=0.005, \lambda_F = 0.035$} & 1.95$\pm$0.00 & 1.31$\pm$0.01 & \cellcolor[HTML]{\bestcolor}0.40$\pm$0.00 & 4.57$\pm$0.01 & 1.47$\pm$0.01\\ \hline

\multicolumn{1}{l|}{$\lambda_B=0.02, \lambda_F = 0.15$} &  \cellcolor[HTML]{\bestcolor}1.87$\pm$0.00 & \cellcolor[HTML]{\bestcolor}1.27$\pm$0.01 & 0.41$\pm$0.01 & \cellcolor[HTML]{\bestcolor}4.36$\pm$0.01 & \cellcolor[HTML]{\bestcolor}1.40$\pm$0.01 \\ \hline
\multicolumn{1}{l|}{$\lambda_B=0.10, \lambda_F = 0.75$} & 2.11$\pm$0.00 & 1.55$\pm$0.01 & 0.48$\pm$0.00 & 4.70$\pm$0.01 & 1.66$\pm$0.01 \\ \hline

\multicolumn{1}{l|}{$\lambda_B=0.25, \lambda_F = 1.85$} & 2.23$\pm$0.00 & 1.61$\pm$0.01 & 0.53$\pm$0.00 & 5.04$\pm$0.01 & 1.77$\pm$ 0.01  \\ \hline

& \multicolumn{5}{c}{\cellcolor[HTML]{\headercolor}Increasing batch size ($n=64$)} \\ \hline

\multicolumn{1}{l|}{w/o increasing} & 1.93$\pm$0.00 & 1.29$\pm$0.01 & 0.42$\pm$0.00 & 4.52$\pm$0.02 & 1.43$\pm$0.01 \\ \hline
\multicolumn{1}{l|}{w/ increasing} &  \cellcolor[HTML]{\bestcolor}1.87$\pm$0.00 & \cellcolor[HTML]{\bestcolor}1.27$\pm$0.01 & \cellcolor[HTML]{\bestcolor}0.41$\pm$0.01 & \cellcolor[HTML]{\bestcolor}4.36$\pm$0.01 &  \cellcolor[HTML]{\bestcolor} 1.40$\pm$0.01  \\ \hline

& \multicolumn{5}{c}{\cellcolor[HTML]{\headercolor}Gradient histogram and $uv$ channels ($n=64$)} \\ \hline

\multicolumn{1}{l|}{w/o gradient histogram} & 2.30$\pm$0.01 & 1.53$\pm$0.01 & 0.45$\pm$0.01 & 5.51$\pm$0.02 & 1.71$\pm$0.02 \\ \hline
\multicolumn{1}{l|}{w/o $uv$} & 2.03$\pm$0.01 & 1.45$\pm$0.01 & 0.44$\pm$0.01 & 4.63$\pm$0.02 & 1.56$\pm$0.01  \\ \hline
\multicolumn{1}{l|}{w/ $uv$ and gradient histogram} &  \cellcolor[HTML]{\bestcolor}1.87$\pm$0.00 & \cellcolor[HTML]{\bestcolor}1.27$\pm$0.01 & \cellcolor[HTML]{\bestcolor}0.41$\pm$0.01 & \cellcolor[HTML]{\bestcolor}4.36$\pm$0.01 & \cellcolor[HTML]{\bestcolor}1.40$\pm$0.01  \\

\end{tabular}}
\end{table}

After mapping the source image to the target white-balanced sensor space, we randomly sample from a cubic curve that has been fit to the $rg$ chromaticity of illuminant colors in the target sensor. Lastly, we apply a chromatic adaptation to generate the augmented image in the target sensor space. This chromatic adaptation is performed by multiplying each color channel of the white-balanced raw image, mapped to the target sensor space, with the corresponding sampled illuminant color channel value; see Figure \ref{C5:fig:sensor_mapping_example} for an example. Additional details can be found in Appendix \ref{ch:appendix1}. This augmentation allows us to generate additional training examples to improve the generalization of our model. More details are provided in Sec.\ \ref{C5:sbusec:results}.

\begin{figure}[!t]
\includegraphics[width=\linewidth]{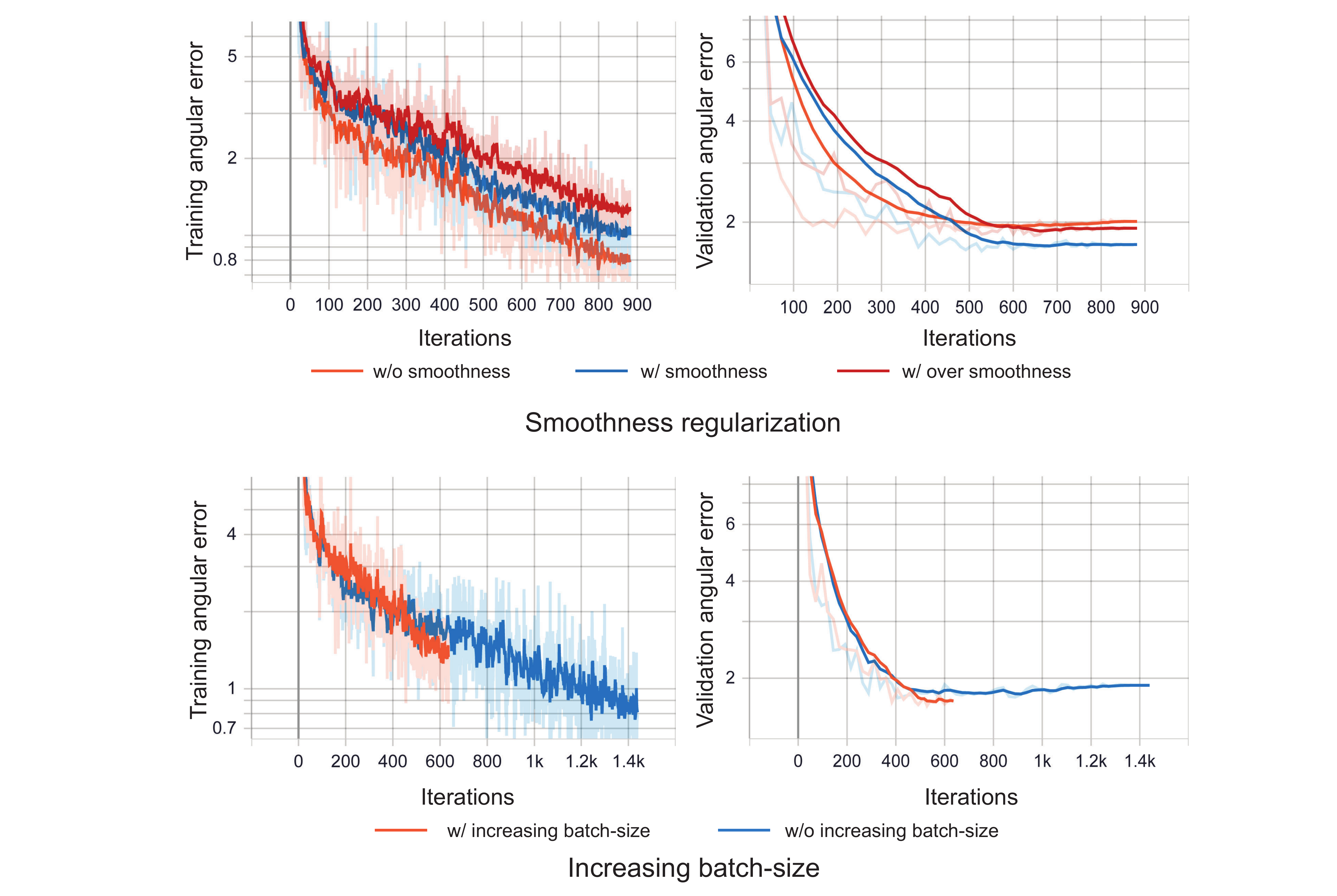}
\vspace{-7mm}
\caption[The impact of  smoothness regularization and of increasing the batch size during training on training/validation accuracy.]{The impact of  smoothness regularization and of increasing the batch size during training on training/validation accuracy. We show the training/validation angular error of training our network on the Gehler-Shi dataset \cite{gehler2008bayesian} for camera-specific color constancy. We set $\lambda_{F}=0.15$, $\lambda_{B}=0.02$ for the experiment labeled with `w/ smoothness', while we used $\lambda_{F}=1.85$, $\lambda_{B}=0.25$ for the experiment labeled with `over smoothness' and $\lambda_{F}=0$, $\lambda_{B}=0$ for the `w/o smoothness' experiments.}
\label{C5:fig:regularization_ablation}
\end{figure}

\subsection{Ablations Studies}\label{C5:sec:ablation}



\begin{table}[!t]
\caption[Angular errors on the Cube+ dataset \cite{banic2017unsupervised} and the Cube+ challenge \cite{challenge}.]{Angular errors on the Cube+ dataset \cite{banic2017unsupervised} and the Cube+ challenge \cite{challenge}. Lowest errors are highlighted in yellow.
$\numinputs$ is the number of additional test-time images used as input, and ``\waugmentation'' indicates if our data augmentation procedure is used. See the text for additional details on model variants. C5 yields state-of-the-art performance.
\label{C5:table:results}}
\centering
\resizebox{0.49\linewidth}{!}{
\begin{tabular}{l|ccccc|c}
\textbf{Cube+ Dataset} & \textbf{Mean} & \textbf{Med.} & \textbf{B. 25\%} & \textbf{W. 25\%} & \textbf{Tri.} & \textbf{Size (MB)} \\ \hline

GW \cite{GW}& 3.52 & 2.55 & 0.60 &  7.98 & 2.82 & - \\
1st-order GE \cite{GE}& 3.06 & 2.05 & 0.55 & 7.22 & 2.32 & - \\
2nd-order GE \cite{GE}& 3.28 & 2.34 & 0.66 & 7.44 & 2.58 & - \\
SoG \cite{SoG}& 3.22 & 2.12 &  0.43 & 7.77 & 2.44 & - \\
Cross-dataset CC \cite{koskinen12cross} & 2.47 & 1.94  & - & - & - & - \\
Quasi-U CC \cite{bianco2019quasi} & 2.69 & 1.76 & 0.49 & 6.45 & 2.00 & 622\\
SIIE (Chapter \ref{ch:ch5}) & 2.14 & 1.44 & 0.44 &  5.06 &  -
& 10.3 \\
FFCC \cite{FFCC} & 2.69 & 1.89 & 0.46 & 6.31 & 2.08 & 0.22 \\ \hdashline
C5 ($\numinputs=1$) & 2.60 & 1.86 & 0.55 & 5.89 & 2.10 & 0.72 \\
C5 ($\numinputs=3$) & 2.28 & 1.50 & 0.59 & 5.19 & 1.74 & 1.05\\
C5 ($\numinputs=5$) & 2.23 & 1.52 & 0.56 & 5.11 & 1.71 & 1.39 \\
C5 ($\numinputs=7$)& 2.10 & 1.38 & 0.49 & 4.97 & 1.56 & 1.74 \\
C5 ($\numinputs=7$, \waugmentation) & \cellcolor[HTML]{\bestcolor} 1.87 & \cellcolor[HTML]{\bestcolor} 1.27 & \cellcolor[HTML]{\bestcolor} 0.41 & \cellcolor[HTML]{\bestcolor} 4.36 & \cellcolor[HTML]{\bestcolor} 1.40 & 1.74\\
C5 ($\numinputs=9$, \waugmentation) & 1.92 & 1.32 & 0.44 & 4.44 & 1.46 & 2.09 \\ 
\end{tabular}}

\vspace{0.11in}

\resizebox{0.49\linewidth}{!}{
\begin{tabular}{l|ccccc}
\textbf{Cube+ Challenge} & \textbf{Mean} & \textbf{Med.} & \textbf{B. 25\%} & \textbf{W. 25\%} & \textbf{Tri.}  \\ \hline

GW \cite{GW} & 4.44 & 3.50 & 0.77 & 9.64 & - \\
1st-order GE \cite{GE} & 3.51 & 2.30 & 0.56 & 8.53 & - \\
Quasi-U CC \cite{bianco2019quasi} & 3.12 & 2.19 & 0.60 & 7.28 & 2.40\\
SIIE (Chapter \ref{ch:ch5}) & 2.89 & 1.72 & 0.71 &  7.06 & - \\
FFCC \cite{FFCC} & 3.25 & 2.04 & 0.64 & 8.22 & 2.09 \\ \hdashline
C5 ($\numinputs=1$) & 2.70 & 2.00 & 0.61  & 6.15 & 2.06\\
C5 ($\numinputs=7$) & 2.55 & 1.63 & 0.54 & 6.21 & 1.79  \\
C5 ($\numinputs=9$) & 2.24 & 1.48 & 0.47 & \cellcolor[HTML]{\bestcolor}5.39 & 1.62 \\
C5 ($\numinputs=9$, another camera model) & 2.97 & 2.47 & 0.78 & 6.11 & 2.52\\
C5 ($\numinputs=9$, dull images) & 2.35 & 1.58 & 0.46 & 5.57 & 1.70 \\
C5 ($\numinputs=9$, vivid images) & \cellcolor[HTML]{\bestcolor}2.19 & \cellcolor[HTML]{\bestcolor}1.39 & \cellcolor[HTML]{\bestcolor}0.43 & 5.44 & \cellcolor[HTML]{\bestcolor}1.54 
\end{tabular}}
\end{table}

In the following ablation experiments, we used the Cube+ dataset \cite{banic2017unsupervised} as our test set and trained our network with seven encoders (i.e., $m=7$) using the following training sets: the NUS dataset \cite{cheng2014illuminant}, the Gehler-Shi dataset \cite{gehler2008bayesian}, and the augmented images after excluding any scene/sensors of the test set. Table \ref{C5:table:ablation} shows the results obtained by models trained using different histogram sizes, using different values of the smoothness factors $\lambda_{B}$ and $\lambda_{F}$, with and without increasing the batch-size during training, and with and without the histogram gradient intensity and the extra $uv$ augmentation channels. Each experiment was repeated ten times and the arithmetic mean and standard deviation of each error metric are reported.

\begin{figure}[!t]
\includegraphics[width=\linewidth]{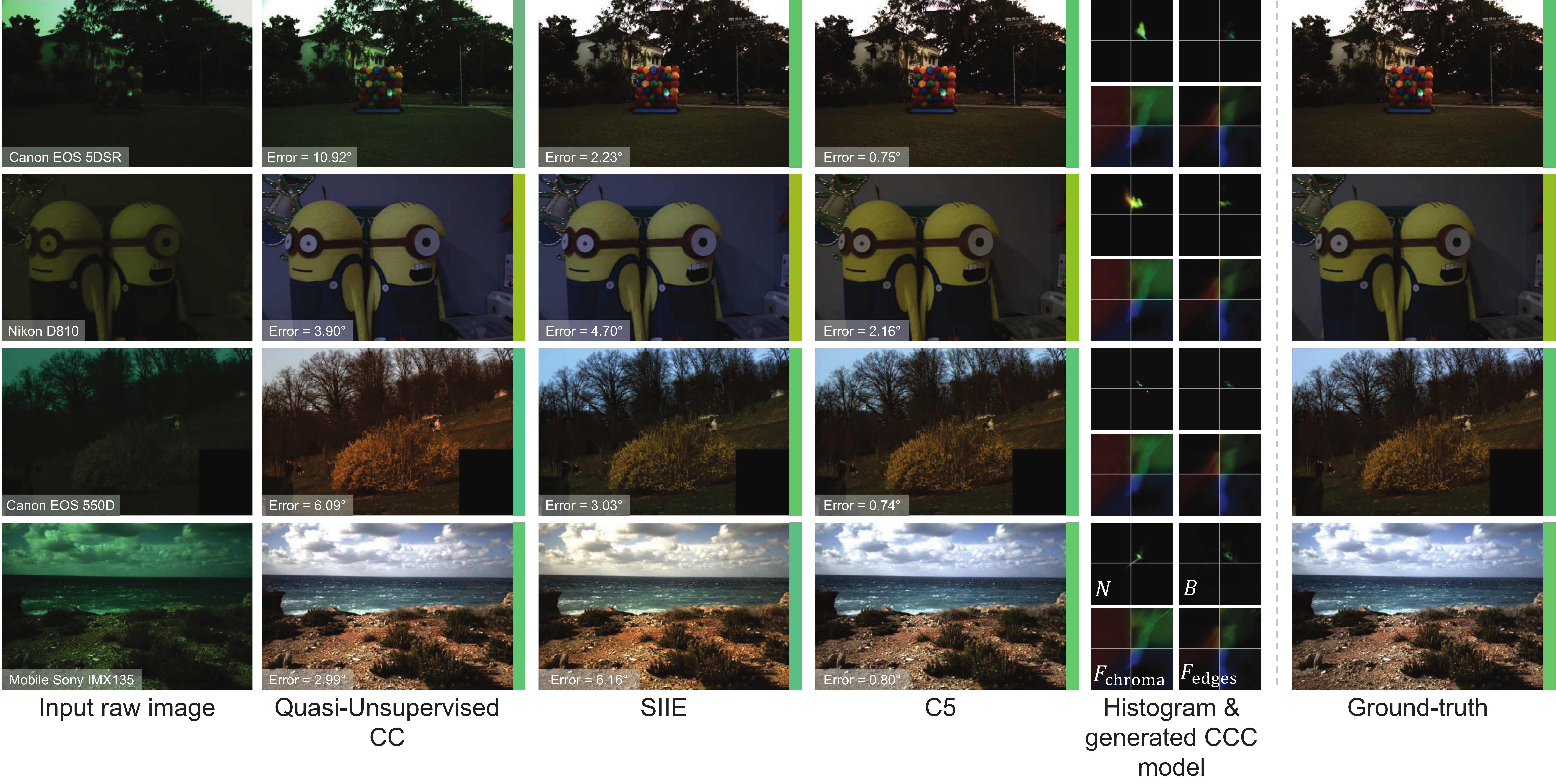}
\vspace{-7mm}
\caption[In this figure, we visualize the performance of our C5 model alongside other camera-independent models]{In this figure, we visualize the performance of our C5 model alongside other camera-independent models: ``quasi-unsupervised CC'' \cite{bianco2019quasi} and SIIE (Chapter \ref{ch:ch5}).
Despite not having seen any images from the test-set camera during training, C5 is able to produce accurate illuminant estimates. The intermediate CCC filters and biases produced by C5 are also visualized.\label{C5:fig:qualitative_comparisons}}
\end{figure}

\begin{figure}[t]
\includegraphics[width=\linewidth]{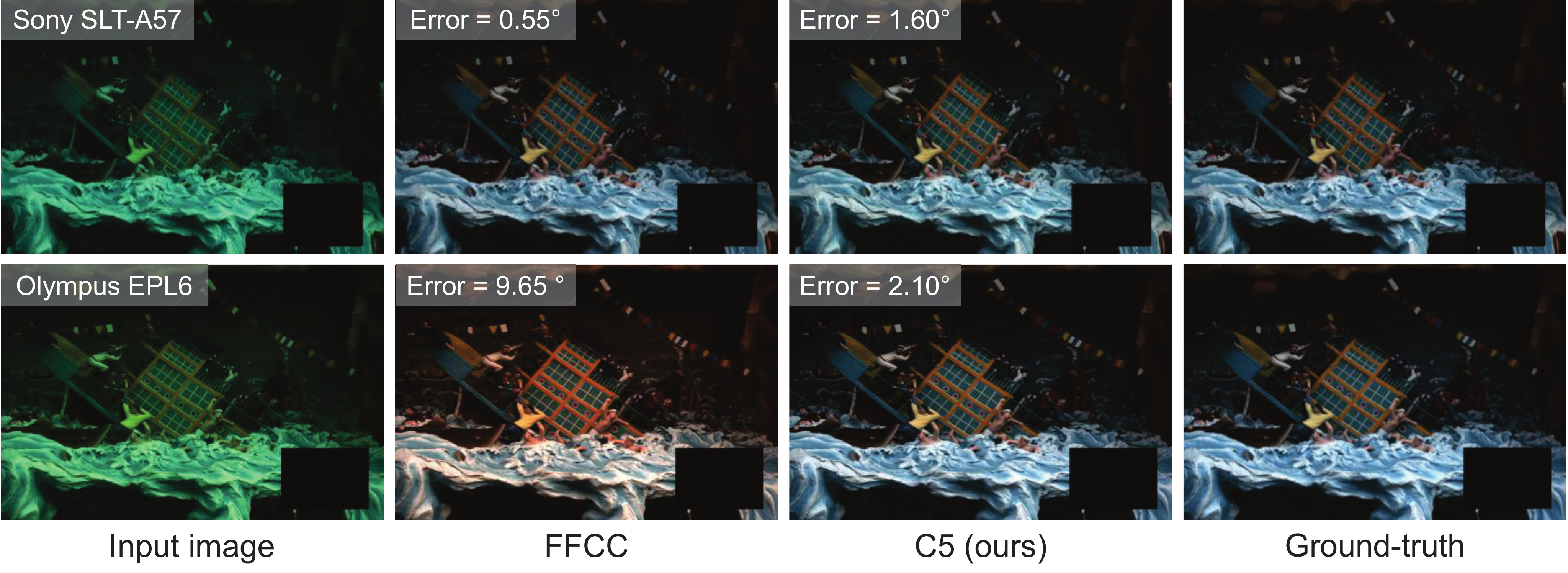}
\vspace{-7mm}
\caption[In this figure, we compare our C5 model against FFCC~\cite{FFCC} on cross-sensor generalization using test-set Sony SLT-A57 images.]{In this figure, we compare our C5 model against FFCC~\cite{FFCC} on cross-sensor generalization using test-set Sony SLT-A57 images from the NUS dataset~\cite{cheng2014illuminant}. If FFCC is trained and tested on images from the same camera, it performs well, as does C5 (top row). But if FFCC is instead tested on a different camera, such as the Olympus EPL6, it generalizes poorly, while C5 retains its performance (bottom row).\label{C5:fig:comparison_w_FFCCC}}
\end{figure}

\begin{table}[!t]
\caption[Angular errors on the INTEL-TAU dataset \cite{laakom2019intel}.]{Angular errors on the INTEL-TAU dataset \cite{laakom2019intel}. Lowest errors are highlighted in yellow.
$\numinputs$ is the number of additional test-time images used as input, and ``\waugmentation'' indicates if our data augmentation procedure is used. See the text for additional details on model variants. C5 yields state-of-the-art performance.
\label{C5:table:results2}}
\centering

\resizebox{0.65\linewidth}{!}{
\begin{tabular}{l|ccccc}
\textbf{INTEL-TAU} & \textbf{Mean} & \textbf{Med.} & \textbf{B. 25\%} & \textbf{W. 25\%} & \textbf{Tri.}  \\ \hline

GW \cite{GW} &  4.7 &  3.7 & 0.9 & 10.0 &  4.0\\
White-Patch \cite{maxRGB} &  7.0  & 5.4 & 1.1 & 14.6 & 6.2\\
1st-order GE \cite{maxRGB} & 5.3 & 4.1 & 1.0 & 11.7 & 4.5\\
SoG \cite{SoG} & 4.0 &  2.9 & 0.7 & 9.0 &  3.2\\
PCA-based B/W Colors \cite{cheng2014illuminant} & 4.6 & 3.4 & 0.7 & 10.3 & 3.7 \\
wGE \cite{WGE} & 6.0 & 4.2 & 0.9 & 14.2 & 4.8\\
Quasi-U CC \cite{bianco2019quasi} & 3.12 & 2.19 & 0.60 & 7.28 & 2.40\\
SIIE (Chapter \ref{ch:ch5}) & 3.42 & 2.42 & 0.73 &  7.80 & 2.64 \\
FFCC \cite{FFCC} & 3.42 & 2.38 & 0.70 & 7.96 & 2.61 \\ \hdashline
C5 ($\numinputs=1$) & 2.99 & 2.18 & 0.66  & 6.71 & 2.36\\
C5 ($\numinputs=7$) & 2.62 & 1.85 & 0.54 & 6.05 & 2.00\\
C5 ($\numinputs=7$, \waugmentation) & \cellcolor[HTML]{\bestcolor}2.49 & \cellcolor[HTML]{\bestcolor}1.66 & \cellcolor[HTML]{\bestcolor}0.51 & \cellcolor[HTML]{\bestcolor}5.93 & \cellcolor[HTML]{\bestcolor}1.83 \\
C5 ($\numinputs=9$, \waugmentation) & 2.52 & 1.70 & 0.52 & 5.96 & 1.86 \\

\end{tabular}}
\end{table}

Figure \ref{C5:fig:regularization_ablation} shows the effect of the smoothness regularization and increasing the batch-size during training on a small training set. We use the first fold of the Gehler-Shi dataset \cite{gehler2008bayesian} as our validation set and the remaining two folds are used for training. In the figure we plot the angular error on the training and validation sets. Each model was trained for 60 epochs as a camera-specific color constancy model (i.e., $m=1$ and without using additional training camera models). As can be seen in Fig. \ref{C5:fig:regularization_ablation}, the smoothness regularization improves the generalization on the test set and increasing the batch size helps the network to reach a lower optimum.

\subsection{Results and Comparisons}\label{C5:sbusec:results}

\begin{table}[!t]
\caption[Angular errors on the Gehler-Shi dataset \cite{gehler2008bayesian}, and the NUS dataset \cite{cheng2014illuminant}.]{Angular errors on the Gehler-Shi dataset \cite{gehler2008bayesian}, and the NUS dataset \cite{cheng2014illuminant}. Lowest errors are highlighted in yellow.
$\numinputs$ is the number of additional test-time images used as input, ``\waugmentation'' indicates if our data augmentation procedure is used, and ``CS'' refers to cross-sensor as used in Chapter \ref{ch:ch5}. See the text for additional details on model variants. C5 yields state-of-the-art performance.
\label{C5:table:results3}}
\centering
\resizebox{0.49\linewidth}{!}{

\begin{tabular}{l|ccccc}
\textbf{Gehler-Shi Dataset} & \textbf{Mean} & \textbf{Med.} & \textbf{B. 25\%} & \textbf{W. 25\%} & \textbf{Tri.}  \\ \hline

PCA-based B/W Colors \cite{cheng2014illuminant}& 3.52 & 2.14 & 0.50 & 8.74  & 2.47 \\
ASM \cite{akbarinia2017colour} & 3.80 & 2.40 & - & - & 2.70 \\
Woo \textit{et al.} \cite{8226796} & 4.30 & 2.86 & 0.71 & 10.14 & 3.31 \\
Grayness Index \cite{GI} & 3.07 & 1.87 &  \cellcolor[HTML]{\bestcolor}0.43 &  7.62 &  2.16 \\
Cross-dataset CC \cite{koskinen12cross} & 2.87 & 2.21 & - & - & - \\
Quasi-U CC \cite{bianco2019quasi} & 3.46 & 2.23 & - & - & - \\
SIIE (Chapter \ref{ch:ch5}) & 2.77 & 1.93 & 0.55 & 6.53 &  -  \\
FFCC \cite{FFCC} & 2.95 & 2.19 & 0.57 & 6.75 & 2.35 \\\hdashline 
C5 ($\numinputs=1$) & 2.98 & 2.05 & 0.54 & 7.13 & 2.25 \\
C5 ($\numinputs=7$, \waugmentation) & \cellcolor[HTML]{\bestcolor}2.36 & \cellcolor[HTML]{\bestcolor}1.61 & 0.44 & \cellcolor[HTML]{\bestcolor}5.60 & \cellcolor[HTML]{\bestcolor}1.74\\
CS ($\numinputs=9$, \waugmentation) & 2.50 & 1.99 & 0.53 & 5.46 & 2.03 \\ 
\end{tabular}
}

\vspace{0.11in}

\centering
\resizebox{0.49\linewidth}{!}{
\begin{tabular}{l|ccccc}
\textbf{NUS Dataset} & \textbf{Mean} & \textbf{Med.} & \textbf{B. 25\%} & \textbf{W. 25\%} & \textbf{Tri.}  \\ \hline

PCA-based B/W Colors \cite{cheng2014illuminant}& 2.93 & 2.33 & 0.78 & 6.13 & 2.42 \\
Grayness Index \cite{GI} & 2.91 & 1.97 &  0.56 & 6.67 &  2.13 \\
Cross-dataset CC \cite{koskinen12cross} & 3.08 & 2.24 & - & - & - \\
Quasi-U CC \cite{bianco2019quasi} & 3.00 & 2.25  & - & - & - \\	
SIIE (CS) (Chapter \ref{ch:ch5}) & 2.05 & 1.50 & 0.52 & 4.48  &  \\
FFCC \cite{FFCC} & 2.87 & 2.14 & 0.71 & 6.23 & 2.30
\\ \hdashline
C5 ($\numinputs=1$) & 2.84 & 2.20 & 0.69 & 6.14 & 2.33 \\
C5 ($\numinputs=7$, \waugmentation) & 2.68 & 2.00 & 0.66 & 5.90 & 2.14 \\
CS ($\numinputs=9$, \waugmentation) & 2.54 & 1.90 & 0.61 & 5.61 & 2.02 \\ 
C5 ($\numinputs=9$, CS) & \cellcolor[HTML]{\bestcolor}1.77 & \cellcolor[HTML]{\bestcolor}1.37 & \cellcolor[HTML]{\bestcolor}0.48 & \cellcolor[HTML]{\bestcolor}3.75 & \cellcolor[HTML]{\bestcolor}1.46 \\

\end{tabular}
}
\end{table}


We validate our model using four public datasets consisting of images taken from one or more camera models: the Gehler-Shi dataset (568 images, two cameras) \cite{gehler2008bayesian}, the NUS dataset (1,736 images, eight cameras) \cite{cheng2014illuminant}, the INTEL-TAU dataset\footnote{This is an updated version of the INTEL-TUT used in Chapter\ \ref{ch:ch5} for evaluation.} (7,022 images, three cameras) \cite{laakom2019intel}, and the Cube+ dataset (2,070 images, one camera) \cite{banic2017unsupervised} which has a separate 2019 ``Challenge'' test set~\cite{challenge}.

As done in Chapter \ref{ch:ch5}, we measure performance by reporting the error statistics commonly used by the community: the mean, median, trimean, and arithmetic means of the first and third quartiles (``best 25\%'' and ``worst 25\%'') of the angular error between the estimated illuminant and the true illuminant.
To evaluate our model's performance at generalizing to new camera models not seen during training, we adopt a leave-one-out cross-validation evaluation approach: for each dataset, we exclude all scenes and cameras used by the test set from our training images. 

To evaluate the improvement of using the additional input images, we report multiple versions of our model in which we vary $\numinputs$, the number of the additional images (and encoders) used ($\numinputs=1$ means that only the query image is used as input). 

As our method randomly selects the additional images, each experiment is repeated ten times and we reported the arithmetic mean of each error metric.

For a fair comparison with FFCC \cite{FFCC}, we trained FFCC using the same leave-one-out cross-validation evaluation approach. Results can be seen in Tables~\ref{C5:table:results}--\ref{C5:table:results3} and qualitative comparisons are shown in Figs.~\ref{C5:fig:qualitative_comparisons} and~\ref{C5:fig:comparison_w_FFCCC}. Even when compared with prior sensor-independent techniques \cite{bianco2019quasi} and our previous SIIE method (Chapter \ref{ch:ch5}), C5 achieves state-of-the-art performance when using ($\numinputs \ge 7$) images, as demonstrated in Tables~\ref{C5:table:results}--\ref{C5:table:results3}.

When evaluating on the two Cube+ ~\cite{banic2017unsupervised,challenge} test sets and the INTEL-TAU~\cite{laakom2019intel} dataset in Tables~\ref{C5:table:results} and \ref{C5:table:results2}, we train our model on the NUS \cite{cheng2014illuminant} and Gehler-Shi \cite{gehler2008bayesian} datasets. When evaluating on the Gehler-Shi \cite{gehler2008bayesian} and the NUS \cite{cheng2014illuminant} datasets in Table~\ref{C5:table:results3}, we train C5 using the INTEL-TAU dataset \cite{laakom2019intel}, the Cube+ dataset \cite{banic2017unsupervised}, and one of the Gehler-Shi \cite{gehler2008bayesian} and the NUS \cite{cheng2014illuminant} datasets after excluding the testing dataset.

\begin{table}[!t]
\centering
\caption[Results of using the gain multiplier, $G$.]{Results of using the gain multiplier, $G$. For each experiment, we used $m=7$ and $n=64$, and trained our network using the same training data explained earlier with augmentation. Lowest errors are highlighted in yellow.\label{C5:table:with_gain}}
\scalebox{0.525}{
\begin{tabular}{l|cccc|cccc|cccc|cccc}
 & \multicolumn{4}{c|}{\textbf{Cube+} \cite{banic2017unsupervised}} & \multicolumn{4}{c|}{\textbf{Cube+ Challenge} \cite{challenge}} & \multicolumn{4}{c|}{\textbf{INTEL-TAU} \cite{laakom2019intel}} & \multicolumn{4}{c}{\textbf{Gehler-Shi} \cite{gehler2008bayesian}}   \\ \cline{2-17} 
 & \textbf{Mean} & \textbf{Med.} & \textbf{B. 25\%} & \textbf{W. 25\%} & \textbf{Mean} & \textbf{Med.} & \textbf{B. 25\%} & \textbf{W. 25\%} & \textbf{Mean} & \textbf{Med.} & \textbf{B. 25\%} & \textbf{W. 25\%} & \textbf{Mean} & \textbf{Med.} & \textbf{B. 25\%} & \textbf{W. 25\%} \\ \hline
w/o $G$ &  1.87 & 1.27 & 0.41 & 4.36 &  2.40 & 1.58 & 0.52 & \cellcolor[HTML]{\bestcolor} 5.76 &  \cellcolor[HTML]{\bestcolor}2.49 & \cellcolor[HTML]{\bestcolor} 1.66 & \cellcolor[HTML]{\bestcolor} 0.51 & \cellcolor[HTML]{\bestcolor} 5.93 & \cellcolor[HTML]{\bestcolor} 2.36 & \cellcolor[HTML]{\bestcolor} 1.61 & \cellcolor[HTML]{\bestcolor}0.44 & 5.60 \\
w/ $G$ & \cellcolor[HTML]{\bestcolor}1.83 & \cellcolor[HTML]{\bestcolor}1.24 & \cellcolor[HTML]{\bestcolor}0.42 & \cellcolor[HTML]{\bestcolor}4.25 & \cellcolor[HTML]{\bestcolor}2.34 & \cellcolor[HTML]{\bestcolor}1.45 & \cellcolor[HTML]{\bestcolor}0.46 & 5.86 & 2.63 & 1.81 & 0.55 & 6.18 & \cellcolor[HTML]{\bestcolor}2.36 & 1.72 & 0.48 & \cellcolor[HTML]{\bestcolor}5.4 \\
\end{tabular}}
\end{table}


\begin{table}[!t]
\centering
\caption[Results using the INTEL-TAU dataset evaluation protocols \cite{laakom2019intel}.]{Results using the INTEL-TAU dataset evaluation protocols \cite{laakom2019intel}. We also show the results of camera-independent methods, including our camera-independent C5 model. Lower errors for each evaluation protocol are highlighted in yellow. The best results are bold-faced.\label{C5:table:intel_tau}}
\scalebox{0.55}{
\begin{tabular}{l|ccccc}
\textbf{\begin{tabular}[c]{@{}l@{}}\textbf{INTEL-TAU} \cite{laakom2019intel} \end{tabular}} & \textbf{Mean} & \textbf{Med.} & \textbf{B. 25\%} & \textbf{W. 25\%} & \textbf{Tri.}  \\ \hline
\multicolumn{6}{c}{\cellcolor[HTML]{\headercolor}\textbf{Camera-specific (10-fold cross-validation protocol \cite{laakom2019intel})}} \\ \hline
Bianco et al.'s CNN \cite{bianco2015color}  &  3.5 &  2.6 & 0.9 & 7.4 & 2.8 \\
C3AE \cite{laakom2019color}  &  3.4 & 2.7 & 0.9 & 7.0 &  2.8 \\
BoCF \cite{laakom2020bag}  & 2.4 &  1.9 & 0.7 & 5.1 & 2.0 \\
FFCC \cite{FFCC} &  2.4 &  1.6  & 0.4 & 5.6 & 1.8 \\
VGG-FC$^4$ \cite{hu2017fc}  &  \cellcolor[HTML]{\bestcolor}\textbf{2.2} & 1.7 & 0.6 &  \cellcolor[HTML]{\bestcolor}\textbf{4.7} & 1.8 \\
\hdashline
C5 ($m=7, n=128$), w/ augmentation & 2.33 & \cellcolor[HTML]{\bestcolor}\textbf{1.55} & \cellcolor[HTML]{\bestcolor}\textbf{0.45} & 5.57 & \cellcolor[HTML]{\bestcolor}\textbf{1.71} \\\hline

\multicolumn{6}{c}{\cellcolor[HTML]{\headercolor}\textbf{Camera-specific (camera invariant protocol \cite{laakom2019intel})}} \\ \hline
Bianco et al.'s CNN \cite{bianco2015color} &  3.4 &  2.5 & 0.8 & 7.2 & 2.7 \\

C3AE \cite{laakom2019color} & 3.4 & 2.7 & 0.9 &  7.0 &  2.8 \\

BoCF \cite{laakom2020bag} &  2.9 & 2.4 & 0.9 &  6.1 & 2.5 \\

VGG-FC$^4$ \cite{hu2017fc}  &  2.6 &  2.0 & 0.7 & 5.5 & 2.2 \\

\hdashline

C5 ($m=9$), \waugmentation & \cellcolor[HTML]{\bestcolor}2.45 & \cellcolor[HTML]{\bestcolor}1.82 & \cellcolor[HTML]{\bestcolor}0.53  & \cellcolor[HTML]{\bestcolor}5.46 & \cellcolor[HTML]{\bestcolor}1.95\\

\hline
\multicolumn{6}{c}{\cellcolor[HTML]{\headercolor}\textbf{Camera-independent}} \\ \hline

GW \cite{GW} &  4.7 &  3.7 & 0.9 & 10.0 &  4.0\\
White-Patch \cite{maxRGB} &  7.0  & 5.4 & 1.1 & 14.6 & 6.2\\
1st-order GE \cite{maxRGB} & 5.3 & 4.1 & 1.0 & 11.7 & 4.5\\
2nd-order GE \cite{maxRGB} & 5.1 & 3.8 & 1.0 & 11.3 & 4.2\\
SoG \cite{SoG} & 4.0 &  2.9 & 0.7 & 9.0 &  3.2\\
PCA-based B/W Colors \cite{cheng2014illuminant} & 4.6 & 3.4 & 0.7 & 10.3 & 3.7 \\
wGE \cite{WGE} & 6.0 & 4.2 & 0.9 & 14.2 & 4.8\\
Quasi-U CC \cite{bianco2019quasi} & 3.12 & 2.19 & 0.60 & 7.28 & 2.40\\
SIIE (Chapter \ref{ch:ch5}) & 3.42 & 2.42 & 0.73 &  7.80 & 2.64 
\\ 

\hdashline
C5 ($m=7$), \waugmentation & \cellcolor[HTML]{\bestcolor}2.49 & \cellcolor[HTML]{\bestcolor}1.66 & \cellcolor[HTML]{\bestcolor}0.51 & \cellcolor[HTML]{\bestcolor}5.93 & \cellcolor[HTML]{\bestcolor}1.83 \\

\end{tabular}}
\end{table}



\begin{table}[!t]
\caption[Results of our C5 trained as a camera-specific model with a single encoder (i.e., $m=1$).]{Results of our C5 trained as a camera-specific model with a single encoder (i.e., $m=1$). In these experiments, we trained our model using a three-fold cross-validation of each dataset, except for the Cube+ challenge \cite{challenge}, where we report our results after training our model on the Cube+ dataset \cite{banic2017unsupervised}. We also show the results of other camera-specific color constancy methods reported in past papers. Lowest angular errors are highlighted in yellow. \label{C5:table:camera_specific}}
\centering
\resizebox{0.6\linewidth}{!}{
\begin{tabular}{l|ccccc}
\textbf{\begin{tabular}[c]{@{}l@{}}\textbf{Cube+ Dataset} \cite{banic2017unsupervised} \end{tabular}} &  \textbf{Mean} & \textbf{Med.} & \textbf{B. 25\%} & \textbf{W. 25\%} & \textbf{Tri.}  \\ \hline

Color Dog \cite{colorDog}& 3.32 & 1.19 & 0.22 & 10.22 & -  \\
APAP \cite{afifi2019projective}& 2.01  &  1.36 & 0.38 & 4.71 &  -  \\
Meta-AWB w/ 20 tuning images \cite{mcdonagh2018formulating} & 1.59 & 1.02 & 0.30 &  3.85 & 1.15 - \\
Color Beaver \cite{kovsvcevic2019color}& 1.49 & 0.77 & 0.21 &
3.94 & -  \\
SqueezeNet-FC$^4$ \cite{hu2017fc} & 1.35 & 0.93 & 0.30 & 3.24 & 1.01  \\
FFCC \cite{FFCC} & 1.38 & \cellcolor[HTML]{\bestcolor}0.74 & \cellcolor[HTML]{\bestcolor}0.19 & 3.67 & \cellcolor[HTML]{\bestcolor}0.89  \\
MDLCC \cite{xiao2020multi} & \cellcolor[HTML]{\bestcolor}1.24 & 0.83 &  0.26 & \cellcolor[HTML]{\bestcolor}2.91 &  0.92 \\
\hdashline
C5 ($n=128$), w/ $G$ & 1.39 & 0.79 & 0.24 & 3.55 & 0.93  \\
\end{tabular}}

\vspace{0.16in}

\centering
\resizebox{0.6\linewidth}{!}{
\begin{tabular}{l|ccccc}
\textbf{\begin{tabular}[c]{@{}l@{}}\textbf{Cube+ Challenge} \cite{challenge}\end{tabular}} & \textbf{Mean} & \textbf{Med.} & \textbf{B. 25\%} & \textbf{W. 25\%} & \textbf{Tri.}  \\ \hline
V Vuk et al., \cite{challenge}& 6.00 & 1.96 & 0.99 & 18.81 & 2.25 \\
A Savchik et al., \cite{savchik2019color} & 2.05 & 1.20 & 0.40 & 5.24 & 1.30 \\
Y Qian et al., (1) \cite{qian2019fast} &  2.48 & 1.56 & 0.44 & 6.11 & - \\
Y Qian et al., (2) \cite{qian2019fast} & 2.27 & 1.26  & 0.39 & 6.02 & 1.35\\
FFCC \cite{FFCC} &  2.1 & 1.23 & 0.47 & 5.38 & - \\
MHCC \cite{hernandez2020multi} & 1.95 & 1.16 & 0.39 & 4.99 & 1.25 \\
K Chen et al., \cite{challenge}& 1.84 & 1.27 & 0.39 & 4.41 & 1.32 \\
\hdashline
C5 ($n=128$), w/ $G$ & \cellcolor[HTML]{\bestcolor}1.72 & \cellcolor[HTML]{\bestcolor}1.07 & \cellcolor[HTML]{\bestcolor}0.36 & \cellcolor[HTML]{\bestcolor}4.27 & \cellcolor[HTML]{\bestcolor}1.15\\
\end{tabular}}
\end{table}

The one deviation from this procedure is for the NUS result labeled ``CS'', where for a fair comparison with our SIIE method, proposed in Chapter \ref{ch:ch5}, we report our results with their cross-sensor (CS) evaluation, in which we only excluded images of the test camera, and repeated this process over all cameras in the dataset.

For experiments labeled ``w/aug" in Table~\ref{C5:table:results}--\ref{C5:table:results3}, we augmented the data used to train the model, adding 5,000 augmented examples generated as described in Sec.\ \ref{C5:subsec:dataaug}. In this process, we used only cameras of the training sets of each experiment as ``target" cameras for augmentation, which has the effect of mixing the sensors and scene content from the training sets only. For instance, when evaluating on the INTEL-TAU~\cite{laakom2019intel} dataset, our augmented images simulate the scene content of the NUS \cite{cheng2014illuminant} dataset as observed by sensors of the Gehler-Shi \cite{gehler2008bayesian} dataset, and vice-versa.

Unless otherwise stated, in our experiments varying $\numinputs$, the additional input images are randomly selected, but from the same camera model as the test image. 
This setting is meant to be equivalent to the real-world use case in which the additional images provided as input are, say, a photographer's previously-captured images that are already present on the camera during inference.
However, for the ``Cube+ Challenge'' table, we provide an additional set of experiments in which the set of additional images are chosen according to some heuristic, rather than randomly. We identified the 20 test-set images with the lowest variation of $uv$ chroma values (``dull images''), the 20 test-set images with the highest variation of $uv$ chroma values (``vivid images''), and we show that using vivid images produces lower error rates than randomly-chosen or dull images. This makes intuitive sense, as one might expect colorful images to be a more informative signal as to the spectral properties of previously-unobserved camera. We also show results where the additional images are taken from a different camera than the test-set camera, and show that this results in error rates that are higher than the $\numinputs=1$ case, as one might expect.

We did not include the ``gain'' multiplier, originally proposed in FFCC \cite{FFCC}, in the main method section as it did not result in a consistent improved performance over all error metrics and datasets. Here, we report results with and without using the gain multiplier map. This gain multiplier map can be generated by our network by adding an additional decoder network with skip connections from the query encoder. This modification increases our model size from 1.74 MB to 1.97 MB using $m=7$. Based on this modification, our convolutional structure can now be described as:
\begin{equation}
\mat{P} = \operatorname{softmax}\bigg(\mat{B} + \mat{G} \circ \sum_i\big(\mat{N}_i * \mat{F}_i\big)\bigg)\,, \label{C5:eq:CCC_w_G}
\end{equation}
where $\{ \mat{F}_i \}$, $\mat{B}$, and $\mat{G}$ are filters, a bias map $\mat{B}(i,j)$, and the gain multiplier map $\mat{G}(i,j)$, respectively. We also change the smoothness regularizer to include the generated gain multiplier as follows:
\begin{align}
S\left(\{ \mat{F}_i \}, \mat{B}, \mat{G}\right) = \lambda_{B} (&\lVert \mat{B} \ast \sobel_u \rVert^2 + \lVert \mat{B} \ast \sobel_v \rVert^2 )  \nonumber \\
+ \lambda_{G} (&\lVert \mat{G} \ast \sobel_u \rVert^2 + \lVert \mat{G} \ast \sobel_v \rVert^2 ) \nonumber \\
+ \lambda_{F} \sum_i (&\lVert \mat{F}_i \ast \sobel_u \rVert^2 + \lVert \mat{F}_i \ast \sobel_v \rVert^2 ) \,, \label{C5:loss:Eq.2_supp}
\end{align}
\noindent where $\sobel_u$ and $\sobel_v$ are $3\!\times\!3$ horizontal and vertical Sobel filters, respectively, and $\lambda_{F}$, $\lambda_{B}$, $\lambda_{G}$ are scalar multipliers to control the strength of the smoothness of each of the filters, the bias, and the gain, respectively. The results of using the additional gain multiplier map are reported in Table \ref{C5:table:with_gain}.

We further trained and tested our C5 model using the INTEL-TAU dataset evaluation protocols \cite{laakom2019intel}. Specifically, the INTEL-TAU dataset introduced two different evaluation protocols: (i) the cross-validation protocol, where the model is trained using a 10-fold cross-validation scheme of images taken from three different camera models, and (ii) the camera invariance evaluation protocol, where the model is trained on a single camera model and then tested on another camera model. This camera invariance protocol is equivalent to the CS evaluation used in Chapter \ref{ch:ch5}, as the models are trained and tested on the same scene set, but with different camera models in the training and testing phases. See Table \ref{C5:table:intel_tau} for comparison with other methods using the INTEL-TAU evaluation protocols. In Table \ref{C5:table:intel_tau}, we also show the results of our C5 model trained on the NUS and Gehler-Shi datasets with augmentation (i.e., our camera-independent model) for completeness. 

Our C5 model achieves reasonable accuracy when used as a camera-specific model. In this scenario, we trained our model on training images captured by the same test camera model with a single encoder (i.e., $m=1$). We found that $n=128$, using the gain multiplier map $G(i,j)$, achieves the best camera-specific results. We report the results of our camera-specific models in Table \ref{C5:table:camera_specific}. 

\section{Summary} \label{C5:sec.conclusion}
We have presented C5, a cross-camera convolutional color constancy method. By embedding the existing state-of-the-art CCC model \cite{CCC, FFCC} into a multi-input hypernetwork approach, C5 can be trained on images from multiple cameras, but at test time synthesize weights for a CCC-like model that is dynamically calibrated to the spectral properties of the previously-unseen camera of the test-set image.
Extensive experimentation demonstrates that C5 achieves state-of-the-art performance on cross-camera color constancy for several datasets.
By enabling accurate illuminant estimation without requiring the tedious collection of labeled training data for every particular camera, we hope that C5 will accelerate the widespread adoption of learning-based white balance by the camera industry.

\part{Addressing Camera White-Balance Errors\label{part:wb-correction}}

\chapter[Correcting Improperly White-Balanced Images]{Correcting Improperly White-Balanced\\Images \label{ch:ch7}}
Virtually all consumer cameras have a set of pre-defined WB settings for common illuminations that the user can manually select (e.g., sunlight, incandescent, fluorescent).  A problem that often arises is when the sRGB images are rendered with the incorrect WB.  This is generally attributed to the incorrect WB setting being erroneously selected by the user or due to errors by the camera illuminant estimation module.  As previously discussed, incorrectly white-balanced images can be extremely difficult to correct in the sRGB-rendered image due to the nonlinear color manipulation applied after the essential WB step---{\it even} when the correct WB settings or a scene reference white can be identified. This part of the thesis discuss this problem -- namely, correcting improperly white-balanced images -- and studies the impact of WB errors on the performance of different computer vision tasks.

\section{Introduction}

Recall that we showed earlier in Fig.~\ref{fig:wrongWB}-(E) and  Fig.~\ref{fig:wrongWB}-(F) attempts at using the diagonal WB corrections using the color rendition chart's patch and bridge scene region as reference white, respectively.  We can see that in both cases, only the selected reference white region is corrected, while the other region remains incorrect. As we clarify that WB is applied early in the processing chain, attempting to correct it using a diagonal matrix will not work. Matlab suggests using an optional pre-linearization step using a 2.2 gamma \cite{anderson1996proposal, ebner2007color}. However, it has long been known that a 2.2 gamma does not reflect the true nature of the camera specific rendering function. There are currently {\it no solutions} that directly address this problem and most images like Fig.~\ref{fig:wrongWB}-(A) are simply discarded.

\begin{figure}[!t]
\includegraphics[width=\linewidth]{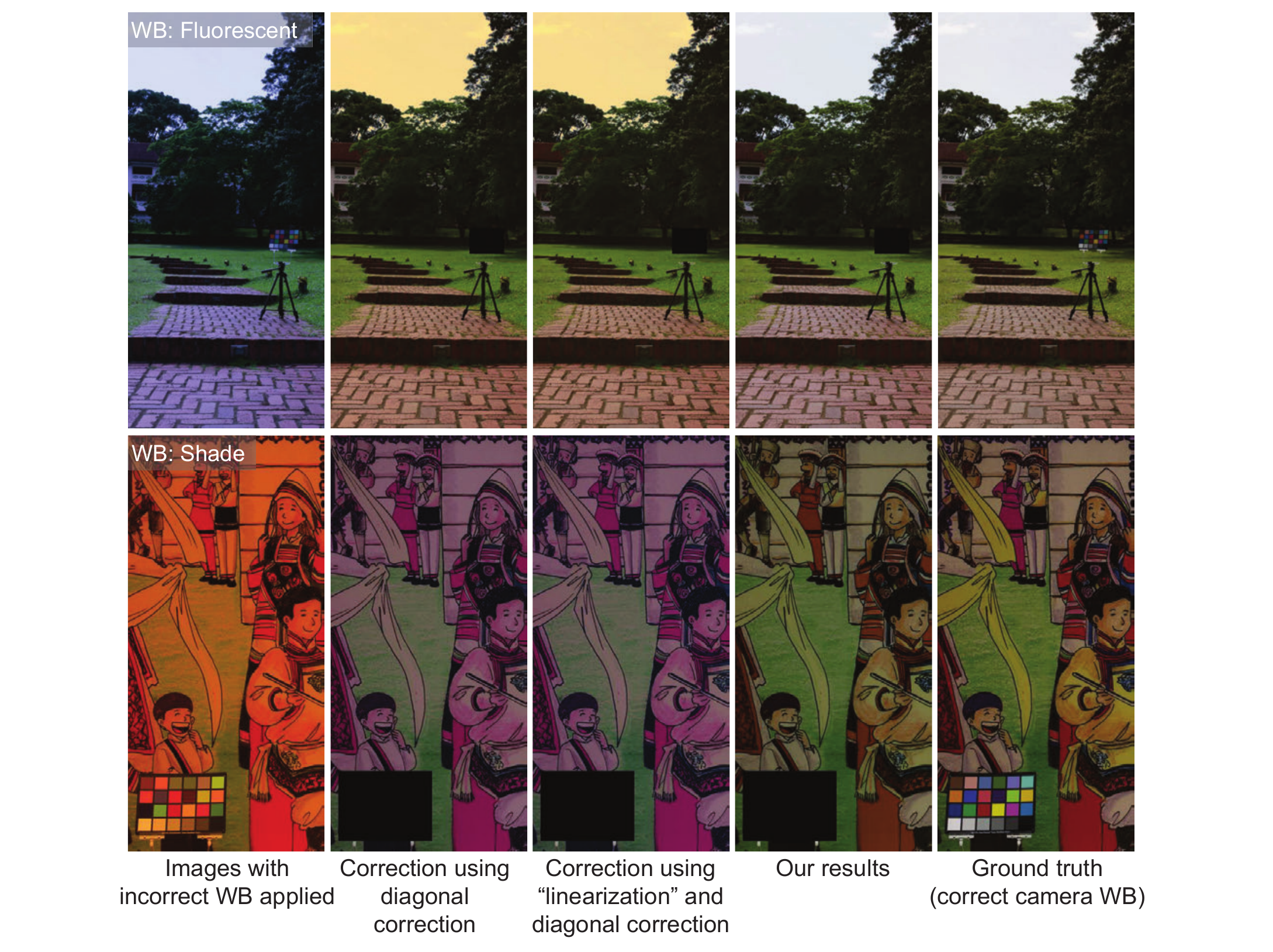}
\vspace{-7mm}
\justifying\caption[This figure shows two incorrectly white-balanced images produced by different cameras.]{This figure shows two incorrectly white-balanced images produced by different cameras. Shown are attempts to correct the images using (1) a linear WB correction, (2) by first applying an erroneously linearization (i.e., a 2.2 gamma \cite{anderson1996proposal, ebner2007color}), and (3) our results. Also shown is the correct output produced by the camera.}
\label{KNN:fig:teaser_1}
\end{figure}

\paragraph{Contribution}~In this chapter, we propose a data-driven approach to correct images that have been improperly white-balanced\footnote{This work was published in \cite{afifi2019color, afifi2020image}: Mahmoud Afifi, Brian Price, Scott Cohen, and Michael S. Brown. When Color Constancy Goes Wrong: Correcting Improperly White-Balanced Images. In IEEE Conference on Computer Vision and Pattern Recognition (CVPR), 2019.}.  As part of this effort, we have generated a new dataset of over 65,000 images from different cameras that have been rendered into a camera's output sRGB image with each camera's different pre-defined WB and photo-finishing settings. The latter is also referred to as picture styles in photography. Each incorrect white-balanced image in the dataset has a corresponding correct white-balanced sRGB image rendered to a standard picture style. Given an improperly white-balanced camera image, we outline a straightforward KNN strategy that is able to find similar incorrectly white-balanced images in the dataset. Based on these similar example images, we describe how to construct a nonlinear color correction transform that is used to remove the color cast.  This idea of using a training set to search for nearest neighbors is commonly used in one-shot and prototype learning \cite{snell2017prototypical, vinyals2016matching}. Our approach gives good results -- see Fig.~\ref{KNN:fig:teaser_1}, and generalizes well to camera makes and models not found in the training data.  In addition, our solution requires a small memory overhead (less than 24 MB) and is computationally fast (less than 1.5 seconds for a 12 mega-pixel image). To the best of our knowledge, this is the first work to explicitly address correcting WB for improperly white-balanced sRGB-rendered images. The dataset and source code of this work are available on GitHub: \href{https://github.com/mahmoudnafifi/WB_sRGB}{https://github.com/mahmoudnafifi/WB$\_$sRGB}.

\section{Methodology}

\subsection{Method Overview}\label{KNN:sec:overview}

\begin{figure*}[!t]
\includegraphics[width=\linewidth]{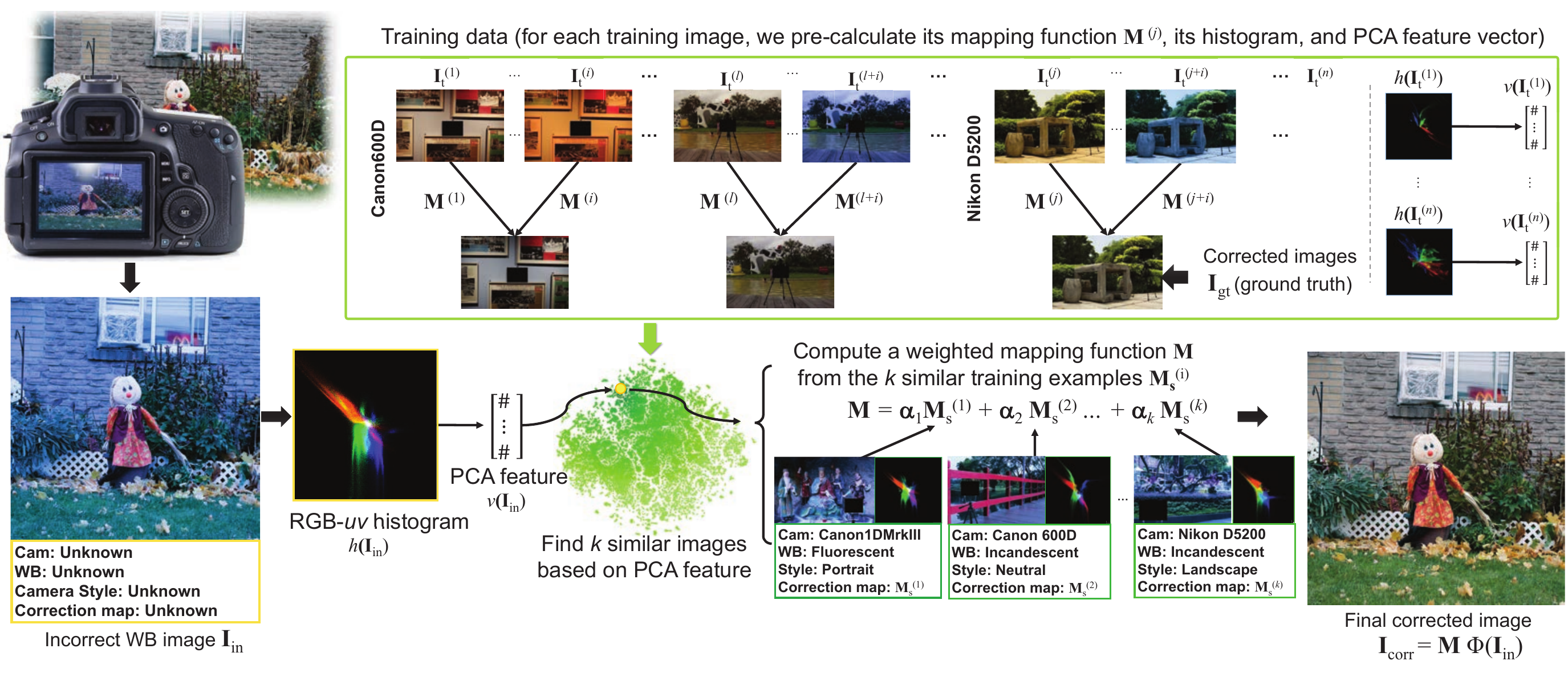}
\vspace{-8mm}
\caption{An overview diagram of our overall procedure. For the input sRGB image and our training data, we first extract the histogram feature of the input image, followed by generating a compact PCA feature to find the most similar $k$ nearest neighbors to the input image in terms of colors. Based on the retrieved similar images, a color transform $\mat{M}$ is computed to correct the input image.}
\label{KNN:fig:overview}
\end{figure*}

We begin with an overview of our approach followed by specific implementation details.  Our method is designed with the additional constraints of fast execution and a small memory overhead to make it suitable for incorporation as a mobile app or as a software plugin. Figure~\ref{KNN:fig:overview} overviews our framework.  Given an incorrectly white-balanced input image, denoted as $\mat{I}_{\textrm{in}}$, our goal is to compute a mapping $\mat{M}$ that can transform the input colors to appear as if the WB was correctly applied.

Our method relies on a large set of $n$ training images expressed as $\mat{I}_{\textrm{t}} = \{\mat{I}^{(1)}_{\textrm{t}},$ $...,$ $\mat{I}^{(n)}_{\textrm{t}}\}$ that have been generated using the {\it incorrect} WB settings. Each training image has a corresponding {\it correct} white-balanced image (or ground truth image), denoted as $\mat{I}^{(i)}_{\textrm{gt}}$.   Note that multiple training images may share the same target ground truth image. Section~\ref{KNN:sec:datageneration} details how we generated this dataset.

For each pair of training image $\mat{I}^{(i)}_{\textrm{t}}$ and its ground truth image $\mat{I}^{(i)}_{\textrm{gt}}$, we compute a \textit{nonlinear} color correction matrix $\mat{M}^{(i)}$ that maps the incorrect image's colors to its target ground truth image's colors. The details of this mapping are discussed in Section~\ref{KNN:sec:MappingFunc}.

Given an input image, we search the training set to find images with similar color distributions.  This image search is performed using compact features derived from input and training image histograms as described in Section~\ref{KNN:sec:similaritymeasurement}. Finally, we obtain a color correction matrix $\mat{M}$ for our input image by blending the associated color correction matrices of the similar training image color distributions, denoted as $\mat{M}_{\textrm{s}}$. This is described in Section~\ref{KNN:sec:finalCorrection}.

\subsection{Dataset Generation}\label{KNN:sec:datageneration}

\begin{figure}[!t]
\includegraphics[width=\linewidth]{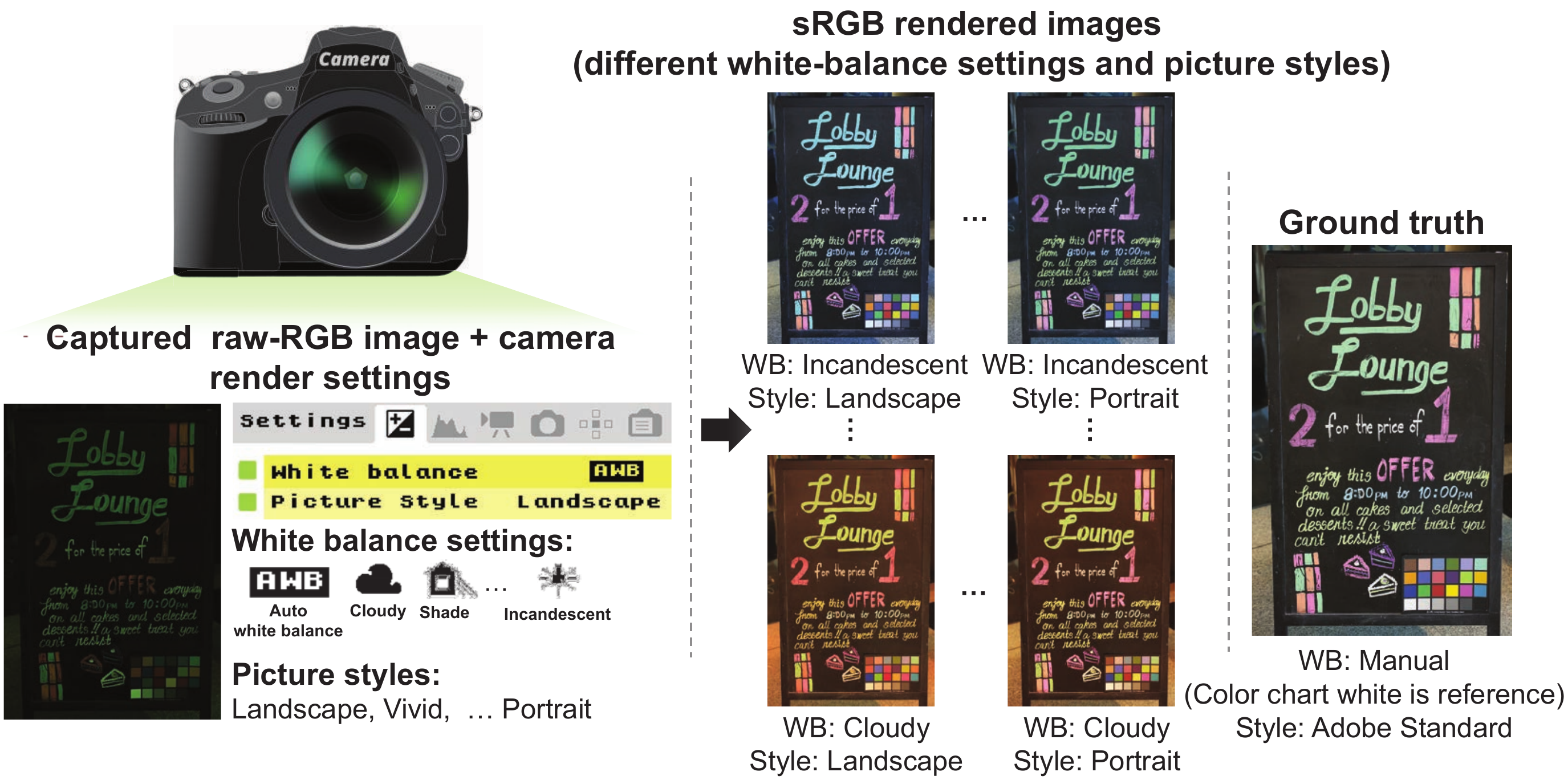}
\caption[Example of rendering an sRGB training image.]{Example of rendering an sRGB training image. Working directly from the raw-RGB camera image, we render sRGB output images using the camera's pre-defined WB settings and different picture styles.  A target WB sRGB image is also rendered using the color rendition chart in the scene to provide the ground truth.}
\label{KNN:fig:dataset}
\end{figure}

Our training images are generated from two publicly available illumination estimation datasets: the NUS dataset~\cite{cheng2014illuminant} and the Gehler dataset~\cite{gehler2008bayesian}.  Images in these datasets are captured using digital single-lens reflex (DSLR) cameras with a color rendition chart placed in the scene that provides ground truth reference for illumination estimation.  Since these datasets are intended for use in illumination estimation, as discussed in Chapters \ref{ch:ch5} and \ref{ch:ch6}, they are captured in raw-RGB format. Because the images are in the camera's raw format, we can convert them to sRGB output emulating different WB settings and picture styles on the camera. To do this, we use the Adobe Camera Raw software development kit (SDK) to render sRGB images using different WB presets in the camera. Adobe Camera Raw accurately emulates the camera imaging pipeline and produces results virtually identical to what the in-camera processing would produce.

In addition, each incorrect WB can be rendered with different camera picture styles (e.g., vivid, standard, neutral, landscape).  Depending on the make and model of the camera, a single raw image can be rendered to more than 25 different camera-specific sRGB images.  These images make up our training images $\{\mat{I}^{(1)}_{\textrm{t}},$ $...,$ $\mat{I}^{(n)}_{\textrm{t}}\}$.

To produce the correct target image, we manually select the ``ground truth'' white from the middle gray patches in the color rendition chart, followed by applying a camera-independent rendering style---namely, {\it Adobe Standard}. This provides the target ground truth sRGB image $\mat{I}^{(i)}_{\textrm{gt}}$. Figure \ref{KNN:fig:dataset} illustrates an example of a raw image from the NUS dataset and the corresponding sRGB images rendered with different WB settings and picture styles.  In the end, we generated 62,535 images from these data sets.

\subsection{Color Correction Transform}\label{KNN:sec:MappingFunc}

After generating our training images, we have $n$ pairs of images representing an incorrectly white-balanced image $\mat{I}^{(i)}_{\textrm{t}}$ and its {\it correct} white-balanced image $\mat{I}^{(i)}_{\textrm{gt}}$. These are represented as $3\!\times\!N$ matrices, where $N$ is the total number of pixels in the image and the three rows represent the red, green, and blue values in the camera's output sRGB color space.

We can compute a color correction matrix $\mat{M}^{(i)}$, which maps  $\mat{I}^{(i)}_{\textrm{t}}$ to $\mat{I}^{(i)}_{\textrm{gt}}$, by minimizing the following equation:
\begin{equation}
\label{KNN:eq1_M}
\underset{\mat{M}^{(i)}}{\argmin} \left\|\mat{M}^{(i)} \textrm{ } \Phi\left(\mat{I}^{(i)}_{\textrm{t}}\right)  - \mat{I}^{(i)}_{\textrm{gt}}\right\|_{\textrm{F}},
\end{equation}
\noindent
where $\left\|.\right\|_{\textrm{F}}$ is the Frobenius norm and $\Phi$ is a kernel function that projects the sRGB triplet to a high-dimensional space.

We have examined several different color transformation mappings and found the polynomial kernel function proposed by Hong et al.~\cite{hong2001study} provided the best results for our task (more details are given in Sec. \ref{KNN:sub:correction_matrices}). Based on \cite{hong2001study}, $\Phi$:$[\textrm{R}$, $\textrm{G}$, $\textrm{B}]^T \rightarrow [\textrm{R}$, $\textrm{G}$, $\textrm{B}$, $\textrm{RG}$, $\textrm{RB}$, $\textrm{GB}$, $\textrm{R}^2$, $\textrm{G}^2$, $\textrm{B}^2$, $\textrm{RGB}$, $1]^T$ and $\mat{M}^{(i)}$ is represented as a $3\!\times\!11$ matrix. 

The color chart in the images was masked out during this process to avoid any bias that may occur from having the same object with a wide range of colors present in the scene. Note that spatial information is not considered when estimating the $\mat{M}^{(i)}$.

\subsection{Image Search}~\label{KNN:sec:similaritymeasurement}

Since our color correction matrix is related to the image's color distribution, our criteria for finding similar images are based on the color distribution.  We also seek compact representation as these features represent the bulk of information that will need to be stored in memory.

We rely on a non-learnable version of the RGB-$uv$ histogram used in Chapter \ref{ch:ch5}. Specifically, we construct a histogram feature from the log-chrominance space, which represents the color distribution of an image $\mat{I}$ as an $m \times m \times 3$ tensor that is parameterized by $uv$. This histogram is generated by the function $h(\mat{I})$ described by the following equations:
\begin{equation}
\label{KNN:eq2_UVHist}
\begin{gathered}
\mat{I}_{y(i)} = \sqrt{\mat{I}_{\textrm{R}(i)}^{2} + \mat{I}_{\textrm{G}(i)}^{2} + \mat{I}_{\textrm{B}(i)}^{2}},
\\
\mat{I}_{u1(i)} = \log{\left(\mat{I}_{\textrm{R}(i)}\right)}  - \log{\left(\mat{I}_{\textrm{G}(i)}\right)}, \\ \mat{I}_{v1(i)} = \log{\left(\mat{I}_{\textrm{R}(i)}\right)} - \log{\left(\mat{I}_{\textrm{B}(i)}\right)},
\\
\mat{I}_{u2} = -\mat{I}_{u1} \textrm{ , }  \mat{I}_{v2} = -\mat{I}_{u1} + \mat{I}_{v1},
\\
\mat{I}_{u3} = -\mat{I}_{v1} \textrm{ , }  \mat{I}_{v3} = -\mat{I}_{v1} + \mat{I}_{u1}.
\end{gathered}
\end{equation}

\begin{equation}
\label{KNN:eq3_UVHist}
\mat{H}\left(\mat{I}\right)_{(u,v,C)} = \sum_{i} \mat{I}_{y(i)} \left[ \left | \mat{I}_{uC(i)} - u \right | \leqslant \frac{\varepsilon}{2} \wedge  \left | \mat{I}_{vC(i)} - v \right | \leqslant \frac{\varepsilon}{2} \right],
\end{equation}

\begin{equation}
\label{KNN:eq4_UVHist}
h(\mat{I})_{(u,v,C)} = \sqrt{\frac{\mat{H}\left(\mat{I}\right)_{(u,v,C)}}{\sum_{u^{'}}{\sum_{v^{'}}{\mat{H}\left(\mat{I}\right)_{(u^{'},v^{'},C)}}}}},
\end{equation}
\noindent
where $i = \{1,...,N\}$, $\textrm{R}, \textrm{G}, \textrm{B}$ represent the color channels in $\mat{I}$, $C \in \{1, 2, 3\}$ represents each color channel in the histogram, and $\varepsilon$ is the histogram bin's width. Taking the square root after normalizing $\mat{H}$ increases the discriminatory ability of our projected histogram feature \cite{CCC, arandjelovic2012three}.

Lastly, the histogram is normalized to represent the color distribution in the new chrominance coordinates ($u_c$, $v_c$).

\begin{figure}[!t]
\includegraphics[width=\linewidth]{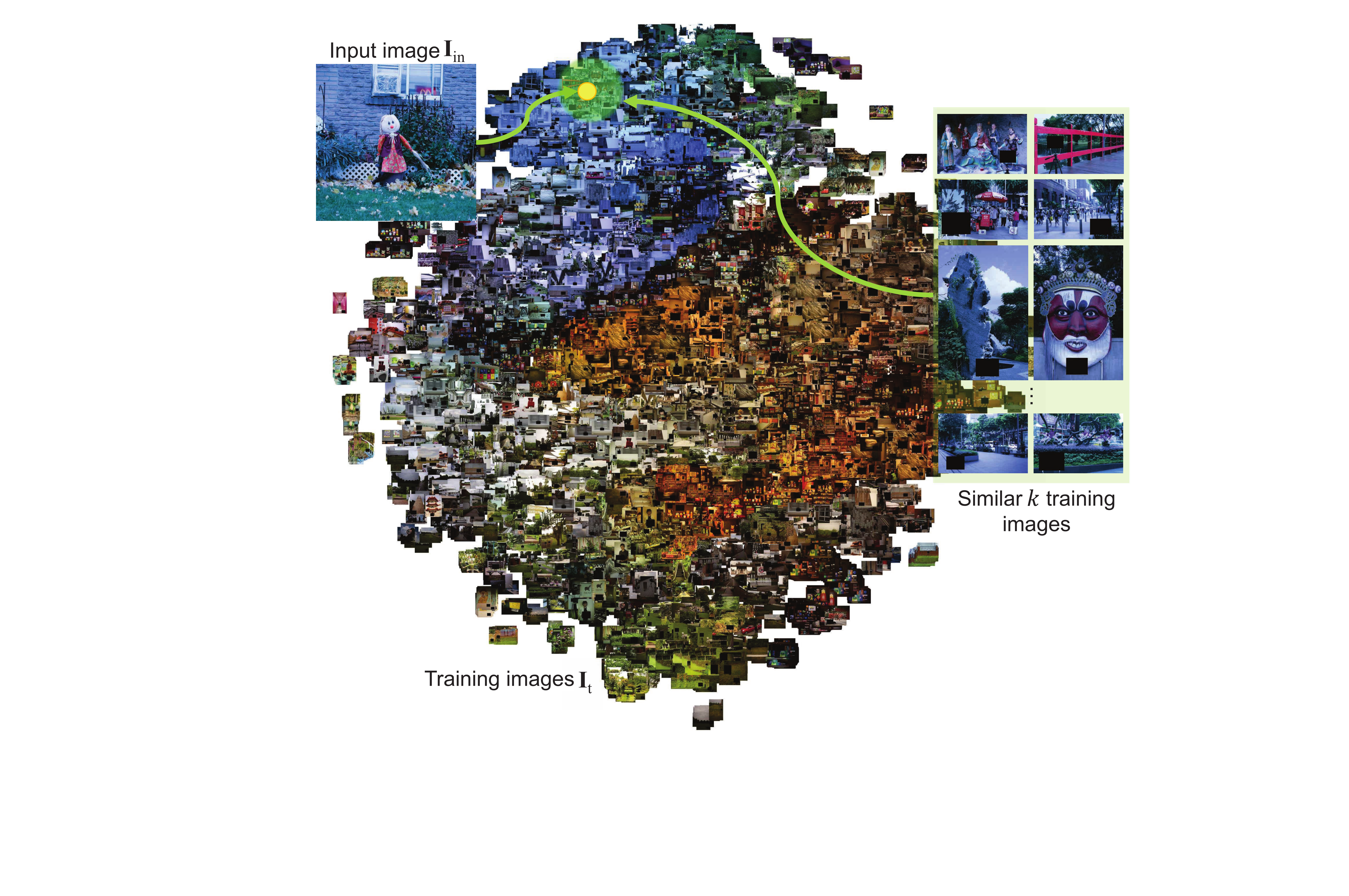}
\vspace{-7mm}
\caption[Visualization of the training images based on their corresponding PCA feature vectors.]{Visualization of the training images based on their corresponding PCA feature vectors. In this figure t-SNE \cite{maaten2008visualizing} is used to aid visualization of the training space. Shown is an example input image and several of the nearest images retrieved using the PCA feature.}
\label{KNN:fig:similarity}
\end{figure}

For the sake of efficiency, we apply a dimensionality reduction step in order to extract a compact feature representing each RGB-$uv$ histogram. We found that the PCA linear transformation is adequate for our task to map the vectorized histogram $\textrm{vec}(h\left(\mat{I}\right)) \in \mathbb{R}^{m\!\times\!m\!\times\!3}$ to a new lower-dimensional space. The PCA feature vector is computed as follows:
\begin{equation}
\label{KNN:eq_PCA}
v(\mat{I}) = \left(\textrm{vec}\left(h\left(\mat{I}\right)\right) - \mat{b}\right)^T \mat{W},
\end{equation}
\noindent where $v\left(\mat{I}\right) \in \mathbb{R}^c$ is the PCA feature vector containing $c$ principal component (PC) scores, $c\ll m\!\times\!m\!\times 3$, $\mat{W} = [\mat{w}_1,\mat{w}_2, ... , \mat{w}_c], \mat{w} \in \mathbb{R}^{m\!\times\!m\!\times 3}$ is the PC coefficient matrix computed by the singular value
decomposition, and $\mat{b} \in \mathbb{R}^{m\!\times\!m\!\times 3}$ is the mean histogram vector. As a result, each training image $\mat{I}_{\textrm{t}}^{(i)}$ can be represented by a small number of PC scores $v\left(\mat{I}_{\textrm{t}}^{(i)}\right)$. The input image is finally represented by $v(\mat{I}_{\textrm{in}})$. The $\textrm{L}_2$ distance is adapted to measure the similarity between the PCA feature vectors.
Figure \ref{KNN:fig:similarity} visualizes the training images based on their corresponding PCA features.

\subsection{Final Color Correction}\label{KNN:sec:finalCorrection}

Given a new input image, we compute its PCA feature and search the training dataset for images with similar features. We extract the set of color correction matrices $\mat{M}_{\textrm{s}}$ associated with the $k$ similar images. The final correction matrix $\mat{M}$ is then computed as a weighted linear combination of the correction matrices $\mat{M}_{\textrm{s}}$ as follows:
\begin{equation} \label{KNN:eq:final_matrix}
\mat{M} = \sum_{j=1}^{k}{\boldsymbol{\alpha}_j \mat{M}^{(j)}_{\textrm{s}}},
\end{equation}
\noindent where $\boldsymbol{\alpha}$ is a weighting vector represented as a RBF:

\begin{equation}
\label{KNN:eq:weighting}
\boldsymbol{\alpha}_j = \frac{\exp\left(-\mat{d}_j^2/2\sigma^2\right)}{\sum_{k^{'}=1}^{k}\exp\left(-\mat{d}_{k^{'}}^2/2\sigma^2\right)}, \text{ } j\in[1,...,k],
\end{equation}

\noindent where $\sigma$ is the radial fall-off factor used in Eq.~\ref{KNN:eq:weighting} and $\mat{d}$ represents a vector containing the $\textrm{L}_2$ distance between the given input feature and the similar $k$ training features.

As shown in Fig. \ref{KNN:fig:overview}, the final color transformation is generated based on correction transformations associated with training images taken from different cameras and render styles.  By blending the mapping functions from images produced by a wide range of different cameras and their different photo-finishing styles, we can interpret $\mat{M}$ correction as mapping the input image to a {\it meta-camera}'s output composed from the most similar images to the input.

Lastly, the corrected image $\mat{I}_{\textrm{corr}}$ is produced by the following equation:
\begin{equation}
\label{KNN:eq:correction}
\mat{I}_{\textrm{corr}} = \mat{M} \textrm{  }\Phi\left(\mat{I}_{\textrm{in}}\right).
\end{equation}

\subsection{Implementation Details}\label{KNN:res:imp}

Our Matlab implementation requires approximately 0.54 seconds to compute the histogram feature. Once the PCA histogram feature is computed, the correction process takes an average of 0.73 seconds; this process includes the PCA feature extraction, the brute-force search of the $k$ nearest neighbors, blending the correction matrix, and the final image correction. All the reported runtimes were computed on an Intel$^\circledR$ Xeon$^\circledR$ E5-1607 @ 3.10 GHz machine and for a 12 mega-pixel image. The accelerated GPU implementation runs on average in 0.12 seconds using GTX 1080 GPU.

Our method requires 23.3 MB to store 62,535 feature vectors, mapping matrices, the PCA coefficient matrix, and the mean histogram vector using single-precision floating-point representation without affecting the accuracy. 

In our implementation, each PCA feature vector was represented by 55 PC scores (i.e., $c=55$), the PC coefficient matrix $\mat{W}$ was represented as a $(60\!\times\!60\!\times 3)\!\times\!55$ matrix (i.e., $m=60$), and the mean vector $\mat{b} \in \mat{R}^{60\!\times\!60\!\times 3}$. We used a fall-off factor $\sigma=0.25$ and $k=25$.

\section{Experimental Results}

\begin{table}[!t]
\centering
\caption[Camera models used in the proposed dataset.]{Camera models used in the proposed dataset. The intrinsic set (Set 1) comprises 62,535 sRGB images (48.7 GB) for seven different cameras. The extrinsic set (Set 2) comprises 2,881 sRGB images (5.43 GB) for one DSLR camera and four different mobile phone cameras. For each image in the dataset, there is a corresponding ground truth sRGB image rendered with a correct WB in Adobe Standard.}
\vspace{2mm}
\label{KNN:Table:dataset}
\scalebox{0.75}
{\begin{tabular}{c|c|c|c|c|c|c|c|c|}
\cline{2-9}
\multicolumn{1}{l|}{} & \multicolumn{8}{c|}{\cellcolor[HTML]{B7E589}\textbf{Intrinsic set (Set 1)}} \\ \hline
\multicolumn{1}{|c|}{\textbf{Camera}} & \begin{tabular}[c]{@{}c@{}}Canon \\ EOS-1Ds\\  Mark III\end{tabular} & \begin{tabular}[c]{@{}c@{}}Canon \\ EOS 600D\end{tabular} & \begin{tabular}[c]{@{}c@{}}Fujifilm \\ X-M1\end{tabular} & \begin{tabular}[c]{@{}c@{}}Nikon \\ D40\end{tabular} & \begin{tabular}[c]{@{}c@{}}Nikon \\ D5200\end{tabular} & \begin{tabular}[c]{@{}c@{}}Canon \\ 1D\end{tabular} & \begin{tabular}[c]{@{}c@{}}Canon \\ 5D\end{tabular} & Total \\ \hline
\multicolumn{1}{|c|}{\textbf{\# of images}} & 10,721 & 9,040 & 5,884 & 10,826 & 8,953 & 2,284 & 14,827 & 62,535 \\ \hline
\multicolumn{1}{|c|}{\textbf{Size}} & 11.00 GB & 8.27 GB & 4.78 GB & 3.4 GB & 10.3 GB & 1.27 GB & 9.68 GB & 48.7 GB \\ \hline
\multicolumn{1}{l|}{} & \multicolumn{8}{c|}{\cellcolor[HTML]{B7E589}\textbf{Extrinsic set (Set 2)}} \\ \hline
\multicolumn{1}{|c|}{\textbf{Camera}} & \multicolumn{2}{c|}{Olympus E-PL6} & \multicolumn{5}{c|}{\begin{tabular}[c]{@{}c@{}}Mobile phone cameras: Galaxy S6 Edge,\\ iPhone 7, LG G4, and Google Pixel\end{tabular}} & Total \\ \hline
\multicolumn{1}{|c|}{\textbf{\# of images}} & \multicolumn{2}{c|}{1,874} & \multicolumn{5}{c|}{1,007} & 2,881 \\ \hline
\multicolumn{1}{|c|}{\textbf{Size}} & \multicolumn{2}{c|}{3.5 GB} & \multicolumn{5}{c|}{1.93 GB} & 5.43 GB \\ \hline

\end{tabular}}
\end{table}

\subsection{Proposed Dataset}
\label{KNN:sub:dataset}

As described in Sec. \ref{KNN:sec:datageneration}, we have generated a dataset of 65,416 sRGB-rendered images that were divided into two sets: intrinsic set (Set 1) and extrinsic set (Set 2). Table \ref{KNN:Table:dataset} shows more details of the camera makes and models used for each set. The size of the original dataset is $\sim$1.14 TB, which was down-sampled by bicubic interpolation to 48.7 GB. For Set 1, the average image width and height are 890.1 and 687.2 pixels, respectively. For Set 2, the average width and height are 1,219.5 and 1,129.9 pixels, respectively.

For both sets, we have used the following WB presets:  Incandescent, Fluorescent, Daylight, Cloudy, and Shade. The corresponding color temperatures of these presets are: 2850 Kelvin (K), 3800K, 5500K, 6500K, and 7500K. For Set 1, we have used the following camera picture styles: Adobe Standard, Faithful, Landscape, Neutral, Portrait, Standard, Vivid, Soft, D2X (mode 1, 2, and 3), and ACR (4.4 and 3.7). For Set 2, we have used the following camera picture styles: (for the Olympus camera) Muted, Portrait, Vivid, Adobe Standard, and (for the mobile cameras) the camera's ``embedded style''.

\begin{table}[!t]

\centering
\caption[Different kernel functions used to study the most suitable color correction matrix for our problem.]{Different kernel functions used to study the most suitable color correction matrix for our problem. The first column represents the dimensions of the output vector of the corresponding kernel function in the second column. The terms PCC and RPC stand for polynomial color correction and root polynomial color correction, respectively.}
\label{KNN:sub:Tabel1}
\vspace{2mm}
\scalebox{0.68}
{
\begin{tabular}{|l|l|}
\hline
\multicolumn{1}{|c|}{\textbf{Dimensions}} & \multicolumn{1}{c|}{\textbf{Kernel function output}} \\ \hline
3 (linear) & $[$R, G, B$]^T$ (identity) \\ \hline
9 (PCC) \cite{finlayson2015color}& $[$R, G, B, $\textrm{R}^2$, $\textrm{G}^2$, $\textrm{B}^2$, RG, RB, GB$]^T$ \\ \hline
11 (PCC) \cite{hong2001study}& $[$R, G, B, RG, RB, GB, $\textrm{R}^2$, $\textrm{G}^2$, $\textrm{B}^2$, RGB,1$]^T$ \\ \hline
19 (PCC) \cite{finlayson2015color}& \begin{tabular}[c]{@{}l@{}}$[$R, G, B, RG, RB, GB, $\textrm{R}^2$, $\textrm{G}^2$, $\textrm{B}^2$, $\textrm{R}^3$, $\textrm{G}^3$, $\textrm{B}^3$, $\textrm{RG}^2$,  $\textrm{RB}^2$, $\textrm{GB}^2, \textrm{GR}^2$, $\textrm{BG}^2$, $\textrm{BR}^2$, RGB$]^T$\end{tabular} \\ \hline
34 (PCC) \cite{finlayson2015color}& \begin{tabular}[c]{@{}l@{}}$[$R, G, B, RG, RB, GB, $\textrm{R}^2$, $\textrm{G}^2$, $\textrm{B}^2$, $\textrm{R}^3$, $\textrm{G}^3$, $\textrm{B}^3$, $\textrm{RG}^2$, $\textrm{RB}^2$, $\textrm{GB}^2, \textrm{GR}^2$, $\textrm{BG}^2$, $\textrm{BR}^2,$ RGB, $\textrm{R}^4$,  $\textrm{G}^4$, $\textrm{B}^4$, $\textrm{R}^3\textrm{G}$,\\ $\textrm{R}^3\textrm{B}$, $\textrm{G}^3\textrm{R}$, $\textrm{G}^3\textrm{B}$, $\textrm{B}^3\textrm{R}$, $\textrm{B}^3\textrm{G}$, $\textrm{R}^2\textrm{G}^2$, $\textrm{G}^2\textrm{B}^2$, $\textrm{R}^2\textrm{B}^2$, $\textrm{R}^2\textrm{GB}$, $\textrm{G}^2\textrm{RB}$,
$\textrm{B}^2\textrm{RB}]^T$\end{tabular} \\ \hline
6 (RPC) \cite{finlayson2015color}& $[$R, G, B, $\sqrt{\textrm{RG}}$, $\sqrt{\textrm{GB}}$, $\sqrt{\textrm{RB}}]^T$ \\ \hline
13 (RPC) \cite{finlayson2015color}& \begin{tabular}[c]{@{}l@{}}$[$R, G, B, $\sqrt{\textrm{RG}}$, $\sqrt{\textrm{GB}}$, $\sqrt{\textrm{RB}}$, $\sqrt[3]{\textrm{RG}^2}$, $\sqrt[3]{\textrm{RB}^2}$,  $\sqrt[3]{\textrm{GB}^2}$, $\sqrt[3]{\textrm{GR}^2}$, $\sqrt[3]{\textrm{BG}^2}$, $\sqrt{\textrm{BR}^2}$, $\sqrt[3]{\textrm{RGB}}]^T$\end{tabular} \\ \hline
22 (RPC) \cite{finlayson2015color}& \begin{tabular}[c]{@{}l@{}}$[$R, G, B, $\sqrt{\textrm{RG}}$, $\sqrt{\textrm{GB}}$, $\sqrt{\textrm{RB}}$, $\sqrt[3]{\textrm{RG}^2}$, $\sqrt[3]{\textrm{RB}^2}$,  $\sqrt[3]{\textrm{GB}^2}$, $\sqrt[3]{\textrm{GR}^2}$, $\sqrt[3]{\textrm{BG}^2}$, $\sqrt{\textrm{BR}^2}$, $\sqrt[3]{\textrm{RGB}}, \sqrt[4]{\textrm{R}^3\textrm{G}}, \sqrt[4]{\textrm{R}^3\textrm{B}},$ 
$\sqrt[4]{\textrm{G}^3\textrm{R}}$,\\ $\sqrt[4]{\textrm{G}^3\textrm{B}}$,
$\sqrt[4]{\textrm{B}^3\textrm{R}}$, $\sqrt[4]{\textrm{B}^3\textrm{G}}$,
$\sqrt[4]{\textrm{R}^2\textrm{GB}}$, $\sqrt[4]{\textrm{G}^2\textrm{RB}}$, $\sqrt[4]{\textrm{B}^2\textrm{RG}}
]^T$\end{tabular} \\ \hline
\end{tabular}
}
\end{table}

\subsection{Hyperparameters Selection}
\label{KNN:sub:correction_matrices}

In order to select the number of nearest neighbors, $k$, and the most appropriate color correction transform, we have evaluated the accuracy of different color correction approaches between an incorrectly white-balanced camera-rendered image and its correctly white-balanced camera-rendered target image.

This study was conducted for quality assessment rather than performance. Accordingly, we have used the RGB-$uv$ histogram features with bandwidth $m=180$ without dimensionality reduction applied.

By taking the square root after normalizing $\mat{H}$, it makes the classic Euclidean $\textrm{L}_2$ distance applicable as a symmetric similarity metric to measure the similarity between two distributions \cite{arandjelovic2012three}. Consistently with the PCA feature similarity measurement, the Hellinger distance \cite{pollard2002user} was used as a similarity metric in this study. The Hellinger distance~\cite{pollard2002user} between two histograms $h\left(\mat{I}_1\right)$ and $h\left(\mat{I}_2\right)$ can be represented as $\left(1/\sqrt{2}\right)\textrm{ L}_2\left(h\left(\mat{I}_1\right), h\left(\mat{I}_2\right)\right)$.

We tested 8 different color correction matrices on Set 1 using three-fold validation. The matrices are: $3\!\times\!3$ full color correction matrices, $3\!\times\!9$, $3\!\times\!11$, $3\!\times\!19$, $3\!\times\!34$ polynomial color correction (PCC) matrices, and $3\!\times\!6$, $3\!\times\!13$, $3\!\times\!22$ root polynomial color correction (RPC) matrices \cite{hong2001study, finlayson2015color}. For each color correction matrix, we have tested different values of $k$. Note that the $3\!\times\!3$ color correction matrix is computed using Eq. \ref{KNN:eq1_M} with $\Phi\left(\mat{I}^{(i)}_{\textrm{t}}\right) = \mat{I}^{(i)}_{\textrm{t}}$---that is, the kernel function is an identity function.

\begin{figure}[!h]
\includegraphics[width=\linewidth]{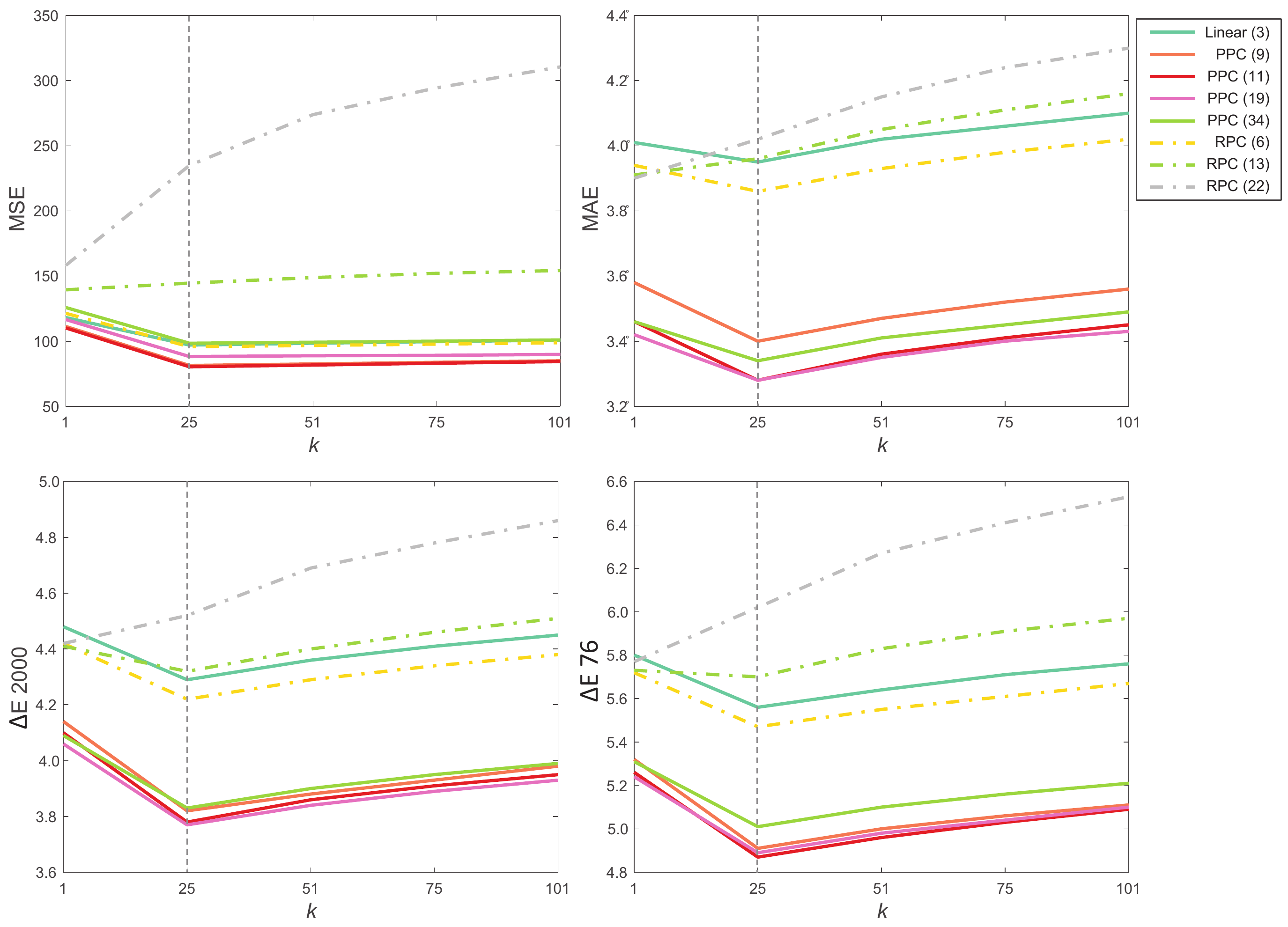}
\vspace{-7mm}
\caption[A study of the accuracy obtained using different color correction matrices.]{A study of the accuracy obtained using different color correction matrices, which are: (i) linear $3\!\times\!3$ full matrix, (ii) $3\!\times\!9$, (iii) $3\!\times\!11$, (iv) $3\!\times\!19$, (v) $3\!\times\!34$ polynomial color correction (PCC) matrices \cite{hong2001study, finlayson2015color}, (vi) $3\!\times\!6$, (vii) $3\!\times\!13$, and (viii) $3\!\times\!22$ root-polynomial color correction (RPC) matrices \cite{finlayson2015color}. The horizontal axis represents the number of nearest neighbors, $k$, and the vertical axis represents the error between the corrected images and the ground truth images using different error metrics.}
\label{KNN:fig:sub_mapping_k}
\end{figure}

Table \ref{KNN:sub:Tabel1} shows the kernel functions used to generate each color correction matrix. Figure \ref{KNN:fig:sub_mapping_k} shows the results obtained using  four different error metrics.
As shown, the $3\!\times\!11$ color correction matrix, described by Hong et al.~\cite{hong2001study}, provided the best results for our task. Also, it is shown that the accuracy increases as the value of $k$ increases. At a certain point, however, increasing $k$ negatively affects the accuracy.

In this set of experiments, adopting the RGB-$uv$ histogram features requires approximately 14.9 GB of memory having $\sim$41K training samples represented as single-precision floating-point values, and runs in 44.6 seconds to correct a 12 mega-pixel image on average. This process includes the RGB-$uv$ histogram feature extraction, the brute-force search of the $k$ nearest neighbors, blending the correction matrix, and the final image correction. In Sec. \ref{KNN:sec:similaritymeasurement}, we extract a compact feature representing each RGB-$uv$ histogram. This compact representation improves the performance (requiring less than 1.5 seconds on CPU to correct a 12 mega-pixel image) and achieves on-par accuracy compared to employing the original RGB-$uv$ histogram features.

\begin{table}[!t]
\caption[Comparisons between our method with  diagonal WB correction using an exact achromatic reference point.]{Comparisons between our method with  diagonal WB correction using an {\it exact achromatic} reference point (ER). We also show results obtained by different illuminant estimation methods. The diagonal correction is applied directly to the sRGB images, denoted as (sRGB) and ``linearized'' RGB \cite{anderson1996proposal, ebner2007color}, denoted as (LRGB). The terms Q1, Q2, and Q3 denote the first, second (median), and third quartile, respectively. The terms MSE and MAE stand for mean square error and mean angular error, respectively. The top results are indicated with yellow and boldface.}
\label{KNN:sTable0}
\centering
\scalebox{0.58}{
\begin{tabular}{|l|c|c|c|c|c|c|c|c|c|c|c|c|}
\hline
\multicolumn{1}{|c|}{} & \multicolumn{4}{c|}{\textbf{MSE}} & \multicolumn{4}{c|}{\textbf{MAE}} & \multicolumn{4}{c|}{\textbf{$\boldsymbol{\bigtriangleup}$\textbf{E} 2000}} \\ \cline{2-13}
\multicolumn{1}{|c|}{\multirow{-2}{*}{\textbf{Method}}} & \textbf{Mean} & \textbf{Q1} & \textbf{Q2} & \textbf{Q3} & \textbf{Mean} & \textbf{Q1} & \textbf{Q2} & \textbf{Q3} & \textbf{Mean} & \textbf{Q1} & \textbf{Q2} & \textbf{Q3}  \\ \hline

\multicolumn{13}{|c|}{\cellcolor[HTML]{B7E589}\textbf{Intrinsic set (Set 1): DSLR multiple cameras (62,535 images)}} \\ \hline

GW (sRGB)\cite{GW} &  282.76 & 70.50 & 180.81 & 380.89 & 7.23\textdegree & 4.14\textdegree & 6.40\textdegree & 9.51\textdegree & 7.99 & 5.08 & 7.47 & 10.40  \\ \hline

GW (LRGB) \cite{GW} &  285.51 & 73.37 & 184.72 & 384.67 & 7.97\textdegree & 4.36\textdegree & 6.91\textdegree & 10.53\textdegree & 8.48 & 5.38 & 7.91 & 11.03\\ \hline

GE-1 (sRGB) \cite{GE} &  193.99 & 43.38 & 119.93 & 267.84 & 6.81\textdegree & 3.22\textdegree & 5.49\textdegree & 9.31\textdegree & 6.86 & 3.92 & 6.30 & 9.17\\ \hline

GE-1 (LRGB) \cite{GE} & 190.59 & 42.51 & 116.75 & 261.6 & 6.54\textdegree & 3.22\textdegree & 5.37\textdegree & 8.78\textdegree & 6.82 & 3.91 & 6.20 & 9.06 \\ \hline

GE-2 (sRGB) \cite{GE} & 208.20 & 44.82 & 126.83 & 285.31 & 7.03\textdegree & 3.30\textdegree & 5.67\textdegree & 9.62\textdegree & 7.06 & 4.01 & 6.44 & 9.44\\ \hline

GE-2 (LRGB) \cite{GE} & 204.93 & 43.86 & 123.19 & 279.15 & 6.76\textdegree & 3.31\textdegree & 5.54\textdegree & 9.12\textdegree & 7.02 & 4.01 & 6.35 & 9.32\\ \hline

wGE (sRGB) \cite{gijsenij2012improving} & 225.87 & 42.76 & 122.39 & 300.48 & 7.15\textdegree & 3.20\textdegree & 5.60\textdegree & 9.69\textdegree & 7.07 & 3.78 & 6.29 & 9.50\\ \hline

wGE (LRGB) \cite{gijsenij2012improving} & 223.30 & 42.40 & 119.47 & 294.22 & 6.90\textdegree & 3.17\textdegree & 5.42\textdegree & 9.26\textdegree & 7.05 & 3.76 & 6.21 & 9.43\\ \hline

max-RGB (sRGB) \cite{maxRGB} & 285.07 & 47.20 & 160.6 & 397.79 & 8.48\textdegree & 3.49\textdegree & 7.03\textdegree & 12.35\textdegree & 8.05 & 3.97 & 7.25 & 11.34\\ \hline

max-RGB (LRGB)  \cite{maxRGB} & 280.86 & 46.21 & 156.09 & 390.99 & 8.17\textdegree & 3.41\textdegree & 6.76\textdegree & 11.83\textdegree & 7.96 & 3.92 & 7.12 & 11.21\\ \hline

SoG (sRGB) \cite{SoG} & 171.30 & 38.31 & 104.85 & 235.57 & 6.06\textdegree & 2.99\textdegree & 5.08\textdegree & 8.21\textdegree & 6.19 & 3.52 & 5.72 & 8.25 \\ \hline

SoG (LRGB) \cite{SoG} & 169.33 & 38.10 & 102.21 & 231.50  & 5.96\textdegree & 3.03\textdegree & 5.03\textdegree & 7.91\textdegree & 6.24 & 3.56 & 5.71 & 8.27 \\ \hline

FC4 (sRGB) \cite{hu2017fc}   & 426.31 & 118.30 & 282.05 & 561.92 & 7.91\textdegree & 4.57\textdegree &  7.33\textdegree & 10.38\textdegree & 9.78 & 6.12 & 9.14 &  12.65 \\ \hline

FC4 (LRGB) \cite{hu2017fc} & 179.55 & 33.89 & 100.09 & 246.50 & 6.14\textdegree & 2.62\textdegree & 4.73\textdegree & 8.40\textdegree &  6.55 & 3.54 & 5.90 & 8.94 \\ \hline

ER (sRGB) & 135.77 & 20.20 & 71.74 & 196.15 & 4.63\textdegree & 1.99\textdegree & 3.56\textdegree & 6.14\textdegree & 4.69 & 2.25 & 4.00 & 6.68 \\ \hline

ER (LRGB) &  130.01 & 19.73 & 68.54 & 183.65 & 4.29\textdegree & 1.85\textdegree & 3.35\textdegree & 5.70\textdegree & 4.59 &  2.24 &  3.89 & 6.51\\ \hline

Ours &  \cellcolor[HTML]{\bestcolor} \textbf{77.79} &  \cellcolor[HTML]{\bestcolor} \textbf{13.74} &  \cellcolor[HTML]{\bestcolor} \textbf{39.62} &  \cellcolor[HTML]{\bestcolor} \textbf{94.01} &  \cellcolor[HTML]{\bestcolor} \textbf{3.06\textdegree} &  \cellcolor[HTML]{\bestcolor} \textbf{1.74\textdegree} &  \cellcolor[HTML]{\bestcolor} \textbf{2.54\textdegree} &  \cellcolor[HTML]{\bestcolor} \textbf{3.76\textdegree} &  \cellcolor[HTML]{\bestcolor} \textbf{3.58} &  \cellcolor[HTML]{\bestcolor} \textbf{2.07} &  \cellcolor[HTML]{\bestcolor} \textbf{3.09} &  \cellcolor[HTML]{\bestcolor} \textbf{4.55} \\ \hline

\end{tabular}
}
\end{table}

\begin{table}[!t]
\caption[Comparisons between our method with  diagonal WB correction using an exact achromatic reference point.]{Comparisons between our method with  diagonal WB correction using an {\it exact achromatic} reference point (ER). We also show results obtained by different illuminant estimation methods. The diagonal correction is applied directly to the sRGB images, denoted as (sRGB) and ``linearized'' RGB \cite{anderson1996proposal, ebner2007color}, denoted as (LRGB). The terms Q1, Q2, and Q3 denote the first, second (median), and third quartile, respectively. The terms MSE and MAE stand for mean square error and mean angular error, respectively. The top results are indicated with yellow and boldface.}
\label{KNN:sTable01}
\centering
\scalebox{0.58}{
\begin{tabular}{|l|c|c|c|c|c|c|c|c|c|c|c|c|}
\hline
\multicolumn{1}{|c|}{} & \multicolumn{4}{c|}{\textbf{MSE}} & \multicolumn{4}{c|}{\textbf{MAE}} & \multicolumn{4}{c|}{\textbf{$\boldsymbol{\bigtriangleup}$\textbf{E} 2000}} \\ \cline{2-13}
\multicolumn{1}{|c|}{\multirow{-2}{*}{\textbf{Method}}} & \textbf{Mean} & \textbf{Q1} & \textbf{Q2} & \textbf{Q3} & \textbf{Mean} & \textbf{Q1} & \textbf{Q2} & \textbf{Q3} & \textbf{Mean} & \textbf{Q1} & \textbf{Q2} & \textbf{Q3}  \\ \hline
\multicolumn{13}{|c|}{\cellcolor[HTML]{B7E589}\textbf{Extrinsic set (Set 2): DSLR and mobile phone cameras (2,881 images)}} \\ \hline

GW (sRGB)\cite{GW} & 500.18 & 173.69 & 332.75 & 615.40 & 8.89\textdegree & 5.82\textdegree & 8.32\textdegree & 11.33\textdegree & 10.74 & 7.92 & 10.29 & 13.12\\ \hline

GW (LRGB) \cite{GW} & 469.86 & 163.07 & 312.28 & 574.85 & 8.61\textdegree & 5.44\textdegree & 7.94\textdegree & 10.93\textdegree & 10.68 & 7.70 & 10.13 & 13.15\\ \hline

GE-1 (sRGB) \cite{GE} & 791.10 & 235.37 & 524.94 & 1052.38 & 12.90\textdegree & 7.98\textdegree & 12.41\textdegree & 17.50\textdegree & 13.09 & 9.17 & 12.98 & 16.68\\ \hline

GE-1 (LRGB) \cite{GE} & 779.27 & 225.36 & 510.03 & 1038.71 & 12.55\textdegree & 7.55\textdegree & 11.87\textdegree & 17.00\textdegree & 12.98 & 9.06 & 12.86 & 16.57\\ \hline

GE-2 (sRGB) \cite{GE} & 841.83 & 239.58 & 542.07 & 1114.86 & 13.16\textdegree & 7.94\textdegree & 12.55\textdegree & 17.76\textdegree & 13.31 & 9.20 & 13.20 & 17.09\\ \hline

GE-2 (LRGB) \cite{GE} & 831.01 & 231.42 & 530.52 & 1099.75 & 12.84\textdegree & 7.64\textdegree & 12.13\textdegree & 17.45\textdegree & 13.22 & 9.00 & 13.13 & 17.01\\ \hline

wGE (sRGB) \cite{gijsenij2012improving} & 999.95 & 236.46 & 587.55 & 1350.41 & 13.80\textdegree & 8.08\textdegree & 12.99\textdegree & 18.80\textdegree & 14.05 & 9.13 & 13.80 & 18.56\\ \hline

wGE (LRGB) \cite{gijsenij2012improving} & 990.20 & 230.38 & 577.62 & 1345.52 & 13.52\textdegree & 7.76\textdegree & 12.62\textdegree & 18.56\textdegree & 14.00 & 9.00 & 13.70 & 18.56\\ \hline

max-RGB (sRGB) \cite{maxRGB} & 791.99 & 263.00 & 572.23 & 1087.14 & 13.47\textdegree & 8.44\textdegree & 12.93\textdegree & 18.50\textdegree & 13.01 & 9.12 & 13.44 & 17.15\\ \hline

max-RGB (LRGB)  \cite{maxRGB} & 780.63 & 256.40 & 560.58 & 1073.22 &  13.18\textdegree &  8.15\textdegree &  12.57\textdegree & 18.12\textdegree & 12.93 & 9.02 & 13.36 & 17.08\\ \hline

SoG (sRGB) \cite{SoG} & 429.35 & 147.05 & 286.84 & 535.72 & 9.54\textdegree & 5.72\textdegree & 8.85\textdegree & 12.65\textdegree & 10.01 & 7.09 & 9.85 & 12.69\\ \hline

SoG (LRGB) \cite{SoG} & 393.85 & 137.21 & 267.37 & 497.40 & 8.96\textdegree & 5.31\textdegree & 8.26\textdegree & 11.97\textdegree & 9.81 & 6.87 & 9.67 & 12.46\\ \hline

FC4 (sRGB) \cite{hu2017fc} & 662.53  & 304.88 & 524.42 & 817.57 & 8.92\textdegree & 5.94\textdegree & 8.03\textdegree & 10.84\textdegree & 12.12 &  8.94& 11.79 &  14.76 \\\hline

FC4 (LRGB) \cite{hu2017fc}  &  505.30 & 142.46 & 307.77 & 635.35 & 10.37\textdegree
& 5.31\textdegree & 9.26\textdegree & 14.15\textdegree & 10.82 & 7.39 & , 10.64 &  13.77   \\ \hline

ER (sRGB) & 422.31 & 110.70 & 257.76 & 526.16 & 7.99\textdegree & 4.36\textdegree & 7.11\textdegree & 10.57\textdegree & 8.53 & 5.52 & 8.38 & 11.11  \\ \hline
ER (LRGB) & 385.23 & 99.05 & 230.86 & 475.72 & 7.22\textdegree & 3.80\textdegree
  & 6.34\textdegree & 9.54\textdegree & 8.15 & 5.07 & 7.88 & 10.68  \\ \hline

Ours & \cellcolor[HTML]{\bestcolor} \textbf{171.09} &  \cellcolor[HTML]{\bestcolor} \textbf{37.04} &  \cellcolor[HTML]{\bestcolor} \textbf{87.04} &  \cellcolor[HTML]{\bestcolor} \textbf{190.88} &  \cellcolor[HTML]{\bestcolor} \textbf{4.48\textdegree} &  \cellcolor[HTML]{\bestcolor} \textbf{2.26\textdegree} &  \cellcolor[HTML]{\bestcolor} \textbf{3.64\textdegree} &  \cellcolor[HTML]{\bestcolor} \textbf{5.95\textdegree} &  \cellcolor[HTML]{\bestcolor} \textbf{5.60} &  \cellcolor[HTML]{\bestcolor} \textbf{3.43} &  \cellcolor[HTML]{\bestcolor} \textbf{4.90} &  \cellcolor[HTML]{\bestcolor} \textbf{7.06}  \\ \hline
\end{tabular}
}
\end{table}

\begin{table}[t]

\caption[Comparisons between our method with the Adobe Photoshop functions: auto-color and auto-tone.]{Comparisons between our method with the Adobe Photoshop functions: auto-color (AC) and auto-tone (AT). The terms Q1, Q2, and Q3 denote the first, second (median), and third quartile, respectively. The terms MSE and MAE stand for mean square error and mean angular error, respectively. The top results are indicated with yellow and boldface.}

\label{KNN:Table1}
\centering
\scalebox{0.7}{
\begin{tabular}{|l|c|c|c|c|c|c|c|c|c|c|c|c|}
\hline
\multicolumn{1}{|c|}{} & \multicolumn{4}{c|}{\textbf{MSE}} & \multicolumn{4}{c|}{\textbf{MAE}} & \multicolumn{4}{c|}{\textbf{$\boldsymbol{\bigtriangleup}$\textbf{E}}} \\ \cline{2-13}

\multicolumn{1}{|c|}{\multirow{-2}{*}{\textbf{Method}}} & \textbf{Mean} & \textbf{Q1} & \textbf{Q2} & \textbf{Q3} & \textbf{Mean} & \textbf{Q1} & \textbf{Q2} & \textbf{Q3} & \textbf{Mean} & \textbf{Q1} & \textbf{Q2} & \textbf{Q3}  \\ \hline

\multicolumn{13}{|c|}{\cellcolor[HTML]{B7E589}\textbf{Intrinsic set (Set 1): DSLR multiple cameras (62,535 images)}} \\ \hline
Photoshop-AC & 780.52 & 157.39 & 430.96 & 991.28 & 7.96\textdegree & 3.43\textdegree & 5.59\textdegree & 10.58\textdegree & 10.06 & 5.75 & 8.92 & 13.30\\ \hline
Photoshop-AT & 1002.93 & 238.33 & 606.74 & 1245.51 & 7.56\textdegree & 3.08\textdegree & 5.75\textdegree & 10.83\textdegree & 11.12 & 6.55 & 10.54 & 14.68\\ \hline
Ours &  \cellcolor[HTML]{\bestcolor} \textbf{77.79} &  \cellcolor[HTML]{\bestcolor} \textbf{13.74} &  \cellcolor[HTML]{\bestcolor} \textbf{39.62} &  \cellcolor[HTML]{\bestcolor} \textbf{94.01} &  \cellcolor[HTML]{\bestcolor} \textbf{3.06\textdegree} &  \cellcolor[HTML]{\bestcolor} \textbf{1.74\textdegree} &  \cellcolor[HTML]{\bestcolor} \textbf{2.54\textdegree} &  \cellcolor[HTML]{\bestcolor} \textbf{3.76\textdegree} &  \cellcolor[HTML]{\bestcolor} \textbf{3.58} &  \cellcolor[HTML]{\bestcolor} \textbf{2.07} &  \cellcolor[HTML]{\bestcolor} \textbf{3.09} &  \cellcolor[HTML]{\bestcolor} \textbf{4.55} \\ \hline
\multicolumn{13}{|c|}{\cellcolor[HTML]{B7E589}\textbf{Extrinsic set (Set 2): DSLR and mobile phone cameras (2,881 images)}} \\ \hline
Photoshop-AC & 745.49 & 240.58 & 514.33 & 968.27 & 10.19\textdegree & 5.25\textdegree & 8.60\textdegree & 14.13\textdegree & 11.71 & 7.56 & 11.41 & 15.00 \\ \hline
Photoshop-AT & 953.85 & 386.7 & 743.84 & 1256.94 & 11.91\textdegree & 7.01\textdegree & 10.70\textdegree & 15.92\textdegree & 13.12 & 9.63 & 13.18 & 16.5 \\ \hline

Ours & \cellcolor[HTML]{\bestcolor} \textbf{171.09} &  \cellcolor[HTML]{\bestcolor} \textbf{37.04} &  \cellcolor[HTML]{\bestcolor} \textbf{87.04} &  \cellcolor[HTML]{\bestcolor} \textbf{190.88} &  \cellcolor[HTML]{\bestcolor} \textbf{4.48\textdegree} &  \cellcolor[HTML]{\bestcolor} \textbf{2.26\textdegree} &  \cellcolor[HTML]{\bestcolor} \textbf{3.64\textdegree} &  \cellcolor[HTML]{\bestcolor} \textbf{5.95\textdegree} &  \cellcolor[HTML]{\bestcolor} \textbf{5.60} &  \cellcolor[HTML]{\bestcolor} \textbf{3.43} &  \cellcolor[HTML]{\bestcolor} \textbf{4.90} &  \cellcolor[HTML]{\bestcolor} \textbf{7.06}  \\ \hline
    \end{tabular}
}
\end{table}

\subsection{Results and Comparisons} \label{KNN:subsec:evaluationWBsRGB}

In this section, our proposed approach is compared with common approaches that are currently used to correct improperly white-balanced rendered sRGB images in the proposed  dataset. We report and discuss both quantitative and qualitative results. Failure cases of our algorithm are also shown. Finally, we provide a time analysis of the proposed approach.

We have used Set 1 of our  dataset for training and evaluation using three-fold validation, such that the three folds are disjointed in regards to the imaged scenes, meaning if the scene (i.e., original raw-RGB image) appears in a fold, it is excluded from the other folds. The color rendition chart is masked out in the image and ignored during training and testing.

\subsubsection{Quantitative results}\label{KNN:res:quan}
We compared our results against a diagonal WB correction that is computed using the center gray patch in the color checker chart placed in the scene. We refer to this as the {\it exact achromatic} reference point, as it represents a true neutral point found in the scene. This exact white point represents the best results that an illumination estimation algorithm could achieve when applied to our input in order to determine the diagonal WB matrix.

For the sake of completeness, we also compared our results against a ``linearized'' diagonal correction that applies an inverse gamma operation \cite{anderson1996proposal, ebner2007color}, then performed WB using the exact reference point, and then reapplied the gamma to produce the result in the sRGB color space. We also include results using Adobe Photoshop corrections---specifically, the auto-color function (AC) and auto-tone function (AT).

We also perform comparisons against the diagonal correction using representative examples of illuminant estimation methods---this means the diagonal matrix is automatically computed and not derived from a selected achromatic patch in the scene. As mentioned in Chapter \ref{ch:ch2}, illumination estimation methods are intended to be applied on raw-RGB images; however, in this case we apply it on the sRGB-rendered image. We use five well-known statistical methods for illumination estimation---namely, the GW~\cite{GW}, GE \cite{GE}, wGE \cite{gijsenij2012improving}, max-RGB \cite{maxRGB}, and SoG \cite{SoG}. The Minkowski norm ($p$) was set to 5 for GE, wGE, and SoG. For GE, we computed the results using the first and second differentiations. For each method, we calculated the diagonal correction with and without the pre-linearization process. We could not find learning-based models trained on sRGB images for illuminant estimation except for the fully convolutional color constancy with confidence-weighted pooling (FC4) model \cite{hu2017fc}. Specifically, we use the FC4 trained model on sRGB-rendered images of Gehler dataset~\cite{gehler2008bayesian} provided by the authors. Other learning-based illuminant estimation methods were excluded, since they were trained on the linear RGB space and re-training them in the sRGB space would use the ERs as ground truth illuminants that were already included in our comparisons. Tables \ref{KNN:sTable0} and \ref{KNN:sTable01} shows the obtained results of the exact achromatic reference point correction and other illuminant estimation algorithms. Table \ref{KNN:Table1} shows our results against the Adobe Photoshop functions for color corrections---namely, auto color and auto tone.

Tables \ref{KNN:sTable0}--\ref{KNN:Table1} show that our proposed method consistently outperforms the other approaches in all metrics.

\subsubsection{Qualitative results and User Study}\label{KNN:res:qual}

\begin{figure}[!t]
\includegraphics[width=\linewidth]{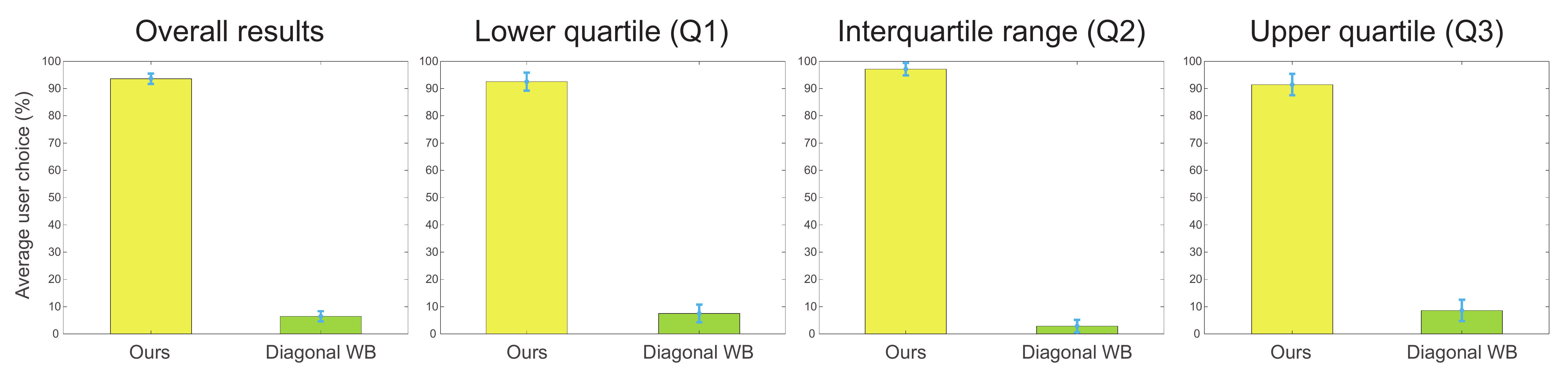}
\vspace{-7mm}
\caption[The results of a user study with 35 people in which users are asked which output is most visually similar to the ground truth image.]{The results of a user study with 35 people in which users are asked which output is most visually similar to the ground truth image.  An equal number of images are selected randomly from the different quartiles. The outcome of the user study is shown via interval plots, with error bars shown at a 99\% confidence interval.}
\label{KNN:fig10}
\end{figure}

\begin{figure}[!t]
\includegraphics[width=\linewidth]{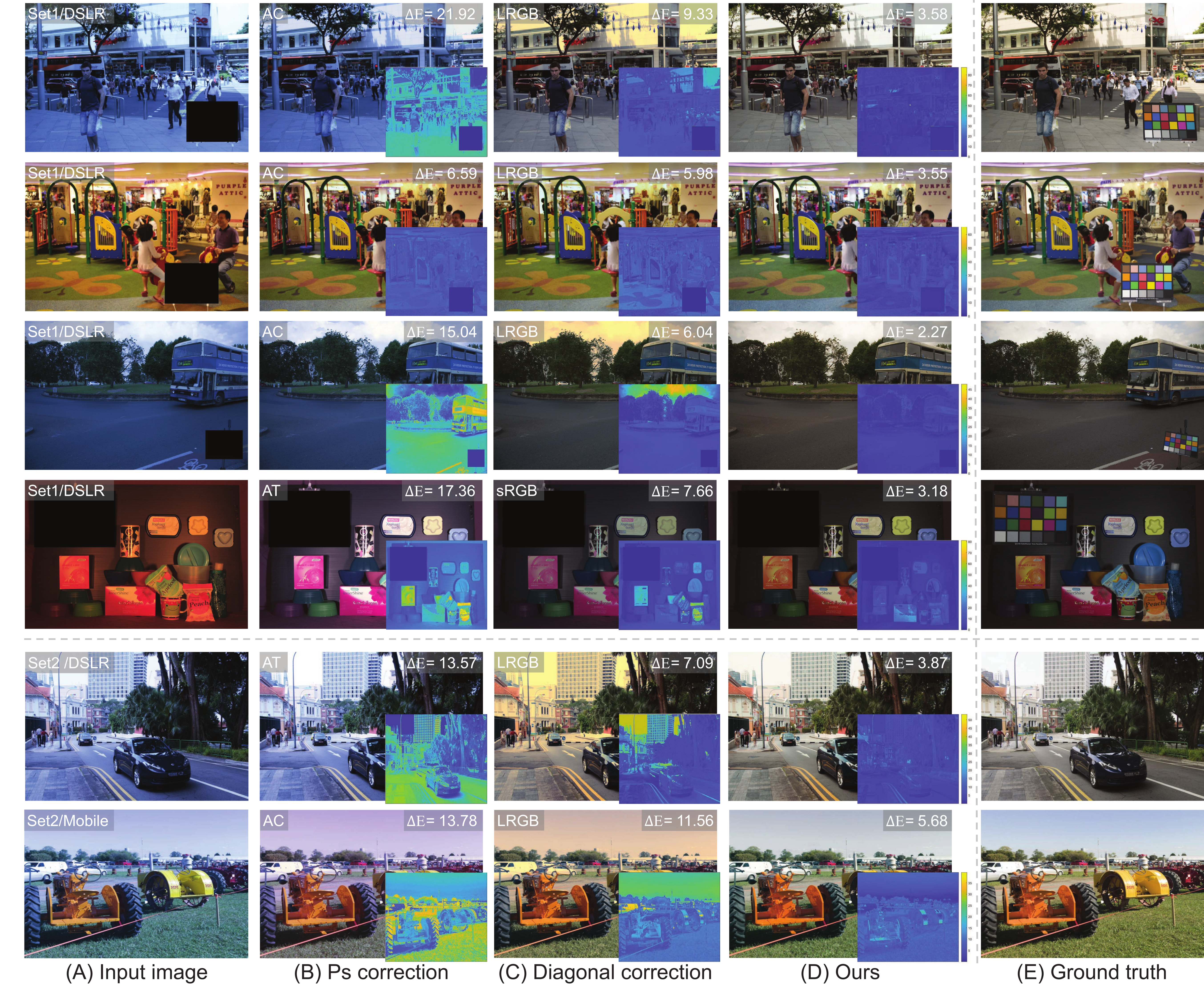}
\vspace{-7mm}
\caption[Comparisons between the proposed approach and other techniques on Set 1 and Set 2.]{Comparisons between the proposed approach and other techniques on \textbf{Set 1} (first four rows) and \textbf{Set 2} (last two rows). (A) Input image in sRGB. (B) Results of Adobe Photoshop (Ps) color correction functions. (C) Results of diagonal correction using the {\it exact} reference point obtained directly from the color chart. (D) Our results. (E) Ground truth images. In (B) and (C) we pick the best result between the auto-color (AC) and auto-tone (AT) functions and between the sRGB (sRGB) and ``linearized'' sRGB (LRGB) \cite{anderson1996proposal} based on $\bigtriangleup$E values, respectively.}
\label{KNN:fig:set1_2_visual_results}
\end{figure}

\begin{figure}[!t]
\includegraphics[width=\linewidth]{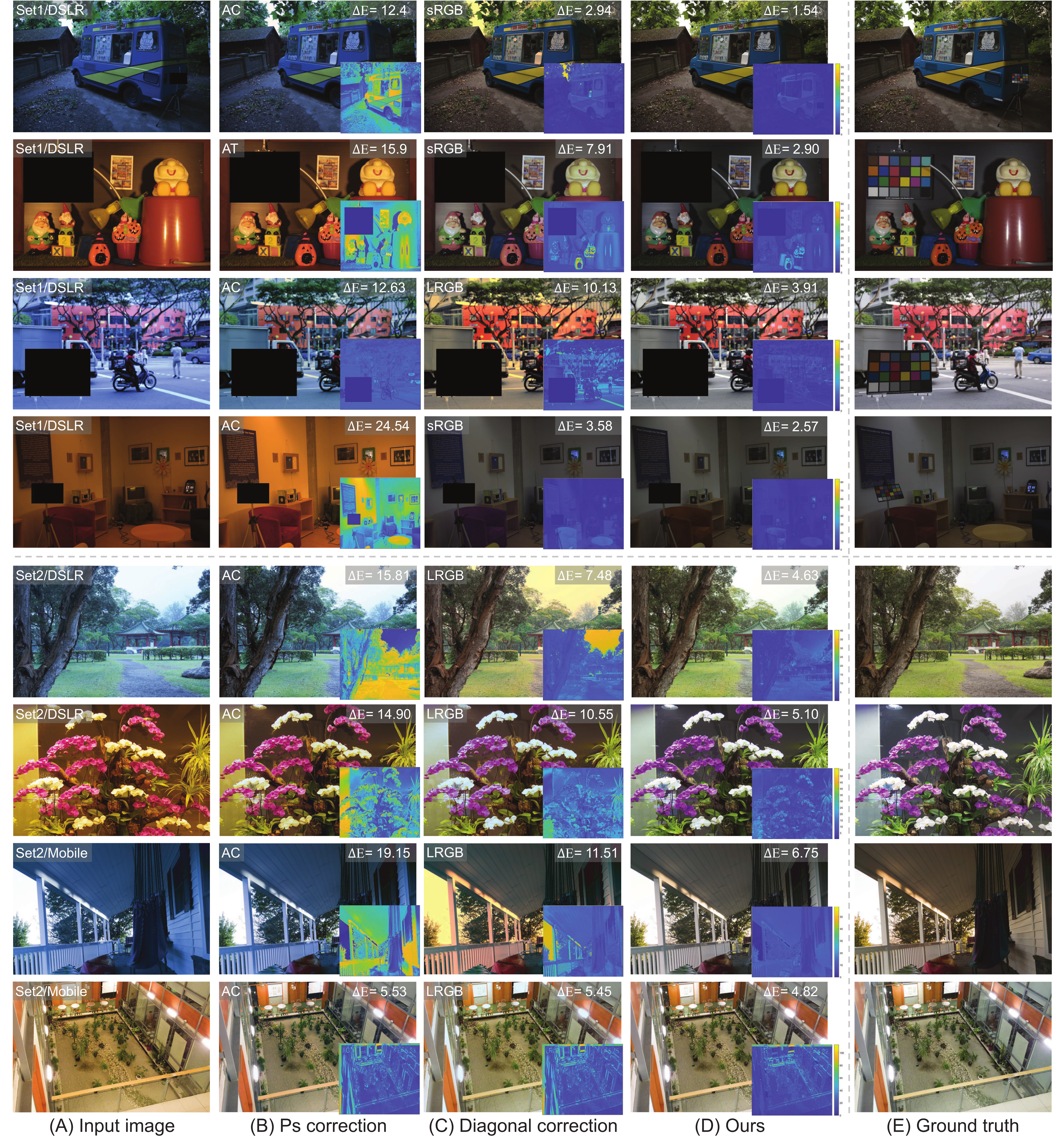}
\vspace{-7mm}
\caption[Additional qualitative comparisons between our method and other techniques on Set 1 and Set 2.]{Additional qualitative comparisons between our method and other techniques on \textbf{Set 1} (first four rows) and \textbf{Set 2} (last four rows). 
}
\label{KNN:sub:additional_results}
\end{figure}

Qualitative visual results for Set 1 and Set 2 are shown in Fig.~\ref{KNN:fig:set1_2_visual_results} and Fig.~\ref{KNN:sub:additional_results}.
It is arguable that our results are most visually similar to the ground truth images.  To confirm this independently, we have conducted a user study of 35 participants (18 males and 17 females), ranging in age from 21 to 46. Each one was asked to choose the most visually similar image to the ground truth image between the results of our method and the diagonal correction with the exact reference point. Experiments were carried out in a controlled environment. The monitor was calibrated using a Spyder5 colorimeter. 

Participants were asked to compare 24 pairs of images, such that for each quartile, based on the MSE of each method, 4 images were randomly picked from Set 1 and Set 2. That means the selected images represent the best, median, and worst results of each method and for each set. On average, 93.69\% of our results were chosen as the most similar to the ground truth images. Figure \ref{KNN:fig10} illustrates that the results of this study are statistically significant with $p$-value $< 0.01$.

\section{Summary}

This chapter has proposed the first method to explicitly address the problem of correcting a camera image that has been improperly white-balanced. This situation occurs when a camera's AWB fails or when the wrong manual WB setting is used. The proposed method is enabled by a dataset we generated with over 65,000 pairs of incorrectly white-balanced images and their corresponding correctly white-balanced image. Given an improperly white-balanced camera image, we outlined a simple KNN strategy that is able to find similar incorrectly white-balanced images in the dataset. Based on these similar examples images, we described how to construct a nonlinear color correction transform that is used to remove the color cast.  The proposed approach requires a small memory overhead (less than 25MB) and is computationally fast ($\sim$1 second for a 12 mega-pixel image).

\chapter{Deep Neural Networks Performance With White-Balance Errors \label{ch:ch8}}

Color correction has a crucial importance not only in photography aesthetics but also for computer vision tasks. In this chapter, we explore how strong color casts caused by incorrectly applied WB negatively impact the performance of DNNs targeting image segmentation and classification\footnote{This work was published in \cite{afifi2019else}: Mahmoud Afifi and Michael S. Brown. What Else Can Fool Deep Learning? Addressing Color Constancy Errors on Deep Neural Network Performance. In IEEE International Conference on Computer Vision (ICCV), 2019.}. In addition, we discuss how existing image augmentation methods used to improve the robustness of DNNs are not well suited for modeling WB errors. To address this problem, a novel augmentation method is proposed that can emulate accurate color constancy degradation. We also explore pre-processing training and testing images with a recent WB correction algorithm to reduce the effects of incorrectly white-balanced images. We examine both augmentation and pre-processing strategies on different datasets and demonstrate notable improvements on the CIFAR-10, CIFAR-100, and ADE20K datasets. The test set and source code of this work are available on GitHub: \href{https://github.com/mahmoudnafifi/WB_color_augmenter}{https://github.com/mahmoudnafifi/WB$\_$color$\_$augmenter}.

\section{Introduction} \label{ICCV:sec:introduction}

\begin{figure}[t]
\includegraphics[width=\linewidth]{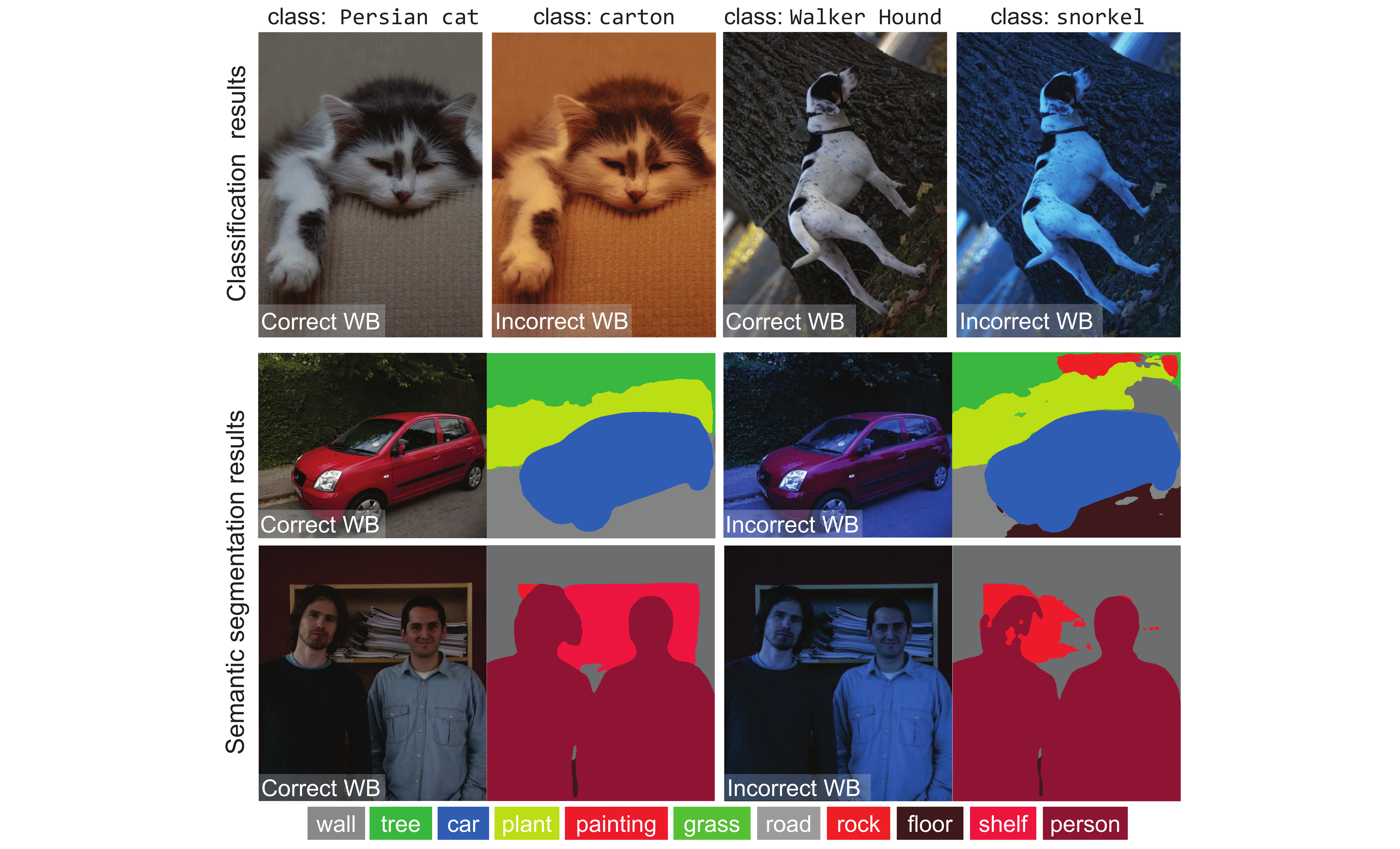}
\vspace{-7mm}
\caption{The effect of correct/incorrect computational color constancy (i.e., white balance) on (top) classification results by ResNet \cite{he2016deep}; and (bottom) semantic segmentation by RefineNet \cite{lin2017refinenet}.\label{ICCV:fig:teaser}}
\end{figure}

There is active interest in local image manipulations that can be used to fool DNNs into producing erroneous results.  Such ``adversarial attacks'' often result in drastic misclassifications.  We examine a less explored problem of \textit{global} image manipulations that can result in similar adverse effects on DNNs' performance.  In particular, we are interested in the role of computational color constancy, which makes up the WB routine on digital cameras.

We focus on computational color constancy because it represents a common source of global image errors found in real images. As discussed in Chapter \ref{ch:ch7}, when WB is applied incorrectly on a camera, it results in an undesirable color cast in the captured image. Images with such strong color casts are often discarded by users.  As a result, online image databases and repositories are biased to contain mostly correctly white-balanced images. This is an implicit assumption that is not acknowledged for datasets composed of images crawled from the web and online.  However, in real-world applications, it is unavoidable that images will, at some point, be captured with the incorrect WB applied.  Images with incorrect WB can have unpredictable results on DNNs trained on white-balanced biased training images, as demonstrated in Fig.~\ref{ICCV:fig:teaser}.

\paragraph{Contribution}~We examine how errors related to computational color constancy can adversely affect DNNs focused on image classification and semantic segmentation.  In addition, we show that image augmentation strategies used to expand the variation of training images are not well suited to mimic the type of image degradation caused by color constancy errors.  To address these problems, we introduce a novel augmentation method that can accurately emulate realistic color constancy degradation. We also examine our WB correction method (discussed in Chapter \ref{ch:ch7}) to pre-process testing and training images. Experiments on CIFAR-10, CIFAR-100, and the ADE20K datasets using the proposed augmentation and pre-processing correction demonstrate notable improvements to test image inputs with color constancy errors.

\section{Related Work} \label{ICCV:sec:relatedwork}

\begin{figure}[t]
\includegraphics[width=\linewidth]{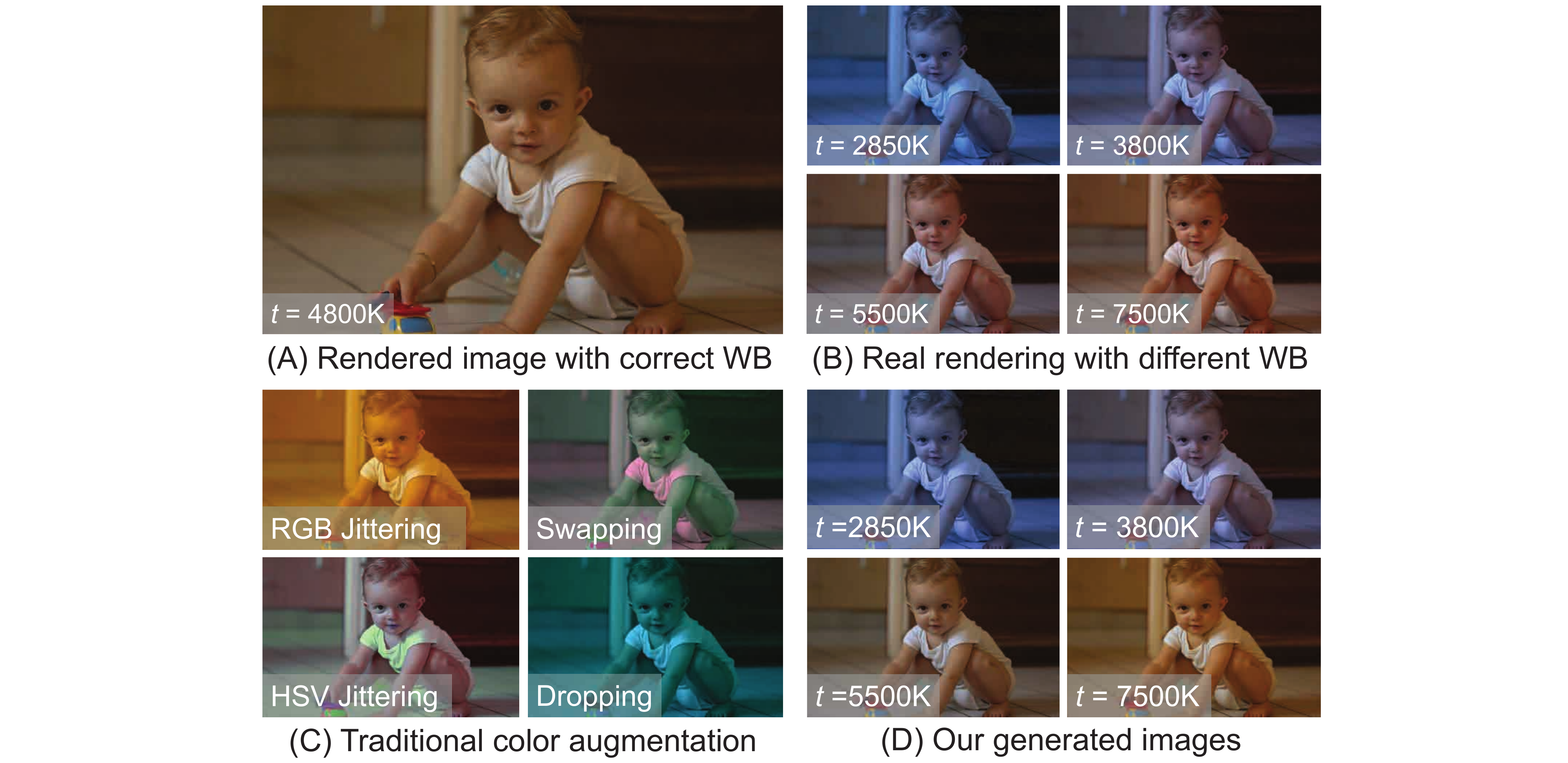}
\vspace{-7mm}
\caption[(A) An sRGB image with the correct WB applied.  (B) Images from the same camera with the incorrect WB color temperatures applied. (C) Images generated by processing image (A) using existing augmentation methods. (D) Images generated from (A) using our proposed method.]{(A) An sRGB image from a camera with the correct WB applied.  (B) Images from the same camera with the incorrect WB color temperatures ($t$) applied. (C) Images generated by processing image (A) using existing augmentation methods---the images clearly do not represent those in (B).  (D) Images generated from (A) using our proposed method detailed in Sec.~\ref{ICCV:sec:method}.}
\label{ICCV:fig:AugExamples}
\end{figure}

\paragraph{Adversarial Attacks}~DNN models are susceptible to adversarial attacks in the form of \textit{local} image manipulation (e.g., see~\cite{szegedy2013intriguing, goodfellow2014explaining, kurakin2016adversarial, cisse2017parseval}). These images are created by adding a carefully crafted imperceptible perturbation layer to the original image \cite{szegedy2013intriguing, goodfellow2014explaining}. Such perturbation layers are usually represented by \textit{local} non-random adversarial noise \cite{szegedy2013intriguing,goodfellow2014explaining,moosavi2016deepfool,xie2017adversarial, akhtar2018threat} or \textit{local} spatial transformations \cite{xiao2018spatially}. Adversarial examples are able to misguide pre-trained models to predict either a certain wrong response (i.e., targeted attack) or any wrong response (i.e., untargeted attack)~\cite{liu2017delving, 7958570,akhtar2018threat}.  While incorrect color constancy is not an explicit attempt at an adversarial attack, the types of failures produced by this \textit{global} modification act much like an untargeted attack and can adversely affect DNNs' performance.

\paragraph{Data Augmentation} To overcome limited training data and to increase the visual variation, image augmentation techniques are applied to training images. Existing image augmentation techniques include: geometric transformations (e.g., rotation, translation, shearing) \cite{hauberg2016dreaming, perez2017effectiveness, hauberg2016dreaming, cubuk2018autoaugment}, synthetic occlusions \cite{zhong2017random}, pixel intensity processing (e.g., equalization, contrast adjustment, brightness, noise) \cite{veeravasarapu2017adversarially, cubuk2018autoaugment}, and color processing (e.g., RGB color jittering and PCA-based shifting, HSV jittering,  color channel dropping, color channel swapping)~\cite{cubuk2018autoaugment,chatfield2014return, ImgaugLib, krizhevsky2012imagenet, redmon2016you, doersch2015unsupervised, movshovitz2016useful, lee2017unsupervised, kalantari2017deep}. Traditional color augmentation techniques randomly change the original colors of training images aiming for better generalization and robustness of the trained model in the inference phase. However, existing color augmentation methods often generate unrealistic colors which rarely happen in reality (e.g., green skin or purple grass). More importantly, the visual appearance of existing color augmentation techniques does not well represent the color casts produced by incorrect WB applied onboard cameras, as shown in Fig.~\ref{ICCV:fig:AugExamples}.  As demonstrated in \cite{andreopoulos2012sensor, diamond2017dirty, carlson2018modeling}, image formation has an important effect on the accuracy of different computer vision tasks. Recently, a simplified version of the camera imaging pipeline was used for data augmentation \cite{carlson2018modeling}. This augmentation method in \cite{carlson2018modeling}, however, explicitly did not consider the effects of incorrect WB due to the subsequent nonlinear operations applied after WB.
To address this issue, we propose a camera-based augmentation technique that can synthetically generates images with realistic WB settings.

\paragraph{DNN Normalization Layers}
Normalization layers are commonly used to improve the efficiency of the training process. Such layers apply simple statistics-based shifting and scaling operations to the activations of network layers. The shift and scale factors can be computed either from the entire mini-batch (i.e., batch normalization \cite{ioffe2015batch}) or from each training instance (i.e., instance normalization \cite{ulyanov2016instance}). Recently, batch-instance normalization (BIN) \cite{nam2018batch} was introduced to ameliorate problems related to styles/textures in training images by balancing between batch and instance normalizations based on the current task. Though the BIN is designed to learn the trade-off between keeping or reducing original training style variations using simple statistics-based operations, the work in~\cite{nam2018batch} does not provide any study regarding incorrect WB settings. The augmentation and pre-processing methods proposed in our work directly target training and testing images and do not require any change to a DNNs architecture or training regime.

\section{Effects of WB Errors on DNNs} \label{ICCV:subsec:pre-trained_evaluation}

\begin{figure}[t]
\includegraphics[width=\linewidth]{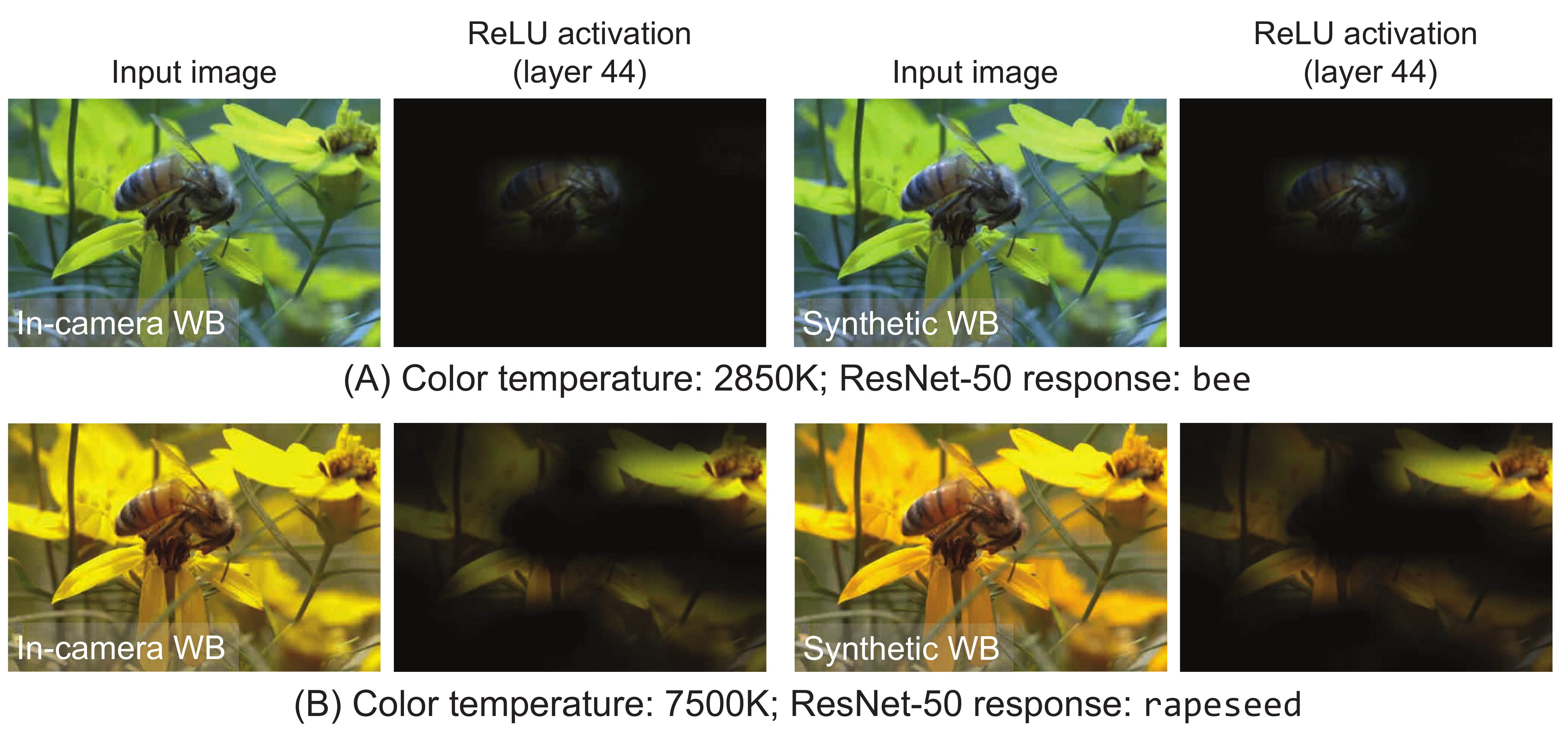}
\vspace{-7mm}
\caption[Image rendered with two different color temperatures (denoted by $t$) using in-camera rendering and our method.]{Image rendered with two different color temperatures (denoted by $t$) using in-camera rendering and our method. (A) Image class is \texttt{bee}. (B) Image class is \texttt{rapeseed}. Classification results were obtained by ResNet-50 \cite{he2016deep}.}
\label{ICCV:fig:attention}
\end{figure}

We begin by studying the effect of incorrectly white-balanced images on pre-trained DNN models for image classification and semantic segmentation. As a motivation, Fig. \ref{ICCV:fig:attention} shows two different WB settings applied to the same image.  Figure \ref{ICCV:fig:attention} shows that the DNN's attention for the same scene is considerably altered by changing the WB setting.

\begin{figure}[t]
\includegraphics[width=\linewidth]{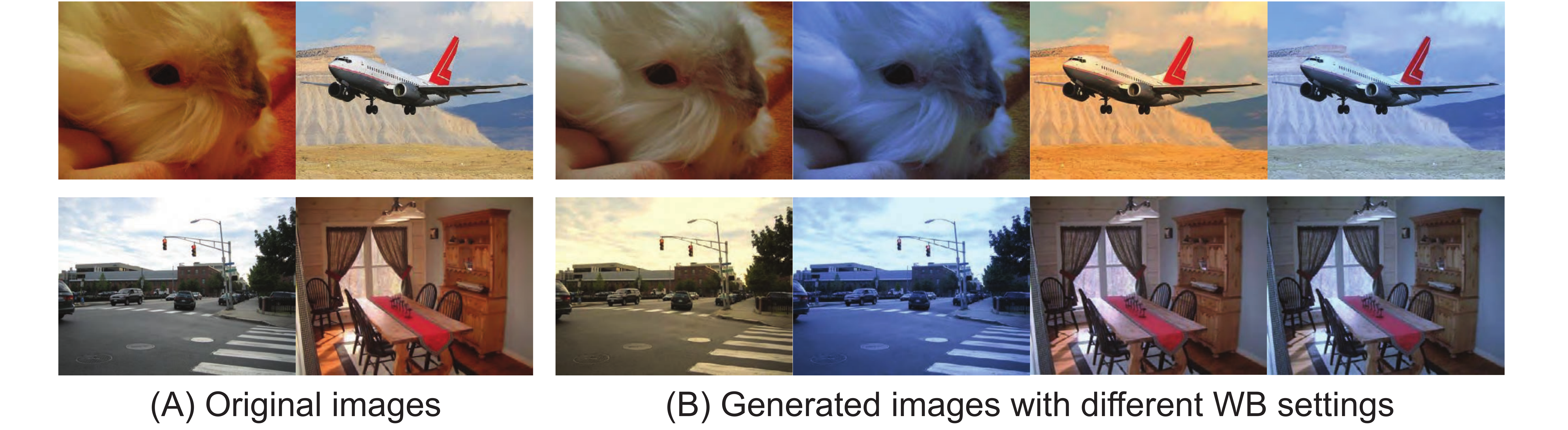}
\vspace{-7mm}
\caption[Examples from ImageNet validation set \cite{deng2009imagenet} and ADE20K validation set \cite{zhou2017scene}.]{Examples from ImageNet validation set \cite{deng2009imagenet} (first row) and ADE20K validation set \cite{zhou2017scene} (second row). (A) Original images. (B) Images with different WB settings produced by our method.}
\label{fig:examples_of_CIFAR10_SceneParsing_testing}
\end{figure}

For quantitative evaluations, we adopted several DNN models trained for the ImageNet Large Scale Visual Recognition Challenge (ILSVRC) 2012 \cite{deng2009imagenet} and the ADE20K Scene Parsing Challenge 2016 \cite{zhou2017scene}. Generating an entirely new labeled testing set composed of images with incorrect WB is an enormous task---ImageNet classification includes 1,000 classes and pixel-accurate semantic annotation requires $\sim$60 minutes per image \cite{richter2016playing}.   In lieu of a new testing set, we applied our method which emulates WB errors to the validation images of each dataset.  Our method will be detailed shortly in Sec.~\ref{ICCV:sec:method}.  Figure \ref{fig:examples_of_CIFAR10_SceneParsing_testing} shows examples of the generated images with different WB settings used in our study.

\begin{figure}[!t]
\includegraphics[width=\linewidth]{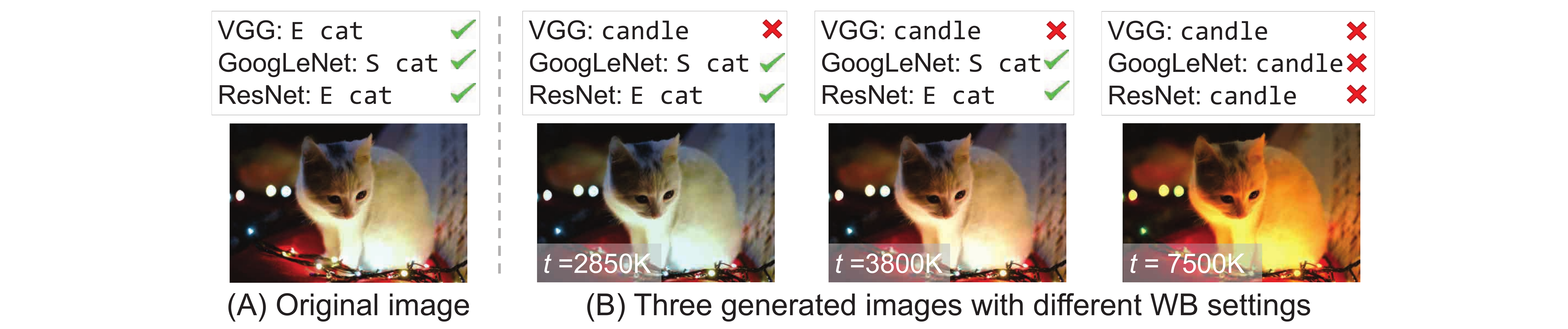}
\vspace{-7mm}
\caption[Pre-trained models are negatively impacted by incorrect WB settings.]{Pre-trained models are negatively impacted by incorrect WB settings. (A) Original image. (B) Generated images with different WB color temperatures (denoted by $t$). Classification results of: VGG-16 \cite{simonyan2014very}, GoogLeNet \cite{szegedy2015going}, and ResNet-50 \cite{he2016deep} are written on top of each image. The terms \texttt{E} and \texttt{S} stand for \texttt{Egyptian} and \texttt{Siamese}, respectively.}
\label{ICCV:fig:effect_synthWB_pre_trained}
\end{figure}

\paragraph{Classification}~We apply our method to ImageNet's validation set to generate images with five different color temperatures and two different photo-finishing styles for a total of ten WB variations for each validation image; 899 grayscale images were excluded from this process. In total, we generated 491,010 images. We examined the following six well-known DNN models, trained on the original ImageNet training images: AlexNet \cite{krizhevsky2012imagenet}, VGG-16 \& VGG-19 \cite{simonyan2014very}, GoogLeNet \cite{szegedy2015going}, and ResNet-50 \& ResNet-101 \cite{he2016deep}. Table \ref{ICCV:Table:resultsOfPretrained-classification} shows the accuracy drop for each model when tested on our generated validation set (i.e., with different WB and photo-finishing settings) compared to the original validation set.  In most cases, there is a drop of $\sim$10\% in accuracy. Figure \ref{ICCV:fig:effect_synthWB_pre_trained} shows an example of the impact of incorrect WB.

\paragraph{Semantic Segmentation} We used the ADE20K validation set for 2,000 images, and generated ten images with different WB/photo-finishing settings for each image. At the end, we generated a total of 20,000 new images. We tested the following two DNN models trained on the original ADE20K training set: DilatedNet \cite{chen2018deeplab, yu2015multi} and RefineNet \cite{lin2017refinenet}. Table \ref{ICCV:Table:resultsOfPretrained-segmentation} shows the effect of improperly white-balanced images on the intersection-over-union (IoU) and pixel-wise accuracy (pxl-acc) obtained by the same models on the original validation set.
While DNNs for segmentation fare better than the results for  classification, we still incur a drop of over 2\% in performance.

\begin{table}
\caption[Adverse classification performance on ImageNet\cite{deng2009imagenet} due to the inclusion of incorrect WB versions of its validation images.]{Adverse classification performance on ImageNet\cite{deng2009imagenet} due to the inclusion of incorrect WB versions of its validation images. The models were trained on the original ImageNet training set. The reported numbers denote the changes in the top-1 accuracy achieved by each model.}
\centering
\scalebox{0.7}
{
\begin{tabular}{|c|c|}
\hline
\textbf{Model}    & \textbf{Effect on top-1 accuracy} \\\hline
AlexNet \cite{krizhevsky2012imagenet} &  -0.112 \\\hline
VGG-16 \cite{simonyan2014very} &  -0.104 \\\hline
VGG-19 \cite{simonyan2014very}& -0.102  \\\hline
GoogLeNet \cite{szegedy2015going}& -0.107  \\\hline
ResNet-50 \cite{he2016deep} & -0.111 \\\hline
ResNet-101 \cite{he2016deep} & -0.109 \\ \hline
\end{tabular}\label{ICCV:Table:resultsOfPretrained-classification}
}
\end{table}

\begin{table}
\caption[Adverse semantic segmentation performance on ADE20K~\cite{zhou2017scene} due to the inclusion of incorrect WB versions of its validation images.]{Adverse semantic segmentation performance on ADE20K~\cite{zhou2017scene} due to the inclusion of incorrect WB versions of its validation images. The models were trained on ADE20K's original training set. The reported numbers denote the changes in intersection-over-union (IoU) and pixel-wise accuracy (pxl-acc) achieved by each model on the original validation.}
\centering
\scalebox{0.7}{
\begin{tabular}{|c|c|c|}
\hline
\textbf{Model} & \textbf{Effect on IoU} & \textbf{Effect on pxl-acc} \\ \hline
 DilatedNet \cite{chen2018deeplab, yu2015multi}  & -0.023  & -0.024  \\ \hline
RefineNet \cite{lin2017refinenet} & -0.031  & -0.026  \\ \hline
\end{tabular}
\label{ICCV:Table:resultsOfPretrained-segmentation}
}
\end{table}

\section{Emulating WB Errors} \label{ICCV:sec:method}

\begin{figure}[t]
\includegraphics[width=\linewidth]{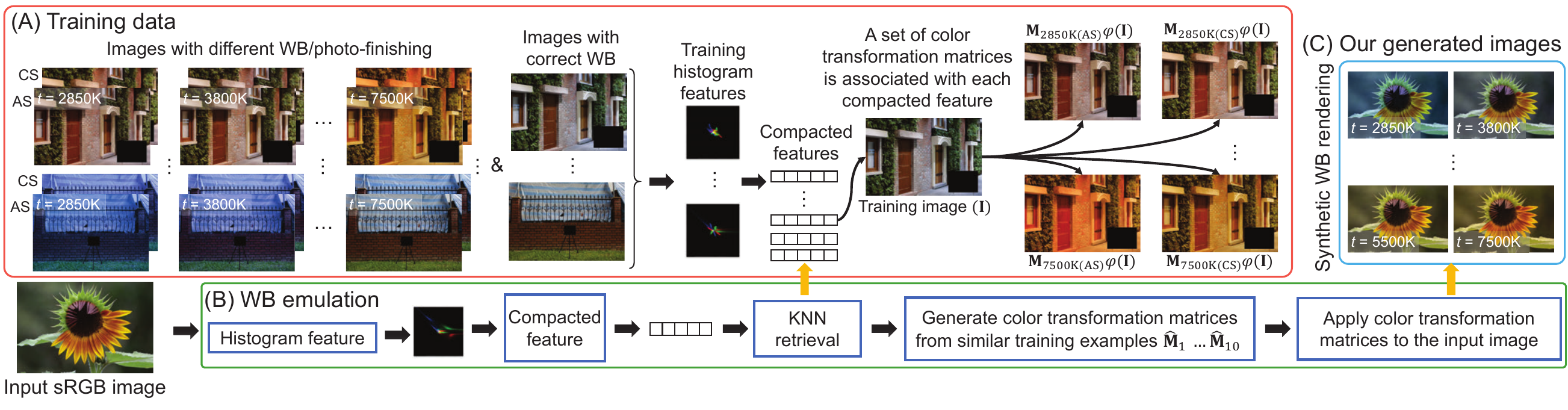}
\vspace{-7mm}
\caption[Our WB emulation framework.]{Our WB emulation framework. (A) A dataset of 1,797 correctly white-balanced sRGB images  (proposed in Chapter \ref{ch:ch7}); each image has ten corresponding sRGB images rendered with five different color temperatures and two photo-finishing styles, Camera Standard (CS) and Adobe Standard (AS). For each white-balanced image, we generate its compact histogram feature and ten color transformation matrices to the corresponding ten images. (B) Our WB emulation pipeline (detailed in Sec. \ref{ICCV:sec:method}). (C) The augmented images for the input image that represent different color temperatures (denoted by $t$) and photo-finishing styles.}
\label{ICCV:fig:overview}
\end{figure}

In this section, we outline our method for emulating WB errors. Our WB emulator heavily relies on our framework presented in Chapter\ \ref{ch:ch7}. Given an sRGB image, denoted as $\mat{I}_{t_\textrm{corr}}$, that is assumed to be white-balanced with the correct color temperature, our goal is to modify $\mat{I}_{t_\textrm{corr}}$'s colors to mimic its appearance as if it were rendered by a camera with different (incorrect) color temperatures, $t$, under different photo-finishing styles.  Since we do not have access to $\mat{I}_{t_\textrm{corr}}$'s original raw-RGB image, we cannot re-render the image from raw-RGB to sRGB using a standard camera pipeline.  Instead, we have modified our data-driven method discussed in Chapter\ \ref{ch:ch7}  to mimic this manipulation directly in the sRGB color space. Figure \ref{ICCV:fig:overview} provides an overview of our modified framework.

\subsection{Dataset} \label{ICCV:subsec:dataset}

Our method relies on our dataset of sRGB images generated in Chapter \ref{ch:ch7}. Recall that this dataset contains images rendered with different WB settings and photo-finishing styles. We have also a ground truth sRGB image (i.e., rendered with the ``correct'' color temperature) associated with each training image. In our WB emulation framework, we used 17,970 images from this dataset (1,797 correct sRGB images each with ten corresponding images rendered with five different color temperatures and two different photo-finishing styles, Camera Standard and Adobe Standard). The five color temperatures are: 2850 Kelvin (K), 3800K, 5500K, 6500K, and 7500K. In addition, each image was rendered using different camera photo-finishing styles.

\subsection{Color Mapping} \label{ICCV:subsec:WBaug}

Next, we compute a mapping between the correct white-balanced sRGB image to each of its ten corresponding images.  We follow the same procedure of the KNN WB method  (Chapter \ref{ch:ch7}) and use the kernel function, $\varphi$, to project RGB colors into a high-dimensional space. Then, we perform polynomial data fitting on these projected values as described in Chapter\ \ref{ch:ch7}.  Afterwards, we compute a color transformation matrix between each pair of correctly white-balanced image and its corresponding target image rendered with a specific color temperature and photo-finishing. In the end, we have \textit{ten} matrices associated with each image in our training data.

\subsection{Color Feature} As shown in Fig.~\ref{ICCV:fig:overview}, when augmenting an input sRGB image to have different WB settings, we search our dataset for similar sRGB images to the input image. This search is not based on scene content, but on the color distribution of the image (i.e., the RGB-$uv$ projected color histogram feature used in  Chapter \ref{ch:ch7}). 

\subsection{KNN Retrieval} Given a new input image, we extract its compacted color feature $\mat{v}$ (Eq. \ref{KNN:eq_PCA}), and then search for training examples with color distributions similar to the input image's color distribution. Similarly to our framework in Chapter\ \ref{ch:ch7}, the $\textrm{L}_2$ distance is adopted as a similarity metric between $\mat{v}$ and the training compacted color features. Afterwards, we retrieve the color transformation matrices associated with the nearest $k$ training images. The retrieved set of matrices is represented by  $\mat{M}_{\textrm{s}} = \{\mat{M}^{(j)}_{\textrm{s}}\}_{j=1}^{j=k}$, where $\mat{M}^{(j)}_{\textrm{s}}$ represents the color transformation matrix that maps the $j^{\textrm{th}}$ white-balanced training image colors to their corresponding image colors rendered with color temperature $t$.

\subsection{Transformation Matrix} After computing the distance vector between $\mat{v}$ and the nearest training features, we compute a weighting vector $\boldsymbol{\alpha}$ to blend between the associated transformation matrices of the nearest neighbor training examples (as described in Eqs. \ref{KNN:eq:weighting} and \ref{KNN:eq:final_matrix}). Lastly, the ``re-rendered'' image $\hat{\mat{I}}_{t}$ with color temperature $t$ is computed as in Eq. \ref{KNN:eq:correction}.

\section{Experiments}

\paragraph{Robustness Strategies}~Our goal is to improve the performance of DNN methods in the face of test images that may have strong global color casts due to computational color constancy errors. Based on the KNN framework  (Chapter \ref{ch:ch7}) and the modified framework discussed in Sec. \ref{ICCV:sec:method}, we examine three strategies to improve the robustness of the DNN models.

\begin{figure}[t]
\includegraphics[width=\linewidth]{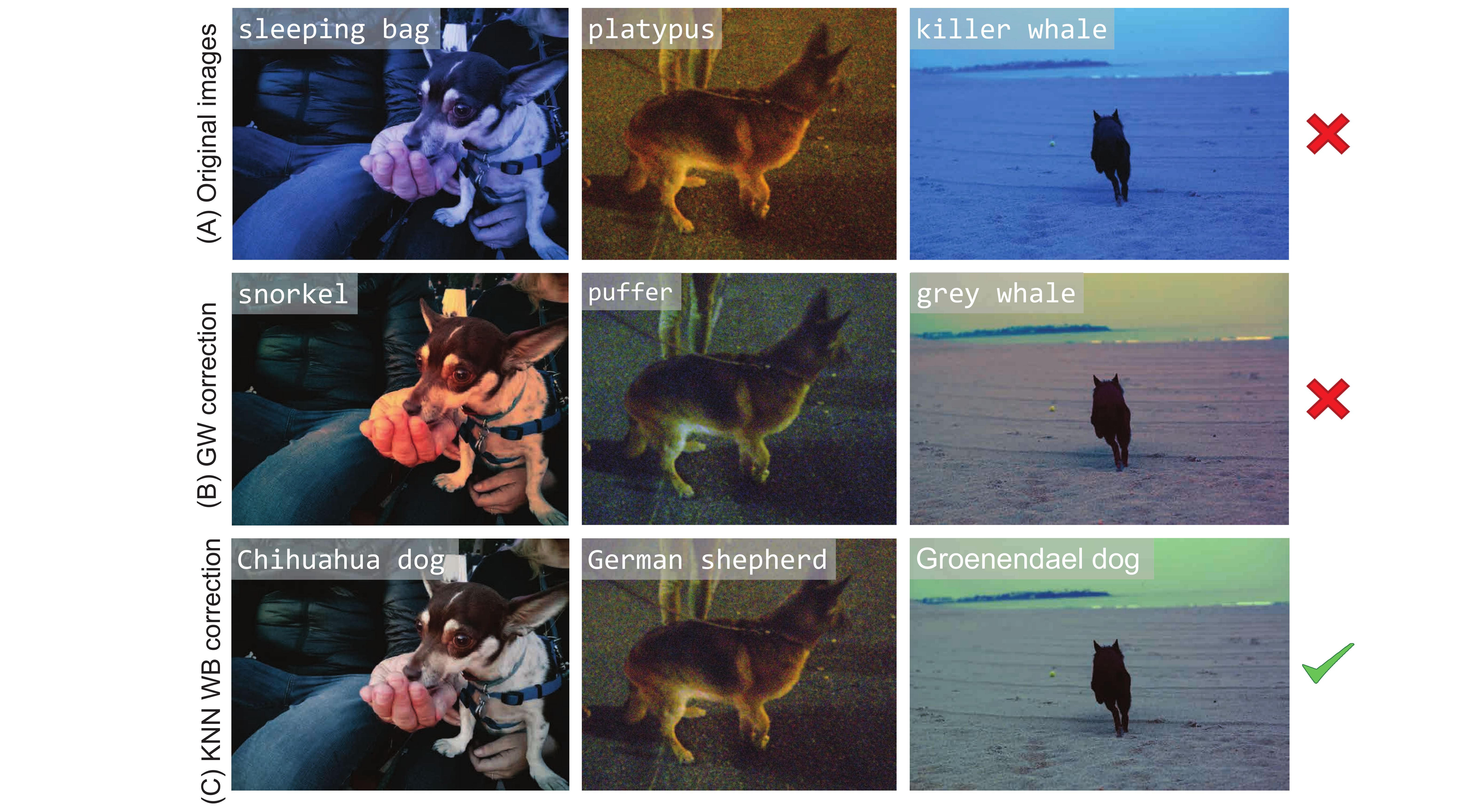}
\vspace{-7mm}
\caption[(A) Images with different categories of ``dogs'' rendered with incorrect WB settings. (B) Corrected images using GW \cite{GW}. (C) Corrected images using the KNN WB method  (Chapter \ref{ch:ch7}).]{(A) Images with different categories of ``dogs'' rendered with incorrect WB settings. (B) Corrected images using GW \cite{GW}. (C) Corrected images using the KNN WB method  (Chapter \ref{ch:ch7}). Predicted class by AlexNet is written on top of each image. Images in (A) and (B) are misclassified.}
\label{ICCV:fig-1}
\end{figure}

\noindent \textbf{(1)}~The first strategy is to apply a WB correction to each testing image in order to remove any unexpected color casts during the inference time.  Note that this approach implicitly assumes that the training images are correctly WB. In our experiments, we used the KNN WB method  (Chapter \ref{ch:ch7}) to correct the test images, because it currently achieves the state-of-the-art on white balancing sRGB rendered images. We examined adapting the simple diagonal-based correction -- which is applied by traditional WB methods that are intended to be applied on raw-RGB images (e.g., GW \cite{GW}) -- but found that they give inadequate results when applied on sRGB images, as also demonstrated in  (Chapter \ref{ch:ch7}). In fact, applying diagonal-based correction directly on the training image is similar to multiplicative color jittering. This is why we need to use a nonlinear color manipulation (e.g., polynomial correction estimated by our method in Chapter \ref{ch:ch7}) for more accurate WB correction for sRGB images. An example of the difference is shown in Fig. \ref{ICCV:fig-1}.

It is worth mentioning that the training data used by the KNN WB method (Chapter \ref{ch:ch7}) has five fixed color temperatures (2850K, 3800K, 5500K, 6500K, 7500K), all with color correction matrices mapping to their corresponding correct WB.  In most cases, one of these five fixed color temperatures will be visually similar to the correct WB.  Thus, if the KNN WB method is applied to an input image that is already correctly white-balanced, the computed transformation will act as an identity.

\noindent \textbf{(2)}~The second strategy considers the case that some of the training images may include some incorrectly white-balanced images. We, therefore, also apply the WB correction step to all the training images as well as testing images. This again uses the KNN WB method  (Chapter \ref{ch:ch7}) on both testing and training images.

\noindent \textbf{(3)}~The final strategy is to augment the training dataset based on our method described in Sec.~\ref{ICCV:sec:method}. Like other augmentation approaches, there is no pre-processing correction required. The assumption behind this augmentation process is that the robustness of DNN models can be improved by training on augmented images that serve as exemplars for color constancy errors.

\begin{figure}[t]
\includegraphics[width=\linewidth]{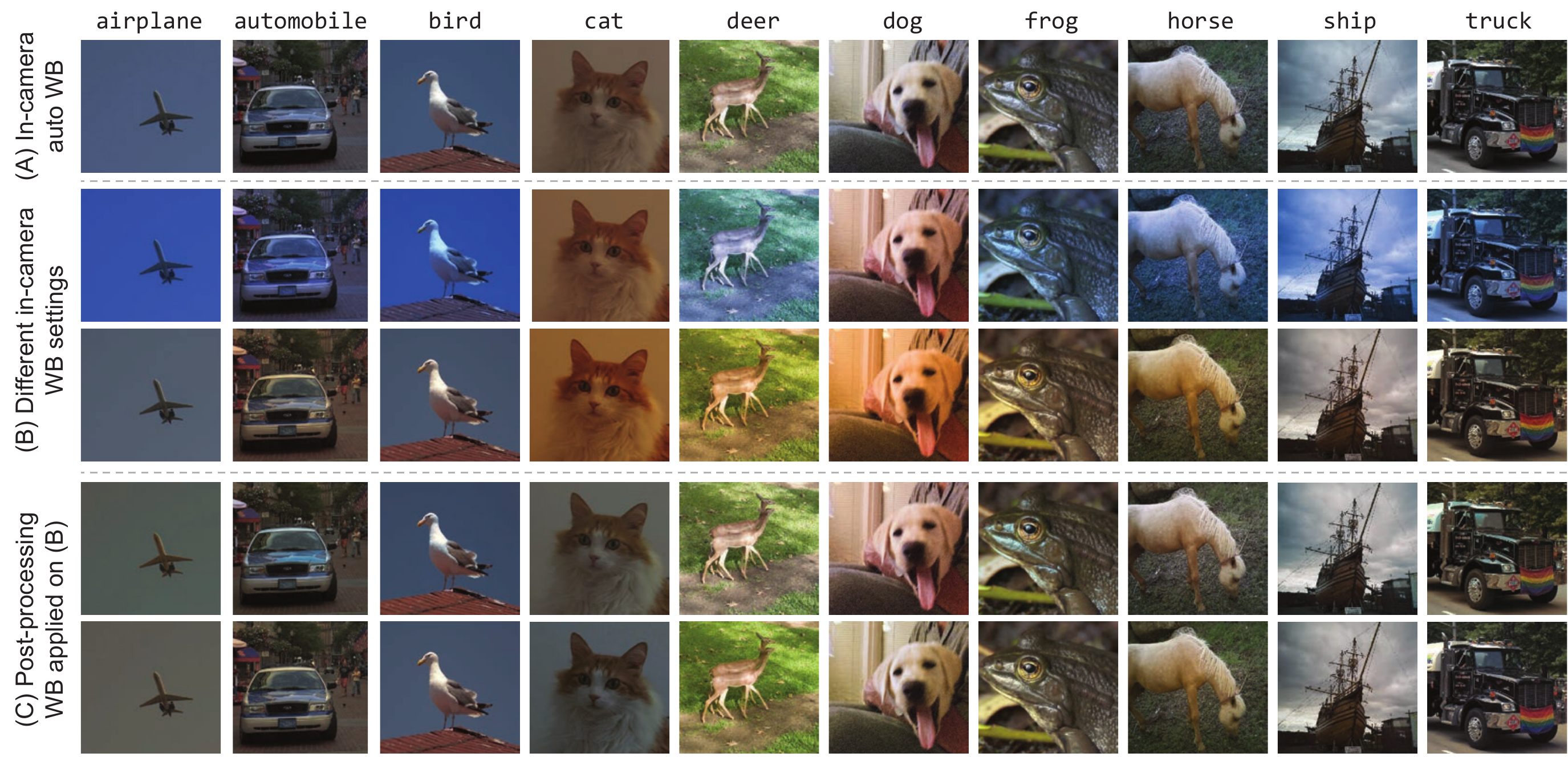}
\vspace{-7mm}
\caption[Examples of sRGB images used in Cat-2 (i.e., the external testing set of in-camera rendered images).]{Examples of sRGB images used in Cat-2 (i.e., the external testing set of in-camera rendered images). We used this set to evaluate trained models on CIFAR-10 dataset \cite{krizhevsky2009learning}. Class labels of CIFAR-10 dataset are written on top of each column. (A) Images were rendered using the in-camera auto WB setting. (B) Images were rendered with different WB settings. (C) Pre-processing WB correction  (Chapter \ref{ch:ch7}) is applied to images in (B).} 
\label{fig:exampleFromRealTestingSet}
\end{figure}

\paragraph{Testing Data Categories}~Testing images are grouped into two categories. In Category 1 (Cat-1), we expand the original testing images in the CIFAR-10, CIFAR-100, and ADE20K datasets by applying our method to emulate camera WB errors (described in Sec.~\ref{ICCV:sec:method}).  Each test image now has ten (10) variations that share the same ground truth labels.  We acknowledge this is less than optimal, given that the same method to modify the testing image is used to augment the training images.  However, we are confident in the proposed method's ability to emulate WB errors that we feel Cat-1 images represents real-world examples.  With that said,  we do not apply strategies 1 and 2 to Cat-1, as the KNN WB method is based on a similar framework used to generate the testing images.   For the sake of completeness, we also include Category  2 (Cat-2), which consists of new datasets generated directly from raw-RGB images.  Specifically, raw-RGB images are rendered using the full in-camera pipeline to sRGB images with in-camera color constancy errors. As a result, Cat-2's testing images exhibit accurate color constancy errors but contain fewer testing images for which we have provided the ground truth labels. Figure \ref{fig:exampleFromRealTestingSet} shows examples from our external testing set.

\subsection{Experimental Setup}

We compare the three above strategies with two existing and widely adopted color augmentation processes:  RGB color jittering and HSV jittering.

\paragraph{Our Method} The nearest neighbor searching was applied using $k=25$.
The proposed WB augmentation model runs in 7.3 sec (CPU) and 1.0 sec (GPU) to generate \textit{ten} 12-mega-pixel images.
The reported runtime was computed using Intel$^\circledR$ Xeon$^\circledR$ E5-1607 @ 3.10 GHz CPU and $\textrm{NVIDIA\texttrademark}$ Titan X GPU.

\paragraph{Existing Color Augmentation}
To the best of our knowledge, there is no standardized approach for existing color augmentation methods. Accordingly, we tested different settings and selected the settings that produce the best results.

For RGB color jittering, we generated ten images with new colors by applying a random shift $x \sim \mathcal{N}(\mu_x,\sigma^2)$ to each color channel of the image. For HSV jittering, we generated ten images with new colors by applying a random shift $x$ to the hue channel and multiplying each of the saturation and value channels by a random scaling factor $s \sim \mathcal{N}(\mu_s,\sigma^2)$. We found that $\mu_x = -0.3$, $\mu_s = 0.7$, and $\sigma = 0.6$ give us the best compromise between having color diversity with low color artifacts during the augmentation process.

\subsection{Network Training} \label{ICCV:subsec:training}

For image classification, training new models on the ImageNet dataset requires unaffordable efforts---for instance, ILSVRC 2012 consists of $\sim$1 million images and would be $\sim$10 million images after applying any of the color augmentation techniques. For that reason, we perform experiments on CIFAR-10 and CIFAR-100 datasets \cite{krizhevsky2009learning} due to a more manageable number of images in each dataset.

We trained SmallNet \cite{perez2017effectiveness} from scratch on CIFAR-10. We also fine-tuned AlexNet \cite{krizhevsky2012imagenet} to recognize the new classes in CIFAR-10 and CIFAR-100 datasets. As the CIFAR dataset contains $32\times 32$ pixels images, SmallNet was implemented to accept images with these dimensions. In order to fine-tune AlexNet, we rescale all images to $227\times 227$ pixels to fit with the input size of the architecture. For SegNet, the input size was $360\times 480$ pixels. For semantic segmentation, we fine-tuned SegNet \cite{badrinarayanan2017segnet} on the training set of the ADE20K dataset \cite{zhou2017scene}.

We train each model on: (i) the original training images, (ii) the KNN WB method  (Chapter \ref{ch:ch7}) applied to the original training images, and (iii) original training images with the additional images produced by color augmentation methods.
For color augmentation, we examined RGB color jittering, HSV jittering, and our WB augmentation. Thus, we trained five models for each CNN architecture, each of which was trained on one of the mentioned training settings.

Training was performed using mini-batch stochastic gradient descent with momentum. In our experiments, we used 0.9 momentum. The $\texttt{L}2$ regularization factor was set to 0.0005.
The mini-batch size was 512 images for SmallNet and AlexNet. For SegNet, the mini-batch size was 4 images due to the GPU memory limitation.

The cross entropy loss was used for image classification (i.e., SmallNet and AlexNet). For image semantic segmentation (i.e., SegNet), we adopted the weighted pixel-wise entropy loss as be suggested by \cite{badrinarayanan2017segnet}. The assigned weights for each class were computed using the median frequency balancing \cite{eigen2015predicting}.

The learning rate $\lambda$ was as follows. For AlexNet's conv1--fc7 layers, we used
$\lambda = 10^{-4}$. For AlexNet's fc8 layer, we used $\lambda = 10^{-4}\times 20$. SmallNet and SegNet were trained using $\lambda = 10^{-3}$. 

For fair comparisons, we trained each model for the same number of iterations. Specifically, the training was for $\sim$29,000 and $\sim$550,000 iterations for image classification and semantic segmentation tasks, respectively. We adjusted the number of epochs to make sure that each model was trained on the same number of mini-batches for fair comparison between training on augmented and original sets.
Note that by using a fixed number of iterations to train models with both original training data and augmented data, we did not fully exploit the full potential of the additional training images when we trained models using additional augmented data.

\begin{figure}[t]
\includegraphics[width=\linewidth]{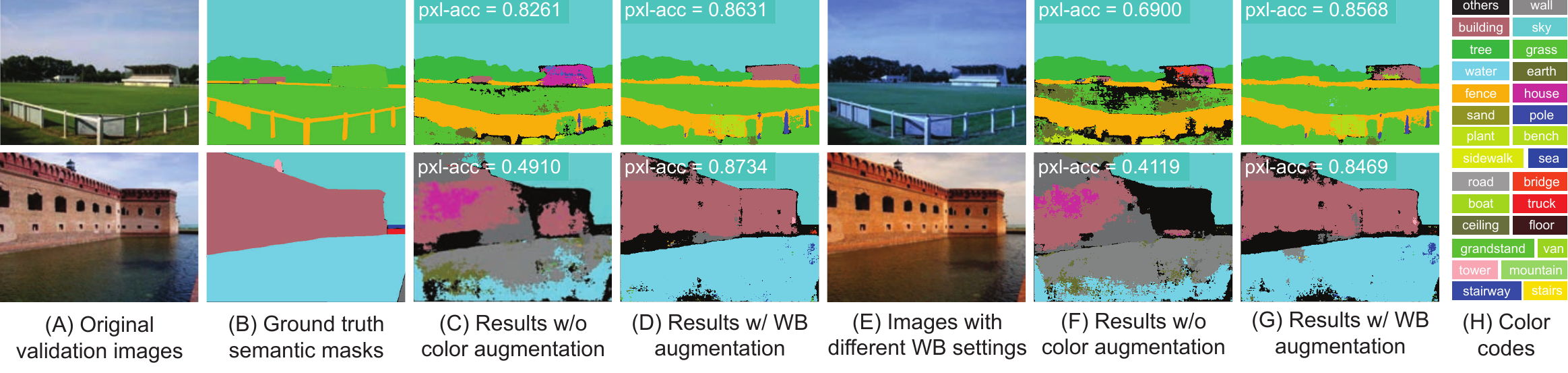}
\vspace{-7mm}
\caption[Results of SegNet \cite{badrinarayanan2017segnet} on the ADE20K validation set \cite{zhou2017scene}.]{Results of SegNet \cite{badrinarayanan2017segnet} on the ADE20K validation set \cite{zhou2017scene}. (A) Original validation image. (B) Ground truth semantic mask. (C) \& (D) Results of trained model wo/w color augmentation using image in (A), respectively. (E) Image with a different WB. (F) \& (G) Results w/o and with color augmentation using image in (E), respectively. (H) Color codes. The term `pxl-acc' refers to pixel-wise accuracy.}
\label{ICCV:fig:segNet}
\end{figure}

\begin{figure}[!t]
\includegraphics[width=\linewidth]{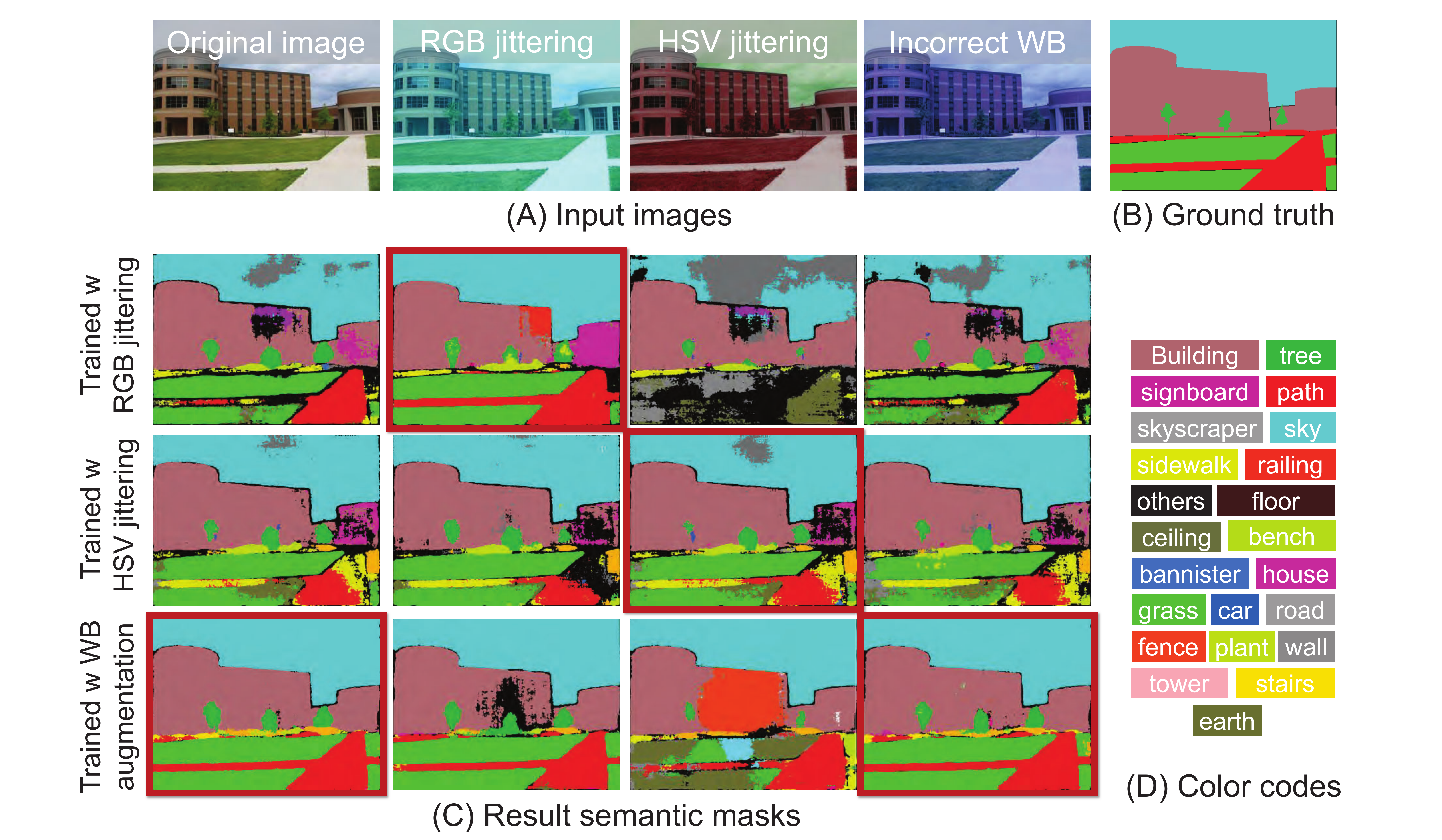}
\vspace{-7mm}
\caption[(A) Original image. (B) Ground truth semantic mask. (C) Generated by RGB and HSV jittering, and our WB emulation method. (D) color codes.]{(A) Original image. (B) Ground truth semantic mask. (C) Generated by RGB and HSV jittering, and our WB emulation method. (D) color codes. Result masks are obtained by training on augmented data using RGB/HSV jittering and our WB emulation method. The best results are shown in red borders.\label{fig:semantic2}}
\end{figure}

\begin{figure}[!t]
\includegraphics[width=\linewidth]{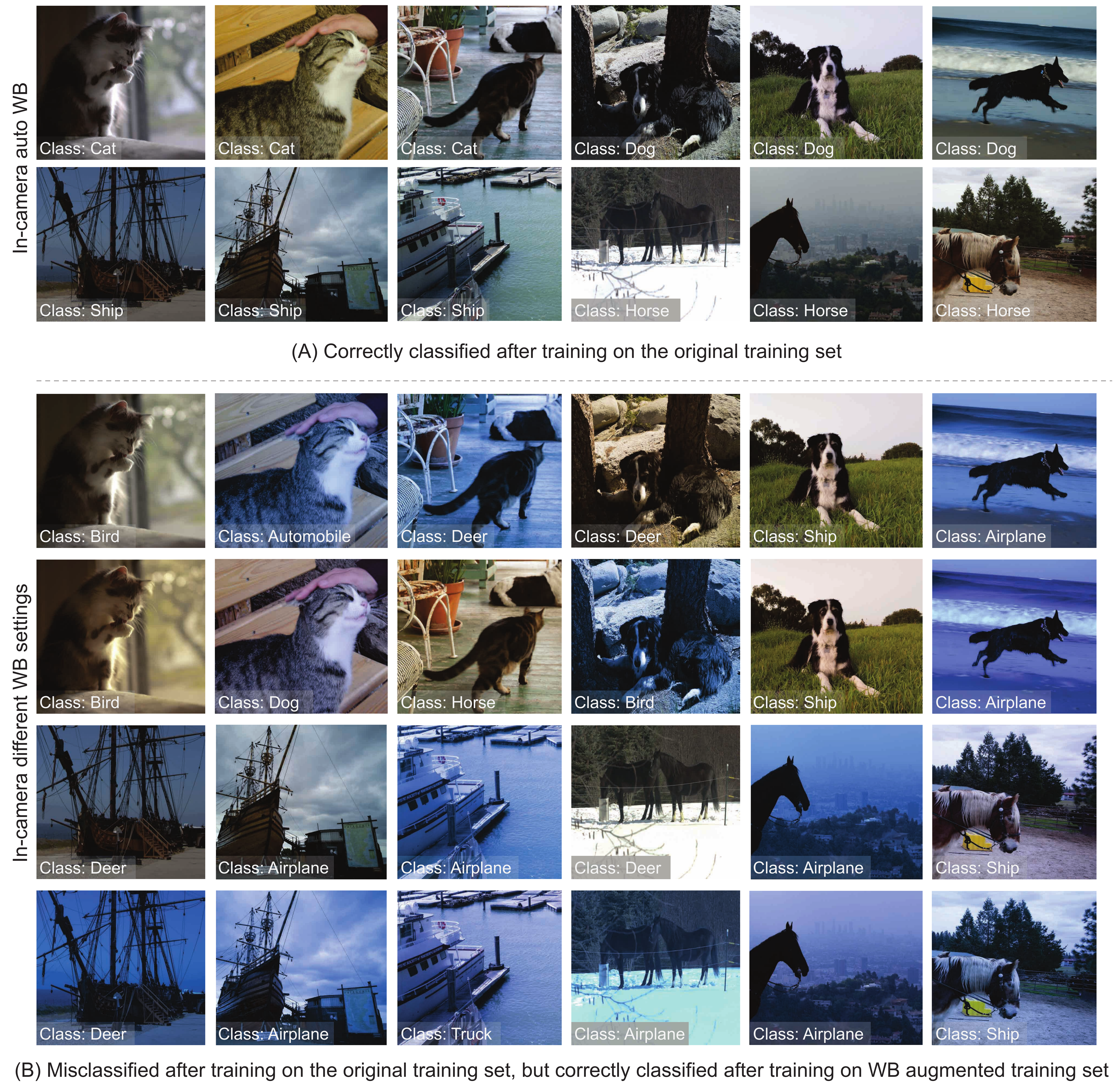}
\vspace{-7mm}
\caption[(A) Correctly classified images rendered with in-camera auto WB. (B) Misclassified images rendered with \textit{in-camera} different WB.]{(A) Correctly classified images rendered with in-camera auto WB. (B) Misclassified images rendered with \textit{in-camera} different WB. Note that all images in (B) are correctly classified by the same model (AlexNet \cite{krizhevsky2012imagenet}) trained on WB augmented data.}
\label{ICCV:fig:realFalseExamples}
\end{figure}

\begin{table}[t]
\caption[Results of SmallNet \cite{perez2017effectiveness} and AlexNet \cite{krizhevsky2012imagenet} on CIFAR dataset \cite{krizhevsky2009learning} (Cat-1).]{[Cat-1] Results of SmallNet \cite{perez2017effectiveness} and AlexNet \cite{krizhevsky2012imagenet} on CIFAR dataset \cite{krizhevsky2009learning}. The shown accuracies obtained by models trained on: original training, ``white-balanced'', and color augmented sets. The testing was performed using: original testing set and testing set with different synthetic WB settings (denoted as diff. WB).
The results of the baseline models (i.e., trained on the original training set) are highlighted in green, while the best result for each testing set is shown bold. We highlight best results obtained by color augmentation techniques in yellow. Effects on baseline model results are shown in parentheses. \label{ICCV:Table:resultsOfSmallNetAlexNet}}\vspace{-3mm}
\centering 
\scalebox{0.6}
{
\begin{tabular}{c|c|c|}
\cline{2-3}
\multicolumn{1}{c|}{\textbf{Cat-1}} & \multicolumn{2}{c|}{\textbf{SmallNet \cite{perez2017effectiveness} on CIFAR-10 \cite{krizhevsky2009learning}}} \\ \hline
\multicolumn{1}{|c|}{\textbf{Training set}} & Original & Diff. WB \\ \hline
\multicolumn{1}{|c|}{Original training set} & \cellcolor[HTML]{9AFF99}0.799 &\cellcolor[HTML]{9AFF99} 0.655 \\ \hline
\multicolumn{1}{|c|}{``White-balanced'' set} & 0.801 (+0.002) & 0.683 (+0.028) \\ \hline
\multicolumn{1}{|c|}{HSV augmented set} & 0.801 (+0.002) & 0.747 (+0.092) \\ \hline
\multicolumn{1}{|c|}{RGB augmented set} & 0.780 (-0.019) & 0.765 (+0.11) \\ \hline
\multicolumn{1}{|c|}{WB augmented set (ours)} & \cellcolor[HTML]{FFFC9E}\textbf{0.809 (+0.010)} &  \cellcolor[HTML]{FFFC9E}\textbf{0.786 (+0.131)} \\ \hline
\cline{2-3}
\multicolumn{1}{c|}{\textbf{Cat-1}} & \multicolumn{2}{c|}{\textbf{AlexNet \cite{krizhevsky2012imagenet} on CIFAR-10 \cite{krizhevsky2009learning}}} \\\hline
\multicolumn{1}{|c|}{Original training set} & \cellcolor[HTML]{9AFF99}\textbf{0.933} & \cellcolor[HTML]{9AFF99} 0.797 \\ \hline
\multicolumn{1}{|c|}{``White-balanced'' set} & 0.932 (-0.001) & 0.811 (+0.014) \\ \hline
\multicolumn{1}{|c|}{HSV augmented set} & 0.923 (-0.010) & 0.864 (+0.067) \\ \hline
\multicolumn{1}{|c|}{RGB augmented set} & 0.922 (-0.011) & 0.872 (+0.075) \\ \hline
\multicolumn{1}{|c|}{WB augmented set (ours)} & \cellcolor[HTML]{FFFC9E}0.926 (-0.007) & \cellcolor[HTML]{FFFC9E}\textbf{0.889 (+0.092)} \\ \hline
\cline{2-3}
\multicolumn{1}{c|}{\textbf{Cat-1}} & \multicolumn{2}{c|}{\textbf{AlexNet \cite{krizhevsky2012imagenet} on CIFAR-100 \cite{krizhevsky2009learning}}} \\ \hline
\multicolumn{1}{|c|}{Original training set} & \cellcolor[HTML]{9AFF99}\textbf{0.768} & \cellcolor[HTML]{9AFF99}0.526 \\ \hline
\multicolumn{1}{|c|}{``White-balanced'' set} & 0.757 (-0.011) & 0.543 (+0.017) \\ \hline
\multicolumn{1}{|c|}{HSV augmented set} & 0.722 (-0.044) & 0.613 (+0.087) \\ \hline
\multicolumn{1}{|c|}{RGB augmented set} & 0.723 (-0.045) & 0.645 (+0.119) \\ \hline
\multicolumn{1}{|c|}{WB augmented set (ours)} & \cellcolor[HTML]{FFFC9E}0.735 (-0.033) & \cellcolor[HTML]{FFFC9E}\textbf{0.670 (+0.144)} \\ \hline
\end{tabular}}
\end{table}

\begin{table}[t]
\caption[Results of SegNet \cite{badrinarayanan2017segnet} on the ADE20K validation set \cite{zhou2017scene} (Cat-1).]{[Cat-1] Results of SegNet \cite{badrinarayanan2017segnet} on the ADE20K validation set \cite{zhou2017scene}. The shown intersection-over-union (IoU) and pixel-wise accuracy (pxl-acc) were achieved by models trained using: original training, ``white-balanced'', and color augmented sets. The testing was performed using: original testing set and testing set with different synthetic WB settings (denoted as diff. WB). Effects on results of SegNet trained on the original training set are shown in parentheses.  Highlight marks are as described in Table \ref{ICCV:Table:resultsOfSmallNetAlexNet}.
\label{ICCV:Table:resultsOfSegNEt}}
\centering
\scalebox{0.7}
{
\begin{tabular}{c|c|c|}
\cline{2-3}
 & \multicolumn{2}{c|}{\textbf{IoU}} \\ \cline{2-3}
\multicolumn{1}{c|}{\textbf{Cat-1}} & Original & Diff. WB \\ \hline
\multicolumn{1}{|c|}{Original training set} & \cellcolor[HTML]{9AFF99}0.208 & \cellcolor[HTML]{9AFF99} 0.180  \\ \hline
\multicolumn{1}{|c|}{``White-balanced'' set} & \textbf{0.210 (+0.002)} & 0.197 (+0.017) \\ \hline
\multicolumn{1}{|c|}{HSV augmented set} & 0.192 (-0.016) & 0.185 (+0.005) \\ \hline
\multicolumn{1}{|c|}{RGB augmented set} & 0.195 (-0.013) & 0.190 (+0.010) \\ \hline
\multicolumn{1}{|c|}{WB augmented set (ours)} & \cellcolor[HTML]{FFFC9E}0.202 (-0.006) & \cellcolor[HTML]{FFFC9E}\textbf{0.199 (+0.019)} \\ \hline
\multicolumn{1}{c|}{\textbf{Cat-1}} & \multicolumn{2}{c|}{\textbf{pxl-acc}} \\ \hline
\multicolumn{1}{|c|}{Original training set} & \cellcolor[HTML]{9AFF99}0.603 & \cellcolor[HTML]{9AFF99} 0.557 \\ \hline
\multicolumn{1}{|c|}{``White-balanced'' set} & \textbf{0.605 (+0.002)} & 0.579 (+0.022) \\ \hline
\multicolumn{1}{|c|}{HSV augmented set} & 0.583 (-0.020) & 0.536 (-0.021) \\ \hline
\multicolumn{1}{|c|}{RGB augmented set} & 0.544 (-0.059) & 0.534 (-0.023) \\ \hline
\multicolumn{1}{|c|}{WB augmented set (ours)} & \cellcolor[HTML]{FFFC9E}0.597 (-0.006) & \cellcolor[HTML]{FFFC9E}\textbf{0.581 (+0.024)} \\ \hline
\end{tabular}}

\end{table}

\begin{table}[!t]
\caption[Results of SmallNet \cite{perez2017effectiveness} and AlexNet \cite{krizhevsky2012imagenet} (Cat-2).]{[Cat-2] Results of SmallNet \cite{perez2017effectiveness} and AlexNet \cite{krizhevsky2012imagenet}. The shown accuracies were obtained using trained models on the original training, ``white-balanced'', and color augmented sets. Effects on results of models trained on the original training set are shown in parentheses.
Highlight marks are as described in Table \ref{ICCV:Table:resultsOfSmallNetAlexNet}.
\label{ICCV:Table:RealDataResults}}
\centering
\scalebox{0.62}{
\begin{tabular}{c|c|c|c|}
\cline{2-4}
\textbf{Cat-2} & \multicolumn{3}{c|}{\textbf{SmallNet}}\\ \hline
\multicolumn{1}{|c|}{\textbf{Training Set}} & In-cam AWB & In-cam Diff. WB & WB pre-processing \\ \hline
\multicolumn{1}{|c|}{Original training set} & \cellcolor[HTML]{9AFF99}0.467 & \cellcolor[HTML]{9AFF99}0.404 & \cellcolor[HTML]{9AFF99}0.461 \\ \hline
\multicolumn{1}{|c|}{``White-balanced'' set} & \textbf{0.496 (+0.029)} & 0.471 (+0.067) & \textbf{0.492 (+0.031)} \\ \hline
\multicolumn{1}{|c|}{HSV augmented set} & 0.477 (+0.001) & 0.462 (+0.058) & 0.481 (+0.02) \\ \hline
\multicolumn{1}{|c|}{RGB augmented set} & 0.474 (+0.007) & 0.475 (+0.071) & 0.470 (+0.009) \\ \hline
\multicolumn{1}{|c|}{WB augmented set (ours)} & \cellcolor[HTML]{FFFC9E}0.494 (+0.027) & \cellcolor[HTML]{FFFC9E}\textbf{0.496 (+0.092)} & \cellcolor[HTML]{FFFC9E}0.484 (+0.023) \\ \hline
\multicolumn{1}{c|}{\textbf{Cat-2}} & \multicolumn{3}{c|}{\textbf{AlexNet}}\\ \hline
\multicolumn{1}{|c|}{Original training set} & \cellcolor[HTML]{9AFF99}0.792 & \cellcolor[HTML]{9AFF99}0.734 & \cellcolor[HTML]{9AFF99}0.772 \\ \hline
\multicolumn{1}{|c|}{``White-balanced'' set} & 0.784 (-0.008) & 0.757 (+0.023) & 0.784 (+0.012) \\ \hline
\multicolumn{1}{|c|}{HSV augmented set} & 0.790 (+0.002) & 0.771 (+0.037) & 0.779 (+0.007) \\ \hline
\multicolumn{1}{|c|}{RGB augmented set} & 0.791 (-0.001) & 0.779 (+0.045) & 0.783 (+0.011) \\ \hline
\multicolumn{1}{|c|}{WB augmented set (ours)} & \cellcolor[HTML]{FFFC9E}\textbf{0.799 (+0.007)} & \cellcolor[HTML]{FFFC9E}\textbf{0.788 (+0.054)} & \cellcolor[HTML]{FFFC9E}\textbf{0.787 (+0.015)} \\ \hline
\end{tabular}}
\end{table}

\subsection{Results on Cat-1}

Cat-1 tests each model using test images that have been generated by our method described in Sec.~\ref{ICCV:sec:method}.

\paragraph{Classification} We used the CIFAR-10 testing set (10,000 images) to test SmallNet and AlexNet models trained on the training set of the same dataset. We also used the CIFAR-100 testing set (10,000 images) to evaluate the AlexNet model trained on CIFAR-100. After applying our WB emulation to the testing sets, we have 100,000 images for each testing set of CIFAR-10 and CIFAR-100. The top-1 accuracies obtained by each trained model are shown in Table \ref{ICCV:Table:resultsOfSmallNetAlexNet}. The best results on our expanded testing images, which include strong color casts, were obtained using models trained on our proposed WB augmented data.

Interestingly, the experiments show that applying WB correction (Chapter \ref{ch:ch7}) on the training data, in most cases, improves the accuracy using both the original and expanded test sets. DNNs that were trained on WB augmented training images achieve the best improvement on the original testing images compared to using other color augmenters.

\paragraph{Semantic Segmentation} We used the ADE20K validation set using the same setup explained in Sec. \ref{ICCV:subsec:pre-trained_evaluation}. Table \ref{ICCV:Table:resultsOfSegNEt} shows the obtained pxl-acc and IoU of the trained SegNet models. The best results were obtained with our WB augmentation; Figure \ref{ICCV:fig:segNet} shows qualitative examples.

It is worth pointing out that when we utilize a certain augmentation technique, we implicitly help the model to expect inputs with similar conditions to what the color augmenter generates. When the testing images having \textit{unrealistic} color manipulations generated by RGB/HSV jittering, we found that trained models on augmented data by these techniques (i.e., RGB/HSV jittering) are more robust than models trained on other types of images (e.g., original or WB augmented training images). However, for images with color casts caused by different WB settings, the trained model with our WB augmentation has more resistance than other models; Figure \ref{fig:semantic2} shows an example.

\subsection{Results on Cat-2} \label{ICCV:subsec:Testing-realdata}

Cat-2 data requires us to generate and label our own testing image dataset using raw-RGB images.   To this end, we collected 518 raw-RGB images containing CIFAR-10 object classes from the following datasets:  HDR+ Burst Photography dataset \cite{hasinoff2016burst}, MIT-Adobe FiveK dataset \cite{bychkovsky2011learning}, and Raise dataset \cite{dang2015raise}. We rendered all raw-RGB images with different color temperatures and two photo-finishing styles using the  Adobe Camera Raw SDK. Adobe Camera Raw accurately emulates the ISP onboard a camera and produces results virtually identical to what the in-camera processing would produce (Chapter \ref{ch:ch7}). Images that contain multiple objects were manually cropped to include only the interesting objects---namely, the CIFAR-10 classes. At the end, we generated 15,098 rendered testing images that reflect real in-camera WB settings. We used the following testing sets in our experiments:

\noindent\textbf{(i) In-camera auto WB} contains images rendered with the AWB correction setting in Adobe Camera Raw, which mimics the camera's AWB functionality. AWB does fail from time to time; we manually removed images that had a noticeable color cast. This set of images is intended to be equivalent to testing images on existing image classification datasets.

\noindent\textbf{(ii) In-camera WB settings} contains images rendered with the different color temperatures and photo-finishing styles.  This set represents testing images that contain WB color cast errors.

\noindent\textbf{(iii) WB pre-processing correction applied to set (ii)} contains images of set (ii) after applying the KNN WB correction  (Chapter \ref{ch:ch7}). This set is used to study the potential improvement of applying a pre-processing WB correction in the inference phase.

Table \ref{ICCV:Table:RealDataResults} shows the top-1 accuracies obtained by SmallNet and AlexNet on the external testing sets. The experiments show the accuracy is reduced by $\sim$6\% when the testing set is images that have been modified with  incorrect WB settings compared with their original accuracies obtained with ``properly'' white-balanced images using the in-camera AWB. We also notice that the best accuracies are obtained by applying either a pre-processing WB on both training/testing images or our WB augmentation in an end-to-end manner. Examples of misclassified images are shown in Fig. \ref{ICCV:fig:realFalseExamples}.

\section{Summary}

This chapter has examined the impact on computational color constancy errors  on DNNs for image classification and semantic segmentation. A new method to perform augmentation that accurately mimics WB errors was introduced. We show that both pre-processing WB correction and training DNNs with our augmented WB images improve the results for DNNs targeting CIFAR-10, CIFAR-100, and ADE20K datasets. We believe our WB augmentation method will be useful for other tasks targeted by DNN where image augmentation is sought (see Appendix \ref{ch:appendix2}).

\part{Post-Capture White-Balance Editing\label{part:wb-editing}}
\chapter{Interactive White-Balance Editing \label{ch:ch9}}
Interactive WB editing allows the user to choose WB settings based on preference rather than the AWB estimation. This interactive mechanism is often performed by allowing the user to manually select different regions in a photo as examples of the illumination for WB correction (e.g., clicking on achromatic objects), or by using a color temperature slider to adjust the WB settings. Such interactive editing is possible only with images saved in a raw image format.  This is because raw images have no photo-rendering operations applied and photo-editing software is able to apply WB and other photo-finishing procedures to render the final image. Interactively editing WB in camera-rendered images is significantly more challenging, as discussed earlier, because the camera hardware has already applied WB to the image and subsequent nonlinear photo-processing routines. These nonlinear rendering operations make it difficult to change the WB post-capture. The goal of the following chapters, including this one, is to allow interactive WB manipulation of camera-rendered images. 

The work in this chapter builds on Chapter \ref{ch:ch7}.  We introduce a new framework\footnote{This work was published in \cite{afifi2020interactive}: Mahmoud Afifi and Michael S. Brown. Interactive White Balancing for Camera-Rendered Images. In Color and Imaging Conference, 2020.} that is able to link the nonlinear color-mapping functions, introduced in Chapter \ref{ch:ch7}, directly to the user's selected colors to allow interactive WB manipulation. In addition, our framework is more efficient in terms of memory and run-time (99\% reduction in memory and 3$\times$ speed-up).  Lastly, we describe how our framework can leverage a simple illumination estimation method (i.e., gray-world) to perform auto-WB correction that is on a par with the WB correction results achieved by our method in Chapter\ \ref{ch:ch7}. The source code of this work is available on GitHub: \href{https://github.com/mahmoudnafifi/WB_sRGB}{https://github.com/mahmoudnafifi/WB$\_$sRGB}.

\section{Introduction}\label{CIC28:sec:introduction} 

\begin{figure}[!t]
\includegraphics[width=\linewidth]{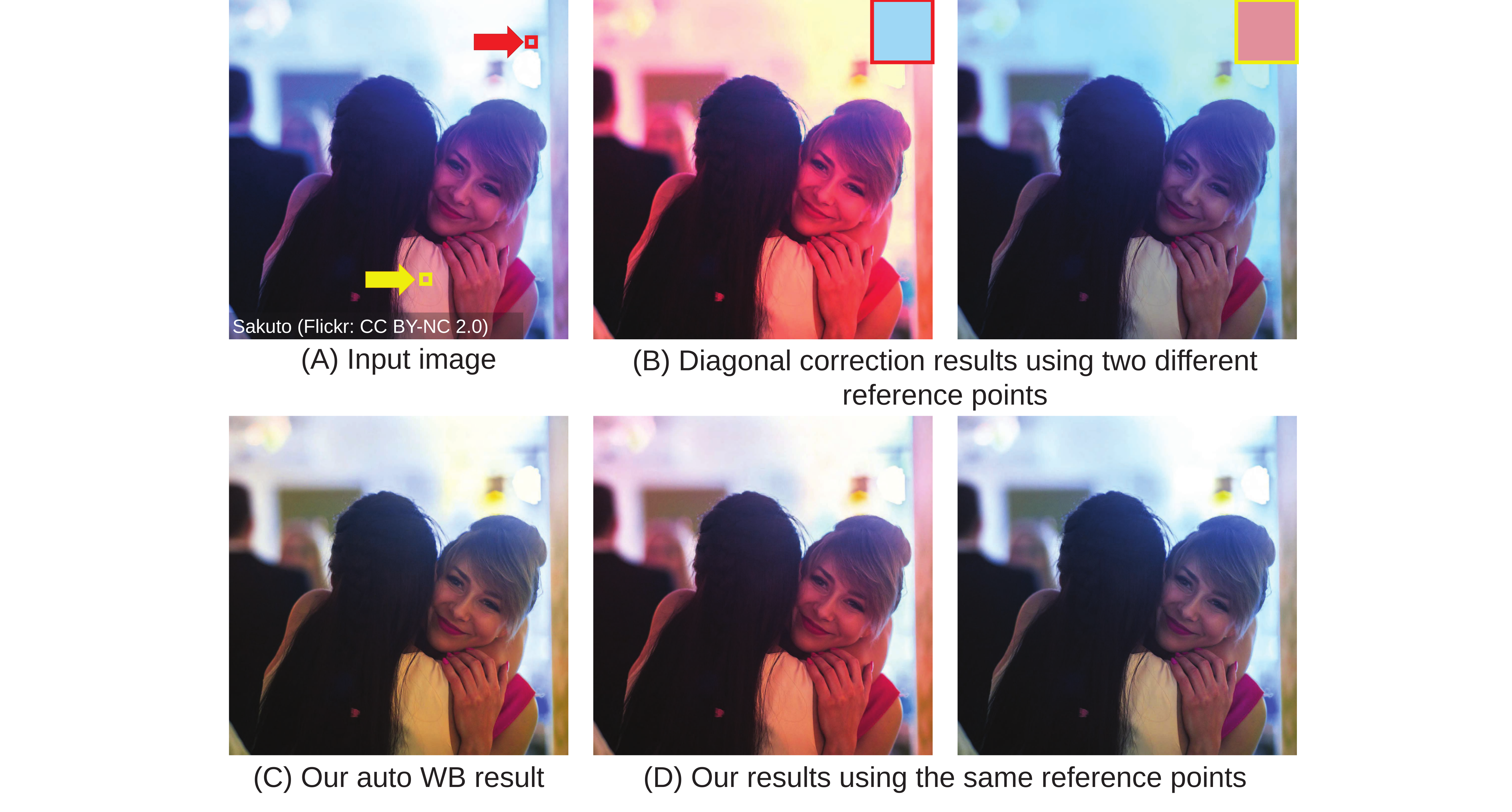}
\vspace{-7mm}
\caption[(A) A camera-rendered image with the wrong WB applied.  The user selects two reference colors in the scene representing achromatic scene materials. (B) The results correction using the conventional diagonal WB correction matrix.   (C) Our method's AWB on (A).  (D) Our method's results using the user-supplied reference colors.]{(A) A camera-rendered image with the wrong WB applied.  The user selects two reference colors in the scene representing achromatic scene materials. (B) The results correction using the conventional diagonal WB correction matrix.  Due to the nonlinear camera-rendering, the conventional approach is not sufficient to correct the WB.  (C) Our method's AWB on (A).  (D) Our method's results using the user-supplied reference colors.}
\label{CIC28:fig:teaser}
\end{figure}

There is photo-editing software (e.g., Adobe Lightroom~\cite{kelby2014adobe},  Skylum~\cite{Skylum}, Affinity Photo~\cite{affinity}) that enables interactive WB manipulation.  Instead of applying AWB, these methods allow the user to select a pixel's RGB values in the image of achromatic scene materials to serve as the estimated illumination color vector.  In some scenes, there may be more than one illuminant present and the users can choose which illumination they prefer to correct (see~Fig.~\ref{CIC28:fig:teaser}). This interactive WB editing, however, is possible only for photos saved in a raw image format.  This is because raw images have no photo-finishing applied---instead, the photo-editing software mimics the onboard camera rendering using the user-supplied parameters.

The goal of this chapter is to allow interactive WB manipulation for camera-rendered images.  As previously mentioned, WB manipulation in camera-rendered images is challenging due to the nonlinear operations applied by the camera hardware.  Our work presented in Chapter \ref{ch:ch7} showed that even when an exact achromatic reference scene point is known in a camera-rendered image, the conventional diagonal correction method cannot sufficiently remove the color casts caused by WB errors (see~Fig.~\ref{CIC28:fig:teaser}). To address this issue, we have proposed in Chapter \ref{ch:ch7} an effective nonlinear polynomial color correction function in lieu of the convention diagonal WB matrix.  The work in Chapter \ref{ch:ch7} used a histogram feature computed from an input image to determine which nonlinear correction function to use to correct a camera-rendered image that had the wrong WB applied.  The work in Chapter \ref{ch:ch7} provided no mechanism to link the nonlinear WB color-mapping functions to the user's selected pixel values.

\paragraph{Contribution}~We build upon the idea in Chapter \ref{ch:ch7} and propose a new framework that allows interactive WB editing in camera-rendered images. Our approach works by associating color-cast vectors with rectification functions that output nonlinear color-mapping functions to correct the camera-rendered image's WB.  This allows the user to supply a color vector from the image that results in a nonlinear color mapping to modify the image's WB based on the specified color vector.  Our framework requires only 1\% of the memory used in our work in Chapter \ref{ch:ch7} and runs 3$\times$ faster, allowing interactive functionality (see~Fig.~\ref{CIC28:fig:teaser}).  In addition, we show that our framework can also be used to perform AWB correction for camera-rendered images that were incorrectly white-balanced with results on a par with our work in Chapter \ref{ch:ch7}.

\section{Methodology}\label{CIC28:sec:method} 

Given an input image, $\mat{I}$, that is rendered with an incorrect or undesired WB setting, the goal is to generate a new output image, $\mat{I}_{\texttt{corr}}$, that represents the input $\mat{I}$ as it would appear if re-rendered with a new (presumably correct or desired) WB setting.  We will refer to the target  ``ground truth'' white-balanced image as $\mat{G}$. In the remaining part of this chapter, each image is represented as $3\!\times\!N$ matrix of the $\texttt{R}$, $\texttt{G}$, $\texttt{B}$ triplets, where $N$ is the total number of pixels in the image.

As discussed in previous chapters, a traditional diagonal-based solution relies on determining a 3D vector $\mat{\gamma} = \left[\mat{\gamma}(\texttt{R}), \mat{\gamma}({\texttt{G}}), \mat{\gamma}({\texttt{B})}\right]^\top$ that represents the scene illuminant color.  Since in our application, we are applying a correction not of the scene illumination but instead to a color cast present in the camera-rendered image with the wrong WB applied, we will refer to $\mat{\gamma}$ as a color-cast vector instead of an illumination vector.  In our case, this color-cast vector is provided manually by the user or based on an algorithm when in AWB mode. Given a color-cast vector, the image is assumed to be corrected using the following equation:
\begin{equation}
\label{CIC28:diagonal_corr}
\mat{I}_{\texttt{corr(diag)}} = diag\left(\mat{\ell}\right)\textrm{ }\mat{I},
\end{equation}
\noindent
where $diag\left(\cdot\right)$ is a $3\!\times\!3$ diagonal matrix of the {\it color-cast-correction} vector, $\mat{\ell}$. The vector $\mat{\ell} = \left[\mat{\gamma}(\texttt{G})/\mat{\gamma}(\texttt{R}), \texttt{ } \texttt{1}, \texttt{ } \mat{\gamma}(\texttt{G})/\mat{\gamma}(\texttt{B})\right]^\top$ and represents a simple modification of the color-cast vector.

Due to the nonlinearity applied to camera-rendered images, this simple scaling operation cannot properly correct WB errors. We proposed in Chapter \ref{ch:ch7} to replace the diagonal correction matrix with a nonlinear color-mapping function that could deal with the nonlinearities in the input $\mat{I}$. This function was computed in the following form:
\begin{equation}
\label{CIC28:eq:polynomial}
\mat{I}_{\texttt{corr(poly)}} = r\left(\mat{M}\right)\textrm{ }\phi\left(\mat{I}\right),
\end{equation}
\noindent
where $\phi$ is a kernel mapping function \cite{hong2001study} used in Chapter\ \ref{ch:ch7} (i.e., $\phi:\phi\left([\texttt{R}, \texttt{ }\texttt{G}, \texttt{ }\texttt{B}]^\top\right) \rightarrow [\texttt{R}$, $\texttt{G}$, $\texttt{B}$, $\texttt{RG}$, $\texttt{RB}$, $\texttt{GB}$, $\texttt{R}^2$, $\texttt{G}^2$, $\texttt{B}^2$, $\texttt{RGB}$, $1]^\top$), $\mat{M} \in \mathbb{R}^{33}$ is a vectorized form of the polynomial mapping matrix, and $r\left(\cdot\right)$ is a reshaping function that constructs the $3\!\times\!11$ matrix from the vectorized version, $\mat{M}$. As done in Chapter~\ref{ch:ch7}, this polynomial matrix can be computed in a closed-form via standard least squares methods.

\begin{figure}[!t]
\includegraphics[width=\linewidth]{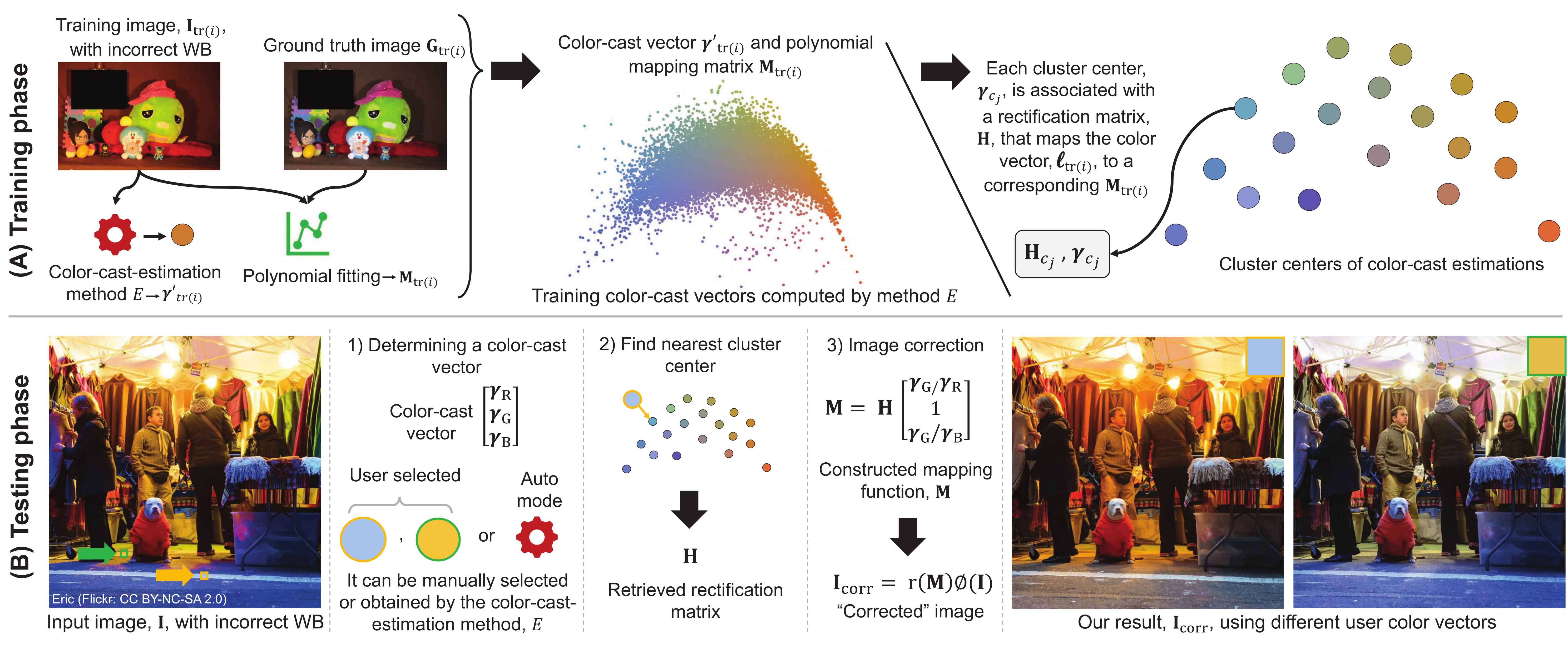}
\vspace{-7mm}
\caption[Overview of our proposed method.]{Overview of our proposed method. (A) [Training Phase] For each image in the training dataset, we estimate a color-cast vector using an off-the-shelf-illumination estimation algorithm $E$ (e.g., GW \cite{GW}).  A nonlinear color-mapping function that maps this training image to a correctly white-balanced image is also computed. The color-cast vectors of each training image are clustered.  A \textit{rectification function} is computed for each cluster that returns a color-mapping function based on a color-cast vector. (B) [Testing Phase] When applying our method, the user either manually provides a color-cast vector or uses the method $E$ to predict a color-cast vector. Using this color-cast vector, the most similar cluster in the training data is found and its rectification function is used to compute a mapping function $\mat{M}$ that is applied to correct the image.}
\label{CIC28:fig:main}
\end{figure}

Though this nonlinear color mapping achieves superior results compared to the diagonal-based correction, it lack a correlation to a color-cast vector. In the following section, we describe how to efficiently associate these color-mapping functions with color vectors that can be used for WB correction in a camera-rendered image $\mat{I}$.   Fig.~\ref{CIC28:fig:main} provides an overview of our framework.

\subsection{Training Phase}  \label{CIC28:subsec:training} 

\begin{figure}[!t]
\includegraphics[width=\linewidth]{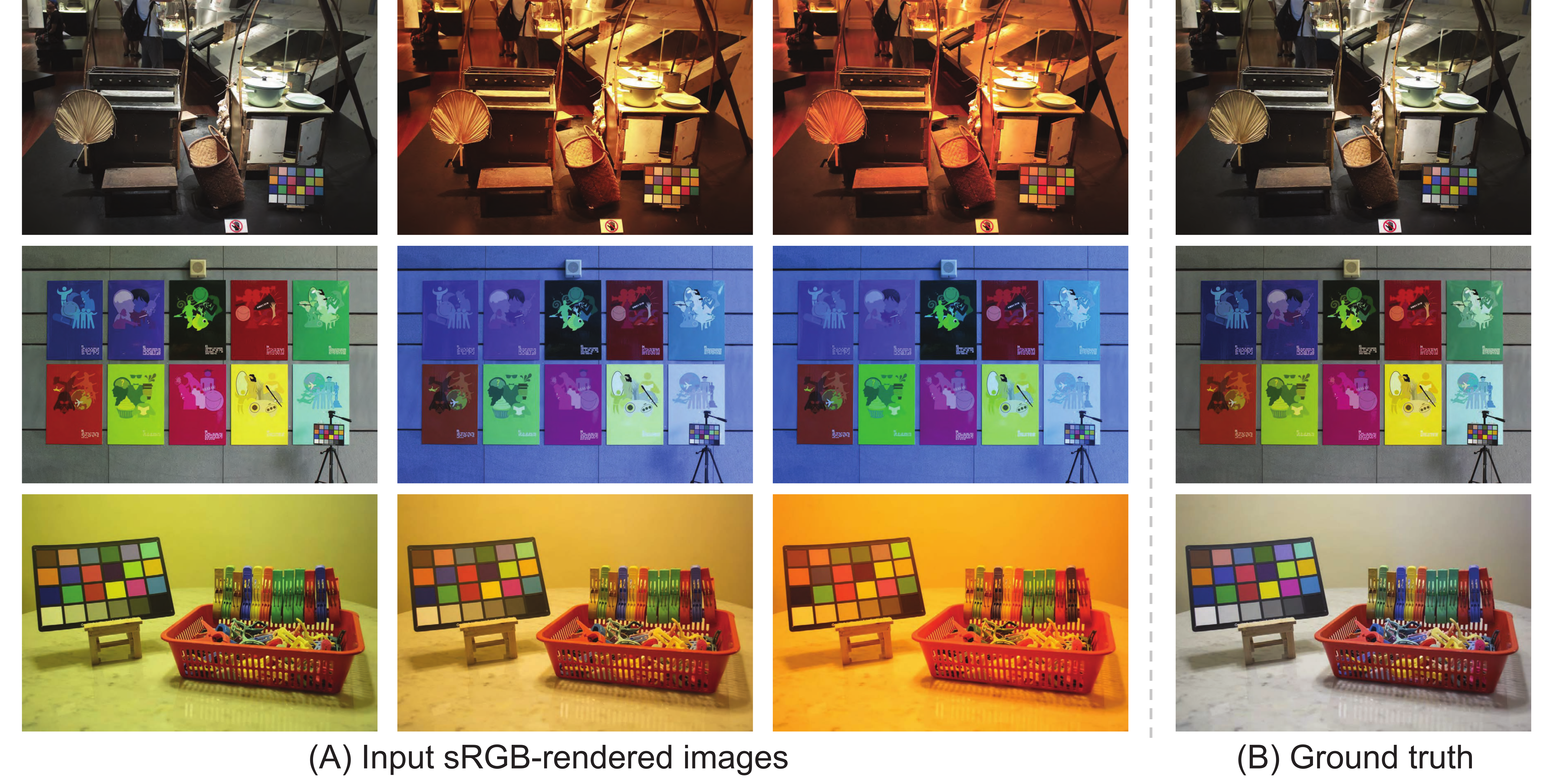}
\vspace{-7mm}
\caption[This figure shows examples from our Rendered WB dataset (proposed in Chapter \ref{ch:ch7}) used in order to generate our training rectification functions.]{This figure shows examples from our Rendered WB dataset (proposed in Chapter \ref{ch:ch7}) used in order to generate our training rectification functions. (A) Three examples of the same scene rendered with different incorrect WB settings. (B) The ground truth white-balanced image.}
\label{CIC28:fig:dataset}
\end{figure}

The first step of our method is to have a large number of training examples that exhibits a wide range of WB errors in camera-rendered images. We used the Rendered WB dataset proposed in Chapter \ref{ch:ch7}, which contains $\sim$65,000 pairs of improperly white-balanced camera-rendered (sRGB) images and their corresponding ground truth white-balanced sRGB images. All images have WB settings applied in the sensor raw space followed by an emulation of in-camera nonlinear rendering operations to get the final camera-rendered images. This dataset consists of two sets: (i) the training set, referred to as Set 1, and (ii) the testing set, referred to as Set 2. Fig. \ref{CIC28:fig:dataset} shows example images taken from the Rendered WB dataset.

We used all training images in Set 1 to compute the correction functions $\mat{M}$ described in Eq.~\ref{CIC28:eq:polynomial}. For each pair, $i$, of an improperly white-balanced image, $\mat{I}_{\texttt{tr}(i)}$, and the corresponding ground truth image, $\mat{G}_{\texttt{tr}(i)}$, we compute our 33-dimensional vectorized polynomial mapping matrix $\mat{M}_{\texttt{tr}(i)}$ as follows:
\begin{equation}
\label{CIC28:minimization_for_M}
\underset{\mat{M}_{\texttt{tr}(i)}}{\argmin} \left\|r\left(\mat{M}_{\texttt{tr}(i)}\right)\textrm{ }\phi\left(\mat{I}_{\texttt{tr}(i)}\right)    - \mat{G}_{\texttt{tr}(i)}\right\|_{\textrm{F}},
\end{equation}
\noindent
where $\left\|.\right\|_{\textrm{F}}$ is the Frobenius norm. Afterwards, each training image, $\mat{I}_{\texttt{tr}(i)}$, is associated with a color-cast vector  $\mat{\gamma}^{'}_{\texttt{tr}(i)}$, which will be later used to compute a color-cast-correction vector $\mat{\ell}_{\texttt{tr}(i)}$. This color-cast vector can be computed using any off-the-shelf illuminant estimation algorithm, $E: E\left(\mat{I}_{\texttt{tr}(i)}\right) \rightarrow \mat{\gamma}^{'}_{\texttt{tr}(i)}$.  

We cluster the training data based on their color-cast vectors into $k$ clusters. In our experiments, we used k-means++ \cite{arthur2006k}  with a cosine similarity distance metric and set $k$ to 50.  Each cluster, noted as $\textbf{c}$, also has a number of $\mat{M}_{i}$ mapping functions associated with it, where $i \in \textbf{c}$.  Instead of storing all of these mapping functions, we derive a single mapping function, termed a {\it rectification function}, that can estimate the color-mapping function based on the polynomial matrix $\mat{M}$. This is described in the following section.

\subsection{ Rectification Function}   \label{CIC28:subsec:rectfunc} 

Our rectification function is inspired by a bias-correction method proposed to rectify scene illuminant estimation errors \cite{MomentCorrection}. Specifically, we propose a rectification function, $\mat{H}$, that maps a color-cast-correction vector, $\mat{\ell}$, directly to a nonlinear correction matrix as follows:
\begin{equation}
\label{CIC28:matrix_decomspition}
\mat{M} = \mat{H}\textrm{ }\mat{\ell},
\end{equation}
\noindent
where $\mat{M}$ is a $33\!\times\!1$ vectorized polynomial matrix computed to map the colors of an incorrectly white-balanced image, $\mat{I}$, into the corresponding colors of the correctly white-balanced image, $\mat{G}$, $\mat{H}$ is our $33\!\times\!3$ rectification matrix, and $\mat{\ell}$ is the color-cast-correction vector computed from the color-cast vector as described earlier.  This type of correction function was used by Finlayson~\cite{MomentCorrection} to correct biases made by illumination estimation algorithms.  In ~\cite{MomentCorrection}, an estimated illumination vector would be mapped to a new illumination vector that was closer to the ground truth based on the training dataset.  In our case, the function $\mat{H}$ maps $\mat{\ell}$ to its corresponding matrix $\mat{M}$, allowing us to connect a color-cast vector directly to a mapping function.

Working from Eq.~\ref{CIC28:matrix_decomspition}, we compute a rectification matrix for each cluster, denoted as $\mat{H}_{\textbf{c}(j)}$ for cluster $j$. Let $n$ be the number of training examples belonging to each cluster $\textbf{c}(j)$, s.t. $j\in\left[1,2...k\right]$. For each cluster, we minimize the following equation to compute its rectification matrix:
\begin{equation}
\label{CIC28:minimization_for_H}
\underset{\mat{H}_{\textbf{c}(j)}}{\argmin} \left\|\sum_{i \in \textbf{c}(j)}\mat{H}_{\textbf{c}(j)}\textrm{ }\mat{\ell}_{i} - \mat{M}_{i}\right\|_{\textrm{F}},
\end{equation}
\noindent
where $\mat{\ell}_{i}$ is a $3\!\times\!1$ color-cast-correction vector of the $i^\text{th}$ training example in cluster $\textbf{c}(j)$, $\mat{M}_{i} \in \mathbb{R}^{33}$ is a vectorized polynomial matrix associated with the $i^\text{th}$ training example in cluster $\textbf{c}(j)$, and $\mat{H}_{\textbf{c}(j)}$ is a $33\!\times\!3$ rectification matrix assigned to cluster $\textbf{c}(j)$.   Eq.~\ref{CIC28:minimization_for_H} essentially estimates a single $\mat{H}$ per cluster that minimizes the error over all $\mat{\ell}_{i}$ associated with this cluster.

After this procedure, each color-cast cluster is now represented by the mean color-cast vector, $\mat{\gamma}_{c_j}$, of all color-cast vectors that belong to it. For each cluster, we store the corresponding rectification matrix, $\mat{H}_{\textbf{c}(j)}$, to be used in the testing phase. This model requires only 0.04 MB memory to encode our 50 rectification functions, compared to $\sim$25 MB required by our method in Chapter \ref{ch:ch7} ($\sim$99\% reduction in memory requirements).

\subsection{Testing Phase}   \label{CIC28:subsec:testing} 

Our method can easily be used in an AWB mode. To do this, given an input image $\mathbf{I}$, the same illuminant estimation algorithm, $E$, used in the training phase to compute the illuminant vector, $\mat{\gamma}^{'}$, is applied.  Afterwards, we search among the pre-computed color-cast cluster centers to find the closest cluster $\textbf{c}(h)$ to our testing color vector. Then, the rectification function of the closest cluster is retrieved and used along with the computed color-cast-correction vector, $\mat{\ell}^{'}$, to correct the testing image as described in the following equation:
\begin{equation}
\label{CIC28:our_correction}
\mat{I}_{\texttt{corr}} = r\left(\mat{H}_{\textbf{c}(h)}\textrm{ } \mat{\ell}^{'}\right)\textrm{ }\phi\left(\mat{I}\right).
\end{equation}

\begin{figure}[t]
\includegraphics[width=\linewidth]{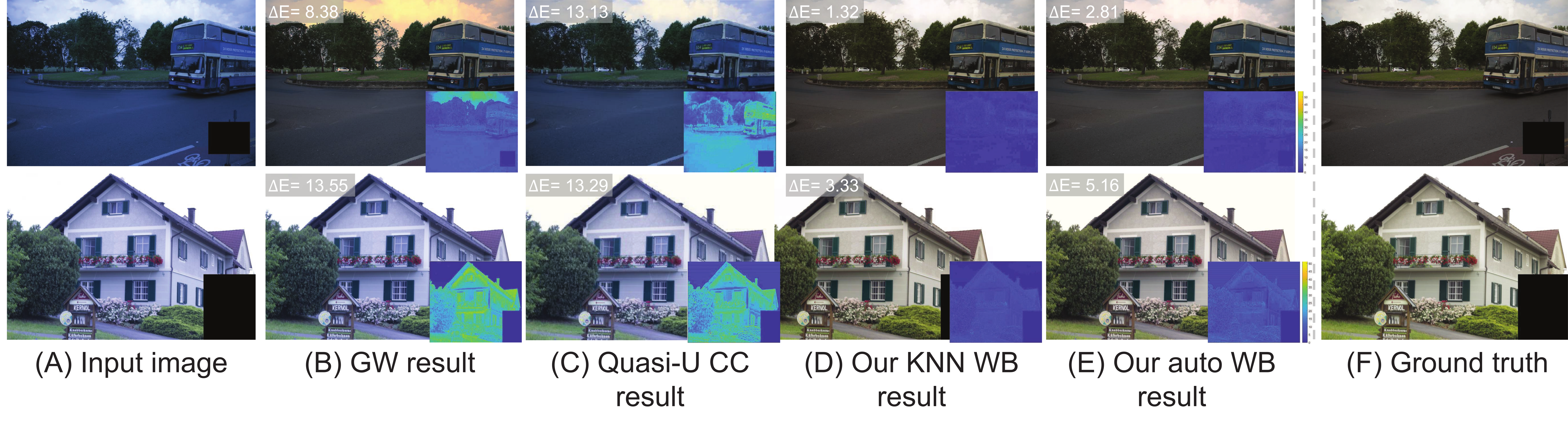}
\vspace{-7mm}
\caption[Qualitative results on our Rendered WB dataset (Chapter \ref{ch:ch7}) and the rendered version of the Cube+ dataset \cite{banic2017unsupervised}.]{Qualitative results on our Rendered WB dataset (Chapter \ref{ch:ch7}) and the rendered version of the Cube+ dataset \cite{banic2017unsupervised}. (A) Input images. (B) Diagonal correction using the GW method \cite{GW}. (C) Diagonal correction using the quasi-unsupervised color constancy (quasi-U CC) method \cite{bianco2019quasi}. (D) Results of the KNN WB method (Chapter \ref{ch:ch7}). (E) Results of applying our rectification function to initial estimation of GW. (F) Ground truth images.}
\label{CIC28:fig:qualitative_results}
\end{figure}

\begin{figure}[!t]
\centering
\includegraphics[width=\linewidth]{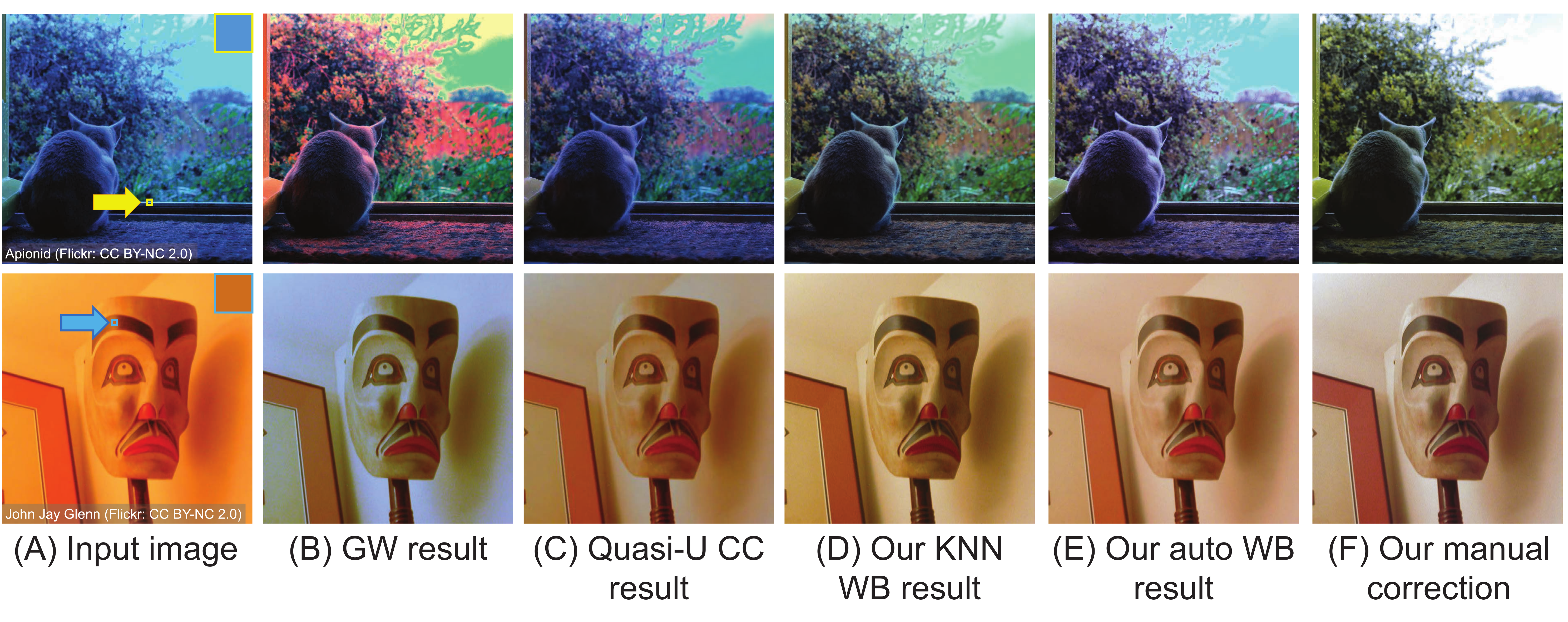}
\vspace{-7mm}
\caption[Our method allows the user to manually select achromatic points in order to improve the results.]{Our method allows the user to manually select achromatic points in order to improve the results. (A) Input images. (B) Results of the GW method \cite{GW}. (C) Results of the quasi-unsupervised color constancy (quasi-U CC) method \cite{bianco2019quasi}. (D) Results of the KNN WB method (Chapter \ref{ch:ch7}). (E) Our results using the GW's estimated illuminant colors. (F) Our results using manually selected achromatic reference points (see the shown arrows).}
\label{CIC28:fig:manual}
\end{figure}

In the user-interactive mode, we use the color-cast vector supplied by the user of some achromatic reference point in the image instead of the estimated color vector $\mat{\gamma}^{'}$. Our method requires $\sim$0.5 seconds to correct a 12-mega-pixel image on an Intel$^\circledR$ Xeon$^\circledR$ E5-1607 @ 3.10 GHz machine, compared to 1.5 seconds required by our method in Chapter \ref{ch:ch7} using a similar machine.  Our approach can significantly improve the results of diagonal-based methods (e.g., \cite{GW, GE, SoG, hu2017fc}) to correct improperly white-balanced images.

\section{Experimental Results}  \label{CIC28:results}
We evaluate our method extensively through quantitative and qualitative comparisons with existing solutions for AWB and user-interactive WB correction. Sec.\ \ref{CIC28:subsec:quanResults} provides quantitative evaluation of our method, while qualitative comparisons are provided in Sec.\ \ref{CIC28:subsec:QaulResults}.

 \begin{table}[!t]
 \caption[Quantitative results on the Rendered WB testing set (Set 2) and the rendered version of the Cube+ dataset \cite{banic2017unsupervised}.]{Quantitative results on the Rendered WB testing set (Set 2) and the rendered version of the Cube+ dataset \cite{banic2017unsupervised}. We applied our rectification function (RF) to the estimated illuminants of different methods. Our RF results are highlighted in gray. The term ``linearized'' refers to applying the standard gamma linearization \cite{anderson1996proposal, ebner2007color} to images before estimating and correcting images. The terms Q1, Q2, and Q3 denote the first, second (median), and third quartile, respectively. The terms MSE and MAE stand for mean square error and mean angular error, respectively.  The best results are highlighted in yellow and boldface.}

 \label{CIC28:Table0}
 \centering
 \scalebox{0.45}{
 \begin{tabular}{|l|c|c|c|c|c|c|c|c|c|c|c|c|}
 \hline
 \multicolumn{1}{|c|}{} & \multicolumn{4}{c|}{\textbf{MSE}} & \multicolumn{4}{c|}{\textbf{MAE}} & \multicolumn{4}{c|}{\textbf{$\boldsymbol{\bigtriangleup}$\textbf{E} 2000}} \\ \cline{2-13}
 \multicolumn{1}{|c|}{\multirow{-2}{*}{\textbf{Method}}} & \textbf{Mean} & \textbf{Q1} & \textbf{Q2} & \textbf{Q3} & \textbf{Mean} & \textbf{Q1} & \textbf{Q2} & \textbf{Q3} & \textbf{Mean} & \textbf{Q1} & \textbf{Q2} & \textbf{Q3}  \\ \hline

 \multicolumn{13}{|c|}{\cellcolor[HTML]{99FF66}\textbf{Testing set of the Rendered WB dataset (Set 2): DSLR and mobile phone cameras (2,881 images)}} \\ \hline
 GW \cite{GW} & 500.18 & 173.69 & 332.75 & 615.40 & 8.89\textdegree & 5.82\textdegree & 8.32\textdegree & 11.33\textdegree & 10.74 & 7.92 & 10.29 & 13.12\\ \hline
 SoG \cite{SoG} & 429.35 & 147.05 & 286.84 & 535.72 & 9.54\textdegree & 5.72\textdegree & 8.85\textdegree & 12.65\textdegree & 10.01 & 7.09 & 9.85 & 12.69\\ \hline
 FC4 \cite{hu2017fc} & 662.53  & 304.88 & 524.42 & 817.57 & 8.92\textdegree & 5.94\textdegree & 8.03\textdegree & 10.84\textdegree & 12.12 &  8.94& 11.79 &  14.76 \\ \hdashline
 GW (linearized) \cite{GW} & 469.86 & 163.07 & 312.28 & 574.85 & 8.61\textdegree & 5.44\textdegree & 7.94\textdegree & 10.93\textdegree & 10.68 & 7.70 & 10.13 & 13.15\\ \hline

 SoG (linearized) \cite{SoG} & 393.85 & 137.21 & 267.37 & 497.40 & 8.96\textdegree & 5.31\textdegree & 8.26\textdegree & 11.97\textdegree & 9.81 & 6.87 & 9.67 & 12.46\\ \hline

 FC4 (linearized) \cite{hu2017fc}  &  505.30 & 142.46 & 307.77 & 635.35 & 10.37\textdegree
 & 5.31\textdegree & 9.26\textdegree & 14.15\textdegree & 10.82 & 7.39 & , 10.64 &  13.77   \\ \hdashline
 \cellcolor[HTML]{CCCCCC}GW \cite{GW} + our RF &  207.13 &  46.71 &  111.89 & 230.10 & 5.35\textdegree & 2.89\textdegree &  4.59\textdegree  &  6.84\textdegree & 6.74 & 4.45 & 6.15 & 8.45\\ \hline
 \cellcolor[HTML]{CCCCCC}SoG \cite{SoG}  + our RF & 256.10 &  55.93 & 132.09 & 266.61 & 6.25\textdegree & 3.28\textdegree &  5.20\textdegree & 8.20\textdegree & 7.27 & 4.77 & 6.61 & 9.05  \\\hline
 \cellcolor[HTML]{CCCCCC}FC4 \cite{hu2017fc} + our RF & 303.99 &  50.01 & 118.88 &  298.44 & 6.61\textdegree & 2.99\textdegree & 4.99\textdegree & 8.40\textdegree & 7.28  & 4.30 & 6.22 &  9.33 \\ \hdashline
 KNN WB (Chapter \ref{ch:ch7}) & \cellcolor[HTML]{\bestcolor} \textbf{171.09} &  \cellcolor[HTML]{\bestcolor} \textbf{37.04} &  \cellcolor[HTML]{\bestcolor} \textbf{87.04} &  \cellcolor[HTML]{\bestcolor} \textbf{190.88} &  \cellcolor[HTML]{\bestcolor} \textbf{4.48\textdegree} &  \cellcolor[HTML]{\bestcolor} \textbf{2.26\textdegree} &  \cellcolor[HTML]{\bestcolor} \textbf{3.64\textdegree} &  \cellcolor[HTML]{\bestcolor} \textbf{5.95\textdegree} &  \cellcolor[HTML]{\bestcolor} \textbf{5.60} &  \cellcolor[HTML]{\bestcolor} \textbf{3.43} &  \cellcolor[HTML]{\bestcolor} \textbf{4.90} &  \cellcolor[HTML]{\bestcolor} \textbf{7.06}  \\ \hline
 \multicolumn{13}{|c|}{\cellcolor[HTML]{99FF66}\textbf{Rendered Cube+ dataset with different WB settings (10,242 images)}} \\ \hline
 GW \cite{GW} & 312.62 &  55.16 & 159.63 & 358.02 & 6.85\textdegree & 3.08\textdegree & 5.76\textdegree & 9.70\textdegree & 9.01 & 5.35 & 8.38 & 12.08 \\ \hline

 SoG \cite{SoG}  &  269.31 & 21.92 & 90.37 & 312.02 & 6.69\textdegree & 2.3\textdegree & 4.63\textdegree & 9.62\textdegree & 7.70 & 3.40 & 6.38 & 11.07\\ \hline
 FC4 \cite{hu2017fc} &  410.01 & 79.26 & 219.05 & 505.71 & 6.7\textdegree & 3.26\textdegree & 5.45\textdegree & 8.7\textdegree & 10.4 & 6.51 & 9.73 & 13.43 \\ \hdashline
 GW (linearized) \cite{GW}  &  244.59 & 32.58 & 121.42 & 300.99 & 6.37\textdegree & 2.51\textdegree & 5.13\textdegree & 9.09\textdegree & 8.05 & 4.18 & 7.25 & 11.08 \\ \hline

 SoG (linearized) \cite{SoG}  & 275.33  & 17.16 & 67.49  & 309.97 & 6.66\textdegree & 2.08\textdegree & 4.17\textdegree & 9.58\textdegree & 7.57 & 3.00 & 5.73 & 11.05 \\ \hline

 FC4 (linearized) \cite{hu2017fc}  & 371.9 & 79.15 & 213.41 & 467.33 & 6.49\textdegree & 3.34\textdegree & 5.59\textdegree & 8.59\textdegree & 10.38 & 6.6 & 9.76 & 13.26 \\ \hdashline
 \cellcolor[HTML]{CCCCCC}GW \cite{GW} + our RF  & \cellcolor[HTML]{\bestcolor} \textbf{159.88} & 21.94 & 54.76 & 125.02 & 4.64\textdegree & 2.12\textdegree & 3.64\textdegree & 5.98\textdegree & 6.2 & 3.28 & 5.17 & 7.45 \\ \hline
 \cellcolor[HTML]{CCCCCC}SoG \cite{SoG} + our RF  & 226.83 & 20.01 & 58.61 & 165.03 & 5.33\textdegree & 2.1\textdegree & 3.83\textdegree & 6.97\textdegree & 6.61 & 3.17 & 5.38 & 8.56 \\ \hline
 \cellcolor[HTML]{CCCCCC}FC4 \cite{hu2017fc} + our RF & 175.73 & \cellcolor[HTML]{\bestcolor} \textbf{17.8} & \cellcolor[HTML]{\bestcolor} \textbf{43.65} & \cellcolor[HTML]{\bestcolor} \textbf{114.65} & 4.67\textdegree & 1.89\textdegree & \cellcolor[HTML]{\bestcolor} \textbf{3.10\textdegree} & 5.48\textdegree & 5.7 & \cellcolor[HTML]{\bestcolor} \textbf{2.95} & 4.63 & 7.05 \\ \hdashline
KNN WB (Chapter \ref{ch:ch7}) &  194.98 &  27.43 &  57.08 &  118.21 &  \cellcolor[HTML]{\bestcolor} \textbf{4.12\textdegree} &  \cellcolor[HTML]{\bestcolor} \textbf{1.96\textdegree} & 3.17\textdegree &  \cellcolor[HTML]{\bestcolor} \textbf{5.04\textdegree} &  \cellcolor[HTML]{\bestcolor} \textbf{5.68} &  3.22 &  \cellcolor[HTML]{\bestcolor} \textbf{4.61} &  \cellcolor[HTML]{\bestcolor} \textbf{6.70} \\\hline
 \end{tabular}
 }
 \end{table}

\begin{figure}[t]
\includegraphics[width=\linewidth]{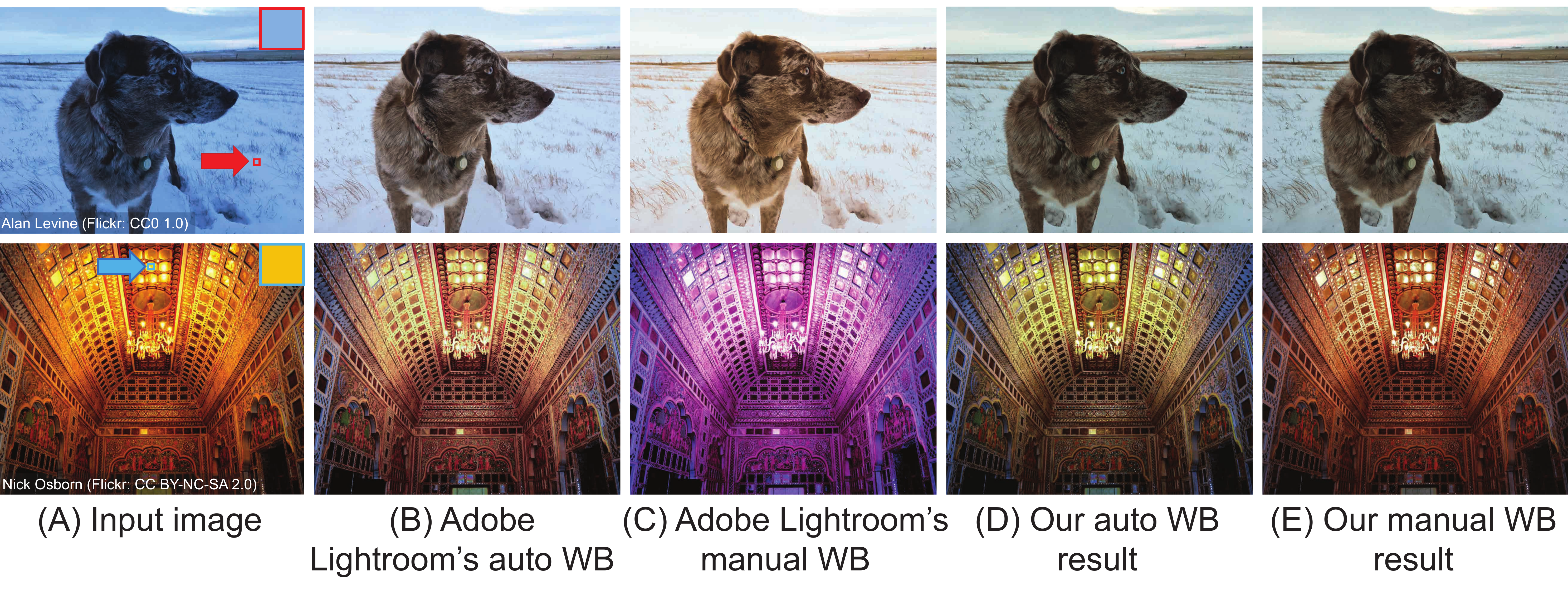}
\vspace{-7mm}
\caption[Comparison with Adobe Lightroom WB correction.]{Comparison with Adobe Lightroom WB correction. (A) Input image. (B) and (C) Adobe Lightroom's auto WB and manual WB correction results, respectively. (D) and (E) Our auto and manual WB correction results, respectively.}
\label{CIC28:fig:lightroom}
\end{figure}

\subsection{Quantitative Evaluation}  \label{CIC28:subsec:quanResults}

We evaluated our AWB correction results using $\sim$13,000 testing images from the Rendered WB testing set (Set 2) (Chapter \ref{ch:ch7}). In addition, we rendered the Cube+ dataset \cite{banic2017unsupervised} in the same way of rendering our dataset in Chapter \ref{ch:ch7}. We also used this rendered version of the Cube+ dataset in our evaluation. As mentioned in Sec. \ref{CIC28:sec:method}, any off-the-shelf illumination estimation method $E$ can be used. We tested different illuminant estimation methods in our AWB framework.  Specifically, we utilized the following methods: the GW \cite{GW}, the SoG \cite{SoG}, and the FC4 \cite{hu2017fc} methods. In each experiment, we use the training data of our Rendered WB dataset (Chapter \ref{ch:ch7}) to compute our rectification functions, as described in Sec.\ \ref{CIC28:subsec:rectfunc}. For each of the illuminant estimation methods, we compare our results with the diagonal correction with and without linearizing the testing images. The linearizing process was performed using the standard de-gamma linearization operation \cite{anderson1996proposal, ebner2007color}. We include this linearization process in our comparisons as it is a common misconception that a simple gamma linearization can remove the nonlinearity applied by cameras.

For the sake of completeness, we also compare our results with our recent nonlinear method for post-capture KNN WB correction proposed in Chapter \ref{ch:ch7}).  Table \ref{CIC28:Table0} shows the first, second, and third quantile and the mean of the error values obtained by each method. We followed the evaluation metrics used in Chapter \ref{ch:ch7}, which are: (i) MSE, (ii) MAE, and (iii) $\boldsymbol{\bigtriangleup}$\textbf{E} 2000 \cite{sharma2005ciede2000}.

As shown in Table \ref{CIC28:Table0}, our rectification function significantly improves the results of diagonal-based methods and achieves results on a par with the WB-sRGB method on both testing sets. As can be seen from the results, our method reduces the MSE by $\sim$55\%, the MAE by $\sim$32\%, and the $\bigtriangleup$E 2000 by $\sim$35 \% on average compared to the gamma linearization process, which reduces the MSE by $\sim$15\%, the MAE by $\sim$5\%, and the $\bigtriangleup$E 2000 by $\sim$5\% on average.

\subsection{Qualitative Evaluation}  \label{CIC28:subsec:QaulResults} 
We qualitatively evaluated our method against different methods, including a commercial photo-editing software, for auto and user-interactive WB correction. Figure \ref{CIC28:fig:qualitative_results} shows a comparison between our results and the diagonal correction of two illuminant estimation methods---namely, the GW method \cite{GW} and the quasi-unsupervised color constancy method \cite{bianco2019quasi}. As shown, our AWB is superior to diagonal WB and achieves similar results to our KNN WB method proposed in Chapter \ref{ch:ch7}.

In contrast to the KNN WB method (Chapter \ref{ch:ch7}), our method allows interactive correction to improve the results by manually adjusting the color-cast color. Figure \ref{CIC28:fig:manual} shows that this manual adjustment feature produces arguably visually superior results compared to the KNN WB method.

We further compared our method against Adobe Lightroom, as it is one of the most common photo-editing software programs that provide the same manual WB correction feature. As can be seen in Fig. \ref{CIC28:fig:lightroom}, our method produces perceptibly superior results in comparison with Adobe Lightroom's results for both auto and manual WB correction.

\section{Summary} 
We have introduced an interactive WB method for use on camera-rendered images which allows the user to directly specify scene points to be used for white balancing.  Previously, this type of interaction could be performed only by photo-editing software operating on raw images.   We have enabled this feature for camera-rendered images that already have a white-balance correction applied as well as additional photo-finishing.  Our method works by efficiently computing nonlinear color-correction mappings based on user-supplied color-cast vectors directly from the camera-rendered image.  We also showed how our method can easily perform auto-WB correction in a camera-rendered image.  Our method enables a new photo-editing feature for color manipulation.

\chapter{Deep White-Balance Editing \label{ch:ch10}}
In this chapter, we introduce a deep learning approach\footnote{Work done while the author was an intern at Samsung AI Center -- Toronto; This work was published in \cite{afifi2020deep, afifi2021apparatus}: Mahmoud Afifi and Michael S. Brown. Deep White-Balance Editing. In IEEE Conference on Computer Vision and Pattern Recognition (CVPR), 2020.} to realistically edit an sRGB image's white balance. As discussed in previous chapters, cameras capture sensor images that are rendered by their ISPs to a sRGB color space encoding. The ISP rendering begins with a white-balance procedure that is used to remove the color cast of the scene's illumination.  The ISP then applies a series of nonlinear color manipulations to enhance the visual quality of the final sRGB image.  In Chapter \ref{ch:ch7}, we showed that sRGB images that were rendered with the incorrect white balance cannot be easily corrected due to the ISP's nonlinear rendering.  The work in Chapter \ref{ch:ch7} proposed a KNN solution based on tens of thousands of image pairs.  In this chapter, we propose to solve this problem with a DNN architecture trained in an end-to-end manner to learn the correct white balance.  Our DNN maps an input image to two additional white-balance settings corresponding to indoor and outdoor illuminations. Our solution not only is more accurate than the KNN approach in terms of correcting a wrong white-balance setting but also provides the user the freedom to edit the white balance in the sRGB image to other illumination settings. The source code of this work is available on GitHub: \href{https://github.com/mahmoudnafifi/Deep_White_Balance}{https://github.com/mahmoudnafifi/Deep$\_$White$\_$Balance}.

\section{Introduction}\label{deepWB:sec:intro_related}

\begin{figure}
\includegraphics[width=\linewidth]{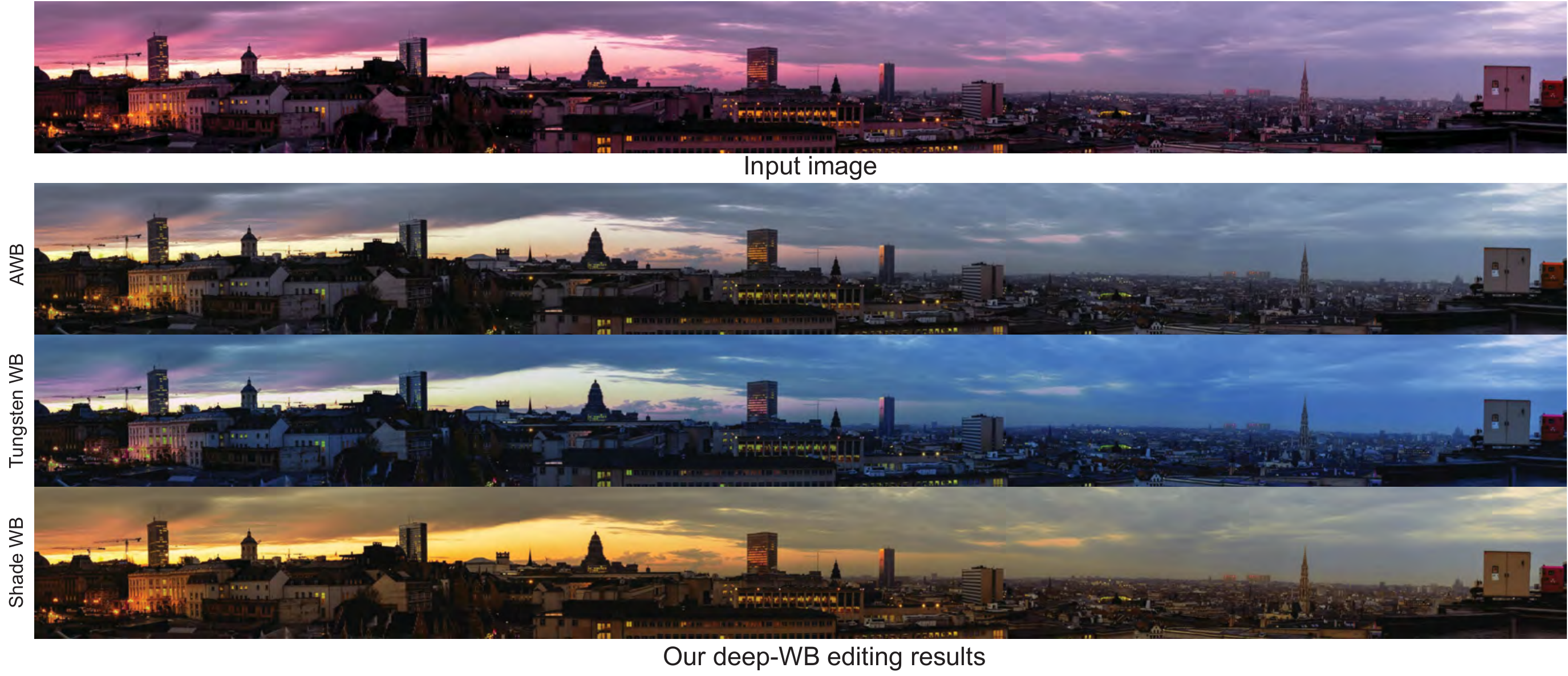}
\vspace{-7mm}
\caption[Our deep white-balance editing framework produces compelling results and generalizes well to images outside our training data (e.g., image above taken from an Internet photo repository).]{Our deep white-balance editing framework produces compelling results and generalizes well to images outside our training data (e.g., image above taken from an Internet photo repository). Top: input image captured with a wrong WB setting. Bottom: our framework's AWB, Incandescent WB, and Shade WB results. Photo credit: \textit{M@tth1eu} Flickr--CC BY-NC 2.0.}
\label{deepWB:fig:teaser}
\end{figure}

While the goal of WB is intended to normalize the effect of the scene's illumination, ISPs often incorporate aesthetic considerations in their color rendering based on photographic preferences. Such preferences do not always conform to the white light assumption and can vary based on different factors, such as cultural preference and scene content~\cite{scuello2004museum, cheng2016two, FFCC, hu2018exposure}.

Most digital cameras provide an option to adjust the WB settings during image capturing. However, once the WB setting has been selected and the image is fully processed by the ISP to its final sRGB encoding it becomes challenging to perform WB editing without access to the original unprocessed raw-RGB image. This problem becomes even more difficult if the WB setting was wrong, which results in a strong color cast in the final sRGB image.

The ability to edit the WB of an sRGB image not only is useful from a photographic perspective but also can be beneficial for computer vision applications, such as object recognition, scene understanding, and color augmentation~\cite{barnard2002comparison, gijsenij2011computational}. Our study in Chapter~\ref{ch:ch8} showed that images captured with an incorrect WB setting produce a similar effect of an untargeted adversarial attack for DNN models.

\paragraph{Contribution}~We present a novel deep learning framework that allows realistic post-capture WB editing of sRGB images. Our framework consists of a single encoder network that is coupled with three decoders targeting the following WB settings: (1) a ``correct'' AWB setting; (2) an indoor WB setting; (3) an outdoor WB setting.   The first decoder allows an sRGB image that has been incorrectly white-balanced image to be edited to have the correct WB.  This is useful for the task of post-capture WB correction.  The additional indoor and outdoor decoders provide users the ability to produce a wide range of different WB appearances by blending between the two outputs.  This supports photographic editing tasks to adjust an image's aesthetic WB properties.  We provide extensive experiments to demonstrate that our method generalizes well to images outside our training data and achieves state-of-the-art results for both tasks.

\begin{figure}[t]
\includegraphics[width=\linewidth]{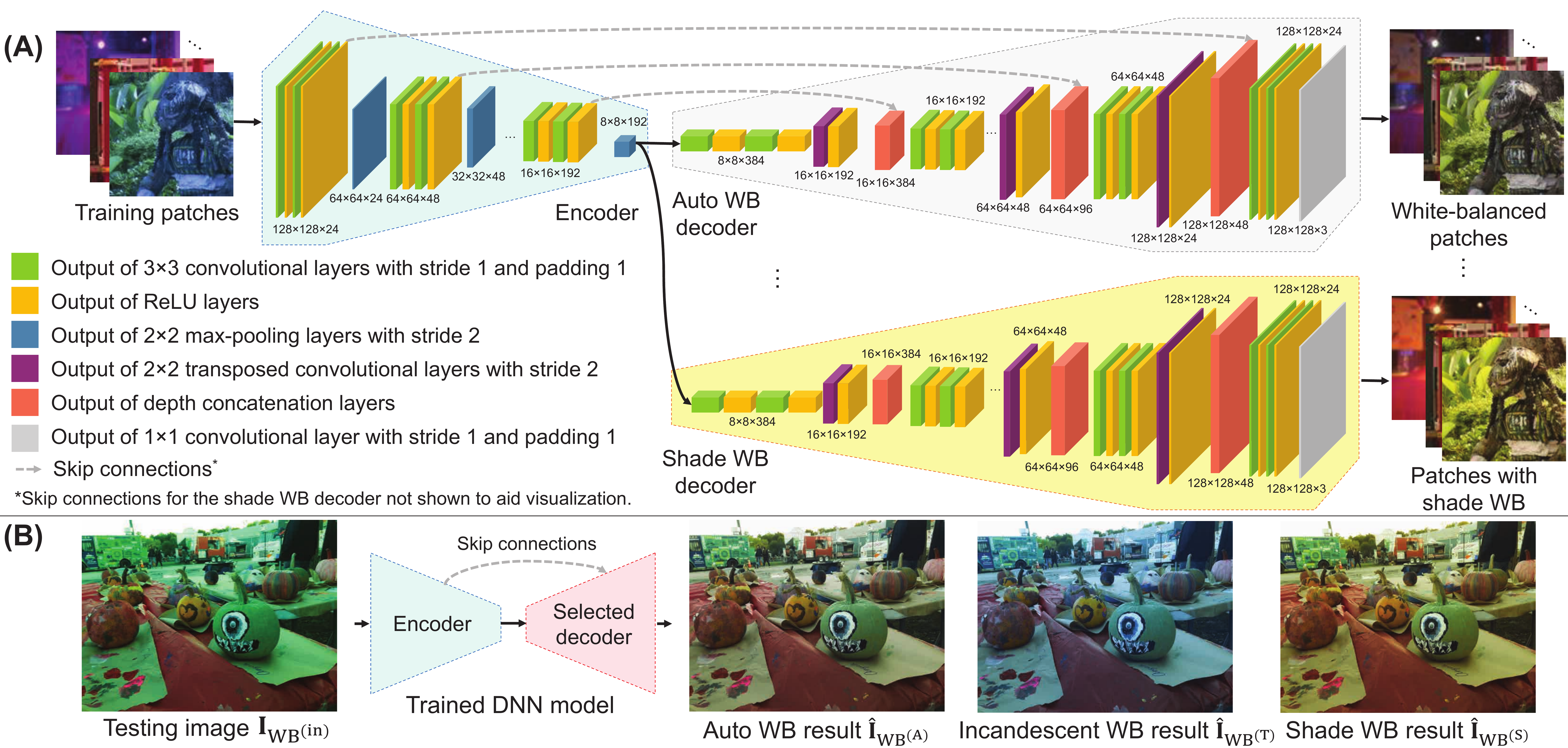}
\vspace{-7mm}
\caption[Proposed multi-decoder framework for sRGB WB editing.]{Proposed multi-decoder framework for sRGB WB editing. (A) Our proposed framework consists of a single encoder and multiple decoders. The training process is performed in an end-to-end manner, such that each decoder ``re-renders'' the given training patch with a specific WB setting, including AWB. For training, we randomly select image patches from the Rendered WB dataset (Chapter \ref{ch:ch7}). (B) Given a testing image, we produce the targeted WB setting by using the corresponding trained decoder. }
\label{deepWB:fig:main}
\end{figure}

\section{Methodology}\label{deepWB:sec:method}

\subsection{Problem formulation} \label{deepWB:subsec:formulation}
Given an sRGB image, $\mat{I}_{\text{WB}^{(\text{in})}}$, rendered through an unknown camera ISP with an arbitrary WB setting $\text{WB}^{(\text{in})}$, our goal is to edit its colors to appear as if it were re-rendered with a target WB setting $\text{WB}^{(t)}$.

As mentioned in Sec.~\ref{deepWB:sec:intro_related}, our task can be accomplished accurately if the original unprocessed raw-RGB image is available. If we could recover the unprocessed raw-RGB values, we can change the WB setting $\text{WB}^{(\text{in})}$ to $\text{WB}^{(t)}$, and then re-render the image back to the sRGB color space with a software-based ISP. This ideal process can be described by the following equation:
\begin{equation}
\label{deepWB:eq1}
\mat{I}_{\text{WB}^{(t)}} =  G\left(F  \left(\mat{I}_{\text{WB}^{(\text{in})}}\right)\right),
\end{equation}

\noindent where $F: \mat{I}_{\text{WB}^{(\text{in})}} \rightarrow  \mat{D}_{\text{WB}^{(\text{in})}}$ is an unknown reconstruction function that reverses the camera-rendered sRGB image $\mat{I}$ back to its corresponding raw-RGB image $\mat{D}$ with the current $\text{WB}^{(\text{in})}$ setting applied and $G: \mat{D}_{\text{WB}^{(\text{in})}} \rightarrow  \mat{I}_{\text{WB}^{(t)}}$ is an unknown camera rendering function that is responsible for editing the WB setting and re-rendering the final image.

\subsection{Method Overview} \label{deepWB:subsec:overview}

\begin{figure}[t]
\includegraphics[width=\linewidth]{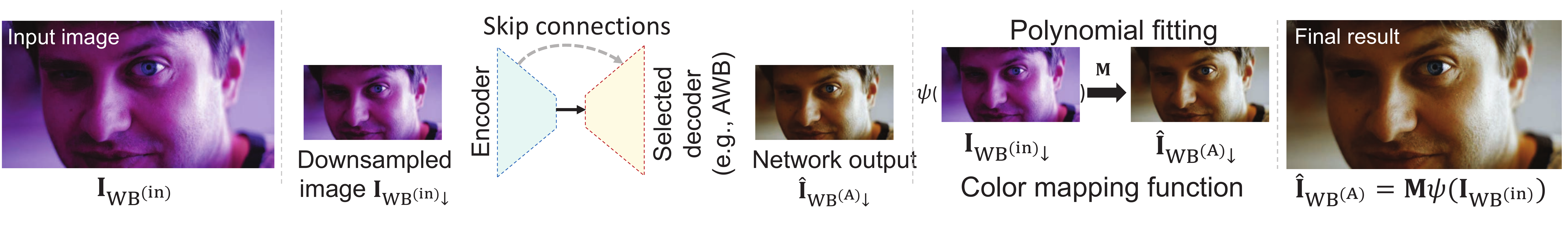}
\vspace{-7mm}
\caption[We consider the runtime performance of our method to be able to run on limited computing resources ($\sim$1.5 seconds on a single CPU to process a 12-megapixel image).]{We consider the runtime performance of our method to be able to run on limited computing resources ($\sim$1.5 seconds on a single CPU to process a 12-megapixel image). First, our DNN processes a downsampled version of the input image, and then we apply a global color mapping to produce the output image in its original resolution. Shown input image is rendered from the MIT-Adobe FiveK dataset \cite{bychkovsky2011learning}.}
\label{deepWB:fig:entier_process}
\end{figure}

Our goal is to model the functionality of $G\left(F\left(\cdot\right)\right)$ to generate $\mat{I}_{\text{WB}^{(t)}}$. We first analyze how the functions $G$ and $F$ cooperate to produce $\mat{I}_{\text{WB}^{(t)}}$. From Eq. \ref{deepWB:eq1}, we see that the function $F$ transforms the input image $\mat{I}_{\text{WB}^{(\text{in})}}$ into an intermediate representation (i.e., the raw-RGB image with the captured WB setting), while the function $G$ accepts this intermediate representation and renders it with the target WB setting to an sRGB color space encoding.

Due to the nonlinearities applied by the ISP's rendering chain, we can think of $G$ as a hybrid function that consists of a set of sub-functions, where each sub-function is responsible for rendering the intermediate representation with a specific WB setting.

Our ultimate goal is \textit{not} to reconstruct/re-render the original raw-RGB values, but rather to generate the final sRGB image with the target WB setting ${\text{WB}^{(t)}}$. Therefore, we can model the functionality of $G\left(F\left(\cdot\right)\right)$ as an encoder/decoder scheme. Our encoder $f$ transfers the input image into a latent representation, while each of our decoders ($g_1$, $g_2$, ...) generate the final images with a different WB setting. Similar to Eq. \ref{deepWB:eq1}, we can formulate our framework as follows:

\begin{equation}
\label{deepWB:eq2}
\hat{\mat{I}}_{\text{WB}^{(\text{t})}} = g_t\left(f \left( \mat{I}_{\text{WB}^{(\text{in})}} \right)\right),
\end{equation}

\noindent where $f: \mat{I}_{\text{WB}^{(\text{in})}} \rightarrow \mat{\mathcal{Z}}$, $g_t: \mat{\mathcal{Z}} \rightarrow \hat{\mat{I}}_{\text{WB}^{(\text{t})}}$, and $\mathcal{Z}$ is an intermediate representation (i.e., latent representation) of the original input image $\mat{I}_{\text{WB}^{(\text{in})}}$.

Our goal is to make the functions $f$ and $g_t$ independent, such that changing $g_t$ with a new function $g_y$ that targets a different WB $y$, does not require any modification in $f$, as it is the case in Eq. \ref{deepWB:eq1}.

In our work, we target three different WB settings: (i) $\text{WB}^{(\text{A})}$: AWB -- representing the correct lighting of the captured image's scene; (ii) $\text{WB}^{(\text{T})}$: Tungsten/ Incandescent -- representing WB for indoor lighting; and (iii) $\text{WB}^{(\text{S})}$: Shade -- representing WB for outdoor lighting.  This gives rise to three different decoders ($g_A$, $g_T$, and $g_S$) that are responsible for generating output images that correspond to AWB, Incandescent WB, and Shade WB.

The Incandescent and Shade WB are specifically selected based on the color properties.  This can be understood when considering the illuminations in terms of their correlated color temperatures. For example, Incandescent and Shade WB settings are correlated to $2850$ Kelvin ($K$) and $7500K$ color temperatures, respectively. This wide range of illumination color temperatures consider the range of pleasing illuminations \cite{kruithof1941tubular, petrulis2018exploring}. Moreover, the wide color temperature range between Incandescent and Shade allows the approximation of images with color temperatures within this range by interpolation. The details of this interpolation process are explained in Sec. \ref{deepWB:subsec:testing_phase}. Note that there is no fixed correlated color temperature for the AWB mode, as it changes based on the input image's lighting conditions.

\subsection{Multi-Decoder Architecture} \label{deepWB:subsec:architecture}

An overview of our DNN's architecture is shown in Fig. \ref{deepWB:fig:main}. We use a U-Net architecture \cite{ronneberger2015u} with multi-scale skip connections between the encoder and decoders. Our framework consists of two main units:  the first is a 4-level encoder unit that is responsible for extracting a multi-scale latent representation of our input image; the second unit includes three 4-level decoders.  Each unit has a different bottleneck and transpose convolutional (conv) layers. At the first level of our encoder and each decoder, the conv layers have 24 channels. For each subsequent level, the number of channels is doubled (i.e., the fourth level has 192 channels for each conv layer).

\subsection{Training Phase} \label{deepWB:subsec:training}

\paragraph{Training Data}

We adopt our Rendered WB dataset produced in Chapter~\ref{ch:ch7} to train and validate our model. This dataset includes $\sim$65,000 sRGB images rendered by different camera models and with different WB settings, including the Shade and Incandescent settings. For each image, there is also a corresponding ground truth image rendered with the correct WB setting (considered to be the correct AWB result). This dataset consists of two subsets: training set (Set 1) and testing set (Set 2). The training set (Set 1) is divided into three folds -- two for training and one for validation. Specifically, we randomly selected 12,000 training images from two folds and 2,000 validation images from the remaining fold. For each training image, we have three ground truth images rendered with: (i) the correct WB (denoted as AWB), (ii) Shade WB, and (iii) Incandescent WB. 
The validation results on Set 1 of the Rendered WB dataset are reported in Sec. \ref{deepWB:sec:results}.

\begin{figure}
\includegraphics[width=\linewidth]{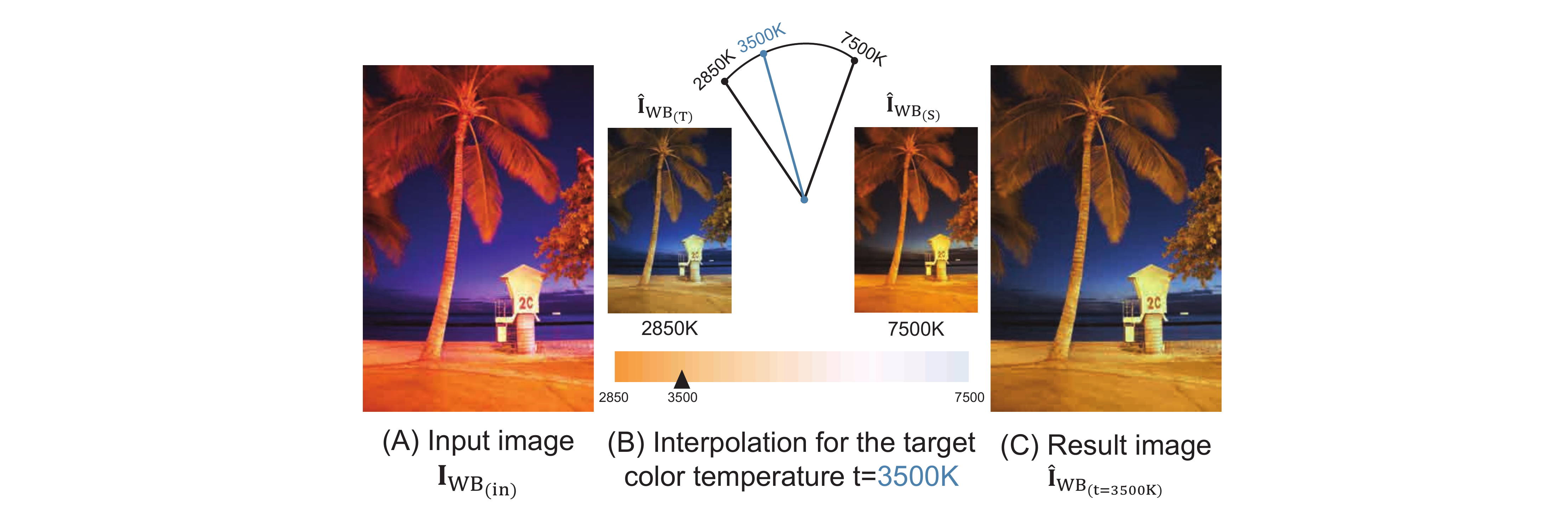}
\vspace{-7mm}
\caption[In addition to our AWB correction, we train our framework to produce two different color temperatures (i.e., Incandescent and Shade WB settings).]{In addition to our AWB correction, we train our framework to produce two different color temperatures (i.e., Incandescent and Shade WB settings). We interpolate between these settings to produce images with other color temperatures. (A) Input image. (B) Interpolation process. (C) Final result. Shown input image is taken from the rendered version of the MIT-Adobe FiveK dataset.}
\label{deepWB:fig:tempblending}
\end{figure}

\paragraph{Data Augmentation} We also augment the training images by rendering an additional 1,029 raw-RGB images, of the same scenes included in our Rendered WB dataset (Chapter\ \ref{ch:ch7}), but with random color temperatures. At each epoch, we randomly select four $128\!\times\!128$ patches from each training image and their corresponding ground truth images for each decoder and apply geometric augmentation (rotation and flipping) as an additional data augmentation to avoid overfitting.

\paragraph{Loss Function}
We trained our model to minimize the $\texttt{L}_1$-norm loss function between the reconstructed and ground truth patches:

\begin{equation}
\label{deepWB:eq3} \sum_{i}\sum_{p=1}^{3hw}\left|\mat{P}_{\text{WB}^{(i)}}(p) - \mat{C}_{\text{WB}^{(i)}}(p)\right|,
\end{equation}

\noindent where $h$ and $w$ denote the patch's height and width, and $p$ indexes into each pixel of the training patch $\mat{P}$ and the ground truth camera-rendered patch $\mat{C}$, respectively. The index $i\in\{\text{A},\text{T},\text{S}\}$ refers to the three target WB settings. We also have examined the squared $\texttt{L}_2$-norm loss function and found that both loss functions work well for our task.

\begin{figure*}[t]
\includegraphics[width=\linewidth]{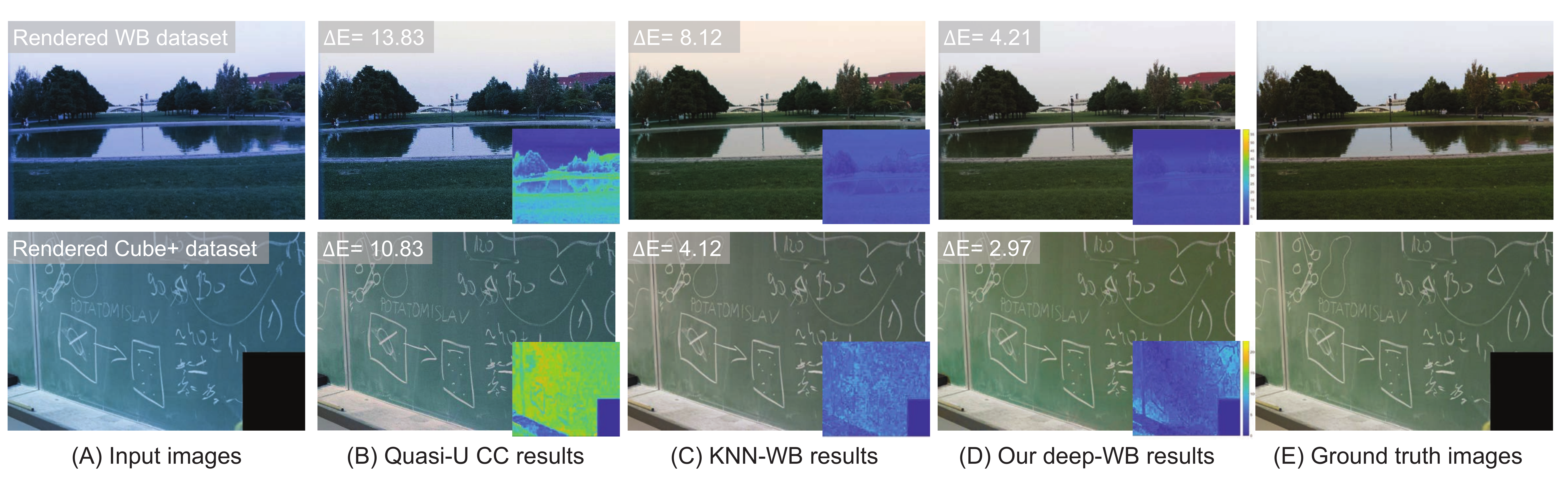}
\vspace{-7mm}
\caption{Qualitative comparison of AWB correction. (A) Input images. (B) Results of quasi-U CC \cite{bianco2019quasi}. (C) Results of KNN WB (Chapter \ref{ch:ch7}). (D) Our results. (E) Ground truth images. Shown input images are taken from our Rendered WB dataset (Chapter\ \ref{ch:ch7}) and the rendered version of Cube+ dataset \cite{banic2017unsupervised} (see Chapter\ \ref{ch:ch9} for more details about the rendered version of Cube+ dataset).}
\label{deepWB:fig:qualitative_AWB}
\end{figure*}

\paragraph{Training Parameters}
We initialized the weights of the conv layers using He's initialization \cite{he2015delving}. The training process is performed for 165,000 iterations using the Adam optimizer \cite{kingma2014adam}, with a decay rate of gradient moving average $\beta_1=0.9$ and a decay rate of squared gradient moving average $\beta_2=0.999$. We used a learning rate of $10^{-4}$ and reduced it by 0.5 every 25 epochs. The mini-batch size was 32 training patches per iteration.

\subsection{Testing Phase} \label{deepWB:subsec:testing_phase}

\paragraph{Color mapping procedure}
Our DNN model is a fully convolutional network and is able to process input images in their original dimensions with the restriction that the dimensions should be multiples of $2^4$, as we use 4-level encoder/decoders with $2\!\times\!2$ max-pooling and transpose conv layers. However, to ensure a consistent run time for any sized input images, we resize all input images to a maximum dimension of 656 pixels. Our DNN is applied on this resized image to produce image $\hat{\mat{I}}_{\text{WB}^{(i)}\downarrow}$ with the target WB setting $i\in\{\text{A},\text{T},\text{S}\}$.

We then compute a color mapping function between our resized input and output image. As done in Chapter \ref{ch:ch7}, we computed a polynomial mapping matrix $\mat{M}$ that globally maps values of $\psi\left(\mat{I}_{\text{WB}^{(\text{in})}\downarrow}\right)$ to the colors of our generated image $\hat{\mat{I}}_{\text{WB}^{(i)}\downarrow}$, where $\psi(\cdot)$ is a polynomial kernel function that maps the image's RGB vectors to a higher 11-dimensional space. This mapping matrix $\mat{M}$ can be computed in a closed-form solution, as demonstrated in Chapters \ref{ch:ch7} and \ref{ch:ch8}.

Once $\mat{M}$ is computed, we obtain our final result in the same input image resolution using the following equation:

\begin{equation}
\label{deepWB:eq5}
\hat{\mat{I}}_{\text{WB}^{(\text{i})}} =  \mat{M}\psi\left(\mat{I}_{\text{WB}^{(\text{in})}}\right).
\end{equation}

Figure \ref{deepWB:fig:entier_process} illustrates our color mapping procedure. Our method requires $\sim$1.5 seconds on an Intel Xeon E5-1607 @ 3.10GHz machine with 32 GB RAM to process a 12-megapixel image for a selected WB setting.

We note that an alternative strategy is to compute the color polynomial mapping matrix directly~\cite{schwartz2018deepisp}).  We conducted preliminary experiments and found that estimating the polynomial matrix directly was less robust than generating the image itself followed by fitting a global polynomial function. The reason is that having small errors in the estimated polynomial coefficients can lead to noticeable color errors (e.g., out-of-gamut values), whereas small errors the estimated image were ameliorated by the global fitting.

\paragraph{Editing by User Manipulation}

Our framework allows the user to choose between generating result images with the three available WB settings (i.e., AWB, Shade WB, and Incandescent WB).  Using the Shade and Incandescent WB, the user can edit the image to a specific WB setting in terms of color temperature, as explained in the following.

To produce the effect of a \textit{new} target WB setting with a color temperature $t$ that is not produced by our decoders, we can interpolate between our generated images with the Incandescent and Shade WB settings. We found that a simple linear interpolation was sufficient for this purpose. This operation is described by the following equations:

\begin{equation}
\label{deepWB:eq6}
\hat{\mat{I}}_{\text{WB}^{(t)}} = b \text{ }\hat{\mat{I}}_{\text{WB}^{(\text{T})}} + (1-b) \text{ } \hat{\mat{I}}_{\text{WB}^{(\text{S})}},
\end{equation}

\noindent where $\hat{\mat{I}}_{\text{WB}^{(\text{T})}}$ and $\hat{\mat{I}}_{\text{WB}^{(\text{S})}}$ are our produced images with Incandescent and Shade WB settings, respectively, and $b$ is the interpolation ratio that is given by $\frac{1/t - 1/t(S)}{1/t(T) - 1/t(S)}$. Figure \ref{deepWB:fig:tempblending} shows an example.

\begin{table*}[t]
\caption[AWB results using our Rendered WB dataset (Chapter\ \ref{ch:ch7}) and the rendered version of the Cube+ dataset \cite{banic2017unsupervised}.]{AWB results using our Rendered WB dataset (Chapter\ \ref{ch:ch7}) and the rendered version of the Cube+ dataset \cite{banic2017unsupervised}. We report the mean, first, second (median), and third quartile (Q1, Q2, and Q3) of MSE, MAE, and $\boldsymbol{\bigtriangleup}$\textbf{E} 2000 \cite{sharma2005ciede2000}. For all diagonal-based methods, gamma linearization \cite{anderson1996proposal, ebner2007color} is applied. The top results are indicated with yellow and boldface.}
\label{deepWB:Table0}
\centering
\scalebox{0.65}{
\begin{tabular}{|l|c|c|c|c|c|c|c|c|c|c|c|c|}
\hline
\multicolumn{1}{|c|}{} & \multicolumn{4}{c|}{\textbf{MSE}} & \multicolumn{4}{c|}{\textbf{MAE}} & \multicolumn{4}{c|}{\textbf{$\boldsymbol{\bigtriangleup}$\textbf{E} 2000}} \\ \cline{2-13}
\multicolumn{1}{|c|}{\multirow{-2}{*}{\textbf{Method}}} & \textbf{Mean} & \textbf{Q1} & \textbf{Q2} & \textbf{Q3} & \textbf{Mean} & \textbf{Q1} & \textbf{Q2} & \textbf{Q3} & \textbf{Mean} & \textbf{Q1} & \textbf{Q2} & \textbf{Q3}  \\ \hline

\multicolumn{13}{|c|}{\cellcolor[HTML]{D4EBF2}\textbf{Rendered WB dataset: Set 1 (21,046 validation images)}} \\ \hline

FC4 \cite{hu2017fc} & 179.55 & 33.89 & 100.09 & 246.50 & 6.14\textdegree & 2.62\textdegree & 4.73\textdegree & 8.40\textdegree &  6.55 & 3.54 & 5.90 & 8.94  \\\hline

Quasi-U CC \cite{bianco2019quasi}   & 172.43 & 33.53 & 97.9 & 237.26 & 6.00\textdegree & 2.79\textdegree & 4.85\textdegree & 8.15\textdegree &	6.04 & 3.24 & 5.27 & 8.11 \\ \hline

KNN WB (Chapter \ref{ch:ch7}) &  \cellcolor[HTML]{\bestcolor}\textbf{77.79} &  13.74& \cellcolor[HTML]{\bestcolor} \textbf{39.62}& \cellcolor[HTML]{\bestcolor}\textbf{94.01 }&  \cellcolor[HTML]{\bestcolor} \textbf{3.06}\textdegree & \cellcolor[HTML]{\bestcolor} \textbf{1.74}\textdegree & \cellcolor[HTML]{\bestcolor} \textbf{2.54}\textdegree &  \cellcolor[HTML]{\bestcolor}\textbf{3.76}\textdegree &  \cellcolor[HTML]{\bestcolor}\textbf{3.58} &  \cellcolor[HTML]{\bestcolor} \textbf{2.07} &  \cellcolor[HTML]{\bestcolor} \textbf{3.09} &  \cellcolor[HTML]{\bestcolor} \textbf{4.55} \\ \hdashline

Ours & 82.55 & \cellcolor[HTML]{\bestcolor} \textbf{13.19} & 42.77 & 102.09 & 3.12\textdegree & 1.88\textdegree & 2.70\textdegree & 3.84\textdegree & 3.77 & 2.16 & 3.30 & 4.86 \\ \hline

\multicolumn{13}{|c|}{\cellcolor[HTML]{D4EBF2}\textbf{Rendered WB dataset: Set 2 (2,881 images)}} \\ \hline

FC4 \cite{hu2017fc}  &  505.30 & 142.46 & 307.77 & 635.35 & 10.37\textdegree
& 5.31\textdegree & 9.26\textdegree & 14.15\textdegree & 10.82 & 7.39 & 10.64 &  13.77   \\ \hline

Quasi-U CC \cite{bianco2019quasi} & 553.54 & 146.85 & 332.42 & 717.61 & 10.47\textdegree & 5.94\textdegree & 9.42\textdegree & 14.04\textdegree & 10.66 & 7.03 & 10.52 & 13.94 \\ \hline

KNN WB (Chapter \ref{ch:ch7}) & 171.09 & 37.04 & 87.04 & 190.88 & 4.48\textdegree &  2.26\textdegree & 3.64\textdegree & 5.95\textdegree & 5.60 & 3.43 &  4.90 & 7.06  \\ \hdashline

Ours & \cellcolor[HTML]{\bestcolor} \textbf{124.97} &
\cellcolor[HTML]{\bestcolor} \textbf{30.13} &
\cellcolor[HTML]{\bestcolor} \textbf{76.32} &
\cellcolor[HTML]{\bestcolor} \textbf{154.44} &
\cellcolor[HTML]{\bestcolor} \textbf{3.75}\textdegree &
\cellcolor[HTML]{\bestcolor} \textbf{2.02}\textdegree &
\cellcolor[HTML]{\bestcolor} \textbf{3.08}\textdegree &
\cellcolor[HTML]{\bestcolor} \textbf{4.72}\textdegree &
\cellcolor[HTML]{\bestcolor} \textbf{4.90} &
\cellcolor[HTML]{\bestcolor} \textbf{3.13} &
\cellcolor[HTML]{\bestcolor} \textbf{4.35}  &
\cellcolor[HTML]{\bestcolor} \textbf{6.08} \\ \hline

\multicolumn{13}{|c|}{\cellcolor[HTML]{D4EBF2}\textbf{Rendered Cube+ dataset with different WB settings (10,242 images)}} \\ \hline

FC4 \cite{hu2017fc}  & 371.9 & 79.15 & 213.41 & 467.33 & 6.49\textdegree & 3.34\textdegree & 5.59\textdegree & 8.59\textdegree & 10.38 & 6.6 & 9.76 & 13.26 \\ \hline

Quasi-U CC \cite{bianco2019quasi} & 292.18& 15.57 & 55.41 & 261.58 & 6.12\textdegree & 1.95\textdegree & 3.88\textdegree & 8.83\textdegree & 7.25 & 2.89 & 5.21 & 10.37 \\ \hline

KNN WB (Chapter \ref{ch:ch7}) &  194.98 &  27.43 &  57.08 &  118.21 &  4.12\textdegree &  1.96\textdegree & 3.17\textdegree & 5.04\textdegree &  5.68 &  3.22 &  4.61 &  6.70
\\ \hdashline

Ours & \cellcolor[HTML]{\bestcolor} \textbf{80.46} &
\cellcolor[HTML]{\bestcolor} \textbf{15.43} &
\cellcolor[HTML]{\bestcolor} \textbf{33.88} &
\cellcolor[HTML]{\bestcolor} \textbf{74.42} &
\cellcolor[HTML]{\bestcolor} \textbf{3.45}\textdegree &
\cellcolor[HTML]{\bestcolor} \textbf{1.87}\textdegree &
\cellcolor[HTML]{\bestcolor} \textbf{2.82}\textdegree &
\cellcolor[HTML]{\bestcolor} \textbf{4.26}\textdegree &
\cellcolor[HTML]{\bestcolor} \textbf{4.59} &
\cellcolor[HTML]{\bestcolor} \textbf{2.68} &
\cellcolor[HTML]{\bestcolor} \textbf{3.81}  &
\cellcolor[HTML]{\bestcolor} \textbf{5.53} \\ \hline

\end{tabular}
}
\end{table*}

\section{Results}\label{deepWB:sec:results}

Our method targets two different tasks: post-capture WB correction and manipulation of the sRGB rendered images to a specific WB color temperature. We achieve the state-of-the-art results for several different datasets for both tasks. We first describe the datasets used to evaluate our method in Sec.~\ref{deepWB:subsec:datasets}. We then discuss our quantitative and qualitative results in Sec.~\ref{deepWB:subsec:quantitative} and Sec. \ref{deepWB:subsec:qualitative}, respectively. We also perform an ablation study to validate our problem formulation and the proposed framework.

\begin{figure}[!t]
\includegraphics[width=\linewidth]{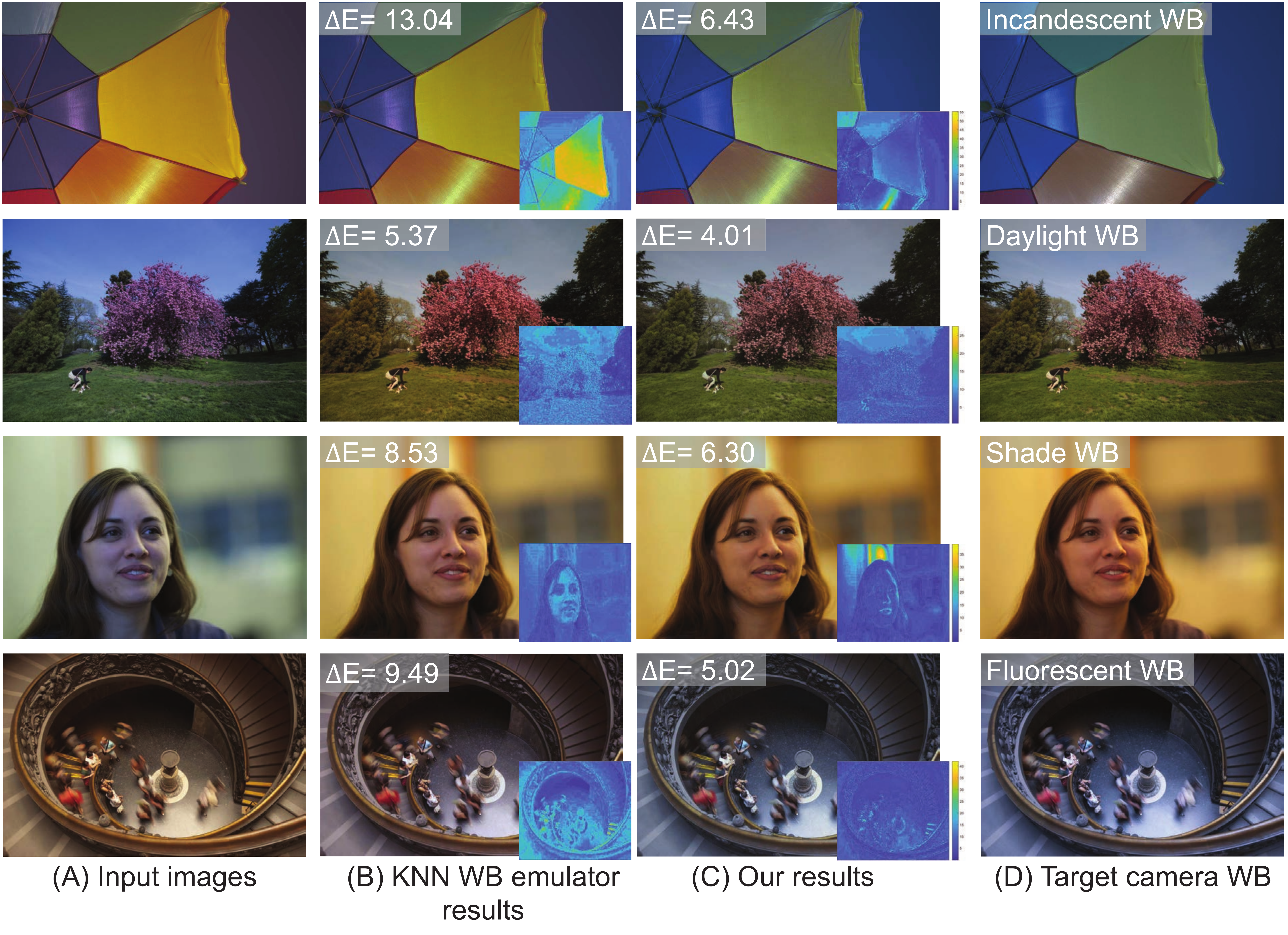}
\vspace{-7mm}
\caption[Qualitative comparison of WB manipulation.]{Qualitative comparison of WB manipulation. (A) Input images. (B) Results of KNN WB emulator (Chapter \ref{ch:ch8}). (C) Our results. (D) Ground truth camera-rendered images with the target WB settings. In this figure, the target WB settings are Incandescent, Daylight, Shade, and Fluorescent. Shown input images are taken from the rendered version of the MIT-Adobe FiveK dataset.}
\label{deepWB:fig:qualitative_WBEditing}
\end{figure}

\subsection{Datasets}\label{deepWB:subsec:datasets}
As previously mentioned, we used Set 1 of our Rendered WB dataset (Chapter\ \ref{ch:ch7}) for training and validation. For testing, we used three datasets not part of training or validation. Two of these additional datasets are as follows: (1) Set 2 of the Rendered WB dataset (2,881 images), and (2) the sRGB rendered version of the Cube+ dataset (10,242 images) \cite{banic2017unsupervised} (see Chapter\ \ref{ch:ch9} for more details). Datasets (1) and (2) are used to evaluate the task of AWB correction. For the WB manipulation task, we used the rendered Cube+ dataset and (3) the rendered version of the MIT-Adobe FiveK dataset (29,980 images) \cite{bychkovsky2011learning}. The rendered version of the MIT-Adobe FiveK dataset was generated similarly to the rendering process used to render the Rendered WB dataset (Chapter\ \ref{ch:ch7}). Specifically, each raw-RGB image was rendered to the sRGB color space with different WB settings. This rendering process was performed in an accurate emulation of advanced camera rendering procedures.

\begin{figure}[t]
\includegraphics[width=\linewidth]{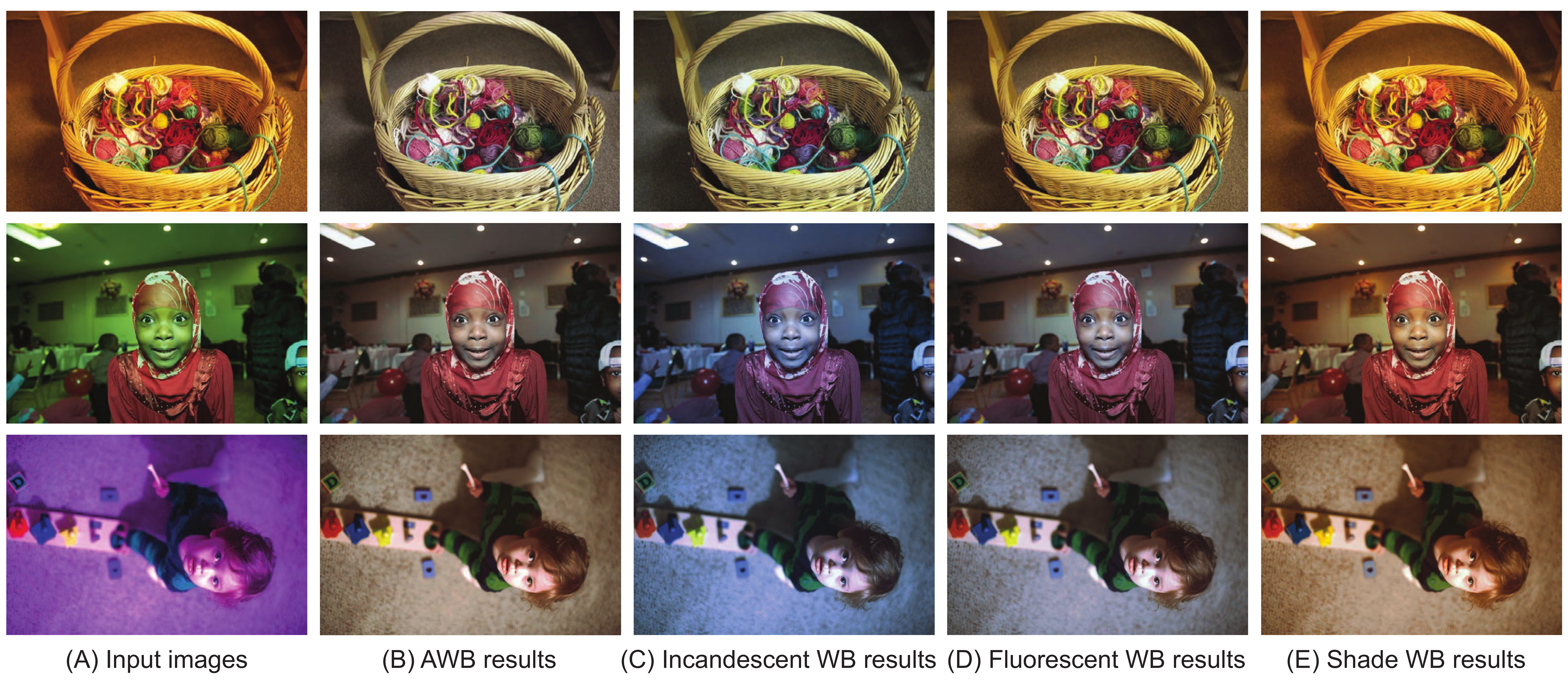}
\vspace{-7mm}
\caption[Qualitative results of our method.]{Qualitative results of our method. (A) Input images. (B) AWB results. (C) Incandescent WB results. (D) Fluorescent WB results. (E) Shade WB Results. Shown input images are rendered from the MIT-Adobe FiveK dataset \cite{bychkovsky2011learning}. }
\label{deepWB:fig:qualitative}
\end{figure}

\begin{figure}[t]
\includegraphics[width=\linewidth]{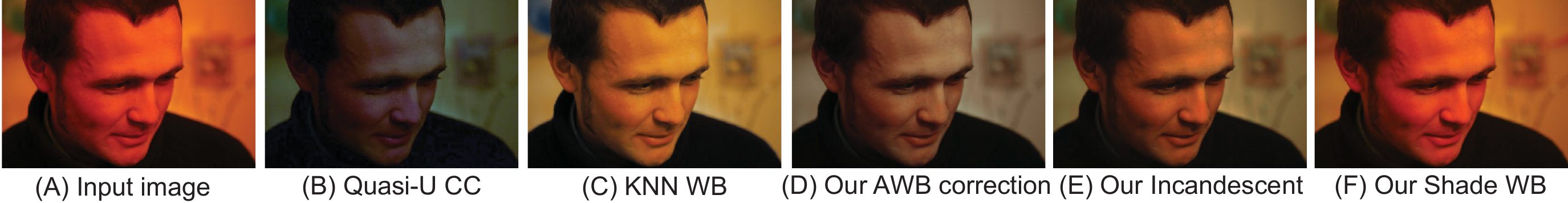}
\vspace{-7mm}
\caption[(A) Input image. (B) Result of quasi-U CC \cite{bianco2019quasi}. (C) Result of KNN WB (Chapter \ref{ch:ch7}). (D)-(F) Our deep-WB editing results.]{(A) Input image. (B) Result of quasi-U CC \cite{bianco2019quasi}. (C) Result of KNN WB (Chapter \ref{ch:ch7}). (D)-(F) Our deep-WB editing results. Photo credit: \textit{Duncan Yoyos} Flickr--CC BY-NC 2.0.}
\label{deepWB:fig:hard_case}
\end{figure}

\begin{figure}[t]
\includegraphics[width=\linewidth]{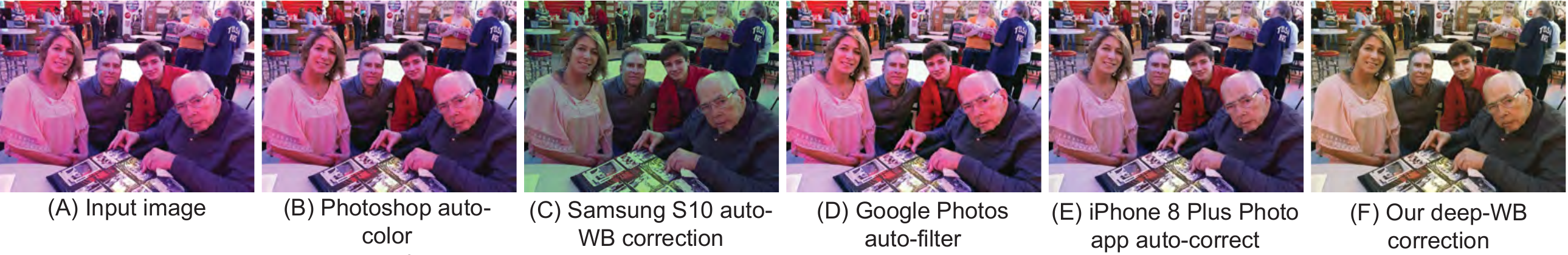}
\vspace{-7mm}
\caption[Strong color casts due to WB errors are hard to correct.]{Strong color casts due to WB errors are hard to correct. (A) Input image rendered with an incorrect WB setting. (B) Result of Photoshop auto-color correction. (C) Result of Samsung S10 auto-WB correction. (D) Result of Google Photos auto-filter. (E) Result of iPhone 8 Plus built-in Photo app auto-correction. (F) Our AWB result using the proposed deep-WB editing framework. Photo credit: \textit{OakleyOriginals} Flickr--CC BY 2.0.}
\label{deepWB:fig:commercial}
\end{figure}

\subsection{Quantitative Results}\label{deepWB:subsec:quantitative}

For both tasks, we follow the same evaluation metrics used Chapter \ref{ch:ch7}. Specifically, we used the following metrics to evaluate our results: MSE, MAE, and $\bigtriangleup$E 2000 \cite{sharma2005ciede2000}. For each evaluation metric, we report the mean, lower quartile (Q1), median (Q2), and the upper quartile (Q3) of the error.

\paragraph{WB Correction}  We compared the proposed method with the KNN WB approach (Chapter \ref{ch:ch7}). We also compared our results against the traditional WB diagonal-correction using recent illuminant estimation methods~\cite{bianco2019quasi, hu2017fc}. We note that methods \cite{bianco2019quasi, hu2017fc} were not designed to correct nonlinear sRGB images. These methods are included, because it is often purported that such methods are effective when the sRGB image has been ``linearized'' using a decoding gamma.


\begin{table}[t]
\caption[Results of WB manipulation using the rendered version of the Cube+ dataset \cite{banic2017unsupervised} and the rendered version of the MIT-Adobe FiveK dataset \cite{bychkovsky2011learning}.]{Results of WB manipulation using the rendered version of the Cube+ dataset \cite{banic2017unsupervised} and the rendered version of the MIT-Adobe FiveK dataset \cite{bychkovsky2011learning}. We report the mean, first, second (median), and third quartile (Q1, Q2, and Q3) of MSE, MAE, and $\boldsymbol{\bigtriangleup}$\textbf{E} 2000 \cite{sharma2005ciede2000}. The top results are indicated with yellow and boldface.}
\label{deepWB:Table1}
\centering
\scalebox{0.62}{
\begin{tabular}{|l|c|c|c|c|c|c|c|c|c|c|c|c|}
\hline
\multicolumn{1}{|c|}{} & \multicolumn{4}{c|}{\textbf{MSE}} & \multicolumn{4}{c|}{\textbf{MAE}} & \multicolumn{4}{c|}{\textbf{$\boldsymbol{\bigtriangleup}$\textbf{E} 2000}} \\ \cline{2-13}
\multicolumn{1}{|c|}{\multirow{-2}{*}{\textbf{Method}}} & \textbf{Mean} & \textbf{Q1} & \textbf{Q2} & \textbf{Q3} & \textbf{Mean} & \textbf{Q1} & \textbf{Q2} & \textbf{Q3} & \textbf{Mean} & \textbf{Q1} & \textbf{Q2} & \textbf{Q3}  \\ \hline

\multicolumn{13}{|c|}{\cellcolor[HTML]{D4EBF2}\textbf{Rendered Cube+ dataset (10,242 images)}} \\ \hline

KNN WB emulator (Chapter \ref{ch:ch8}) & 317.25 & 50.47 & 153.33 & 428.32 &  7.6\textdegree &  3.56\textdegree & 6.15\textdegree & 10.63\textdegree &  7.86 & 4.00 & 6.56 &  10.46
 \\ \hdashline
 					
Ours & \cellcolor[HTML]{\bestcolor} \textbf{199.38} &
\cellcolor[HTML]{\bestcolor} \textbf{32.30} &
\cellcolor[HTML]{\bestcolor} \textbf{63.34} &
\cellcolor[HTML]{\bestcolor} \textbf{142.76} &
\cellcolor[HTML]{\bestcolor} \textbf{5.40}\textdegree &
\cellcolor[HTML]{\bestcolor} \textbf{2.67}\textdegree &
\cellcolor[HTML]{\bestcolor} \textbf{4.04}\textdegree &
\cellcolor[HTML]{\bestcolor} \textbf{6.36}\textdegree &
\cellcolor[HTML]{\bestcolor} \textbf{5.98} &
\cellcolor[HTML]{\bestcolor} \textbf{3.44} &
\cellcolor[HTML]{\bestcolor} \textbf{4.78}  &
\cellcolor[HTML]{\bestcolor} \textbf{7.29} \\ \hline

\multicolumn{13}{|c|}{\cellcolor[HTML]{D4EBF2}\textbf{Rendered MIT-Adobe FiveK dataset (29,980 images)}} \\ \hline

KNN WB emulator (Chapter \ref{ch:ch8}) & 249.95 & 41.79 & 109.69 & 283.42 & 7.46\textdegree &  3.71\textdegree & 6.09\textdegree & 9.92\textdegree &  6.83 & 3.80 & 5.76 & 8.89 \\ \hdashline

Ours & \cellcolor[HTML]{\bestcolor} \textbf{135.71} &
\cellcolor[HTML]{\bestcolor} \textbf{31.21} &
\cellcolor[HTML]{\bestcolor} \textbf{68.63} &
\cellcolor[HTML]{\bestcolor} \textbf{151.49} &
\cellcolor[HTML]{\bestcolor} \textbf{5.41}\textdegree &
\cellcolor[HTML]{\bestcolor} \textbf{2.96}\textdegree &
\cellcolor[HTML]{\bestcolor} \textbf{4.45}\textdegree &
\cellcolor[HTML]{\bestcolor} \textbf{6.83}\textdegree &
\cellcolor[HTML]{\bestcolor} \textbf{5.24} &
\cellcolor[HTML]{\bestcolor} \textbf{3.32} &
\cellcolor[HTML]{\bestcolor} \textbf{4.57}  &
\cellcolor[HTML]{\bestcolor} \textbf{6.41} \\ \hline

\end{tabular}
}
\end{table}

Table \ref{deepWB:Table0} reports the error between corrected images obtained by each method and the corresponding ground truth images. In Table \ref{deepWB:Table0}, we show results on the validation set (i.e., Set 1 in the Rendered WB dataset) and three testing sets, for a total of 13,123 unseen sRGB images rendered with different camera models and WB settings. For the diagonal-correction results, we pre-processed each testing image by first applying the 2.2 gamma linearization \cite{anderson1996proposal, ebner2007color}, and then we applied the gamma encoding after correction. As we can see, our method generalizes well for the testing sets, achieving state-of-the-art results compared to the other approaches in all evaluation metrics. We have on par results with the state-of-the-art method (Chapter \ref{ch:ch7}) on the validation set.

\paragraph{WB Manipulation} The goal of this task is to change the input image's colors to appear as they were rendered using a target WB setting. We compare our result with the work in Chapter \ref{ch:ch8} that proposed a KNN WB emulator that mimics WB effects in the sRGB space. We used the same WB settings produced by the KNN WB emulator. Specifically, we selected the following target WB settings: Incandescent ($2850K$), Fluorescent ($3800K$), Daylight ($5500K$), Cloudy ($6500K$), and Shade ($7500K$). As our decoders were trained to generate only Incandescent and Shade WB settings, we used Eq.\ \ref{deepWB:eq6} to produce the other WB settings (i.e., Fluorescent, Daylight, and Cloudy WB settings).

Table \ref{deepWB:Table1} shows the obtained results using our method and the KNN WB emulator. Table \ref{deepWB:Table1} demonstrates that our method outperforms the KNN WB emulator (Chapter \ref{ch:ch8}) over a total of 40,222 testing images captured with different camera models and WB settings using all evaluation metrics.

\subsection{Qualitative Results}\label{deepWB:subsec:qualitative}

In Fig. \ref{deepWB:fig:qualitative_AWB} and Fig. \ref{deepWB:fig:qualitative_WBEditing}, we provide a visual comparison of our results against the most recent work proposed for WB correction \cite{bianco2019quasi} and our work in Chapters \ref{ch:ch7} and \ref{ch:ch8}, respectively. On top of each example, we show the $\bigtriangleup$E 2000 error between the result image and the corresponding ground truth image (i.e., rendered by the camera using the target setting). It is clear that our results have the lower $\bigtriangleup$E 2000 and are the most similar to the ground truth images.

Figure \ref{deepWB:fig:qualitative} shows additional examples of our results. As shown, our framework accepts input images with arbitrary WB settings and re-renders them with the target WB settings, including the AWB correction.

We tested our method with several images taken from the Internet to check its ability to generalize to images typically found online. Figure \ref{deepWB:fig:hard_case} and Fig. \ref{deepWB:fig:commercial} show examples. As it is shown, our method produces compelling results compared with other methods and commercial software packages for photo editing, even when input images have strong color casts.

\subsection{Comparison With a Vanilla U-Net}

As explained earlier, our framework employs a single encoder to encode input images, while each decoder is responsible for producing a specific WB setting. Our architecture aims to model Eq. \ref{deepWB:eq1} in the same way cameras would produce colors for different WB settings from the same raw-RGB captured image.

Intuitively, we can re-implement our framework using a multi-U-Net architecture \cite{ronneberger2015u}, such that each encoder/decoder model will be trained for a single target of the WB settings.

In Table \ref{deepWB:Table2}, we provide a comparison between our proposed framework against vanilla U-Net models. We train our proposed architecture and three U-Net models (each U-Net model targets one of our WB settings) for 88,000 iterations. The results validate our design and make evident that our shared encoder not only reduces the required number of parameters but also gives better results.

\begin{table}[t]
\caption[Average of mean square error and $\boldsymbol{\bigtriangleup}$\textbf{E} 2000 \cite{sharma2005ciede2000} obtained by our framework and the traditional U-Net architecture \cite{ronneberger2015u}.]{Average of mean square error and $\boldsymbol{\bigtriangleup}$\textbf{E} 2000 \cite{sharma2005ciede2000} obtained by our framework and the traditional U-Net architecture \cite{ronneberger2015u}. Shown results on Set 2 of our Rendered WB dataset (Chapter \ref{ch:ch7}) for AWB and the rendered version of the Cube+ dataset \cite{banic2017unsupervised} for WB manipulation (see Chapter \ref{ch:ch9} for more details regarding this rendered version of the Cube+ dataset). The top results are indicated with yellow and boldface.}
\label{deepWB:Table2}
\centering
\scalebox{0.7}{
\begin{tabular}{|l|c|c|c|c|}
\hline
& \multicolumn{2}{|c|}{\cellcolor[HTML]{D4EBF2}\textbf{AWB}} & \multicolumn{2}{|c|}{\cellcolor[HTML]{D4EBF2}\textbf{WB editing}} \\ \cline{2-5}
\multicolumn{1}{|c|}{\multirow{-2}{*}{\textbf{Method}}} & \textbf{MSE} & \textbf{$\boldsymbol{\bigtriangleup}$\textbf{E} 2000} & \textbf{MSE} & \textbf{$\boldsymbol{\bigtriangleup}$\textbf{E} 2000} \\\hline

Multi-U-Net \cite{ronneberger2015u}  & 187.25
 & 6.23 & 234.77 &  6.87  \\\hdashline

Ours & \cellcolor[HTML]{\bestcolor}\textbf{124.47} &   \cellcolor[HTML]{\bestcolor}\textbf{4.99} & \cellcolor[HTML]{\bestcolor}\textbf{206.81} &  \cellcolor[HTML]{\bestcolor}\textbf{6.23} \\ \hline

\end{tabular}
}
\end{table}

\section{Summary}\label{deepWB:sec:conclusion}

We have presented a deep learning framework for editing the WB of sRGB camera-rendered images.
Specifically, we have proposed a DNN architecture that uses a single encoder and multiple decoders, which are trained in an end-to-end manner. Our framework allows the direct correction of images captured with wrong WB settings. Additionally, our framework produces output images that allow users to manually adjust the sRGB image to appear as if it was rendered with a wide range of WB color temperatures.    Quantitative and qualitative results demonstrate the effectiveness of our framework against recent data-driven methods.

\chapter{Color Temperature Tuning\label{ch:ch11}}
In this chapter, we propose an imaging framework\footnote{This work was published in \cite{afifi2019colortemp, afifi2020systems}: Mahmoud Afifi, Abhijith Punnappurath, Abdelrahman Abdelhamed, Hakki Can Karaimer, Abdullah Abuolaim, and Michael S. Brown. Color Temperature Tuning: Allowing Accurate Post-Capture White-Balance Editing. In Color and Imaging Conference, 2019.} that renders a small number of ``tiny versions'' of the original image (e.g., 0.1\% of the full-size image), each with different WB color temperatures.  Rendering these tiny images requires minimal overhead from the camera pipeline. These tiny images are sufficient to allow color mapping functions to be computed that can map the full-sized sRGB image to appear as if it was rendered with any of the tiny images' color temperatures. Moreover, by blending the color mapping functions, we can map the output sRGB image to appear as if it was rendered through the full pipeline with {\it any color temperature}. These mapping functions can be stored as a JPEG comment with less than 6 KB overhead. We demonstrate that this capture framework can significantly outperform any existing solution targeting post-capture WB editing. The source code of this work is available on GitHub: \href{https://github.com/mahmoudnafifi/ColorTempTuning}{https://github.com/mahmoudnafifi/ColorTempTuning}.

\section{Introduction} \label{CIC27:sec:intro}

As discussed in previous chapters, WB is applied to the raw-RGB sensor image and aims to remove the color cast due to the scene's illumination, which is often described by its correlated color temperature.   An image's WB temperature can either be specified by a manual setting (e.g., Tungsten, Daylight), or be estimated from the image using the camera's AWB function.

\begin{figure}[t!]
\includegraphics[width=\linewidth]{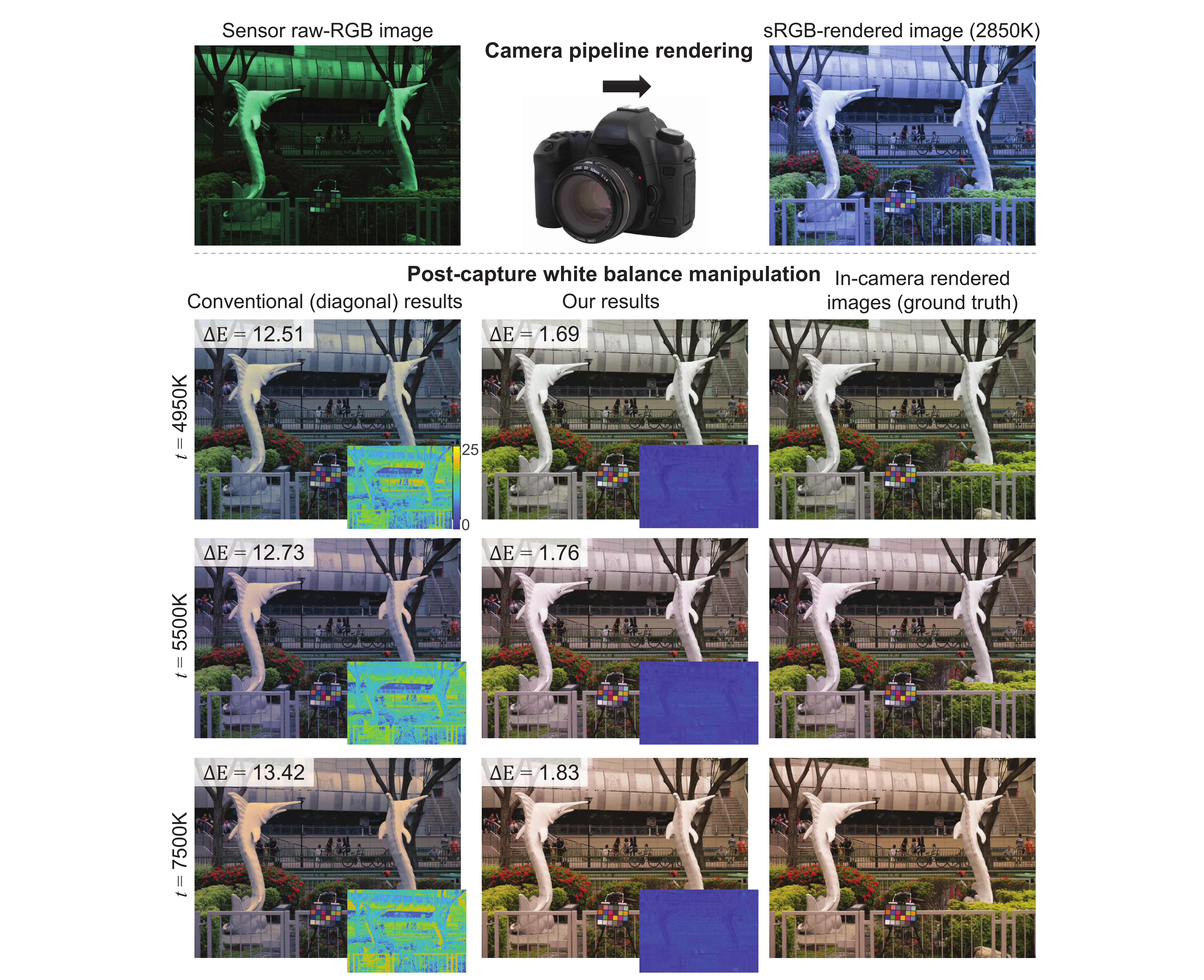}
\vspace{-7mm}
\caption[An sRGB image rendered using our imaging framework with the \emph{Tungsten} WB setting (i.e., 2850K).]{An sRGB image rendered using our imaging framework with the \emph{Tungsten} WB setting (i.e., 2850K). Using data embedded within the rendered image, our method can modify this sRGB image's WB post-capture to \emph{any} target WB color temperature (e.g., 4950K, 5500K, 7500K) producing results (column 2) almost identical to what the actual camera pipeline would have generated (column 3). Error maps ($\triangle$E) insets, and average ($\triangle$E) demonstrate that our method is far superior to the conventional post-capture WB manipulation (column 1).}
\label{CIC27:fig:teaser}
\end{figure}

After the WB step, the camera pipeline applies several nonlinear camera-specific photo-finishing operations to convert the image from the raw-RGB color space to sRGB. These nonlinear operations make it very challenging to modify the WB post-capture. This is particularly troublesome if the WB setting was incorrect, resulting in the captured image having an undesirable color cast. Existing methods for post-capture WB manipulation attempt to reverse the camera pipeline and map the sRGB colors back to raw-RGB. Not only does this process necessitate careful camera calibration, but also it requires reapplying the pipeline to get back to the sRGB space after modifying the WB in the raw-RGB space.

In our work in Chapter\ \ref{ch:ch7} and Chapter\ \ref{ch:ch9}, we proposes to white balance improperly white-balanced sRGB-rendered images by estimating polynomial mapping functions from a large set of training data. Our work in this chapter is close to the work discussed in Chapters \ref{ch:ch7} and \ref{ch:ch9} in the sense that we also use a set of polynomial mapping functions to manipulate the WB of sRGB-rendered images. In contrast, our work here embeds the required mapping functions within the final rendered images during the rendering process.

\paragraph{Contributions} We advocate an image capture framework to enable accurate post-capture WB manipulation \emph{directly} in the sRGB space without having to revert to the raw-RGB space. Our proposed approach is to create a tiny downsampled version of the raw-RGB image and render it through the camera pipeline multiple times using a set of pre-selected WB settings. We can then use these tiny sRGB images to compute nonlinear color mapping functions that can transform the full-sized sRGB output image to appear as if it was rendered through the pipeline with the color temperature of any of the tiny images. More importantly, blending these mapping functions provides the ability to navigate the full parameter space of WB settings---that is, we can produce a full-sized sRGB output image with \emph{any} color temperature that is almost identical to what the camera pipeline would have produced. Figure \ref{CIC27:fig:teaser} shows an example. The mapping functions themselves can be efficiently computed and easily stored in the sRGB-rendered image (e.g., as a JPEG comment with less than 6 KB overhead).

\begin{figure*}[!t]
\includegraphics[width=\linewidth]{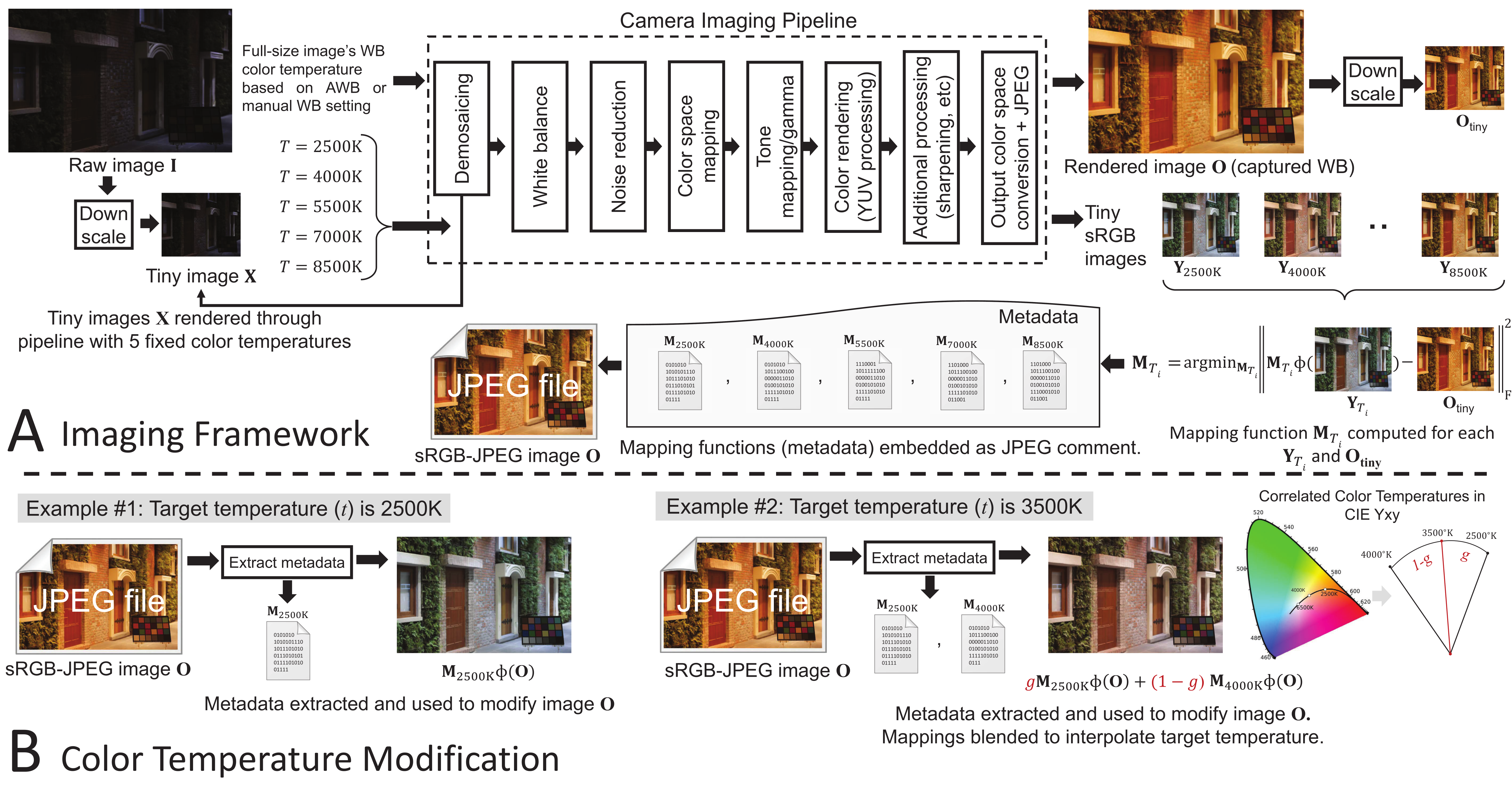}
\vspace{-7mm}
\caption[Our proposed image capture framework.]{(A) Our proposed image capture framework.  The raw-RGB image $\mat{I}$ is downsampled to produce the tiny raw-RGB image $\mat{X}$. Image $\mat{X}$ is then rendered through the pipeline $N$ times, each time with a different pre-selected WB preset $T_i$ applied, to produce the tiny sRGB-rendered images $\{\mat{Y}_{T_i} \}_{i=1}^N$. Mapping functions $\{\mat{M}_{T_i}\}_{i=1}^N$ are computed from these tiny images $\{\mat{Y}_{T_i}\}_{i=1}^N$ and a downsampled version of the sRGB output image $\mat{O}$ denoted as $\mat{O}_{\textrm{tiny}}$. The mappings $\{\mat{M}_{T_i}\}_{i=1}^N$ are stored inside the JPEG comment field of the sRGB output image $\mat{O}$. (B) Using the metadata to modify an sRGB image.  Example 1 shows a case where the sRGB image $\mat{O}$ is mapped to one of the color temperatures in the metadata. The corresponding color mapping function can be extracted and used to modify the image.  Example 2 shows an example where the target color temperature is not in the metadata. In this case, the target temperature mapping function is interpolated by blending between the two closest mapping functions in the metadata.
\label{CIC27:fig:ProposedpipeLine}}
\end{figure*}

\section{Methodology} \label{CIC27:sec:method}

Our proposed method is based on the computation of a set of nonlinear color mapping functions between a set of tiny downsampled camera-rendered sRGB images. We first discuss, in Sec.\ \ref{CIC27:subsec:tiny_images},
how these tiny images are rendered. Sec.\ \ref{CIC27:subsec:MF} describes how to compute the mapping functions. Finally, in Sec.\ \ref{CIC27:subsec:PP}, we elaborate on how these mappings enable us to explore the parameter space of WB settings, and allow for accurate post-capture WB manipulation. An overview of the image framework is shown in Fig. \ref{CIC27:fig:ProposedpipeLine}-(A).

In this chapter, we represent each image as a $3\!\times\!P$ matrix that contains the image's RGB triplets, where $P$ is the total number of pixels in the image.

\subsection{Rendering Tiny Images} \label{CIC27:subsec:tiny_images}

The first step of our imaging framework is to create a tiny downsampled copy of the raw-RGB image $\mat{I}$. The tiny version is denoted as $\mat{X}$ and, in our experiments, is only $150\!\times\!150$ pixels, as compared to, say, a 20-megapixel full-sized image $\mat{I}$. Our tiny raw-RGB image $\mat{X}$, which is approximately 0.1\% of the full-size image, can be stored in memory easily. The full-sized raw-RGB image $\mat{I}$ is first rendered through the camera pipeline with some WB color temperature to produce the full-sized sRGB output image $\mat{O}$. This color temperature is either obtained from a manually selected WB setting or estimated by the camera's AWB. The tiny image $\mat{X}$ is then processed by the camera pipeline $N$ times, each time with a different WB color temperature setting $\{T_i \}_{i=1}^N$. These settings $\{T_i \}_{i=1}^N$ can correspond to the camera's preset WB options, such as Tungsten, Daylight, Fluorescent, Cloudy, and Shade, or their color temperature values, such as 2850K, 3800K, 5500K, 6500K, and 7500K, or any other chosen set of color temperatures. The resulting tiny sRGB images processed by the camera pipeline are denoted as $\{\mat{Y}_{T_i}\}_{i=1}^N$ corresponding to the WB settings $\{T_i \}_{i=1}^N$.

\subsection{Mapping Functions} \label{CIC27:subsec:MF}

\begin{figure}[!t]
\includegraphics[width=\linewidth]{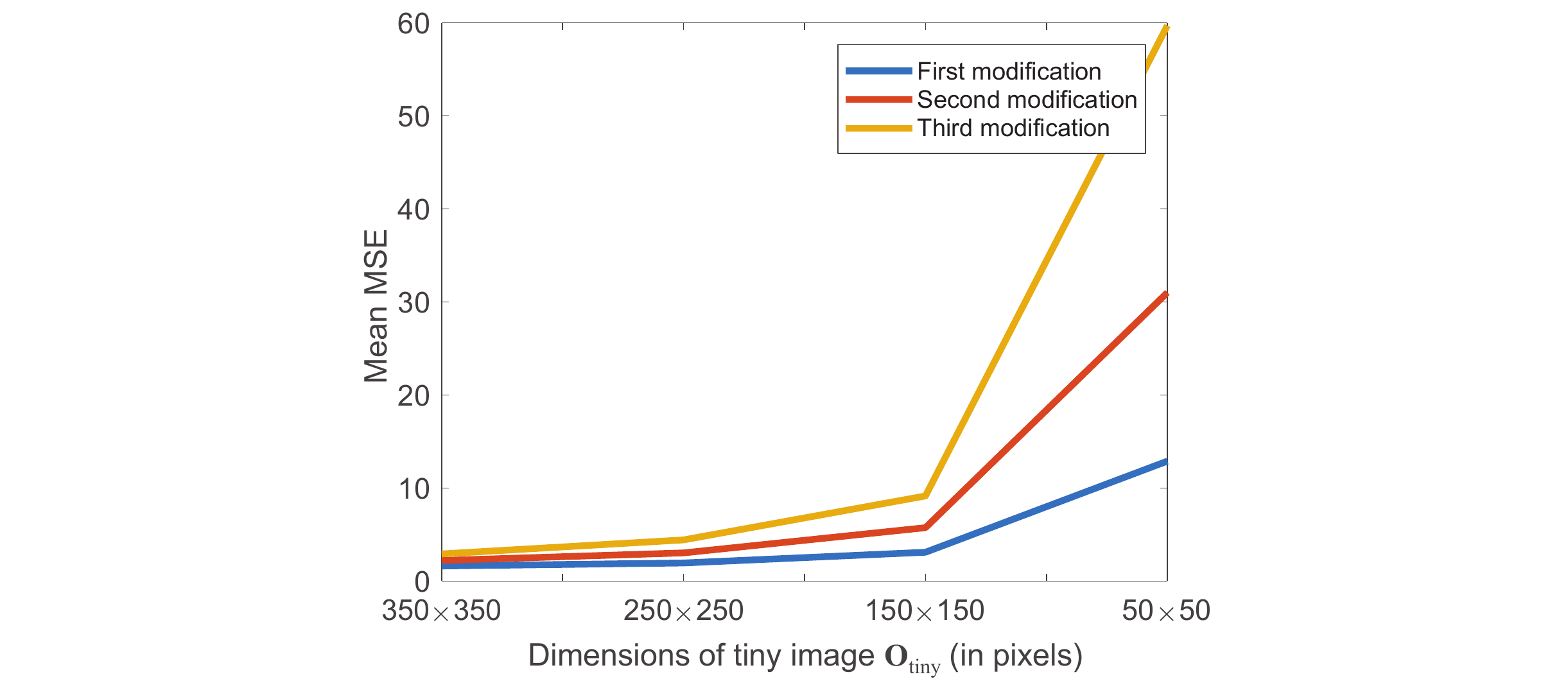}
\vspace{-7mm}
\caption[The effect of different downsampling sizes on our results.]{The effect of different downsampling sizes on our results. In this figure, we also show the mean squared error (MSE) between our results and the target in-camera ground truth images after updating our metadata for three successive WB modifications with different target color temperatures. The details of the update process are discussed in Sec.\ \ref{CIC27:subsec:PP}.}
\label{CIC27:fig:quantative_results}
\end{figure}


The full-sized sRGB output image $\mat{O}$ is downsampled to have the same dimensions as $\{\mat{Y}_{T_i} \}_{i=1}^N$ and is denoted as $\mat{O}_{\textrm{tiny}}$. For each tiny image $\{\mat{Y}_{T_i} \}_{i=1}^N$, we compute a nonlinear mapping function $\mat{M}_{T_i}$, which maps $\mat{O}_{\textrm{tiny}}$ to $\mat{Y}_{T_i}$, by solving the following minimization problem:
\begin{equation}
\label{CIC27:eq_M}
\underset{\mat{M}_{T_i}}{\argmin} \parallel\mat{M}_{T_i}\textrm{ }\Phi\left(\mat{O}_{\textrm{tiny}}\right)    - \mat{Y}_{T_i} \parallel_{\textrm{F}}^2,
\end{equation}
\noindent
where $\Phi: \mathbb{R}^3 \rightarrow \mathbb{R}^u$ is a kernel function that transforms RGB triplets to a $u$-dimensional space, where $u>3$, and $\parallel.\parallel_{\textrm{F}}$ denotes the Frobenius norm. For each image $\mat{Y}_{T_i}$, this equation finds an $\mat{M}_{T_i}$ that minimizes the errors between the RGB colors in the downsampled image $\mat{O}_{\textrm{tiny}}$ and its corresponding image $\mat{Y}_{T_i}$.

In general, relying on kernel functions based on high-degree polynomials can hinder generalization; however, in our case, the mapping function is computed specifically for a pair of images. Hence, a kernel function with a higher degree is preferable. In our experiments, we adopted a polynomial kernel function given in \cite{finlayson2015color}, where $\Phi: \mathbb{R}^3 \rightarrow \mathbb{R}^{34}$. Hence, each mapping function is represented by a $3\times 34$ matrix. Once the mapping functions are computed, the set of downsampled images $\mat{X}$, $\{\mat{Y}_{T_i} \}_{i=1}^N$, and $\mat{O}_{\textrm{tiny}}$ is no longer needed and can be discarded.

For all of our experiments in this chapter, we rendered tiny images using five (5) color temperature values, 2500K, 4000K, 5500K, 7000K, and 8500K, and computed the corresponding mapping functions $\mat{M}_{T_i}$.  The five functions require less than 6 KB of metadata to represent and can be saved inside the final JPEG image $\mat{O}$ as a comment field.

\subsection{Color Temperature Manipulation} \label{CIC27:subsec:PP}

Once the mapping functions have been computed, we can use them to post-process the sRGB output image $\mat{O}$ to appear as if it was rendered through the camera pipeline with any of the WB settings $\{T_i\}_{i=1}^N$. This process can be described using the following equation:
\begin{equation}
\label{CIC27:eq_PP}
\mat{O}_{\textrm{modified}} = \mat{M}_{T_i}\textrm{ }\Phi\left(\mat{O}\right),
\end{equation}
where $\mat{O}_{\textrm{modified}}$ is the full-resolution sRGB image as if it was ``re-rendered'' with the WB setting $T_i$. This is demonstrated in Example 1 of Fig. \ref{CIC27:fig:ProposedpipeLine}-(B).

More importantly, by blending between the mapping functions, we can post-process the sRGB output image $\mat{O}$ to appear as if it was processed by the camera pipeline using \emph{any} color temperature value, and not just the settings $\{T_i \}_{i=1}^N$ the tiny images $\{\mat{Y}_{T_i}\}_{i=1}^N$ were rendered out with.

Given a new target WB setting with a color temperature $t$, we can interpolate between the nearest pre-computed mapping functions to generate a new mapping function for $t$ as follows:
\begin{equation}
\label{CIC27:eq_WB1}
\mat{M}_{t} = g \mat{M}_{a}\textrm{ } + (1-g) \mat{M}_{b},
\end{equation}
\begin{equation}
\label{CIC27:eq_WB2}
g = \frac{1/t - 1/b}{1/a - 1/b},
\end{equation}
where $a$, $b$ $\in \{T_i \}_{i=1}^N$ are the nearest pre-computed color temperatures to $t$, such that $a<t<b$, and $\mat{M}_{a}$ and $\mat{M}_{b}$ are the corresponding mapping functions computed for temperatures $a$ and $b$, respectively.
The final modified image $\mat{O}_{\textrm{modified}}$ is generated by using $\mat{M}_{t}$ instead of $\mat{M}_{T_i}$ in Eq. \ref{CIC27:eq_PP}. Example 2 of Fig.~\ref{CIC27:fig:ProposedpipeLine}-(B) demonstrates this process.

\subsubsection{Recomputing the Mapping Functions}
Our metadata was computed for the original sRGB-rendered image $\mat{O}$. After $\mat{O}$ has been modified to $\mat{O}_{\textrm{modified}}$, this metadata has to be updated to facilitate future modifications of $\mat{O}_{\textrm{modified}}$. The update is performed so as to map $\mat{O}_{\textrm{modified}}$, with color temperature $t$, to our preset color temperatures $\{T_i \}_{i=1}^N$. To that end, each pre-computed mapping function $\{\mat{M}_{T_i}\}_{i=1}^N$ is updated based on the newly generated image $\mat{O}_{\textrm{modified}}$ as follows:
\begin{equation}\label{CIC27:eq_WBUpdate}
\underset{\mat{M}_{T_i}}{\argmin} \parallel\mat{M}_{T_i}\Phi\left(\mat{M}_{t}\Phi\left(\mat{O}_{\textrm{tiny}}\right)\right)    -  \mat{M}_{T_i\textrm{(old)}}\Phi\left(\mat{O}_{\textrm{tiny}}\right) \parallel_{\textrm{F}}^2,
\end{equation}
where $\mat{M}_{T_i}$ is the new mapping function and $\mat{M}_{T_i\textrm{(old)}}$ is the old mapping function for the $i^{\textrm{th}}$ WB setting before the current WB modification.

\begin{figure}[!t]
\includegraphics[width=\linewidth]{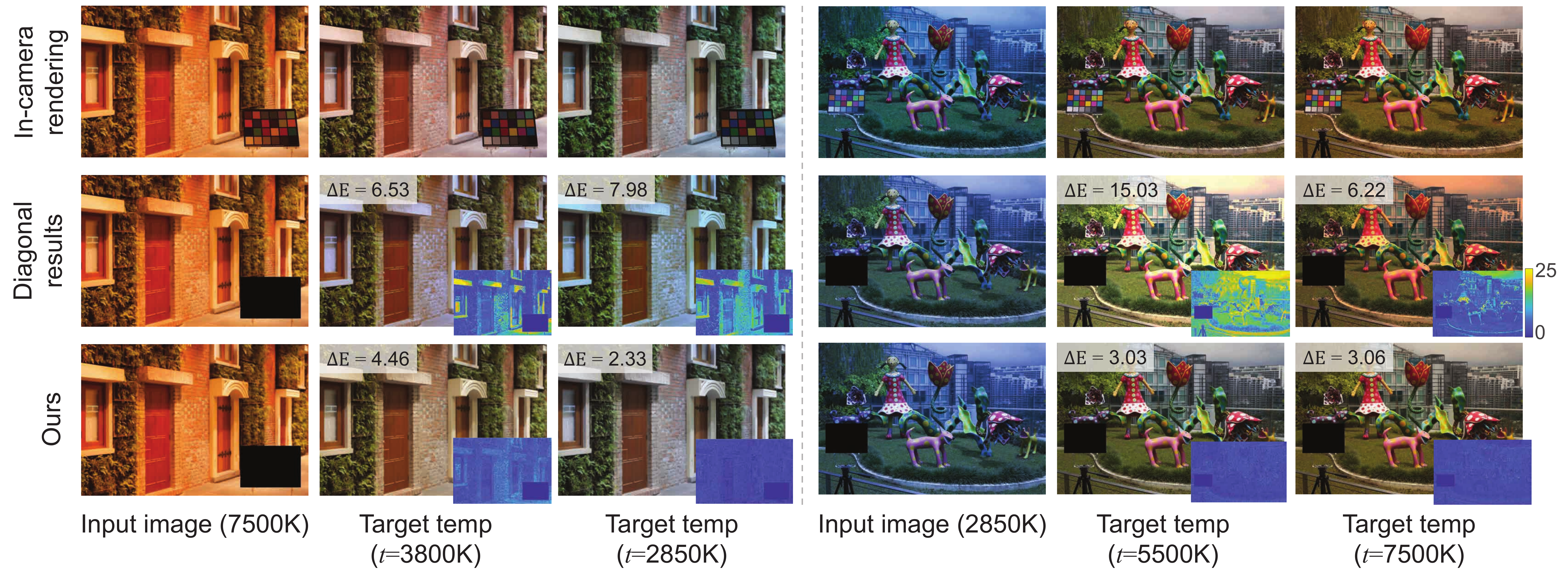}
\vspace{-7mm}
\caption[First row: in-camera sRGB images rendered with different color temperatures. Second row: results obtained by diagonal manipulation using the \textit{exact} achromatic patch from the color chart. Third row: our results.]{First row: in-camera sRGB images rendered with different color temperatures. Second row: results obtained by diagonal manipulation using the \textit{exact} achromatic patch from the color chart. Third row: our results. $\Delta$E error of each result is reported and shown as an error map.}
\label{CIC27:fig:results2}
\end{figure}

\begin{figure}[!t]
\includegraphics[width=\linewidth]{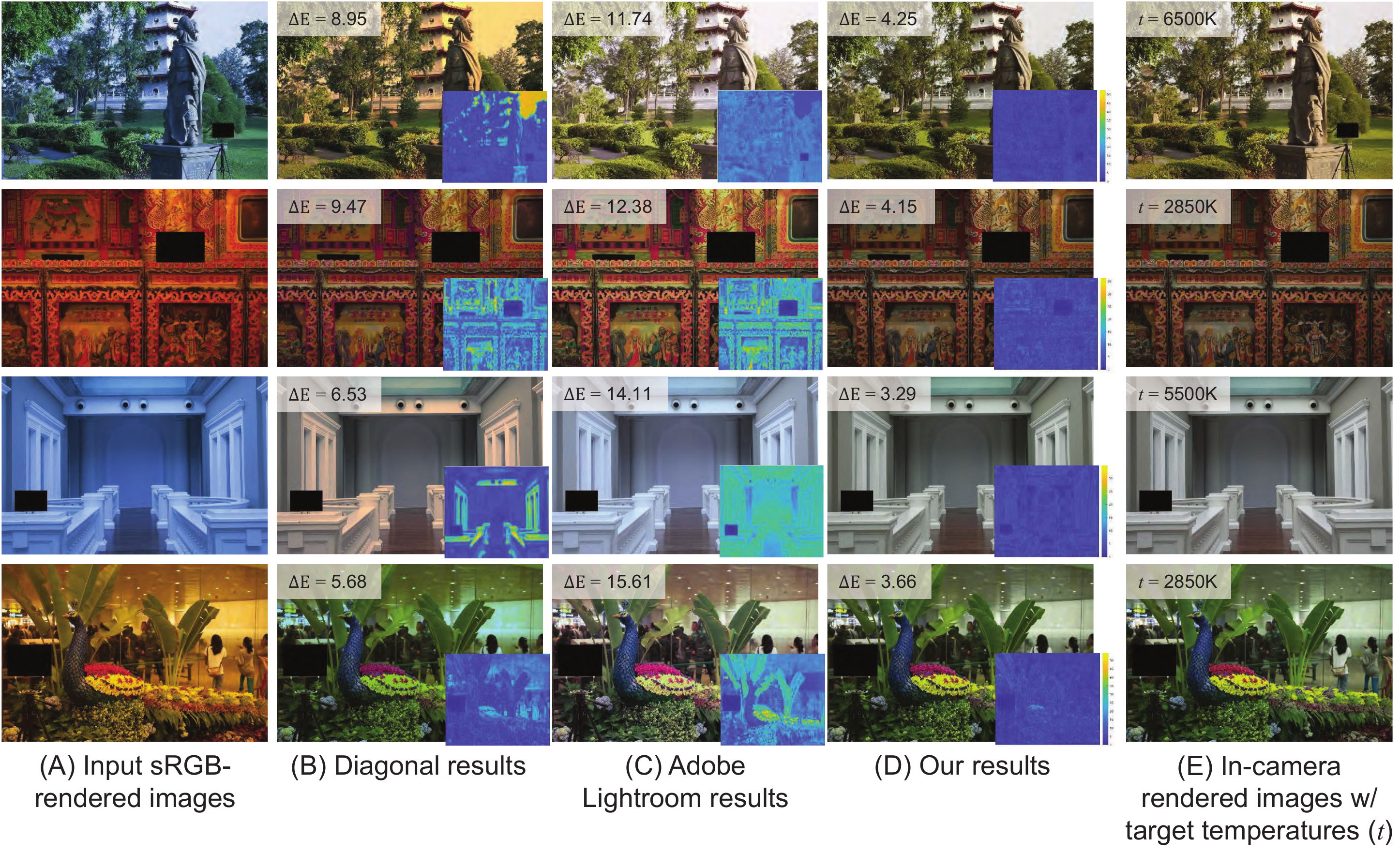}
\vspace{-7mm}
\caption[(A) Input sRGB-rendered image. (B) Diagonal manipulation result. (C) Adobe Lightroom result. (D) Our result. (E) In-camera sRGB image rendered with the target color temperature.]{(A) Input sRGB-rendered image. (B) Diagonal manipulation result. (C) Adobe Lightroom result. (D) Our result. (E) In-camera sRGB image rendered with the target color temperature $t$. The average $\Delta$E error is shown for each image.}
\label{CIC27:fig:results3}
\end{figure}

\section{Results} \label{CIC27:sec:results}

\begin{table}[t]
\caption[Quantitative results on the NUS dataset~\cite{cheng2014illuminant}.]{Quantitative results on the NUS dataset~\cite{cheng2014illuminant}. We compare our results against diagonal WB manipulation, denoted as Diag, using an \emph{exact achromatic} reference point obtained from the color chart in the scene. The diagonal manipulation is applied directly on the sRGB images, and on the ``linearized'' sRGB~\cite{anderson1996proposal,ebner2007color}. The terms Q1, Q2, and Q3 denote the first, second (median), and third quartile, respectively. The terms MSE and MAE stand for mean square error and mean angular error, respectively. The best results are indicated in boldface and highlighted in yellow.
\label{CIC27:tab:quantitative}}
\centering
\scalebox{0.56}{
\begin{tabular}{|c|c|c|c|c|c|c|c|c|c|c|c|c|c|c|c|c|}
\hline
\multirow{2}{*}{Method}                                                  & \multicolumn{4}{c|}{MSE} & \multicolumn{4}{c|}{MAE} & \multicolumn{4}{c|}{$\Delta$ E 2000} & \multicolumn{4}{c|}{$\Delta$ E 76}\\
\cline{2-17}
& Mean  & Q1  & Q2  & Q3  & Mean  & Q1  & Q2  & Q3  & Mean  & Q1  & Q2 & Q3 & Mean  & Q1  & Q2 & Q3 \\
\hline
Diag WB  & 1196.93 &  455.21 & 825.38 & 1441.58 & 5.75& 2.24 & 4.85 & 8.03 & 8.98 & 4.88 & 7.99 & 11.73 & 13.53 & 7.18 & 11.84 & 17.98   \\
\hline
\begin{tabular}[c]{@{}c@{}}Diag WB w linearization\end{tabular} &     1160.16  &   428.18  &   771.78  &   1400.49  &    5.49   &    2.19 &    4.56 &  7.61   &   8.61    &  4.62   &   7.58 &  11.09 &    12.87   & 6.82    &  11.22  & 16.99 \\
\hline
Ours  & \cellcolor[HTML]{\bestcolor}\textbf{75.58}& \cellcolor[HTML]{\bestcolor}\textbf{35.51}& \cellcolor[HTML]{\bestcolor}\textbf{61.05}&     \cellcolor[HTML]{\bestcolor}\textbf{98.68}& \cellcolor[HTML]{\bestcolor}\textbf{2.04} &   \cellcolor[HTML]{\bestcolor}\textbf{1.42} &   \cellcolor[HTML]{\bestcolor}\textbf{1.86} &   \cellcolor[HTML]{\bestcolor}\textbf{2.45} &  \cellcolor[HTML]{\bestcolor}\textbf{3.09} &   \cellcolor[HTML]{\bestcolor}\textbf{2.33} &  \cellcolor[HTML]{\bestcolor}\textbf{3.00} & \cellcolor[HTML]{\bestcolor}\textbf{3.74} &   \cellcolor[HTML]{\bestcolor}\textbf{4.39} &  \cellcolor[HTML]{\bestcolor}\textbf{3.21} &  \cellcolor[HTML]{\bestcolor}\textbf{4.26} & \cellcolor[HTML]{\bestcolor}\textbf{5.30}\\
\hline
\end{tabular}}
\end{table}

We evaluated our proposed method on six cameras from the NUS dataset~\cite{cheng2014illuminant}. Images in this dataset were captured using digital single-lens reflex (DSLR) cameras with a color chart placed in the scene. All images in the dataset have been saved in the raw-RGB format, and so we can convert them to sRGB format using the conventional in-camera pipeline \cite{karaimer2016software}. Specifically, we rendered out sRGB images with five different color temperature values, which are: 2850K, 3800K, 5500K, 6500K, and 7500K, corresponding approximately to the common WB presets available on most cameras---namely, Tungsten, Fluorescent, Daylight, Cloudy, and Shade. For our chosen six cameras totaling 1,340 images, this process yields $1340\!\times\!5=6700$ sRGB-rendered images. These images can be considered as ground truth since these are produced by emulating the in-camera pipeline. See Fig. \ref{CIC27:fig:teaser}, for example. We also render the same images using our proposed framework with the color chart masked out. During this rendering process, mapping functions corresponding to our pre-selected color temperature values (i.e., 2500K, 4000K, 5500K, 7000K, and 8500K) are computed and stored as metadata corresponding to each image.

For each image generated with a particular WB preset, we choose the other four WB settings as the target WB values. Following the procedure described in Sec.\ \ref{CIC27:subsec:PP}, we then process the input image using our pre-computed mapping functions to appear as though it was rendered through the camera pipeline with the four target WB settings. For example, given an input image originally rendered out with the WB Tungsten preset, we generate four WB modified versions with the following WB settings: Fluorescent, Daylight, Cloudy, and Shade. In this manner, we generate $6700\!\times\!4=26800$ WB modified images using our proposed approach. These modified images can be compared with their corresponding ground truth images.

We adopt four commonly used error metrics for quantitative evaluation: (i) MSE, (ii) MAE, (iii) $\Delta$E 2000 \cite{sharma2005ciede2000}, and (iv) $\Delta$E 76. In Table \ref{CIC27:tab:quantitative}, the mean, lower quartile (Q1), median (Q2), and the upper quartile (Q3) of the error between the WB modified images and the corresponding ground truth images are reported. It can be observed that our method consistently outperforms the diagonal manipulation, both with and without the commonly used pre-linearization step using the 2.2 inverse gamma \cite{ebner2007color}, in all metrics.

Figure \ref{CIC27:fig:quantative_results} shows the effect of different downsampling sizes on the post-processing WB modification compared with rendering the original raw-RGB image with each target color temperature (i.e., ground truth). In this example, we randomly selected 300 raw-RGB images from the NUS dataset. Each image is rendered using our method, and then modified to different target WB settings. This process was repeated three times to study the error propagation of our post-processing modifications.

Representative qualitative results from the NUS dataset are shown in Figs. 
\ref{CIC27:fig:results2} and \ref{CIC27:fig:results3}. As can be observed, our method produces better results compared to the sRGB diagonal WB manipulation and Adobe Lightroom.

\section{Summary}

We have described an imaging framework that enables users to accurately modify the WB color temperature of an sRGB-rendered image.  Such functionality is currently not possible with the conventional imaging pipeline.  With our method, images typically discarded due to the camera having the wrong WB setting, can be easily fixed by changing to the correct color temperature.  In addition, our approach enables editing for aesthetic manipulation of the image's appearance. Our approach requires a minor modification to the existing imaging pipeline and produces metadata that can be easily stored in an image file (e.g. JPEG) with only 6 KB of overhead.

\part{Color Enhancement\label{part:enhancement}}

\chapter{Color Enhancement Through the CIE XYZ Space \label{ch:ch12}}
Previous chapters (Chapters \ref{ch:ch5}--\ref{ch:ch11}) discussed methods for computational CC and image white balancing in sensor-raw and sRGB color spaces. This part of the thesis focuses on other degradation  factors that can effect the colors of captured photographs. Specifically, we focus on low-light images that either are captured using under-exposure settings or have scenes with low-lighting conditions. We further discuss a less explored area of research, where we propose a method to correct colors of over-exposed images in the next chapter.

The work in this chapter treats the problem of enhancing low-light captured images from a different angle, where we employ a scene-referred image state of the cameras ISP.  Cameras currently allow access to two image states: (i) a minimally processed linear raw-RGB image state (i.e., raw sensor data) or (ii) a highly-processed nonlinear image state (e.g., sRGB).  There are many computer vision tasks that work best with a linear image state, such as image dehazing and deblurring.  Unfortunately, the vast majority of images are saved in the nonlinear image state.  Because of this, a number of methods have been proposed to ``unprocess'' nonlinear images back to a raw-RGB state.  However, existing unprocessing methods have a drawback because raw-RGB images are sensor-specific.  As a result, it is necessary to know which camera produced the sRGB output and use a method or network tailored for that sensor to properly unprocess it.  This chapter addresses this limitation by exploiting another camera image state that is not available as an output, but it is available inside the camera pipeline\footnote{This work was published in \cite{afifi2020cie}: Mahmoud Afifi, Abdelrahman Abdelhamed, Abdullah Abuolaim, Abhijith Punnappurath, Michael S. Brown. CIE XYZ Net: Unprocessing Images for Low-Level Computer Vision Tasks. IEEE Transactions on Pattern Analysis and Machine Intelligence (TPAMI), 2021.}.  In particular, cameras apply a colorimetric conversion step to convert the raw-RGB image to a device-independent space based on the CIE XYZ color space before they apply the nonlinear photo-finishing.   Leveraging this canonical image state, we propose a deep learning framework, CIE XYZ Net, that can unprocess a nonlinear image back to the canonical CIE XYZ image.   This image can then be processed by any low-level computer vision operator and re-rendered back to the nonlinear image.  We demonstrate the usefulness of the CIE XYZ Net on enhancing low-light images and show significant gains that can be obtained by this processing framework.  The source code and dataset of this work are available on GitHub: Code and dataset are publicly available at \href{https://github.com/mahmoudnafifi/CIE_XYZ_NET}{https://github.com/mahmoudnafifi/CIE$\_$XYZ$\_$NET}.

\section{Introduction}\label{xyz:sec:intro}

As discussed earlier, an a camera's ISP hardware processes the initial captured sensor image in a pipeline fashion, with routines being applied one after the other.  The ISP used by consumer cameras performs operations as two distinct stages.  First, a ``front-end'' stage applies linear operations, such as white
balance and color adaptation, to convert the sensor-specific raw-RGB image to a device-independent color
space (e.g., CIE XYZ or its wide-gamut representation, ProPhoto)~\cite{Le2020gamut}. The image states associated with the front-end process are called a \textit{scene-referred} image because the image remains related directly to initial recorded sensor values related to the physical scene. Next, a ``photo-finishing'' stage is performed that applies nonlinear steps and local operators to produce a visually
pleasing photograph.  For example, selective color manipulation is often applied to enhance skin tone or make the overall colors more vivid, while local tone manipulation increases local contrast within the image.    After the photo-finishing stage, the image is encoded in an output color space (e.g., sRGB, AdobeRGB, or Display P3).  The image states associated with the photo-finishing process are referred to as \textit{display-referred} as they are encoded for visual display. Cameras currently allow access only to either the
minimally processed scene-referred image state (i.e., raw-RGB image) or the final display-referred image state (e.g., sRGB, AdobeRGB, or Display P3).  Unfortunately, these two image states are not ideal for low-level computer vision tasks.

\begin{figure}[t]
\includegraphics[width=\linewidth]{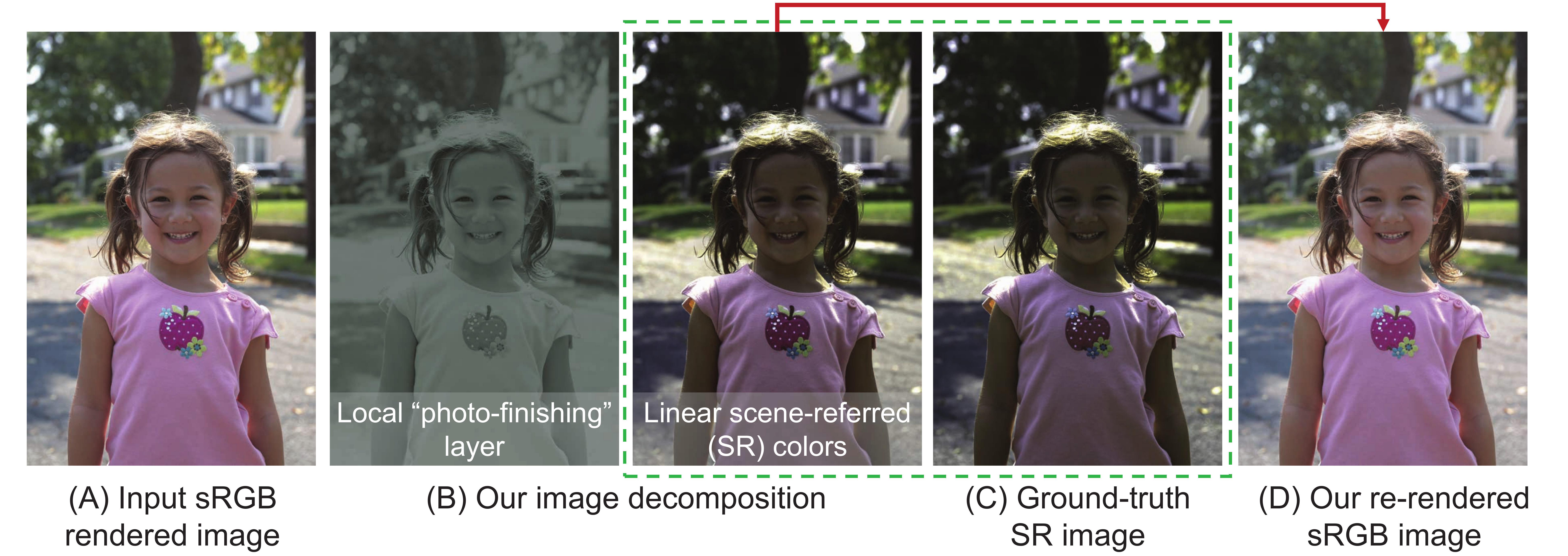}
\vspace{-7mm}
\caption[We propose a cycle framework that can unprocess sRGB images back to the linear CIE XYZ color space and re-render the CIE XYZ images into the nonlinear sRGB color space.]{We propose a cycle framework that can unprocess sRGB images back to the linear CIE XYZ color space and re-render the CIE XYZ images into the nonlinear sRGB color space. (A) The input camera-rendered sRGB image. (B) Our image decomposition (left: residual photo-finishing layer, right: scene-referred CIE XYZ reconstruction). (C) The ground-truth scene-referred CIE XYZ image. (D) Our re-rendering result from the reconstructed CIE XYZ image. To aid visualization, CIE XYZ images are scaled by a factor of two. Input image is taken from the MIT-Adobe FiveK dataset \cite{bychkovsky2011learning}.}
\label{xyz:fig:teaser}
\end{figure}

The raw-RGB image state preserves the linear relationship of incident scene radiance. This linear image formation makes raw-RGB images suitable for a wide range of low-level computer vision tasks, such as image deblurring, image dehazing, image denoising, and various types of image enhancement~\cite{tai2013nonlinear, nguyen2016raw, brooks2018unprocessing, zamir2020cycleisp}.  However, the drawback of raw-RGB is that the physical color filter arrays that make up the sensor's Bayer pattern are sensor-specific.  This means raw-RGB values captured of the same scene but with different sensors are significantly different~\cite{nguyen2014raw}.  This often requires learning-based methods to be trained per sensor or camera make and model (e.g.,~\cite{diamond2017dirty, nam2017modelling, hu2017fc, brooks2018unprocessing, afifi2019sensor}).

The more common display-referred image state (in this chapter, assumed to be in the sRGB color space) also has drawbacks.  While this image state is the most widely used and is suitable for display,  cameras apply their own proprietary photo-finishing to enhance the visual quality of the image.  This means images captured of the same scene but using different camera models (and sometimes the same camera but with different settings) will produce images that have significantly different sRGB values~\cite{kim2012new, karaimer2016software, nguyen2016raw}.

As previously discussed, the front-end processor of a typical camera ISP performs a colorimetric conversion to map the raw-RGB image to a standard perceptual colorspace---namely, CIE 1931 XYZ~\cite{karaimer2016software}.  While there exists no formal image encoding for this image state, it is possible to convert existing raw-RGB images stored in digital negative (DNG) format to this intermediate state by applying a software camera ISP (e.g.,~\cite{abdelhamed2018high, karaimer2016software}). This provides a mechanism to standardize all images into a canonical linear scene-referred image state and is the impetus of our work.

\paragraph{Contribution}~We propose a method to decompose non-linear sRGB images into two parts: 1) a canonical linear scene-referred image state in the CIE XYZ color space and 2) a residual image layer that resembles additional non-linear and local photo-finishing operations. Through such decomposition strategy, we learn a model that can accurately map back and forth between non-linear sRGB and linear CIE XYZ images. An example is shown in Fig.~\ref{xyz:fig:teaser}. Unlike raw-RGB, the CIE XYZ color space is {\it device-independent}, and as a result, helps with model generalization. Furthermore, CIE XYZ images can be encoded as standard three-channel images that can be easily handled by existing computer vision frameworks. We show that our proposed model maps images back to the CIE XYZ color space more accurately compared to alternative approaches.  In addition, we perform experiments on low-light image enhancement to show that employing our proposed CIE XYZ model provides the performance boost anticipated from using linear images (additional computer vision applications of our CIE XYZ model are provided in Appendix \ref{ch:appendix0}).

\section{Related Work}\label{xyz:sec:related}

In this section, we review various methods proposed for linearization of camera-rendered images. More details about camera imaging pipeline and linearization methods are provided in Chapter \ref{ch:ch2}.

\subsection{Camera-Rendered Image Linearization} \label{xyz:subsec:related-lin}

To obtain a linear image from its camera-rendered version, we need to reverse the nonlinear camera-rendering stage in the pipeline. Many methods have been proposed to model a parametric relationship that maps from the camera-rendered image (i.e., sRGB image) back to its raw-RGB version (e.g.,~\cite{nguyen2016raw}). However, raw-RGB space is camera-dependent and requires having a separate model per camera.
As discussed in Chapter \ref{ch:ch2}, other approaches involve simple linearization by inverting the global tone mapping and the gamma compression followed by applying a linearization matrix to obtain a linear sRGB or CIE XYZ image~\cite{brooks2018unprocessing}. Such approaches are too simple and do not account for the local processing or dynamic range adjustments.
Unlike prior approaches, instead of trying only to obtain a linear image, our approach is to decompose the nonlinear image into globally processed and locally processed layers. The locally processed layer represents local color processing, such as local tone mapping. Then, we learn a global mapping from the globally processed image to the linear image.
Another line of research targeting the problem of image linearization is radiometric calibration~\cite{lin2011revisiting, chakrabarti2014modeling} (see as Chapter \ref{ch:ch2} for more details). Unlike our approach, radiometric calibration methods do not target a specific, well-defined color space, and do not address the problem of local processing.

\section{Methodology}\label{xyz:sec:method}

This section describes our overall framework, including network architecture, dataset generation, and training details.

\subsection{Formulation} \label{xyz:subsec:formulation}

Inside a camera imaging pipeline, a raw-RGB image $\imraw \in \mathbb{R}^{h \times w}$ undergoes a sequence of processing stages to be transformed to the final output sRGB image $\imsrgb \in \mathbb{R}^{h \times w \times 3}$, where $h$ and $w$ represent the image height and width, respectively.

As mentioned earlier, the raw-RGB image $\imraw$ is in a camera-dependent color space that is linear with respect to scene light irradiance falling on the sensor.  One of the early steps in the camera processing pipeline is to convert the camera-dependent color space to a device-independent color space---namely, CIE XYZ.  Based on this observation, instead of modeling the whole pipeline back to the raw-RGB image, we choose to model an intermediate representation of the image in the CIE XYZ color space $\imxyz \in \mathbb{R}^{h \times w \times 3}$ that is still linear with respect to scene irradiance, but is in a canonical color space.  We are interested in the on-camera rendering procedures that map the CIE XYZ images into the final display-referred (i.e., photo-finished) sRGB color space. This operation can be described as
\begin{equation}
\imsrgb = \pipe(\imxyz).
\end{equation}

In our method, instead of relying on a single function to model the pipeline stages between sRGB and CIE XYZ, we decompose this mapping into two parts: 1) \textit{global processing}, denoted collectively as $\pipeglob(\cdot)$, that is globally applied to all image pixels and 2) \textit{local processing}, denoted collectively as $\pipeloc(\cdot)$, that represents local photo-finishing operations, such as local tone mapping and selective color adjustments.

Such design is largely motivated by the fact that actual camera ISPs perform both global and local image processing. Global processing can be easily modeled by a polynomial color mapping, and hence, we train a CNN to estimate such polynomial function coefficients. Local processing is more challenging to model, and hence, we chose to model it as a residual fully-convolutional neural network.
Breaking the process into two parts, global and local, has the added advantage that it enables image enhancement methods to selectively manipulate either part independently, and improves the performance of various photo-finishing tasks, as we will demonstrate in our experiments.

Our forward pipeline from $\imxyz$ to $\imsrgb$ can be represented as a cascade of the global and the local processes.
The global processing stage is represented as
\begin{equation}
\matfwd = \pipeglob(\imxyz)  ,
\label{xyz:eq:matfwd}
\end{equation}
\begin{equation}
\imglob = \reshape \left( \matfwd \ \kernelSixbyN(\imxyz) \right) ,
\label{xyz:eq:imglob-fwd}
\end{equation}

\noindent where $\matfwd \in \mathbb{R}^{3 \times 6}$ is a global transformation matrix and $\imglob$ is the globally processed image layer. The operator $\kernelSixbyN(\cdot)$ reshapes the image to be $6 \times n$ where $n$ is the number of pixels in the image and each pixel is transformed from three to six dimensions: $[R, G, B] \rightarrow [R, G, B, R^2, G^2, B^2]$, while the operator $\reshape$ reshapes the image from $3 \times n$ back to $h \times w \times 3$.
We chose $\matfwd$ to be nonlinear to capture global color processing operations, such as gamma compression.

As most consumer cameras locally process the captured scene-referred images to improve the quality of final rendered images \cite{hasinoff2016burst}, such global color processing may not be able to effectively model the function $\pipe$. To that end, we use a residual learning mechanism where we model the residual layer $\imres$ between the locally and globally processed layers of the image as follows:

\begin{equation}
\imres = \pipeloc(\imglob) ,
\label{xyz:eq:imres-fwd}
\end{equation}%
\begin{equation}
\imsrgb = \imglob + \imres .
\label{xyz:eq:imsrgb-fwd}
\end{equation}%

Now, the decomposition process applies the inverse process of Eqs.~\ref{xyz:eq:matfwd} -- \ref{xyz:eq:imsrgb-fwd} as follows:
\begin{equation}
\imres = \invpipeloc(\imsrgb) , \label{xyz:eq:imres-inv}
\end{equation}%
\begin{equation}
\imglob = \imsrgb - \imres , \label{xyz:eq:imglob-inv}
\end{equation}%
\begin{equation}
\matinv = \invpipeglob(\imglob), \label{xyz:eq:matinv}
\end{equation}%
\begin{equation}
\imxyz = \reshape \left( \matinv \ \kernelSixbyN(\imglob) \right) ,
\label{xyz:eq:imxyz-inv}
\end{equation}
where $\invpipeloc(\cdot)$ represents the inverse of residual local processing layer and $\invpipeglob(\cdot)$ is constrained to produce a global transformation matrix $\matinv \in \mathbb{R}^{3 \times 6}$ that represents the inverse global processing stage.

\begin{figure}[t]
\includegraphics[width=\linewidth]{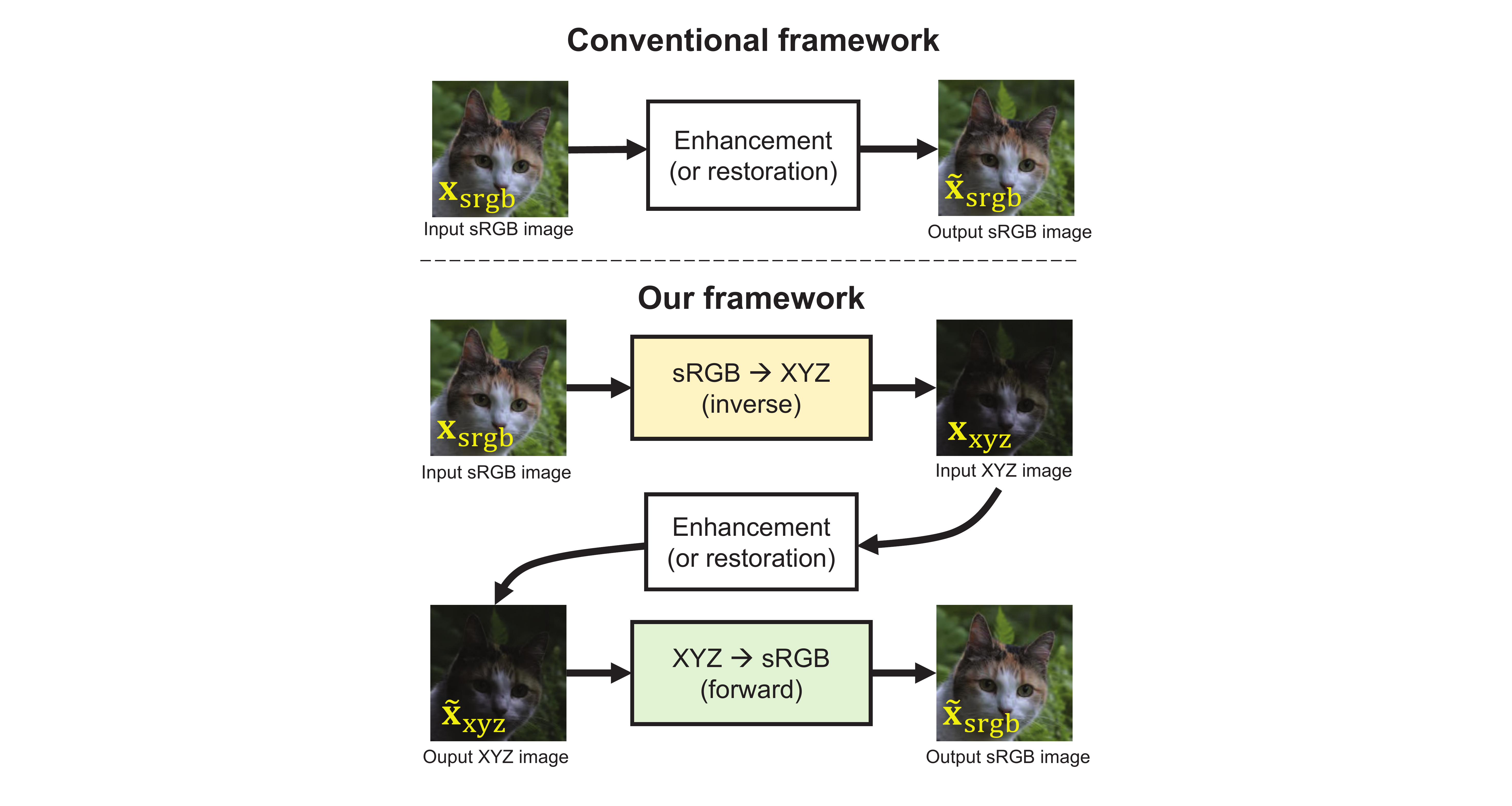}
\vspace{-7mm}
\caption[An illustration of using our inverse and forward image processing pipelines in an sRGB image restoration/enhancement framework.]{An illustration of using our inverse and forward image processing pipelines in an sRGB image restoration/enhancement framework.}
\label{xyz:fig:framework}
\end{figure}

Our ultimate goal is to allow the manipulation of the reconstructed CIE XYZ image by arbitrary image restoration/enhancement algorithms between the inverse and forward pipeline stages (see Fig. \ref{xyz:fig:framework}).
It is, however, non-trivial to infer the inverse functions $\invpipeloc^{-1}(\cdot)$ and $\invpipeglob^{-1}(\cdot)$ to render back the reconstructed image, as its values may be changed by the image restoration or enhancement algorithms. To that end, we model each of $\pipeglob(\cdot)$, $\pipeloc(\cdot)$, $\invpipeglob(\cdot)$, and $\invpipeloc(\cdot)$ by a neural network.

\subsection{Network Design} \label{xyz:subsec:network}

\begin{figure*}[t]
\includegraphics[width=\linewidth]{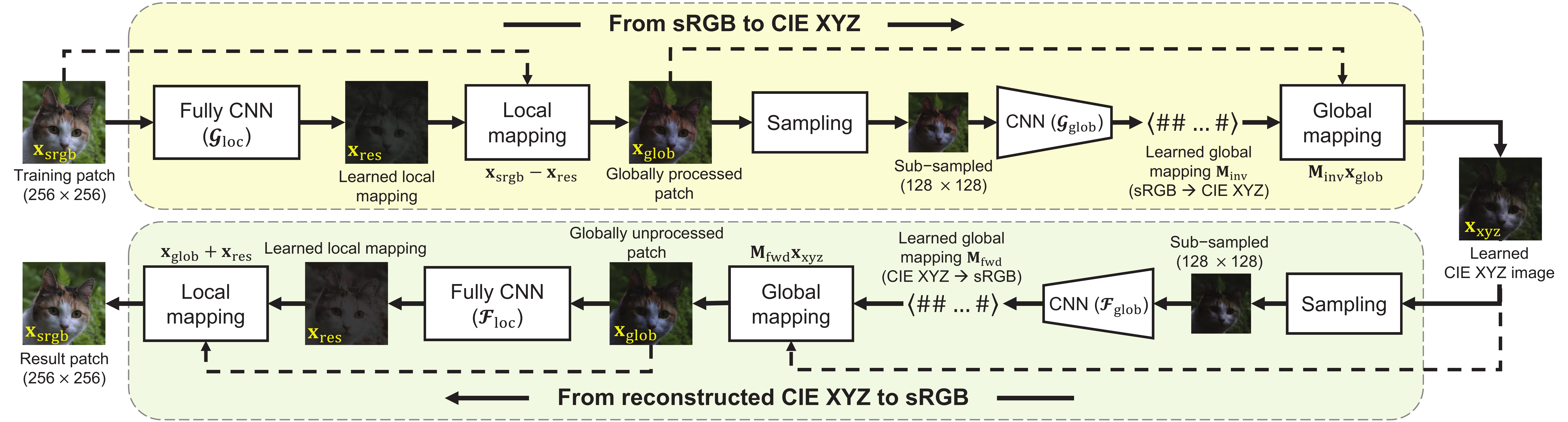}
\vspace{-7mm}
\caption[Our CIE XYZ image pipeline.]{Our CIE XYZ image pipeline. The upper part is the inverse pipeline that unprocesses an sRGB image into a CIE XYZ image. The lower part is the forward pipeline that processes a CIE XYZ image into its equivalent sRGB image. The full framework is trainable end-to-end. The CIE XYZ images are scaled 2x to aid visualization.}
\label{xyz:fig:overview}
\end{figure*}

Imitating this division of the camera imaging pipeline, we build our network architecture to include two sub-networks for modeling both the global and local processing parts for the forward and inverse directions of the imaging pipeline. As shown in Fig.~\ref{xyz:fig:overview}, we start with the inverse pipeline where the first part is a fully-convolutional neural network (CNN) that models the local processing applied to an input non-linear image (i.e., sRGB image) by predicting the residual image $\imres$ (Eq.~\ref{xyz:eq:imres-inv}). Once the local processing layer is predicted, it can be subtracted from the input image $\imsrgb$ to get the globally processed image $\imglob$ (Eq.~\ref{xyz:eq:imglob-inv}). Then, $\imglob$ is fed to another sub-network that predicts a global transformation $\matinv$ that inverts $\imglob$ back to the linear CIE XYZ image $\imxyz$ (Eq.~\ref{xyz:eq:imxyz-inv}). With this inverse pipeline, we decompose the input image $\imsrgb$ into two image layers, $\imres$ and $\imglob$, which represent local and  global processing, respectively, and finally output the linear CIE XYZ image $\imxyz$.

As discussed in Section~\ref{xyz:sec:intro}, there are computer vision tasks, such as image restoration, that are best processed in a linear image state.  A use case of the framework is to convert the input image $imxyz$, process the $imxyz$ image, and then render the image back.  In this scenario, after decomposing an image and applying an image restoration task to the linear XYZ image, we now need to merge these image layers back to produce the fully processed sRGB image. To model this forward pass of our pipeline, as shown in Fig.~\ref{xyz:fig:overview}, we use two sub-networks. The first sub-network predicts a global transformation $\matfwd$ that maps $\imxyz$ to $\imglob$ (Eq.~\ref{xyz:eq:imglob-fwd}). The second sub-network predicts the residual local processing $\imres$ that needs to be applied to $\imglob$ to obtain the final sRGB image $\imsrgb$ (Eq.~\ref{xyz:eq:imsrgb-fwd}). This framework is illustrated in Fig.~\ref{xyz:fig:framework} and compared to the conventional way of directly processing the sRGB image.

\begin{figure}[t]
\includegraphics[width=\linewidth]{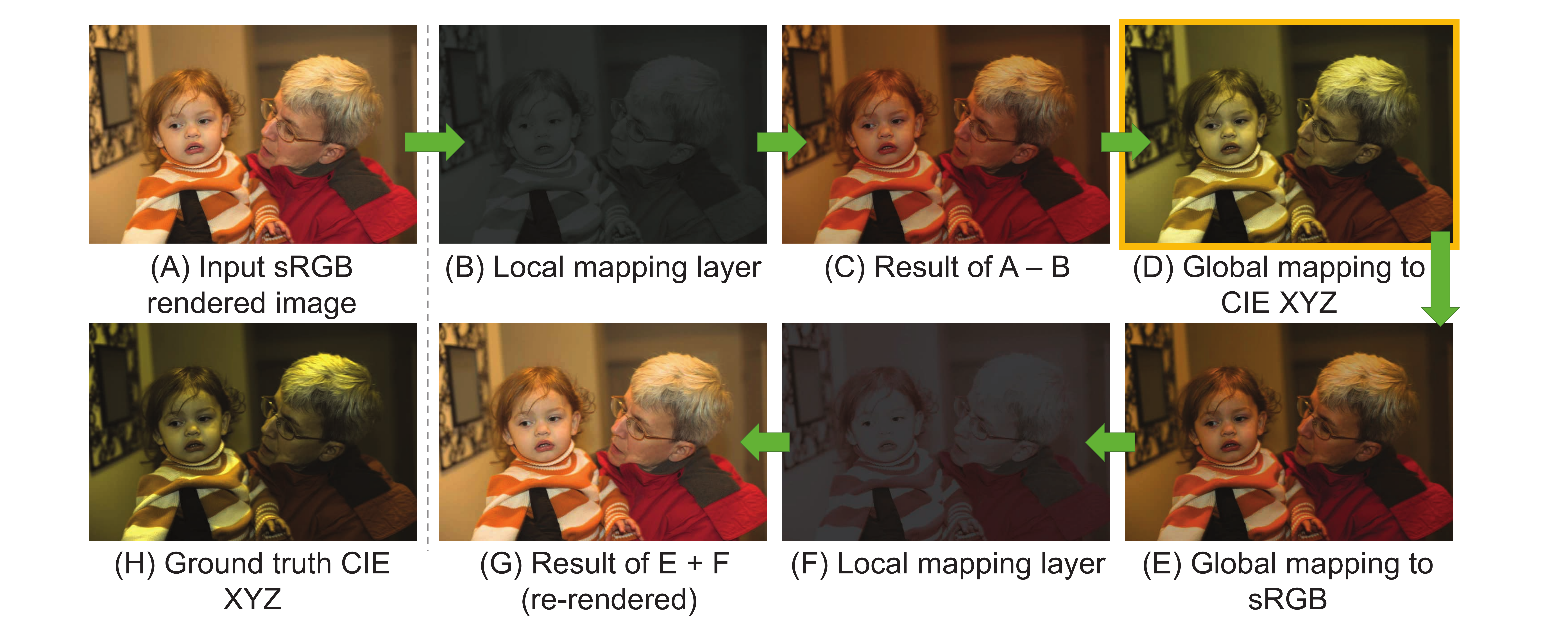}
\vspace{-7mm}
\caption[Our inverse pipeline decomposites a given camera-rendered sRGB image into a local processed layer and the corresponding CIE XYZ image, while our forward pipeline maps the reconstructed CIE XYZ image to the sRGB color space in an inverse way of our decomposition.]{Our inverse pipeline decomposites a given camera-rendered sRGB image into a local processed layer and the corresponding CIE XYZ image, while our forward pipeline maps the reconstructed CIE XYZ image to the sRGB color space in an inverse way of our decomposition. The shown image is taken from our testing set. To aid visualization, CIE XYZ images are scaled by a factor of two.}
\label{xyz:fig:analysis}
\end{figure}

In order to allow the networks $\invpipeglob(\cdot)$ and $\invpipeloc(\cdot)$ to separate the globally and locally processed image layer without having ground truth for both $\imglob$ and $\imres$, we apply a scaling factor to the output of the local processing networks $\invpipeglob(\cdot)$, in both inverse and forward passes, such that the values of $\imres$ are much smaller than $\imglob$. In our experiments, we set this scaling factor to $0.25$. Figure \ref{xyz:fig:analysis} shows an example of the output of each sub-network.

It is challenging to evaluate our global and local processing modules separately; mainly because camera ISP global and local processing modules are typically proprietary and not accessible, and hence, we cannot obtain ground truth data for evaluation. 

\subsection{Loss Function}
The objective of the whole network is to minimize the mean absolute error: 1) between the predicted XYZ image $\imxyzpred$ and its ground truth $\imxyzgt$ in the inverse pipeline and 2) between the predicted sRGB image $\imsrgbpred$ and its ground truth $\imsrgbgt$ in the forward pipeline, as follows:
\begin{equation}\label{xyz:eq:loss}
\lambda \abs{\imxyzpred - \imxyzgt} + \abs{\imsrgbpred - \imsrgbgt},
\end{equation}%
where $\lambda$ is a weighting factor that we use to deal with the fact that XYZ images generally have lower intensity compared to sRGB images; so this weight can balance the learning behavior between the forward and inverse pipelines. In our experiments, we set $\lambda = 1.5$.

\subsection{Sub-Networks Architecture} 
Our local processing sub-networks ($\pipeloc$ and $\invpipeloc$) each consist of 15 blocks of $3\!\times\!3$ convolutional (conv)--LReLU layers. Each conv layer has 32 output channels, with stride of 1 and padding of 1. The last layer of these sub-networks has a single conv layer with three output channels, followed by a $\tanh$ operator.
As our global processing sub-networks are not fully convolutional, we use a fixed size of input by introducing a differentiable subsampling module that uniformly subsamples $128\!\times\!128$ color values of the processed image by the previous sub-network.
Our global sub-network includes five blocks of $3\!\times\!3$ conv--LReLU--$2\!\times\!2$ max pooling layers. The conv layers have stride and padding of 1, while the max pooling layers have a stride factor of 2 with no padding. Then, we added a fully connected layer with 1024 output neurons, followed by a dropout layer with a factor of 0.5. The last layer of our global sub-network has a fully connected layer with 18 output neurons to formulate our $3\!\times\!6$ polynomial mapping function.

Our entire framework is a light-weight model with a total of 2,697,578 learnable parameters ($\sim$11MB of memory) for both sRGB-to-XYZ and XYZ-to-sRGB models, and it is fully differentiable for end-to-end training.

\subsection{Dataset} \label{xyz:subsec:dataset}

To train our proposed model, we need a dataset of sRGB images with their corresponding linear images in the CIE XYZ color space. To do so, we start from raw-RGB images taken from the MIT-Adobe FiveK~\cite{bychkovsky2011learning}. We then process the raw-RGB images twice to obtain both the sRGB and XYZ versions of each image. For processing raw-RGB images into the XYZ color space, we used the camera pipeline from~\cite{abdelhamed2018high}. This pipeline provides an access to the CIE XYZ values after processing the sensor raw-RGB using the color space transformation (CST) matrices provided with the raw-RGB image.
To obtain the camera-like sRGB images, we followed the same procedure explained in Chapter\ \ref{ch:ch7} to generate our sRGB dataset for improperly white-balanced images. Specifically, we used the Adobe Camera RAW SDK, which accurately emulates the nonlinearity applied by consumer cameras \cite{afifi2019color}. In contrast to the dataset generated in Chapter\ \ref{ch:ch7}, we used the illuminant color estimated by each camera's ISP in the AWB mode.

The MIT-Adobe FiveK~\cite{bychkovsky2011learning} dataset contains images captured with different cameras. As a result, the CIE XYZ and sRGB images are rendered with different processing profiles according to the metadata from each camera. 

Our method's CIE XYZ fidelity evaluations are tied to the used camera models in the MIT-Adobe FiveK~\cite{bychkovsky2011learning}. However, this  does not take away from the advantages that the standard canonical CIE XYZ space offers over other linear spaces as a target space for our supervision learning. Our assumption is that by training on images rendered by a broad range of ISP emulations for different camera models, we can learn to unprocess generic processing applied by most cameras in order to achieve a better linearization.
Our dataset includes $\sim$1,200 pairs of sRGB and camera CIE XYZ images.
Our dataset will be publicly available upon acceptance.  

\subsection{Training} \label{xyz:subsec:method-training}

We divided our dataset into a training set of 971 pairs, a validation set of 50 pairs, and a testing set of 244 pairs. We trained our framework in an end-to-end manner on patches of size $256\!\times\!256$ pixels randomly extracted from our training set, with a mini-batch of size 4. We applied random geometric augmentation (i.e., scaling and reflection) to the extracted patches.

Our framework was trained in an end-to-end manner for 300 epochs using Adam optimizer~\cite{kingma2014adam} with gradient decay factor $\beta_1= 0.9$ and squared gradient decay factor $\beta_2 =0.999$. We used a learning rate of $10^{-4}$ with a drop factor of 0.5 every 75 epochs. We added an $L_2$ regularization with a weight of $\lambda_{\mathrm{reg}} = 10^{-3}$ to our loss in Eq.\ \ref{xyz:eq:loss} to avoid overfitting.

\section{Experimental Results}\label{xyz:sec:results}
In this section, we first validate the effectiveness of our proposed model in mapping from camera-rendered sRGB images to CIE XYZ, and processing CIE XYZ images back to sRGB. Next, we demonstrate our method's utility on low-light image enhancement task. We refer the reader to Appendix \ref{ch:appendix0} for additional applications. 

\subsection{From Camera-Rendered sRGB to CIE XYZ, and Back} \label{xyz:subsec:results-XYZ}
We first verify our network's ability to unprocess sRGB images to CIE XYZ. We also demonstrate our ability to reconstruct from CIE XYZ back to sRGB. We test our mapping to sRGB using our reconstructed CIE XYZ results as a starting point, and also using the ground-truth CIE XYZ images.

We compared our method with three existing methods. First, we compare with the \textit{standard CIE XYZ mapping}~\cite{anderson1996proposal, ebner2007color}, which applies a simple 2.2 gamma tone curve. Second, we compare with the recent \textit{unprocessing technique (UPI)} from~\cite{brooks2018unprocessing}. The UPI unprocessing technique is non-trainable and inverts the camera ISP, step-by-step, through a series of transformations, such as gamma expansion and inverting color correction matrices. This unprocessing module is then used to generate realistic training data for the task for image denoising. It is only the denoising module of~\cite{brooks2018unprocessing} that is a CNN, and that is trainable. For a fair comparison, we compare our results with results of UPI obtained at the CIE XYZ stage.

Third, we compare with CycleISP~\cite{zamir2020cycleisp}, a recent network architecture that aims at simulating the camera ISP mapping between raw and sRGB stages. Since CycleISP is targeting a camera-specific sRGB-to-raw mapping, we had to introduce some slight modifications to their architecture to make it suitable to our objective, i.e., mapping between sRGB and CIE XYZ. In particular, we omitted their noise injection and color correction modules and modified their RGB2RAW and RAW2RGB networks to have 3-channel inputs and outputs. We retrained the CycleISP with the same training settings used to train our model for a fair comparison.

\begin{figure}[!t]
\includegraphics[width=\linewidth]{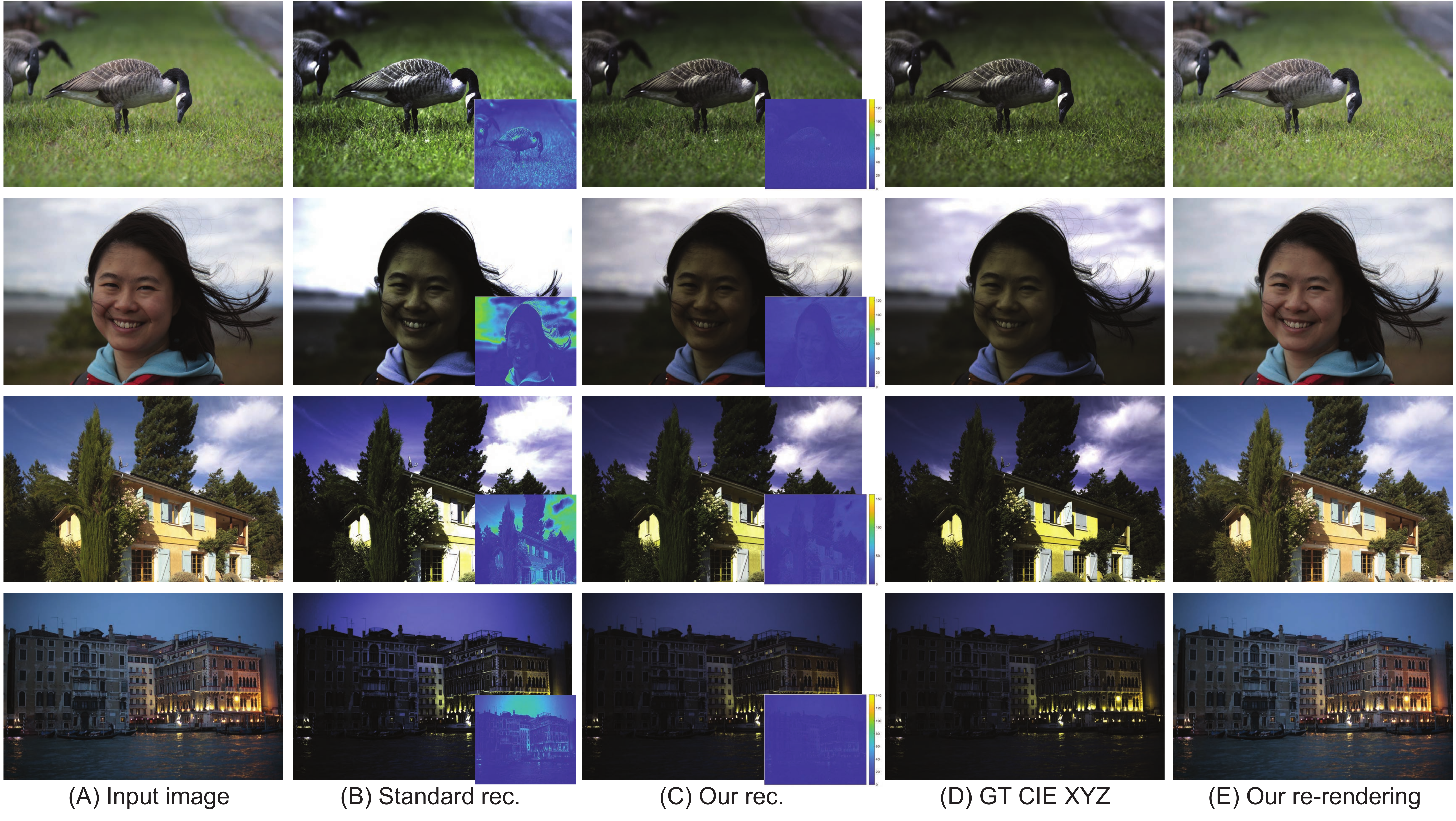}
\vspace{-7mm}
\caption[Qualitative comparisons for CIE XYZ reconstruction and rendering.]{Qualitative comparisons for CIE XYZ reconstruction and rendering. (A) The input sRGB rendered image. (B) Standard display-referred CIE XYZ reconstruction \cite{anderson1996proposal, ebner2007color}. (C) Our reconstruction. (D) The ground-truth scene-referred CIE XYZ image. (E) Our re-rendering result from the reconstructed CIE XYZ image. To aid visualization, CIE XYZ images are scaled by a factor of two. Input images are taken from the MIT-Adobe FiveK dataset \cite{bychkovsky2011learning}.}
\label{xyz:fig:qualitative}
\end{figure}

Table \ref{xyz:Table0} shows peak signal-to-noise ratio (PSNR) results averaged over 244 unseen testing images from the MIT-Adobe FiveK dataset~\cite{bychkovsky2011learning}. The terms Q1, Q2, and Q3 refer to the first, second (median), and third quantile, respectively, of the PSNR values obtained by each method. For the standard XYZ, the results of mapping from the reconstructed CIE XYZ images back to sRGB are not reported because standard XYZ uses an invertible transform. The sRGB reconstruction error from the UPI model~\cite{brooks2018unprocessing} is high due to the fact that the tone mapping is not perfectly invertible.
It can be observed from the results that we outperform all competing methods by a sound margin. Qualitative comparisons are provided in Fig. \ref{xyz:fig:qualitative}.

As shown in Table~\ref{xyz:Table0}, the mapping to sRGB from reconstructed CIE XYZ is better than mapping from ground-truth CIE XYZ. For our method, this behavior is expected because the forward model is trained on the reconstructed CIE XYZ, not the ground truth. Also, for the UPI method, as it is based on matrix inversion, the mapping from the reconstructed CIE XYZ makes the transformation more accurate than mapping from the ground truth.

Lastly, we did experiment with different choices for the global mapping polynomial to validate the polynomial kernel used in our method (see Table~\ref{xyz:tab:poly-ablation}) %
We found that our chosen ($[R, G, B, R^2, G^2, B^2]$) polynomial yielded the best results.

\begin{table}[!t]
\caption[Results of camera-rendered sRGB $\leftrightarrow$ CIE XYZ mapping.]{Results (in terms of PSNR) of camera-rendered sRGB $\leftrightarrow$ CIE XYZ mapping. We compare our results against the standard XYZ mapping (the 2.2 gamma tone curve)~\cite{anderson1996proposal, ebner2007color}, the recent unprocessing technique (UPI)~\cite{brooks2018unprocessing}, and CycleISP~\cite{zamir2020cycleisp}. Average PSNR (dB) results are reported on 244 unseen testing pairs (camera-rendered sRGB and corresponding CIE XYZ images) from the MIT-Adobe FiveK dataset~\cite{bychkovsky2011learning}. We show results of mapping from both reconstructed (Rec.) CIE XYZ images and ground truth (GT) CIE XYZ images to the corresponding camera-rendered sRGB images. Highest PSNR values are shown in boldface and highlighted in yellow.\label{xyz:Table0}}
\centering
\scalebox{0.65}{
\begin{tabular}{|l|c|c|c|c|c|c|c|c|c|c|c|c|}
\hline
\multicolumn{1}{|c|}{} & \multicolumn{4}{c|}{\textbf{sRGB $\rightarrow$ XYZ}} & \multicolumn{4}{c|}{\textbf{Rec. XYZ $\rightarrow$ sRGB}} & \multicolumn{4}{c|}{\textbf{GT XYZ $\rightarrow$ sRGB}}\\ \cline{2-13}
\multicolumn{1}{|c|}{\multirow{-2}{*}{\textbf{Method}}} & \textbf{Avg.} & \textbf{Q1} & \textbf{Q2} & \textbf{Q3} & \textbf{Avg.} & \textbf{Q1} & \textbf{Q2} & \textbf{Q3} & \textbf{Avg.} & \textbf{Q1} & \textbf{Q2} & \textbf{Q3}  \\ \hline

Standard \cite{anderson1996proposal, ebner2007color} &  21.84 & 16.88 & 20.91 &  25.24 & - & - & -  & - & 22.22 & 19.19 & 21.79 &  24.37 \\ \hline

Unprocessing \cite{brooks2018unprocessing}  & 22.19 & 19.31 & 22.12 & 24.75 & 37.72 & 37.78 & 40.56 & 41.88 & 18.04 & 15.67 & 17.79 & 20.02 \\ \hline

CycleISP~\cite{zamir2020cycleisp} & 28.29 & 23.63 & 28.08 & 31.98 & 34.78 & 30.60 & 34.20 & 37.22 & 20.91 & 18.36 & 21.42 & 24.31
\\ \hline

Ours & \cellcolor[HTML]{\bestcolor}\textbf{29.66} & \cellcolor[HTML]{\bestcolor}\textbf{23.77} & \cellcolor[HTML]{\bestcolor}\textbf{29.57} & \cellcolor[HTML]{\bestcolor}\textbf{34.71} & \cellcolor[HTML]{\bestcolor}\textbf{43.82} & \cellcolor[HTML]{\bestcolor}\textbf{41.43} & \cellcolor[HTML]{\bestcolor}\textbf{43.94} & \cellcolor[HTML]{\bestcolor}\textbf{46.58} & \cellcolor[HTML]{\bestcolor}\textbf{27.44} & \cellcolor[HTML]{\bestcolor}\textbf{23.57} & \cellcolor[HTML]{\bestcolor}\textbf{28.32} & \cellcolor[HTML]{\bestcolor}\textbf{30.88} \\ \hline
\end{tabular}
}
\end{table}

\begin{figure}[!t]
\includegraphics[width=\linewidth]{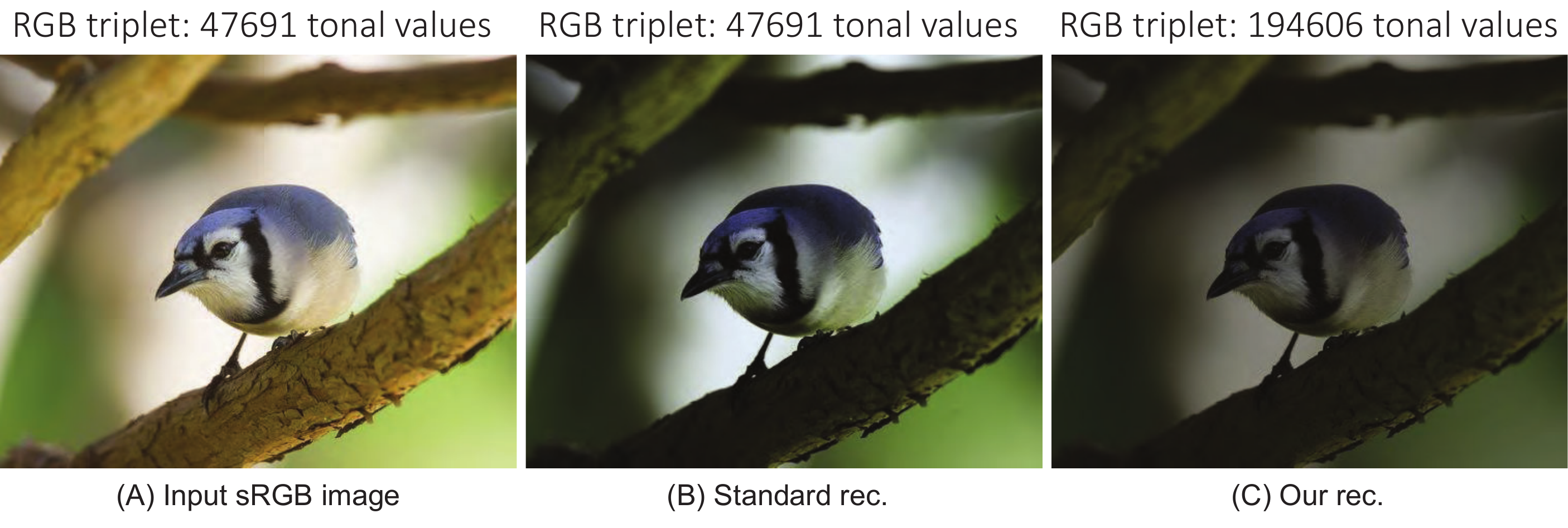}
\vspace{-7mm}
\caption[Our XYZ reconstruction provides a wider range of tonal values compared to the standard CIE XYZ mapping~\cite{anderson1996proposal, ebner2007color}.]{Our XYZ reconstruction provides a wider range of tonal values compared to the standard CIE XYZ mapping~\cite{anderson1996proposal, ebner2007color}. (A) The input sRGB image. (B) Standard XYZ reconstruction \cite{anderson1996proposal, ebner2007color}. (C) Our XYZ reconstruction. The input image is taken from~\cite{wah2011caltech}.}
\label{xyz:fig:tonal_values}
\end{figure}

\begin{table}[!t]
\caption[Effect of different polynomial terms on global mapping results for mapping camera-rendered sRGB to CIE XYZ mapping; and mapping from both reconstructed CIE XYZ images and ground truth CIE XYZ images to the corresponding camera-rendered sRGB images.]{Effect of different polynomial terms on global mapping results (in terms of PSNR) for mapping camera-rendered sRGB to CIE XYZ mapping; and mapping from both reconstructed (Rec.) CIE XYZ images and ground truth (GT) CIE XYZ images to the corresponding camera-rendered sRGB images. Highest PSNR values are shown in boldface and highlighted in yellow.\label{xyz:tab:poly-ablation}}
\centering
\scalebox{0.6}{
\begin{tabular}{|l|c|c|c|c|c|c|c|c|c|c|c|c|}
\hline
\multicolumn{1}{|c|}{} & \multicolumn{4}{c|}{\textbf{sRGB $\rightarrow$ XYZ}} & \multicolumn{4}{c|}{\textbf{Rec. XYZ $\rightarrow$ sRGB}} & \multicolumn{4}{c|}{\textbf{GT XYZ $\rightarrow$ sRGB}}\\ \cline{2-13}
\multicolumn{1}{|c|}{\multirow{-2}{*}{\textbf{Polynomial terms}}} & \textbf{Avg.} & \textbf{Q1} & \textbf{Q2} & \textbf{Q3} & \textbf{Avg.} & \textbf{Q1} & \textbf{Q2} & \textbf{Q3} & \textbf{Avg.} & \textbf{Q1} & \textbf{Q2} & \textbf{Q3}  \\ \hline
			
Linear $[R, G, B]$ &  26.85 & 22.07 & 26.72 & 31.58 & \cellcolor[HTML]{\bestcolor}\textbf{44.23} & {41.06} & \cellcolor[HTML]{\bestcolor}\textbf{44.53} & \cellcolor[HTML]{\bestcolor}\textbf{47.72} & 23.06 & 19.49 & 22.97 & 26.18 \\ \hline
			
$[R, G, B, R^2, G^2, B^2]$ (our choice) & \cellcolor[HTML]{\bestcolor}\textbf{29.66} & \cellcolor[HTML]{\bestcolor}\textbf{23.77} & \cellcolor[HTML]{\bestcolor}\textbf{29.57} & \cellcolor[HTML]{\bestcolor}\textbf{34.71} & 43.82 & \cellcolor[HTML]{\bestcolor}\textbf{41.43} & 43.94 & 46.58 & \cellcolor[HTML]{\bestcolor}\textbf{27.44} & \cellcolor[HTML]{\bestcolor}\textbf{23.57} & \cellcolor[HTML]{\bestcolor}\textbf{28.32} & \cellcolor[HTML]{\bestcolor}\textbf{30.88} \\ \hline
			
$[R, G, B, R^2, G^2, B^2, RG, RB, GB]$ & 27.89 & 23.47 & 27.04 & 33.28 & 39.64 & 38.18 & 41.27 & 43.47 & 25.04 & 21.32 & 25.43 & 30.01
\\ \hline
\end{tabular}
}
\end{table}

\begin{figure*}[!t]
\includegraphics[width=\linewidth]{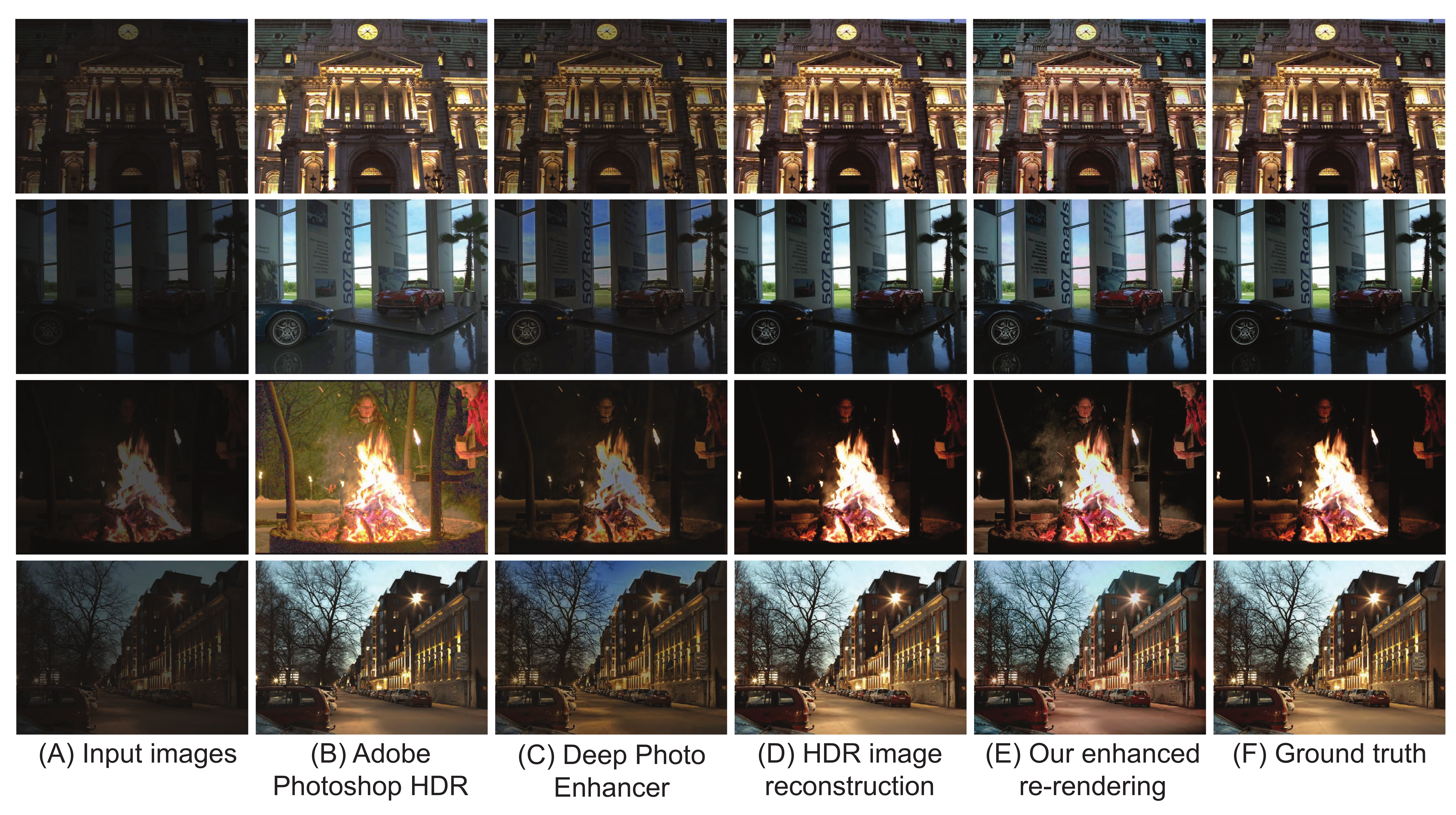}
\vspace{-7mm}
\caption{(A) The input sRGB rendered image. (B) Adobe Photoshop HDR results. (C) Deep Photo Enhancer results \cite{DPE}. (D) HDR result of ~\cite{HDRCNN}. (E) Our re-rendered images after photo-finishing enhancement. (F) Ground-truth images. Input images are taken from~\cite{HDRCNN}.}
\label{xyz:fig:HDR}
\end{figure*}

\begin{figure}[!t]
\includegraphics[width=\linewidth]{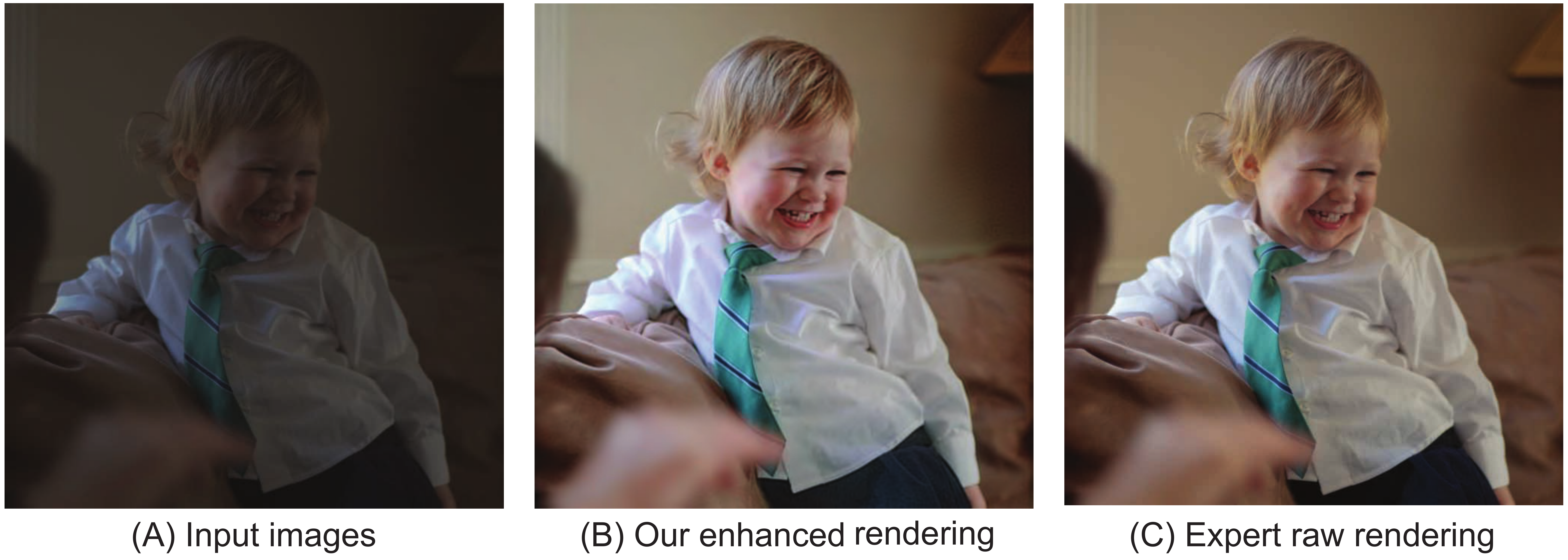}
\vspace{-7mm}
\caption[Example from the under-exposure testing set \cite{DeepUPE}.]{Example from the under-exposure testing set \cite{DeepUPE}. (A) Input image. (B) Our enhanced rendered image. (C) Expert-retouched image. \label{xyz:fig:low-light}}%
\end{figure}

\section{Comparison with U-Net Baseline} \label{sec:comparisonWithUnet}
We compare our proposed network against a U-Net-based baseline. This baseline consists of two U-Net-like \cite{ronneberger2015u} models trained in an end-to-end manner using the same training settings used to train our network (i.e., epochs, training patches, and loss function). Each U-Net model consists of a 3-level encoder/decoder with skip connections. The output channels of the first conv layer in the encoder unit has 28 channels. The two U-Net models have a total of 2,949,246 learnable parameters, compared to 2,697,578 learnable parameters in our network, and they were trained to map from sRGB to XYZ and from XYZ back to sRGB, similar to our model.

Table \ref{Table:unetcomparison} shows the results obtained by the U-Net baseline and our network on our testing set. 

\begin{table}[!t]
\caption[Comparison between our network and two U-Net models trained in an end-to-end manner to map from sRGB to CIE XYZ and back.]{Comparison between our network and two U-Net models trained in an end-to-end manner to map from sRGB to CIE XYZ and back. Both networks, ours and the two U-Net models, have approximately the same number of learnable parameters and both were trained using the same training settings. The best PSNR (dB) values are shown in boldface and highlighted in yellow.\label{Table:unetcomparison}}
\centering
\scalebox{0.7}{
\begin{tabular}{|l|c|c|c|c|c|c|c|c|}
\hline
\multicolumn{1}{|c|}{} & \multicolumn{4}{c|}{\textbf{sRGB $\rightarrow$ XYZ}} & \multicolumn{4}{c|}{\textbf{Rec. XYZ $\rightarrow$ sRGB}} \\ \cline{2-9}
\multicolumn{1}{|c|}{\multirow{-2}{*}{\textbf{Method}}} & \textbf{Avg.} & \textbf{Q1} & \textbf{Q2} & \textbf{Q3} & \textbf{Avg.} & \textbf{Q1} & \textbf{Q2} & \textbf{Q3}  \\ \hline

U-Net \cite{ronneberger2015u} & 20.05 & 16.84 & 19.76 & 22.78 &  43.39 & 40.56 & 43.40 & 45.91 \\ \hline

Ours & \textbf{29.66} & \cellcolor[HTML]{\bestcolor}\textbf{23.77} & \cellcolor[HTML]{\bestcolor}\textbf{29.57} & \cellcolor[HTML]{\bestcolor}\textbf{34.71} & \cellcolor[HTML]{\bestcolor}\textbf{43.82} & \cellcolor[HTML]{\bestcolor}\textbf{41.43} & \cellcolor[HTML]{\bestcolor}\textbf{43.94} & \cellcolor[HTML]{\bestcolor}\textbf{46.58}  \\ \hline

\end{tabular}
}
\end{table}

\begin{figure}[!t]
\includegraphics[width=\linewidth]{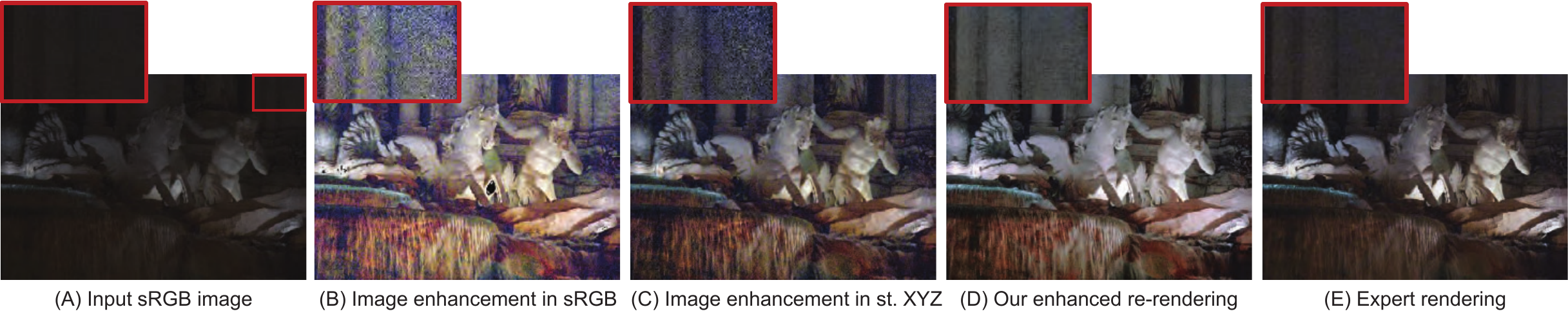}
\vspace{-7mm}
\caption[(A) The input image. (B) Image enhancement in sRGB. (C) Image enhancement in standard XYZ reconstruction. (D) Our enhanced re-rendering. (E) Expert enhancement.]{(A) The input image. (B) Image enhancement in sRGB. (C) Image enhancement in standard XYZ reconstruction. (D) Our enhanced re-rendering. (E) Expert enhancement. The enhancement is based on fusion of ``multi-exposed'' images \cite{mertens2009exposure} and local details enhancement \cite{paris2011local}. The image is from the under-exposure testing set \cite{DeepUPE}.}
\label{xyz:fig:supp_ours_vs_working_in_srgb}
\end{figure}

\begin{figure}[!t]
\includegraphics[width=\linewidth]{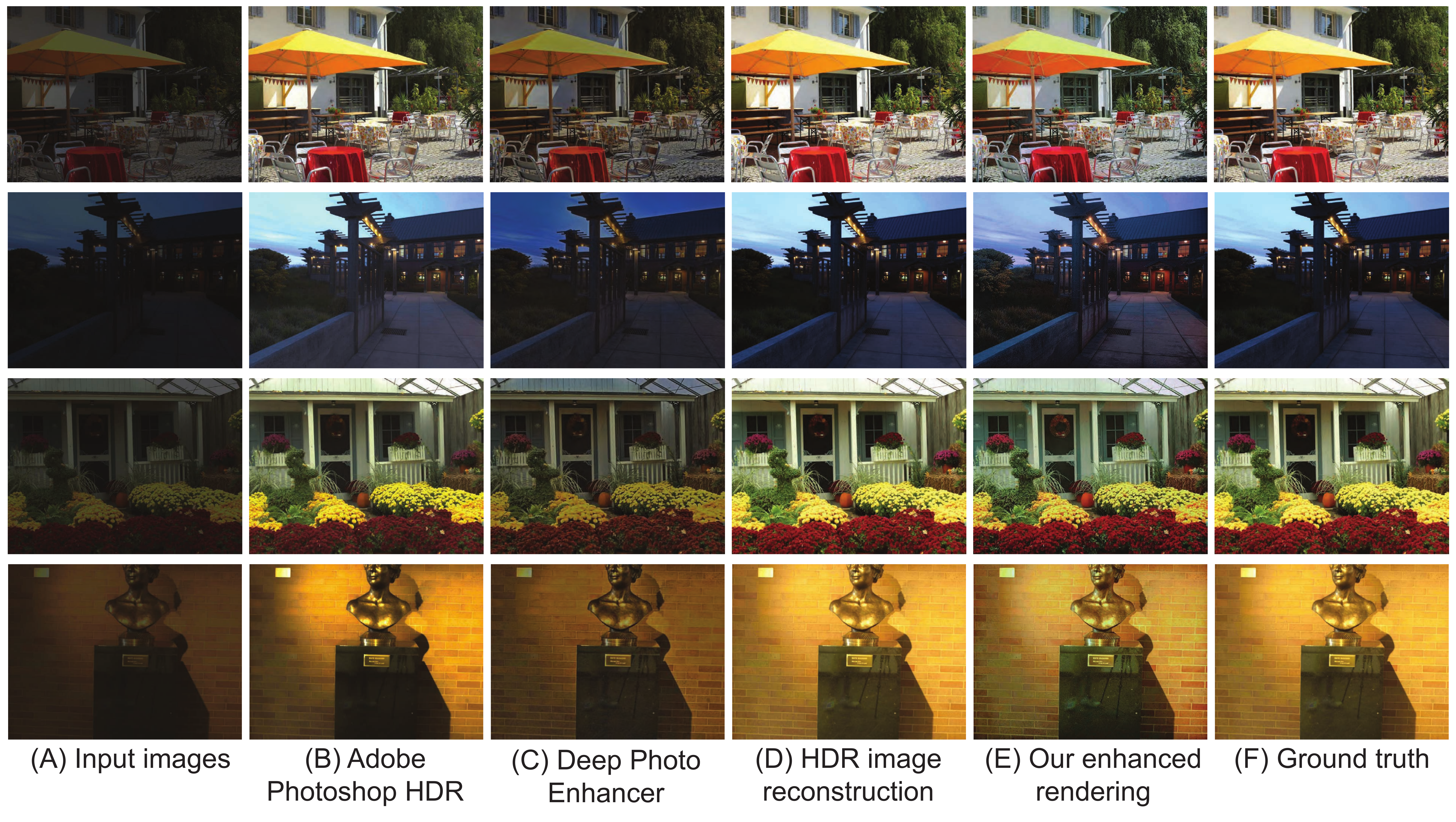}
\vspace{-7mm}
\caption[\hspace{0.5mm} Low-light image enhancement application.]{Low-light image enhancement application. (A) Input sRGB rendered image. (B) Adobe Photoshop HDR results. (C) Deep Photo Enhancer results \cite{DPE}. (D) HDR result of ~\cite{HDRCNN}. (E) Our re-rendered images after photo-finishing enhancement. (F) Ground truth images. Input images are taken from~\cite{HDRCNN}.}
\label{fig:supp_hdr}
\end{figure}

\begin{figure}[!t]
\includegraphics[width=\linewidth]{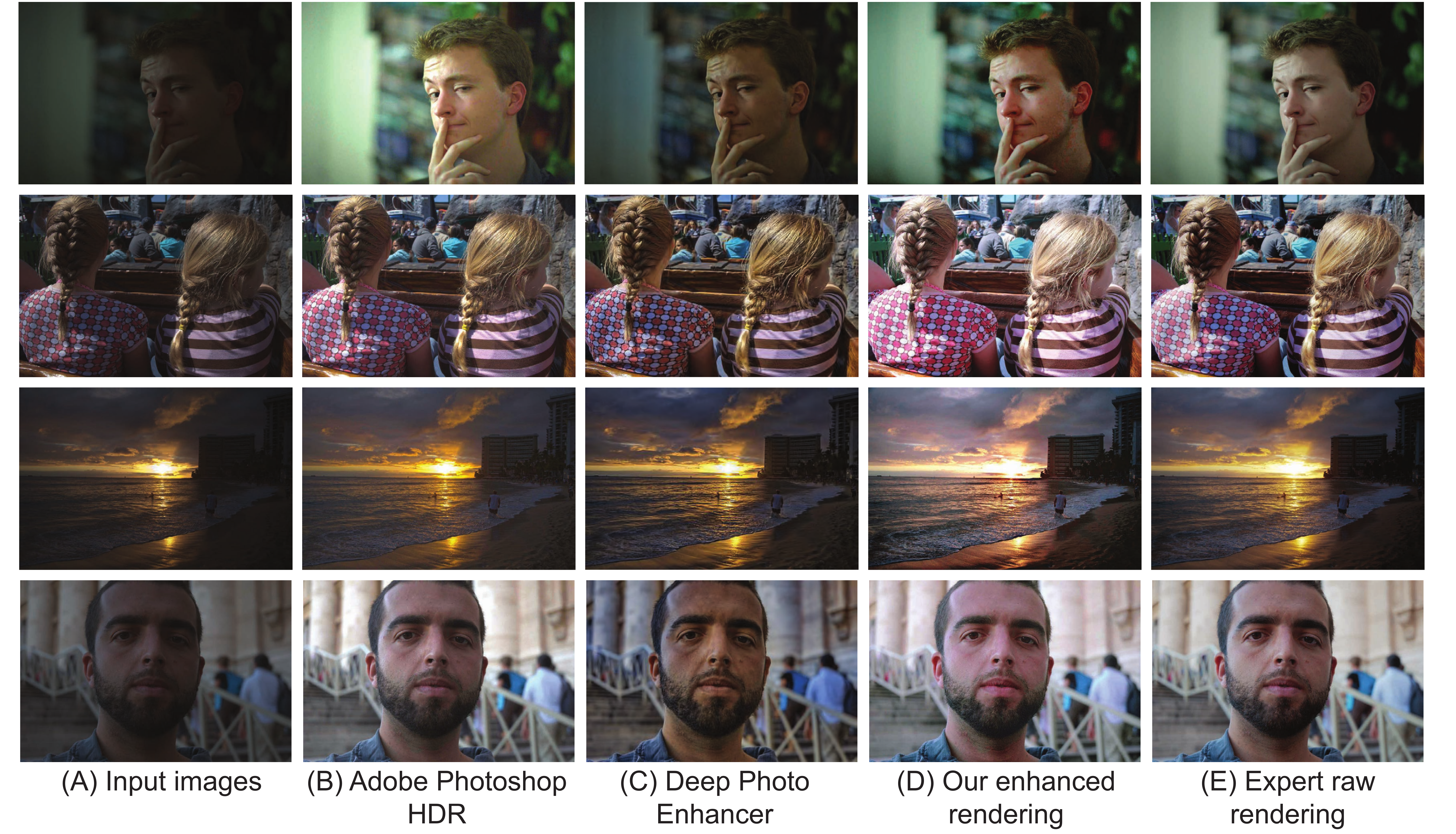}
\vspace{-7mm}
\caption[\hspace{0.5mm} Qualitative comparison for low-light image enhancement task.]{Qualitative comparison for low-light image enhancement task. Images are taken from the under-exposure testing set \cite{DeepUPE}. (A) Input image. (B) Adobe Photoshop HDR results. (C) Results of deep photo enhancer \cite{DPE}. (D) Our enhanced rendered image. (E) Expert-retouched image.}
\label{fig:supp_underexp}
\end{figure}

\begin{table}[!t]
\caption[Quantitative results of the photo-finishing enhancement application using 500 under-exposure images provided in \cite{DeepUPE}.]{Quantitative results of the photo-finishing enhancement application using 500 under-exposure images provided in \cite{DeepUPE}.\label{xyz:table:low-light}}
\centering
\scalebox{0.7}{
\begin{tabular}{|c|c|} \hline
\textbf{Method} & \textbf{PSNR} \\ \hline
White-Box \cite{hu2018exposure}& 18.57 \\
Distort-and-Recover \cite{park2018distort} & 20.97 \\
HDRNet \cite{HDRNET}& 21.96 \\
Deep photo enhancer \cite{DPE}& 22.150\\
DeepUPE \cite{DeepUPE}& 23.04 \\ \hline
Enhanced in sRGB & 16.92  \\
Enhanced in rec. standard XYZ & 18.41 \\\hdashline
Our enhanced re-rendering & 21.03 \\ \hline
\end{tabular}}
\end{table}

\subsection{Low-Light Image Enhancement}\label{xyz:sec:photo-finish}
Many photographers prefer to edit photographs in the linear raw-RGB sensor space rather than the nonlinear 8-bit sRGB space, due to the fact that raw-RGB images provide higher tonal values compared to sRGB camera-rendered images \cite{schewe2015digital}. Similar to the raw-RGB space, the CIE XYZ space is linear scene-referred with higher tonal values compared to the final sRGB space. Thus, we can also benefit from our linear CIE XYZ space for image enhancement tasks.

In this set of experiments, we present a set of simple operations that can achieve results on par with recent methods designed for low-light image enhancement. Specifically,  we apply the following set of heuristic operations to perform low-light image enhancement. As our reconstructed XYZ image has a wider range of tonal values (see Fig. \ref{xyz:fig:tonal_values}), we apply a set of synthetic digital gains to simulate multi-exposure settings. This simulation does not introduce any new information that did not exist in the original image; however, it allows us to better explore the range of tonal values provided in our reconstructed image---we can think of this operation as an ISO gain that is applied on board cameras to amplify the captured image signal. To that end, we multiply the reconstructed image by four different factors. These factors can be tuned in an interactive manner based on each image, but we preferred to fix these hyperparameters over all experiments. In particular, we multiplied our reconstructed XYZ image by (0.1, 1.4, 2.7, 4.0) to generate four different versions of our reconstructed XYZ image. Following this, we apply an off-the-shelf exposure-fusion algorithm~\cite{mertens2009exposure} to create the modified XYZ layer. To enhance the local details, we apply a local details enhancement method~\cite{paris2011local} on our forward local sRGB reconstructed layer. Figure \ref{xyz:fig:HDR} shows examples of our results. As can be seen, we achieve on par results with state-of-the-art methods designed specifically for the given image enhancement task.

We further evaluated this simple pipeline on 500 under-exposed images taken from \cite{DeepUPE}. Figure \ref{xyz:fig:low-light} shows a qualitative example. We show a quantitative comparison in Table \ref{xyz:table:low-light}. 
Applying digital gain to our reconstructed space provides better results compared to using the standard XYZ reconstruction or the nonlinear sRGB space. This is due to the fact that our reconstructed images have a better linearization with a high tonal range; see Fig. \ref{xyz:fig:supp_ours_vs_working_in_srgb}. We provide additional results in Figs. \ref{fig:supp_hdr} and \ref{fig:supp_underexp}.

\section{Summary}\label{xyz:sec:conclusion}

We have proposed a method and DNN model that can map back and forth between non-linear sRGB and linear CIE XYZ images more accurately compared to alternative approaches. Our method is based on learning a decomposition of sRGB images into a globally processed and locally processed image layers. The learned globally processed image layer is then used to learn a mapping to the device independent CIE XYZ color space.  By utilizing the decomposed image layers produced by our method, we show that our model can be used to perform low-light image enhancement.

\chapter{Correcting Colors in Exposure Errors \label{ch:ch13}}
The previous chapter presented a linearization method that could be used to enhance low-light and under-exposed image colors. As discussed earlier, exposure problems are categorized as either: (i) overexposed, where the camera exposure was too long, resulting in bright and washed-out image regions, or (ii) underexposed, where the exposure was too short, resulting in dark regions. \textit{Both} under- and overexposure greatly reduce the contrast and visual appeal of an image. Prior work mainly focuses on underexposed images or general image enhancement. In contrast, this chapter presents a method targeting both over- and underexposure errors in photographs\footnote{Work done while the author was an intern at Samsung AI Center -- Toronto; this work is under review. This work was published in \cite{afifi2020learning, afifi2021network}: Mahmoud Afifi, Konstantinos G Derpanis, Bj\"{o}rn Ommer, and Michael S. Brown. Learning Multi-Scale Photo Exposure Correction. In IEEE Conference on Computer Vision and Pattern Recognition (CVPR), 2021.}. We formulate the exposure correction problem as two main sub-problems: (i) \textit{color enhancement} and (ii) \textit{detail enhancement}. Accordingly, we propose a coarse-to-fine DNN model, trainable in an end-to-end manner, that addresses each sub-problem separately. A key aspect of our solution is a new dataset of over 24,000 images exhibiting the broadest range of exposure values to date with a corresponding properly exposed image. Our method achieves results on par with existing state-of-the-art methods on underexposed images and yields significant improvements for images suffering from overexposure errors. The source code and dataset of this work are available on GitHub: \href{https://github.com/mahmoudnafifi/Exposure_Correction}{https://github.com/mahmoudnafifi/Exposure$\_$Correction}.

\section{Introduction}
\label{exposure:sec:intro}

As discussed in Chapter \ref{ch:ch3}, digital cameras adjust cature exposure to control the overall brightness levels in the image. This adjustment can be controlled manually or performed automatically in the AE mode, where cameras uses TTL metering that measures the amount of light received from the scene to adjust the EV to compensate for high/low level of brightness in the captured scene  \cite{peterson2016understanding}. 

We have also showed that in Chapter \ref{ch:ch3} prior work mainly focuses on underexposed images or general image enhancement. In contrast to the majority of prior work, our work is the first deep learning method to explicitly correct {\it both} overexposed and underexposed photographs with a single model.

Our method is enabled by generating a large dataset of images with exposure errors. Unlike existing datasets for exposure correction, our dataset is rendered with a wide range of exposure errors to cover both cases of exposure errors---i.e., over- and under-exposure errors. Figure \ref{exposure:fig:dataset} shows a comparison between our dataset and the LOL dataset in terms of the number of images and the variety of exposure errors in each dataset. The LOL dataset covers a relatively small fraction of the possible exposure levels, as compared to our introduced dataset.  Our dataset is based on the MIT-Adobe FiveK dataset~\cite{bychkovsky2011learning} and is accurately rendered by adjusting the high tonal values provided in camera sensor raw-RGB images to realistically emulate camera exposure errors.

\begin{figure}[!t]
\includegraphics[width=\linewidth]{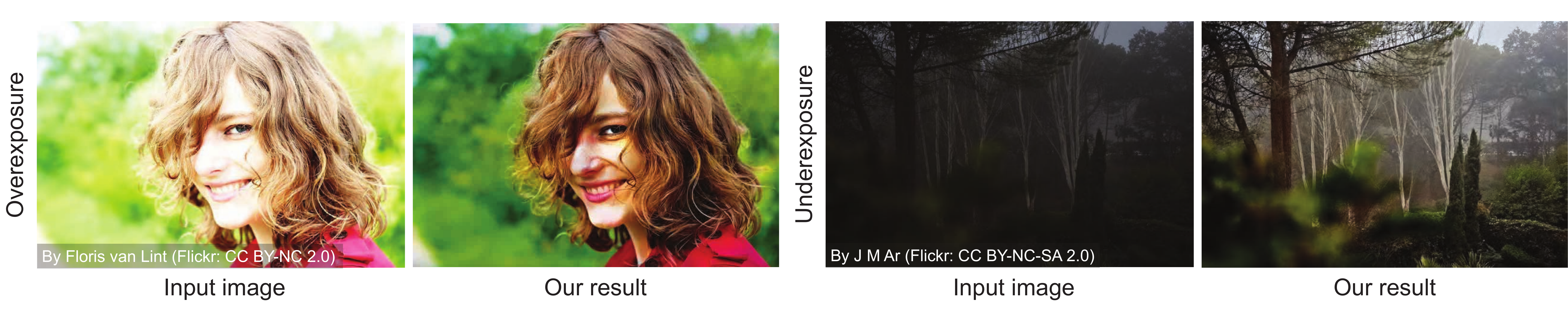}
\vspace{-7mm}
\caption[Photographs with over- and underexposure errors and the results of our method using a \textit{single} model for exposure correction.]{Photographs with over- and underexposure errors and the results of our method using a \textit{single} model for exposure correction. These sample input images are taken from outside our dataset to demonstrate the generalization of our trained model.}
\label{exposure:fig:teaser}
\end{figure}

\paragraph{Contributions}~We propose a coarse-to-fine deep learning method for exposure error correction of \textit{both} over- and underexposed sRGB images. Our approach formulates the exposure correction problem as two main sub-problems: (i) color and (ii) detail enhancement. We propose a coarse-to-fine DNN model, trainable in an end-to-end manner, that
begins by correcting the global color information and subsequently refines the image details.
In addition to our DNN model, a key contribution to the exposure correction problem is a new dataset containing over 24,000 images rendered from raw-RGB to sRGB with different exposure settings with broader exposure ranges than previous datasets. Each image in our dataset is provided with a corresponding properly exposed reference image. Lastly, we present an extensive set of evaluations and ablations of our proposed method with comparisons to the state of the art.  We demonstrate that our method achieves results on par with previous methods dedicated to underexposed images and yields significant improvements on overexposed images. Furthermore, our model generalizes well to images outside our dataset.

\begin{figure}[!t]
\includegraphics[width=\linewidth]{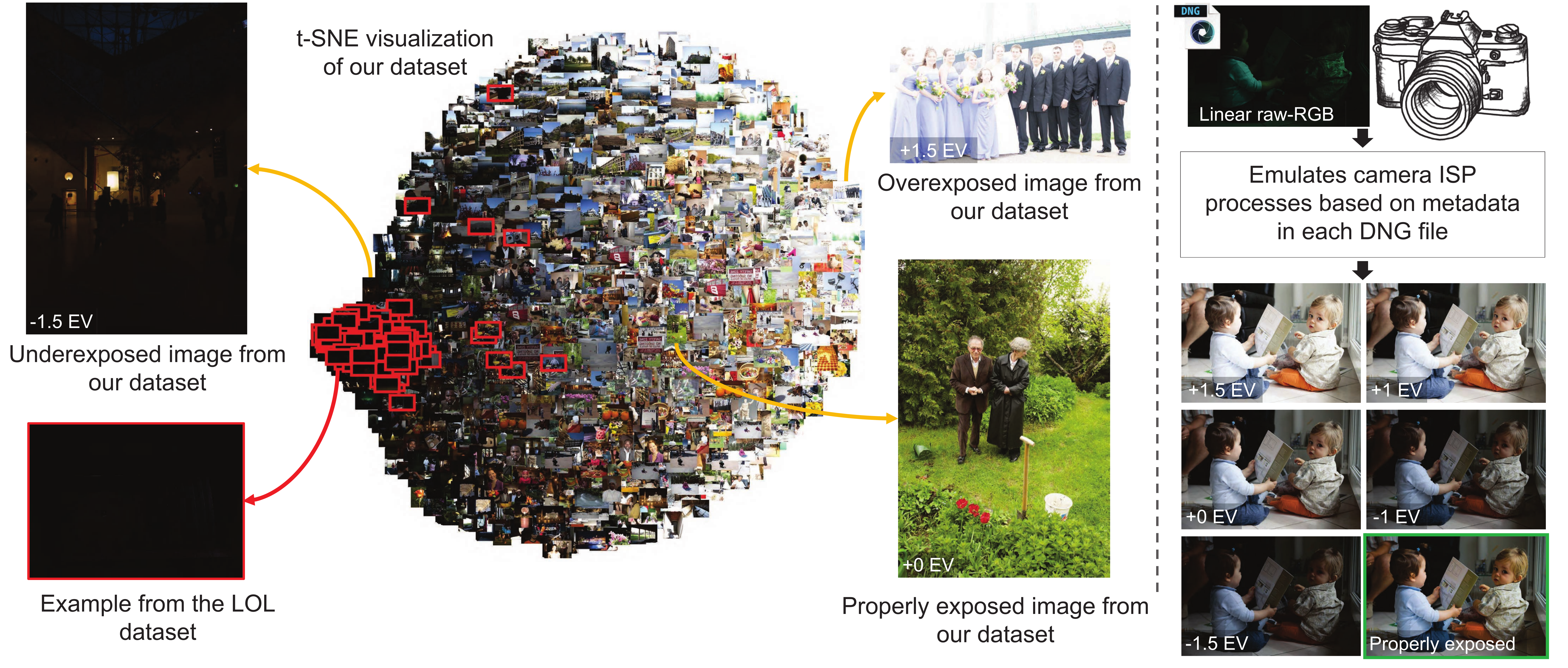}
\vspace{-7mm}
\caption[Our dataset contains images with different exposure error types and their corresponding properly exposed reference images.]{Our dataset contains images with different exposure error types and their corresponding properly exposed reference images. Shown is a t-SNE visualization \cite{maaten2008visualizing} of all images in our dataset and the low-light (LOL) paired dataset (outlined in red) \cite{Chen2018Retinex}.  Notice that LOL covers a relatively small fraction of the possible exposure levels, as compared
to our introduced dataset.
Our dataset was rendered from linear raw-RGB images taken from the MIT-Adobe FiveK dataset \cite{bychkovsky2011learning}. Each image was rendered with different relative exposure values (EVs) by an accurate emulation of the camera ISP processes.}
\label{exposure:fig:dataset}
\end{figure}

\begin{figure}[!t]
\begin{center}
\includegraphics[width=0.95\linewidth]{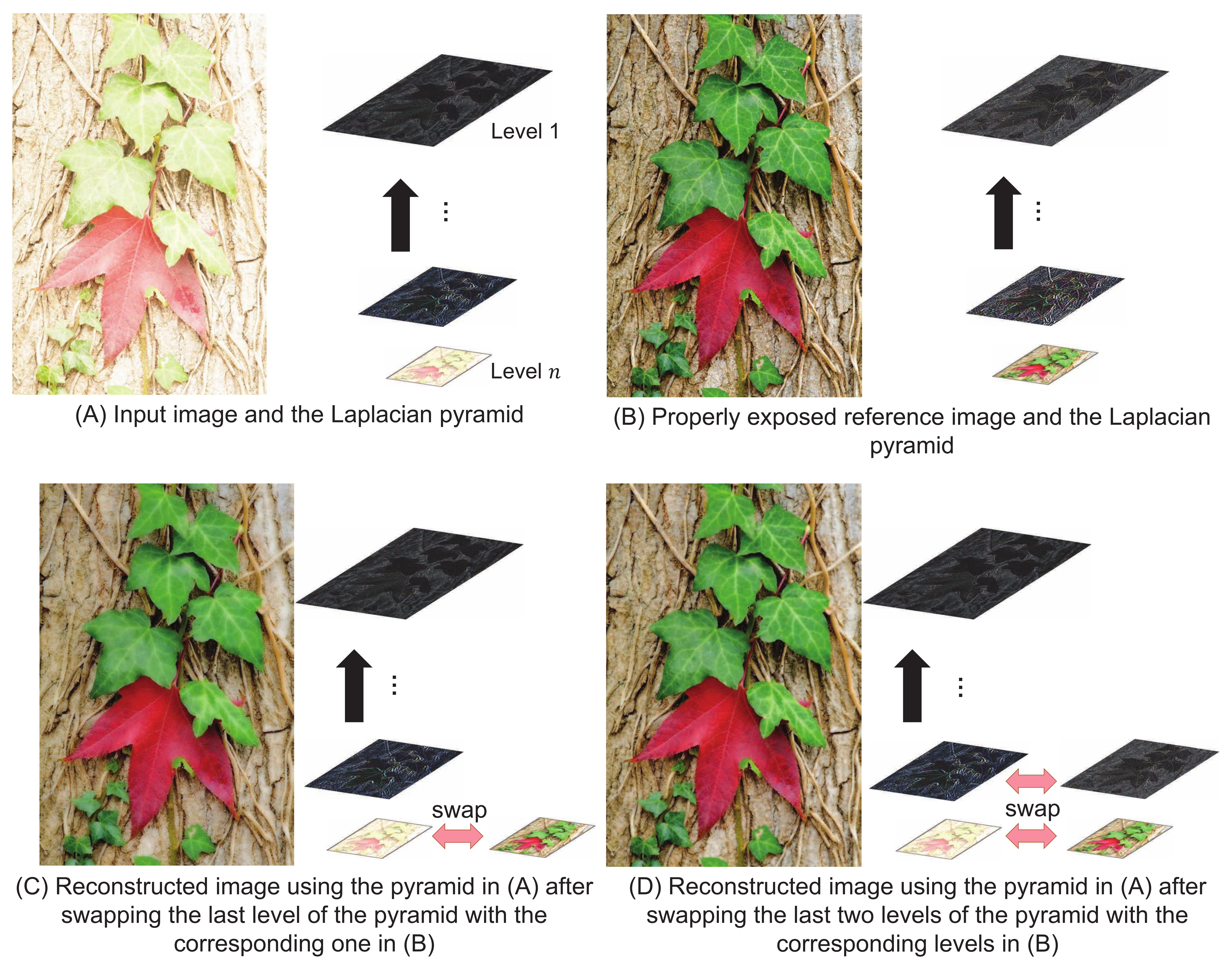}
\end{center}
\vspace{-7mm}
\caption[Motivation behind our coarse-to-fine exposure correction approach.]{Motivation behind our coarse-to-fine exposure correction approach. Example of an overexposed image and its corresponding properly exposed image shown in (A) and (B), respectively. The Laplacian pyramid decomposition allows us to enhance the color and detail information sequentially, as shown in (C) and (D), respectively.}
\label{exposure:fig:idea}
\end{figure}

\section{Our Dataset} \label{exposure:subsec:data}

To train our model, we need a large number of training images rendered with realistic over-  and underexposure errors and corresponding properly exposed ground truth images.
As discussed in Chapter \ref{ch:ch3}, such datasets are currently not publicly available to support exposure correction research. For this reason, our first task is to create a new dataset.  As done in Chapter \ref{ch:ch12}, our dataset is rendered from the MIT-Adobe FiveK dataset~\cite{bychkovsky2011learning}, which has 5,000 raw-RGB images and corresponding sRGB images rendered manually by five expert photographers~\cite{bychkovsky2011learning}.

For each raw-RGB image, we use the Adobe Camera Raw SDK~\cite{CameraRaw} to emulate different EVs as would be applied by a camera~\cite{schewe2010real}. We render each raw-RGB image with the AWB settings and with different digital EVs to mimic real exposure errors. Specifically, we use the  relative EVs $-1.5$, $-1$, $+0$, $+1$, and $+1.5$ to render images with underexposure errors, a zero gain of the original EV, and overexposure errors, respectively. The zero-gain relative EV is equivalent to the original exposure settings applied onboard the camera during capture time.

As the ground truth images, we use images that were manually retouched by an expert photographer (referred to as Expert C in \cite{bychkovsky2011learning}) as our target correctly exposed images, rather than using our rendered images with $+0$ relative EV. The reason behind this choice is that a significant number of images contain backlighting or partial exposure errors in the original exposure capture settings. The expert adjusted images were performed in ProPhoto RGB color space \cite{bychkovsky2011learning} (rather than raw-RGB), which we converted to a standard 8-bit sRGB color space encoding.

In total, our dataset contains 24,330 8-bit sRGB images with different digital exposure settings. We discarded a small number of images that had misalignment with their corresponding ground truth image. These misalignments are due to different usage of the DNG crop area metadata by Adobe Camera Raw SDK and the expert. Our dataset is divided into three sets: (i) training set of 17,675 images, (ii) validation set of 750 images, and (iii) testing set of 5,905 images.   The training, validation, and testing sets, use different images taken from the FiveK dataset. This means the training, validation, and testing images do not share any images in common. Figure \ref{exposure:fig:dataset} shows examples of our generated 8-bit sRGB images and the corresponding properly exposed 8-bit sRGB reference images. We acknowledge that digital exposure used to produce our dataset does not consider the noise characteristics that could change based on the camera exposure settings.

\begin{figure*}[!t]
\includegraphics[width=\linewidth]{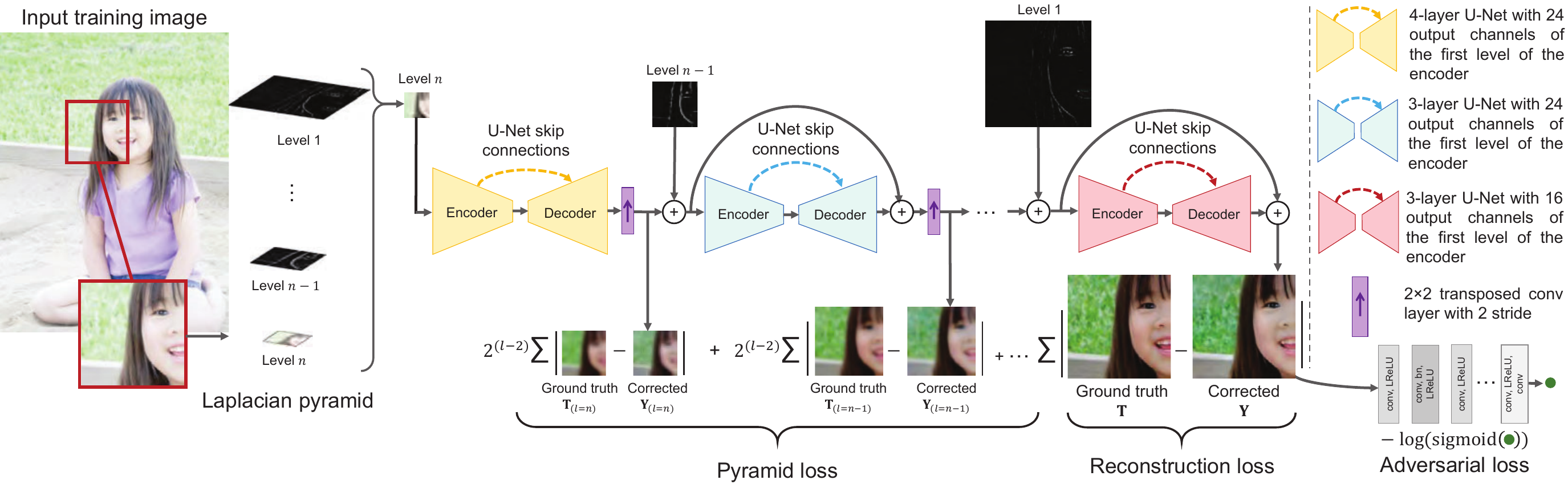}
\vspace{-7mm}
\caption{Overview of our image exposure correction architecture. We propose a coarse-to-fine deep network to progressively correct exposure errors in 8-bit sRGB images.
Our network first corrects the global color captured at the final level of the Laplacian pyramid and then the subsequent frequency layers.}
\label{exposure:fig:main}
\end{figure*}

\section{Methodology} \label{exposure:sec:method}
Given an 8-bit sRGB input image, $\mat{I}$, rendered with the incorrect exposure setting, our method aims to produce an output image, $\mat{Y}$, with fewer exposure errors than those in $\mat{I}$. As we simultaneously target both over- and underexposed errors, our input image, $\mat{I}$, is expected to contain regions of nearly over- or under-saturated values with corrupted color and detail information.
We propose to correct color and detail errors of $\mat{I}$ in a sequential manner. Specifically, we process a multi-resolution representation of $\mat{I}$, rather than directly dealing with the original form of $\mat{I}$. We use the Laplacian pyramid \cite{burt1983laplacian} as our multi-resolution decomposition, which is derived  from the Gaussian pyramid of $\mat{I}$.

\subsection{Coarse-to-Fine Exposure Correction}

Let $\mat{X}$ represent the Laplacian pyramid of $\mat{I}$ with $n$ levels, such that  $\mat{X}_{(l)}$ is the $l^\texttt{th}$ level of $\mat{X}$. The last level of this pyramid (i.e., $\mat{X}_{(n)}$) captures low-frequency information of $\mat{I}$, while the first level (i.e., $\mat{X}_{(1)}$) captures the high-frequency information. Such frequency levels can be categorized into: (i) global color information of $\mat{I}$ stored in the low-frequency level and (ii) image coarse-to-fine details stored in the mid- and high-frequency levels. These levels can be later used to reconstruct the full-color image $\mat{I}$.

Figure \ref{exposure:fig:idea} motivates our coarse-to-fine approach to exposure correction.  Figures \ref{exposure:fig:idea}-(A) and (B) show an example overexposed image and its corresponding well-exposed target, respectively.
As observed, a significant exposure correction can be obtained by  using only the low-frequency layer (i.e., the global color information) of the target image in the Laplacian pyramid reconstruction process, as shown in Fig.~\ref{exposure:fig:idea}-(C). We can then improve the final image by enhancing the details in a sequential way by correcting each level of the Laplacian pyramid, as shown in Fig.~\ref{exposure:fig:idea}-(D). Practically, we do not have access to the properly exposed image in Fig.~\ref{exposure:fig:idea}-(B) at the inference stage, and thus our goal is to predict the missing color/detail information of each level in the Laplacian pyramid.

Inspired by this observation and the success of coarse-to-fine architectures for various other computer vision tasks (e.g., \cite{denton2015deep, shaham2019singan, lai2017deep, ma2020efficient}), we design a DNN that corrects the global color and detail information of $\mat{I}$ in a sequential manner using the Laplacian pyramid decomposition.
The remaining parts of this section explain the technical details of our model (Sec.\ \ref{exposure:subsec:network}), including details of the losses (Sec.\ \ref{exposure:subsec:losses}),
inference phase (Sec.\ \ref{exposure:subsec:Inference}), and training (Sec.\ \ref{exposure:subsec:training_details}).

\subsection{Coarse-to-Fine Network} \label{exposure:subsec:network}
Our image exposure correction architecture sequentially processes the $n$-level Laplacian pyramid, $\mat{X}$, of the input image, $\mat{I}$, to produce the final corrected image, $\mat{Y}$. The proposed model consists of $n$ sub-networks. Each of these sub-networks is a U-Net-like architecture \cite{ronneberger2015u} with untied weights. We allocate the network capacity in the form of weights based on how significantly each sub-problem (i.e., global color correction and detail enhancement) contributes to our final result.
Figure \ref{exposure:fig:main} provides an overview of our network. As shown, the largest (in terms of weights) sub-network in our architecture is dedicated to processing the global color information in $\mat{I}$ (i.e., $\mat{X}_{(n)}$). This sub-network (shown in yellow in Fig.\ \ref{exposure:fig:main}) processes the low-frequency level $\mat{X}_{(n)}$ and produces an upscaled image $\mat{Y}_{(n)}$. The upscaling process scales up the output of our sub-network by a factor of two using strided transposed convolution with trainable weights. Next, we add the first mid-frequency level $\mat{X}_{(n-1)}$ to $\mat{Y}_{(n)}$ to be processed by the second sub-network in our model. This sub-network enhances the corresponding details of the current level and produces a residual layer that is then added to $\mat{Y}_{(n)} + \mat{X}_{(n-1)}$ to reconstruct image $\mat{Y}_{(n-1)}$, which is equivalent to the corresponding Gaussian pyramid level $n-1$. This refinement-upsampling process proceeds until the final output image, $\mat{Y}$, is produced. Our network is fully differentiable and thus can be trained in an end-to-end manner.

\subsection{Losses}\label{exposure:subsec:losses}

We train our model end-to-end to minimize the following loss function:
\begin{equation}
\label{exposure:eq:final_loss}
\mathcal{L} = \mathcal{L}_{\text{rec}} + \mathcal{L}_{\text{pyr}} + \mathcal{L}_{\text{adv}},
\end{equation}
where  $\mathcal{L}_{\text{rec}}$ denotes the reconstruction loss, $\mathcal{L}_{\text{pyr}}$ the pyramid loss, and $\mathcal{L}_{\text{adv}}$ the adversarial loss. The individual losses are defined next.

\paragraph{Reconstruction Loss:}
We use the $\texttt{L}_1$ loss function between the reconstructed and properly exposed reference images. This loss can be expressed as follows:
\begin{equation}
\label{exposure:eq:wo_pyramid_loss}
\mathcal{L}_{\text{rec}} = \sum_{p=1}^{3hw} \left|\mat{Y}(p) - \mat{T}(p)\right|,
\end{equation}
where $h$ and $w$ denote the height and width of the training image, respectively, and
$p$ is the index of each pixel in our corrected image, $\mat{Y}$, and the
corresponding properly exposed reference image, $\mat{T}$, respectively.

\begin{figure}[!t]
\includegraphics[width=\linewidth]{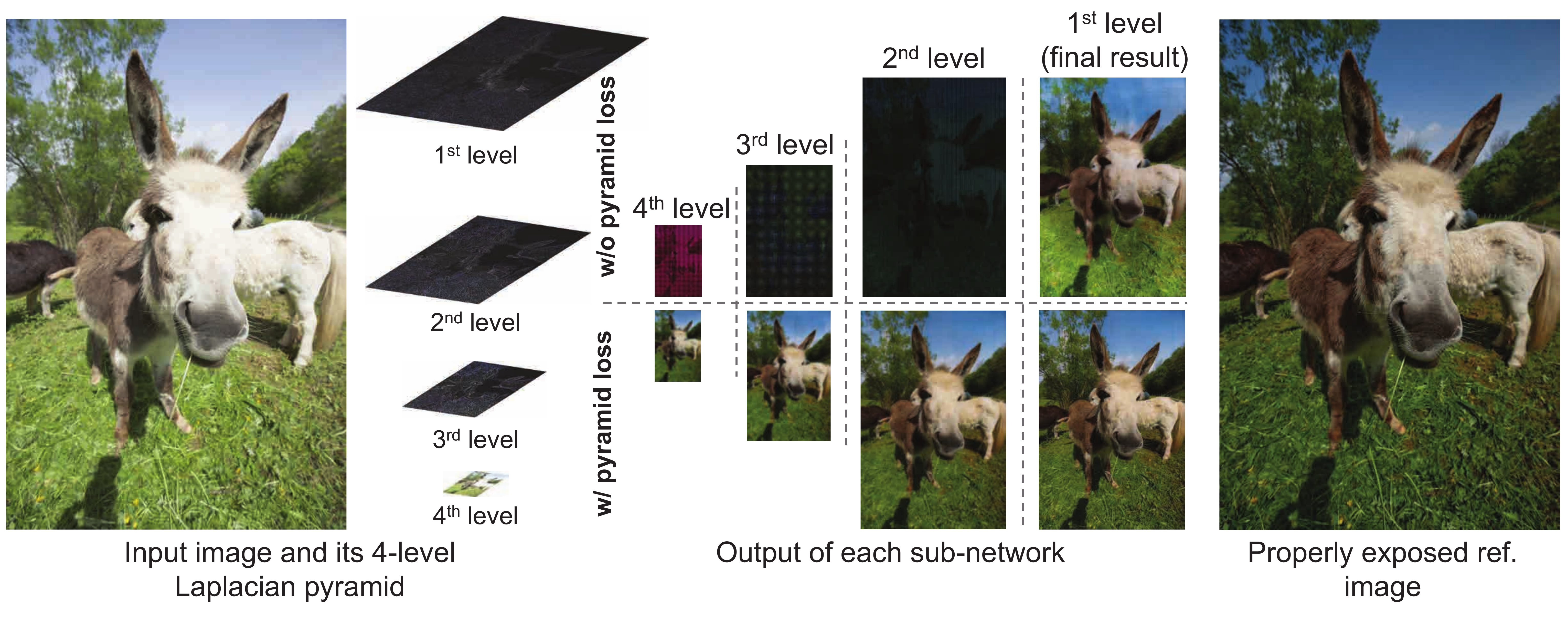}
\vspace{-7mm}
\caption[Multiscale losses. Shown are the output of each sub-net trained with and without the pyramid loss (Eq.\ \ref{exposure:eq:w_pyramid_loss}).]{Multiscale losses. Shown are the output of each sub-net trained with and without the pyramid loss (Eq.\ \ref{exposure:eq:w_pyramid_loss}).\label{exposure:fig:how_it_works}}
\end{figure}

\paragraph{Pyramid Loss:}
To guide each sub-network to follow the Laplacian pyramid reconstruction procedure, we introduce dedicated losses at each pyramid level.  Let $\mat{T}_{(l)}$ denote the  $l^{\texttt{th}}$ level of the Gaussian pyramid of our reference image, $\mat{T}$, after upsampling by a factor of two. We use the standard pyramid upsampling operation \cite{burt1983laplacian}. Our pyramid loss is computed as follows:
\begin{equation}
\label{exposure:eq:w_pyramid_loss}
\mathcal{L}_{\text{pyr}} = \sum_{l=2}^{n}2^{(l-2)}\sum_{p=1}^{3h_lw_l} \left|\mat{Y}_{(l)}(p) - \mat{T}_{(l)}(p)\right|,
\end{equation}
where $h_l$ and $w_l$ are twice the height and width of the $l^{\texttt{th}}$ level in the Laplacian pyramid of the training image, respectively, and
$p$ is the index of each pixel in our corrected image at the $l^{\texttt{th}}$ level $\mat{Y}_{(l)}$ and the
properly exposed reference image at the same level $\mat{T}_{(l)}$, respectively. The pyramid loss not only gives a principled interpretation of the task of each sub-network but also results in less visual artifacts compared to training using only the reconstruction loss, as can be seen in Fig.~\ref{exposure:fig:how_it_works}. Notice that without the intermediate pyramid losses, the multi-scale reconstructions, shown in Fig.\ \ref{exposure:fig:how_it_works} (right-top), deviate widely from the intermediate Gaussian targets compared to using the pyramid loss at each scale, as shown in Fig.\ \ref{exposure:fig:how_it_works} (right-bottom).

\paragraph{Adversarial Loss:} To perceptually enhance the reconstruction of the corrected image output in terms of realism and appeal, we also consider an adversarial loss as a regularizer.
This adversarial loss term can be described by the following equation \cite{goodfellow2014generative}:

\begin{equation}
\label{exposure:eq:w_adv_loss}
\mathcal{L}_{\text{adv}} = -3hwn \log\left(\mathcal{S}\left(\mathcal{D}\left(\mat{Y}\right)\right)\right),
\end{equation}
where $\mathcal{S}$ is the sigmoid function and $\mathcal{D}$ is a discriminator DNN that is trained together with our main network. 

\begin{figure}[t]
\includegraphics[width=\textwidth]{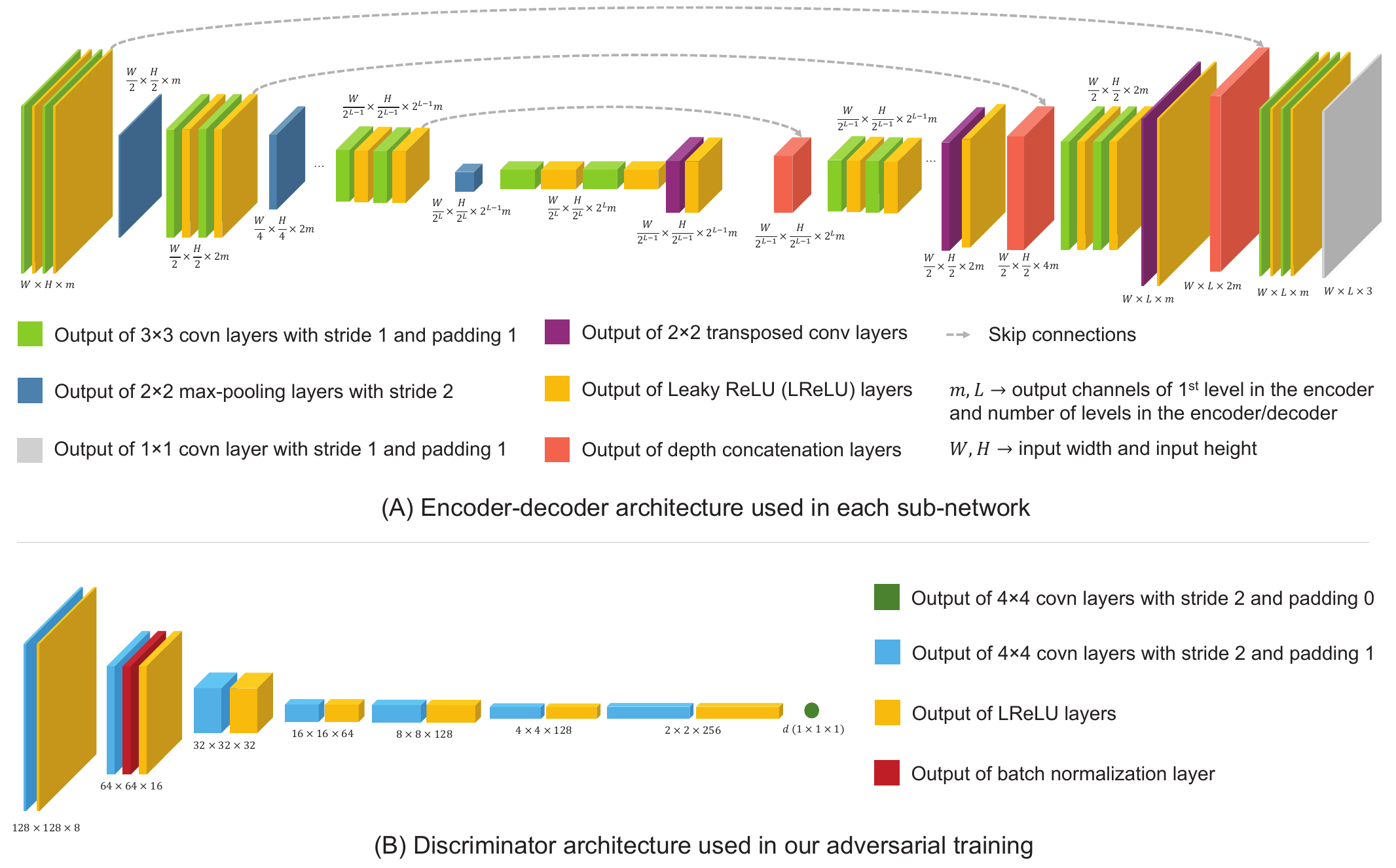}
\vspace{-7mm}
\caption[Details of the architectures used in our work.]{Details of the architectures used in our work. (A) Encoder-decoder architecture \cite{ronneberger2015u} used to design our sub-networks in the main network. (B) Discriminator architecture.\label{exposure:fig:arch}}
\end{figure}

\subsection{Network Details} \label{exposure:network_details}
Our main network consists of four sub-networks with $\sim$7M parameters trained in an end-to-end manner. The largest network capacity is dedicated to the first sub-network with decreasing amounts of capacity as we move from coarse-to-fine scales. Each sub-network accepts a different representation of the input image extracted from the Laplacian pyramid decomposition. The first sub-network is a four-layer encoder-decoder network with skip connections (i.e., U-Net-like architecture  \cite{ronneberger2015u}). The output of the first conv layer has 24 channels. Our first sub-network has $\sim$4.4M learnable parameters and accepts the low-frequency band level of the Laplacian pyramid, i.e., $\mat{X}_{(4)}$. The result of the first sub-network is then upscaled using a $2\!\times\!2\!\times\!3$ transposed conv layer with three output channels and a stride of two. This processed layer is then added to the first mid-frequency band level of the Laplacian pyramid (i.e., $\mat{X}_{(3)}$) and is fed to the second sub-network. 

The second sub-network is a three-layer encoder-decoder network with skip connections. It has 24 channels in the first conv layer of the encoder, with a total of $\sim$1.1M learnable parameters. The second sub-network processes the upscaled input from the first sub-network and outputs a residual layer, which is then added back to the input to the second sub-network followed by a $2\!\times\!2\!\times\!3$ transposed conv layer with three output channels and a stride of two. The result is added to the second mid-frequency band level of the Laplacian pyramid (i.e., $\mat{X}_{(2)}$) and is fed to the third sub-network, which generates a new residual that is added back again to the input of this sub-network. 

The third sub-network has the same design as the second network.  Finally, the result is added to the high-frequency band level of the Laplacian pyramid (i.e., $\mat{X}_{(1)}$) and is fed to the fourth sub-network to produce the final processed image.

The final sub-network is a three-layer encoder-decoder network with skip connections and has $\sim$482.2K learnable parameters, where the output of the first conv layer in its encoder has 16 channels. We provide the details of the main encoder-decoder architecture of each sub-network in Fig.\ \ref{exposure:fig:arch}-(A).

In the adversarial training of our network, we use a light-weight discriminator network with $\sim$1M learnable parameters. We provide the details of the discriminator in Fig.\ \ref{exposure:fig:arch}-(B). Notice that unlike our main network, we resize all input image patches to have $256\!\times\!256$ pixels before being processed by the discriminator. The output of the last layer in our discriminator is a single scalar value which is then used in our loss during the optimization.

\begin{figure}[t]
\includegraphics[width=\linewidth]{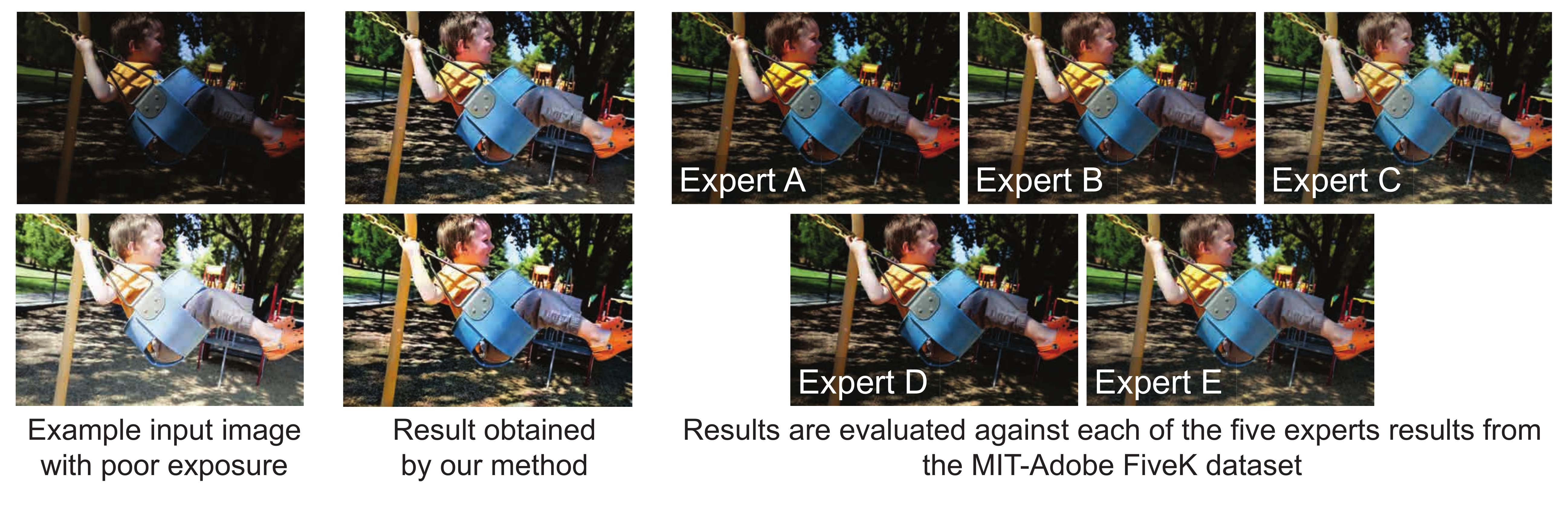}
\vspace{-7mm}
\caption[We evaluate the results of input images against all five expert photographers' edits from the FiveK dataset~\cite{bychkovsky2011learning}.]{We evaluate the results of input images against all five expert photographers' edits from the FiveK dataset~\cite{bychkovsky2011learning}.\label{exposure:fig:experts}}
\end{figure}

\begin{figure}[!t]
\centering
\includegraphics[width=0.95\linewidth]{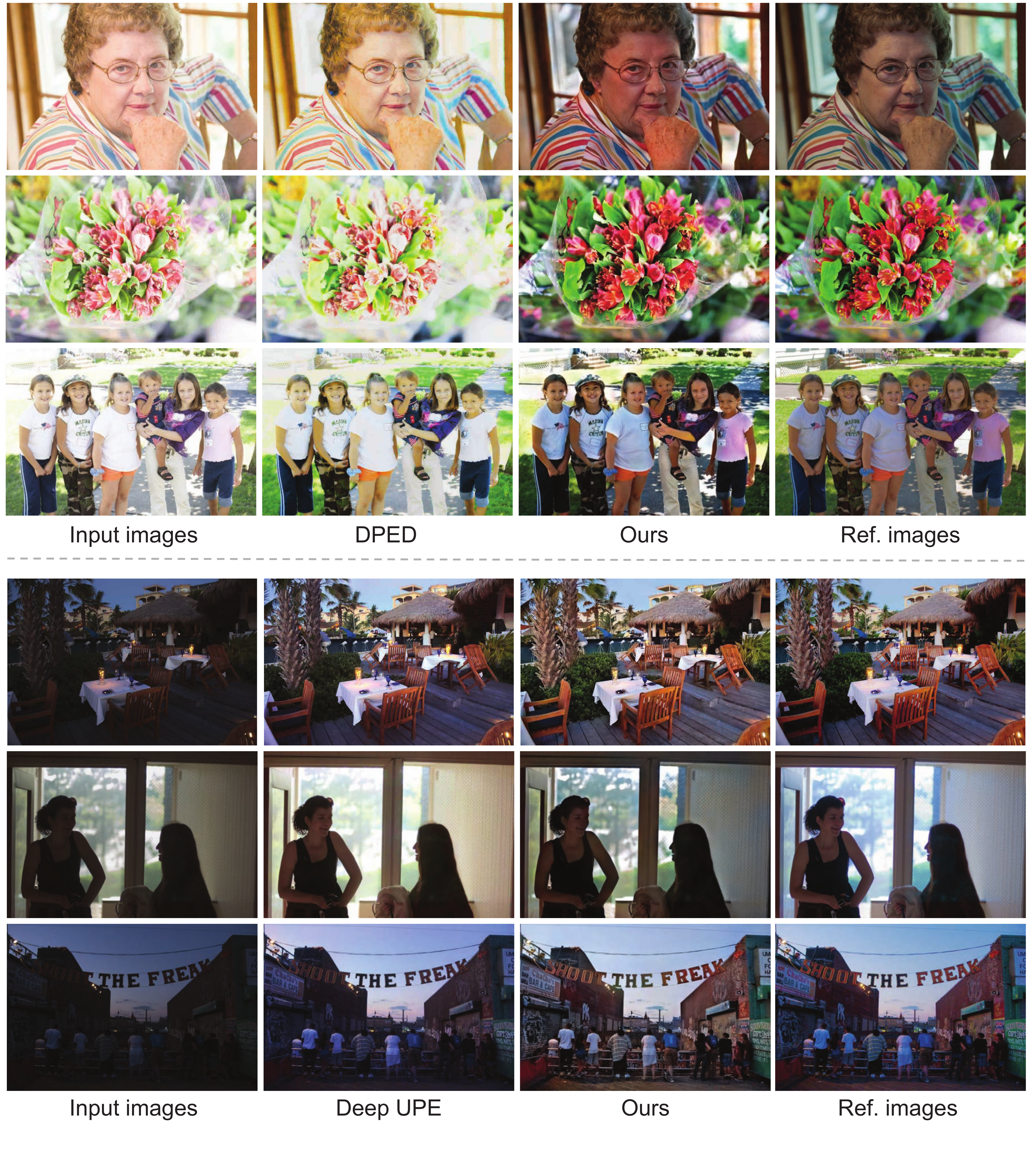}
\vspace{-7mm}
\caption[Qualitative results of correcting images with exposure errors.]{Qualitative results of correcting images with exposure errors. Shown are the input images from our test set, results from the DPED \cite{DPED}, results from the Deep UPE \cite{DPE}, our results, and the corresponding ground truth images.\label{exposure:fig:qualitative_our_set}}
\end{figure}

\subsection{Inference Stage}\label{exposure:subsec:Inference}

Our network is fully convolutional and can process input images with different resolutions. While our model requires a reasonable memory size ($\sim$7M parameters), processing high-resolution images requires a high computational power that may not always be available. Furthermore, processing images with considerably higher resolution (e.g., 16-megapixel) than the range of resolutions used in the training process can affect our model's robustness with large homogeneous image regions. This issue arises because our network was trained on a certain range of effective receptive fields, which is very low compared to the receptive fields required for images with very high resolution. To that end, we use the bilateral guided upsampling method \cite{chen2016bilateral} to process high-resolution images. First, we resize the input test image to have a maximum dimension of 512 pixels. Then, we process the downsampled version of the input image using our model, followed by applying the fast upsampling technique~\cite{chen2016bilateral} with a bilateral grid of $22\!\times\!22\!\times\!8$ cells. This process allows us to process a 16-megapixel image in $\sim$4.5 seconds on average. This time includes $\sim$0.5 seconds to run our network on an NVIDIA$^\circledR$ GeForce GTX 1080$^\texttt{TM}$ GPU and $\sim$4 seconds on an Intel$^\circledR$ Xeon$^\circledR$ E5-1607 @ 3.10 GHz machine for the guided upsampling process. Note the runtime of the guided upsampling step can be significantly improved with a Halide implementation \cite{HALIDE}.

\begin{figure}[t]
\includegraphics[width=\linewidth]{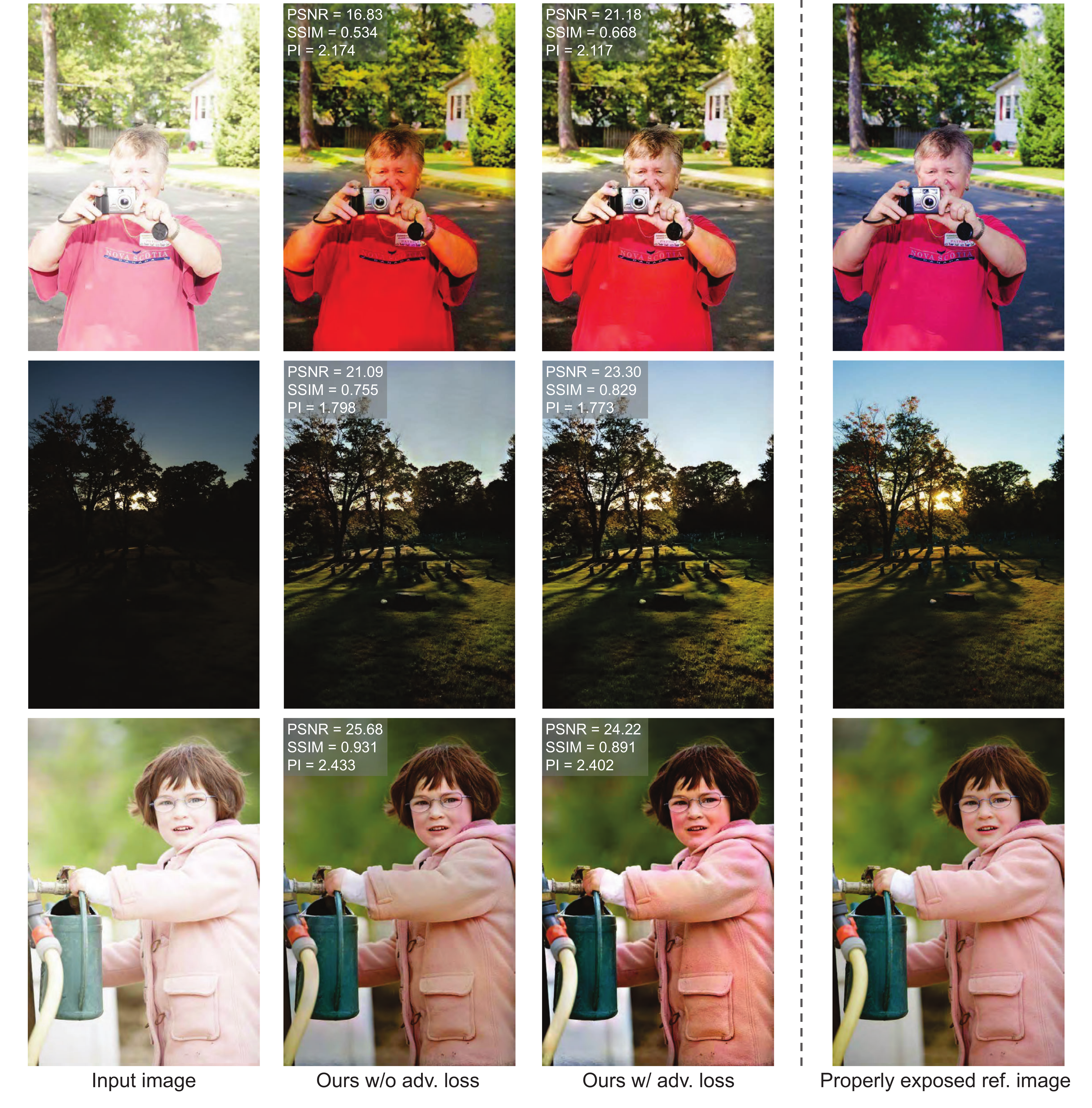}
\vspace{-7mm}
\caption[Comparisons between our results with (w/) and without (w/o) the adversarial loss for training.]{Comparisons between our results with (w/) and without (w/o) the adversarial loss for training. The PSNR, structural similarity index measure (SSIM) \cite{wang2004image}, and perceptual index (PI) \cite{blau20182018} are shown for each result. Notice that higher PSNR and SSIM values are better, while lower PI values indicate better perceptual quality. The input images are taken from our test set.} \label{exposure:fig:ablation-with_without_discriminator}
\end{figure}

\subsection{Training Details}\label{exposure:subsec:training_details}

\begin{figure}[!t]
\includegraphics[width=\linewidth]{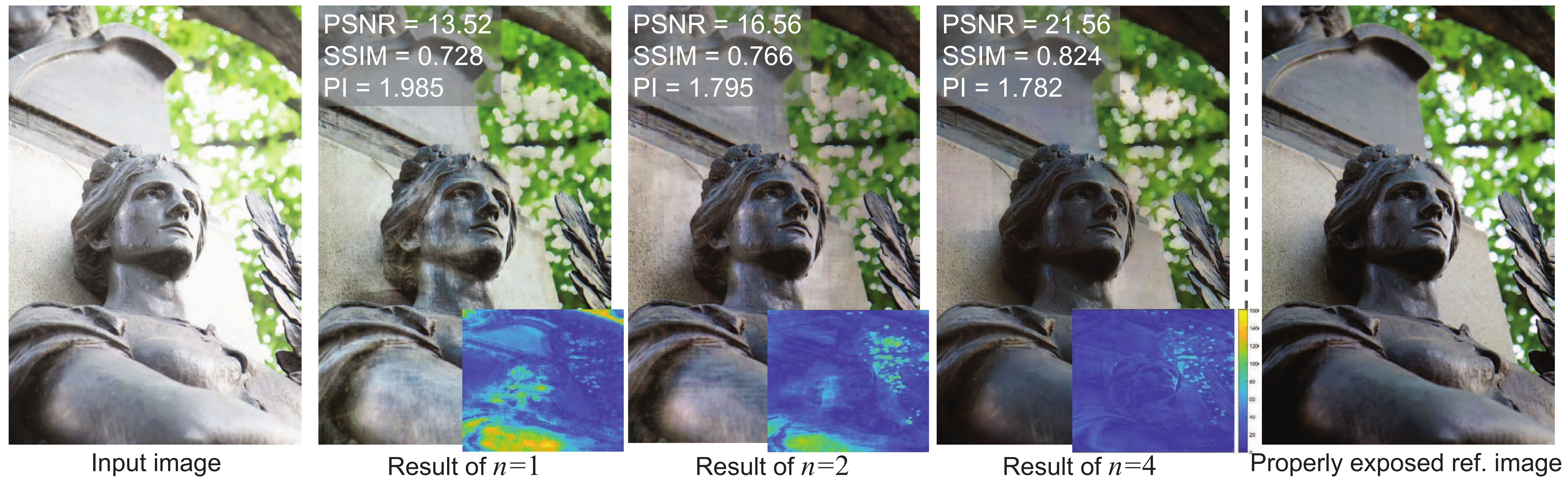}
\vspace{-7mm}
\caption[\hspace{0.5mm} Comparison of results by varying the number of Laplacian pyramid levels.]{Comparison of results by varying the number of Laplacian pyramid levels. Notice that higher PSNR and SSIM values are better, while lower PI values indicate better perceptual quality. The input image is taken from our validation set.}
\label{exposure:fig:ablation1}
\end{figure}

In our implementation, we use a Laplacian pyramid with four levels (i.e., $n=4$) and thus we have four sub-networks in our model---an ablation study evaluating the effect on the
number of Laplacian levels, including a comparison with a vanilla U-Net architecture, is presented in Sec. \ref{exposure:sec:ablations}. We trained our model on patches randomly extracted from training images with different dimensions. We begin our training without $\mathcal{L}_{\text{adv}}$ on 176,590 patches with dimensions of $128\!\times\!128$ pixels extracted randomly from our training images for 40 epochs. The mini-batch size is set to 32. The learning rate is decayed by a factor of 0.5 after the first 20 epochs. Then, we continue training on another 105,845 patches with dimensions of $256\!\times\!256$ pixels for 30 epochs with a mini-batch size of eight. At this stage, we train our main network without $\mathcal{L}_{\text{adv}}$ for 15 epochs and continue training for another 15 epochs with $\mathcal{L}_{\text{adv}}$. The learning rates for the main network and the discriminator network are decayed by a factor of 0.5 every 10 epochs. Finally, we fine-tune the trained networks on another 69,515 training patches with dimensions of $512\!\times\!512$ pixels for 20 epochs with a mini-batch size of four and a learning rate decay of 0.5 applied every five epochs.

We use the Adam optimizer \cite{kingma2014adam} to minimize our loss function in Eq.\ \ref{exposure:eq:final_loss}.
Inspired by previous work \cite{ma2017pose}, we initially train without the adversarial loss term $\mathcal{L}_{\text{adv}}$ to speed up the convergence of our main network. Upon convergence, we then add the adversarial loss term $\mathcal{L}_{\text{adv}}$ and fine-tune our network to enhance our initial results.

We use He et al.'s method \cite{he2015delving} to initialize the weights of our encoder and decoder conv layers, while the bias terms are initialized to zero. We minimize our loss functions using the Adam optimizer \cite{kingma2014adam} with a decay rate $\beta_1 = 0.9$ for the exponential moving averages of the gradient and a decay rate $\beta_2 = 0.999$ for the squared gradient. We use a learning rate of $10^{-4}$ to update the parameters of our main network and a learning rate of $10^{-5}$ to update our discriminator's parameters.

We discard any training patches that have an average intensity less than 0.02 or higher than 0.98. We also discard homogeneous patches that have an average gradient magnitude less than 0.06. We randomly left-right flip training patches for data augmentation.

In the adversarial training, we optimize both the main network and the discriminator in an iterative manner. At each optimization step, the learnable parameters of each network are updated to minimize its own loss function. The discriminator is trained to minimize the following loss function \cite{goodfellow2014generative}:

\begin{equation}
\label{exposure:eq:dsc_loss}
\mathcal{L}_{\text{dsc}} = r\left(\mat{T}\right) + c\left(\mat{Y}\right),
\end{equation}
where $r\left(\mat{T}\right)$ refers to the discriminator loss of recognizing the properly exposed reference image $\mat{T}$, while $c\left(\mat{Y}\right)$ refers to the discriminator loss of recognizing our corrected image $\mat{Y}$. The $r\left(\mat{T}\right)$ and $c\left(\mat{Y}\right)$ loss functions are given by the following equations: 
\begin{equation}
\label{exposure:eq:r_loss}
r\left(\mat{T}\right) =
-\log\left(\mathcal{S}\left(\mathcal{D}\left(\mat{T}\right)\right)\right), 
\end{equation}

\begin{equation}
\label{exposure:eq:e_loss}
c\left(\mat{Y}\right) = 
-\log\left(1-\mathcal{S}\left(\mathcal{D}\left(\mat{Y}\right)\right)\right),
\end{equation}    

\noindent where $\mathcal{S}$ denotes the sigmoid function and $\mathcal{D}$ is the discriminator network described in Fig.\ \ref{exposure:fig:arch}-(B).

\begin{figure}[!t]
\centering
\includegraphics[width=0.95\linewidth]{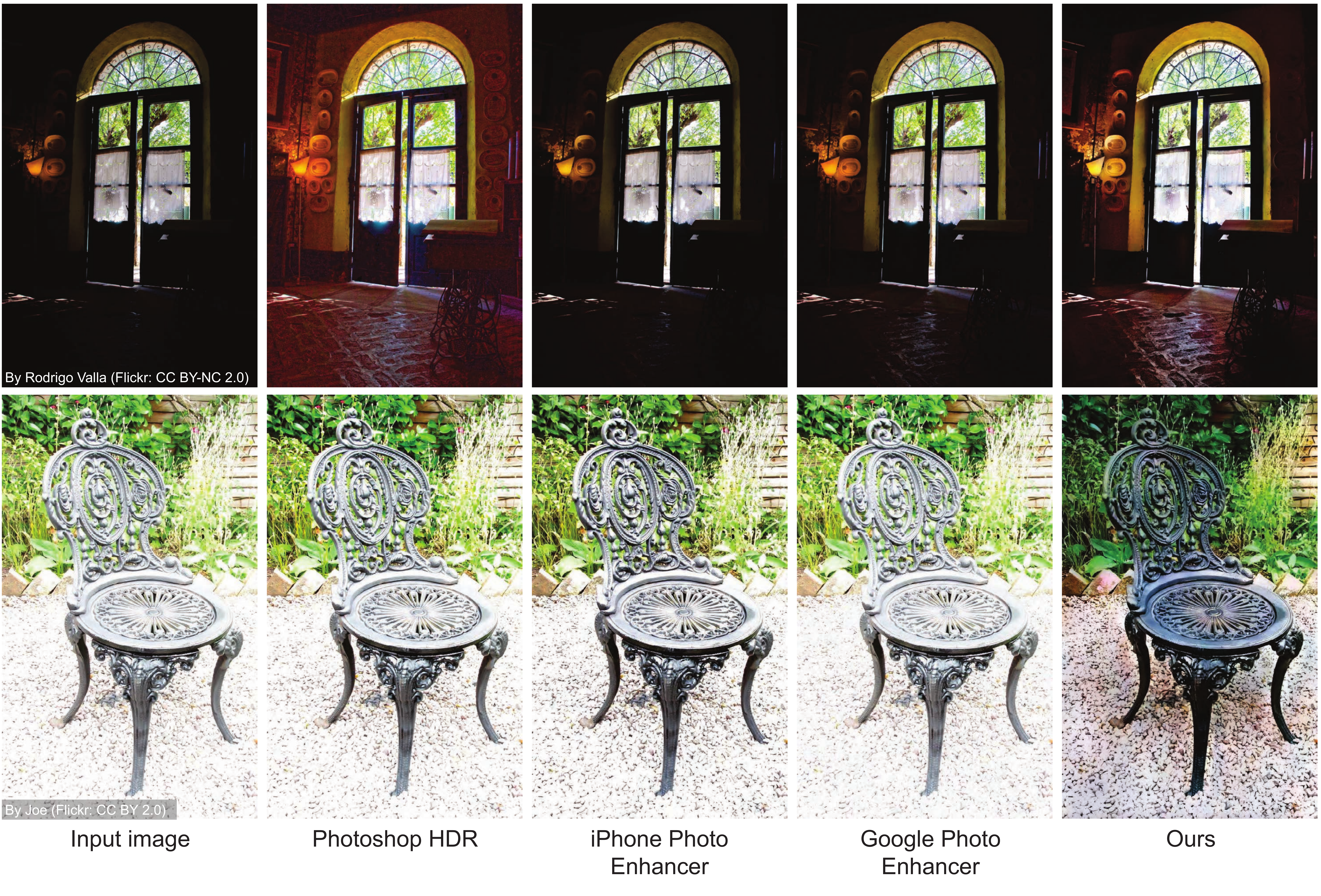}
\vspace{-7mm}
\caption[\hspace{0.5mm} Comparisons with commercial software packages.]{Comparisons with commercial software packages. 
The input images are taken from Flickr.}
\label{exposure:fig:ours_vs_commercial_sw_supp}
\end{figure}

\section{Ablation Studies}\label{exposure:sec:ablations}

\subsection{Loss Function}

Our loss function  includes three main terms. The first term is the standard reconstruction loss (i.e., $\texttt{L}_1$ loss). The second and third terms consist of the pyramid and adversarial losses, respectively, which are introduced to further improve the reconstruction and perceptual quality of the output images. In the following, we discuss the effect of these loss terms.

\begin{table}[t]
\caption[Results of our ablation study on 500 images randomly selected from our validation set.]{Results of our ablation study on 500 images randomly selected from our validation set. We show the effects of: (i) the pyramid loss, $\mathcal{L}_{\text{pyr}}$, and (ii) the number of Laplacian levels, $n$, in the main network. For each experiment, we show the values of the PSNR and SSIM \cite{wang2004image}. The best PSNR/SSIM values are indicated with bold for each experiment.\vspace{-7mm}}\label{exposure:table:ablation}
\centering
\scalebox{0.92}{
\begin{tabular}{l|c|c:c|c|c|}
\cline{2-6}
 & \multicolumn{2}{c:}{\cellcolor[HTML]{FFDDDA} Pyramid loss $\mathcal{L}_{\text{pyr}}$} & \multicolumn{3}{c|}{\cellcolor[HTML]{CCECEB} Number of levels $n$} \\ \cline{2-6} 
 &  w/o & w/ & $n=1$ & $n=2$ & $n=4$ \\ \hline
\multicolumn{1}{|l|}{PSNR} & 18.041 & \textbf{18.385}  & 16.984 & 17.442 & \textbf{18.385}  \\ \hline
\multicolumn{1}{|l|}{SSIM} & 0.746 & \textbf{0.749}  & 0.723 & 0.734 & \textbf{0.749}  \\ \hline
\end{tabular}}
\end{table}

\subsubsection{Pyramid Loss Impact} \label{exposure:subsubsec:pyramid}

In this ablation study, we aim to quantitatively evaluate the effect of the pyramid loss on our final results.

We train two light-weight models of our main network with and without our pyramid loss term. Each model has four 3-layer U-Nets with a total of $\sim$4M learnable parameters, where the number of output channels of the first encoder in each U-Net is set to 24. 

The training is performed on a sub-set of our training data for $\sim$150,000 iterations on 80,000 $128\!\times\!128$ patches, $\sim$100,000 iterations on 40,000 $256\!\times\!256$ patches, and $\sim$25,000 iterations on 25,000 $512\!\times\!512$ patches. Table\ \ref{exposure:table:ablation} shows the results on 500 randomly selected images from our validation set. The results show that the pyramid loss not only helps in providing a better interpretation of the task of each sub-network but also improves the final results.

\subsubsection{Adversarial Loss Impact}

In Tables \ref{exposure:table:results1}--\ref{exposure:table:results3}, we show quantitative results of our method with and without the adversarial loss term. Our trained model with the adversarial loss term achieves better perceptual quality (i.e., lower perceptual index (PI) values \cite{blau20182018}) than training without the adversarial loss term.

Figure \ref{exposure:fig:ablation-with_without_discriminator} shows qualitative comparisons of our results with and without the adversarial loss. As shown, the network trained without the adversarial training tends to produce darker images with slightly unrealistic colors in some cases, while the adversarial regularization improves the perceptual quality of our results.

\begin{figure}[t]
\includegraphics[width=\linewidth]{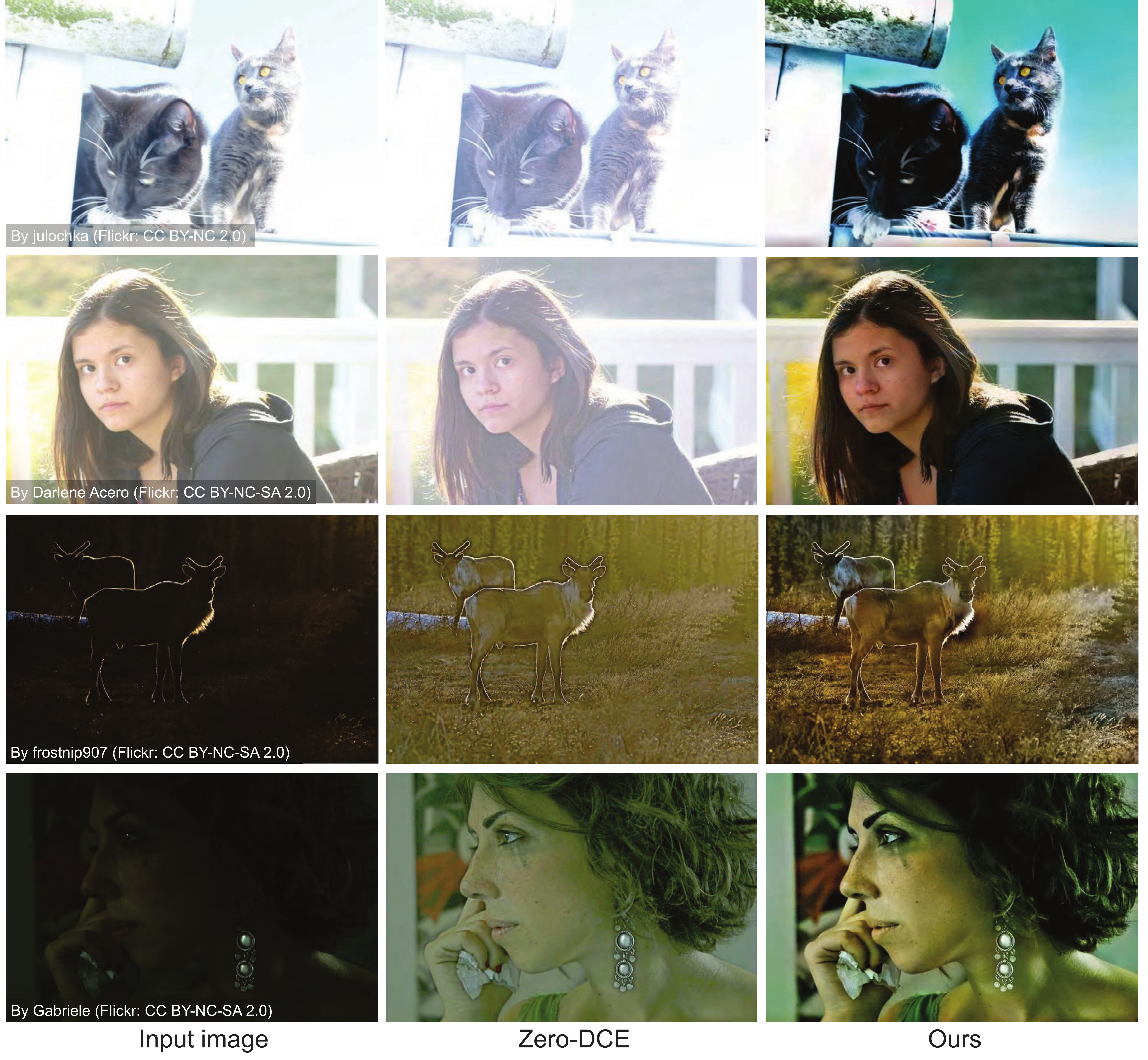}
\vspace{-7mm}
\caption[\hspace{0.5mm} Comparison with the recent Zero-DCE method \cite{guo2020zero} using images taken from Flickr.]{Comparison with the recent Zero-DCE method \cite{guo2020zero} using images taken from Flickr.}
\label{exposure:fig:flickr}
\end{figure}

\subsection{Number of Laplacian Pyramid Levels}
We repeat the same experimental setup described in Sec.\ \ref{exposure:subsubsec:pyramid} with a varying number of Laplacian pyramid levels (sub-networks). Specifically, we train a network with $n=1$ levels---this network is equivalent to a vanilla U-Net-like architecture \cite{ronneberger2015u}. Additionally, we train another network with $n=2$ (i.e., two sub-networks).

For a fair comparison, we fix the total number of parameters in each model by changing the number of filters in the conv layers. Specifically, we set the number of output channels of the first layer in the encoder to 48 for the trained model with $n=1$, while we decrease it to 34 for the two-sub-net model (i.e., $n=2$) to have approximately the same number of learnable parameters.  Thus, the trained model in Sec.\ \ref{exposure:subsubsec:pyramid}, used to study the pyramid loss impact, and the additional two trained models have approximately the same number of parameters.

Table\ \ref{exposure:table:ablation} shows the results obtained by each model on the same random validation image subset used to study the pyramid loss impact in Sec.\ \ref{exposure:subsubsec:pyramid}. Figure \ref{exposure:fig:ablation1} shows a qualitative comparison. As can be seen, the best quantitative and qualitative results are obtained using the four-sub-net model  (i.e., $n=4$ levels).

\section{Empirical Evaluation} \label{exposure:sec:results}
We compare our method against a broad range of existing methods for exposure correction and image enhancement. We first present quantitative results and comparisons in Sec.\ \ref{exposure:subsec:quan-results}, followed by qualitative comparisons in Sec.\ \ref{exposure:subsec:qual-results}.

\subsection{Quantitative Results} \label{exposure:subsec:quan-results}

\begin{table}[t]
\caption[Quantitative evaluation on our introduced over-exposure test set.]{Quantitative evaluation on our introduced over-exposure test set. \textbf{The best results are highlighted with green and bold. The second- and third-best results are highlighted in yellow and red, respectively.} We denote methods designed for underexposure correction in gray. Non-deep learning methods are marked by $\textasteriskcentered$.}\label{exposure:table:results1}
\centering
\scalebox{0.55}{
\begin{tabular}{|l|c|c|c|c|c|c|c|c|c|c|c|c|c|}
\cline{1-14}
\multirow{2}{*}{Method} & \multicolumn{2}{c|}{Expert A} & \multicolumn{2}{c|}{Expert B} & \multicolumn{2}{c|}{Expert C} & \multicolumn{2}{c|}{Expert D} & \multicolumn{2}{c|}{Expert E} & \multicolumn{2}{c|}{Avg.} & \multirow{2}{*}{$\texttt{PI}$ $\downarrow$} \\ \cline{2-13}
 & PSNR $\uparrow$ & SSIM $\uparrow$ & PSNR & SSIM & PSNR & SSIM & PSNR & SSIM & PSNR & SSIM & PSNR & SSIM & \\ \cline{1-14}

\multicolumn{14}{|c|}{\cellcolor[HTML]{CCECEB}$+0$, $+1$, and $+1.5$ relative EVs (3,543 properly exposed and overexposed images)}\\ \hline
HE \cite{10.5555/559707} $\textasteriskcentered$& 16.140 & \cellcolor[HTML]{FFCCCB}0.686 & 16.277 & 0.672 & 16.531 & 0.699 & 16.643 & 0.669 & 17.321  & 0.691 & 16.582 & 0.683 & 2.351\\
CLAHE \cite{adaptivehisteq} $\textasteriskcentered$& 13.934 & 0.568 & 14.689 & 0.586 & 14.453 &  0.584 & 15.116 & 0.593 & 15.850 & 0.612 & 14.808 & 0.589 & 2.270\\
WVM \cite{fu2016weighted} $\textasteriskcentered$& 12.355 & 0.624 & 13.147 & 0.656 & 12.748 & 0.645 & 14.059 & 0.669 & 15.207 & 0.690 & 13.503 & 0.657 & 2.342\\
\cellcolor[HTML]{D5D5D5}LIME \cite{guo2016lime, guo2017lime} $\textasteriskcentered$& 09.627 & 0.549 & 10.096 & 0.569 & 9.875 & 0.570 & 10.936 & 0.597 & 11.903 & 0.626 & 10.487 & 0.582 & 2.412\\
HDR CNN \cite{HDRCNN} w/ RHT \cite{yang2018image}& 13.151 & 0.475 & 13.637 & 0.478 & 13.622 & 0.497 & 14.177 & 0.479 & 14.625 & 0.503 & 13.842 & 0.486 & 4.284\\
HDR CNN \cite{HDRCNN} w/ PS \cite{dayley2010photoshop}& 14.804  & 0.651 & 15.622 & 0.689 & 15.348 & 0.670 & 16.583  & 0.685 & 18.022 & \cellcolor[HTML]{FFCCCB}0.703 & 16.076 & 0.680 & \cellcolor[HTML]{FFCCCB}2.248\\
DPED (iPhone) \cite{DPED}& 12.680  &  0.562 & 13.422 & 0.586 & 13.135 & 0.581 & 14.477 & 0.596 & 15.702 & 0.630 & 13.883 & 0.591 & 2.909 \\
DPED (BlackBerry) \cite{DPED} & 15.170 & 0.621 & 16.193 & 0.691 &  15.781 & 0.642 & 17.042 & 0.677 & 18.035 & 0.678 & 16.444 & 0.662 & 2.518\\
DPED (Sony) \cite{DPED}& \cellcolor[HTML]{FFCCCB}16.398 & 0.672 & \cellcolor[HTML]{FFCCCB}17.679 & \cellcolor[HTML]{FFCCCB}0.707 & \cellcolor[HTML]{FFCCCB}17.378 & 0.697 & \cellcolor[HTML]{FFCCCB}17.997 & \cellcolor[HTML]{FFCCCB}0.685 & \cellcolor[HTML]{FFCCCB}18.685 & 0.700 & \cellcolor[HTML]{FFCCCB}17.627 & \cellcolor[HTML]{FFCCCB}0.692 & 2.740\\
DPE (HDR) \cite{DPE} & 14.399 & 0.572 & 15.219 & 0.573 & 15.091 & 0.593 & 15.692 & 0.581 & 16.640 & 0.626 & 15.408 & 0.589 & 2.417\\
DPE (U-FiveK) \cite{DPE} & 14.314 & 0.615 & 14.958 & 0.628 & 15.075 & 0.645 & 15.987 & 0.647 & 16.931 & 0.667 & 15.453 & 0.640 & 2.630\\
DPE (S-FiveK) \cite{DPE} & 14.786 & 0.638 & 15.519 & 0.649 & 15.625 &  0.668 & 16.586 & 0.664 &  17.661 & 0.684 & 16.035 & 0.661 & 2.621\\
\cellcolor[HTML]{D5D5D5}HQEC \cite{HQEC} $\textasteriskcentered$& 11.775 & 0.607 & 12.536 & 0.631 & 12.127 & 0.627 & 13.424 & 0.652 & 14.511 & 0.675 & 12.875 & 0.638 & 2.387\\
\cellcolor[HTML]{D5D5D5}RetinexNet \cite{Chen2018Retinex} & 10.149 & 0.570 & 10.880 & 0.586 & 10.471 & 0.595 & 11.498 & 0.613 & 12.295 & 0.635 & 11.059 & 0.600 & 2.933\\
\cellcolor[HTML]{D5D5D5}Deep UPE \cite{DeepUPE} & 10.047 & 0.532 & 10.462 & 0.568 & 10.307 & 0.557 & 11.583 & 0.591 & 12.639 & 0.619 & 11.008 & 0.573 & 2.428\\
\cellcolor[HTML]{D5D5D5}Zero-DCE \cite{guo2020zero}  & 10.116 & 0.503 & 10.767 & 0.502 & 10.395 & 0.514 & 11.471 & 0.522 & 12.354 & 0.557 & 11.0206 & 0.5196 & 2.774 \\
\hdashline
Our method w/o $\mathcal{L}_{\text{adv}}$ & \cellcolor[HTML]{79CC7A}\textbf{18.976} & \cellcolor[HTML]{79CC7A}\textbf{0.743} & \cellcolor[HTML]{79CC7A}\textbf{19.767} & \cellcolor[HTML]{79CC7A}\textbf{0.731} & \cellcolor[HTML]{79CC7A}\textbf{19.980} & \cellcolor[HTML]{79CC7A}\textbf{0.768} & \cellcolor[HTML]{79CC7A}\textbf{18.966} & \cellcolor[HTML]{79CC7A}\textbf{0.716} & \cellcolor[HTML]{79CC7A}\textbf{19.056} & \cellcolor[HTML]{79CC7A}\textbf{0.727} & \cellcolor[HTML]{79CC7A}\textbf{19.349} & \cellcolor[HTML]{79CC7A}\textbf{0.737} & \cellcolor[HTML]{FFFBA3}2.189\\
Our method w/ $\mathcal{L}_{\text{adv}}$ & \cellcolor[HTML]{FFFBA3}18.874 & \cellcolor[HTML]{FFFBA3}0.738 & \cellcolor[HTML]{FFFBA3}19.569 & \cellcolor[HTML]{FFFBA3}0.718 & \cellcolor[HTML]{FFFBA3}19.788 & \cellcolor[HTML]{FFFBA3}0.760 & \cellcolor[HTML]{FFFBA3}18.823 & \cellcolor[HTML]{FFFBA3}0.705 & \cellcolor[HTML]{FFFBA3}18.936 & \cellcolor[HTML]{FFFBA3}0.719 & \cellcolor[HTML]{FFFBA3}19.198 & \cellcolor[HTML]{FFFBA3}0.728 & \cellcolor[HTML]{79CC7A}\textbf{2.183}\\\hline
\end{tabular}
}
\end{table}

\begin{table}[t]
\caption[Quantitative evaluation on our introduced under-exposure test set set.]{Quantitative evaluation on our introduced under-exposure test set set. \textbf{The best results are highlighted with green and bold. The second- and third-best results are highlighted in yellow and red, respectively.} We denote methods designed for underexposure correction in gray. Non-deep learning methods are marked by $\textasteriskcentered$.}\label{exposure:table:results2}
\centering
\scalebox{0.55}{
\begin{tabular}{|l|c|c|c|c|c|c|c|c|c|c|c|c|c|}
\cline{1-14}
\multirow{2}{*}{Method} & \multicolumn{2}{c|}{Expert A} & \multicolumn{2}{c|}{Expert B} & \multicolumn{2}{c|}{Expert C} & \multicolumn{2}{c|}{Expert D} & \multicolumn{2}{c|}{Expert E} & \multicolumn{2}{c|}{Avg.} & \multirow{2}{*}{$\texttt{PI}$ $\downarrow$} \\ \cline{2-13}
 & PSNR $\uparrow$ & SSIM $\uparrow$ & PSNR & SSIM & PSNR & SSIM & PSNR & SSIM & PSNR & SSIM & PSNR & SSIM & \\ \cline{1-14}

\multicolumn{14}{|c|}{\cellcolor[HTML]{CCECEB}$-1$ and $-1.5$ relative EVs (2,362 underexposed images)}\\ \hline
HE \cite{10.5555/559707} $\textasteriskcentered$& 16.158 & 0.683 & 16.293 & 0.669 & 16.517  & 0.692 & 16.632 &  0.665 & 17.280 & 0.684 & 16.576 & 0.679 & 2.486\\
CLAHE \cite{adaptivehisteq} $\textasteriskcentered$& 16.310 & 0.619 & 17.140 & 0.646 & 16.779 & 0.621 & 15.955 & 0.613 & 15.568  & 0.608 & 16.350 & 0.621 & 2.387\\
WVM \cite{fu2016weighted} $\textasteriskcentered$& 17.686 & 0.728 & 19.787 & \cellcolor[HTML]{79CC7A}\textbf{0.764} & 18.670 & 0.728 & 18.568 & \cellcolor[HTML]{FFFBA3}0.729 & 18.362 & \cellcolor[HTML]{FFCCCB}0.724 & 18.615 & 0.735 & 2.525\\
\cellcolor[HTML]{D5D5D5}LIME \cite{guo2016lime, guo2017lime} $\textasteriskcentered$& 13.444 & 0.653 & 14.426 & 0.672 & 13.980 & 0.663 & 15.190 & 0.673 & 16.177 & 0.694 & 14.643 & 0.671 & 2.462\\
HDR CNN \cite{HDRCNN} w/ RHT \cite{yang2018image}& 14.547 & 0.456 & 14.347 & 0.427 & 14.068 & 0.441 & 13.025 & 0.398 &  11.957 & 0.379 & 13.589 & 0.420 & 5.072\\
HDR CNN \cite{HDRCNN} w/ PS \cite{dayley2010photoshop}& 17.324 & 0.692 & 18.992 & 0.714 & 18.047 & 0.696 & 18.377 & 0.689 & \cellcolor[HTML]{FFFBA3}19.593 & 0.701 & 18.467 & 0.698 & \cellcolor[HTML]{79CC7A}\textbf{2.294}\\
DPED (iPhone) \cite{DPED}& 18.814 & 0.680 & \cellcolor[HTML]{FFFBA3}21.129 & 0.712 & 20.064 & 0.683 &  \cellcolor[HTML]{79CC7A}\textbf{19.711}  & 0.675 & \cellcolor[HTML]{FFCCCB}19.574  & 0.676 & \cellcolor[HTML]{FFFBA3}19.858 & 0.685 & 2.894\\
DPED (BlackBerry) \cite{DPED} & \cellcolor[HTML]{FFFBA3}19.519 & 0.673 & \cellcolor[HTML]{79CC7A}\textbf{22.333} & 0.745 & 20.342 & 0.669 & 19.611 & 0.683 & 18.489 & 0.653 & \cellcolor[HTML]{79CC7A}\textbf{20.059} & 0.685 & 2.633\\
DPED (Sony) \cite{DPED}& 18.952 & 0.679 & 20.072 & 0.691 & 18.982 &  0.662 & 17.450 & 0.629 & 15.857 & 0.601 & 18.263 & 0.652 & 2.905\\
DPE (HDR) \cite{DPE} & 17.625 & 0.675 & 18.542 & 0.705 & 18.127  & 0.677 & 16.831 & 0.665 & 15.891 & 0.643 & 17.403 & 0.673 & \cellcolor[HTML]{FFFBA3}2.340\\
DPE (U-FiveK) \cite{DPE} & 19.130 & 0.709 & 19.574 & 0.674 & 19.479 & 0.711 & 17.924 & 0.665 & 16.370 & 0.625 & 18.495 & 0.677 & 2.571\\
DPE (S-FiveK) \cite{DPE} & \cellcolor[HTML]{79CC7A}\textbf{20.153} & \cellcolor[HTML]{FFCCCB}0.738 & \cellcolor[HTML]{FFCCCB}20.973 & 0.697 & \cellcolor[HTML]{79CC7A}\textbf{20.915} & 0.738 & \cellcolor[HTML]{FFCCCB}19.050 & 0.688 & 17.510 & 0.648 & \cellcolor[HTML]{FFCCCB}19.720 & 0.702  & 2.564\\
\cellcolor[HTML]{D5D5D5}HQEC \cite{HQEC} $\textasteriskcentered$& 15.801 & 0.692 & 17.371 & 0.718 & 16.587 & 0.700 & 17.090 & 0.705 & 17.675 & 0.716 & 16.905 & 0.706 & 2.532\\
\cellcolor[HTML]{D5D5D5}RetinexNet \cite{Chen2018Retinex}  & 11.676 & 0.607 & 12.711 & 0.611 & 12.132 & 0.621 & 12.720 & 0.618 & 13.233 & 0.637 & 12.494 & 0.619 & 3.362\\
\cellcolor[HTML]{D5D5D5}Deep UPE \cite{DeepUPE} & 17.832 & 0.728 & 19.059 & \cellcolor[HTML]{FFFBA3}0.754 & 18.763 & \cellcolor[HTML]{FFCCCB}0.745 & \cellcolor[HTML]{FFFBA3}19.641 & \cellcolor[HTML]{79CC7A}\textbf{0.737} & \cellcolor[HTML]{79CC7A}\textbf{20.237} & \cellcolor[HTML]{79CC7A}\textbf{0.740} & 19.106 & \cellcolor[HTML]{FFFBA3}0.741 & 2.371\\ 
\cellcolor[HTML]{D5D5D5}Zero-DCE \cite{guo2020zero}  & 13.935 & 0.585 & 15.239 & 0.593 & 14.552 & 0.589 & 15.202 & 0.587 & 15.893 & 0.614 & 14.9642 & 0.5936 & 3.001 \\
\hdashline
Our method w/o $\mathcal{L}_{\text{adv}}$ & \cellcolor[HTML]{FFCCCB}19.432 & \cellcolor[HTML]{FFFBA3}0.750 & 20.590 & \cellcolor[HTML]{FFCCCB}0.739 & \cellcolor[HTML]{FFFBA3}20.542 & \cellcolor[HTML]{79CC7A}\textbf{0.770} & 18.989 & \cellcolor[HTML]{FFCCCB}0.723 & 18.874 & \cellcolor[HTML]{FFFBA3}0.727 & 19.685 & \cellcolor[HTML]{79CC7A}\textbf{0.742} & 2.344 \\
Our method w/ $\mathcal{L}_{\text{adv}}$ & 19.475 & \cellcolor[HTML]{79CC7A}\textbf{0.751} & 20.546 & 0.730 & \cellcolor[HTML]{FFCCCB}20.518 & \cellcolor[HTML]{FFFBA3}0.768 & 18.935 & 0.715 & 18.756 & 0.719 & 19.646 & \cellcolor[HTML]{FFCCCB}0.737 &  \cellcolor[HTML]{FFCCCB}2.342\\\hline
\end{tabular}
}
\end{table}

\begin{figure}[!t]
\includegraphics[width=\linewidth]{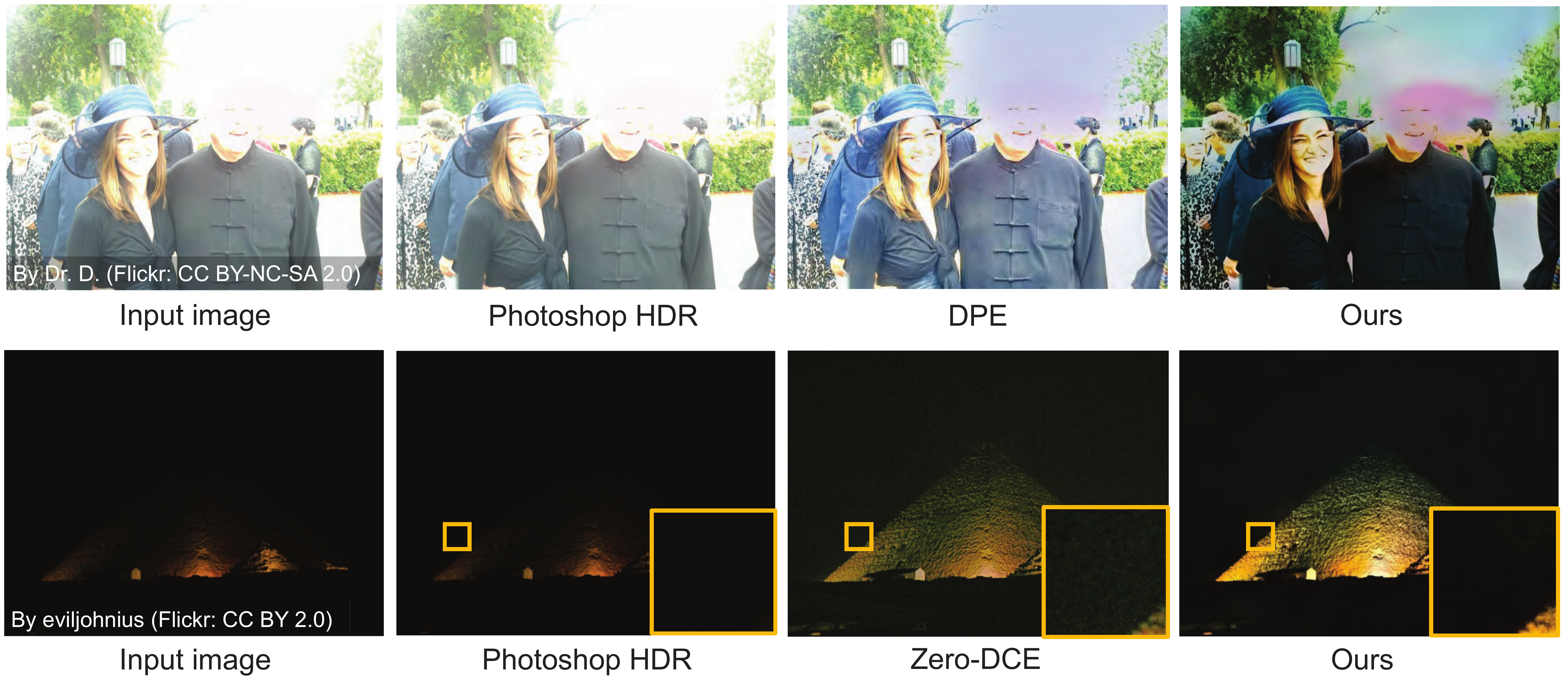}
\vspace{-7mm}
\caption[\hspace{0.5mm} Failure examples of correcting overexposed and underexposed images.]{Failure examples of correcting (top) overexposed and (bottom) underexposed images. The input images are taken from Flickr.\vspace{-7mm}} 
\label{exposure:fig:failureExamples}
\end{figure}

\begin{figure}[!t]
\includegraphics[width=\linewidth]{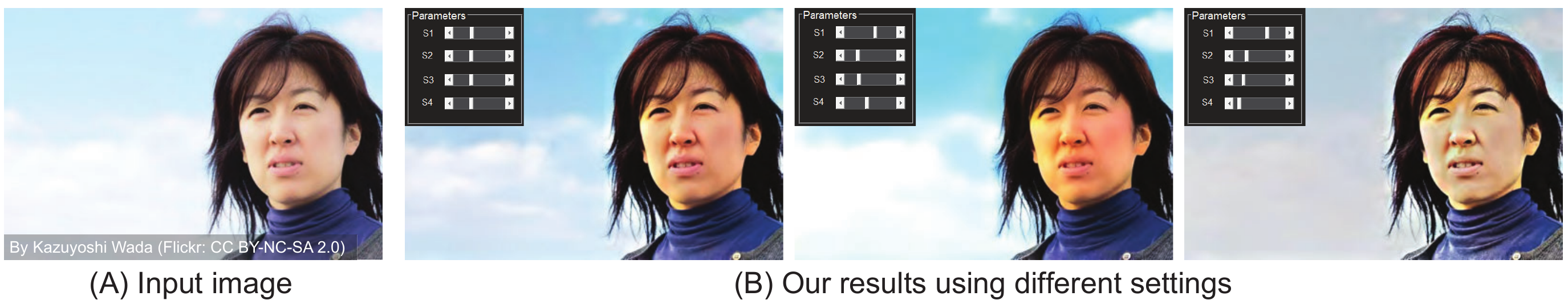}
\vspace{-7mm}
\caption[\hspace{0.5mm} Our GUI photo editing tool.]{Our GUI photo editing tool. (A) Input image. (B) Our results using different pyramid level scaling settings set by the user in an interactive way. The input image is taken from Flickr.}
\label{exposure:fig:gui}
\end{figure}

To evaluate our method, we use our test set, which consists of 5,905 images rendered with different exposure settings, as described in Sec.\ \ref{exposure:subsec:data}. Specifically, our test set includes 3,543 well-exposed/overexposed images rendered with $+0$, $+1$, and $+1.5$ relative EVs, and 2,362 underexposed images rendered with $-1$ and $-1.5$ relative EVs.

We adopt the following three standard metrics to evaluate the pixel-wise accuracy and the perceptual quality of our results: (i)  PSNR, (ii)  SSIM \cite{wang2004image}, and (iii)  perceptual index ($\texttt{PI}$) \cite{blau20182018}. The $\texttt{PI}$ is given by:

\begin{equation}
\label{exposure:eq:PI}
\texttt{PI} = 0.5 (10 - \texttt{Ma} + \texttt{NIQE}),
\end{equation}
where both $\texttt{Ma}$ \cite{ma2017learning} and $\texttt{NIQE}$  \cite{mittal2012making} are \textit{no-reference} image quality metrics.

\begin{table}[!t]
\caption[Quantitative evaluation on our introduced test set (both under- and over-exposed images).]{Quantitative evaluation on our introduced test set (both under- and over-exposed images). \textbf{The best results are highlighted with green and bold. The second- and third-best results are highlighted in yellow and red, respectively.} We denote methods designed for underexposure correction in gray. Non-deep learning methods are marked by $\textasteriskcentered$.}\label{exposure:table:results3}
\centering
\scalebox{0.55}{
\begin{tabular}{|l|c|c|c|c|c|c|c|c|c|c|c|c|c|}
\cline{1-14}
\multirow{2}{*}{Method} & \multicolumn{2}{c|}{Expert A} & \multicolumn{2}{c|}{Expert B} & \multicolumn{2}{c|}{Expert C} & \multicolumn{2}{c|}{Expert D} & \multicolumn{2}{c|}{Expert E} & \multicolumn{2}{c|}{Avg.} & \multirow{2}{*}{$\texttt{PI}$ $\downarrow$} \\ \cline{2-13}
 & PSNR $\uparrow$ & SSIM $\uparrow$ & PSNR & SSIM & PSNR & SSIM & PSNR & SSIM & PSNR & SSIM & PSNR & SSIM & \\ \cline{1-14}

\multicolumn{14}{|c|}{\cellcolor[HTML]{CCECEB}Combined over and underexposed images (5,905 images)}\\ \hline
HE \cite{10.5555/559707} $\textasteriskcentered$& 16.148 & \cellcolor[HTML]{FFCCCB}0.685 & 16.283 & 0.671 & 16.525 & \cellcolor[HTML]{FFCCCB}0.696  & 16.639 & 0.668 & 17.305 & 0.688 & 16.580 & 0.682 & 2.405\\
CLAHE \cite{adaptivehisteq} $\textasteriskcentered$& 14.884 & 0.589 & 15.669 &  0.610 & 15.383 & 0.599 & 15.452 & 0.601 & 15.737 & 0.610 & 15.425 & 0.602 & 2.317\\
WVM \cite{fu2016weighted} $\textasteriskcentered$&  14.488 & 0.665 & 15.803 & \cellcolor[HTML]{FFCCCB}0.699 & 15.117 & 0.678 & 15.863 & \cellcolor[HTML]{FFCCCB}0.693 & 16.469 & \cellcolor[HTML]{FFCCCB}0.704 & 15.548 & 0.688 & 2.415\\
\cellcolor[HTML]{D5D5D5}LIME \cite{guo2016lime, guo2017lime} & 11.154 & 0.591 & 11.828 & 0.610 & 11.517 & 0.607 & 12.638 & 0.628 & 13.613 & 0.653 & 12.150 & 0.618 & 2.432\\
HDR CNN \cite{HDRCNN} w/ RHT \cite{yang2018image}& 13.709 & 0.467 & 13.921 & 0.458 & 13.800 & 0.474 & 13.716 & 0.446 & 13.558 & 0.454 & 13.741 & 0.460 & 4.599\\
HDR CNN \cite{HDRCNN} w/ PS \cite{dayley2010photoshop}  & 15.812 & 0.667 & 16.970 & 0.699 & 16.428 & 0.681 & 17.301 & 0.687 & 18.650  & 0.702 & 17.032 & \cellcolor[HTML]{FFCCCB}0.687 & \cellcolor[HTML]{FFCCCB}2.267\\
DPED (iPhone) \cite{DPED}& 15.134 & 0.609 & 16.505 & 0.636 & 15.907 & 0.622 & 16.571 & 0.627 & 17.251 & 0.649 & 16.274 & 0.629 & 2.903\\
DPED (BlackBerry) \cite{DPED} & 16.910 & 0.642 & \cellcolor[HTML]{FFCCCB}18.649 & \cellcolor[HTML]{FFCCCB}0.713 & 17.606 & 0.653 & \cellcolor[HTML]{FFCCCB}18.070 & 0.679 & 18.217 & 0.668 & \cellcolor[HTML]{FFCCCB}17.890 & 0.671 & 2.564\\
DPED (Sony) \cite{DPED}& \cellcolor[HTML]{FFCCCB}17.419  & 0.675 & 18.636 & 0.701 & \cellcolor[HTML]{FFCCCB}18.020  & 0.683 &  17.554 & 0.660 & \cellcolor[HTML]{FFCCCB}17.778 & 0.663 & 17.881 & 0.676 & 2.806\\
DPE (HDR) \cite{DPE} & 15.690 & 0.614 & 16.548 & 0.626 & 16.305 & 0.626 & 16.147 & 0.615 & 16.341 & 0.633 & 16.206 & 0.623 & 2.417\\
DPE (U-FiveK) \cite{DPE} & 16.240 & 0.653 & 16.805 & 0.646 & 16.837 & 0.671 &  16.762 & 0.654 & 16.707 & 0.650 & 16.670 & 0.655 & 2.606\\
DPE (S-FiveK) \cite{DPE} & 16.933 & 0.678 & 17.701 & 0.668 & 17.741 & \cellcolor[HTML]{FFCCCB}0.696 & 17.572 & 0.674 & 17.601 & 0.670 & 17.510 & 0.677 & 2.621\\
\cellcolor[HTML]{D5D5D5}HQEC \cite{HQEC} $\textasteriskcentered$& 13.385 & 0.641 & 14.470 & 0.666 & 13.911 & 0.656 & 14.891 & 0.674 & 15.777 & 0.692 & 14.487 & 0.666 & 2.445\\
\cellcolor[HTML]{D5D5D5}RetinexNet \cite{Chen2018Retinex}  & 10.759 & 0.585 & 11.613 & 0.596 & 11.135 & 0.605 & 11.987 & 0.615 & 12.671 & 0.636 & 11.633 & 0.607 & 3.105\\
\cellcolor[HTML]{D5D5D5}Deep UPE \cite{DeepUPE} & 13.161 & 0.610 & 13.901 & 0.642 & 13.689 & 0.632 & 14.806 & 0.649 & 15.678 & 0.667 & 14.247 & 0.640  & 2.405\\ 
\cellcolor[HTML]{D5D5D5}Zero-DCE \cite{guo2020zero}  & 11.643  & 0.536 & 12.555 & 0.539 & 12.058 & 0.544 & 12.964  & 0.548 & 13.769 & 0.580 & 12.5978 & 0.5494 & 2.865 \\
\hdashline
Our method w/o $\mathcal{L}_{\text{adv}}$ & \cellcolor[HTML]{79CC7A}\textbf{19.158} & \cellcolor[HTML]{79CC7A}\textbf{0.746} &  \cellcolor[HTML]{79CC7A}\textbf{20.096} & \cellcolor[HTML]{79CC7A}\textbf{0.734} & \cellcolor[HTML]{79CC7A}\textbf{20.205} & \cellcolor[HTML]{79CC7A}\textbf{0.769} & \cellcolor[HTML]{79CC7A}\textbf{18.975} & \cellcolor[HTML]{79CC7A}\textbf{0.719} & \cellcolor[HTML]{79CC7A}\textbf{18.983} & \cellcolor[HTML]{79CC7A}\textbf{0.727} & \cellcolor[HTML]{79CC7A}\textbf{19.483} & \cellcolor[HTML]{79CC7A}\textbf{0.739} & \cellcolor[HTML]{FFFBA3}2.251\\
Our method w/ $\mathcal{L}_{\text{adv}}$ & \cellcolor[HTML]{FFFBA3}19.114 & \cellcolor[HTML]{FFFBA3}0.743 & \cellcolor[HTML]{FFFBA3}19.960 & \cellcolor[HTML]{FFFBA3}0.723 & \cellcolor[HTML]{FFFBA3}20.080 & \cellcolor[HTML]{FFFBA3}0.763 & \cellcolor[HTML]{FFFBA3}18.868 & \cellcolor[HTML]{FFFBA3}0.709 & \cellcolor[HTML]{FFFBA3}18.864 & \cellcolor[HTML]{FFFBA3}0.719 & \cellcolor[HTML]{FFFBA3}19.377 & \cellcolor[HTML]{FFFBA3}0.731 & \cellcolor[HTML]{79CC7A}\textbf{2.247} \\\hline
\end{tabular}
}
\end{table}

For the pixel-wise error metrics -- namely, PSNR and SSIM -- we compare the results not only against the properly exposed rendered images by Expert C but also with \textit{all} five expert photographers in the MIT-Adobe FiveK dataset \cite{bychkovsky2011learning}. Though the expert photographers may render the same image in different ways due to differences in the camera-based rendering settings (e.g., white balance and tone mapping), a common characteristic over all rendered images by the expert photographers is that they all have fairly proper exposure settings \cite{bychkovsky2011learning} (see Fig.~\ref{exposure:fig:experts}). For this reason, we evaluate our method against the \textit{five} expert rendered images as they all represent satisfactory exposed reference images.

We also evaluate a variety of previous non-learning and learning-based methods on our test set for comparison: histogram equalization (HE) \cite{10.5555/559707}, 
contrast-limited adaptive histogram equalization (CLAHE) \cite{adaptivehisteq}, the weighted variational model (WVM) \cite{fu2016weighted}, the low-light image enhancement method (LIME) \cite{guo2016lime, guo2017lime}, HDR CNN \cite{HDRCNN},  DPED models \cite{DPED},  deep photo enhancer (DPE) models \cite{DPE}, the high-quality exposure correction method (HQEC) \cite{HQEC}, RetinexNet \cite{Chen2018Retinex}, deep underexposed photo enhancer (UPE) \cite{DeepUPE}, and the zero-reference deep curve estimation method (Zero-DCE) \cite{guo2020zero}. To render the reconstructed HDR images generated by the HDR CNN method \cite{HDRCNN} back into LDR, we tested both the deep reciprocating HDR transformation method (RHT) \cite{yang2018image}, and Adobe Photoshop's (PS) HDR tool \cite{dayley2010photoshop}.

Tables \ref{exposure:table:results1}--\ref{exposure:table:results3} summarizes the quantitative results obtained by each method. As shown in the top portion of the table, our method achieves the best results for overexposed images under
all metrics. In the underexposed image correction setting,
our results (middle portion of table) are on par with the state-of-the-art methods. Finally, in contrast to most of the existing methods, the results in the bottom portion of the table show that our method can effectively deal with \textit{both} types of exposure errors.

\noindent\textbf{Generalization} We further evaluate the generalization ability of our method on the following standard image datasets used by previous low-light image enhancement methods: (i)  LIME (10 images) \cite{guo2017lime}, (ii)  NPE (75 images) \cite{wang2013naturalness}, (iii)  VV (24 images) \cite{VVDataset}, and  DICM (44 images) \cite{lee2012contrast}. Note that in these experiments, we report results of our model trained on our training set without further tuning or re-training on any of these datasets. Similar to previous methods, we use the $\texttt{NIQE}$ perceptual score \cite{mittal2012making} for evaluation. Table\ \ref{exposure:table:results_extra_datasets} compares results by our method and the following methods: LIME \cite{guo2016lime, guo2017lime}, WVM \cite{fu2016weighted}, RetinexNet (RNet) \cite{Chen2018Retinex}, ``kindling the darkness'' (KinD) \cite{zhang2019kindling}, enlighten GAN (EGAN) \cite{jiang2019enlightengan}, and deep bright-channel prior (BCP) \cite{8955834}. As can be seen in Table\ \ref{exposure:table:results_extra_datasets}, our method generally achieves perceptually superior results in correcting low-light 8-bit images of other datasets.

\begin{figure}[!t]
\includegraphics[width=\linewidth]{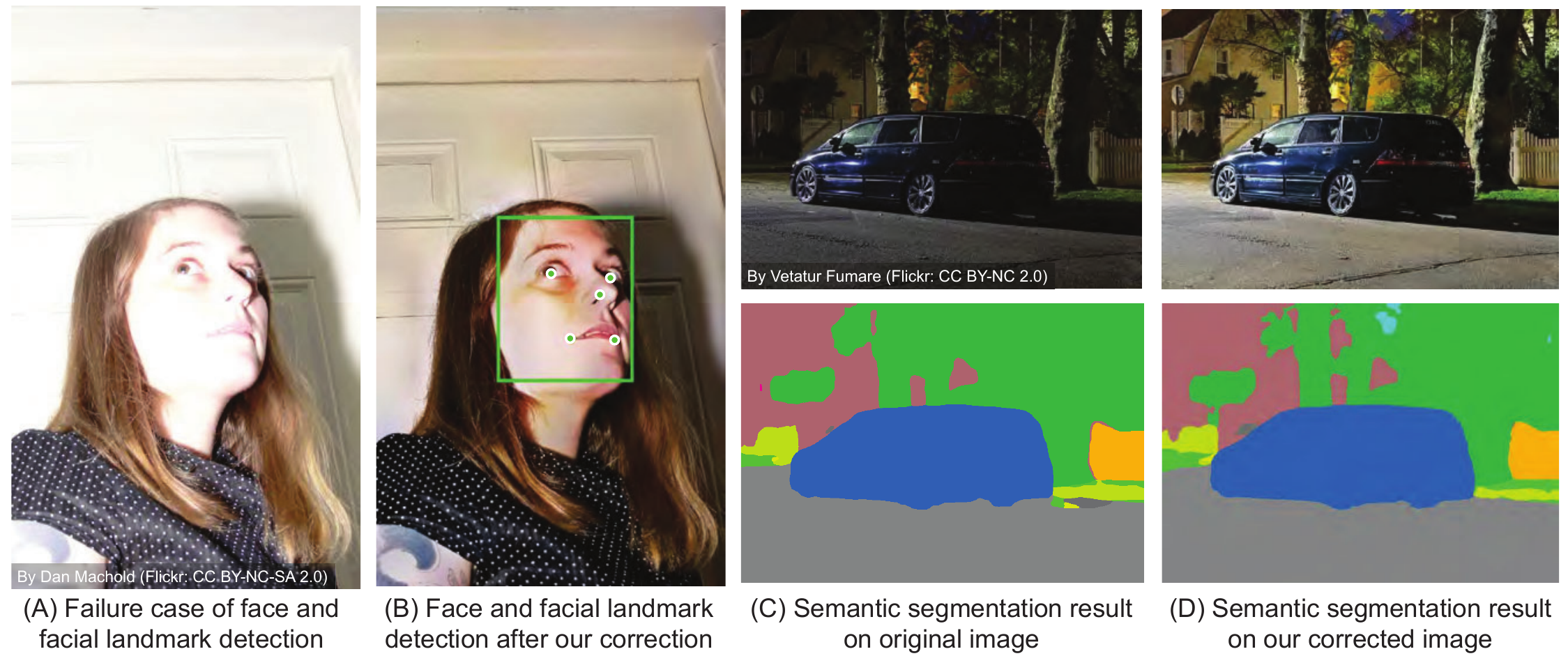}
\vspace{-7mm}
\caption[\hspace{0.5mm} Applying our method as a pre-processing step can improve results of different computer vision tasks.]{Applying our method as a pre-processing step can improve results of different computer vision tasks. (A) False negative result of face and facial landmark detection due to the overexposure error in the input image. (B) Our corrected image and the results of face and facial landmark detection. (C) Underexposed input image and its semantic segmentation mask. (D) Our corrected image and its semantic segmentation mask. We use the cascaded convolutional networks proposed in \cite{zhang2016joint} for face and facial landmark detection. For image semantic segmentation, we use RefineNet \cite{lin2017refinenet, lin2019refinenet}. The input images are taken from Flickr.}
\label{exposure:fig:applications}
\end{figure}

\subsection{Qualitative Results}\label{exposure:subsec:qual-results}

We compare our method qualitatively with a variety of previous methods. Note we show results using the model trained with the adversarial loss term, as it produces perceptually superior results (see the perceptual metric results in Tables \ref{exposure:table:results1}--\ref{exposure:table:results3}).

Figure~\ref{exposure:fig:qualitative_our_set} shows our results on different overexposed and underexposed images. As shown, our results are arguably visually superior to the other methods, even when input images have hard backlight conditions, as shown in the second row in Fig.~\ref{exposure:fig:qualitative_our_set} (right).

\noindent\textbf{Generalization} We also ran our model on several images from Flickr that are outside our introduced dataset, as shown in Figs.\ \ref{exposure:fig:teaser} and \ref{exposure:fig:ours_vs_commercial_sw_supp}.  As with the images from our introduced dataset, our results on the Flickr images are arguably superior to the compared methods. Additional qualitative comparisons using images taken from Flickr are shown in Fig.\ \ref{exposure:fig:flickr}.

\subsection{Limitations}\label{exposure:subsec:limitations}

Our method produces unsatisfactory results in regions that have insufficient semantic information, as shown in Fig.~\ref{exposure:fig:failureExamples}. For example, the input image shown in the first row in Fig.\ \ref{exposure:fig:failureExamples} is completely saturated and contains almost no details in the region of the man's face. We can see that our network cannot constrain the color inside the face region due to the lack of semantic information. In that way, one can control the output results to reduce such color bleeding problems. It also can be observed that our method may introduce noise when the input image has extreme dark regions, as shown in the second example in Fig.\ \ref{exposure:fig:failureExamples}. These challenging conditions prove difficult for other methods as well.

\begin{table}[!t]
\caption[Perceptual quality evaluation.]{Perceptual quality evaluation.  Summary of $\texttt{NIQE}$ scores \cite{mittal2012making} on different \textit{low-light} image datasets. In these dataset, there are no ground-truth images provided for full-reference quality metrics (e.g., PSNR). Highlights are in the same format as Table\ \ref{exposure:table:results1}}
\label{exposure:table:results_extra_datasets}
\centering
\scalebox{0.67}{
\begin{tabular}{|l|c|c|c|c|c|}
\hline
Method & LIME \cite{guo2017lime} & NPE \cite{wang2013naturalness} & VV \cite{VVDataset} & DICM \cite{lee2012contrast} & Avg. \\ \hline
NPE \cite{wang2013naturalness} $\textasteriskcentered$ & 3.91 & 3.95 & \cellcolor[HTML]{FFCCCB}2.52 & 3.76 & 3.54 \\ \hline
LIME \cite{guo2017lime} $\textasteriskcentered$ & 4.16 & 4.26 & \cellcolor[HTML]{FFFBA3}2.49 & 3.85 & 3.69 \\ \hline
WVM \cite{fu2016weighted} $\textasteriskcentered$ & 3.79 & 3.99 & 2.85 & 3.90 & 3.63 \\ \hline
RNet \cite{Chen2018Retinex} & 4.42 & 4.49 & 2.60 & 4.20 & 3.93 \\ \hline
KinD \cite{zhang2019kindling} & \cellcolor[HTML]{79CC7A}\textbf{3.72} & \cellcolor[HTML]{FFCCCB}3.88 & - & - & 3.80 \\ \hline
EGAN \cite{jiang2019enlightengan} & \cellcolor[HTML]{79CC7A}\textbf{3.72} & 4.11 & 2.58 & - & 3.50 \\ \hline
DBCP \cite{8955834} & \cellcolor[HTML]{FFCCCB}3.78 & \cellcolor[HTML]{79CC7A}\textbf{3.18} & - & \cellcolor[HTML]{FFCCCB}3.57 & \cellcolor[HTML]{FFCCCB}3.48 \\ \hdashline
Ours w/o $\mathcal{L}_{\text{adv}}$ & \cellcolor[HTML]{FFFBA3}3.76 & \cellcolor[HTML]{FFFBA3}3.20 & \cellcolor[HTML]{79CC7A}\textbf{2.28} & \cellcolor[HTML]{FFFBA3}2.55 & \cellcolor[HTML]{FFFBA3}2.95 \\ \hline
Ours w/ $\mathcal{L}_{\text{adv}}$ & \cellcolor[HTML]{FFFBA3}3.76 & \cellcolor[HTML]{79CC7A}\textbf{3.18} & \cellcolor[HTML]{79CC7A}\textbf{2.28} & \cellcolor[HTML]{79CC7A}\textbf{2.50} & \cellcolor[HTML]{79CC7A}\textbf{2.93} \\ \hline
\end{tabular}
}
\end{table}

\section{Potential Applications}\label{exposure:sec:applications}

In this section, we highlight two potential applications of our method: (i) photo editing and (ii) image preprocessing.
 
\paragraph{Photo Editing}
The main potential application of the proposed method is to post-capture correct exposure errors in images. This correction process can be performed in a fully automated way or can be performed in an interactive way with the user. Specifically, we introduce a scale vector $\mat{S} = \left[S_1, S_2, S_3, S_4\right]^\top$ that can be used to independently scale each level in the pyramid $\mat{X}$ in the inference stage. The scale vector $\mat{S}$ is introduced to produce different visual effects in the final result $\mat{Y}$. In particular, this scaling operation is performed as a pre-processing of each level in the pyramid $\mat{X}$ as follows: $\mat{S}_{(l=i)}\mat{X}_{(l=i)},$ s.t. $i \in \left\{1, 2, 3, 4\right\}$. 
The values of the scale vector $\mat{S}$ can be interactively controlled by the user to edit our network results. Figure\ \ref{exposure:fig:gui} shows different results obtained by our network in an interactive way through our graphical user interface (GUI). Our GUI can be used as a photo editing tool to apply different visual effects and filters on the input images. 

\paragraph{Image Preprocessing}
Our method can also improve the results of computer vision tasks by using it as a pre-processing step to correct exposure errors in input images. Figure \ref{exposure:fig:applications} shows example applications. In these examples, we show results of face and facial landmark detection of the work in \cite{zhang2016joint} and image semantic segmentation results obtained by the work in \cite{lin2017refinenet, lin2019refinenet}. As shown, the results of face detection and semantic segmentation are improved by pre-processing the input images using our method. In future work, we plan to investigate the impact of our exposure correction method on a variety of computer vision tasks.

\section{Summary}

We proposed a single coarse-to-fine deep learning model for overexposed and underexposed image correction. We employed the Laplacian pyramid decomposition to process input images in different frequency bands. Our method is designed to sequentially correct each of the Laplacian pyramid levels in a multi-scale manner, starting with the global color in the image and progressively addressing the image details. 

Our method is enabled by a new dataset of over 24,000 images rendered with the broadest range of exposure errors to date. Each image in our introduced dataset has a reference image properly rendered by a well-trained photographer with well-exposure compensation. Through extensive evaluation, we showed that our method produces compelling results compared to available solutions for correcting images rendered with exposure errors and it generalizes well. We believe that our dataset will help future work on improving exposure correction for photographs.

\part{Image Recoloring \label{part:recoloring}}
\chapter[Recoloring Based on Object Color Distributions]{Recoloring Based on Object Color\\Distributions \label{ch:ch14}}
This part of the thesis focuses on an auto color editing, where our goal is to manipulate an image's RGB color values to produce a new appearance that conveys a different ``look and feel'' of the image. This procedure of manipulating an image's color in this manner is often referred to as {\it image recoloring}.  We are interested in achieving realistic auto recolor images with minimal user interaction. To that end, we will outline two methods for auto image recoloring. 

In this chapter, we present a method to perform automatic image recoloring based on the distribution of colors associated with objects present in an image\footnote{This work was published in \cite{afifi2019image}: Mahmoud Afifi, Brian Price, Scott Cohen, and Michael S. Brown. Image Recoloring Based on Object Color Distributions. In Eurographics -- Short Paper, 2019.}. For example, when recoloring an image containing a sky object, our method incorporates the observation that objects of class `sky' have a color distribution with three dominant modes for blue (daytime), yellow/red (dusk/dawn), and dark (nighttime). Our work leverages recent deep-learning methods that can perform reasonably accurate object-level segmentation.  By using the images in datasets used to train deep-learning object segmentation methods, we are able to model the color distribution of each object class in the dataset.  Given a new input image and its associated semantic segmentation (i.e., object mask), we perform color transfer to map the input image color histogram to a set of target color histograms that were constructed based on the learned color distribution of the objects in the image.  We show that our framework is able to produce compelling color variations that are often more interesting and unique than results produced by existing methods. The source code of this work is available on GitHub: \href{https://github.com/mahmoudnafifi/Image_recoloring}{https://github.com/mahmoudnafifi/Image$\_$recoloring}.

\begin{figure}
\includegraphics[width=\linewidth]{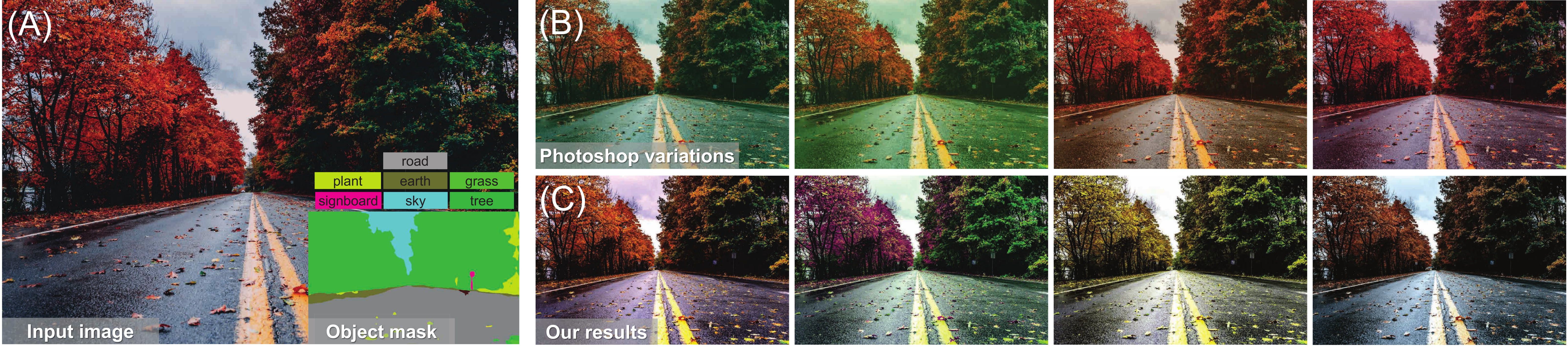}
\vspace{-7mm}
\caption[(A) An input image and its semantic segmentation. (B) Recolored images produced by Photoshop's variation tool. (C) Recolored images from our method that considers the color distribution of objects in the image.]{(A) An input image and its semantic segmentation (object mask) obtained by RefineNet~\cite{lin2017refinenet}. (B) Recolored images produced by Photoshop's variation tool. (C) Recolored images from our method that considers the color distribution of objects in the image.}
\label{DoD:fig:teaser}
\end{figure}

\section{Introduction}

Image recoloring aims at transferring the colors of a given input image to share the same colors and ``feel'' with some target colors. Color manipulation for this purpose is achieved in different ways, such as color transfer (e.g., \cite{faridul2016colour}), appearance transfer (e.g., \cite{laffont2014transient}), and style transfer (e.g.,\cite{LuanCVPR17}). Image editing software, such as Photoshop, provides tools for automatic image recoloring as a way to provide users with interesting variations on an input image. Figure \ref{DoD:fig:teaser} shows a typical case of image recoloring, where an input image is manipulated automatically to produce several recolored variations.

The vast majority of existing recoloring methods require a target image that is specified by the user (e.g., ~\cite{faridul2016colour, gatys2016image, LuanCVPR17}).  There are several methods that perform automatic transfer (e.g., \cite{huang2014learning, lee2016automatic}). To the best of our knowledge, these existing methods do not explicitly consider the objects present in the image in the recoloring process. Such object-level semantic information can be useful in guiding the recoloring effort to produce interesting and plausible variations on the input image. For example, if the input image has a \textit{sky object}, the recolored image should exploit observations from the training data that a sky object can be blue, but not green.  Moreover, if the original input image already has a sky object with a blue color appearance, we can use a dissimilar color associated with the sky object class (e.g., a reddish appearance) to provide more variety in the recolored results.

\begin{figure}[!t]
\includegraphics[width=\linewidth]{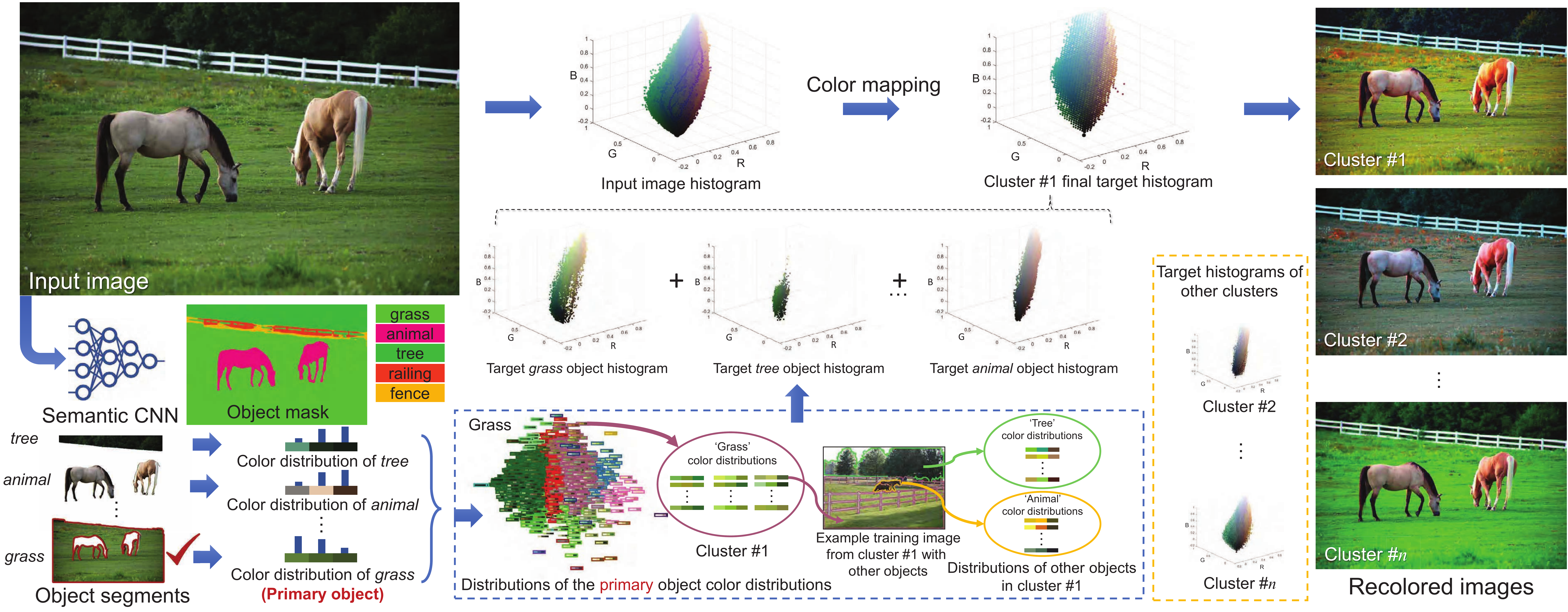}
\vspace{-7mm}
\caption[Our image recoloring framework.]{Our image recoloring framework. We use a deep-learning method to obtain the input image's semantic segmentation as an object mask. We generate a color distribution for each object as a color palette and select one object in the image as the primary object. Next, we visit each cluster in the training images of the primary object class.   For each cluster, we select an object instance with the most \textit{dissimilar} colors to our input object's colors and add its colors to our target histogram. Within this same cluster, we search for the other (non-primary) objects found in the input image and use the most dissimilar instance in the target histogram. Lastly, color transfer is applied to map the input image to our generated target histogram. This is repeated for each cluster, producing several variations on the input image. The term $w_{\text{obj}}$ refers to each object's pixel ratio in the input image.}
\label{DoD:fig:main}
\end{figure}

\paragraph{Contribution}~We propose a data-driven framework to automatically recolor an input image that incorporates information about the color distributions of objects present in the image. Specifically, we show how to model the color distributions of different object classes from thousands of images with labeled objects. Given a new input image and its object segmentation, we outline a procedure to produce a diverse set of recolored images. We show that our results produce compelling examples that provide more interesting variations than existing methods.

\begin{figure}[b]
\includegraphics[width=\linewidth]{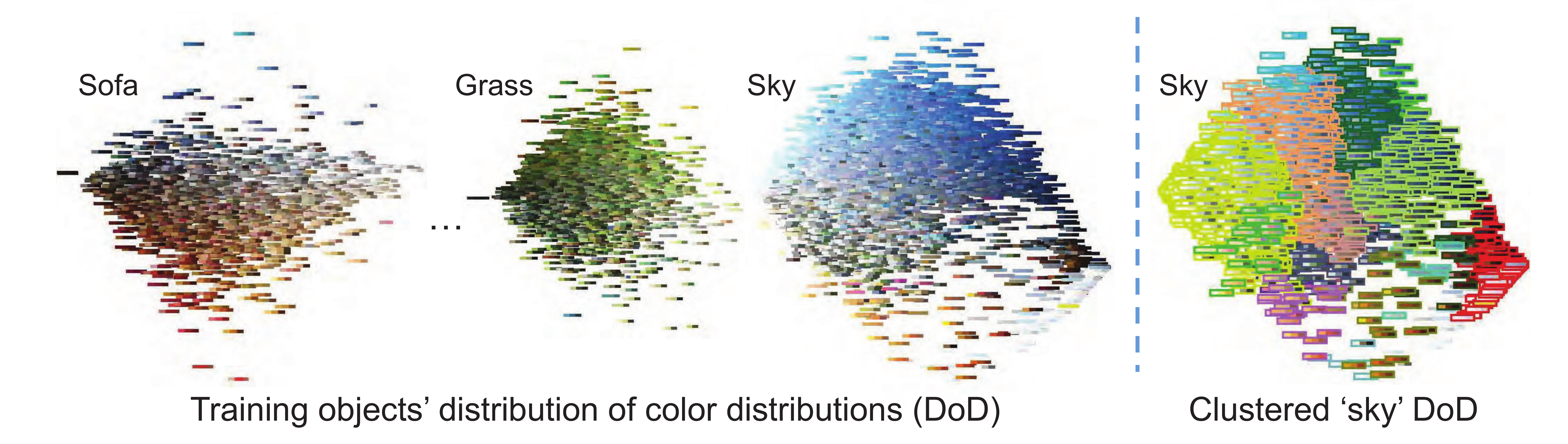}
\vspace{-7mm}
\caption[For each object class in our training data, we extract each instance of an object of that class that appears in the training images and represent its color appearance as a distribution of $k$ colors. We then construct a distribution of color distributions for each object class, termed a DoD (distribution of color distributions).]{For each object class in our training data, we extract each instance of an object of that class that appears in the training images and represent its color appearance as a distribution of $k$ colors in the form of a color palette.  We then construct a distribution of color distributions for each object class, termed a DoD (distribution of color distributions).  The DoD is computed by clustering the color palettes into $n$ clusters.}
\label{DoD:fig:DoD}
\end{figure}

\begin{figure}
\includegraphics[width=\linewidth]{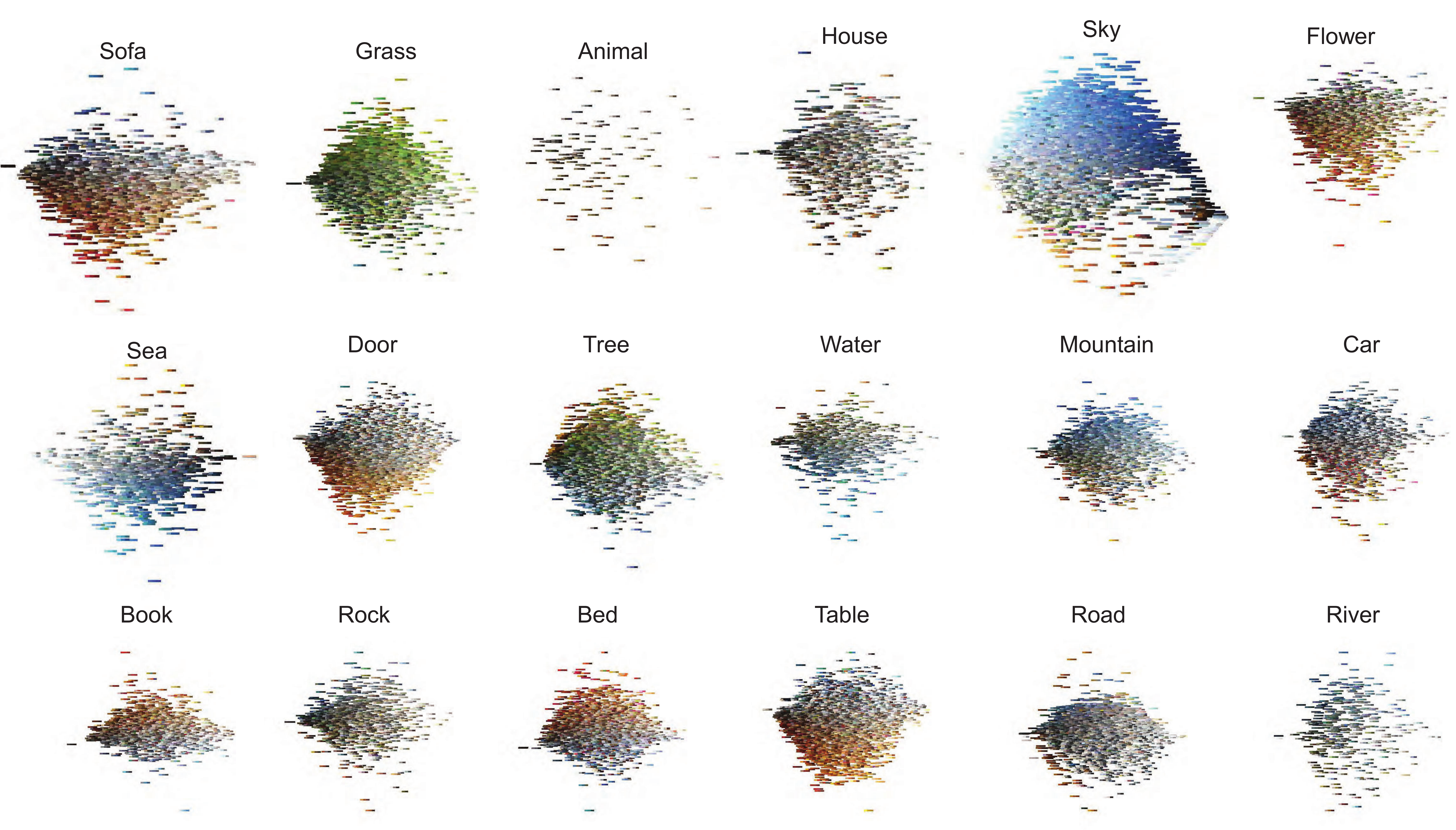}
\vspace{-7mm}
\caption[Examples of the distribution of color distributions (DoD) of different object classes.]{Examples of the distribution of color distributions (DoD) of different object classes.}
\label{DoD:fig:DoDExamples}
\end{figure}

\section{Methodology}

Figure \ref{DoD:fig:main} shows a diagram of our procedure. We start by describing the data preparation followed by the details of our method.

\subsection{Training Data}\label{DoD:sec:trainingdata}

\noindent Our method requires a large source of images with labeled object masks in order to build a color distribution for each class (type) of objects. To this end, we use the MIT Scene Parsing Benchmark (SPB) \cite{zhou2017scene}, which contains 20,210 images with pixel masks for 150 different object classes.

For each object class $b$, we extract the RGB pixel values for all object instances for this class in the training images. The object class is represented as a set of color distributions $\mathcal{C}_b$ = $\{\mathbf{c}_b^{(m)}\}_{m=1}^M$, where $M$ is the number of training images containing an instance of the object of class $b$.  An individual object instance's color distribution $\mathbf{c}_b^{(i)}$ is modeled as a color palette, with $k$ colors, generated by the method proposed in \cite{chang2015palette}.  The color palette representation also maintains the ratio of the number of pixels associated with each color in the palette.

\begin{figure}[b]
\includegraphics[width=\linewidth]{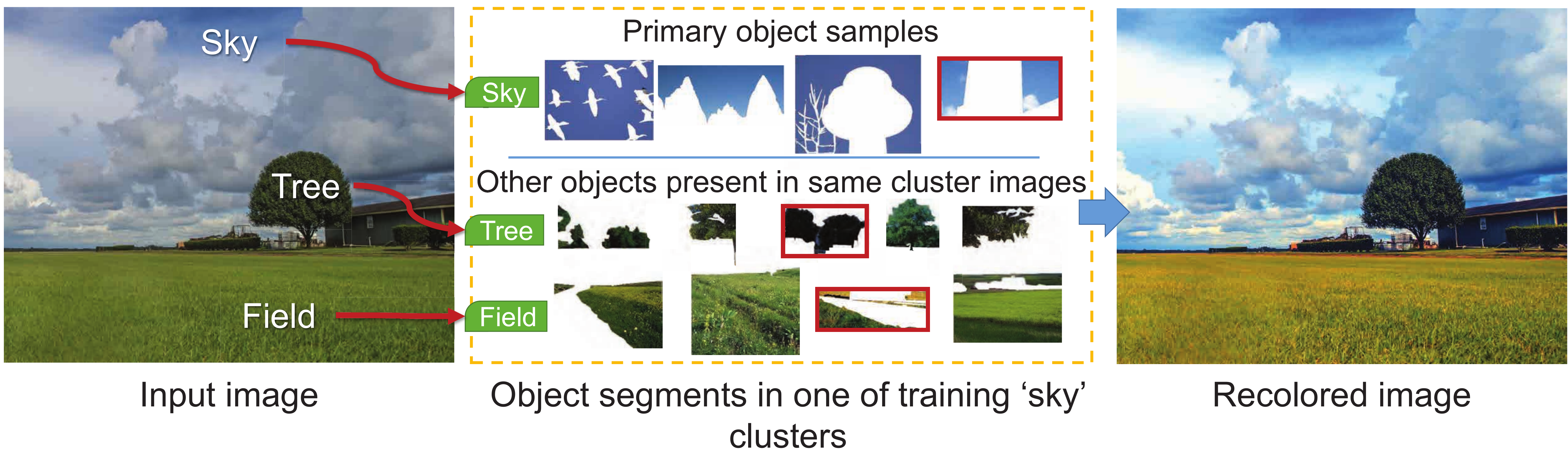}
\vspace{-7mm}
\caption[For each cluster of the primary object, we search for the most dissimilar training samples of the input image's semantic objects.]{For each cluster of the primary object, we search for the most dissimilar training samples (highlighted with red borders) of the input image's semantic objects.}
\label{DoD:fig:cluster}
\end{figure}

We use the set $\mathcal{C}_b$ to build a ``distribution of color distributions'' (DoD) for each object class $b$. The DoD of an object $b$ is generated by calculating the earth mover's distance (EMD)~\cite{rubner2000earth} between each pair of color palettes $\mathbf{c}_b^{(i)}$ and $\{\mathbf{c}_b^{(j)}\}_{j=1}^M$ of object instances, where $i = [1,...,M]$ and  $j\neq i$. Based on the EMDs between all color distributions in $\mathcal{C}_b$, we generate $n$ clusters of the color distributions using an agglomerative (bottom-up) hierarchical clustering with Ward's minimum variance \cite{ward1963hierarchical}.

By clustering all individual objects in the training images of object class $b$ based on their color palettes, each object instance associated with a cluster shares a similar overall color appearance. Figure \ref{DoD:fig:DoD} shows a visualization of this clustering procedure.  Figure \ref{DoD:fig:DoDExamples} shows more examples of the DoD of different object classes to provide an idea the different color appearances present in te DoDs. Within each cluster, we maintain a list of all other objects that appeared in the training images.  This latter point is important as it gives us a way to find other objects that have appeared in the images for a particular object class $b$.

 \begin{algorithm}[t]
 \caption{Given an input image $\mathbf{I}$ and its object mask $\mathbf{M}$, we generate $n$ recolored images ${\{\mathbf{R}_{i}\}}_{i=1}^n$.\label{DoD:algo}}
 \begin{algorithmic}

 \STATE $p \gets \text{select the primary object}$
 
 \STATE \textbf{for} each object $j$ in
 $\mathbf{M}$, \textbf{do}:
 \STATE $\text{  }$ $\mathbf{c}_{\text{obj}_j}^{\text{input}} \gets \text{get color palette (CP) of object } j \text{ in } \mathbf{I}$
 \STATE \textbf{end}
 
 \STATE \textbf{for} each training cluster $i$ of primary object $p$, \textbf{do}:
 
 \STATE $\text{  }$ $\mathbf{H}^{(i)} \gets \text{\{\}}$ \texttt{\color{blue}{//Initialize empty target histogram}}
 
 \STATE $\text{  }$ \textbf{for} each object $j$ in
 $\mathbf{M}$, \textbf{do}:

 \STATE $\text{  }\text{  }\text{  }$ \textbf{if} {$\mathcal{C}_{\text{obj}_j}^{\text{clust}(i)} \text{is emtpy}$}, \textbf{then}
 \STATE $\text{  }\text{  }\text{  }\text{  }\text{  }\text{  }$ $\mathbf{S}_{\text{obj}_j} \gets \mathbf{I}_{\text{obj}_j}$ \texttt{\color{blue}{//Original input object pixels}}
 
 \STATE $\text{  }\text{  }\text{  }$  \textbf{else}
 
 \STATE $\text{  }\text{  }\text{  }\text{  }\text{  }\text{  }$ $\mathbf{d}_{\text{obj}_j} \gets \text{calculate EMD between } \mathbf{c}_{\text{obj}_j}^{\text{input}} \text{ and CPs in } \mathcal{C}_{\text{obj}_j}^{\text{clust}(i)}$
 
 \STATE $\text{  }\text{  }\text{  }\text{  }\text{  }\text{  }$ $\mathbf{S}_{\text{obj}_j} \gets \text{retrieve training sample with max value in } \mathbf{d}_{\text{obj}_j}$
 
 \STATE $\text{  }\text{  }\text{  }$ \textbf{end}
 
 \STATE $\text{  }\text{  }\text{  }$ $\mathbf{H}_{\text{obj}_j}^{(i)} \gets \text{generate color histogram of } \mathbf{S}_{\text{obj}_j}$.
 
 \STATE $\text{  }\text{  }\text{  }$ $\mathbf{H}^{(i)} \gets \text{add } \mathbf{H}_{\text{obj}_j}^{(i)} \text{ to } \mathbf{H}^{(i)}$ .
 
 \STATE $\text{  }\text{  }$ \textbf{end}
 
 \STATE $\text{  }\text{  }$ $\mathbf{R}_{i} \gets \text{color mapping} \left(\mathbf{H}^{\text{input}}, \mathbf{H}^{(i)}\right)$
 
 \STATE \textbf{end}

 \end{algorithmic}
 \end{algorithm}

\subsection{Recoloring Procedure}\label{DoD:sec:constructhistogram}

Given an input image $\mathbf{I}$, we compute its semantic object mask $\mathbf{M}$ using RefineNet \cite{lin2017refinenet}.  Note that this mask can be noisy (i.e., RefineNet reports $\sim 0.79$ pixel-wise accuracy rate on the SPB dataset).
The objects in $\mathbf{I}$ are ranked based on each object's pixel ratio in $\mathbf{I}$ ($w_{\text{obj}}$) and the ratio of the training samples with that object class ($r_{\text{obj}}$) in the training dataset.  Both of these terms are normalized to range between 0 and 1. We select the primary object $p$ as the object with the maximum score $w_{\text{obj}_p} + r_{\text{obj}_p}$.

For the primary object and all other objects in $\mathbf{I}$, we compute their color distributions (i.e., a color palette with associated pixel weights). We then visit each cluster of the primary object's class as described in Sec.~\ref{DoD:sec:trainingdata}.  Within a cluster, we find the most dissimilar object instance based on the EMD of the primary object's color palette and each example in the cluster.  The colors of this dissimilar object instance are added to the target histogram.  Within the same cluster, this procedure is repeated for all other objects present in the image.   Again, we seek the most dissimilar color palette to the input image's object within the cluster using the EMD metric.  Figure \ref{DoD:fig:cluster} provides an illustrative example. These dissimilar colors from the training data are added to the target histogram. If an object found in the input image does not exist in a cluster, we copy the input object's colors to the target histogram. Note that the construction of the target histogram was performed by a weighted summation based on each object's pixel ratio in $\mathbf{I}$. We have purposely chosen to use objects with the most dissimilar colors to provide notably different recolored images; however, we note the criteria for selecting target object instances can be adjusted to employ different strategies.

Once a target histogram has been constructed, we map the input histogram to the target histogram. This mapping produces the recolored result.  In our experiments, we used the method by Pitie and Kokaram~\cite{Pitie2007} to transfer the input image's color histogram to the target color histogram. Then, we scale any out-of-gamut pixels to fall in the range [0-255].

The procedure described above is repeated for each cluster in the primary object's class, producing $n$ output images.  Alg.~\ref{DoD:algo} provides pseudocode for our procedure.
\vspace{-3mm}
\section{Results}

In our experiments, the number of clusters used in the DoD is set to $n=20$, the input and target color histogram bins are resized to resolution $32\!\times\!32\!\times\!32$,
and our color palettes are fixed to have $k=20$ colors. We show color palettes with three colors in Fig. \ref{DoD:fig:main} and Fig. \ref{DoD:fig:DoD} to simplify the visualization.

\begin{figure}[t]
\includegraphics[width=\linewidth]{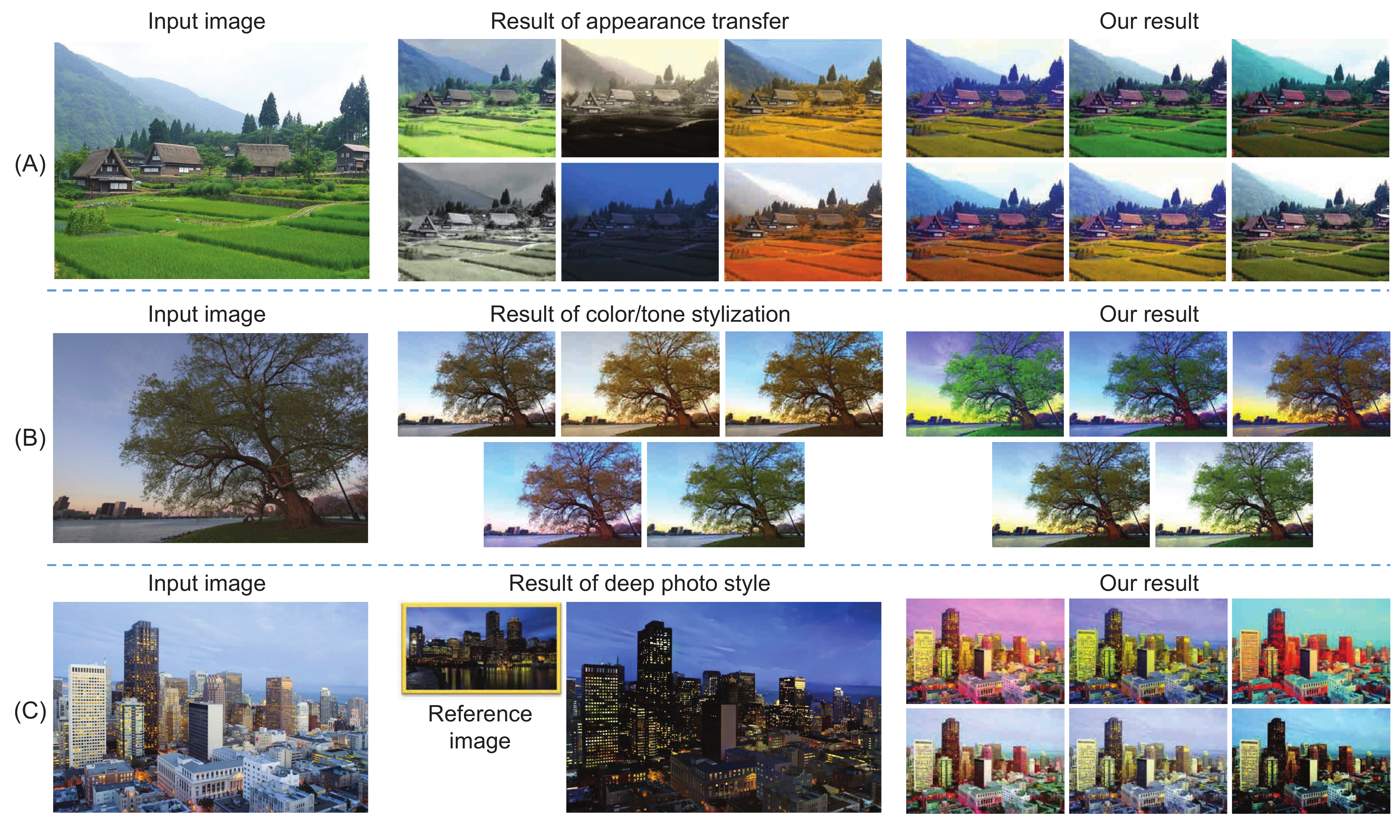}
\vspace{-7mm}
\caption[Comparisons with existing style transfer methods.]{Comparisons with existing style transfer methods. For each input image (first column), we show results of other methods (second column) and our results (third column). The other methods are: (A) appearance transfer \cite{laffont2014transient}, (B) auto content-aware color and tone stylization \cite{lee2016automatic}, and (C) a reference-based deep style transfer \cite{LuanCVPR17}.}
\label{DoD:fig:comparisons}
\end{figure}

\begin{figure} [t]
\includegraphics[width=\linewidth]{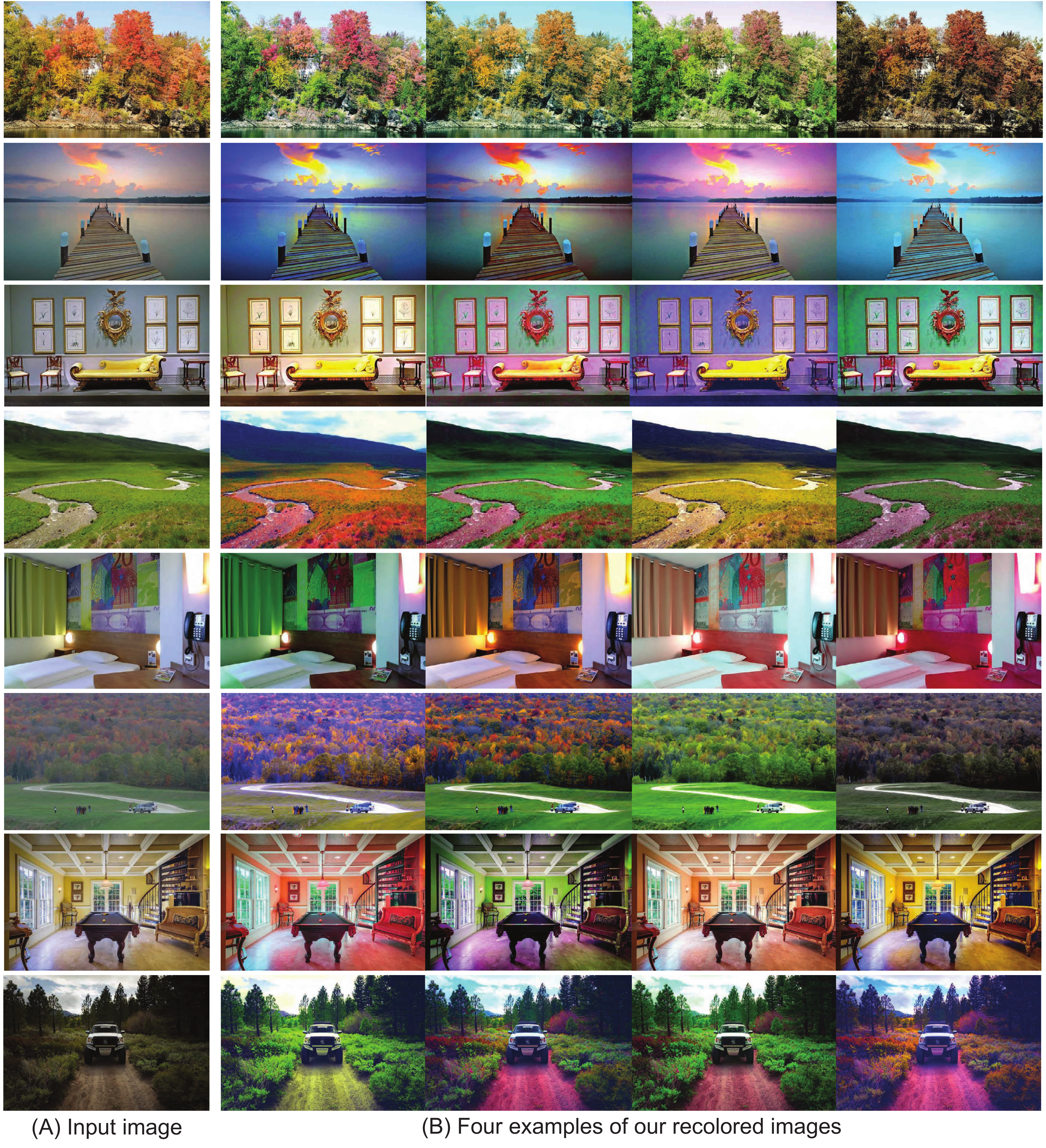}
\vspace{-7mm}
\caption[Qualitative results of the proposed method.]{Qualitative results of the proposed method. (A) Input image. (B) Four examples of our recolored images.}
\label{DoD:fig:result}
\end{figure}

\begin{figure} [t]
\includegraphics[width=\linewidth]{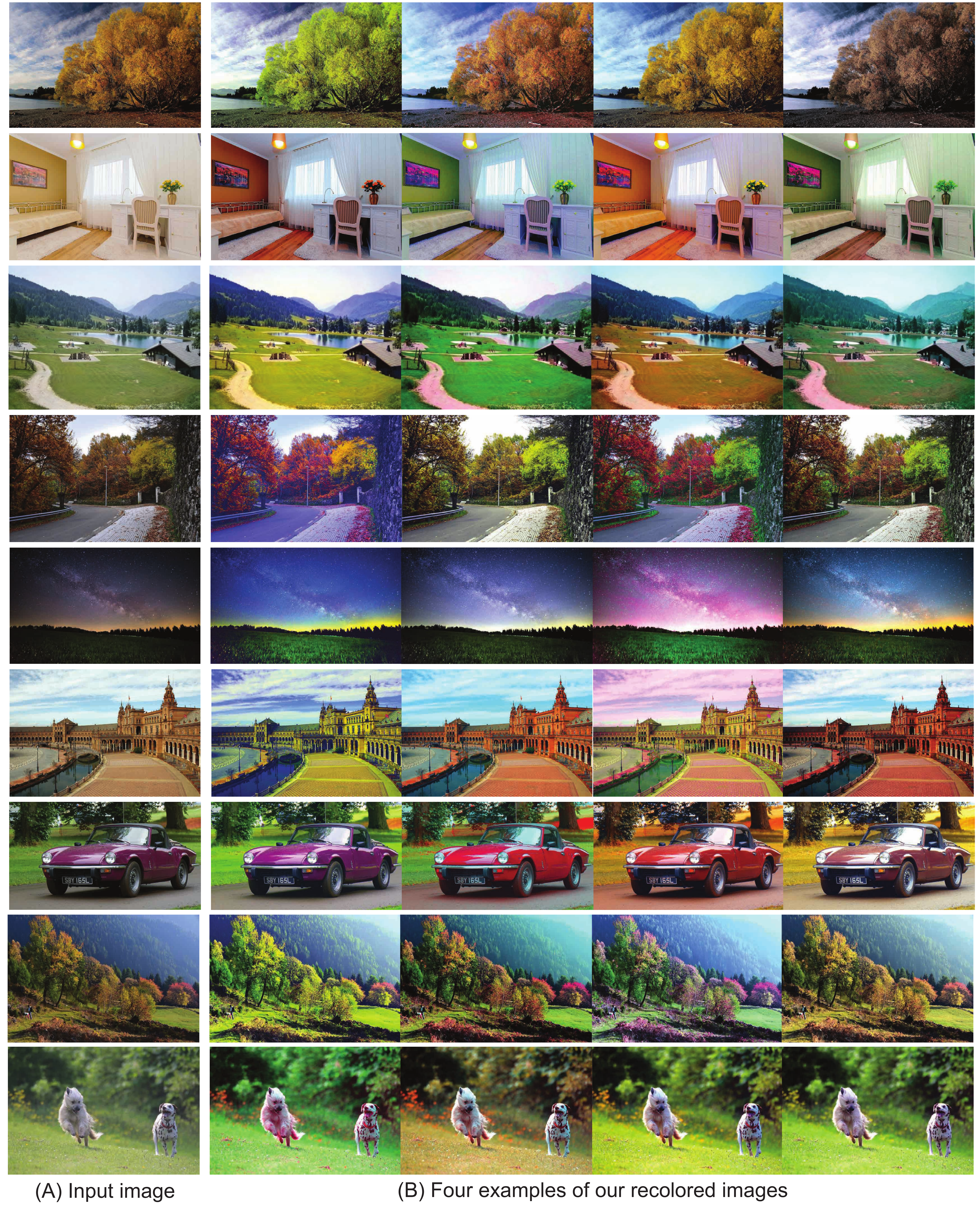}
\vspace{-7mm}
\caption[Additional qualitative results of the proposed method.]{Additional qualitative results of the proposed method. (A) Input image. (B) Four examples of our recolored images.}
\label{DoD:fig:result1}
\end{figure}

Figure \ref{DoD:fig:comparisons} shows comparisons between results of our method and three other methods. The first method \cite{laffont2014transient} was proposed to transfer the appearance of outdoor images to different scene appearances specified by a set of attributes selected by the user. The second method \cite{lee2016automatic} is an auto color/style transfer method which employs deep features extracted from a pre-trained CNN on the ImageNet dataset \cite{deng2009imagenet} in order to produce content-aware color stylization. The third method \cite{LuanCVPR17} is a CNN-based method which requires a reference image in order to transfer its style to the input image. The results show that our method produces compelling results without requiring any user interaction or reference images. Additional qualitative results are shown in Figs. \ref{DoD:fig:result} and \ref{DoD:fig:result1}.

\section{Summary}

We have proposed an automated data-driven method to generate recolored images. Our method leverages semantic object information of the input image and these objects' associated color distributions that have been modeled from thousands of training images. We demonstrate the effectiveness of the proposed method on several input images and compare our results with other strategies to produce color variations.


\chapter{Controlling Colors via Color Histograms \label{ch:ch15}}
In Chapter \ref{ch:ch14}, we have presented an auto recoloring framework based on object color distribution. In this chapter, we propose a generative adversarial network (GAN)-based method for image recoloring. Our method is not only designed to recolor real images but also to control colors of GAN-generated images. While GANs can successfully produce high-quality images, they can be challenging to control.\  Simplifying GAN-based image generation is critical for their adoption in graphic design and artistic work.\  This goal has led to significant interest in methods that can intuitively control the appearance of images generated by GANs.\ In this chapter, we present HistoGAN\footnote{This work was published in \cite{afifi2020histogan}: Mahmoud Afifi, Marcus A. Brubaker, Michael S. Brown. HistoGAN: Controlling Colors of GAN-Generated and Real Images via Color Histograms. In IEEE Conference on Computer Vision and Pattern Recognition (CVPR), 2021.}, a color histogram-based method for controlling GAN-generated images' colors.\ We focus on color histograms as they provide an intuitive way to describe image color while remaining decoupled from domain-specific semantics. Specifically, we introduce an effective modification of the recent StyleGAN architecture~\cite{karras2020analyzing} to control the colors of GAN-generated images specified by a target color histogram feature.\ We then describe how to expand HistoGAN to recolor real images.\  For image recoloring, we jointly train an encoder network along with HistoGAN.\ The recoloring model, ReHistoGAN, is an unsupervised approach trained to encourage the network to keep the original image's content while changing the colors based on the given target histogram. We demonstrate that this histogram-based approach offers a better way to control GAN-generated and real images' colors while producing more compelling results compared to existing alternative strategies. The source code and dataset of this work are available on GitHub: \href{https://github.com/mahmoudnafifi/HistoGAN}{https://github.com/mahmoudnafifi/HistoGAN}.

\begin{figure}[!t]
\includegraphics[width=\textwidth]{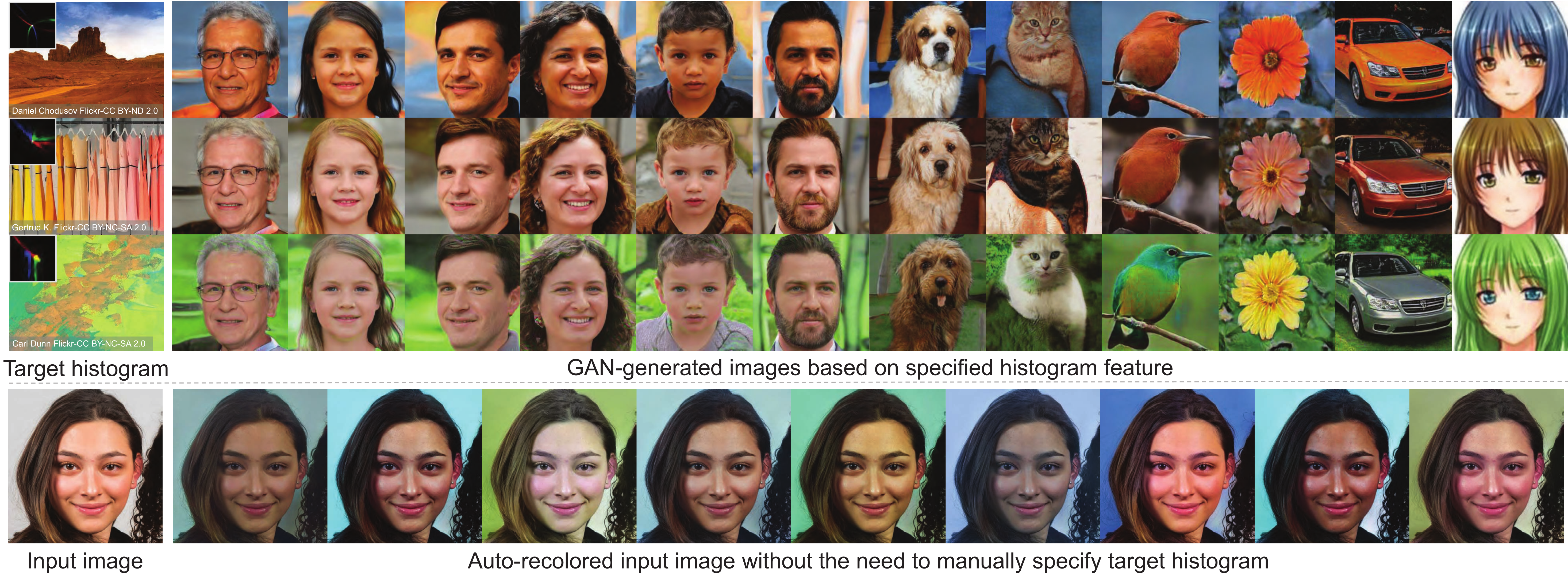}
\vspace{-7mm}
\caption[HistoGAN is a generative adversarial network (GAN) that learns to manipulate image colors based on histogram features.]{HistoGAN is a generative adversarial network (GAN) that learns to manipulate image colors based on histogram features. Top: GAN-generated images with color distributions controlled via target histogram features (left column). Bottom: Results of ReHistoGAN, an extension of HistoGAN to recolor real images, using sampled target histograms.\label{histogan:fig:teaser}}
\end{figure}

\section{Introduction} \label{histogan:sec.intro}

Color histograms are an expressive and convenient representation of an image's color content. Color histograms are routinely used by conventional color transfer methods (e.g., \cite{reinhard2001color, xiao2006color, nguyen2014illuminant, faridul2016colour}).
These color transfer methods aim to manipulate the colors in an input image to match those of a target image, such that the images share a similar ``look and feel''. As discussed in Chapter\ \ref{ch:ch3} there are various forms of color histograms used in the color transfer literature to represent the color distribution of an image, such as a direct 3D histogram~\cite{reinhard2001color, xiao2006color,faridul2016colour}, color palettes \cite{chang2015palette, zhang2017palette} or color triads \cite{shugrina2020nonlinear}. Despite the effectiveness of color histograms for color transfer, recent deep learning methods almost exclusively rely on image-based examples to control colors.
While image exemplars impact the final colors of generative adversarial network (GAN)-generated images and deep recolored images, they also affect other style attributes, such as texture information and tonal values~\cite{gatys2015neural, gatys2016image, johnson2016perceptual, ulyanov2016instance, isola2017image, luan2017deep, sheng2018avatar}.
Consequently, the quality of the results produced by these methods often depends on the semantic similarity between the input and target images, or between a target image and a particular domain~\cite{sheng2018avatar, he2019progressive}.

In this chapter, our attention is focused explicitly on controlling only the color attributes of images---this can be considered a sub-category of image style transfer. Specifically, our method does not require shared semantic content between the input/GAN-generated images and a target image or guide image. Instead, our method aims to assist the deep network through color histogram information only.  With this motivation, we first explore using color histograms to control the colors of images generated by GANs.

\begin{figure}[!t]
\begin{center}
\includegraphics[width=0.98\linewidth]{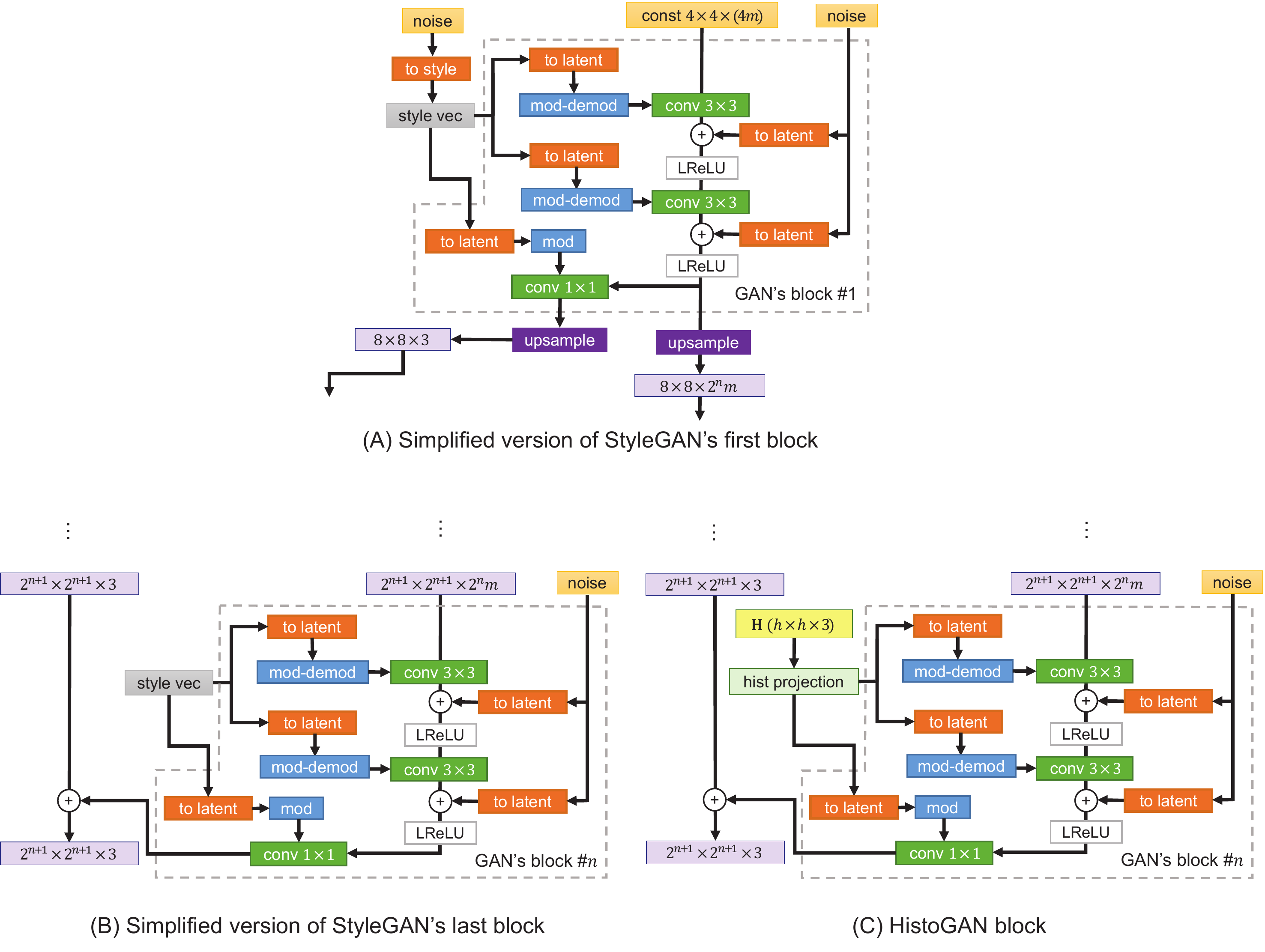}
\end{center}
\vspace{-7mm}
\caption[We inject our histogram into StyleGAN \cite{karras2020analyzing} to control the generated image colors.]{We inject our histogram into StyleGAN \cite{karras2020analyzing} to control the generated image colors. (A) and (B) are simplified versions of the StyleGAN's first and last blocks. We modified the last two blocks of the StyleGAN by projecting our histogram feature into each block's latent space, as shown in (C). The parameter $m$ controls the capacity of the model.}
\label{histogan:fig:GAN-design}	
\end{figure}

\paragraph{Controlling Color in GAN-Generated Images}

GANs are often used as ``black boxes'' that can transform samples from a simple distribution to a meaningful domain distribution without an explicit ability to control the details/style of the generated images \cite{goodfellow2014generative, radford2015unsupervised, karras2017progressive, arjovsky2017wasserstein, liu2019wasserstein}.
Recently, methods have been proposed to control the style of the GAN-generated images.
For example, StyleGAN~\cite{karras2019style, karras2020analyzing} proposed the idea of ``style mixing'', where different latent style vectors are progressively fed to the GAN to control the style and appearance of the output image.
To transfer a specific style in a target image to GAN-generated images, an optimization process can be used to project the target image to the generator network's latent space to generate images that share some properties with the target image~\cite{abdal2019image2stylegan, karras2020analyzing}.
However, this process requires expensive computations to find the latent code of the target image.
Another direction is to jointly train an encoder-generator network to learn this projection \cite{pidhorskyi2020adversarial, li2020mixnmatch, choi2020stargan}.
More recently, methods have advocated different approaches to control the output of GANs, such as using the normalized flow \cite{abdal2020styleflow}, latent-to-domain-specific mapping  \cite{choi2020stargan}, deep classification features \cite{shocher2020semantic}, few-shot image-to-image translation \cite{saito2020coco}, and a single-image training strategy \cite{shaham2019singan}.
Despite the performance improvements, most of these methods are limited to work with a single domain of both target and GAN-generated images \cite{li2020mixnmatch, pidhorskyi2020adversarial}.

We seek to control GAN-generated images using color histograms as our specified representation of image style.
Color histograms enable our method to accept target images taken from \textit{any} arbitrary domain.
Figure \ref{histogan:fig:teaser}-top shows GAN-generated examples using our method.
As shown in Fig.~\ref{histogan:fig:teaser}, our generated images share the same color distribution as the target images without being restricted to, or influenced by, the semantic content of the target images.

\paragraph{Recoloring Real Images}
In addition to controlling the GAN-generated images, we seek to extend our approach to perform image recoloring within the GAN framework. In this context, our method accepts a real input image and a target histogram to produce an output image with the fine details of the input image but with the same color distribution given in the target histogram.
Our method is trained in a fully unsupervised fashion, where no ground-truth recolored image is required.
Instead, we propose a novel adversarial-based loss function to train our network to extract and consider the color information in the given target histogram while producing realistic recolored images.
One of the key advantages of using the color histogram representation as our target colors can be shown in Fig.\ \ref{histogan:fig:teaser}-bottom, where we can {\it automatically recolor} an image without directly having to specify a target color histogram.
As discussed in Chapter \ref{ch:ch3}, auto-image recoloring is a less explored research area with only a few attempts in the literature (e.g., \cite{laffont2014transient, deshpande2017learning,  anokhin2020high}).

\section{HistoGAN} \label{histogan:sec.method}

We begin by describing the histogram feature used by our method (Sec.\ \ref{histogan:subsec.histoblock}). Afterwards, we discuss the proposed modification to StyleGAN \cite{karras2020analyzing} to incorporate our histogram feature into the generator network (Sec.\ \ref{histogan:subsec.method-coloring-GAN-images}). Lastly, we explain how this method can be expanded to control colors of real input images to perform image recoloring~(Sec.\ \ref{histogan:subsec.method-recoloring}).

\subsection{Histogram feature} \label{histogan:subsec.histoblock}
The histogram feature used by HistoGAN is borrowed from the color constancy literature (see Chapters \ref{ch:ch5}--\ref{ch:ch7})  and is constructed to be a differentiable histogram of colors in the log-chroma space due to better invariance to illumination changes~\cite{finlayson2001color, eibenberger2012importance}. The feature is a 2D histogram of an image's colors projected into a log-chroma space. This 2D histogram is parameterized by $uv$ and conveys an image's color information while being more compact than a typical 3D histogram defined in RGB space.   A log-chroma space is defined by the intensity of one channel, normalized by the other two, giving three possible options of how it is defined.
Instead of selecting only one such space, all three options can be used to construct three different histograms which are combined together into a histogram feature, $\mat{H}$, as an $h\!\times\!h\!\times 3$ tensor.

As explained in Chapters \ref{ch:ch5}--\ref{ch:ch7}, the histogram is computed from a given input image, $\I$, by first converting it into the log-chroma space.
For instance, selecting the $\cR$ color channel as primary and normalizing by $\cG$ and $\cB$ gives:
\begin{equation}
\I_{uR}(\pX) = \log{\left(\frac{ \I_{\cR}(\pX)+ \epsilon}{\I_{\cG}(\pX)+ \epsilon}\right)} \textrm{ , }  \I_{vR}(\pX) = \log{\left(\frac{ \I_{\cR}(\pX)+ \epsilon}{\I_{\cB}(\pX)+ \epsilon}\right)},
\end{equation}
where the $\cR, \cG, \cB$ subscripts refer to the color channels of the image $\I$, $\epsilon = 10^{-8}$ is a small constant added for numerical stability, $\pX$ is the pixel index, and $(uR, vR)$ are the $uv$ coordinates based on using $\cR$ as the primary channel.
The other components $\I_{uG}$, $\I_{vG}$, $\I_{uB}$, $\I_{vB}$ are computed similarly by projecting the $\cG$ and $\cB$ color channels to the log-chroma space.
In Chapter \ref{ch:ch7}, the RGB-$uv$ histogram is computed by thresholding colors to a bin and computing the contribution of each pixel based on the intensity $\Iy(\pX) = \sqrt{\I_{\cR}^{2}(\pX) + \I_{\cG}^{2}(\pX) + \I_{\cB}^{2}(\pX)}$.  In order to make the representation differentiable, we replaced the thresholding operator with a kernel weighted contribution to each bin in Chapter \ref{ch:ch5}.  The final unnormalized histogram is computed as:
\begin{equation}
\mat{H}(u,v,c) \propto \sum_{\pX} k(\I_{uc}(\pX), \I_{vc}(\pX), u, v) \I_{y}(\pX),
\end{equation}
where $c \in {\{\textrm{R}, \textrm{G}, \textrm{B}\}}$ and $k(\cdot)$ is a pre-defined kernel.
While a Gaussian kernel was originally used in Chapter \ref{ch:ch5}, we found that the inverse-quadratic kernel significantly improved training stability in our current task.  The inverse-quadratic kernel is defined as:
\begin{equation}
k(\I_{uc}, \I_{vc}, u, v) = \left(1+\left(\left| \I_{uc} - u \right|/\tau\right)^2\right)^{-1} \times \left(1 + \left(\left| \I_{vc} - v \right|/\tau\right)^2\right)^{-1},
\end{equation}
\noindent
where  $\tau$ is a fall-off parameter to control the smoothness of the histogram's bins.
Finally, the histogram feature is normalized to sum to one, i.e., $\sum_{u,v,c}\mat{H}(u,v,c)=1$.

\begin{figure}[!t]
\includegraphics[width=\linewidth]{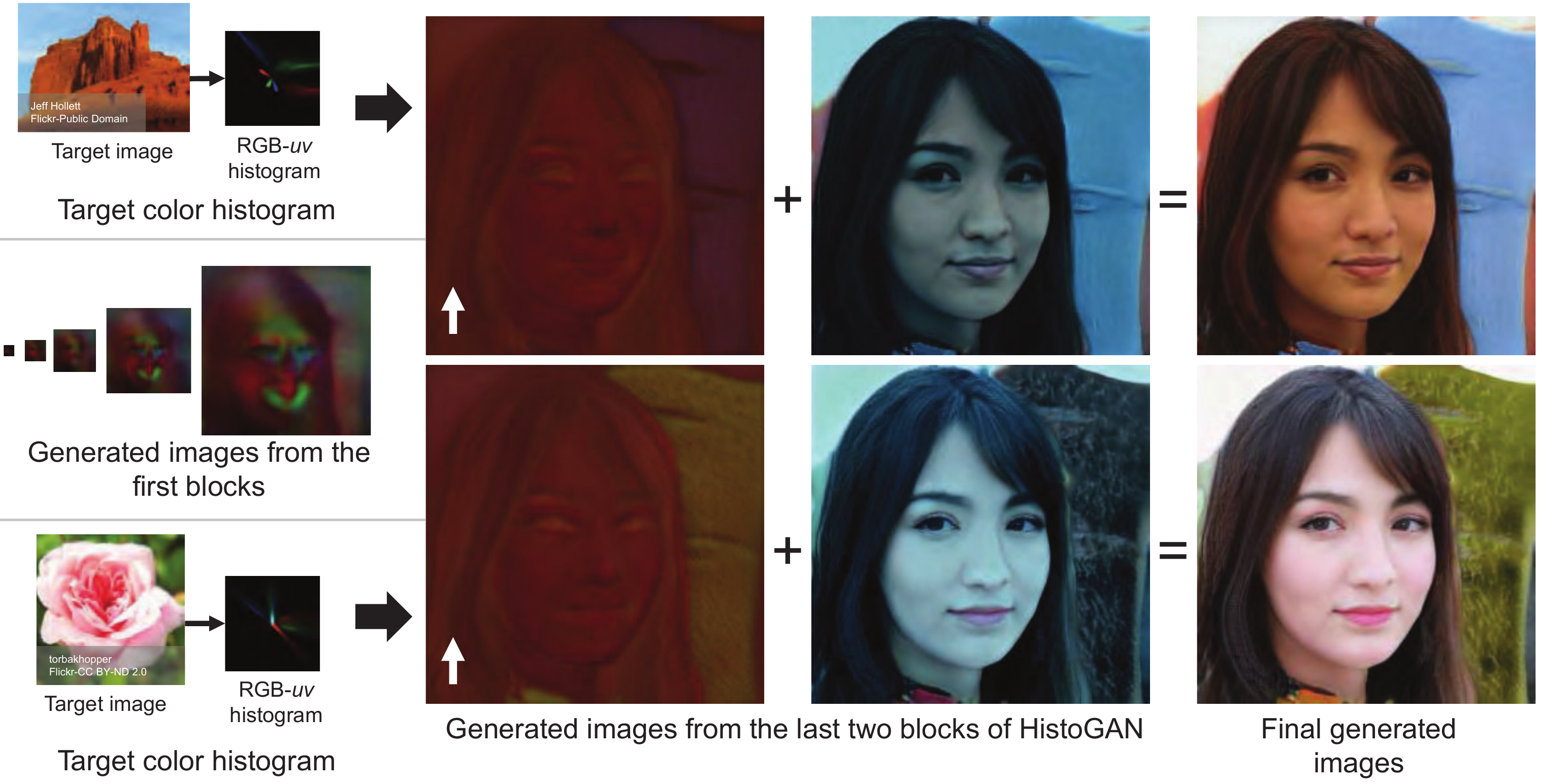}
\vspace{-7mm}
\caption[Progressively generated images using the HistoGAN modifications.]{Progressively generated images using the HistoGAN modifications.}
\label{histogan:fig:analysis}
\end{figure}

\subsection{Color-controlled Image Generation}
\label{histogan:subsec.method-coloring-GAN-images}

Our histogram feature is incorporated into an architecture based on StyleGAN~\cite{karras2020analyzing}.
Specifically, we modified the original design of StyleGAN (Fig.\ \ref{histogan:fig:GAN-design}-[A] and [B]) such that we can ``inject'' the histogram feature into the progressive construction of the output image.
The last two blocks of the StyleGAN (Fig.\ \ref{histogan:fig:GAN-design}-[B]) are modified by replacing the fine-style vector with the color histogram feature.
The histogram feature is then projected into a lower-dimensional representation by a ``histogram projection'' network (Fig.\ \ref{histogan:fig:GAN-design}-[C]).
This network consists of eight fully connected layers with a leaky ReLU (LReLU) activation function \cite{maas2013rectifier}.
The first layer has 1,024 units, while each of the remaining seven layers has 512.
The ``to-latent'' block, shown in orange in Fig.\ \ref{histogan:fig:GAN-design}, maps the projected histogram to the latent space of each block.  This ``to-latent'' block consists of a single fc layer with $2^n m$ output neurons, where $n$ is the block number, and $m$ is a parameter used to control the entire capacity of the network.

To encourage generated images to match the target color histogram, a color matching loss is introduced to train the generator.
Because of the differentiability of our histogram representation, the loss function, $C(\mat{H}_g,\mat{H}_t)$, can be any differentiable metric of similarity between the generated and target histograms $\mat{H}_g$ and $\mat{H}_t$, respectively.
For simplicity, we use the Hellinger distance defined as:
\begin{equation}\label{histogan:eq:hellinger-distance}
C\left(\mat{H}_g, \mat{H}_t\right) = \frac{1}{\sqrt{2}} \left\Vert \mat{H}_g^{1/2} - \mat{H}_t^{1/2} \right\Vert_2,
\end{equation}
where $\Vert \cdot \Vert_2$ is the standard Euclidean norm and $\mat{H}^{1/2}$ is an element-wise square root.

This color-matching histogram loss function is combined with the discriminator to give the generator network loss:
\begin{equation}\label{histogan:eq:gan-loss}
{\mathcal{L}}_g = D\left(\I_g\right) + \alpha C\left(\mat{H}_g, \mat{H}_t\right),
\end{equation}
where $\I_g$ is the GAN-generated image, $D\left(\cdot\right)$ is our discriminator network that produces a scalar feature given an image (see Appendix \ref{ch:appendix3} for more details), $\mat{H}_t$ is the target histogram feature (injected into the generator network), $\mat{H}_g$ is the histogram feature of $\I_g$, $C\left(\cdot\right)$ is our histogram loss function, and $\alpha$ is a scale factor to control the strength of the histogram loss term.

\begin{figure}[!t]
\includegraphics[width=\linewidth]{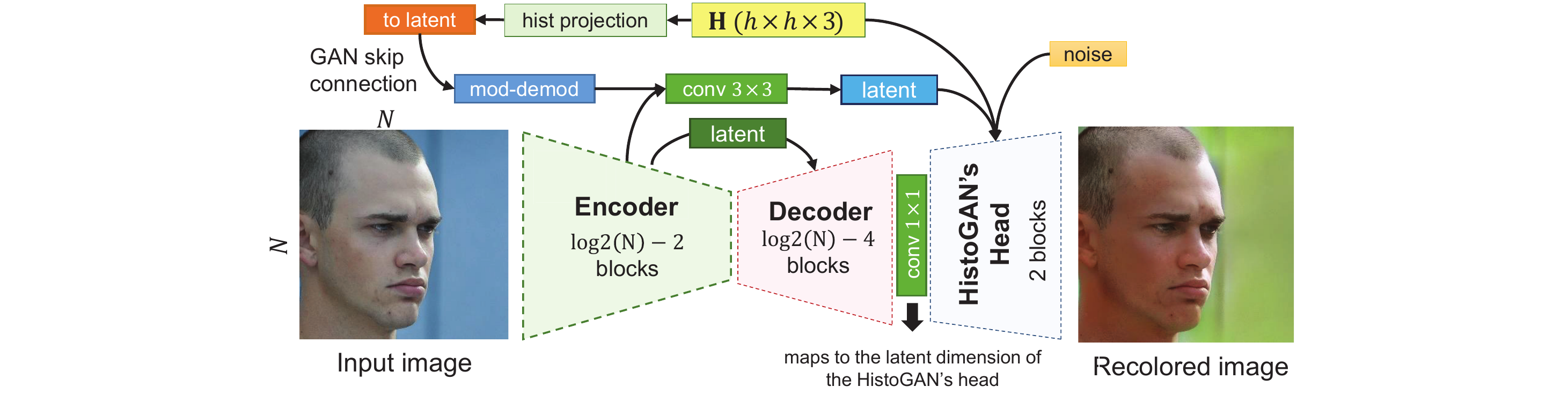}
\vspace{-7mm}
\caption[Our Recoloring-HistoGAN (ReHistoGAN) network.]{Our Recoloring-HistoGAN (ReHistoGAN) network. We map the input image into the HistoGAN's latent space using an encoder-decoder network with skip connections between each encoder and decoder blocks. Additionally, we pass the latent feature of the first two encoder blocks to our GAN's head after processing it with the histogram's latent feature.}
\label{histogan:fig:recoloring-design}
\end{figure}

As our histogram feature is computed by a set of differentiable operations, our loss function (Eqs. \ref{histogan:eq:hellinger-distance} and \ref{histogan:eq:gan-loss}) can be optimized using SGD.
During training, different target histograms $\mat{H}_t$ are required. To generate these for each generated image, we randomly select two images from the training set, compute their histograms $\mat{H}_1$ and $\mat{H}_2$, and then randomly interpolate between them. Specifically, for each generated image during training, we generate a random target histogram as follows:
\begin{equation}
\label{histogan:eq.target_hist}
\mat{H}_t = \delta \mat{H}_1 + \left(1 - \delta \right) \mat{H}_2,
\end{equation}
where $\delta \sim U(0,1)$ is sampled uniformly.

With this modification to the original StyleGAN architecture, our method can control the colors of generated images using our color histogram features.\ Figure~\ref{histogan:fig:analysis} shows the progressive construction of the generated image by HistoGAN.  As can be seen, the outputs of the last two blocks are adjusted to consider the information conveyed by the target histogram to produce output images with the same color distribution represented in the target histogram.

\begin{figure}[!t]
\includegraphics[width=\linewidth]{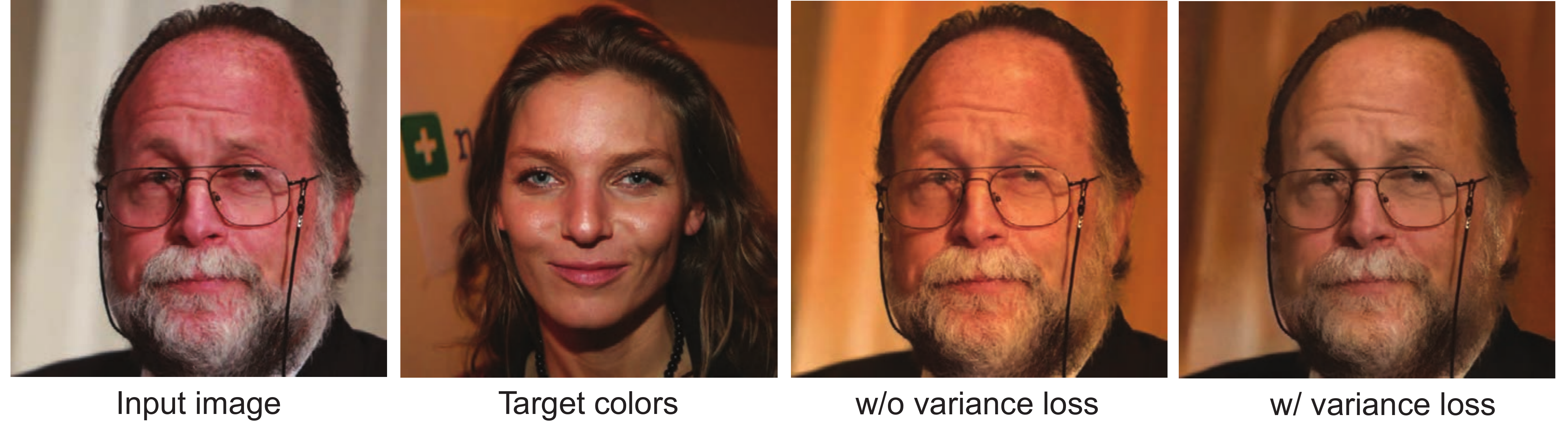}
\vspace{-7mm}
\caption[Results of training ReHistoGAN with and without the variance loss term described in Eq. \ref{histogan:eq.variance-loss}.]{Results of training ReHistoGAN with and without the variance loss term described in Eq. \ref{histogan:eq.variance-loss}.}
\label{histogan:fig:variance_loss}
\end{figure}

\begin{figure}[!t]
\includegraphics[width=\linewidth]{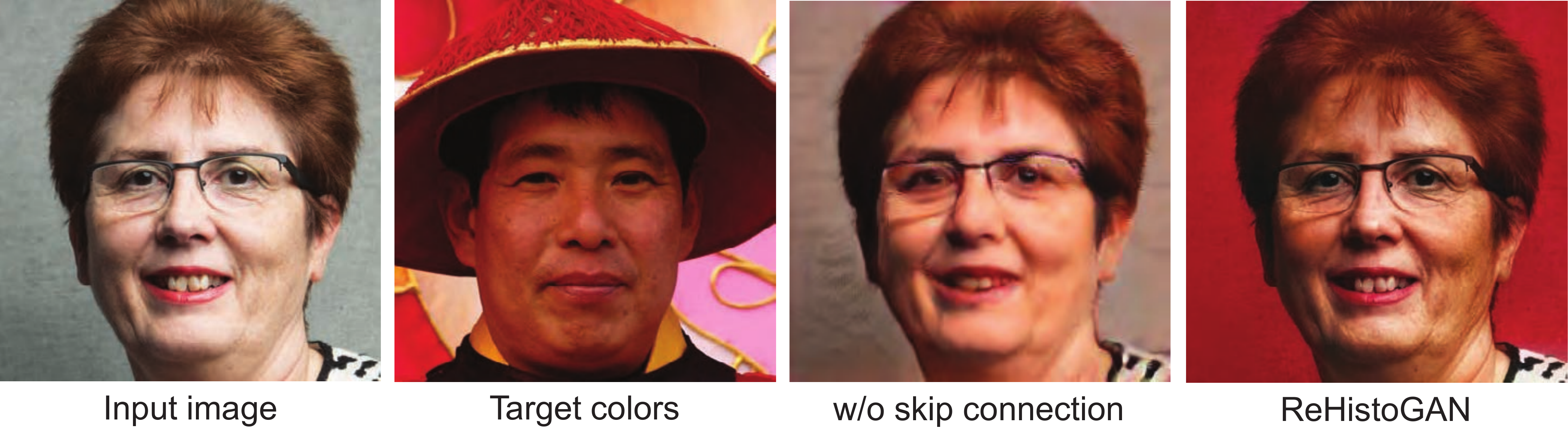}
\vspace{-7mm}
\caption[Results of image recoloring using the encoder-GAN reconstruction without skip connections and our ReHistoGAN using our proposed loss function.]{Results of image recoloring using the encoder-GAN reconstruction without skip connections and our ReHistoGAN using our proposed loss function.}
\label{histogan:fig:comparison_with_projection}
\end{figure}

\begin{figure}[!t]
\centering
\includegraphics[width=0.97\linewidth]{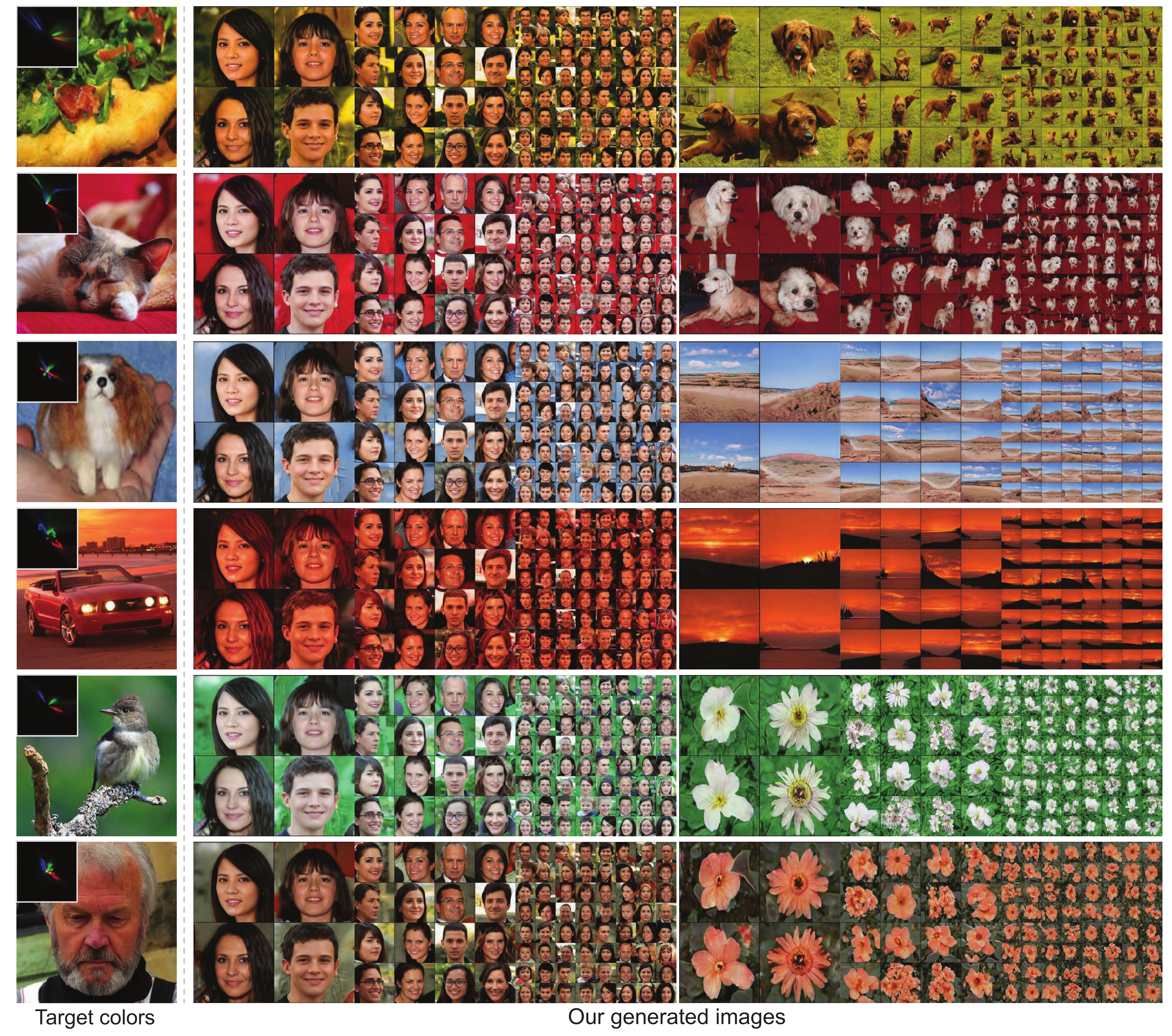}
\vspace{-2mm}
\caption[Images generated by HistoGAN.]{Images generated by HistoGAN. For each input image shown in the left, we computed the corresponding target histogram (shown in the upper left corner of the left column) and used it to control colors of the generated images in each row.}
\label{histogan:fig:GAN_results}
\end{figure}

\subsection{Image Recoloring} \label{histogan:subsec.method-recoloring}

We can also extend HistoGAN to recolor an input image, as shown in Fig.\ \ref{histogan:fig:teaser}-bottom. Recoloring an existing input image, $\I_i$, is not straightforward because the randomly sampled noise and style vectors are not available as they are in a GAN-generated scenario.  As shown in Fig.\ \ref{histogan:fig:analysis}, the head of HistoGAN (i.e., the last two blocks) are responsible for controlling the colors of the output image.
Instead of optimizing for noise and style vectors that could be used to generate a given image $\I_i$, we propose to train an encoding network that maps the input image into the necessary inputs of the head of HistoGAN.

With this approach, the head block can be given different histogram inputs to produce a wide variety of recolored versions of the input image.
We dub this extension the ``Recoloring-HistoGAN'' or ReHistoGAN for short.
The architecture of ReHistoGAN is shown in Fig.\ \ref{histogan:fig:recoloring-design}.
The ``encoder'' has a U-Net-like structure \cite{ronneberger2015u} with skip connections.
To ensure that fine details are preserved in the recolored image, $\I_r$, the early latent feature produced by the first two U-Net blocks are further provided as input into the HistoGAN's head through skip connections.

The target color information is passed to the HistoGAN head blocks as described in Sec.\ \ref{histogan:subsec.method-coloring-GAN-images}. Additionally, we allow the target color information to influence through the skip connections to go from the first two U-Net-encoder blocks to the HistoGAN's head.
We add an additional histogram projection network, along with a ``to-latent'' block, to project our target histogram to a latent representation.
This latent code of the histogram is processed by weight modulation-demodulation operations \cite{karras2020analyzing} and is then convolved over the skipped latent of the U-Net-encoder's first two blocks.
We modified the HistoGAN block, described in Fig.\ \ref{histogan:fig:GAN-design}, to accept this passed information (see Appendix \ref{ch:appendix3} for more information).
The leakage of the target color information helps ReHistoGAN to consider information from both the input image content and the target histogram in the recoloring process.

We initialize our encoder-decoder network using He's initialization \cite{he2015delving}, while the weights of the HistoGAN head are initialized based on a previously trained HistoGAN model (trained in Sec.\ \ref{histogan:subsec.method-coloring-GAN-images}).
The entire ReHistoGAN is then jointly trained to minimize the following loss function:
\begin{equation}
\label{histogan:eq.recoloring-loss}
{\mathcal{L}}_r = \beta R\left(\I_i, \I_r\right) + \gamma D\left(\I_r\right) + \alpha C\left(\mat{H}_r, \mat{H}_t\right)
\end{equation}
where $R\left(\cdot\right)$ is a reconstruction term, which encourages the preservation of image structure and $\alpha$, $\beta$, and $\gamma$ are hyperparameters used to control the strength of each loss term (see Appendix \ref{ch:appendix3} for associated ablation study).
The reconstruction loss term, $R\left(\cdot\right)$, computes the L1 norm between the second order derivative of our input and recolored images as:
\begin{equation}
	R\left(\I_i, \I_r\right) = \left\Vert \I_i \ast \mat{L} - \I_r \ast \mat{L} \right\Vert_1
\end{equation}
where $\ast \mat{L}$ denotes the application of the Laplacian operator.
The idea of employing the image derivative was used initially to achieve image seamless cloning \cite{perez2003poisson, afifi2015mpb}, where this Laplacian operator suppressed image color information while keeping the most significant perceptual details.
Intuitively, ReHistoGAN is trained to consider the following aspects in the output image: (i) having a similar color distribution to the one represented in the target histogram, this is considered by $C\left(\cdot\right)$, (ii) being realistic, which is the goal of  $D\left(\cdot\right)$, and (iii) having the same content of the input image, which is the goal of $R\left(\cdot\right)$.

\begin{figure}[!t]
\includegraphics[width=\linewidth]{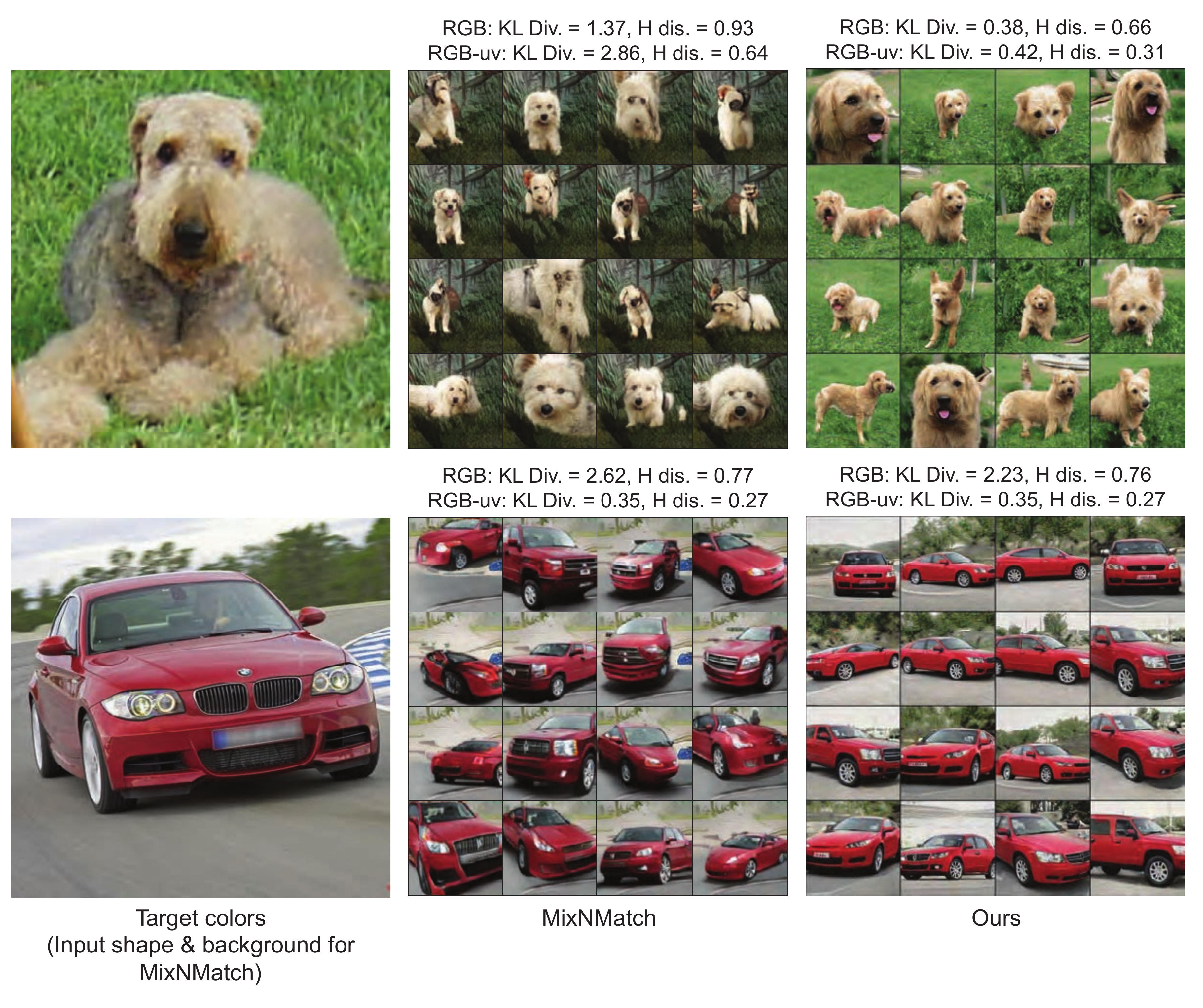}
\vspace{-7mm}
\caption[Comparison with the MixNMatch method \cite{li2020mixnmatch}.]{Comparison with the MixNMatch method \cite{li2020mixnmatch}. In the shown results, the target images are used as input shape and background images for the MixNMatch method \cite{li2020mixnmatch}.\label{histogan:fig:GAN_comparison_w_MixNMatch}}
\end{figure}

\begin{figure}[!t]
\includegraphics[width=\linewidth]{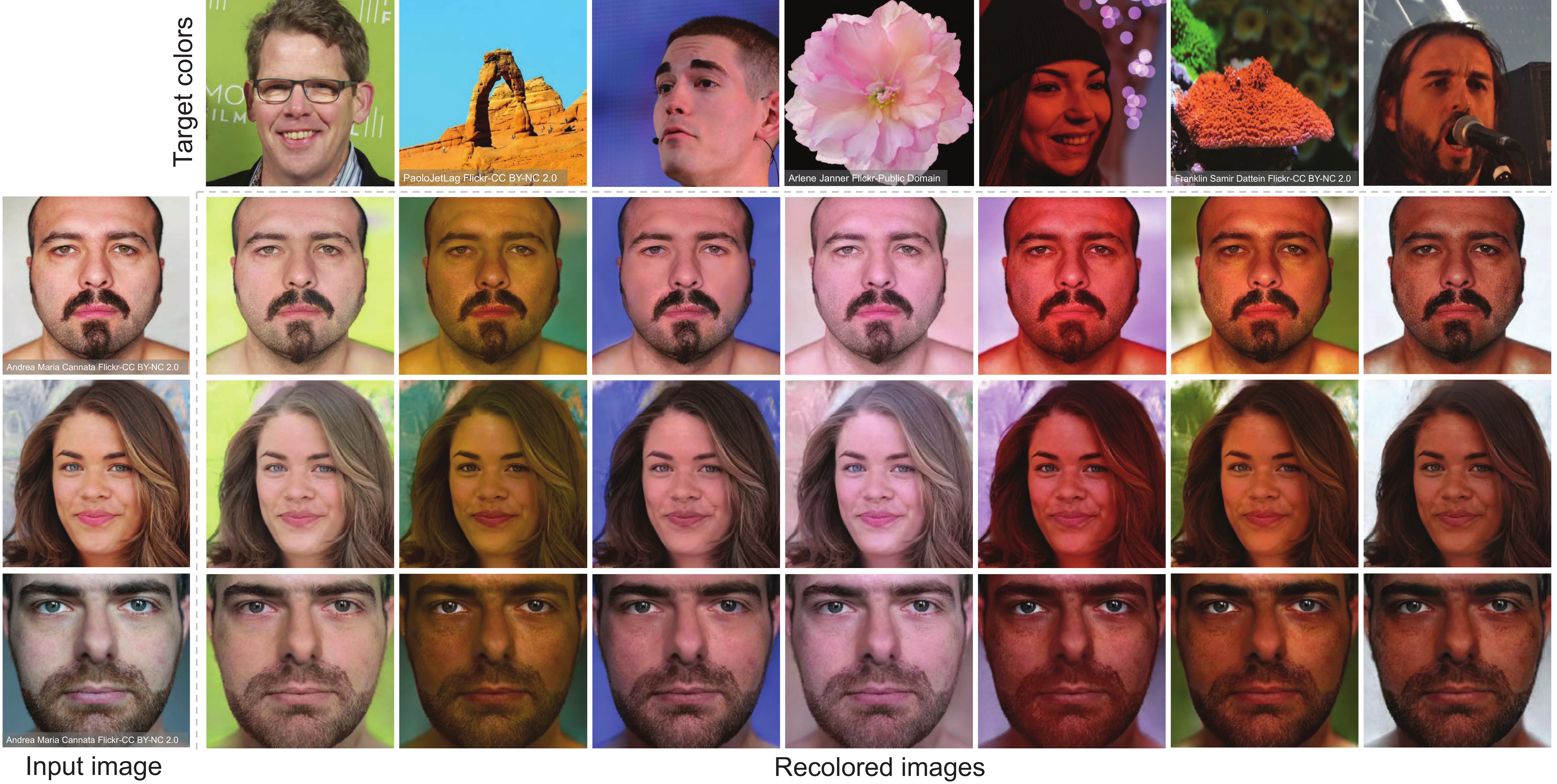}
\vspace{-7mm}
\caption[Results of our ReHistoGAN.]{Results of our ReHistoGAN. The shown results are after recoloring input images (shown in the left column) using the target colors (shown in the top row).\label{histogan:fig:face_recoloring}}
\end{figure}

Our model trained using the loss function described in Eq.\ \ref{histogan:eq.recoloring-loss} produces reasonable recoloring results.
However, we noticed that, in some cases, our model tends to only apply a global color cast (i.e., shifting the recolored image's histogram) to minimize $C\left(\cdot\right)$.
To mitigate this behavior, we added variance loss term to Eq.\ \ref{histogan:eq.recoloring-loss}.
The variance loss can be described as:
\begin{equation}
\label{histogan:eq.variance-loss}
V(\I_i, \I_r) = - w \sum_{c\in\{\cR,\cG,\cB\}}{\left| \sigma\left(\I_{ic} \ast \mat{G}\right) - \sigma\left(\I_{rc} \ast \mat{G}\right) \right|},
\end{equation}
where $\sigma\left(\cdot\right)$ computes the standard deviation of its input (in this case the blurred versions of $\I_i$ and $\I_r$ using a Gaussian blur kernel, $\mat{G}$, with a scale parameter of $15$), and $w = \Vert \mat{H}_t - \mat{H}_i \Vert_1$ is a weighting factor that increases as the target histogram and the input image's histogram, $\mat{H}_t$ and $\mat{H}_i$, become dissimilar and the global shift solution becomes more problematic.

The variance loss encourages the network to avoid the global shifting solution by increasing the differences between the color variance in the input and recolored images. The reason behind using a blurred version of each image is to avoid having a contradiction between the variance loss and the reconstruction loss---the former aims to increase the differences between the variance of the \textit{smoothed} colors in each image, while the latter aims to retain the similarity between the fine details of the input and recolored images. Figure \ref{histogan:fig:variance_loss} shows recoloring results of our trained models with and without the variance loss term.

We train ReHistoGAN with target histograms sampled from the target domain dataset, as described earlier in Sec.\ \ref{histogan:subsec.method-coloring-GAN-images} (Eq.\ \ref{histogan:eq.target_hist}).

A simpler architecture was experimented with initially, which did not make use of the skip connections and the end-to-end fine tuning (i.e., the weights of the HistoGAN head were fixed).
However, this approach gave unsatisfactory result, and generally failed to retain fine details of the input image.
A comparison between this approach and the above ReHistoGAN architecture can be seen in Fig.\ \ref{histogan:fig:comparison_with_projection}.

Our ReHistoGAN processes a single $256\!\times\!256$ image in $\sim$0.5 seconds using a single GTX 1080 GPU. To process images with higher resolutions, we use the guided upsampling procedure \cite{chen2016bilateral} (as described in Chapter\ \ref{ch:ch13}) with additional $\sim$21 seconds using an unoptimized implementation of the guided upsampling.

\section{Results and Discussion} \label{histogan:sec.results}

\begin{figure}[!t]
\includegraphics[width=\linewidth]{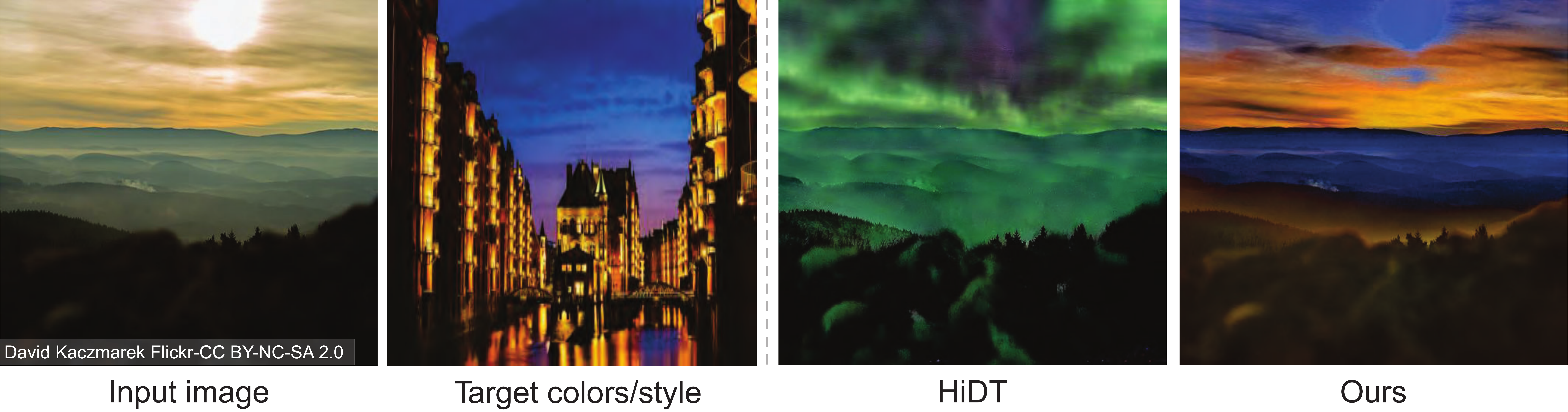}
\vspace{-7mm}
\caption[\hspace{0.5mm} Comparison with the high-resolution daytime translation (HiDT) method \cite{anokhin2020high}.]{Comparison with the high-resolution daytime translation (HiDT) method \cite{anokhin2020high}.\label{histogan:fig:comparison_w_HiDT}}
\end{figure}

\begin{figure}[!t]
\includegraphics[width=\linewidth]{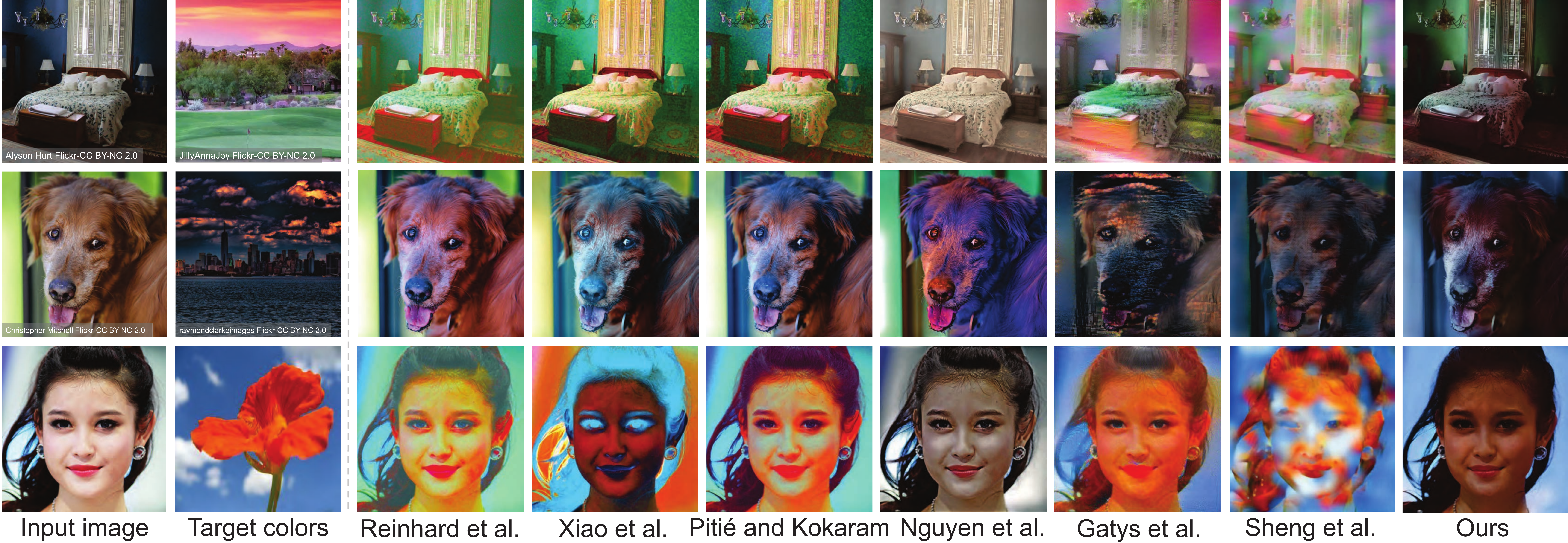}
\vspace{-7mm}
\caption[\hspace{0.5mm} Comparisons between our ReHistoGAN and other image color/style transfer methods.]{Comparisons between our ReHistoGAN and other image color/style transfer methods, which are: Reinhard et al., \cite{reinhard2001color}, Xiao et al., \cite{xiao2006color}, Piti\'e and Kokaram \cite{Pitie2007}, Nguyen et al., \cite{nguyen2014illuminant}, Gatys et al., \cite{gatys2016image}, and Sheng et al., \cite{sheng2018avatar}.\label{histogan:fig:compariosns_recoloring_class_based}}
\end{figure}

\begin{figure}[!t]
\includegraphics[width=\linewidth]{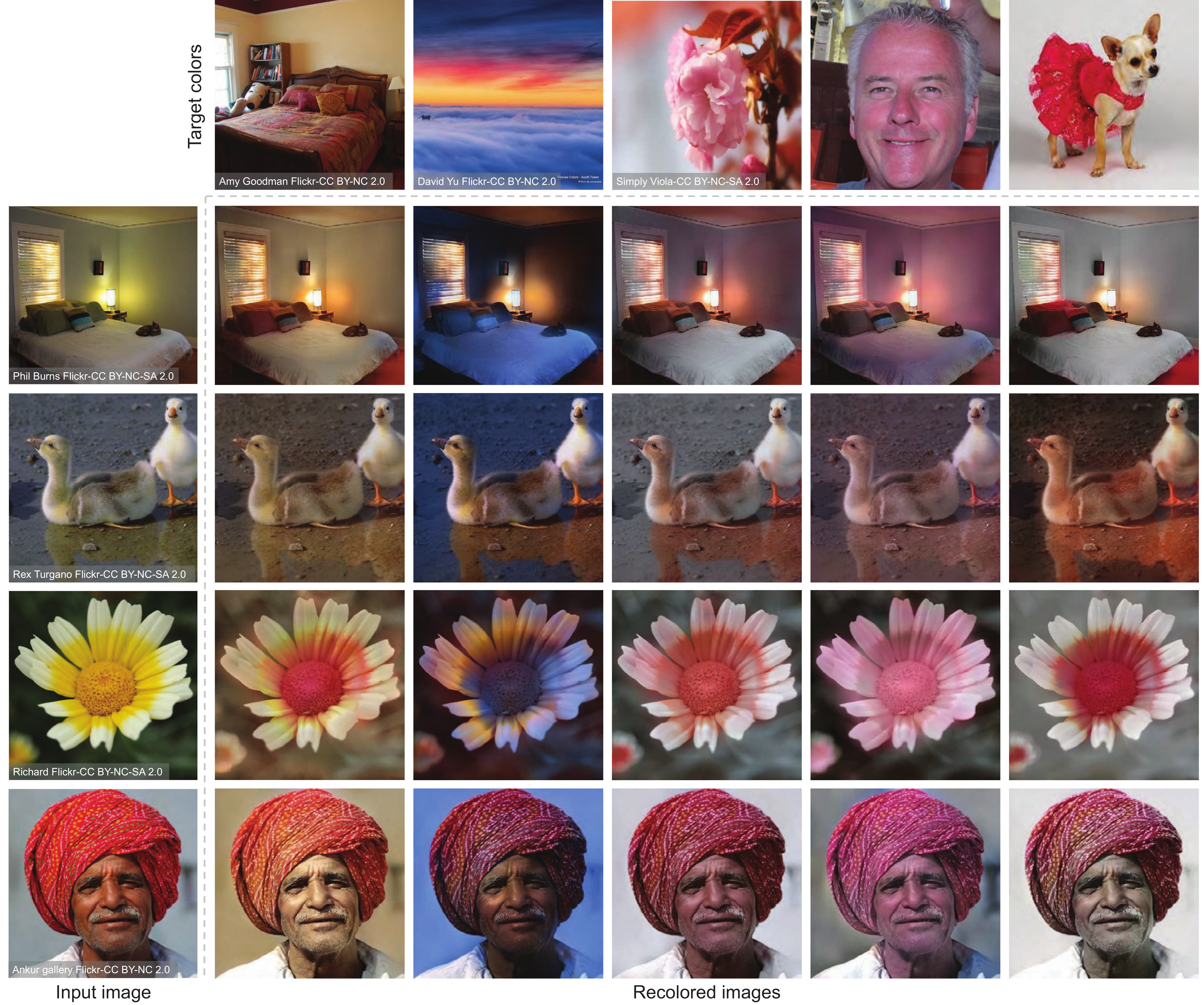}
\vspace{-7mm}
\caption[\hspace{0.5mm} Additional results for image recoloring.]{Additional results for image recoloring. We recolor input images, shown in the right by feeding our network with the target histograms of images shown in the top.\label{histogran_appendix:fig:qualitative_recoloring_class_based}}
\end{figure}

\begin{figure}[!t]
\includegraphics[width=\linewidth]{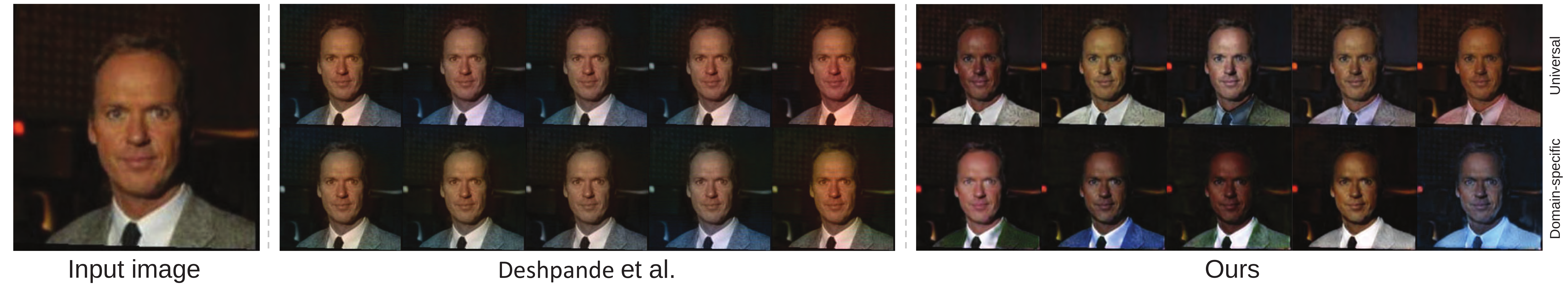}
\vspace{-7mm}
\caption[\hspace{0.5mm} Comparisons between our ReHistoGAN and the diverse colorization method proposed by Deshpande et al., \cite{deshpande2017learning}.]{Comparisons between our ReHistoGAN and the diverse colorization method proposed by Deshpande et al., \cite{deshpande2017learning}. For our ReHistoGAN, we show the resutls of our domain-specific and universal models.\label{histogran_appendix:fig:comparison_with_diverse_colorization}}
\end{figure}

\begin{figure}[!t]
\includegraphics[width=\linewidth]{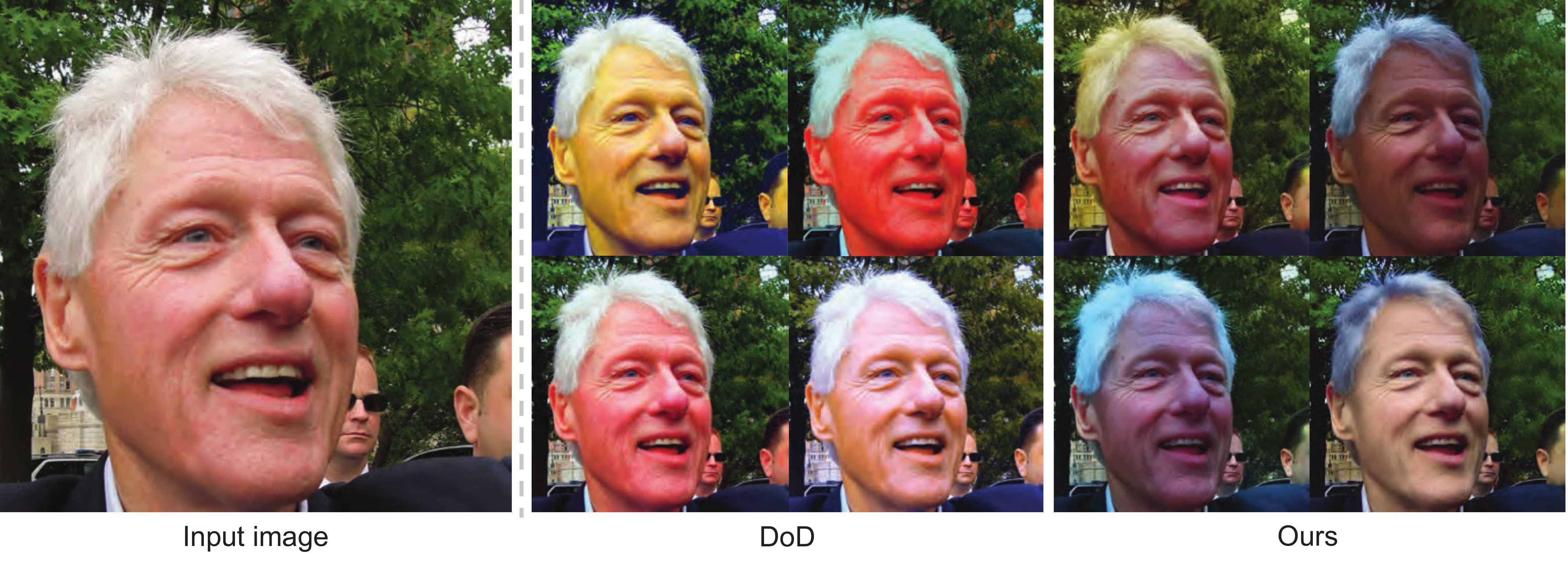}
\vspace{-7mm}
\caption[\hspace{0.5mm} Automatic recoloring comparison with our method in Chapter \ref{ch:ch14}.]{Automatic recoloring comparison with our method in Chapter \ref{ch:ch14}.\label{histogan:fig:compariosns_auto_recoloring}}
\end{figure}

\paragraph{Image Recoloring}

This section discusses our results and comparisons with alternative methods proposed in the literature for controlling color.
Due to hardware limitations, we used a lightweight version of the original StyleGAN \cite{karras2020analyzing} by setting $m$ to 16, shown in Fig.\ \ref{histogan:fig:GAN-design}.
We begin by presenting our image generation results, followed by our results on image recoloring.
Additional results, comparisons, and discussion are also available in Appendix \ref{ch:appendix3}.

\begin{table}[t]
\caption[Comparison with StyleGAN \cite{karras2020analyzing}.]{Comparison with StyleGAN \cite{karras2020analyzing}. The term `w/ proj.' refers to projecting the target image colors into the latent space of StyleGAN. We computed the similarity between the target and generated histograms in RGB and projected RGB-$uv$ color spaces. For each dataset, we report the number of training images. Note that StyleGAN results shown here \textit{do not} represent the actual output of \cite{karras2020analyzing}, as the used model here has less capacity ($m=16$).\label{histogan:table:results}}
\centering
\scalebox{0.54}{
\begin{tabular}{|c|c|c|c|c|c|c|c|c|c|c|c|}
\hline
\multirow{3}{*}{Dataset} & \multicolumn{6}{c|}{StyleGAN \cite{karras2020analyzing}} & \multicolumn{5}{c|}{HistoGAN (ours)} \\ \cline{2-12}
 & \multicolumn{2}{c|}{FID} & \multicolumn{2}{c|}{RGB hist. (w/ proj.)} & \multicolumn{2}{c|}{RGB-$uv$ hist. (w/ proj.)} & \multirow{2}{*}{FID} & \multicolumn{2}{c|}{RGB hist. (w/ proj.)} & \multicolumn{2}{c|}{RGB-$uv$ hist. (w/ proj.)} \\ \cline{2-7} \cline{9-12}
 & w/o proj. & w/ proj. & KL Div. & H dis. & KL Div. & H dis. &  & KL Div. & H dis. & KL Div. & H dis. \\ \hline
Faces (69,822) \cite{karras2019style} & 9.5018 & 14.194 & 1.3124 & 0.9710 & 1.2125 & 0.6724 & \cellcolor[HTML]{FFFFC7}{\textbf{8.9387}} & \cellcolor[HTML]{FFFFC7}{\textbf{0.9810}} & \cellcolor[HTML]{FFFFC7}{\textbf{0.7487}} & \cellcolor[HTML]{FFFFC7}{\textbf{0.4470}} & \cellcolor[HTML]{FFFFC7}{\textbf{0.3088}} \\ \hline
Flowers (8,189) \cite{nilsback2008automated} & 10.876 & 15.502 & 1.0304 & 0.9614 & 2.7110 & 0.7038 & \cellcolor[HTML]{FFFFC7}{\textbf{4.9572}} & \cellcolor[HTML]{FFFFC7}{\textbf{0.8986}} & \cellcolor[HTML]{FFFFC7}{\textbf{0.7353}} & \cellcolor[HTML]{FFFFC7}{\textbf{0.3837}} & \cellcolor[HTML]{FFFFC7}{\textbf{0.2957}} \\ \hline
Cats (9,992) \cite{catdataset} & \cellcolor[HTML]{FFFFC7}{\textbf{14.366}} & 21.826 & 1.6659 & 0.9740 & 1.4051 & 0.5303 & 17.068 & \cellcolor[HTML]{FFFFC7}{\textbf{1.0054}} & \cellcolor[HTML]{FFFFC7}{\textbf{0.7278}} & \cellcolor[HTML]{FFFFC7}{\textbf{0.3461}} & \cellcolor[HTML]{FFFFC7}{\textbf{0.2639}}\\ \hline
Dogs (20,579) \cite{khosla2011novel} & \cellcolor[HTML]{FFFFC7}{\textbf{16.706}} & 30.403 & 1.9042 & 0.9703 & 1.4856 & 0.5658 & 20.336 & \cellcolor[HTML]{FFFFC7}{\textbf{1.3565}} & \cellcolor[HTML]{FFFFC7}{\textbf{0.7405}} & \cellcolor[HTML]{FFFFC7}{\textbf{0.4321}} & \cellcolor[HTML]{FFFFC7}{\textbf{0.3058}} \\ \hline
Birds (9,053) \cite{wah2011caltech} & 3.5539 & 12.564 & 1.9035 & 0.9706 & 1.9134 & 0.6091 & \cellcolor[HTML]{FFFFC7}{\textbf{3.2251}} & \cellcolor[HTML]{FFFFC7}{\textbf{1.4976}} & \cellcolor[HTML]{FFFFC7}{\textbf{0.7819}} & \cellcolor[HTML]{FFFFC7}{\textbf{0.4261}} & \cellcolor[HTML]{FFFFC7}{\textbf{0.3064}} \\ \hline
Anime (63,565) \cite{animedataset}& \cellcolor[HTML]{FFFFC7}{\textbf{2.5002}} & 9.8890 & 0.9747 & 0.9869 & 1.4323 & 0.5929 & 5.3757 & \cellcolor[HTML]{FFFFC7}{\textbf{0.8547}} & \cellcolor[HTML]{FFFFC7}{\textbf{0.6211}} & \cellcolor[HTML]{FFFFC7}{\textbf{0.1352}} & \cellcolor[HTML]{FFFFC7}{\textbf{0.1798}} \\ \hline
Hands (11,076) \cite{afifi201911k} & 2.6853
 & 2.7826 & 0.9387 & 0.9942 & 0.3654 & 0.3709 & \cellcolor[HTML]{FFFFC7}{\textbf{2.2438}}
 & \cellcolor[HTML]{FFFFC7}{\textbf{0.3317}}
 & \cellcolor[HTML]{FFFFC7}{\textbf{0.3655}}
 & \cellcolor[HTML]{FFFFC7}{\textbf{0.0533}}
 & \cellcolor[HTML]{FFFFC7}{\textbf{0.1085}} \\ \hline
Landscape (4,316) & 24.216 & 29.248 & 0.8811 & 0.9741 & 1.9492 & 0.6265 & \cellcolor[HTML]{FFFFC7}{\textbf{23.549}} & \cellcolor[HTML]{FFFFC7}{\textbf{0.8315}} & \cellcolor[HTML]{FFFFC7}{\textbf{0.8169}} & \cellcolor[HTML]{FFFFC7}{\textbf{0.5445}} & \cellcolor[HTML]{FFFFC7}{\textbf{0.3346}} \\ \hline
Bedrooms (303,116) \cite{yu2015lsun} & 10.599 & 14.673 & 1.5709 & 0.9703 & 1.2690 & 0.5363 & \cellcolor[HTML]{FFFFC7}{\textbf{4.5320}} & \cellcolor[HTML]{FFFFC7}{\textbf{1.3774}} & \cellcolor[HTML]{FFFFC7}{\textbf{0.7278}} & \cellcolor[HTML]{FFFFC7}{\textbf{0.2547}} & \cellcolor[HTML]{FFFFC7}{\textbf{0.2464}} \\ \hline
Cars (16,185) \cite{krause20133d}& 21.485 & 25.496 & 1.6871 & 0.9749 & 0.7364 & 0.4231
& \cellcolor[HTML]{FFFFC7}{\textbf{14.408}}
& \cellcolor[HTML]{FFFFC7}{\textbf{1.0743}}
& \cellcolor[HTML]{FFFFC7}{\textbf{0.7028}}
& \cellcolor[HTML]{FFFFC7}{\textbf{0.2923}}
& \cellcolor[HTML]{FFFFC7}{\textbf{0.2431}}\\ \hline
Aerial Scenes (36,000) \cite{maggiori2017can} & \cellcolor[HTML]{FFFFC7}{\textbf{11.413}} & 14.498 & 2.1142 & 0.9798 & 1.1462 & 0.5158 & 12.602 & \cellcolor[HTML]{FFFFC7}{\textbf{0.9889}} & \cellcolor[HTML]{FFFFC7}{\textbf{0.5887}} & \cellcolor[HTML]{FFFFC7}{\textbf{0.1757}} & \cellcolor[HTML]{FFFFC7}{\textbf{0.1890}}\\ \hline
\end{tabular}}
\end{table}

\paragraph{Image Generation} \label{histogan:sec.results-generated-images}

Figure\ \ref{histogan:fig:GAN_results} shows examples of our HistoGAN-generated images.\
Each row shows samples generated from different domains using the corresponding input target colors.\
For each domain, we fixed the style vectors responsible for the coarse and middle styles to show our HistoGAN's response to changes in the target histograms.
Qualitative comparisons with the recent MixNMatch method \cite{li2020mixnmatch} are provided in Fig.\ \ref{histogan:fig:GAN_comparison_w_MixNMatch}.

To evaluate the potential improvement/degradation of the generated-image diversity and quality caused by our modification to StyleGAN, we trained StyleGAN~\cite{karras2020analyzing} with $m=16$ (i.e., the same as our model capacity) without our histogram modification.
We evaluated both models on different datasets, including our collected set of landscape images.
For each dataset, we generated 10,000 $256\!\times\!256$ images using the StyleGAN and our HistoGAN.
We evaluated the generated-image quality and diversity using the Frech\'et inception distance (FID) metric \cite{heusel2017gans} using the second max-pooling features of the Inception model~\cite{szegedy2015going}.

We further evaluated the ability of StyleGAN to control colors of GAN-generated images by training a regression deep neural network (ResNet \cite{he2016deep}) to transform generated images back to the corresponding fine-style vectors.
These fine-style vectors are used by the last two blocks of StyleGAN and are responsible for controlling delicate styles, such as colors and lights \cite{karras2019style, karras2020analyzing}.

The training was performed for each domain separately using 100,000 training StyleGAN-generated images and their corresponding ``ground-truth'' fine-style vectors.
In the testing phase, we used the trained ResNet to predict the corresponding fine-style vectors of the target image---these target images were used to generate the target color histograms for HistoGAN's experiments. We then generated output images based on the predicted fine-style vectors of each target image.
In the evaluation of StyleGAN and HistoGAN, we used randomly selected target images from the same domain.

The Hellinger distance and KL divergence were used to measure the color errors between the histograms of the generated images and the target histogram; see Table \ref{histogan:table:results}.

Figure \ref{histogan:fig:face_recoloring} shows examples of image recoloring using our ReHistoGAN. A comparison with the recent high-resolution daytime translation (HiDT) method \cite{anokhin2020high} is shown in Fig.\ \ref{histogan:fig:comparison_w_HiDT}.
Additional comparisons with image recoloring and style transfer methods are shown in Fig.\ \ref{histogan:fig:compariosns_recoloring_class_based}.
Arguably, our ReHistoGAN produces image recoloring results that are visually more compelling than the results of other methods for image color/style transfer.
As shown in Fig.\ \ref{histogan:fig:compariosns_recoloring_class_based}, our ReHistoGAN produces realistic recoloring even when the target image is from a different domain than the input image, compared to other image style transfer methods (e.g., \cite{gatys2016image, sheng2018avatar}). Additional results are shown in Fig.\ \ref{histogran_appendix:fig:qualitative_recoloring_class_based}.

Another strategy for image recoloring is to learn a diverse colorization model. That is, the input image is converted to grayscale, and then a trained method for diverse colorization can generate different colorized versions of the input image. In Fig.\ \ref{histogran_appendix:fig:comparison_with_diverse_colorization}, we show a qualitative comparison with the diverse colorization method proposed by Deshpande et al., \cite{deshpande2017learning}.

Lastly, we provide a qualitative comparison with our auto-recoloring method presented in Chapter \ref{ch:ch14} in Fig.\ \ref{histogan:fig:compariosns_auto_recoloring}.
In the shown example, our target histograms were dynamically generated by sampling from a pre-defined set of histograms and applying a linear interpolation between the sampled histograms (see Eq.\ \ref{histogan:eq.target_hist}).

\begin{figure}[!t]
\includegraphics[width=\linewidth]{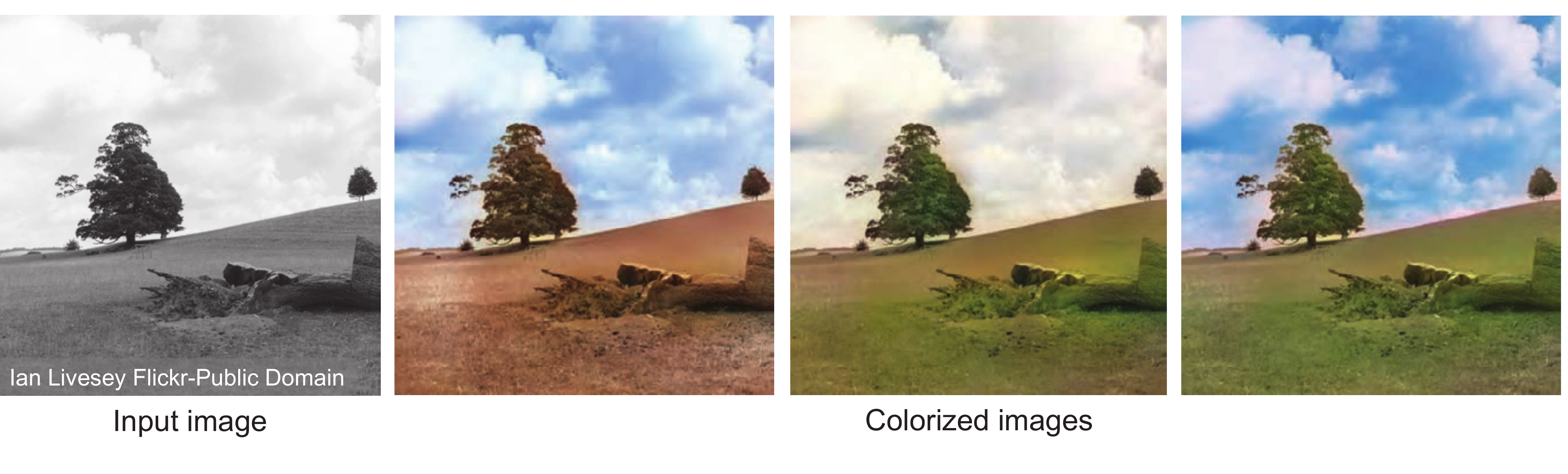}
\vspace{-7mm}
\caption[\hspace{0.5mm} Results of using our ReHistoGAN for a diverse image colorization.]{Results of using our ReHistoGAN for a diverse image colorization.\label{histogan:fig:colorization}}
\end{figure}

\paragraph{What is Learned?}
Our method learns to map color information, represented by the target color histogram, to an output image's colors with a realism consideration in the recolored image.  Maintaining realistic results is achieved by learning proper matching between the target colors and the input image's semantic objects (e.g., grass can be green, but not blue). To demonstrate this, we examine a trained ReHistoGAN model for an image colorization task, where the input image is grayscale.  The input of a grayscale image means that our ReHistoGAN model has no information regarding objects' colors in the input image.  Figure\ \ref{histogan:fig:colorization} shows outputs where the input has been ``colorized''.  As can be seen, the output images have been colorized with good semantic-color matching based on the image's content.

\section{Summary} \label{histogan:sec.conclusion}

We have presented HistoGAN, a simple, yet effective, method for controlling colors of GAN-generated images.
Our HistoGAN framework learns how to transfer the color information encapsulated in a target histogram feature to the colors of a generated output image.\
To the best of our knowledge, this is the first work to control the color of GAN-generated images directly from color histograms.
Color histograms provide an abstract representation of image color that is decoupled from spatial information.
This allows the histogram representation to be less restrictive and suitable for GAN-generation across arbitrary domains.

We have shown that HistoGAN can be extended to control colors of real images in the form of the ReHistoGAN model.
Our recoloring results are visually more compelling than currently available solutions for image recoloring.
Our image recoloring also enables ``auto-recoloring'' by sampling from a pre-defined set of histograms.
This allows an image to be recolored to a wide range of visually plausible variations.
HistoGAN can serve as a step towards intuitive color control for GAN-based graphic design and artistic endeavors.

\part{Conclusions and Future Work\label{part:conclusion}}
\chapter{Conclusions and Future Work \label{ch:ch16}}
\section{Concluding Remarks}

This thesis has presented a number of works targeting research related to color correction, enhancement, and editing. The thesis discussed color correction and editing from the standpoint of the camera's ISP. We first reviewed different standard color spaces and the details of camera ISPs in Chapter\ \ref{ch:ch2}. We then presented a survey of prior work for color constancy, enhancement, and editing in Chapter\ \ref{ch:ch3}. 
Afterwards, we discussed methods for computational color correction onboard cameras in Chapters \ref{ch:ch5} and \ref{ch:ch6}. Specifically, we have proposed two sensor-independent learning-based methods for illuminant estimation.

In Chapter \ref{ch:ch7}, we have presented the first method to correct WB errors appears in camera-rendered photographs. Through an extensive evaluation, we showed that our method outperforms existing solutions in correcting such WB errors in camera-rendered images. We further discussed a modification to our WB method (presented in Chapter \ref{ch:ch7}) to augment training data used by DNNs for different computer vision tasks in Chapter \ref{ch:ch8}. We showed that our WB augmenter improves the robustness of DNN methods for image classification and image semantic segmentation against WB errors.  

We have discussed a set of methods designed for color editing based on WB editing in Chapters \ref{ch:ch9}--\ref{ch:ch11}. In particular, Chapter \ref{ch:ch9} extends our method proposed in Chapter \ref{ch:ch7} to allow for user interactive WB editing in post-capture stage. Then, we have proposed a DNN framework for WB editing in Chapter \ref{ch:ch10}. Lastly, we discussed a simple modification to camera ISPs in Chapter \ref{ch:ch11} to achieve accurate WB editing in camera-rendered images.

We also discussed camera exposure errors and how they can significantly affect the quality of colors in camera-rendered photographs. To that end, we have proposed two methods to enhance under- and over-exposed images in Chapters \ref{ch:ch12} and \ref{ch:ch13}. Our first method in Chapter \ref{ch:ch12} was built on the idea of image linearization, where we discussed the benefit of accurately unprocess camera photo-finishing to enhance low-light images. In Chapter \ref{ch:ch13}, we proposed a post-capture DNN enhancement method to correct colors of under- and over-exposed photographs. 

Lastly, we have proposed two methods for auto color editing in camera-rendered photographs, where we have presented a data-driven approach for auto recoloring (Chapter \ref{ch:ch14}) and a GAN-based method for controlling colors in synthetic images produced by GANs and real photographs (Chapter \ref{ch:ch15}).

\subsection{Broader Impact}

We discussed several methods in this thesis that target improving the quality of photographs. When compared with other alternatives, our methods introduce more flexible, practical, and accurate image color correction, enhancement, and editing solutions. While the impact of our work is often evaluated from an aesthetic point of view, color correction and
image enhancement can be used as a pre-processing step to enhance crucial computer vision applications. These applications may include skin cancer diagnosis \cite{karaimer2019customized}, diabetic retinopathy detection/classification \cite{smailagic2019learned}, image forensics \cite{de2013exposing}, and object tracking \cite{yilmaz2006object}. The importance of color correction for computer vision tasks was discussed in Chapter \ref{ch:ch8}, where we presented,
through a thorough evaluation, an experimental analysis of color correction's impact on image classification and image semantic segmentation tasks. Chapter \ref{ch:ch8} also discussed the idea of WB augmentation, which can improve the accuracy of other critical computer vision tasks.

We have also proposed methods for image recoloring that can produce realistic versions of images after changing their original colors. This recoloring tool can be used as a data augmenter to automatically recolor (or augment) training images to improve the robustness and accuracy of other deep learning techniques for different tasks. These recoloring techniques, though, like two sides of a coin, carry both the negative and the positive sides, where image recoloring may result in changes in images intended to deceive the observer (for example, day-to-night image translation or alter evidence). Future work is required to ensure that the authenticity of images is verifiable after such modifications have been performed.

\section{Future Research Directions}

The work in this thesis focused on color processing and discussed two main factors that significantly affect the quality of colors in camera-rendered photographs, namely: (i) WB settings and (ii) exposure settings. The work presented in this thesis was modeled based on our knowledge of camera ISP pipeline design. This knowledge allows us to generate accurate datasets to develop our methods. Through our discussion on WB, we showed that WB editing is not only required to correct improperly white-balanced images but can also help to satisfy user preference, especially in challenging scenes (e.g., multi-illuminant scenes). We further discussed another direction of color editing by modeling/learning a prior knowledge of semantic-color matches to achieve a realistic auto image recoloring based on the image's content. We then extended this data-driven idea to attain more compelling results using GANs. Through our discussion in previous chapters, we can summarize the future research directions into the following points: (i) deep learning in camera ISPs, (ii) multi-illuminant CC, (iii) user preferences in WB, (iv) enhancing  over-exposed images, (v) and universal image recoloring.

\subsection{Deep Learning in Camera ISPs}
As discussed in Chapter \ref{ch:ch2}, there is a sequence of processing steps applied onboard camera ISPs to render the final sRGB image. Each module of camera ISP is responsible for a particular task to enhance the perceptual quality of the captured image. Due to the astounding results achieved by deep learning on a broad range of computer vision tasks, researchers proposed several methods to replace specific modules in camera ISPs with DNNs showing impressive results compared to traditional approaches (e.g., \cite{liu2020joint, 9296313, afifi2020cross, wang2020practical, HDRNET, punnappurath2019learning}). More recently, a few methods proposed to replace the entire camera ISP with a single DNN model to be trained in an end-to-end fashion (e.g., \cite{schwartz2018deepisp, ignatov2020replacing}). While this idea is promising---in the sense that a single end-to-end trained model would diminish the effort required to tune the parameters of each camera ISP's module to get the final desirable results---it has some drawbacks. First, the computational power required by currently used DNNs usually exceeds the memory and computing limits in camera hardware units. Second, such DNNs are often treated as black boxes making it hard to have insights into the reason behinds any failure cases. Third, although employing deep learning offers image quality as good as that is produced by traditional camera ISPs without the tedious manual parameter tuning process required to deploy traditional camera ISP, training such DNNs usually requires collecting paired training data, which may require more effort than the camera ISP's parameter tuning procedure. 

With that said, DNNs not only allow the generation of good-quality sRGB images, they can also be designed to have additional benefits that are hard to be attempted by traditional camera ISPs. For example, the recent work in \cite{xing21invertible} proposed an invertible DNN to emulate almost the entire camera ISP's tasks (i.e., denoising, white balancing, etc.), where the DNN used to replace the camera ISP consists of invertible blocks, which means the entire rendering process can be reversed to reconstruct the original linear raw images. 

Following the idea of using invertible DNNs to emulate the camera rendering procedure \cite{xing21invertible}, there is an interesting research direction to not only utilize deep learning to replace camera ISPs but also to have a more effective diagnosis of failure cases. One may think of designing a single deep learning framework with different invertible sub-networks, each of which is designed for a specific rendering task. That is, each step in the rendering (e.g., denoising, white balancing, color mapping, etc.) can be inverted or modified in the post-capture stage. Another interesting research direction related to this point would be designing a learning-based camera ISP that learns user preferences and updates the ISP color rendering based on the post-capture edits performed by the user. 

\subsection{Multi-Illuminant Color Constancy}
We have discussed and proposed several methods for color correction in order to achieve CC in photographs in Chapters \ref{ch:ch3}--\ref{ch:ch7}. These methods, including our methods (Chapters \ref{ch:ch5}--\ref{ch:ch7}), target scenes with a single uniform lighting. In many real scenarios, however, scenes have mixed light sources and, thus, these methods are expected to correct colors only for one of these illuminants and ignore other illuminants. For that reason, we propose the WB editing methods in Chapters \ref{ch:ch9}--\ref{ch:ch11} to circumvent the issue by  allowing the user to manually edit WB settings in the post-capture. 

Though this interactive approach gives a reasonable way to deal with such cases, auto multi-illuminant scene color constancy would offer a more comfortable and effective solution. In the literature, as discussed in Chapter \ref{ch:ch3}, there are a few attempts proposed to deal with multi-illuminant scenarios. These methods, however, could work only under certain conditions and generally are not accurate enough. While deep learning solutions are expected to achieve more accurate results for multi-illuminant color constancy, there are no paired multi-illuminant datasets that represent realistic scenarios of scenes with mixed lighting for supervised learning. A promising research direction is to generate realistic multi-illuminant datasets to accelerate the adoption of learning-based multi-illuminant color constancy by the research community and the camera industry.

\subsection{User Preferences in White Balancing}
\begin{figure}[!t]
\includegraphics[width=\linewidth]{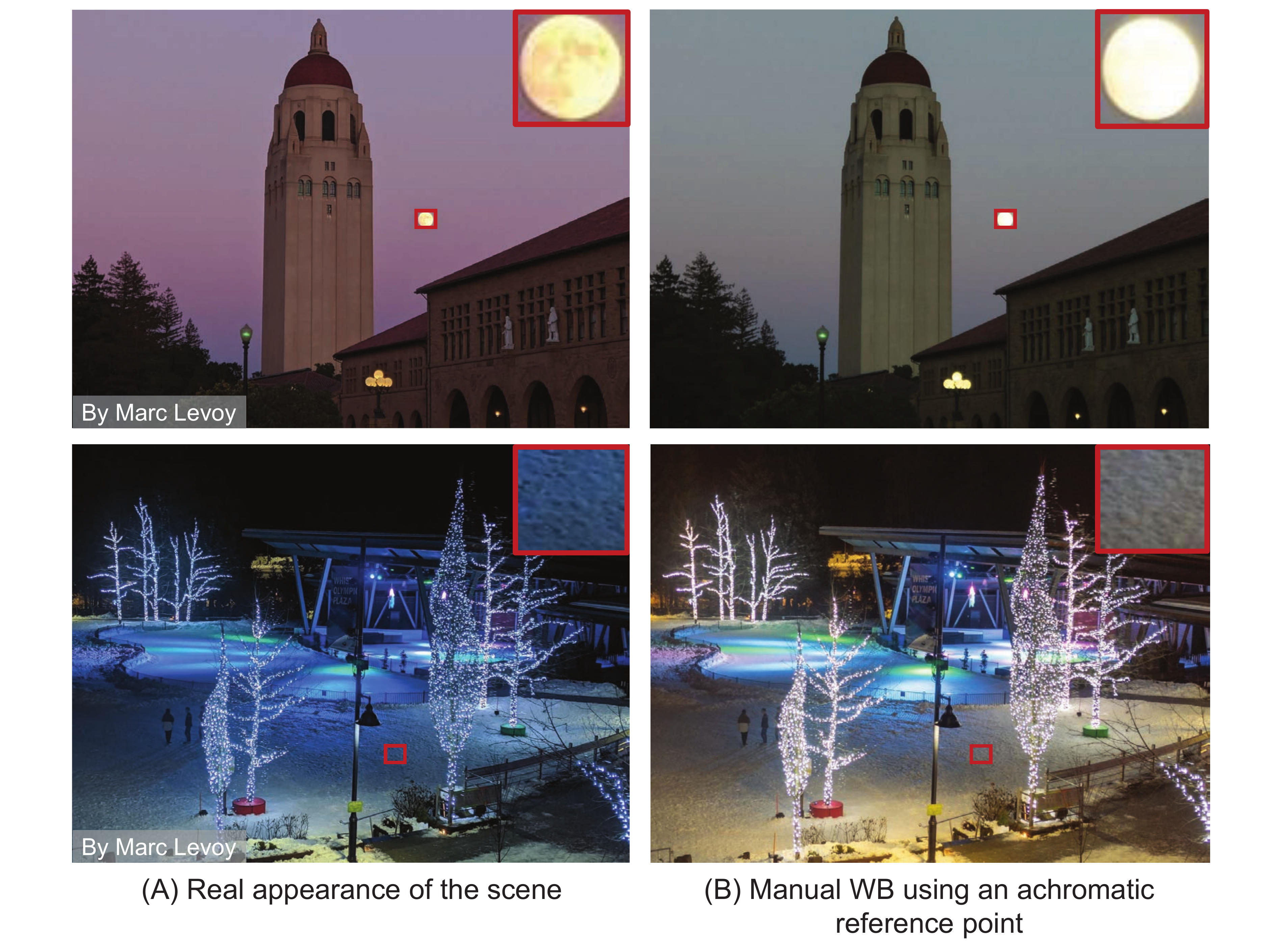}
\vspace{-8mm}
\caption[Image white balancing does not always match what we see in reality.]{Image white balancing does not always match what we see in reality. (A) Real scene appearance. (B) WB result. In each example, we rendered the DNG file of each image using Adobe Camera Raw in Photoshop. For WB results, an achromatic reference point were manually selected from each raw image. The red boxes show the selected reference point used for each image. Photo credit: Marc Levoy.}
\label{future_work:fig:user_preference}
\end{figure}

As discussed in Chapter \ref{ch:ch3}, WB aims to normalize the color cast caused by scene illumination, such that achromatic object's RGB vectors should lie along the achromatic white line in the 3D color space (i.e., R=G=B). In many cases, however, what we observe in real scenes do not match this goal. Figure \ref{future_work:fig:user_preference} shows two examples, where the real appearance of the scene does not match the result of accurate WB correction using manually selected achromatic reference point. As can be seen, the WB procedure changes the colors to make achromatic objects look white. Though this makes
intuitive sense, as any WB algorithm would remove any color casts caused by scene illuminant, these results seem incorrect to many users as it shows completely different colors than what is observed in reality. Potential research directions may include performing a comprehensive user study to define user preferences under diverse lighting conditions. Based on this analysis of user preferences, one could consider user preferences in the ground-truth labels of color constancy datasets to train DNN models that achieve more compelling WB results.

\subsection{Enhancing Over-Exposed Images}
In Chapter \ref{ch:ch13}, we have discussed the first work that employs deep learning towards explicitly enhancing over-exposed images and showed that most of the prior work focuses only on low-light image enhancement and ignores the over-exposure errors. In many cases, over-exposed images may have no information in several regions due to camera exposure errors. Thus, a DNN should hallucinate the missing content in order to correct this image. This hallucination should have different treatments based on the semantic objects in the photographs---for example, completing a corrupted grass region with a reasonable inpainting  quality can be accepted, while the situation is different for other objects, such as faces that require a more careful image completion. A potential research direction may include semantic-aware inpainting methods to correct over-exposed images.

\subsection{Universal Image Recoloring}
We discussed a GAN-based DNN method for image recoloring in Chapter \ref{ch:ch15}. While our method offers domain-specific recoloring models, we show in Appendix \ref{ch:appendix3} that by training on a large dataset, we can train a universal model for image recoloring. The results of this universal recoloring model, however, are less realistic than the domain-specific trained models. Another direction of future research work may include a pre-classification stage to determine the class of each image before recoloring in order to utilize a proper recoloring model for each image.

\part{Bibliography\label{part:reference}}
\bibliographystyle{plain}
\bibliography{references}

\part{Appendices\label{part:appendex}}

\begin{appendices}

\chapter{Applications of CIE XYZ Net\label{ch:appendix0}}
In Chapter \ref{ch:ch12}, we demonstrated the usefulness of the CIE XYZ Net on low-light image enhancement. Here, we show additional low-level vision tasks that can benefit from our CIE XYZ reconstruction. Specifically, we discuss how we can reconstruct raw-like images from the reconstructed CIE XYZ images. Further, we show significant gains that can be obtained by this processing framework for additional computer vision tasks. 

\section{Raw-RGB Image Reconstruction}

\begin{figure}[t]
\includegraphics[width=\linewidth]{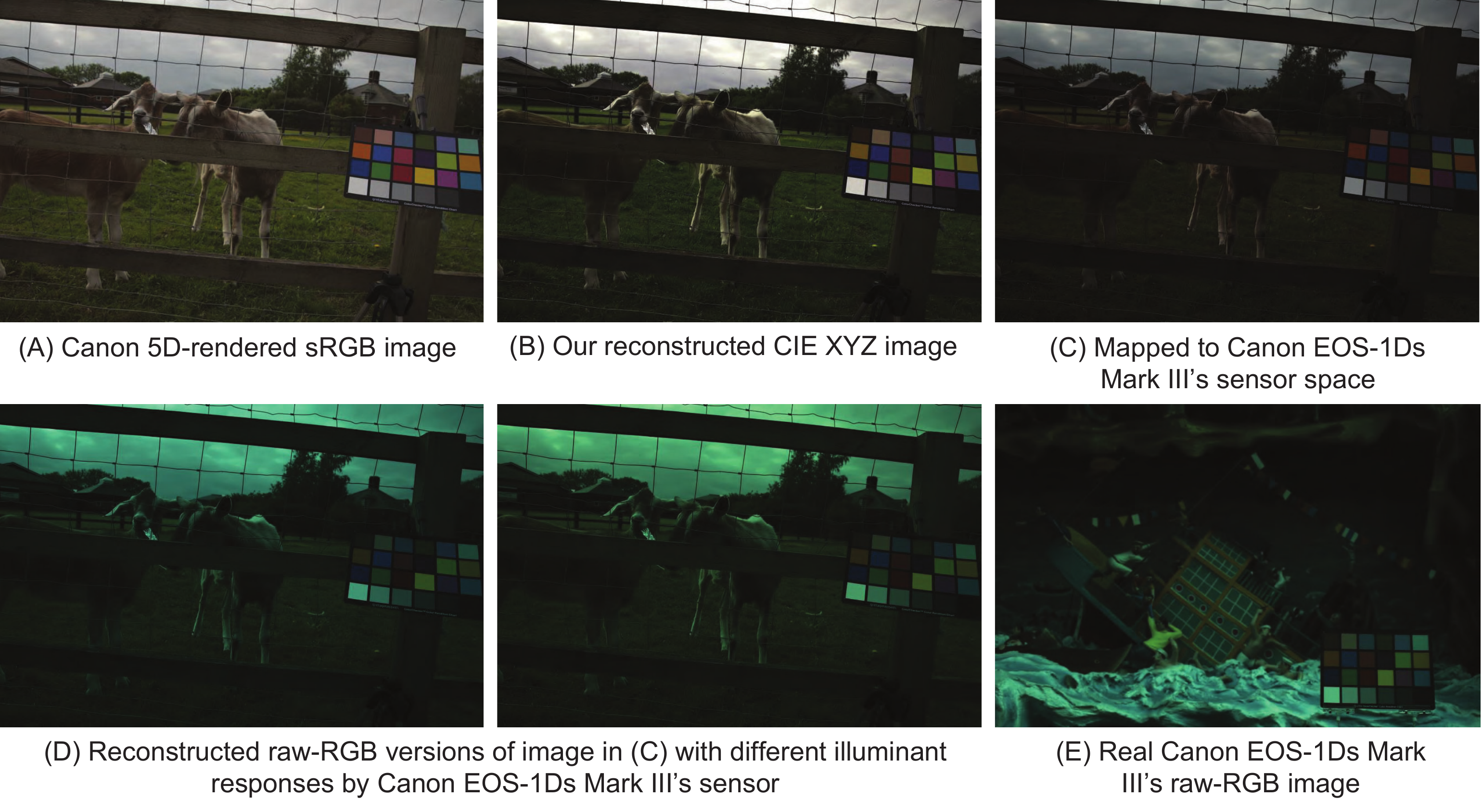}
\vspace{-7mm}
\caption[Sensor raw-RGB image reconstruction.]{Sensor raw-RGB image reconstruction. (A) An sRGB image rendered by Canon 5D from Gehler-Shi \cite{gehler2008bayesian}. (B) Our reconstructed CIE XYZ image. (C) Our reconstructed raw image in the raw-RGB space of the Canon EOS-1Ds Mark III. (D) Two generated raw-RGB images with different illuminant responses in the Canon EOS-1Ds Mark III's sensor space. (E) A real raw-RGB image captured by the Canon EOS-1Ds Mark III taken from the eight-camera NUS dataset \cite{cheng2014illuminant}. To aid visualization, the shown images are scaled by a factor of two.}
\label{xyz:fig:raw_reconstruction}
\end{figure}

One of the advantages of accurately reconstructing scene-referred images is the ability to map the reconstructed images further into a sensor raw-RGB space. Specifically, we can synthetically generate raw-RGB images in any target sensor space by capturing an image with a color rendition calibration chart placed in the scene. The captured image is saved in both the camera's sensor raw-RGB space and the camera-rendered sRGB color space. As the CIE XYZ space is defined for correctly white-balanced colors, we first correct the white balance of the raw-RGB image using the color rendition chart. We then reconstruct the XYZ image using our XYZ network and compute a $3\!\times\!3$ matrix to map our reconstructed image into the sensor space. We refer to this matrix as the XYZ$\rightarrow$raw matrix.

Note that in order to achieve an accurate mapping from the CIE XYZ space to the sensor raw space, this matrix should be calibrated under different illuminant conditions. For simplicity, we here compute a single global matrix to approximate this mapping.
This XYZ$\rightarrow$raw matrix is then used to map any arbitrary image into this sensor space by first reconstructing the corresponding XYZ image, followed by mapping it into the sensor space. The assumption here is that as our method achieves superior linearization to the available solutions (see Table \ref{xyz:Table0}), this calibration process would result in a better sRGB$\rightarrow$raw-RGB mapping.

To validate this assumption, we compare between the raw-RGB reconstruction based on our reconstruction against the raw-RGB reconstruction that is computed based on the standard XYZ mapping~\cite{anderson1996proposal, ebner2007color}. We examine the data augmentation task for illuminant estimation. Scene illuminant estimation is a well-studied problem in computer vision literature. Briefly, we can describe this problem as follows. Given a linear raw-RGB image $\mat{I}_{\text{raw}}$ captured by a specific camera sensor, the goal is to determine a 3D vector $\mat{\ell}$ that represents the illuminant color in the captured scene. Recent work achieves promising results using deep learning to estimate the illuminant vector $\mat{\ell}$ by training deep models that can be later used in the inference phase to estimate illumination colors of given testing images captured by the same sensor used in the training stage \cite{gehler2008bayesian}.

There is currently a challenge in the available datasets for the illuminant estimation task, which is the limited number of available training images captured by the same sensor---for example, the eight-camera NUS dataset \cite{cheng2014illuminant}, one of the common datasets used for illuminant estimation, has 200 images on average for each camera sensor. In this experiment, we examine our raw-like reconstructed images to serve as a data augmenter to train deep learning models for illuminant estimation. Specifically, we train a simple deep learning model to estimate the scene illuminant of a given raw-RGB image captured by Canon EOS-1Ds Mark III \cite{cheng2014illuminant}.

The model is designed to accept a $150\!\times\!150$ raw-RGB image (similar to prior work that proposed to use thumbnail images for the illuminant estimation task \cite{FFCC, afifi2019sensor}). The model includes a sequence of conv, LReLU, BN, and fc layers. In particular, the model consists of two conv--LReLU--conv--BN--LReLU blocks, followed by a conv--LReLU--FC--LReLU--dropout--FC--LReLU--FC block. All conv layers have $3\!\times\!3$ filters with a different number of output channels and stride steps. The first, second, and third conv layers have 64 output channels, while the fourth and fifth conv layers have 128 and 256 output channels, respectively. The stride steps were set to 2 for the first three conv layers. For the last two conv layers, we used a stride step of 3. The first two FC layers have 256 output neurons, while the last FC layer has 3 output neurons. We trained each model for 50 epochs to minimize the angular error between the estimated illuminant vector and the ground truth illuminant. The training process was performed with a learning rate of $10^{-4}$ and mini-batch of 32 using the Adam optimizer \cite{kingma2014adam} with a decay rate of gradient moving average 0.9 and a decay rate of squared gradient moving average 0.999.

There are only 256 original raw-RGB images captured by Canon EOS-1Ds Mark III in the NUS dataset \cite{cheng2014illuminant}. For each image, there is a ground-truth scene illuminant vector extracted from the color rendition chart. During training and testing processes, the color chart is masked out in each image to avoid any bias. To augment the data, we first computed the $3\!\times\!3$ XYZ$\rightarrow$raw calibration matrix as described earlier for our XYZ reconstruction and the standard XYZ mapping. This reconstruction process was performed using a single image captured by the Canon EOS-1Ds Mark III camera with a color rendition chart. %

Afterwards, we used 3,752 white-balanced camera-rendered sRGB images captured by ten different camera models other than our Canon EOS-1Ds Mark III. These images were taken from the Rendered WB dataset \cite{afifi2019color}. Each sRGB image is converted to the CIE XYZ space using our method and the standard XYZ mapping, followed by mapping each reconstructed image to the Canon EOS-1Ds Mark III sensor space using the calibration matrix computed for each XYZ reconstruction method, respectively.

As the calibration matrices map from the reconstructed XYZ space to the white-balanced sensor raw-RGB space, we can apply illuminant color casts, randomly selected from the ground-truth illuminant vectors provided in the Canon EOS-1Ds Mark III's set, to synthetically generate additional training data to train the deep model. Figure \ref{xyz:fig:raw_reconstruction} shows an example.  This process is inspired by previous work in \cite{BMVC1, fourure2016mixed}, which randomly selected illuminant vectors from the ground-truth set and applied chromatic adaptation to augment the training set. These methods, however, use the same images (256 images in the case of the Canon EOS-1Ds Mark III's set) without introducing new image content to the trained model.

We randomly selected 50 testing images from the original 256 images provided in the NUS dataset for the Canon EOS-1Ds Mark III camera. We fixed this testing set over all experiments and excluded these images from any training processes. Table \ref{xyz:table:illuminant_estimation} shows the angular error of the trained model using the following training sets: (i) real training data, (ii) reconstructed raw-like images using the standard XYZ mapping, (iii) real training data and reconstructed raw-like images using the standard XYZ mapping, (iv) reconstructed raw-like images using our XYZ reconstruction, and (v) real training data and reconstructed raw-like images using our XYZ reconstruction. As can be seen, the best results were obtained by using our raw-like reconstruction and real training data. Notice that training only on our raw-like reconstruction gives better results compared with the results obtained by training on real data or reconstructed raw-like images using the standard XYZ mapping. Additional training details are given in the supplementary materials.

\begin{table}[t]
\caption[Angular error of illuminant estimating using the image set captured by the Canon EOS-1Ds Mark III in the NUS dataset \cite{cheng2014illuminant}.]{Angular error of illuminant estimating using the image set captured by the Canon EOS-1Ds Mark III in the NUS dataset \cite{cheng2014illuminant}. We compare the results obtained by training a deep neural network on real raw-RGB training images, reconstructed (rec.) raw-RGB training images based on the standard XYZ reconstruction, and our CIE XYZ reconstruction. The best results are shown in bold.}\label{xyz:table:illuminant_estimation}
\centering
\scalebox{0.7}{
\begin{tabular}{|l|c|c|c|c|}
\hline

Training data & Mean & Median & Best 25\% & Worst 25\% \\ \hline
Real & 4.15 & 3.89 & 1.13 & 7.85  \\ \hline
Rec. (standard) & 3.37 & 3.03 & 1.05 & 6.68 \\ \hline
Real and rec. (standard) & 2.72 & 2.60 & 0.72 & 4.99 \\ \hline
Rec. (ours) & 3.00 & 2.61 & 0.83 & 5.37 \\ \hline
Real and rec. (ours) & \textbf{2.41} & \textbf{2.03} & \textbf{0.65} & \textbf{4.66} \\ \hline
\end{tabular}}
\end{table}

\subsection{Additional Applications} \label{sec:applications}

\begin{figure}[t]
\includegraphics[width=\linewidth]{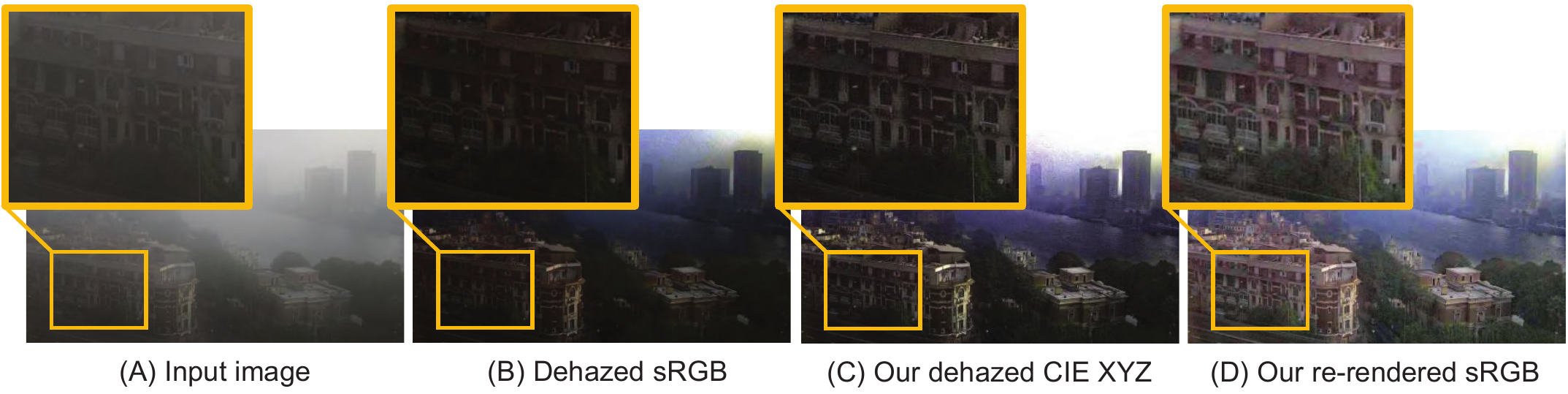}
\vspace{-7mm}
\caption[Dehazing is one of the potential applications that can benefit from our unprocessing method.]{Dehazing is one of the potential applications that can benefit from our unprocessing method. (A) Input image taken from Flickr (by Mike Rivera, CC BY-NC-SA 2.0). (B) Dehazing applied in sRGB space. (C) Dehazing applied to our CIE XYZ image. (D) Our final result in sRGB space. In this example, we used the dehazing method from \cite{he2010single}.}
\label{xyz:fig:dehaze}
\end{figure}

A hazy image is expressed using a linear model as  $\mat{I}(\mat{x})=\mat{J}(\mat{x})t(\mat{x})+\mat{A}(1-t(\mat{x}))$~\cite{he2010single}, where $\mat{I}$ is the observed intensity, $\mat{J}$ is the scene radiance, $\mat{A}$
is the global atmospheric light, and $t$ is the medium transmission describing the portion of the light that is not scattered and reaches the camera. Just as with motion deblurring, this linear relationship is broken by the camera's photo-finishing stages. Therefore, it is desirable to perform dehazing on linearized images.
In Fig. \ref{xyz:fig:dehaze}, we show the result of dehazing an sRGB image versus dehazing our linear CIE XYZ image and then re-rendering to sRGB. The improvement in visual quality can be clearly observed from the zoomed-in regions.

\begin{figure}[!t]
\includegraphics[width=\linewidth]{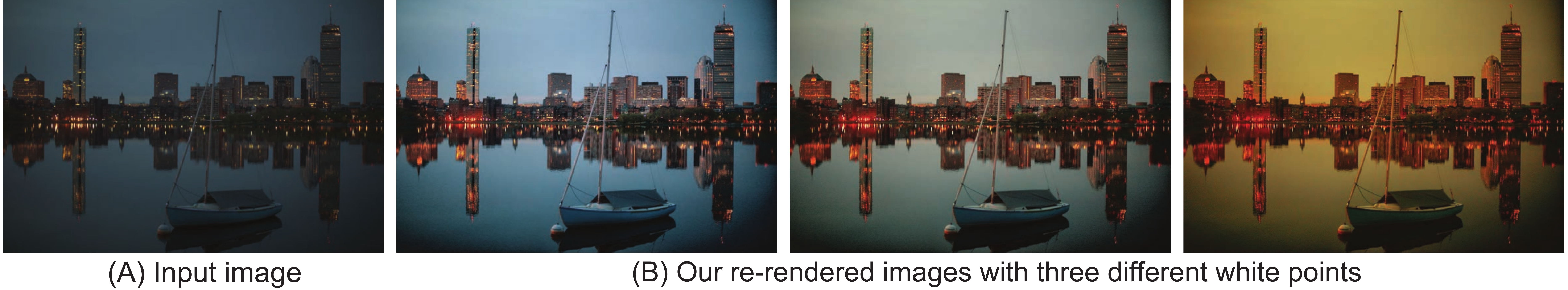}
\vspace{-7mm}
\caption[(A) Input sRGB rendered image. (B) Our re-rendered images after enhancement.]{(A) Input sRGB rendered image. (B) Our re-rendered images after enhancement. In this example, we applied chromatic adaptation to three different reference white points. Input image is taken from the under-exposure set \cite{DeepUPE} of the MIT-Adobe FiveK dataset \cite{bychkovsky2011learning}.}
\label{fig:chromAdapt}
\end{figure}

\begin{figure}[!t]
\includegraphics[width=\linewidth]{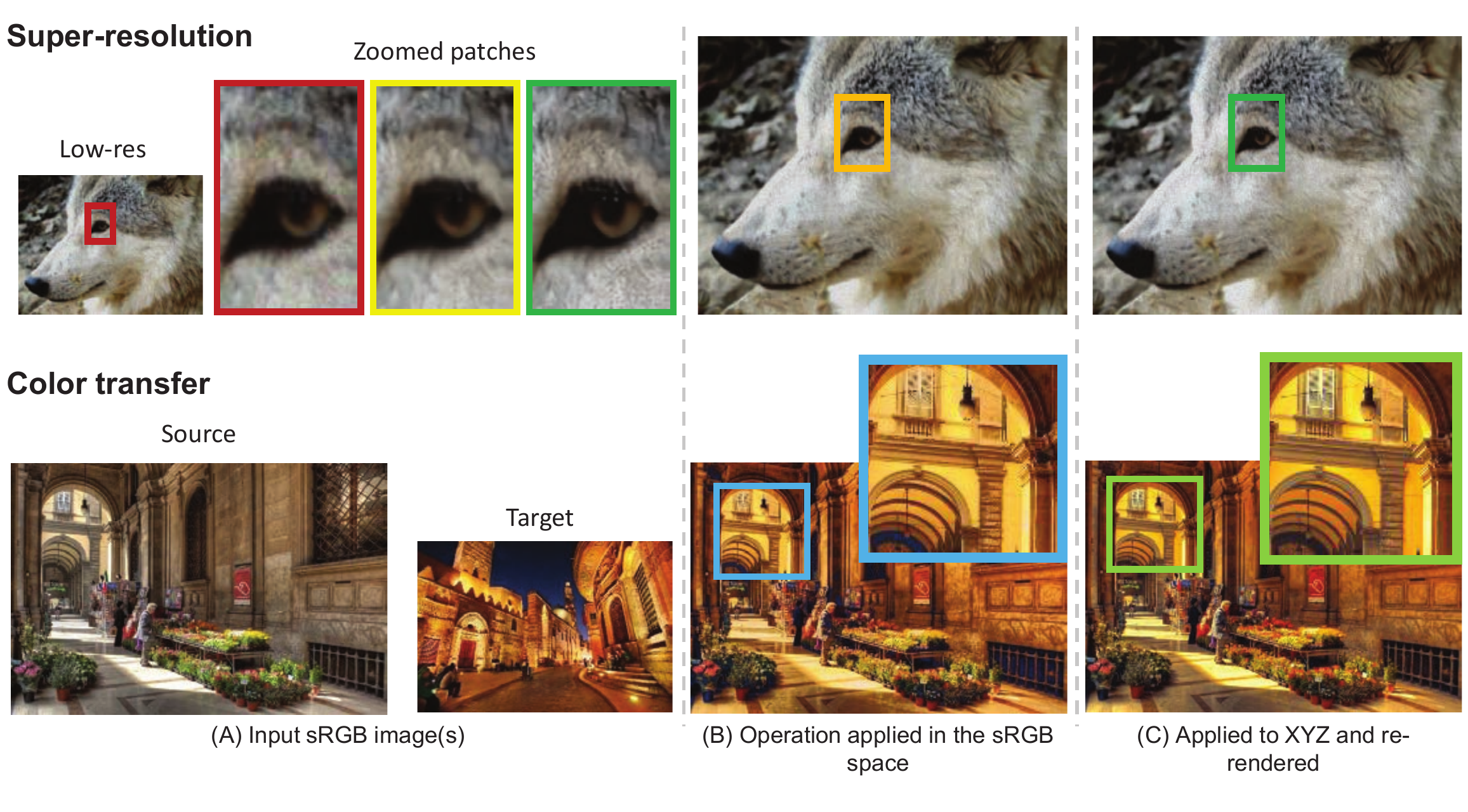}
\vspace{-7mm}
\caption[Additional potential applications of our method.]{Additional potential applications of our method. (A) The input sRGB image. (B) Super-resolution and color transfer applied in the sRGB space. (C) Super-resolution and color transfer applied in our reconstructed CIE XYZ space followed by re-rendering. In this example, we used the deep learning super-resolution model proposed in \cite{zhang2018learning} and the color transfer method in \cite{Pitie2007}. The input image in the first row is taken from the DIV2K dataset \cite{Agustsson_2017_CVPR_Workshops, Timofte_2018_CVPR_Workshops}, while the second input image is taken from Flickr--CC BY-NC 2.0 (by Chris Ford and Giuseppe Moscato, respectively).}
\label{fig:otherapplication}
\end{figure}

\begin{figure}[!t]
\includegraphics[width=\linewidth]{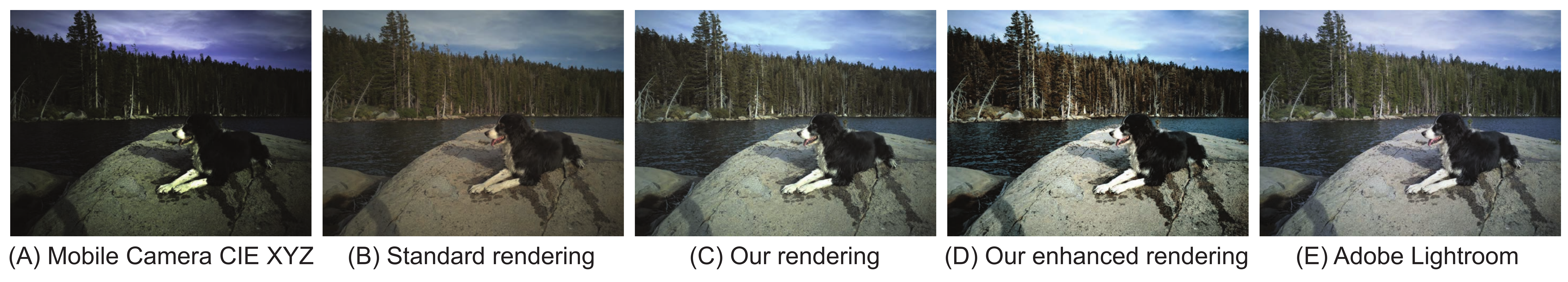}
\vspace{-7mm}
\caption[Our rendering network generalizes well for unseen CIE XYZ input images and produces pleasing results that are close to Adobe Lightroom's quality.]{Our rendering network generalizes well for unseen CIE XYZ input images and produces pleasing results that are close to Adobe Lightroom's quality. (A) Input smartphone camera CIE XYZ image. (B) Standard rendering \cite{anderson1996proposal, ebner2007color}. (C) Our rendering. (D) Our rendering after enhancing the local layer using the local Laplacian filter \cite{paris2011local}. (E) Adobe Lightroom rendering. To aid visualization, CIE XYZ images are scaled by a factor of two. Input image is taken from the HDR+ burst photography dataset \cite{hasinoff2016burst}.}
\label{fig:mobileCIE-XYZ}
\end{figure}

Another application of our CIE XYZ reconstructed images is chromatic adaptation. When we work in our reconstructed space (i.e., XYZ), we have a sound interpretation of post-capture white-balance editing using standard white points (e.g., D65, D50) and standard chromatic adaptation transforms (e.g., Bradford CAT \cite{hunt2011metamerism}, Sharp CAT \cite{finlayson1994spectral}), which are originally designed to work in the camera CIE XYZ space. Fig. \ref{fig:chromAdapt} shows examples of our enhanced rendering with applying chromatic adaptation \cite{finlayson1994spectral} in our reconstructed XYZ space.

Additional potential applications are shown in Fig. \ref{fig:otherapplication}. In the first row of Fig. \ref{fig:otherapplication}, we show super-resolution results obtained directly by working in the sRGB space and in our reconstructed CIE XYZ space followed by applying our re-rendering process. The last row of Fig. \ref{fig:otherapplication} shows an arguably better color transfer result by applying the color transfer process in our reconstructed space.

Lastly, our rendering network can be used as an alternative way to produce aesthetic photographs from raw-RGB DNG files, as shown in Fig. \ref{fig:mobileCIE-XYZ}. In this example, we first used the DNG metadata to map the raw-RGB values into the CIE XYZ space. Then, we used our rendering network and a local Laplacian filter to generate the shown output images.

\chapter{Data Augmentation for Color Constancy\label{ch:appendix1}}

In this appendix, we describe in detail the data augmentation procedure described in Chapter \ref{ch:ch6}. We begin with the steps used to map a color temperature to the corresponding CIE XYZ value. We then elaborate on the process required to map from camera sensor raw to the CIE XYZ color space. Afterwards, we describe the details of the scene retrieval process mentioned in Chapter \ref{ch:ch6}. Finally, we discuss experiments performed to evaluate our data augmentation and compare it with other color constancy augmentation techniques used in the literature.

\section{From Color Temperature to CIE XYZ} \label{C5:subsec:temp_to_xyz}

According to Planck's radiation law \cite{wyszecki1982color}, the SPD of a blackbody radiator at a given wavelength range $[\lambda, \partial\lambda]$ can be computed using the color temperature $q$ as follows:

\begin{equation}\label{C5:xyz_conv:Eq.1}
S_{\lambda}d_{\lambda}= \frac{f_1\lambda^{-5}}{\exp{(f_2/\lambda q}) - 1}\partial\lambda,
\end{equation}

\noindent  where, $f_1= 3.741832^10^{-16}$ $Wm^2$ is the first radiation constant,  $f_2=1.4388^{10-2} mK$ is the second radiation constant, and $q$ is the blackbody temperature, in Kelvin. \cite{johnwalker, li2016accurate}. Once the SPD is computed, the corresponding CIE tristimulus values can be approximated in the following discretized form:

\begin{equation} \label{C5:xyz_conv:Eq.2}
X =\Delta\lambda \sum_{\lambda=380}^{\lambda=780}x_{\lambda} S_{\lambda},
\end{equation}

\noindent where the value of $x_{\lambda}$ is the standard CIE color match value \cite{cie1932commission}. The values of $Y$ and $Z$ are computed similarly. The corresponding chromaticity coordinates of the computed XYZ tristimulus are finally computed as follows:
\begin{equation}
\begin{aligned} \label{C5:xyz_conv:Eq.3}
x = X/(X+Y+Z),\\
y = Y/(X+Y+Z),\\
z = Z/(X+Y+Z).\\
\end{aligned}
\end{equation}

\section{From Raw to CIE XYZ} \label{C5:subsec:raw_to_xyz}

Most DSLR cameras provide two pre-calibrated matrices, $C_1$ and $C_2$, to map from the camera sensor space to the CIE 1931 XYZ 2-degree standard observer color space. These pre-calibrated CST matrices are usually provided as a low color temperature (e.g., Standard-A) and a higher correlated color temperature (e.g., D65) \cite{DNG}.  

Given an illuminant vector $\mat{\light}$, estimated by an illuminant estimation algorithm, the CIE XYZ mapping matrix associated with $\mat{\light}$ is computed as follows \cite{can2018improving}:

\begin{equation}\label{C5:raw_to_xyz_conv:Eq.1}
C_{T_{\mat{\light}}}=C_2 + \left(1-\alpha\right)C_1,
\end{equation}
\begin{equation}\label{C5:raw_to_xyz_conv:Eq.2}
\alpha=(1/q_{\mat{\light}} - 1/q_1)/(1/q_2 - 1/q_1),
\end{equation}

\noindent  where $q_1$ and $q_2$ are the correlated color temperature associated to the pre-calibrated matrices $C_1$ and $C_2$, and $q_{\mat{\light}}$ is the color temperature of the illuminant vector $\mat{\light}$. Here, $q_{\mat{\light}}$ is unknown, and unlike the standard mapping from color temperature to the CIE XYZ space (Sec.~\ref{C5:subsec:temp_to_xyz}), there is no standard conversion from a camera sensor raw space to the corresponding color temperature. Thus, the conversion from the sensor raw space to the CIE XYZ space is a chicken-and-egg problem---computing the correlated color temperature $q_{\mat{\light}}$ is necessarily to get the CST matrix $C_{q_{\mat{\light}}}$, while knowing the mapping from a camera sensor raw to the CIE XYZ space inherently requires knowledge of the correlated color temperature of a given raw illuminant.

This problem can be solved by a trial-and-error strategy as follows. We iterate over the color temperature range of 2500K to 7500K. For each color temperature $q_i$ , we first compute the corresponding CST matrix $C_{q_i}$ using Eqs. \ref{C5:raw_to_xyz_conv:Eq.1} and \ref{C5:raw_to_xyz_conv:Eq.2}. Then, we convert $q_i$ to the corresponding xyz chromaticity triplet using Eqs. \ref{C5:xyz_conv:Eq.1}--\ref{C5:xyz_conv:Eq.3}. 

Afterwards, we map the xyz chromaticity triplet to the sensor raw space using the following equation:

\begin{equation}\label{C5:raw_to_xyz_conv:Eq.3}
\mat{\light}_{\texttt{raw}(q_i)} = C_{q_i}^{-1} \lambda_{\texttt{xyz}(q_i)}.
\end{equation}

We repeated this process for all color temperatures and selected the color temperature/CST matrix that achieves the minimum angular error between $\mat{\light}$ and the reconstructed illuminant color in the sensor raw space. 

The accuracy of our conversion depends on the pre-calibrated matrices provided by the manufacturer of the DSLR cameras. Other factors that may affect the accuracy of the mapping includes the precision of the standard mapping from color temperature to XYZ space defined by \cite{cie1932commission}, and the discretization process in Eq. \ref{C5:xyz_conv:Eq.2}.

\section{Raw-to-raw mapping}

Here, we describe the details of the mapping mentioned in Chapter \ref{ch:ch6}. Let $A$=$\{\mat{a}_1, \mat{a}_2, ...\}$ represent the ``source'' set of demosaiced raw images taken by different camera models with the associated capture metadata. Let $T=\{\mat{t}_1, \mat{t}_2 , ...\}$ represent our ``target'' set of metadata of captured scenes by the target camera model. Here, the capture metadata includes exposure time, aperture size, ISO gain value, and the global scene illuminant color in the camera sensor space. We also assume that we have access to the pre-calibration CST matrices for each camera model in the sets $A$ and $T$ (available in most DNG files of DSLR images \cite{DNG}).

Our goal here is to map all raw images in $A$, taken by different camera models, to the target camera sensor space in $T$. To that end, we map each image in $A$ to the device-independent CIE XYZ color space \cite{cie1932commission}. This mapping is performed as follows. We first compute the correlated color temperature, $q^{(i)}$, of the scene illuminant color vector, $\mat{\light}^{(i)}_{\texttt{raw}(A)}$, of each raw image, $I^{(i)}_{\texttt{raw}(A)}$, in the set $A$ (see Sec. \ref{C5:subsec:raw_to_xyz}). 
Then, we linearly interpolate between the pre-calibrated CST matrices provided with each raw image to compute the final CST mapping matrix, $C_{q^{(i)}}$, \cite{can2018improving}. 
Afterwards, we map each image, $I^{(i)}_{\texttt{raw}(A)}$, in the set $A$ to the CIE XYZ space. Note that here we represent each image $I$ as matrices of the color triplets (i.e., $I = \{\mat{c}^{(k)}\}$), where $k$ is the total number of pixels in the image $I$. We map each raw image to the CIE XYZ space as follows:

\begin{equation}\label{C5:data_aug:Eq.1}
I^{(i)}_{\texttt{xyz}(A)}= C_{q^{(i)}}D_{\mat{\light}^{(i)}}I^{(i)}_{\texttt{raw}(A)},
\end{equation}

\noindent where  $D_{\mat{\light}^{(i)}}$ is the white-balance diagonal correction matrix constructed based on the illuminant vector $\mat{\light}^{(i)}_{\texttt{raw}(A)}$.

Similarly, we compute the inverse mapping from the CIE XYZ space back to the target camera sensor space based on the illuminant vectors and pre-calibration matrices provided in the target set $T$. The mapping from the source sensor space to the target one in $T$ can be performed as follows:

\begin{equation}\label{C5:data_aug:Eq.2}
I^{(i)}_{\texttt{raw}(T)}= D^{-1}_{	\jmath^{(i)}}M^{-1}_{q^{(i)}}I^{(i)}_{\texttt{xyz}(A)},
\end{equation}

\noindent where $\jmath^{(i)}_{\texttt{raw}(T)}$ is the corresponding illuminant color to the correlated color temperature, $q^{(i)}$, in the target sensor space (i.e., the ground-truth illuminant for image $I^{(i)}_{\texttt{raw}(T)}$ in the illuminant estimation task), and $M^{-1}_{q^{(i)}}$ is the CST matrix that maps from the target sensor space to the CIE XYZ space. 

The described steps so far assume that the spectral sensitivities of all sensors in $A$ and $T$ satisfy the Luther condition\footnote{The Luther condition is satisfied when camera spectral sensitivities are linearly related to the CIE XYZ space.} \cite{nakamura2017image}. Prior studies, however, showed that this assumption is not always satisfied, and this can affect the accuracy of the pre-calibration matrices \cite{jiang2013space, karaimer2019beyond}. According to this, we rely on Eqs. \ref{C5:data_aug:Eq.1} and \ref{C5:data_aug:Eq.2} only to map the original colors of captured objects in the scene (i.e., white-balanced colors) to the target camera model. For the values of the global color cast, $\jmath^{(i)}_{\texttt{raw}(T)}$, we do not rely on $M^{-1}_{q^{(i)}}$ to map $\mat{\light}^{(i)}_{\texttt{raw}(A)}$ to the target sensor space of $T$. Instead, we follow a $K$-nearest neighbor strategy to get samples from the target sensor's illuminant color space.

\section{Scene Sampling} \label{C5:subsec:scene_retrieval}
As described in Chapter \ref{ch:ch6}, we retrieve metadata of similar scenes in the target set $T$ for illuminant color sampling. This sampling process should consider the source scene capture conditions to sample suitable illuminant colors from the target camera model space---i.e., having indoor illuminant colors as ground-truth for outdoor scenes may affect the training process. To this end, we introduce a retrieval feature $v_A^{(i)}$ to represent the capture settings of the image $I^{(i)}_{\texttt{raw}(A)}$. This feature includes the correlated color temperature and auxiliary capture settings. These additional capture settings are used to retrieve scenes captured with similar settings of $I^{(i)}_{\texttt{raw}(A)}$. 

Our feature vector is defined as follows:
\begin{equation}\label{C5:data_aug:Eq.3}
v_A^{(i)} = [q_{\texttt{norm}}^{(i)} \,,  h_{\texttt{norm}}^{(i)}, p_{\texttt{norm}}^{(i)}\,, e_{\texttt{norm}}^{(i)}],  
\end{equation}
where $q_{\texttt{norm}}^{(i)}$, $ h_{\texttt{norm}}^{(i)}$, $p_{\texttt{norm}}^{(i)}$, and $e_{\texttt{norm}}^{(i)}$ are the normalized color temperature, gain value, aperture size, and scaled exposure time, respectively. The gain value and the scaled exposure time are computed as follows:
\begin{equation}\label{C5:data_aug:Eq.4}
h^{(i)} = \texttt{BLN}^{(i)} \texttt{ISO}^{(i)}\,,  
\end{equation}
\begin{equation}\label{C5:data_aug:Eq.5}
e^{(i)} = \sqrt{2^{\texttt{BLE}^{(i)}}} l^{(i)}\,, 
\end{equation}
where $\texttt{BLE}$, $\texttt{BLN}$, $\texttt{ISO}$, and $l$ are the baseline exposure, baseline noise, digital gain value, and exposure time (in seconds), respectively.

\paragraph{Illuminant Color Sampling}

A naive sampling from the associated illuminant colors in $T$ does not introduce new illuminant colors over the Planckian locus of the target sensor. For this reason, we first fit a cubic polynomial to the $rg$ chromaticity of illuminant colors in the target sensor $T$. Then, we compute a new $r$ chromaticity value for each query vector as follows:
\begin{equation}\label{C5:data_aug:Eq.6}
\centering
r_v = \sum_{j\in K}{w_jr_j + x}\,,
\end{equation}
where $w_j = \exp(1-d_j)/\sum_{k}^{K}{\exp(1-d_k)}$ is a weighting factor,  $x=\lambda_r \mathcal{N}(0, \sigma_r)$ is a small random shift, $\lambda_r$ is a scalar factor to control the amount of divergence from the ideal Planckian curve, $\sigma_r$ is the standard deviation of the $r$ chromaticity values in the retrieved $K$ metadata of the target camera model, $T_K$, and $d_j$ is the normalized L2 distance between $v_{S(i)}$ and the corresponding $j^\texttt{th}$ feature vector in $T_K$, respectively. In our experiments, we retrieved the nearest 15 sample in from target camera model to our retrieval feature $v_A^{(i)}$ (i.e., $K$ includes 15 samples from the target camera model). The CST matrix $M$ (Eq. \ref{C5:data_aug:Eq.2}) is constructed by linearly interpolating between the corresponding CST matrices associated with each sample in $T_K$ using $w_j$. After computing $r_v$, the corresponding $g$ chromaticity value is computed as:
\begin{equation}\label{C5:data_aug:Eq.8}
g_v=[r_v, r^2_v, r^3_v] [\xi_1, \xi_2, \xi_3]^\top + y\,,   
\end{equation}
where $[\xi_1, \xi_2, \xi_3]$ are the cubic polynomial coefficients, $y$ is a random shift, and $\sigma_g$ is the standard deviation of the $g$ chromaticity values in $T_K$. In our experiments, we set $\lambda_r = 0.7$ and $\lambda_g = 1$. The final illuminant color $\jmath^{(i)}_{\texttt{raw}(T)}$ can be represented as follows:
\begin{equation}\label{C5:data_aug:Eq.9}
\jmath^{(i)}_{\texttt{raw}(T)} = [r_v, g_v, 1 - r_v- g_v]^\top\,.
\end{equation}

\begin{figure}
\includegraphics[width=\linewidth]{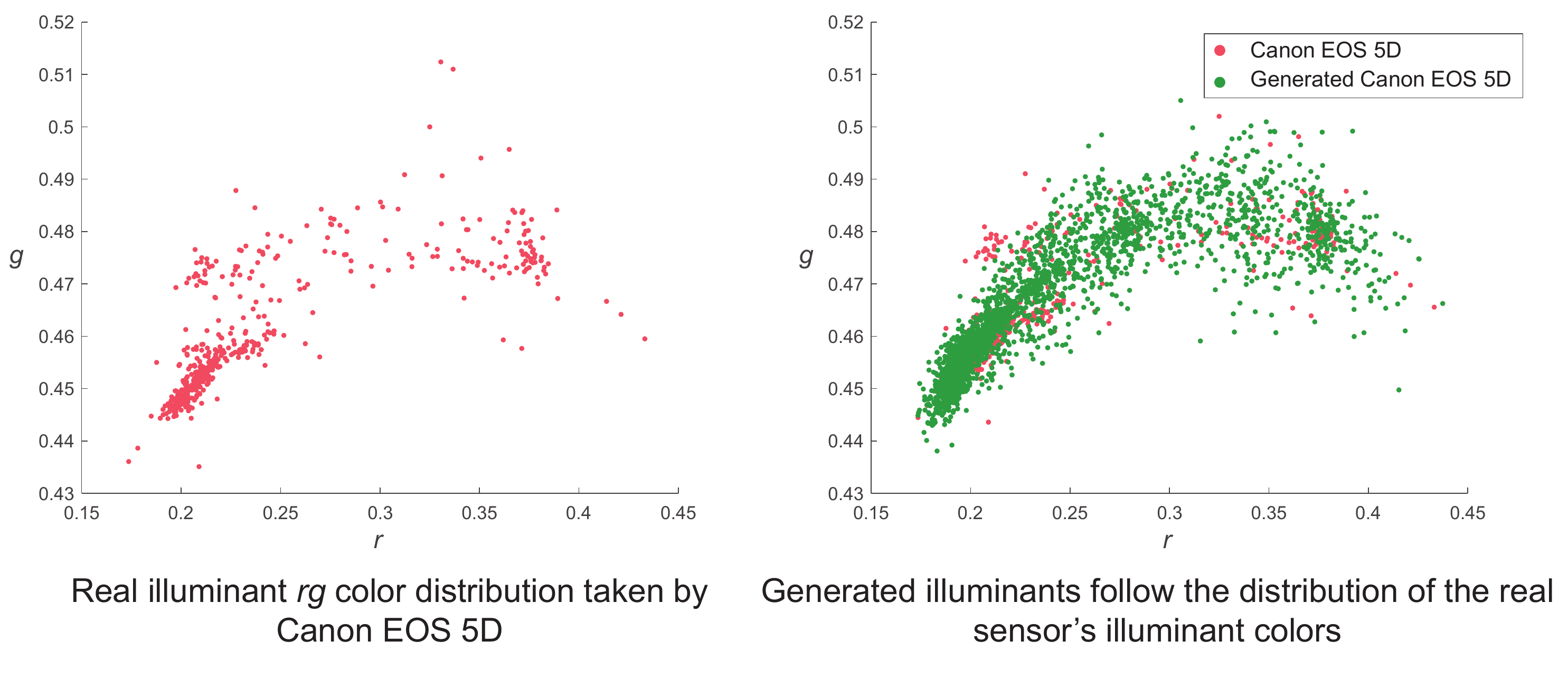}
\vspace{-7mm}
\caption[Synthetic illuminant samples of Canon EOS 5D camera model in the Gehler-Shi dataset \cite{gehler2008bayesian}.]{Synthetic illuminant samples of Canon EOS 5D camera model in the Gehler-Shi dataset \cite{gehler2008bayesian}. The shown generated illuminant colors are then applied to sensor-mapped raw images, originally were taken by different camera models, for augmentation purpose (Chapter \ref{ch:ch6}).}
\label{C5:fig:synthetic_ill_plot}
\end{figure}

\begin{figure}
\includegraphics[width=\linewidth]{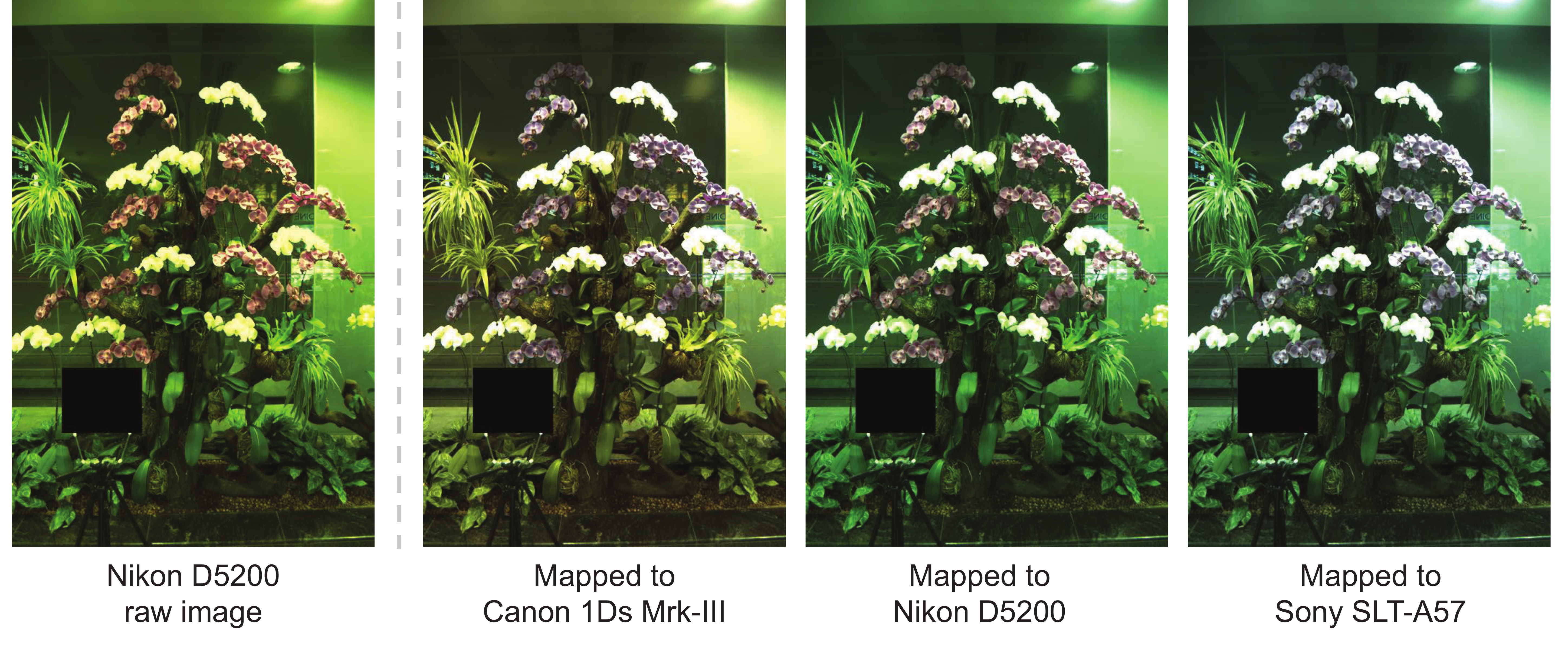}
\vspace{-7mm}
\caption[Example of camera augmentation used to train our network.]{Example of camera augmentation used to train our network. The shown left raw image is captured by Nikon D5200 camera \cite{cheng2014illuminant}. The next three images are the results of our mapping to different camera models. \label{C5:fig:augmented_images_example}}

\end{figure}

To avoid any bias towards the dominant color temperature in the source set, $A$, we first divide the color temperature range of the source set $A$ into different groups with a step of 250K. Then, we uniformly sample examples from each group to avoid any bias towards specific type of illuminants. Figure \ref{C5:fig:synthetic_ill_plot} shows examples of the sampling process. As shown, the sampled illuminant chromaticity values follow the original distribution over the Planckian curve, while introducing new illuminant colors of the target sensors that were not included in the original set. Finally, we apply random cropping to introduce more diversity in the generated images. Figure \ref{C5:fig:augmented_images_example} shows examples of synthetic raw-like images of different target camera models.

\section{Evaluation} \label{C5:subsec:synthetic_data_eval}

\begin{table}[t]
\caption[A comparison of different augmentation methods for illuminant estimation.]{A comparison of different augmentation methods for illuminant estimation. All results were obtained by using training images captured by the Canon EOS 5D camera model \cite{gehler2008bayesian} as the source and target sets for augmentation. Lowest errors are highlighted in yellow.\label{C5:table:augmentation_1}}
\centering
\scalebox{0.75}{
\begin{tabular}{l|c|c|c|c}
\textbf{Training set} & \textbf{Mean} & \textbf{Med.} & \textbf{B. 25\%}& \textbf{W. 25\%}\\ \hline
Original set  & 1.81 & 1.12 & 0.35 & 4.43 \\ 
Augmented (clustering \& sampling) \cite{BMVC1}  & 1.68 & \cellcolor[HTML]{\bestcolor}0.97 & \cellcolor[HTML]{\bestcolor}0.25	& 4.31\\
Augmented (sampling) \cite{fourure2016mixed}  & 1.79 & 1.09 & 0.33 & 4.34 
 \\ \hdashline
Augmented (ours)  & \cellcolor[HTML]{\bestcolor}1.55 & 0.98 & 0.28 & \cellcolor[HTML]{\bestcolor}3.68 \\
\end{tabular}}
\end{table}

\begin{table}[t]
\caption[A comparison of techniques for generating new sensor-mapped raw-like images that were originally captured by different sensors than the training camera model.]{A comparison of techniques for generating new sensor-mapped raw-like images that were originally captured by different sensors than the training camera model. The term `synthetic' refers to training FFCC \cite{FFCC} without including any of the original training examples, while the term `augmented' refers to training on synthetic and real images. The best results are bold-faced. Lowest errors of synthesized and augmented sets are highlighted in red and yellow, respectively.\label{C5:table:augmentation_2}}
\centering
\scalebox{0.75}{
\begin{tabular}{l|c|c|c|c}

\textbf{Training set} & \textbf{Mean} & \textbf{Med.} & \textbf{B. 25\%} & \textbf{W. 25\%} \\ \hline
Synthetic (Appendix \ref{ch:appendix0})  & 4.17 & 3.06 & 0.78 & 9.39\\ 
Augmentation (Appendix \ref{ch:appendix0}) & 2.64 & 1.95 & 0.45 & 5.97\\ \hdashline
Synthetic (ours)  & \cellcolor[HTML]{\secondbestcolor} 2.44 & \cellcolor[HTML]{\secondbestcolor} 1.89 & \cellcolor[HTML]{\secondbestcolor} 0.42 & \cellcolor[HTML]{\secondbestcolor} 5.40\\  
Augmented (ours)  & \cellcolor[HTML]{\bestcolor} \textbf{1.75} & \cellcolor[HTML]{\bestcolor} \textbf{1.28} & \cellcolor[HTML]{\bestcolor} \textbf{0.35} & \cellcolor[HTML]{\bestcolor} \textbf{4.15} \\ 
\end{tabular}}
\end{table}

\begin{table}[t]

\caption[Results of FFCC \cite{FFCC} trained on synthetic raw-like images after they are mapped to the target camera model.]{Results of FFCC \cite{FFCC} trained on synthetic raw-like images after they are mapped to the target camera model. In this experiment, the raw images are mapped from the Canon EOS-1Ds Mark III camera sensor (taken from the NUS dataset \cite{cheng2014illuminant}) to the target Canon EOS 5D camera in the Gehler-Shi dataset \cite{gehler2008bayesian}. The shown results were obtained with and without the intermediate CIE XYZ mapping step to generate the synthetic training set. Lowest errors are highlighted in yellow.\label{C5:table:CIEXYZ_ablation}}
\centering
\scalebox{0.75}{
\begin{tabular}{l|c|c|c|c}

\textbf{Synthetic training set} & \textbf{Mean} & \textbf{Med.} & \textbf{B. 25\%}& \textbf{W. 25\%} \\ \hline
w/o CIE XYZ  & 3.30 & 2.55 & 0.60 & 7.21 \\ 
w/ CIE XYZ & \cellcolor[HTML]{\bestcolor}3.04 &\cellcolor[HTML]{\bestcolor}2.36 & \cellcolor[HTML]{\bestcolor}0.56 & \cellcolor[HTML]{\bestcolor}6.58 \\ 
\end{tabular}}
\end{table}

\begin{table}[t]
\caption[Results of FFCC \cite{FFCC} trained on the Canon EOS 5D camera  \cite{gehler2008bayesian} and tested on images taken by different camera models from the NUS dataset \cite{cheng2014illuminant} and the Cube+ challenge set \cite{banic2017unsupervised}.]{Results of FFCC \cite{FFCC} trained on the Canon EOS 5D camera  \cite{gehler2008bayesian} and tested on images taken by different camera models from the NUS dataset \cite{cheng2014illuminant} and the Cube+ challenge set \cite{banic2017unsupervised}. The synthetic sets refer to testing images generated by our data augmentation approach, where these images were mapped from the Canon EOS 5D set (used for training) to the target camera models. \label{C5:table:sensor_errors}}
\centering
\scalebox{0.75}{
\begin{tabular}{l|c|c|c|c|c|c}

\multirow{2}{*}{\textbf{Testing sensor}} & \multicolumn{3}{c|}{\textbf{Real camera images}} & \multicolumn{3}{c}{\textbf{Synthetic camera images}} \\ \cline{2-7} 
 & \textbf{Mean} & \textbf{Med.} & \textbf{Max} & \textbf{Mean} & \textbf{Med.} & \textbf{Max} \\ \hline
Canon EOS 1D \cite{gehler2008bayesian}& 3.88 & 2.66 & 16.32 & 4.68 & 3.80 & 22.83 \\ 
Fujifilm XM1 \cite{cheng2014illuminant} & 4.22 & 3.05 & 47.87 & 2.91 & 2.06 & 38.93 \\
Nikon D5200 \cite{cheng2014illuminant}& 4.45 & 3.45 & 36.762 & 3.36  & 2.10 & 41.23 \\
Olympus EPL6 \cite{cheng2014illuminant}& 4.35 & 3.56 & 19.89 & 3.28 & 2.27 & 38.81  \\
Panasonic GX1 \cite{cheng2014illuminant}& 2.83 & 2.03 & 16.58 & 3.24 & 2.29 & 17.07 \\
Samsung NX2000 \cite{cheng2014illuminant}& 4.41 & 3.73 & 17.69 & 3.44 & 2.64 & 18.79 \\
Sony A57 \cite{cheng2014illuminant}& 3.84 & 3.02 & 19.38 & 3.04	 & 1.34 & 39.67 \\
Canon EOS 550D \cite{banic2017unsupervised}& 3.83 & 2.49 & 46.55 & 3.14 & 1.98 & 36.30 \\ 
\end{tabular}}
\end{table}

In prior work, several approaches for training data augmentation for illuminant estimation have been attempted \cite{BMVC1, fourure2016mixed}. These approaches first white-balance the training raw images using the associated ground-truth illuminant colors associated with each image. Afterwards, illuminant colors are sampled from the ``ground-truth'' illuminant colors over the entire training set to be applied to the white-balanced raw images. These sampled illuminant colors can be taken randomly from the ground-truth illuminant colors \cite{fourure2016mixed} or after clustering the ground-truth illuminant colors \cite{BMVC1}. These methods, however, are limited to using the same set of scenes as is present in the training dataset. Another approach for data augmentation has been proposed in Appendix \ref{ch:appendix0} by mapping sRGB white-balanced images to a learned normalization space that is is learned based on the CIE XYZ space. Afterwards, a pre-computed global transformation matrix is used to map the images from this normalization space to the target white-balanced raw space. In contrast, the augmentation method described in Chapter \ref{ch:ch6} uses an accurate mapping from the camera sensor raw space to the CIE XYZ using the pre-calibration matrices provided by camera manufacturers. 

In the following set of experiments, we use the baseline model FFCC \cite{FFCC} to study the potential improvement of our chosen data augmentation strategy and alternative augmentation techniques proposed in \cite{BMVC1, fourure2016mixed}. Additionally, we include the results of our augmentation discussed in Appendix \ref{ch:appendix0}. We use the Canon EOS 5D images from in the Gehler-Shi dataset \cite{gehler2008bayesian} for comparisons. For our test set, we randomly select 30\% of the total number of images in the Canon EOS 5D set. The remaining 70\% of images are used for training. We refer to this set as ``real training set'', which includes 336 raw images.

Note that, except for the augmentation proposed in Appendix \ref{ch:appendix0}, none of these methods apply a sensor-to-sensor mapping, as they use the raw images of the ``real training set'' as the source and target set for augmentation. For this reason and for a fair comparison, we provide the results of two different set of experiments. In the first experiment, we use the CIE XYZ images taken by the Canon EOS 5D sensor as our source set $A$, while in the second experiment, we use a different set of four sensors rather than the Canon EOS 5D sensor. The former is comparable to the augmentation methods used in \cite{BMVC1, fourure2016mixed} (see Table \ref{C5:table:augmentation_1}), while the latter is comparable to the augmentation approach proposed in Appendix \ref{ch:appendix0}, which performs ``raw mapping'' in order to introduce new scene content in the training data (see Table \ref{C5:table:augmentation_2}). The shown results obtained by generating 500 synthetic images by each augmentation method, including our augmentation approach. As shown in Tables \ref{C5:table:augmentation_1} and \ref{C5:table:augmentation_2}, our augmentation approach achieves the best improvement of the FFCC results.

In order to study the effect of the CIE XYZ mapping used by our augmentation approach, we trained FFCC \cite{FFCC} on a set of 500 synthetic raw images of the target camera model---namely, the Canon EOS 5D camera model in the Gehler-Shi dataset \cite{gehler2008bayesian}. These synthetic raw images were originally captured by the Canon EOS 1Ds Mark III camera sensor (in the NUS dataset \cite{cheng2014illuminant}), then these images are mapped to the target sensor using our augmentation approach. Table \ref{C5:table:CIEXYZ_ablation} shows the results of FFCC trained on synthetic raw images with and without the intermediate CIE XYZ mapping step (Eqs. \ref{C5:data_aug:Eq.1} and \ref{C5:data_aug:Eq.2}). As shown, using the CIE XYZ mapping achieves better results, which are further improved by increasing the scene diversity of the source set by including additional scenes from other datasets, as shown in Table \ref{C5:table:augmentation_2}.

For a further evaluation, we use our approach to map images from the Canon EOS 5D camera's set (the same set that was used to train the FFCC model) to different target camera models. Then, we trained and tested a FFCC model on these mapped images. This experiment was performed to gauge the ability of our data augmentation approach to have similar negative effects on camera-specific methods that were trained on a different camera model. To that end, we randomly selected 150 images from the Canon EOS 5D sensor set, which was used to train the FFCC model, as our source image set $A$. Then, we mapped these images to different target camera models using our approach. That means that the training and our synthetic testing set share the same scene content. We report the results in Table \ref{C5:table:sensor_errors}. We also report the testing results on real image sets captured by the same target camera models. As shown in Table \ref{C5:table:sensor_errors}, both real and synthetic sets negatively affect the accuracy of the FFCC model (see Table \ref{C5:table:augmentation_1} for results of the FFCC on a testing set taken by the same training sensor).

\chapter{White-Balance Augmentation for Image Relighting\label{ch:appendix2}}
Image relighting has multiple applications both in research and in practice, and is recently witnessing an increased interest. A single-image relighting method would allow aesthetic enhancement applications, such as photo montage of images taken under different illuminations, and illumination retouching without human expert work. Very importantly, in computer vision research image relighting can be leveraged for data augmentation, enabling the trained methods to be robust to changes in light source position or color temperature. It could also serve for domain adaptation, by normalizing input images to a unique set of illumination settings that the down-stream computer vision method was trained on. 

We employed our white-balance techniques for correction (Chapter \ref{ch:ch7}) and data augmentation (Chapter \ref{ch:ch8}) to develop our image relighting framework, Norm-Relighting-U-Net and illuminant setting estimation network\footnote{This work was published in \cite{helou2020aim}: Majed El Helou, Ruofan Zhou, Sabine Süsstrunk, Radu Timofte, Mahmoud Afifi,
Michael S Brown, et al. AIM 2020: Scene relighting and illumination estimation challenge. In European Conference on Computer Vision (ECCV) Workshops, 2020.}. Our frameworks achieved the Running-Up Award over all tasks in the AIM 2020 challenge for Scene relighting \cite{helou2020aim}.   The source code of our method is available in GitHub: \href{https://github.com/mahmoudnafifi/image_relighting}{https://github.com/mahmoudnafifi/image$\_$relighting}.

\section{Challenge Tasks}
The AIM2020 challenge includes three tasks, which are: (i) one-to-one relighting, (ii) one-to-any relighting, and (iii) Illumination settings estimation. In this section, we provide details of each task, the dataset and evaluation protocol used in this challenge. 

\paragraph{One-to-One Relighting} The relighting task is pre-determined and fixed for all validation and test samples. In other words, the objective is to manipulate an input image from one pre-defined set of illumination settings (namely, North, 6500K) to another pre-defined set (East, 4500K). The images are in $1024\!\times\!1024 $resolution, both input and output, and nothing other than the input image is provided.

\paragraph{Any-to-Any Relighting} This track is a generalization of the first track. The objective is to relight an input image (both color temperature and light source position manipulation) from any arbitrary illumination settings to any arbitrary illumination settings. The latter settings are dictated by a second input guide image, as in style transfer applications. The participants were allowed to make use of their solutions to the first two tracks to develop a solution for this track. The images are in $512\!\times\!512$ resolution to ease computations, as this track is very challenging.

\paragraph{Illumination Setting Estimation} The goal of this track is to estimate, from a single input image, the illumination settings that were used in rendering it. Given the input image, the output should estimate the color temperature of the illuminant as well as the orientation, i.e. the position of the light source. The input images are also
$1024\!\times\!1024$ and no other input is given than the 2D image.

\paragraph{Data}
We used the VIDIT dataset \cite{elhelou2020vidit, helou2020aim}, which is a well-controlled setup to provide full-reference evaluation through a set of virtual scenes. The VIDIT dataset contains 300 virtual scenes used for training, where every scene is captured 40 times in total: from 8 equally-spaced azimuthal angles, each lit with 5 different illuminants. Every image is $1024\!\times\!1024$, but the images are downsampled by a factor of 2, with bicubic interpolation over $4\!\times\!4$ windows.

\begin{figure}[t]
\includegraphics[width=\linewidth]{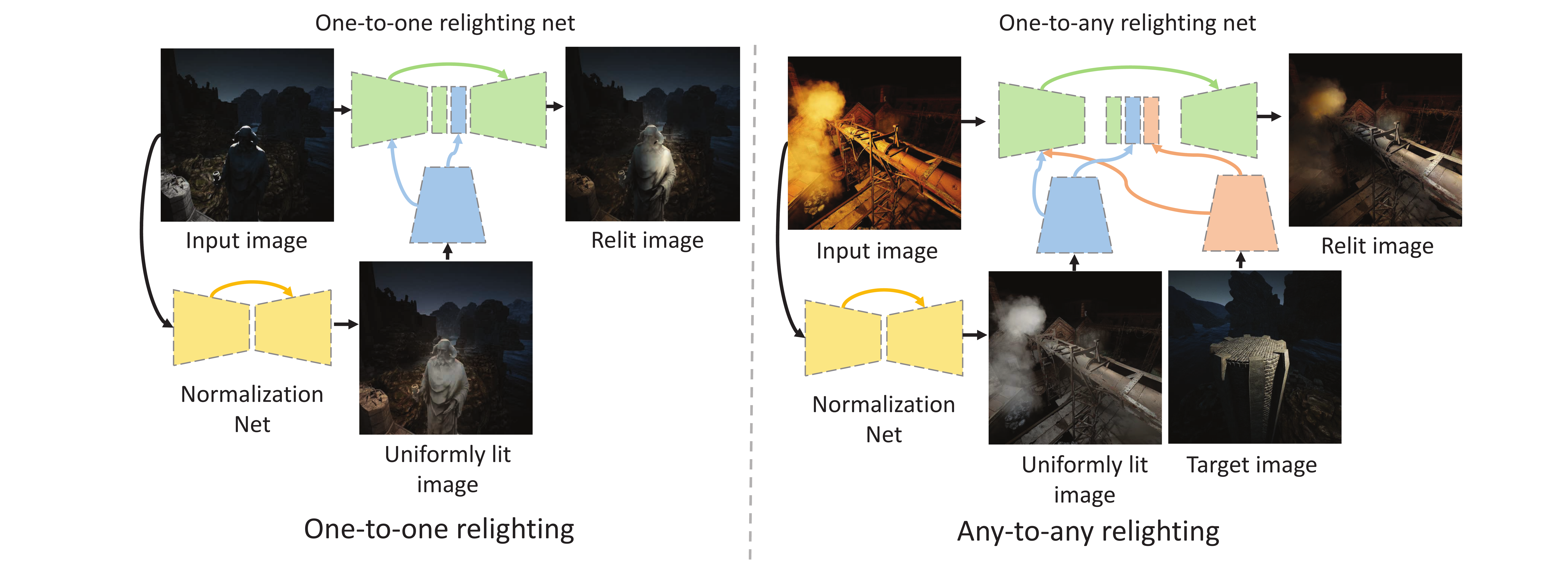}
\vspace{-7mm}
\caption[Overview of our Norm-Relighting-U-Net used for one-to-one and any-to-any relighting tasks.]{Overview of our Norm-Relighting-U-Net used for one-to-one and any-to-any relighting tasks.}
\label{fig:relighting}
\end{figure}

\paragraph{Evaluation Protocol} To evaluate the image relighting results, the PSNR and SSIM \cite{wang2004image} metrics are used. For the final ranking, the Mean Perceptual Score (MPS) is used. The MPS is defined as the average of the normalized SSIM and LPIPS \cite{zhang2018unreasonable} scores, themselves averaged across the entire test set. This MPS can be described by the following equation
\begin{equation}
0.5 (S + (1-L)),
\end{equation}
\noindent where $S$ is the SSIM score, and $L$ is the LPIPS score. While the evaluation of the illuminant setting estimation task is based on the accuracy of predictions following this formula for the loss:
\begin{equation}\label{AIM:eq}
\sqrt{\sum_{i=0}^{N-1}\left(\frac{|\hat{\phi_i} - \phi_i|mod180}{180}\right)^2 + (\hat{T_i} - T)^2},
\end{equation}
\noindent where $\hat{\phi_i}$ is the predicted angle (0-360) for test sample $i$, $\phi_i$ is the ground-truth value for that sample, $\hat{T_i}$ is the temperature prediction for test sample $i$, and $T_i$ is the ground-truth value for that sample. The temperature $T_i$ takes values equal to [0, 0.25, 0.5, 0.75, 1], which correspond to the color temperature values [2500K, 3500K, 4500K, 5500K, 6500K].

\begin{figure}[t]
\includegraphics[width=\linewidth]{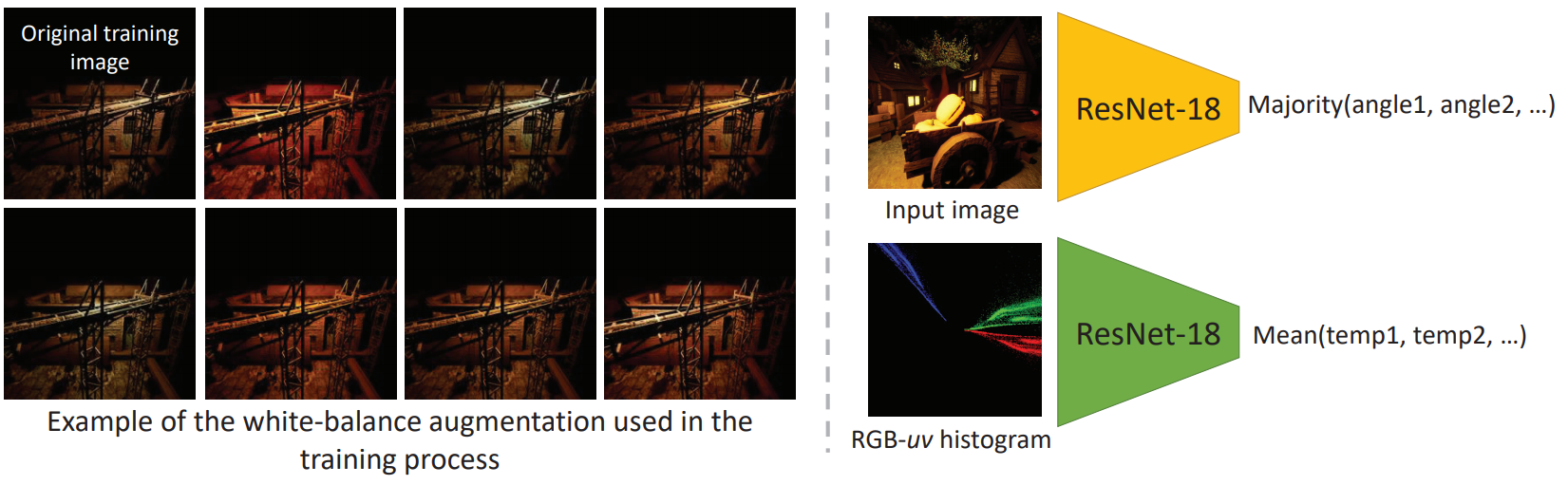}
\vspace{-7mm}
\caption{Overview of our Illuminant-ResNet, with the white-balance augmentation.}
\label{fig:AIMestimation}
\end{figure}

\section{Norm-Relighting-U-Net}
For image relighting tasks, we adopt a U-Net architecture \cite{ronneberger2015u} as the main backbone of our framework. Our framework consists of two networks: (i) the normalization net, which is responsible for producing uniformly lit white-balanced images, and (ii) the relighting network, which performs the image relighting task. We apply an instance normalization \cite{ulyanov2016instance} after each stage in the encoder of the normalization network, while we used batch normalization for the encoder of the relighting network. The relighting network is fed by the input image and the latent representations of the guide image and the uniformly lit image produced by our normalization network. We used our white-balance augmenter, described in Chapter \ref{ch:ch8}, to augment the training data used for the normalization network. To produce the ground-truth of the normalization network, we use the training data, which provide us with a set of images taken from each scene under different lighting directions. Specifically, we first white-balance all images using our KNN WB method described in Chapter \ref{ch:ch7}. Then, we compute the average image over all images of each scene set. We trained two models: the first model is trained on $256\!\times\!256$ random patches; the second model is trained on $256\!\times\!256$ resized images. The final result is generated by taking the mean of the two generated relit images. The details of our framework for each task are shown in Fig. \ref{fig:relighting}.

\section{Illuminant-ResNet}
For the illuminant setting task, we treat the task as two independent classification tasks: (i) illuminant temperature classification and (ii) illuminant angle classification. We adopt the ResNet-18 model \cite{he2016deep} trained on ImageNet \cite{deng2009imagenet}. The last fully connected layer is replaced with a new layer with $n$ neurons, where $n$ is the number of output classes for each task. The Adam optimizer \cite{kingma2014adam} is used with cross entropy loss. For angle classification, we applied our white-balance
augmenter described in Chapter \ref{ch:ch8} to augment the training data. For temperature classification, we use image histogram features instead of the 2D input image. Specifically, we feed the network with 2D RGB-uv projected histogram features (described in Chapter \ref{ch:ch7}), instead of the original training images. This histogram-based training, rather than image-based, improves
the model's generalization. Figure \ref{fig:AIMestimation} shows an overview of the team's solution, including the white-balance augmentation process.

\begin{table}[t]
\caption[Results of Image Relighting Challenge for the one-to-one relighting task.]{Results of Image Relighting Challenge for the one-to-one relighting task. The MPS, used to determine the final ranking.\label{table:AIM_results_1}}
\centering
\scalebox{0.8}{
\begin{tabular}{c|cccc}
Team & \textbf{MPS} $\uparrow$ & SSIM $\uparrow$ & LPIPS $\downarrow$ & PSNR $\uparrow$ \\ \hline
CET$\_$CVLab \cite{helou2020aim}& 0.6451 (1) & 0.6362 (1) & 0.3460 (3) & 16.8927 (6)  \\
lyl \cite{helou2020aim}& 0.6436 (2) & 0.6301 (3) & 0.3430 (2) & 16.6801 (8) \\
YorkU (ours) & 0.6216 (3) & 0.6091 (4) & 0.3659 (5) & 16.8196 (7) \\
IPCV$\_$IITM \cite{helou2020aim}& 0.5897 (4) & 0.5298 (7) & 0.3505 (4) & 17.0594 (3) \\
DeepRelight \cite{helou2020aim}& 0.5892 (5) &  0.5928 (6) & 0.4144 (7)  & 17.4252 (1) \\
Withdrawn \cite{helou2020aim}& 0.5603 (6)  & 0.5236 (8) & 0.4029 (6) & 16.5136 (9) \\
Hertz \cite{helou2020aim}& 0.5339 (7) & 0.5666 (6) & 0.4989 (8) & 16.9234 (4) \\
Image Lab \cite{helou2020aim}& 0.3746 (8) & 0.3769 (9) & 0.6278 (9) & 16.8949 (5)
\end{tabular}}
\end{table}

\begin{table}[t]
\caption[Results of Image Relighting Challenge for the any-to-any relighting task.]{Results of Image Relighting Challenge for the any-to-any relighting task. The MPS, used to determine the final ranking.\label{table:AIM_results_2}}
\centering
\scalebox{0.8}{
\begin{tabular}{c|cccc}
Team & \textbf{MPS} $\uparrow$ & SSIM $\uparrow$ & LPIPS $\downarrow$ & PSNR $\uparrow$ \\ \hline
NPU-CVPG \cite{helou2020aim}& 0.6484 (1) & 0.6353 (1) & 0.3386 (3) & 18.5436 (2) \\
YorkU (ours) & 0.6428 (2) & 0.6195 (2) & 0.3338 (2) & 18.2384 (4) \\
IPCV$\_$IITM \cite{helou2020aim}& 0.6424 (3) & 0.6042 (3) & 0.3194 (1) & 19.3559 (1) \\
lyl \cite{helou2020aim}& 0.6213 (4) & 0.5881 (4) & 0.3455 (4) & 17.6314 (5) \\
AiRiA$\_$CG \cite{helou2020aim}& 0.5258 (5) & 0.4451 (5) & 0.3936 (5) & 18.3493 (3) \\
RGETH \cite{helou2020aim}& 0.3465 (6) & 0.4123 (6) & 0.7192 (6) & 10.4483 (6)\\
\end{tabular}}
\end{table}

\begin{table}[t]
\caption[Results of Image Relighting Challenge for the Illumination setting estimation task.]{Results of Image Relighting Challenge for the Illumination setting estimation task. The loss is computed based on the angle and color temperature
predictions, as described in Eq. \ref{AIM:eq}.\label{table:AIM_results_3}}
\centering
\scalebox{0.8}{
\begin{tabular}{c|ccc}
Team & \textbf{Loss} $\downarrow$ & AngLoss $\downarrow$ & TempLoss $\downarrow$ \\ \hline
AiRiA$\_$CG \cite{helou2020aim}& 0.0875 (1) & 0.0722 (3) &  0.0153 (1) \\
YorkU (ours) & 0.0887 (2) & 0.0639 (2) & 0.0248 (2) \\
Image Lab \cite{helou2020aim}& 0.0984 (3) & 0.0513 (1) & 0.0471 (5) \\
debut$\_$kele \cite{helou2020aim}& 0.1431 (4) & 0.1125 (4) &  0.0306 (3) \\
RGETH \cite{helou2020aim}& 0.1708 (5) & 0.1347 (5) & 0.0361 (4)  \\
\end{tabular}}
\end{table}

\section{Results}
There were 20 teams participated in the AIM challenge for relighting and illuminant setting estimation \cite{helou2020aim}. Tables \ref{table:AIM_results_1}--\ref{table:AIM_results_3} shows the results for each task in the challenge. As can be seen, our method achieves the second/third rank over all tasks.

\chapter{Additional Details of HistoGANs\label{ch:appendix3}}
\begin{figure}[t]
\includegraphics[width=\linewidth]{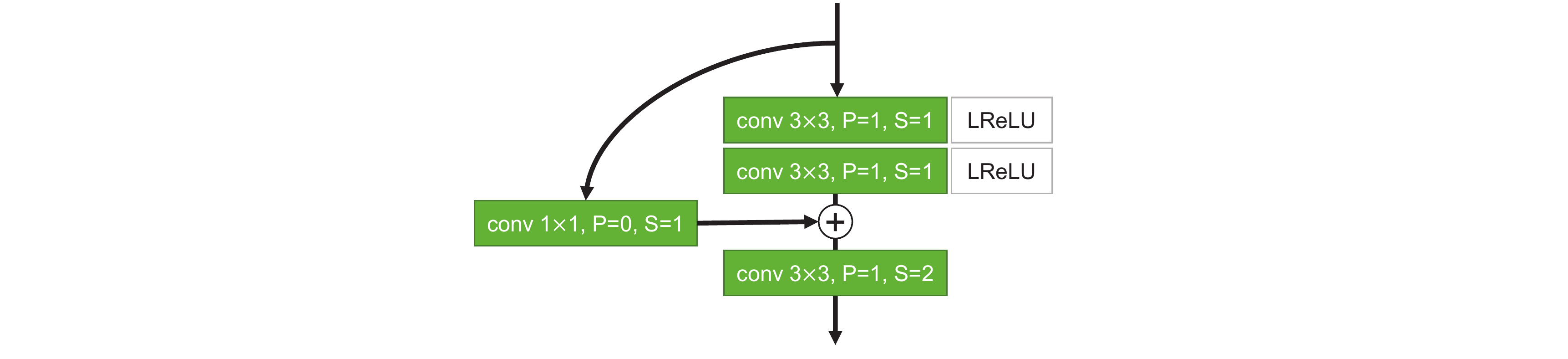}
\vspace{-7mm}
\caption[Details of the residual discriminator block used to reconstruct our discriminator network.]{Details of the residual discriminator block used to reconstruct our discriminator network. The term P and S refer to the padding and stride used in each layer.\label{histogran_appendix:fig:discriminator_block}}
\end{figure}

\begin{figure}[!t]
\includegraphics[width=0.95\linewidth]{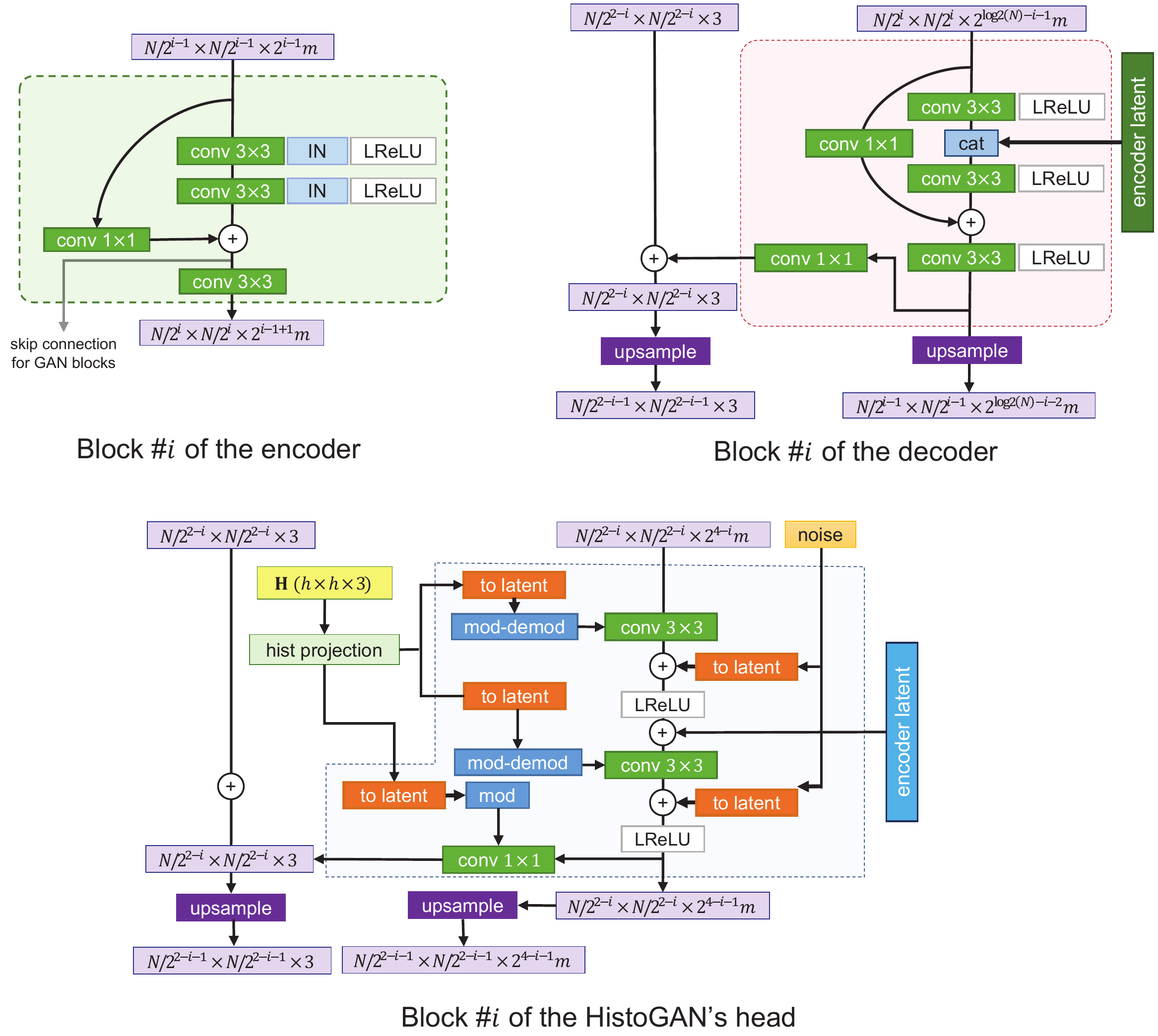}
\vspace{-1mm}
\caption[Details of our ReHistoGAN network.]{Details of our ReHistoGAN network. We modified the last two blocks of our HistoGAN by adding a gate for the processed skipped features from the first two blocks of our encoder.\label{histogran_appendix:fig:recoloring-design}}
\end{figure}

In this appendix, we provide additional details and results of our work described in Chapter \ref{ch:ch15}. We first discuss additional details of our networks in Sec.\ \ref{histogran_appendix:sec:network}. Then, we explain the training details in Sec.\ \ref{histogran_appendix:sec:training}. Afterwards, Sec.\ \ref{histogran_appendix:sec:ablations} presents ablation experiments carried out to validate our choice of hyperparameters and loss terms. In Sec.\ \ref{histogran_appendix:sec:universal}, we present our experiments performed to train a ``universal'' recoloring model to recolor images taken from arbitrary domains. Sec.\ \ref{histogran_appendix:sec:limitations} discusses failure cases of our method. Such failure cases can often be mitigated by applying simple post-processing. The post-processing details in Sec.\ \ref{histogran_appendix:sec:post-processing}. Sec.\ \ref{histogran_appendix:sec:post-processing} also discuss post-processing to deal with high-resolution images.

\section{Details of Our Networks} \label{histogran_appendix:sec:network}

Our discriminator network, used in all of our experiments, consists of a sequence of $\log_2(N) - 1$ residual blocks, where $N$ is the image width/height, and the last layer is an fully connected (fc) layer that produces a scalar feature. The first block accepts a three-channel input image and produce $m$ output channels. Then, each block $i$ produces $2m_{i-1}$ output channels (i.e., duplicate the number of output channels of the previous block). The details of the residual blocks used to build our discriminator network are shown in Fig.\ \ref{histogran_appendix:fig:discriminator_block}.

Figure~\ref{histogran_appendix:fig:recoloring-design} provides the details of our encoder, decoder and GAN blocks used in our ReHistoGAN (used for image recoloring). As shown, we modified the last two blocks of our HistoGAN's to accept the latent feature passed from the first two blocks of our encoder. This modification helps our HistoGAN's head to consider both information of the input image structure and the target histogram in the recoloring process.

\section{Training Details} \label{histogran_appendix:sec:training}

We train our networks using an NVIDIA TITAN X (Pascal) GPU. For HistoGAN training, we optimized both the generator and discriminator networks using the diffGrad optimizer \cite{8939562}. In all experiments, we set the histogram bin, $h$, to 64 and the fall-off parameter of our histogram's bins, $\tau$, was set to 0.02. We adopted the exponential moving average of generator network's weights \cite{karras2019style, karras2020analyzing} with the path length penalty, introduced in StyleGAN \cite{karras2020analyzing}, every 32 iterations to train our generator network. Due to the hardware limitation, we used mini-batch of 2 with accumulated gradients every 16 iteration steps and we set the image's dimension, $N$, to 256. We set the scale factor of the Hellinger distance loss, $\alpha$, to 2 (see Sec.\ \ref{histogran_appendix:sec:ablations} for an ablation study).

\begin{figure}[t]
\centering
\includegraphics[width=0.9\linewidth]{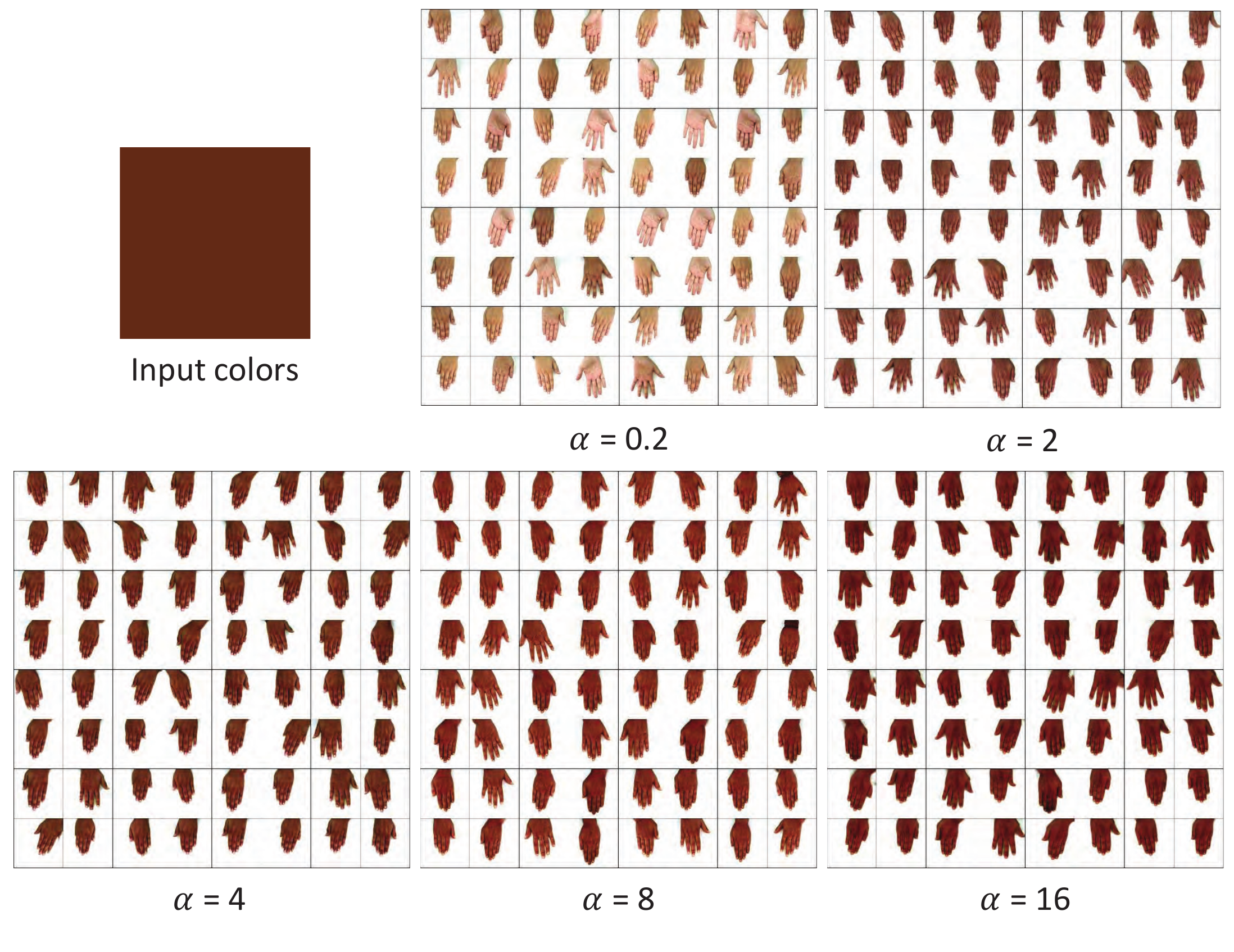}
\vspace{-1mm}
\caption[Results obtained by training our HistoGAN in hand images \cite{afifi201911k} using different values of $\alpha$.]{Results obtained by training our HistoGAN in hand images \cite{afifi201911k} using different values of $\alpha$.\label{histogran_appendix:fig:alpha_ablation}}
\end{figure}

As mentioned in Chapter \ref{ch:ch15}, we trained our HistoGAN using several domain datasets, including: human faces \cite{karras2019style}, flowers \cite{nilsback2008automated}, cats \cite{catdataset}, dogs \cite{khosla2011novel}, birds \cite{wah2011caltech}, anime faces \cite{animedataset}, human hands \cite{afifi201911k}, bedrooms \cite{yu2015lsun}, cars \cite{krause20133d}, and aerial scenes \cite{maggiori2017can}. We further trained our HistoGAN using 4,316 landscape images collected from Flickr. The collected images have one of the following copyright licenses: no known copyright restrictions, Public Domain Dedication (CC0), or Public Domain Mark. 

\begin{figure}[t]
\includegraphics[width=\linewidth]{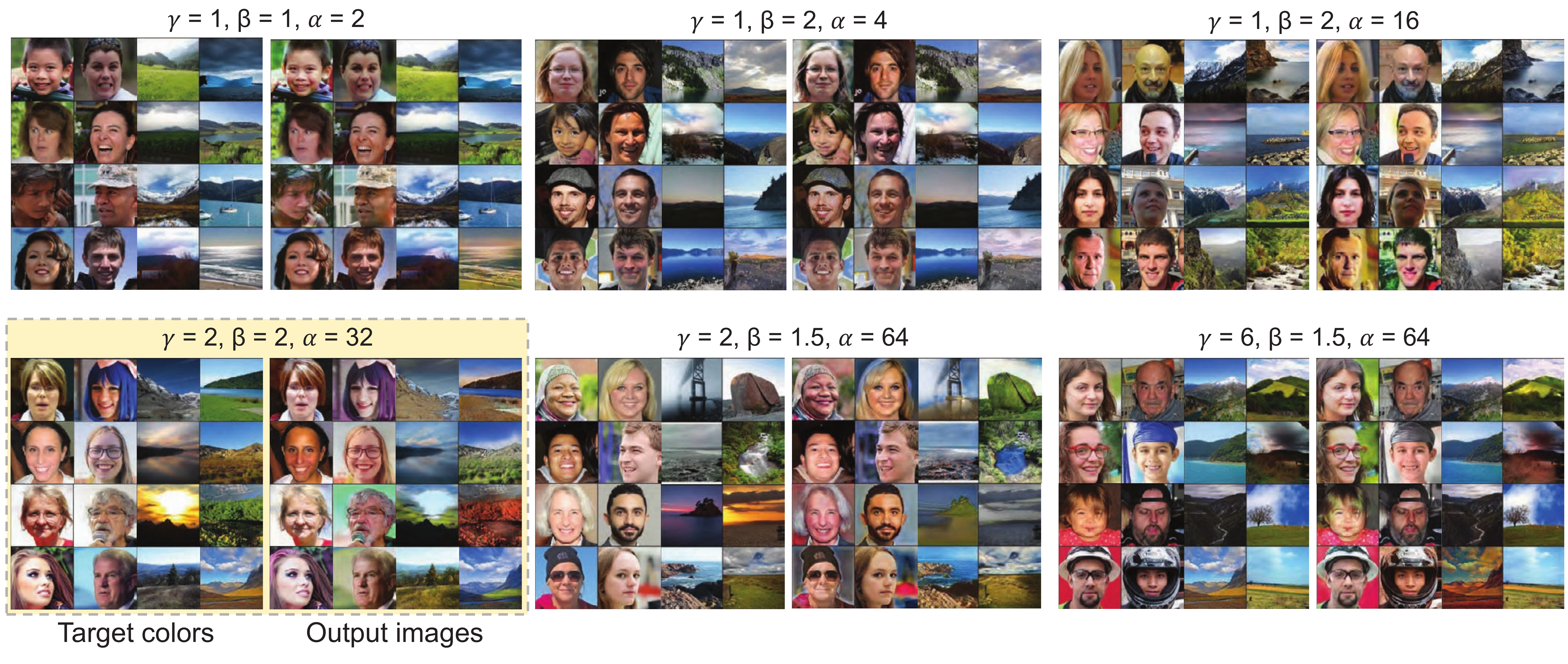}
\vspace{-7mm}
\caption[Results of recoloring by training our recoloring network using different values of $\alpha$, $\beta$, and $\gamma$ hyperparameters.]{Results of recoloring by training our recoloring network using different values of $\alpha$, $\beta$, and $\gamma$ hyperparameters. The highlighted results refer to the settings used to produce the reported results in Chapter \ref{ch:ch15} and this appendix.\label{histogran_appendix:fig:alpha_beta_gamma_ablation}}
\end{figure}

\begin{figure}[t]
\includegraphics[width=\linewidth]{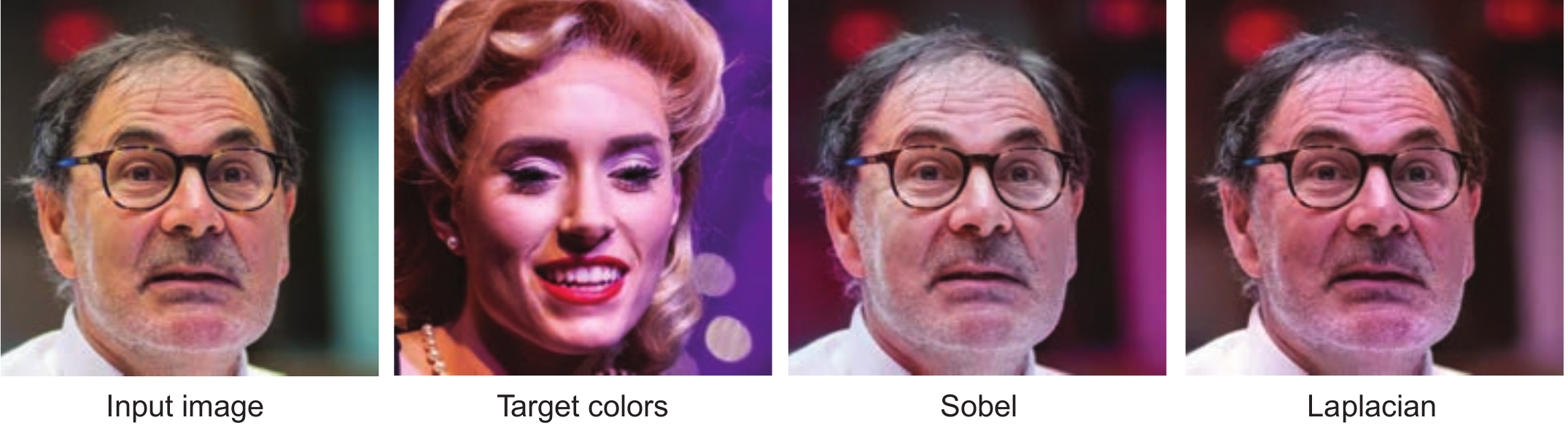}
\vspace{-7mm}
\caption[Results of two different kernels used to compute the reconstruction loss term.]{Results of two different kernels used to compute the reconstruction loss term.\label{histogran_appendix:fig:sobel_vs_laplacian}}
\end{figure}

To train our ReHistoGAN, we used the diffGrad optimizer \cite{8939562} with the same mini-batch size used to train our HistoGAN. We trained our network using the following hyperparameters $\alpha=2$, $\beta=1.5$, $\gamma=32$ for 100,000 iterations. Then, we continued training using $\alpha=2$, $\beta=1$, $\gamma=8$ for additional 30,000 iterations to reduce potential artifacts in recoloring.

\section{Ablation Studies} \label{histogran_appendix:sec:ablations}

We carried out a set of ablation experiments to study the effect of different values of hyperparameters used in Chapter \ref{ch:ch15}. Additionally, we show results obtained by variations in our loss terms.

We begin by studying the effect of the scale factor, $\alpha$, used in the loss function to train our HistoGAN. This scale factor was used to control strength of the histogram loss term. In this set of experiments, we used the 11K Hands dataset \cite{afifi201911k} to be our target domain and trained our HistoGAN with the following values of $\alpha$: 0.2, 2, 4, 8, and 16. Table \ref{histogran_appendix:table:ablation_results} shows the evaluation results using the Frech\'et inception distance (FID) metric \cite{heusel2017gans}, the KL divergence, and Hellinger distance. The KL divergence and Hellinger distance were used to measure the similarity between the target histogram and the histogram of GAN-generated images. Qualitative comparisons are shown in Fig.\ \ref{histogran_appendix:fig:alpha_ablation}

\begin{table}[t]
\caption[Results of our HistoGAN using different values of $\alpha$.]{Results of our HistoGAN using different values of $\alpha$. In this set of experiments, we used the Hands dataset \cite{afifi201911k} as our target domain. The term FID stands for the Frech\'et inception distance metric \cite{heusel2017gans}. The term KL Div. refers to the KL divergence between the histograms of the input image and generated image, while the term H. dis. refers to Hellinger distance.\label{histogran_appendix:table:ablation_results}}
\centering
\scalebox{0.7}{
\begin{tabular}{|c|c|c|c|}
\hline
 &  & \multicolumn{2}{c|}{RGB-$uv$ hist.} \\ \cline{3-4}
\multirow{-2}{*}{$\alpha$} & \multirow{-2}{*}{FID} & KL Div. & H dist. \\ \hline
0.2 & 1.9950 & 0.3935 & 0.3207 \\ \hline
\rowcolor[HTML]{FFFFC7}
2 & \textbf{2.2438} & \textbf{0.0533} & \textbf{0.1085} \\ \hline
4 & 6.8750 & 0.0408 & 0.0956 \\ \hline
8 & 9.4101 & 0.0296 & 0.0822 \\ \hline
16 & 15.747 & 0.0237 & 0.0743 \\ \hline
\end{tabular}}\vspace{2mm}
\end{table}

\begin{figure}[!t]
\includegraphics[width=\linewidth]{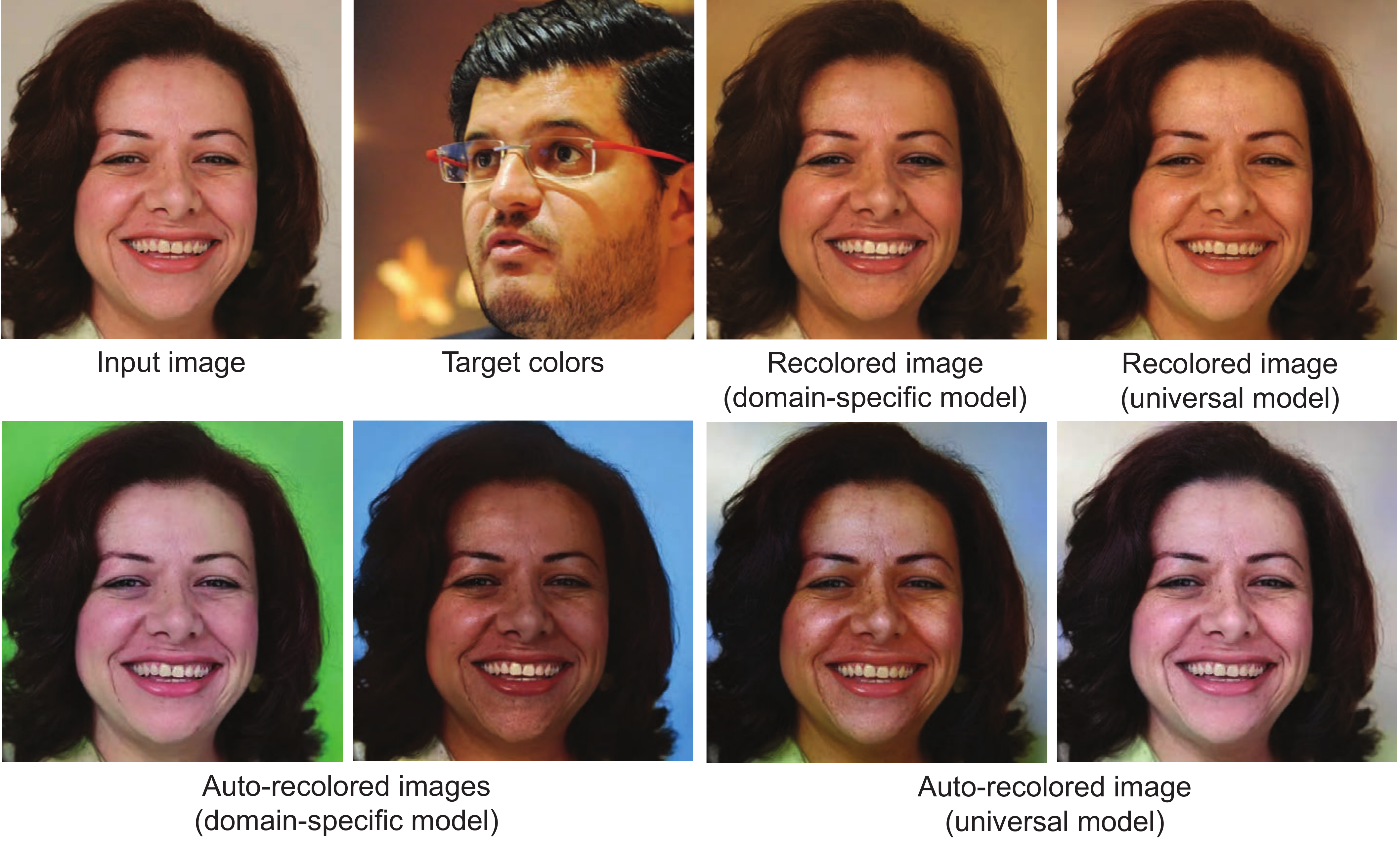}
\vspace{-7mm}
\caption[Results of domain-specific and universal ReHistoGAN models.]{Results of domain-specific and universal ReHistoGAN models. We show results of using a given target histogram for recoloring and two examples of the auto recoloring results of each model.}
\label{histogran_appendix:fig:object_specific_vs_universal_recoloring}
\end{figure}

Figure \ref{histogran_appendix:fig:alpha_beta_gamma_ablation} shows examples of recoloring results obtained by trained ReHistoGAN models using different combination values of $\alpha$, $\beta$, $\gamma$. As can be seen, a lower value of the scale factor, $\alpha$, of the histogram loss term results in ignoring our network to the target colors, while higher values of the scale factor, $\gamma$, of the discriminator loss term, make our method too fixated on producing realistic output images, regardless of achieving the recoloring (i.e., tending to re-produce the input image as is).

\begin{figure}[!t]
\includegraphics[width=\linewidth]{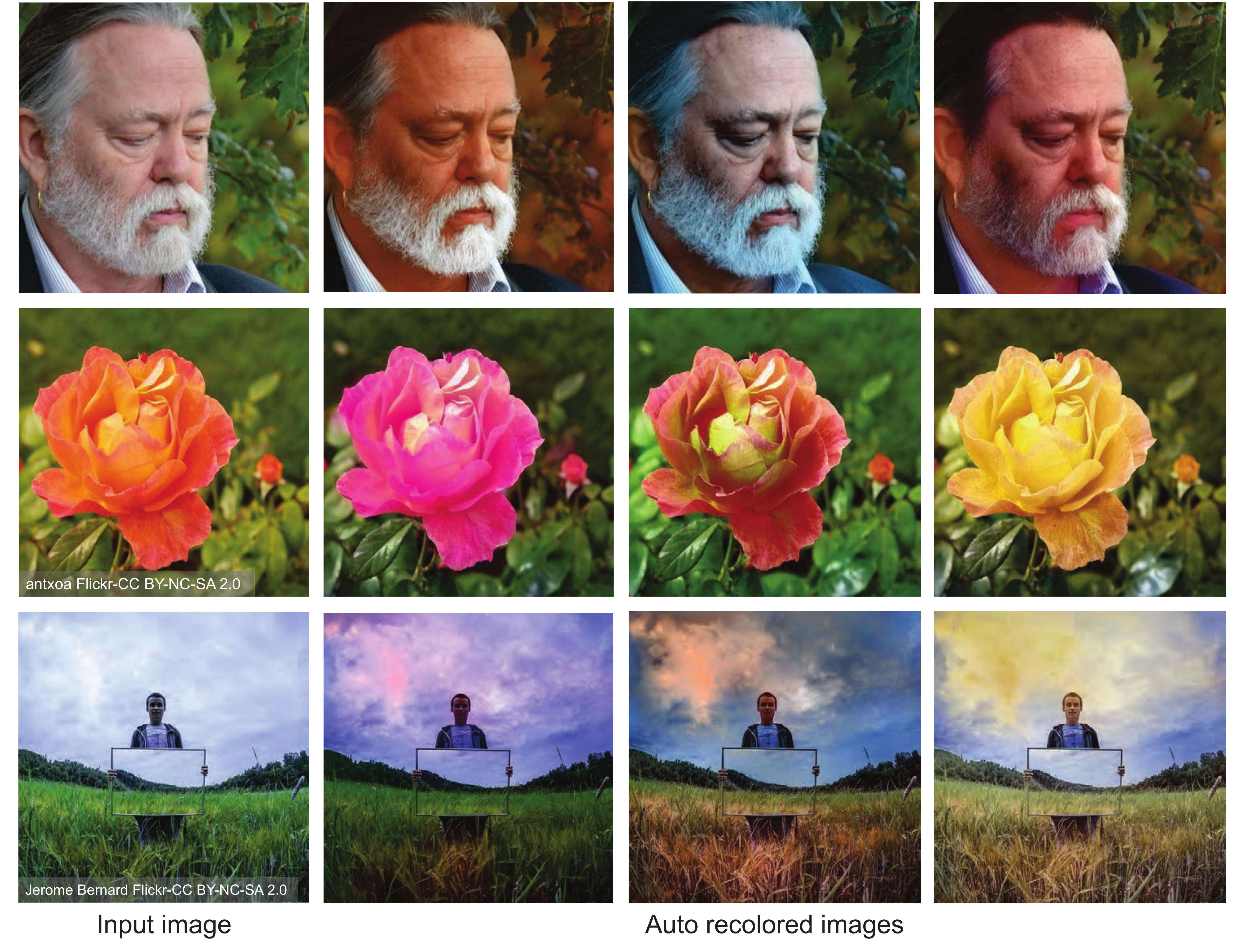}
\vspace{-8mm}
\caption[Auto recoloring using our universal ReHistoGAN model.]{Auto recoloring using our universal ReHistoGAN model.\label{histogran_appendix:fig:universal_model_auto_results}}
\end{figure}

\begin{figure}[!t]
\includegraphics[width=\linewidth]{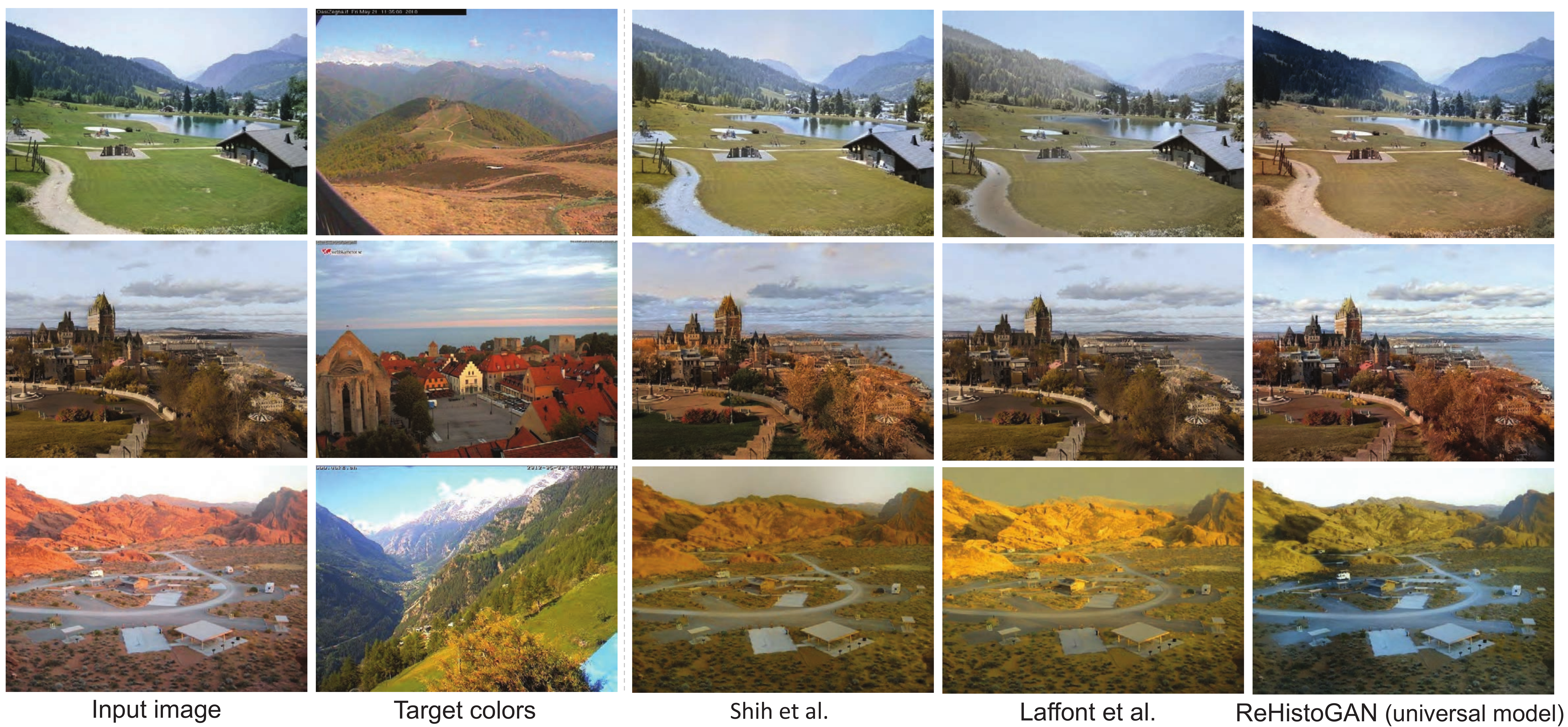}
\vspace{-7mm}
\caption[Comparisons between our universal ReHistoGAN, and the methods proposed by Shih et al., \cite{shih2013data} and Laffont et al., \cite{laffont2014transient} for color transfer.]{Comparisons between our universal ReHistoGAN, and the methods proposed by Shih et al., \cite{shih2013data} and Laffont et al., \cite{laffont2014transient} for color transfer.\label{histogran_appendix:fig:comparions_universal_model}}
\end{figure}

In the recoloring loss, we used a reconstruction loss term to retain the input image's spatial details in the output recolored image. Our reconstruction loss is based on the derivative of the input image. We have examined two different kernels, which are: the vertical and horizontal $3\!\times\!3$ Sobel kernels (i.e., the first-order directional derivative approximation) and the $3\!\times\!3$ Laplacian kernel (i.e., the second-order isotropic derivative). We found that training using both kernels give reasonably good results, while the Laplacian kernel produces more compelling results in most cases; see Fig.\ \ref{histogran_appendix:fig:sobel_vs_laplacian} for an example.

\section{Universal ReHistoGAN Model}\label{histogran_appendix:sec:universal}

As the case of most GAN methods, our ReHistoGAN targets a specific object domain to achieve the image recoloring task. This restriction may hinder the generalization of our method to deal with images taken from arbitrary domains. To deal with that, we collected images from a different domain, aiming to represent the ``universal'' object domain.

\begin{figure}[!t]
\includegraphics[width=\linewidth]{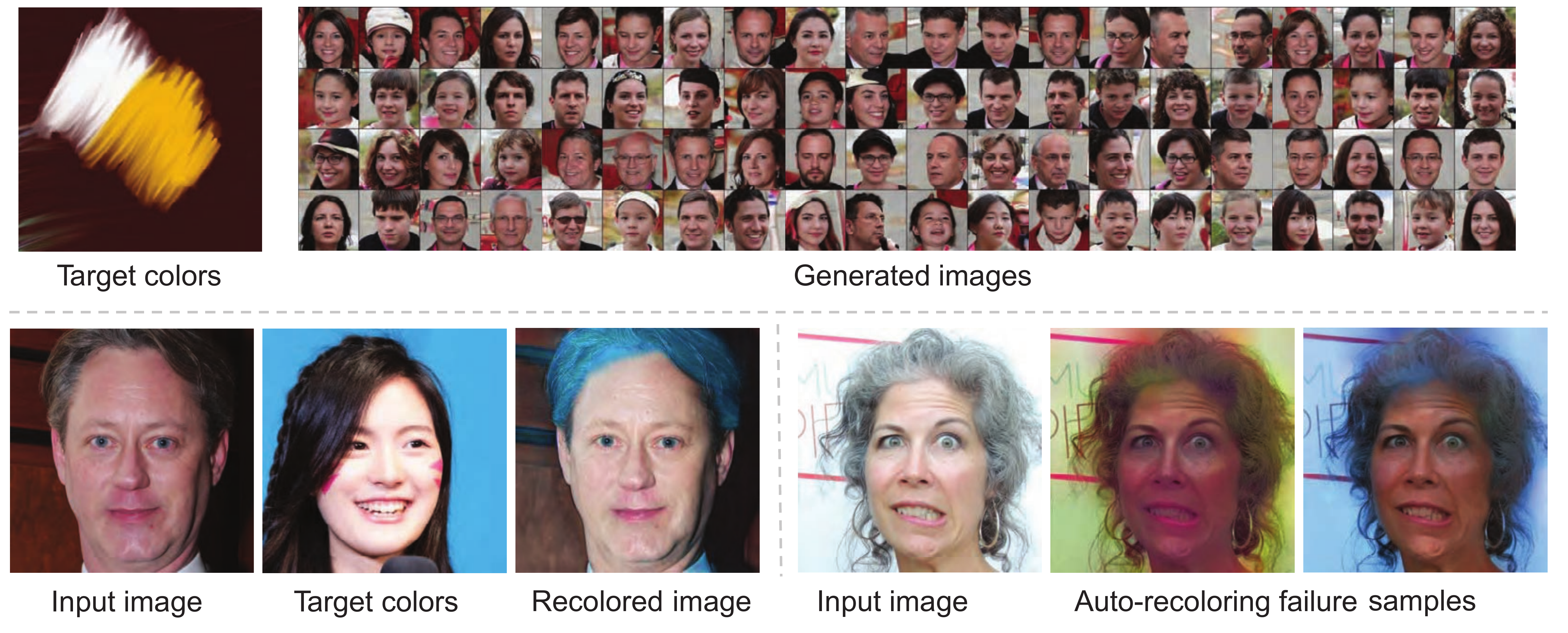}
\vspace{-7mm}
\caption[Failure cases of HistoGAN and ReHistoGAN.]{Failure cases of HistoGAN and ReHistoGAN. Our HistoGAN fails sometimes to consider all colors of target histogram in the generated image. Color bleeding is another problem that could occur in ReHistoGAN's results, where our network could not properly allocate the target (or sampled) histogram colors in the recolored image.\label{histogran_appendix:fig:failure_cases}}
\end{figure}

Specifically, our training set of images contains $\sim$2.4 million images collected from different image datasets. These datasets are: collection from the Open Images dataset \cite{kuznetsova2020open}, the MIT-Adobe FiveK dataset \cite{bychkovsky2011learning}, the Microsoft COCO dataset \cite{lin2014microsoft}, the CelebA dataset \cite{liu2015deep}, the Caltech-UCSD birds-200-2011 dataset \cite{wah2011caltech}, the Cats dataset \cite{catdataset}, the Dogs dataset \cite{khosla2011novel}, the Cars dataset \cite{krause20133d}, the OxFord Flowers dataset \cite{nilsback2008automated}, the LSUN dataset \cite{yu2015lsun}, the ADE20K dataset \cite{zhou2017scene}, and the FFHQ dataset \cite{karras2019style}. We also added Flickr images collected using the following keywords: $\texttt{landscape}$, $\texttt{people}$, $\texttt{person}$, $\texttt{portrait}$, $\texttt{field}$, $\texttt{city}$, $\texttt{sunset}$, $\texttt{beach}$, $\texttt{animals}$, $\texttt{living room}$, $\texttt{home}$, $\texttt{house}$, $\texttt{night}$, $\texttt{street}$, $\texttt{desert}$, $\texttt{food}$. We have excluded any grayscale image from the collected image set.

\begin{figure}[!t]
\includegraphics[width=\linewidth]{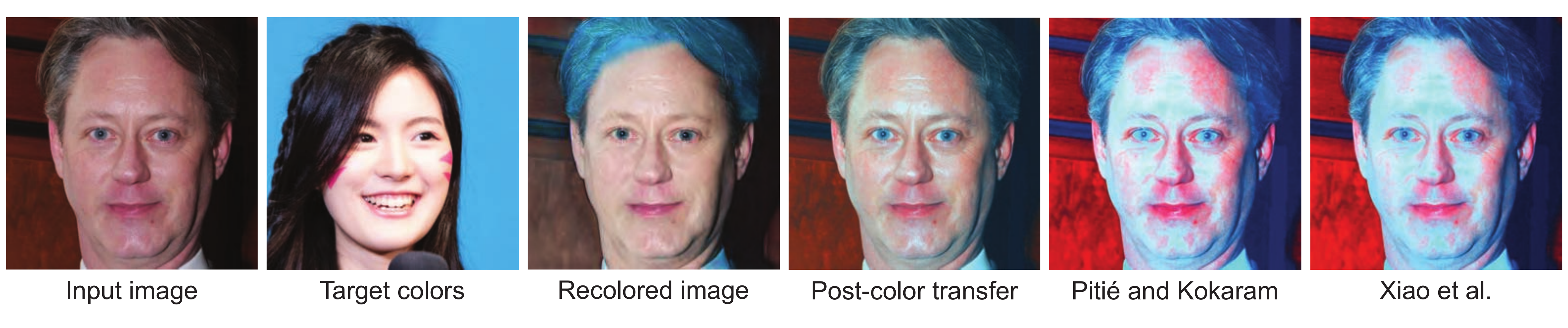}
\vspace{-7mm}
\caption[To reduce potential color bleeding artifacts, it is possible to apply a post-color transfer to our initial recolored image colors to the input image.]{To reduce potential color bleeding artifacts, it is possible to apply a post-color transfer to our initial recolored image colors to the input image. The results of adopting this strategy are better than applying the color transfer to the input image in the first place. Here, we use the color transfer method proposed by Piti\'e and Kokaram \cite{Pitie2007} as our post-color transfer method. We also show the results of directly applying Piti\'e and Kokaram's \cite{Pitie2007} method to the input image.\label{histogran_appendix:fig:fixing_failure_cases}}
\end{figure}

\begin{figure}[!t]
\centering
\includegraphics[width=0.95\linewidth]{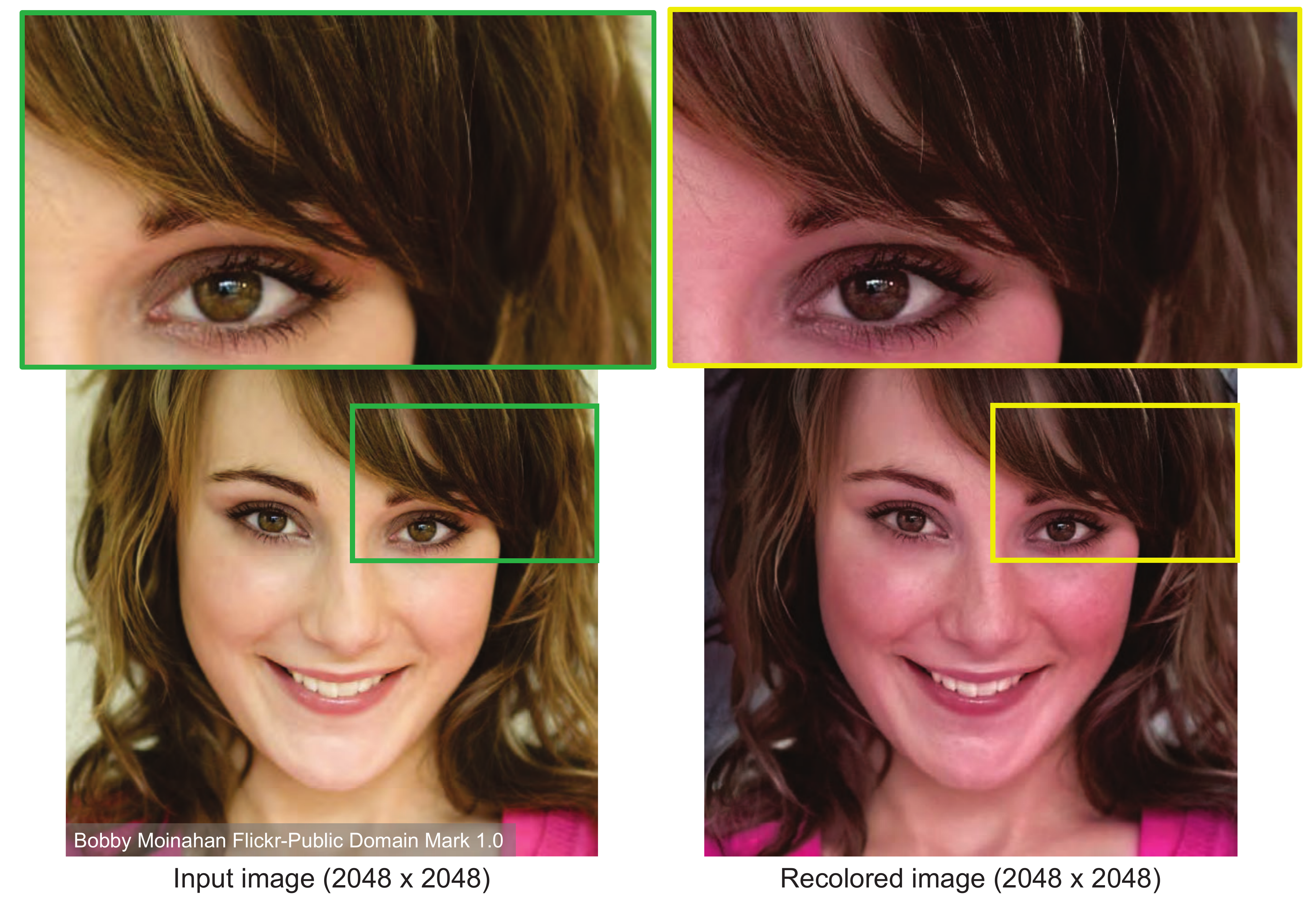}
\vspace{-1mm}
\caption[We apply the bilateral guided upsampling \cite{chen2016bilateral} as a post-processing to reduce potential artifacts of dealing with high-resolution images in the inference phase.]{We apply the bilateral guided upsampling \cite{chen2016bilateral} as a post-processing to reduce potential artifacts of dealing with high-resolution images in the inference phase. In the shown example, we show our results of recoloring using an input image with $2048\!\times\!2048$ pixels.\label{histogran_appendix:fig:dealing_with_any_size}}
\end{figure}

We trained our ``universal'' model using $m=18$ on this collected set of 2,402,006 images from several domains. The diffGrad optimizer \cite{8939562} was used to minimize the same generator loss described in Chapter \ref{ch:ch15} using the following hyperparameters $\alpha=2$, $\beta=1.5$, $\gamma=32$ for 150,000 iterations. Then, we used $\alpha=2$, $\beta=1$, $\gamma=8$ to train the model for additional 350,000 iterations. We set the mini-batch size to 8 with an accumulated gradient every 24 iterations. Figure \ref{histogran_appendix:fig:object_specific_vs_universal_recoloring} show results of our domain-specific and universal models for image recoloring. As can be seen, both models produce realistic recoloring, though the universal model tends to produce recolored images with less vivid colors compared to our domain-specific model. Additional examples of auto recoloring using our universal model are shown in Fig. \ref{histogran_appendix:fig:universal_model_auto_results}.

In Fig.\ \ref{histogran_appendix:fig:comparions_universal_model}, we show qualitative comparisons of the recoloring results using our universal ReHistoGAN and the method proposed in \cite{laffont2014transient}.

\section{Limitations} \label{histogran_appendix:sec:limitations}

Our method fails in some cases, where the trained HistoGAN could not properly extract the target color information represented in the histogram feature. This problem is due to the inherent limitation of the 2D projected representation of the original target color distribution, where different colors are mapped to the same chromaticity value in the projected space. This is shown in Fig.\ \ref{histogran_appendix:fig:failure_cases}-top, where the GAN-generated images do not have all colors in the given target histogram. Another failure case can occur in image recoloring, where the recolored images could have some color-bleeding artifacts due to errors in allocating the target/sampled histogram colors in the recolored image. This can be shown in Fig. \ref{histogran_appendix:fig:failure_cases}-bottom

\section{Post-Processing} \label{histogran_appendix:sec:post-processing}

As discussed in Sec. \ref{histogran_appendix:sec:limitations}, our method produces, in some times, results with color bleeding, especially when the target histogram feature has unsuitable color distribution for the content of the input image. This color-bleeding problem can be mitigated using a post-process color transfer between the input image and our initial recoloring. Surprisingly, this post-processing mapping produces results better than adopting the mapping in the first place---namely, applying the color transfer mapping without having our intermediate recoloring result.

Figure \ref{histogran_appendix:fig:fixing_failure_cases} shows an example of applying Piti\'e, and Kokaram's method \cite{Pitie2007} as a post-processing color transfer to map the colors of the input image to the colors of our recolored image. In the shown figure, we also show the result of using the same color transfer method -- namely, Piti\'e and Kokaram's method \cite{Pitie2007} -- to transfer the colors of the input image directly to the colors of the target image. As shown, the result of using our post-process strategy has a better perceptual quality.

Note that except for this figure (i.e., Fig.\ \ref{histogran_appendix:fig:fixing_failure_cases}), we \textit{did not} adopted this post-processing strategy to produce the reported results in Chapter \ref{ch:ch15} or this appendix. We discussed it here as a solution to reduce the potential color bleeding problem for completeness.

As our image-recoloring architecture is a fully convolutional network, we can process testing images in any arbitrary size. However, as we trained our models on a specific range of effective receptive fields (i.e., our input image size is 256), processing images with very high resolution may cause artifacts. To that end, we follow the post-processing approach used in Chapter \ref{ch:ch13} to deal with high-resolution images (e.g., 16-megapixel) without affecting the quality of the recolored image.

Specifically, we resize the input image to $256\!\times\!256$ pixels before processing it with our network. Afterward, we apply the bilateral guided upsampling \cite{chen2016bilateral} to construct the mapping from the resized input image and our recoloring result. Then, we apply the constructed bilateral grid to the input image in its original dimensions. Figure \ref{histogran_appendix:fig:dealing_with_any_size} shows an example of our recoloring result for a high-resolution image ($2048\!\times\!2048$ pixels). As can be seen, our result has the same resolution as the input image with no artifacts.

\chapter{List of Publications\label{ch:appendix4}}

\section*{Patents and Patent Applications}

\begin{enumerate}

\item \textbf{Mahmoud Afifi},  Michael S. Brown, Brian Price, and Scott Cohen. \href{https://patents.google.com/patent/US20200389635A1/en}{Image White Balancing.} \textbf{US Patent} No. 11,044,450, 2021.

\item \textbf{Mahmoud Afifi} and Michael S. Brown. \href{https://patents.google.com/patent/US20210160470A1}{Apparatus and Method for White Balance Editing}. \textbf{US Patent Application} No. 17/077,837, 2021.

\item \textbf{Mahmoud Afifi} and Michael S. Brown. \href{https://patents.justia.com/patent/20210058596}{Sensor-Independent Illuminant Estimation for Deep Learning Models.} \textbf{US Patent Application} No. 16/993,841, 2021.

\item Michael S. Brown, \textbf{Mahmoud Afifi}, Abdelrahman
Abdelhamed, Hakki Can Karaimer, Abdullah Abuolaim, and Abhijith Punnappurath. \href{https://patents.google.com/patent/WO2020206539A1/}{System and Method of Processing of a Captured Image to Facilitate Post-processing Modification.} \textbf{Worldwide Patent Application} No. PCT/CA2020/050465, 2020.

\item \textbf{Mahmoud Afifi}, Michael S. Brown, Konstantinos Derpanis, and Bj{\"o}rn Ommer. Network for Correcting Overexposed and Underexposed Images. \textbf{US Patent Application} -- to appear.

\end{enumerate}

\section*{Publications}

\begin{enumerate}
\item \textbf{Mahmoud Afifi}, Jonathan T. Barron, Chloe LeGendre, Yun-Ta Tsai, and Francois Bleibel. \href{https://arxiv.org/pdf/2011.11890.pdf}{Cross-Camera Convolutional Color Constancy}. In \href{http://iccv2021.thecvf.com} {International Conference on Computer Vision (\textbf{ICCV})} 2021 -- to appear.

\item \textbf{Mahmoud Afifi}, Marcus A. Brubaker, and Michael S. Brown. \href{https://arxiv.org/pdf/2011.11731.pdf}{HistoGAN: Controlling Colors of GAN-Generated and Real Images via Color Histograms}. In \href{http://cvpr2021.thecvf.com/} {IEEE Conference on Computer Vision and Pattern Recognition (\textbf{CVPR})}, 2021.

\item \textbf{Mahmoud Afifi}, Konstantinos G. Derpanis, Bj{\"o}rn Ommer, and Michael S. Brown. \href{https://arxiv.org/pdf/2003.11596.pdf}{Learning Multi-Scale Photo Exposure Correction}. In \href{http://cvpr2021.thecvf.com/} {IEEE Conference on Computer Vision and Pattern Recognition (\textbf{CVPR})}, 2021.

\item \textbf{Mahmoud Afifi}, Abdelrahman Abdelhamed, Abdullah Abuolaim, Abhijith Punnappurath, and Michael S. Brown. \href{https://ieeexplore.ieee.org/document/9394803/}{CIE XYZ Net: Unprocessing Images for Low-Level Computer Vision Tasks}. \href{https://ieeexplore.ieee.org/xpl/RecentIssue.jsp?punumber=34}{IEEE Transactions on Pattern Analysis and Machine Intelligence (\textbf{TPAMI})}, 2021.

\item \textbf{Mahmoud Afifi} and Michael S. Brown. \href{http://openaccess.thecvf.com/content_CVPR_2020/papers/Afifi_Deep_White-Balance_Editing_CVPR_2020_paper.pdf}{Deep White-Balance Editing.} In \href{http://cvpr2020.thecvf.com/} {IEEE Conference on Computer Vision and Pattern Recognition (\textbf{CVPR})}, 2020.

\item Majed El Helou, Ruofan Zhou, Sabine S\"{u}sstrunk, Radu Timofte, \textbf{Mahmoud Afifi}, Michael S. Brown, et al.. \href{https://link.springer.com/chapter/10.1007\%2F978-3-030-67070-2_30}{AIM 2020: Scene Relighting and Illumination Estimation Challenge}. In \href{https://eccv2020.eu/} {European Conference on Computer
Vision Workshops (\textbf{ECCVW})}, 2020 (\textbf{Runner-Up Award}).

\item \textbf{Mahmoud Afifi} and Michael S. Brown. \href{https://www.ingentaconnect.com/contentone/ist/cic/2020/00002020/00000028/art00022}{Interactive White Balancing for Camera-Rendered Images}. In \href{https://www.imaging.org/site/IST/Conferences/Color_and_Imaging/CIC2020/IST/Conferences/CIC/CIC_Home.aspx?hkey=ff455d92-345d-4ba1-8843-c4bedd9dcead} {Color and Imaging Conference (\textbf{CIC})}, 2020.

\item \textbf{Mahmoud Afifi} and Michael S. Brown. \href{http://openaccess.thecvf.com/content_ICCV_2019/papers/Afifi_What_Else_Can_Fool_Deep_Learning_Addressing_Color_Constancy_Errors_ICCV_2019_paper.pdf}{What Else Can Fool Deep Learning? Addressing Color Constancy Errors on Deep Neural Network Performance.} In \href{http://iccv2019.thecvf.com} {International Conference on Computer Vision (\textbf{ICCV})}, 2019.

\item \textbf{Mahmoud Afifi}, Brian Price, Scott Cohen, and Michael S. Brown. \href{http://openaccess.thecvf.com/content_CVPR_2019/papers/Afifi_When_Color_Constancy_Goes_Wrong_Correcting_Improperly_White-Balanced_Images_CVPR_2019_paper.pdf}{When Color Constancy Goes Wrong: Correcting Improperly White-Balanced Images.} In \href{http://cvpr2019.thecvf.com/} {IEEE Conference on Computer Vision and Pattern Recognition (\textbf{CVPR})}, 2019.

\item \textbf{Mahmoud Afifi}, Abhijith Punnappurath, Abdelrahman Abdelhamed, Hakki Can Karaimer, Abdullah Abuolaim, and Michael S. Brown. \href{https://www.ingentaconnect.com/content/ist/cic/2019/00002019/00000001/art00002}{Color Temperature Tuning: Allowing Accurate Post-Capture White-Balance Editing.} In \href{https://www.imaging.org/site/IST/Conferences/Color_and_Imaging/CIC27__2019_/IST/Conferences/CIC/CIC_Home.aspx?hkey=0fb3eb64-061f-4609-8d3c-7b8e6c0c4b90} {Color and Imaging Conference (\textbf{CIC})}, 2019 (\textbf{Best Paper Award}).

\item \textbf{Mahmoud Afifi}, Brian Price, Scott Cohen, and Michael S. Brown. \href{https://diglib.eg.org/handle/10.2312/egs20191008}{Image Recoloring Based on Object Color Distributions}. In \href{https://www.eurographics2019.it/} {\textbf{Eurographics - Short Papers}}, 2019.

\item \textbf{Mahmoud Afifi} and Michael S. Brown. \href{https://bmvc2019.org/wp-content/uploads/papers/0105-paper.pdf}{Sensor-Independent Illumination Estimation for DNN Models.} In \href{https://bmvc2019.org/} {British Machine Vision Conference (\textbf{BMVC})}, 2019.

\item \textbf{Mahmoud Afifi}, Abhijith Punnappurath, Graham Finlayson, and Michael S. Brown. \href{https://www.osapublishing.org/josaa/abstract.cfm?uri=josaa-36-1-71}{As-Projective-As-Possible Bias Correction for Illumination Estimation Algorithms.} \href{https://www.osapublishing.org/josaa/home.cfm} {Journal of the Optical Society of America A (\textbf{JOSA A})}, 2019.

\end{enumerate}

\end{appendices}

\end{document}